%% file: main.tex
\documentclass[10pt,twocolumn,letterpaper]{article}

\usepackage[pagenumbers]{conf}
\usepackage{graphicx}
\usepackage{wrapfig}
\usepackage{dblfloatfix}
\usepackage{cuted}
\usepackage{nicefrac}
\usepackage{microtype}
\usepackage[dvipsnames]{xcolor}
\usepackage{colortbl}
\definecolor{rankA}{HTML}{F7A8A8}\definecolor{rankB}{HTML}{FBD3A4}\definecolor{rankC}{HTML}{FFF3A3}
\newcommand{\rankA}[1]{\cellcolor{rankA}#1}\newcommand{\rankB}[1]{\cellcolor{rankB}#1}\newcommand{\rankC}[1]{\cellcolor{rankC}#1}
\newcommand{\ranklegend}{\colorbox{rankA}{best}, \colorbox{rankB}{second}, \colorbox{rankC}{third}}
\newcommand{\bestlegend}{\colorbox{rankA}{best}}
\usepackage{amsmath}
\usepackage{algpseudocode}
\algrenewcommand\algorithmicrequire{\textbf{Input:}}
\algrenewcommand\algorithmicensure{\textbf{Output:}}
\usepackage{pifont}
\usepackage{dblfloatfix}
\usepackage{cuted}
\usepackage{pgfplots}
\pgfplotsset{compat=1.18}
\usepgfplotslibrary{fillbetween}
\usepackage{tikz}
\usetikzlibrary{positioning}
\usetikzlibrary{pgfplots.groupplots}
\usetikzlibrary{calc}
\usetikzlibrary{shapes}
\usetikzlibrary{backgrounds}
\usetikzlibrary{arrows.meta}
\pgfdeclarelayer{bg}
\pgfdeclarelayer{fg}
\pgfsetlayers{bg,main,fg}

\input{preamble}

\definecolor{linkblue}{rgb}{0.21,0.49,0.74}

\usepackage[pagebackref,breaklinks,colorlinks,allcolors=linkblue]{hyperref}
\title{FlowSurf4D: Dynamic Surface Reconstruction with Dense Correspondences from Multi-View Video}

\author{Mohamed Ebbed$^{1,2}$ \qquad Daniele Baieri$^{1,2}$ \qquad Vladislav Golyanik$^{3}$ \qquad Zorah L\"ahner$^{1,2}$\\
{\normalsize $^{1}$University of Bonn \qquad $^{2}$Lamarr Institute \qquad $^{3}$Max Planck Institute for Informatics, SIC}
}

\usepackage{placeins}\usepackage{float}\usepackage{needspace}

\begin{document}
\twocolumn[{\renewcommand\twocolumn[1][]{#1}\maketitle
\centering
\vspace{-20pt}
{\small\definecolor{urlpink}{RGB}{232,46,130}\hypersetup{urlcolor=urlpink}\href{https://mohamed-ebbed.github.io/FlowSurf4D}{\texttt{mohamed-ebbed.github.io/FlowSurf4D}}}\par\vspace{2pt}
\input{figures/teaser}
\vspace{-5pt}
\captionof{figure}{\textbf{FlowSurf4D} reconstructs a dynamic surface from multi-view video (left; from TaoGS~\cite{taogs}, with challenging motion and fine detail) as high-fidelity meshes in dense correspondence over time. The correspondences come from bidirectional 3D flow fields between adjacent frames; chained, they track every point of the first frame through all later frames (dense 3D point tracks). They are accurate enough to transfer a texture from the first frame to all others and to relight the meshes with an albedo estimated through them, and the textured meshes can be placed in virtual environments. See the project-page video for dynamic visualizations.
} 
\label{fig:teaser}
\vspace{3mm}
}]

\input{sec/0_abstract}    
\input{sec/1_intro}
\input{sec/2_relwork}

\input{sec/3_method}

\input{sec/4_experiments}
\input{sec/conclusions}

{
    \small
    \bibliographystyle{ieeenat_fullname}
    \bibliography{main}
}

\clearpage
\appendix
\input{supp_floats}\setcounter{figure}{0}\setcounter{table}{0}\setcounter{equation}{0}
\maketitlesupplementary
\input{sec/appendix}

\end{document}

%% file: preamble.tex
\colorlet{Renderpath}{Melon}
\colorlet{Flowpath}{TealBlue}
\colorlet{SDFpath}{BrickRed}
\colorlet{Past}{Gray}

%% file: figures/teaser.tex
\centering
\definecolor{tzIn}{RGB}{74,82,99}\definecolor{tzOurs}{RGB}{14,128,229}\definecolor{tzApp}{RGB}{226,104,38}
\definecolor{tzPink}{RGB}{232,46,130}\definecolor{tzTeal}{RGB}{32,178,160}\definecolor{tzInk}{RGB}{24,28,40}
\resizebox{\textwidth}{!}{\begin{tikzpicture}[x=1in,y=1in,font=\small]
\useasboundingbox (0,0) rectangle (8.4205,-2.0067);
\begin{scope}[xshift=0.0325in]
\node[anchor=north west,inner sep=0pt] at (0.1433,-0.1700) {\includegraphics[width=0.95in]{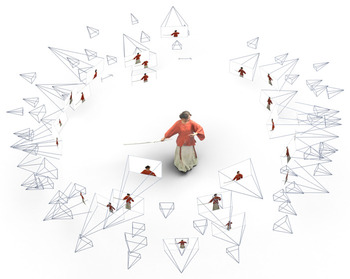}};
\fill[tzInk!85,rounded corners=1pt] (0.1205,-0.9800) rectangle (1.1161,-1.5900);
\foreach \j in {0,1,2}
  \node[anchor=north west,inner sep=0pt] at ({0.1555+\j*0.3202},-1.0550) {\includegraphics[height=0.46in]{imgs/teaser/in_strip_\j}};
\foreach \k in {0,...,13} {
  \fill[white,rounded corners=0.4pt] ({0.1420+\k*0.0700},-1.0025) rectangle +(0.032,-0.030);
  \fill[white,rounded corners=0.4pt] ({0.1420+\k*0.0700},-1.5375) rectangle +(0.032,-0.030); }
\draw[-{Latex[length=2.5pt]},tzIn,line width=0.5pt] (0.1205,-1.6300) -- (1.1161,-1.6300) node[midway,below,inner sep=1pt,text=tzInk,font=\fontsize{7}{8}\selectfont] {time};
\node[anchor=north west,inner sep=0pt] at (1.4250,-0.20) {\includegraphics[width=0.53in]{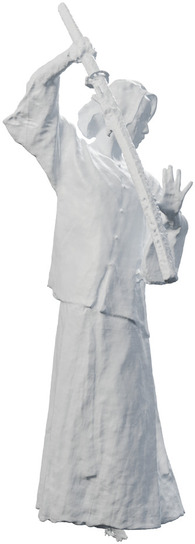}};
\node[anchor=north west,inner sep=0pt] (inset-1) at (2.0017,-0.21) {\includegraphics[width=0.74in]{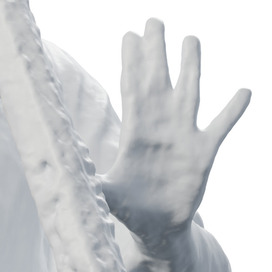}};
\node[anchor=north west,inner sep=0pt] (inset-2) at (2.0017,-0.9884) {\includegraphics[width=0.74in]{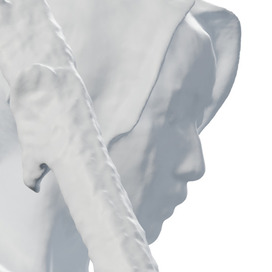}};
\node[anchor=north west,inner sep=0pt] at (2.8000,-0.1383) {\includegraphics[width=1.2850in]{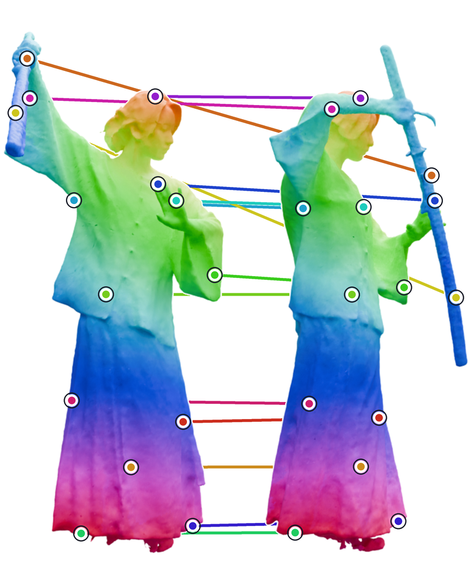}};
\node[anchor=north west,inner sep=0pt] at (4.08,-0.15) {\includegraphics[width=0.6833in]{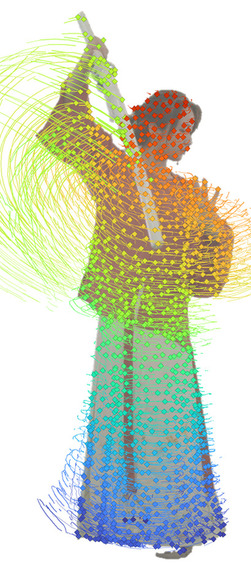}};
\node[anchor=north west,inner sep=0pt] at (5.0333,-0.20) {\includegraphics[width=0.53in]{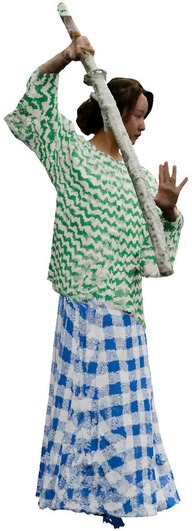}};
\node[anchor=north west,inner sep=0pt] at (5.6383,-0.20) {\includegraphics[width=0.53in]{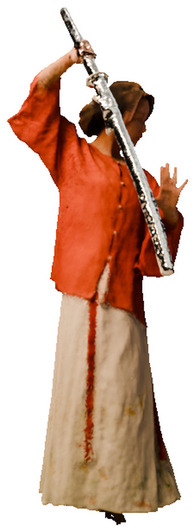}};
\node[anchor=north west,inner sep=0pt] at (6.2067,-0.20) {\includegraphics[width=0.53in]{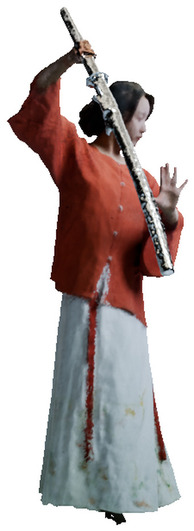}};
\node[anchor=north west,inner sep=0pt] at (6.8000,-0.20) {\includegraphics[width=0.72in]{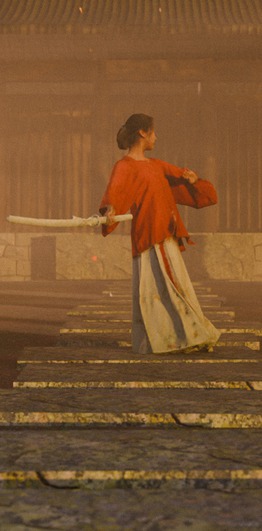}};
\node[anchor=north west,inner sep=0pt] at (7.5600,-0.20) {\includegraphics[width=0.72in]{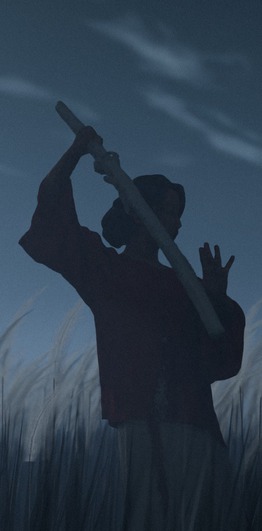}};
\draw[tzPink,line width=0.9pt] (1.8073,-0.5977) rectangle (1.9887,-0.7791);
\draw[tzTeal,line width=0.9pt] (1.7129,-0.4013) rectangle (1.8942,-0.5826);
\draw[tzPink,line width=0.9pt] (inset-1.south west) rectangle (inset-1.north east);
\draw[tzTeal,line width=0.9pt] (inset-2.south west) rectangle (inset-2.north east);
\draw[tzIn,line width=0.8pt,rounded corners=4pt] (0.0055,-0.1175) rectangle (1.2311,-2.0011);
\node[anchor=west,fill=tzIn,text=white,rounded corners=2.2pt,inner xsep=3pt,inner ysep=2pt] at (0.0833,-0.1175) {\bfseries Input};
\draw[tzOurs,line width=0.8pt,rounded corners=4pt] (1.3955,-0.1175) rectangle (4.8395,-2.0011);
\node[anchor=west,fill=tzOurs,text=white,rounded corners=2.2pt,inner xsep=3pt,inner ysep=2pt] at (1.4733,-0.1175) {\bfseries FlowSurf4D};
\draw[tzApp,line width=0.8pt,rounded corners=4pt] (5.0039,-0.1175) rectangle (8.3480,-2.0011);
\node[anchor=west,fill=tzApp,text=white,rounded corners=2.2pt,inner xsep=3pt,inner ysep=2pt] at (5.0817,-0.1175) {\bfseries Applications};
\node[anchor=north,align=center,text=tzInk,inner sep=0pt] at (0.6183,-1.8) {Posed multi-view video};
\node[anchor=north,align=center,text=tzInk,inner sep=0pt] at (2.0917,-1.8) {4D reconstruction};
\node[anchor=north,align=center,text=tzInk,inner sep=0pt] at (3.4425,-1.7281) {Long-term\\correspondences};
\node[anchor=north,align=center,text=tzInk,inner sep=0pt] at (4.4683,-1.8) {Point tracks};
\node[anchor=north,align=center,text=tzInk,inner sep=0pt] at (5.3150,-1.7281) {Texture\\transfer};
\node[anchor=north,align=center,text=tzInk,inner sep=0pt] at (6.2067,-1.7281) {Relighting and\\material editing};
\node[anchor=north,align=center,text=tzInk,inner sep=0pt] at (7.5400,-1.8) {Virtual environments};
\draw[tzOurs,line width=1.6pt] (1.2783,-0.9150) -- (1.2989,-0.9150);
\fill[tzOurs] (1.3483,-0.9150) -- (1.2789,-0.8719) -- (1.2789,-0.9581) -- cycle;
\draw[tzApp,line width=1.6pt] (4.8867,-0.9150) -- (4.9072,-0.9150);
\fill[tzApp] (4.9567,-0.9150) -- (4.8872,-0.8719) -- (4.8872,-0.9581) -- cycle;
\end{scope}
\end{tikzpicture}}

%% file: sec/0_abstract.tex
\begin{abstract} 
Reconstructing a deforming surface from multi-view video that is both accurate in every frame and in dense correspondence over time remains an open problem.
Existing methods trade one for the other: reconstructing every frame independently gives the most accurate surfaces but no correspondences, template-based methods come with correspondences by design but miss fine details and motions that the template does not capture, dynamic Gaussian methods track primitives that do not lie on a surface, and canonical-deformation methods relate the frames only through deformations into a canonical space that are fit to each frame separately and need not agree.
We present FlowSurf4D, which represents a deforming surface by a canonical signed distance field, deformation fields, and bidirectional 3D flow fields between adjacent frames that make the correspondences between frames explicit.
The flow fields are supervised by optical flow, regularize the normals of adjacent frames, and are optimized jointly with the surfaces in a sliding window.
Chaining the flow fields yields dense correspondences between all frames, more accurate than those obtained through the canonical space of the same model, which enables texture editing and relighting of the reconstructed 4D surfaces.
On ActorsHQ and DeformingThings4D, FlowSurf4D reconstructs the most accurate surfaces among dynamic methods on every scene, with the lowest temporal consistency error on DeformingThings4D, and is competitive with per-frame static methods.
On DeformingThings4D, its correspondences have a 43\% lower mean geodesic error than those provided by the next best dynamic method, whose surfaces are less accurate on every scene, and on real captures they agree best with 2D point tracks in held-out cameras.
\end{abstract}

%% file: sec/1_intro.tex
\section{Introduction} \label{sec:intro}

Capturing a moving person or object from multi-view video should produce more than a sequence of shapes: the surfaces should be accurate in every frame \emph{and} in correspondence over time, so that a texture painted on the first frame follows the surface through the whole sequence and the 4D reconstruction can be edited and relit.
Neural surface reconstruction recovers static scenes with remarkable detail~\cite{wang2023neuslearningneuralimplicit,yariv2021volume,li2023neuralangelo}, but for a deforming surface, obtaining both remains open: existing methods typically trade one for the other.
The most accurate geometry today is obtained by reconstructing every frame independently with a strong static method~\cite{Huang2DGS2024,chen2024pgsr,yu2024gaussian,guedon2025milo}.
Each frame is optimized individually on its own multi-view images, so the meshes of different frames have unrelated vertices.

Template-based performance capture~\cite{deAguiar2008, habermann2020deepcap} and dynamic Gaussian splatting~\cite{Luiten2024, huang2024scgs} track their elements over time.
However, templates limit detail, and Gaussian primitives do not lie on a surface, giving less accurate geometry. 
Canonical-deformation methods~\cite{park2021nerfies,tretschk2024scenerflow,ebbed2026moangelo} instead map every frame into a canonical shape. 
Points mapped to the same canonical location correspond only implicitly: each deformation is fit to its own frame, and nothing makes two deformations agree.
These correspondences are hence approximate and cannot be used during training: the deformations are not invertible, so finding a correspondence requires a nearest-neighbor search over the whole surface of the other frame, whereas training samples mini-batches of rays.

We present FlowSurf4D, the first method that provides both accurate surfaces and dense correspondences.
Our key idea is to make the correspondences an explicit part of the model: the model outputs them directly and uses them during training to regularize the normals of adjacent frames.
We represent the sequence by a neural signed distance field (SDF), deformation fields into its canonical space, and 3D flow fields that map each frame to its previous and next frame.
Since the flow fields are local in time, we optimize the frames in a sliding window: adjacent frames are fit together, and memory does not grow with the sequence length.
The optical flow of the input videos~\cite{Poggi_2025_ICCV} supervises the flow fields and concentrates the rays on moving regions, and a loss compares the normals of corresponding points in adjacent frames.
Chaining the flow fields yields dense correspondences between all frames, which, \eg, transfer a texture from the first frame to the whole sequence (see \cref{fig:teaser}).

We evaluate FlowSurf4D on real captures from ActorsHQ~\cite{isik2023humanrf} and the TaoGS dataset~\cite{taogs} and on synthetic characters from DeformingThings4D~\cite{li2021deformingthings4d}.
Among dynamic methods, it reconstructs the most accurate surfaces on every scene with ground-truth geometry, and its correspondences have the lowest mean error on DeformingThings4D and agree best with 2D point tracks on real captures.

\paragraph{Contributions.} Our technical contributions are as follows: 
\begin{itemize}
    \item Joint dynamic surface reconstruction and dense correspondence estimation from multi-view video, with explicit 3D flow fields between adjacent frames (\cref{sec:dynamic_scene}) that act on the reconstructed surface and are read out as dense correspondences between its meshes (\cref{sec:correspondence}). On the same surfaces, they are 18\% more accurate than the correspondences through the canonical space of the same model.
    \item A sliding-window optimization (\cref{sec:window}) in which adjacent frames are fit together and constrain each other, with memory that does not grow with the sequence length: a normal consistency loss between corresponding points (\cref{sec:optimization}) reduces the normal error and the temporal inconsistency, and motion-guided ray sampling driven by the optical flow (\cref{sec:sampling}) reduces the Chamfer distance and the temporal consistency error.
    \item A reconstruction and evaluation benchmark for both geometry and correspondences (\cref{sec:experiments}), created from DeformingThings4D: 10 animated characters rendered from 50 cameras, whose ground-truth meshes share their vertices over time; on real captures, we evaluate the correspondences against 2D point tracks in held-out cameras. Our correspondences have a 43\% lower mean error than those of the next best dynamic method, and our surfaces are the most accurate among dynamic methods on every scene.
\end{itemize}
The code and benchmark will be released soon.

%% file: sec/2_relwork.tex
\section{Related Work} \label{sec:rw}

\subsection{Accurate Dynamic Surface Reconstruction}

Classical 4D techniques~\cite{Starck2007, deAguiar2008, Tung2009, Oswald2015, Mustafa2022} reconstruct mesh sequences with multi-view stereo, relying on 2D matches that are brittle under occlusions. 
Neural surface reconstruction optimizes geometry and appearance jointly by differentiable rendering, with SDFs~\cite{yariv2021volume, wang2023neuslearningneuralimplicit, li2023neuralangelo} or 3D Gaussian Splatting (3DGS)~\cite{kerbl3Dgaussians} aligned with the surface or converted to 
a mesh~\cite{guedon2023sugar, Huang2DGS2024, chen2024pgsr, yu2024gaussian, guedon2025milo, Yang2025, Son2025, Held2026}.
Applied to every frame, possibly initialized from the previous one~\cite{wang2023neus2}, these methods provide the most accurate per-frame geometry, but do not reconstruct or maintain point associations across individual meshes. 

In contrast, dynamic representations model the whole sequence globally. 
Most of them target novel-view synthesis~\cite{Li2022, Cao2023HexPlane, Xu2024, Sun2024, 4dgaussians, Luiten2024}, where the geometric accuracy is not decisive. 
3DGS extensions that explicitly aim at accurate geometry use surface-aligned primitives over time~\cite{zhang2024dynamic2dgaussiansgeometrically, chen2025atgs}, deformed from a canonical space with depth and normal priors~\cite{wang2024st2dgs} or kept locally rigid~\cite{cai2024dynasurfgs}, or extract a mesh from deforming Gaussians~\cite{Liu2025}, but lack the detail of static methods. 
4DSurf~\cite{wu20264dsurf} aligns the motion of Gaussians with the surface through a flow of the SDF, used as a regularizer rather than as an explicit map between adjacent surfaces. Binding Gaussians to a mesh~\cite{zheng2025gaustar, ma2024magsreconstructingsimulatingdynamic} limits the detail to that mesh and requires guide meshes for the entire sequence. 
Next, neural SDFs have been extended to non-rigid scenes with a canonical shape and per-frame deformations~\cite{Tretschk2021, Johnson2023, tretschk2024scenerflow}, and MoAngelo~\cite{ebbed2026moangelo} tracks such a canonical SDF one frame at a time and obtains detailed meshes. 
In all of these methods, the per-frame geometry is constrained only by the corresponding images of that timestep; letting adjacent frames constrain each other requires explicit 3D point correspondences over time. 

\input{figures/cd_vs_corr}

\subsection{Correspondences and Flow} 
Canonical-deformation methods~\cite{park2021nerfies, Tretschk2021, tretschk2024scenerflow, ebbed2026moangelo} relate two frames only through their separate deformations into the canonical space, which are fit to the images of their own frames and need not agree there, so the correspondences obtained through it are approximate and do not enter the optimization.
Invertible deformation fields~\cite{cai2022neuralsurface, Lei2022CaDeX, baieri2025implicitarap} make these correspondences consistent by construction, at the price of restricting the deformations they can represent: a general deformation field, such as a hash grid followed by an MLP~\cite{mueller2022instant}, cannot be inverted.
Dynamic 3D Gaussians~\cite{Luiten2024} and deformable Gaussians~\cite{yang2024deformable, huang2024scgs, 4dgaussians} move their primitives over time, as do volumetric-video methods for humans, which target novel-view synthesis and compression~\cite{jiang2024hifi4g, jiang2024robust, taogs}, but the Gaussians do not lie on a smooth surface, so their tracks are correspondences of primitives rather than of surface points.
Neural Scene Flow Fields (NSFF)~\cite{Li2021} represent motion by forward and backward scene flow between adjacent time steps, supervised by 2D optical flow and regularized by cycle consistency, for view synthesis from monocular video with a density field and no canonical space. 
Like the scene flow of an evolving SDF~\cite{mao2024sdfflow}, this flow is not read out as surface correspondences.
Performance capture deforms a pre-scanned template~\cite{deAguiar2008, cagniart2010probabilistic, habermann2020deepcap, habermann2021ddc}, tracks keyframe meshes that are re-anchored over time~\cite{collet2015high}, or warps a canonical model fused from depth with a deformation graph~\cite{sumner2007embedded, newcombe2015dynamicfusion, dou2016fusion4d, dou2017motion2fusion, yu2021function4d, bozic2021ndg}; DynoSurf~\cite{yao2024dynosurf} learns a template from point clouds.
These give correspondences by construction, but need a template, depth or point clouds, limit the detail to the tracked shape, or lose the correspondences at each re-anchoring.
Shape matching~\cite{zhuravlev2026star, kim2011blended, ovsjanikov2012functional, ren2018continuous, melzi2019zoomout, cao2023ulrssm, bastianxie2024hybrid} computes correspondences assuming meshes as inputs, implying that its accuracy strongly depends on provided surfaces it has no influence on. 

In contrast, our resulting flow fields (3D correspondences over time) are estimated jointly with the surfaces, which reinforce each other and are constantly evaluated for consistency during optimisation. 
Unlike matching through a canonical space, they relate adjacent frames directly, and unlike scene flow~\cite{Li2021, mao2024sdfflow}, they are read out as dense correspondences between the reconstructed surfaces (see \cref{fig:cd_vs_corr}).

%% file: figures/cd_vs_corr.tex
\begin{figure}[t]
\centering
\includegraphics[width=\columnwidth]{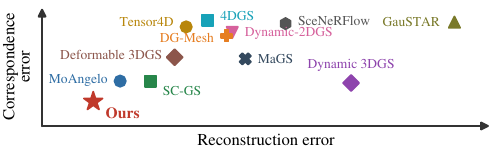}
\vspace{-14pt}
\caption{\textbf{Surface vs.\ correspondence accuracy} on DeformingThings4D (detailed plot: App.~\ref{sec:supp_corr}).}
\label{fig:cd_vs_corr}
\vspace{-8pt}
\end{figure}

%% file: sec/3_method.tex
\section{Methodology}

Given multi-view RGB videos, we reconstruct a dynamic surface together with dense correspondences between its frames (\Cref{fig:pipeline}).
We first optimize a canonical representation of geometry and appearance on the first frame (\cref{sec:template}).
Motion is modeled by deformation fields that map each frame to the canonical space and, unlike previous canonical-deformation methods, by bidirectional flow fields between adjacent frames, which are optimized jointly in a sliding window (\cref{sec:dynamic_scene}).
The flow fields are supervised by optical flow~\cite{Poggi_2025_ICCV}, provide the correspondences for temporal regularization (\cref{sec:rendering,sec:optimization}) and, chained, the dense correspondences between frames (\cref{sec:correspondence}); the optical flow also guides the sampling of rays (\cref{sec:sampling}).

\input{figures/method_fig}

\subsection{Canonical Scene Representation}\label{sec:template}
Our canonical geometry is a neural SDF $d = f_{\text{sdf}}(x, \mathbf{z})$, where $\mathbf{z} = \text{HashGrid}(x)$ are the features of a 16-level multi-resolution hash grid~\cite{mueller2022instant} at the 3D point $x$, followed by a shallow MLP with a single hidden layer.
The canonical appearance is a radiance field $f_{\text{rgb}}$ that maps a surface position $x$, viewing direction $v$, features $\mathbf{z}$ and normal $n = \nabla_x f_{\text{sdf}}(x) / \lVert\nabla_x f_{\text{sdf}}(x)\rVert$ to an RGB color $c$.
The resulting model $f_{\text{sdf,rgb}}^{(0)}$ is fit to the first frame following~\cite{li2023neuralangelo}.

\subsection{Dynamic Scene Representation}\label{sec:dynamic_scene}

\paragraph{Canonical Deformation.} \label{sec:canonical_deform}
After optimizing the canonical scene, we iteratively track its motion across the frames. As in canonical-deformation methods~\cite{park2021nerfies,Tretschk2021,ebbed2026moangelo}, we model the motion of each frame with a deformation field $\delta_t\colon\mathbb{R}^3\to\mathbb{R}^3$ that maps 3D points $x_t$ at frame $t$ to their counterparts ${x}_c$ in the canonical space.
Each deformation field $\delta_t$ is parameterized by a multi-resolution 3D hash grid encoding followed by a shallow MLP with a single hidden layer.
The deformation field predicts a transformation $\mathbf{\xi}_t\in\mathfrak{se}(3)$, which is mapped to a rotation $\mathbf{R}_t \in \mathrm{SO}(3)$ and translation $\mathbf{T}_t\in\mathbb{R}^3$ via the exponential map. The transformed point is then obtained as
\begin{equation}\label{eq:to-canonical}
{x}_t^\star = \mathbf{R}_t \, {x}_t + \mathbf{T}_t.
\end{equation}
The deformation fields relate each frame only to the canonical template, so the motion between two frames,
and with it the dense correspondences between them, is unknown. Hence, no loss can compare a point to its
counterpart in an adjacent frame.
The next sections introduce our solution.

\paragraph{Sliding Window Motion Tracking.}\label{sec:window}

To provide these counterparts, we introduce flow fields that map a point to its corresponding points in the adjacent frames, in both directions. Like the deformation fields, each flow field predicts a transformation in $\mathfrak{se}(3)$, $\phi(x)=\mathbf{R}(x)\,x+\mathbf{T}(x)$.
To use them for temporal regularization, the deformation fields of the neighboring frames must be optimized together with the current one. 
Tracking one frame at a time does not allow this: each frame would be optimized only once, with its neighbors fixed or not yet available.
We therefore jointly optimize each triplet of frames $(t-1, t, t+1)$, denoted $\mathrm{prev}$, $\mathrm{curr}$ and $\mathrm{next}$ (see \cref{fig:pipeline}), with deformation fields $\delta^{t}_{\mathrm{prev}}$, $\delta^{t}_{\mathrm{curr}}$, $\delta^{t}_{\mathrm{next}}$ and flow fields $\phi^t_{\mathrm{curr}\to\mathrm{next}}$, $\phi^t_{\mathrm{curr}\to\mathrm{prev}}$, $\phi^t_{\mathrm{next}\to\mathrm{curr}}$, $\phi^t_{\mathrm{prev}\to\mathrm{curr}}$.
Once optimized, the window shifts forward by one frame. The deformation and flow fields of the two frames that remain in the window are kept, while those of the new frame are initialized from its predecessor (App.~\cref{fig:init}). 
The canonical representation (\cref{sec:template}) is not frozen: the losses of each window also update its parameters, so it is fine-tuned as the window moves, and we denote its state in the window centered at $t$ by $f^{(t)}_{\{\text{sdf,rgb}\}}$.
Each iteration renders three frames instead of one, which roughly triples the computation per frame, but the required GPU memory does not grow with the sequence length, since only the three frames of the current window are optimized (runtimes: App.~\ref{sec:supp_impl}).

\subsection{Motion-Guided Importance Sampling}\label{sec:sampling}
Casting rays uniformly gives each region a share of rays proportional to its size, so a waving arm could receive relatively few rays while being the region with the most change in successive frames.
We therefore sample rays according to the motion between frames $t-1$ and $t$, measured by the precomputed optical flow (App.~\ref{sec:supp_sampling}). Unlike DyNeRF~\cite{Li2022}, which favors pixels whose color changes over time, our sampling measures motion with the optical flow and therefore also favors moving regions of uniform color, such as a shirt or skin, whose color barely changes.
Most rays ($90\%$) are cast through the dilated union of the foreground masks of the two frames, where the geometry initialized from frame $t-1$ (\cref{sec:window}) must be updated, and each pixel is weighted by its forward and backward optical-flow motion, so that moving pixels receive a target share of half of these rays even when they cover only a small part of the image (the emphasis on motion is capped so that static foreground pixels are still fitted, App.~\ref{sec:supp_sampling}); the remaining rays cover the rest of the image, which keeps the geometry stable and removes floaters.
Thus, typically about $45\%$ (half of the $90\%$ foreground rays) of all rays land on moving pixels, whereas uniform sampling would give them only the share of the image they cover, for example about $4\%$ on an ActorsHQ sequence in which mainly the arms move.
We apply this sampling independently to each frame of the window (\cref{sec:window}).

\subsection{Volume Rendering of Dynamic Scenes}\label{sec:rendering}

In the window centered at $t$, we sample pixels in each frame $\tau \in \{\mathrm{prev}, \mathrm{curr}, \mathrm{next}\}$ independently (\cref{sec:sampling}) and sample $N$ points $\mathbf{x}_{\tau}$ along their rays following~\cite{li2023neuralangelo}.
The points are mapped to the canonical space by $\delta^t_{\tau}$ (\Cref{eq:to-canonical}), where $f^{(t)}_{\{\text{sdf,rgb}\}}$ gives their signed distances $d_{\tau}$, normals $n_{\tau}$ and colors $c_{\tau}$. 
Lastly, we convert distances $d_{\tau}$ to volume densities $\sigma_{\tau}$ and compute volume rendering weights $w_{\tau}$ following \cite{li2023neuralangelo}.
The color $C_{\tau,j}$ and the soft mask $O_{\tau,j}$ of the pixel of ray $j$ are
\begin{equation}\label{eq:render-color}
C_{\tau,j} = \sum_{i=1}^{N} w^{i}_{\tau,j} \, c^{i}_{\tau,j} \quad\text{and}\quad O_{\tau,j} = \sum_{i=1}^{N} w^{i}_{\tau,j}.
\end{equation}

Additionally, given $\tau'\in\{\mathrm{prev}, \mathrm{curr}, \mathrm{next}\}$, we can use our flow fields $\phi^t_{\tau\to\tau'}$ to render 2D optical flows of our dynamic scene. We composite the samples of ray $j$ and their positions moved by the flow field with the volume rendering weights, and project both into the image of frame $\tau$:
\begin{equation}\label{eq:optiflow}
    V_{\tau \to \tau',j} = \pi_{\tau}\Big(\sum_{i=1}^{N} w^{i}_{\tau,j}\, \phi^t_{\tau\to\tau'}(\mathbf{x}^i_{\tau,j})\Big) - \pi_{\tau}\Big(\sum_{i=1}^{N} w^{i}_{\tau,j}\, \mathbf{x}^i_{\tau,j}\Big),
\end{equation}
where $\pi_{\tau}$ is the perspective projection of the camera that observes ray $j$ (with known parameters). As in NSFF~\cite{Li2021}, 2D flow is rendered from expected 3D positions; unlike NSFF, the flow acts on a deformed SDF and is read out as surface correspondences.

\subsection{Optimization}\label{sec:optimization}

The losses fall into three groups. Reconstruction losses fit each frame of the window to its own images, and motion regularization keeps the deformation and flow fields locally rigid and smooth. Since the flow fields make the correspondences between adjacent frames explicit, temporal regularization can compare neighboring frames directly, unlike a model that relates frames only through the canonical space~\cite{park2021nerfies,Tretschk2021,ebbed2026moangelo}.

\subsubsection{Reconstruction}
For each frame $\tau$ of the window $W = \{t-1, t, t+1\}$, we sample a mini-batch of $R$ rays; an $\ell_1$ loss $\mathcal{L}_{\text{rgb}}$ compares the rendered colors (\Cref{eq:render-color}) with the input images, and a binary cross-entropy loss $\mathcal{L}_{\text{mask}}$ compares the rendered opacities with the foreground masks, giving the data term $\mathcal{L}_{\text{data}} = \lambda_{\text{rgb}}\mathcal{L}_{\text{rgb}} + \lambda_{\text{mask}}\mathcal{L}_{\text{mask}}$.
Eikonal losses keep the SDF valid both after deformation ($\mathcal{L}_{\text{sdf}}$) and in canonical space ($\mathcal{L}^{\star}_{\text{sdf}}$), where the gradient is normalized to obtain the normals used in $\mathcal{L}_{\mathrm{norm}}$ (equations in App.~\ref{sec:supp_losses}).

\subsubsection{Motion Regularization}\label{sec:motion-reg}
At the volume-rendered surface point of each ray, an as-rigid-as-possible (ARAP) term~\cite{sorkine2007arap}, related to the elastic regularization of~\cite{park2021nerfies}, keeps the local motion of the deformation and flow fields close to a rotation, and a smoothness term penalizes abrupt changes of the motion between nearby points ($\mathcal{L}^{\delta}_{\mathrm{arap}}$ and $\mathcal{L}^{\delta}_{\mathrm{smooth}}$ for the three deformation fields, $\mathcal{L}^{\phi}_{\mathrm{arap}}$ and $\mathcal{L}^{\phi}_{\mathrm{smooth}}$ for the four flow fields, equations in App.~\ref{sec:supp_losses}).

\subsubsection{Temporal Regularization}
Each frame is fit only to its own images, so nothing makes adjacent frames agree. The flow fields provide the correspondences that let us compare them.
\paragraph{Flow supervision.}
To predict correspondences between adjacent frames, we supervise the rendered optical flow (\Cref{eq:optiflow}) with optical flow precomputed by FlowSeek~\cite{Poggi_2025_ICCV} for each of the four directions $(\tau,\tau') \in \mathcal{P} = \{(\mathrm{curr},\mathrm{prev}), (\mathrm{curr},\mathrm{next}), (\mathrm{prev},\mathrm{curr}), (\mathrm{next},\mathrm{curr})\}$:
\begin{equation}\label{eq:lflow}
\mathcal{L}_{\mathrm{flow}} = \frac{1}{4} \sum_{(\tau,\tau') \in \mathcal{P}} \frac{1}{R}\sum_{j=1}^{R} \left\| V_{\tau\to\tau', j} - V^{*}_{\tau\to\tau', j} \right\|_{1},
\end{equation}
where $V^{*}_{\tau\to\tau',j}$ is the FlowSeek optical flow at the pixel of ray $j$.
$\mathcal{L}_{\mathrm{flow}}$ uses no occlusion mask and no cycle consistency. At occlusion boundaries, where the optical flow is unreliable, the ARAP and smoothness terms (\cref{sec:motion-reg}) keep the flow fields close to the motion of their surroundings.

\paragraph{Normal consistency.}
The flow fields then provide correspondences across frames, which we use to enforce geometric consistency over time.
They allow comparing any surface property at corresponding points of adjacent frames, and we compare the normals.
For a ray $j$ of frame $\tau$ with samples $x^i_{\tau,j}$, we compare two canonical-space normals: one obtained through the frame's own deformation field, and one obtained by first moving the samples to frame $\tau'$ with the flow field and then deforming them with that frame's deformation field, both composited with the (fixed) rendering weights of ray $j$:
\begin{equation}\label{eq:composited-normals}
\begin{aligned}
\bar{n}^{\star}_{\tau,j} &= \sum_{i=1}^{N} w^{i}_{\tau,j}\, n^{\star}\!\big(\delta^t_{\tau}(x^i_{\tau,j})\big), \\
\bar{n}^{\star}_{\tau\to\tau',j} &= \sum_{i=1}^{N} w^{i}_{\tau,j}\, n^{\star}\!\big(\delta^t_{\tau'}(\phi^t_{\tau\to\tau'}(x^i_{\tau,j}))\big),
\end{aligned}
\end{equation}
where $n^{\star}(x^\star) = \nabla_{x^\star} f_{\text{sdf}}(x^\star) / \lVert \nabla_{x^\star} f_{\text{sdf}}(x^\star) \rVert$ is the normal of the canonical SDF.
If the deformation and flow fields agree, both routes reach the same canonical point and the two normals coincide.
We compare normals: as derivatives of the canonical SDF, they are sensitive to small differences in the local shape of the surface, whereas which points correspond is set by $\mathcal{L}_{\mathrm{flow}}$.
We measure their disagreement as $e_{\tau\to\tau',j} = 1 - \bar{n}^{\star\top}_{\tau,j}\, \bar{n}^{\star}_{\tau\to\tau',j}$.
Since the weights receive no gradient, $e_{\tau\to\tau',j}$ acts only on the normals; the composites are not normalized, so $e_{\tau\to\tau',j}$ is minimal when the normals agree both between the two routes and along the ray.
We penalize it for each pair $(\tau,\tau') \in \mathcal{P}$, weighting each ray by the (fixed) $\omega_{\tau\to\tau',j} = O_{\tau,j}\, s_{\tau\to\tau',j}$, the product of its rendered opacity $O_{\tau,j}$ and a motion factor $s_{\tau\to\tau',j}$ that decreases with the optical-flow magnitude at the ray (App.~\ref{sec:supp_motion_factor}):
\begin{equation}\label{eq:lnorm}
\mathcal{L}_{\mathrm{norm}} = \frac{1}{4} \sum_{(\tau,\tau') \in \mathcal{P}}
\frac{\sum_{j=1}^{R} \omega_{\tau\to\tau',j}\, e_{\tau\to\tau',j}}{\sum_{j=1}^{R} \omega_{\tau\to\tau',j}}.
\end{equation}
Since $s_{\tau\to\tau',j}$ decreases with the flow magnitude, fast-moving regions, where the optical flow is less accurate, are regularized less.

Within each pair, we hold the normal of the more converged frame fixed.
Frames are ordered $\mathrm{prev} \succ \mathrm{curr} \succ \mathrm{next}$ by how long they have been optimized.
Gradients thus reach only the route to the less converged frame: its deformation field, the flow field leading to it, and, through the normals, the shared canonical SDF.
The loss therefore carries the normals of the frame optimized in more windows over to the frame still being fit, instead of changing a frame that is already well reconstructed.
Since $\mathcal{L}_{\mathrm{flow}}$ ties the flow field to the observed motion, the flow cannot slide to points whose normals merely agree.

We compare normals rather than positions for two reasons: the visible detail depends on them, and they are insensitive to small offsets along the surface, which the optical flow cannot resolve on weakly textured regions.

\paragraph{Total loss.} We weight each loss term $\mathcal{L}_{\star}$ by $\lambda_{\star}$ and minimize their sum (weights in App.~\ref{sec:supp_impl}).

\input{figures/main_results}
\subsection{Dense Correspondence Estimation}\label{sec:correspondence}

Per-frame methods give no correspondences, canonical-deformation methods can only match two frames by a nearest-neighbor search in the canonical space, since their deformations are not invertible, and Gaussian-based methods track primitives that do not lie on the extracted surface. Our flow fields instead map the surface of one frame directly onto the next.
After optimization, we extract a mesh $\mathcal{M}_t$ with vertex set $\mathcal{V}_t$ for every frame $t$ by running Marching Cubes~\cite{lorensen1987marchingcubes} on the deformed SDF $f^{(t+1)}_{\mathrm{sdf}}(\delta_t(\mathbf{x}))$.
Frame $t$ is optimized for the last time in the window centered at $t+1$, so we take $f^{(t+1)}_{\mathrm{sdf}}$, $\delta_t = \delta^{t+1}_{\mathrm{prev}}$ and the forward flow $\phi_{t\to t+1} = \phi^{t+1}_{\mathrm{prev}\to\mathrm{curr}}$ from that window, and the flow is consistent with the mesh it is applied to.

\input{tables/main_comparison_ahq_avg}\paragraph{Adjacent frames.}
Marching Cubes extracts each mesh independently, so the vertices of $\mathcal{M}_t$ and $\mathcal{M}_{t+1}$ differ in number and connectivity and cannot be matched by index.
The flow field supplies the missing link: it moves every point of frame $t$ to its position in frame $t+1$.
We move the vertices of $\mathcal{M}_t$ to frame $t+1$ with the flow field, and assign each vertex of $\mathcal{M}_{t+1}$ the vertex of $\mathcal{M}_t$ whose moved position is closest to it:
\begin{equation}
\pi_{t+1\to t}(\mathbf{u}) = \operatorname*{arg\,min}_{\mathbf{v} \in \mathcal{V}_t} \big\lVert \phi_{t\to t+1}(\mathbf{v}) - \mathbf{u} \big\rVert, \qquad \mathbf{u} \in \mathcal{V}_{t+1}.
\label{eq:corr-step}
\end{equation}
The moved vertices lie close to, but not exactly on, $\mathcal{M}_{t+1}$; the nearest-vertex step absorbs this small offset.
Matching in this direction gives every vertex of the later mesh a correspondence, which leaves no holes when attributes are transferred.
Pushing the vertices of $\mathcal{M}_t$ forward instead would leave vertices of $\mathcal{M}_{t+1}$ unmatched wherever the surface stretches.

\paragraph{Distant frames.}
For any two frames $t < s$, we compose the per-frame maps,
\begin{equation}
\pi_{s\to t} = \pi_{t+1\to t} \circ \pi_{t+2\to t+1} \circ \dots \circ \pi_{s\to s-1},
\label{eq:corr-chain}
\end{equation}
and transfer any per-vertex attribute $a_t$ of $\mathcal{M}_t$, such as texture coordinates or colors, by $a_s(\mathbf{u}) = a_t(\pi_{s\to t}(\mathbf{u}))$.
Because each step ends on the vertices of an extracted mesh, the transferred attributes stay on the reconstructed surface instead of drifting away from it.
A point $\mathbf{x}$ of frame $t$ is tracked forward, as in our evaluation, by composing the forward flow fields and reading off the nearest vertex of $\mathcal{M}_s$.

\paragraph{Lite variant.}\label{sec:geo-variant}
When accurate correspondences are not needed, \emph{Ours (Lite)} removes the flow fields, and the normal consistency loss compares pixels matched by the precomputed optical flow instead (App.~\ref{sec:supp_geo}). It needs about a third less training time at almost the same geometric accuracy, with less accurate correspondences (\cref{sec:exp_reconstruction}).

%% file: figures/method_fig.tex
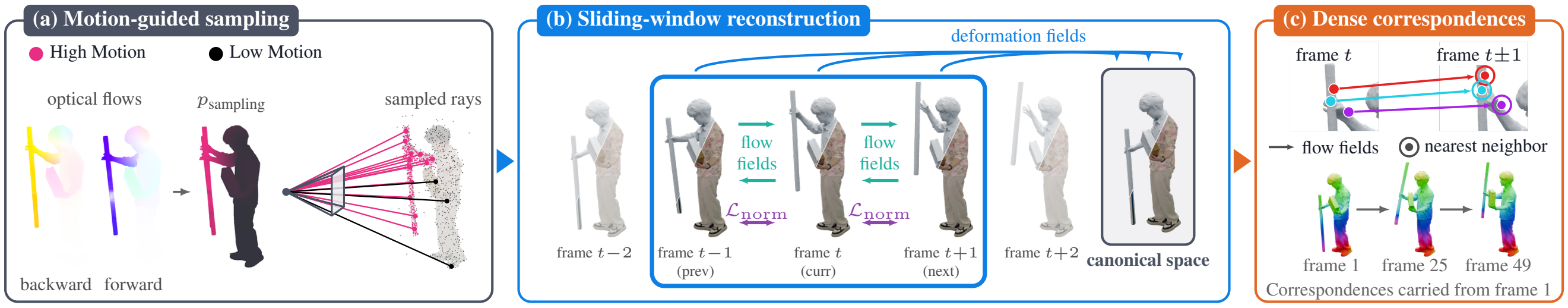
\begin{figure*}[t]
\centering
\usetikzlibrary{arrows.meta}
\resizebox{\textwidth}{!}{\input{figures/method_overview}}
\caption{\textbf{Overview of our FlowSurf4D approach.}
  (a)~Motion-guided ray sampling. (b)~Flow fields between adjacent frames and deformation fields to the canonical space, in a sliding window. (c)~Correspondences: vertices moved by the flow field (ring) and matched to the next mesh (dot).
  The visualised scene is from Jiang et al.~\cite{taogs}. 
}
\label{fig:pipeline}
\vspace{-8pt}\end{figure*}

%% file: figures/method_overview.tex
\providecommand{\methoddir}{imgs/method}\newcommand{\mimg}[2]{\includegraphics[height=#1]{\methoddir/#2}}\definecolor{tzIn}{RGB}{74,82,99}\definecolor{tzOurs}{RGB}{14,128,229}\definecolor{tzApp}{RGB}{226,104,38}\definecolor{tzPink}{RGB}{232,46,130}\definecolor{tzTeal}{RGB}{32,178,160}\definecolor{tzInk}{RGB}{24,28,40}\colorlet{MovePink}{tzPink}\colorlet{mFlow}{tzTeal}\colorlet{mDef}{tzOurs}\colorlet{mSdf}{tzIn}\definecolor{mReg}{RGB}{142,68,173}\begin{tikzpicture}[x=1cm, y=1cm, font=\scriptsize, text=tzInk,
    panel/.style={rounded corners=4pt, line width=0.5pt},
    tab/.style={anchor=west, text=white, rounded corners=2.2pt, inner xsep=3pt, inner ysep=2pt, font=\scriptsize\bfseries},
    lbl/.style={inner sep=1pt, font=\scriptsize},
    tny/.style={inner sep=1pt, font=\tiny},
    lab/.style={inner sep=1pt, font=\fontsize{6}{7}\selectfont},
    img/.style={inner sep=0pt, outer sep=0pt, anchor=south west},
    flowA/.style={-{Latex[length=3pt]}, mFlow, line width=0.8pt},
    defA/.style={-{Latex[length=3pt]}, mDef, line width=0.8pt},
    sdfA/.style={-{Latex[length=3pt]}, mSdf, line width=0.8pt},
    regA/.style={{Latex[length=3pt]}-{Latex[length=3pt]}, mReg, line width=1pt},
    stepA/.style={-{Latex[length=3.5pt]}, black!60, line width=0.7pt},
    loss/.style={rounded corners=2pt, draw=black!25, fill=white, line width=0.4pt}]
\def\mH{2.75cm}

\begin{scope}[shift={(0,3.2)}]
\draw[panel, tzIn, line width=0.8pt] (0,0) rectangle (5.55,3.2);
\node[tab, fill=tzIn] at (0.2,3.2) {(a) Motion-guided sampling};

\def\ix{4.45}\def\iy{0.35}\def\ih{1.8}\def\iw{0.855}\def\ly{2.24}
\coordinate (O) at (3.2,1.25);
\node[img] (FlowB) at (0.15,\iy) {\includegraphics[height=\ih cm]{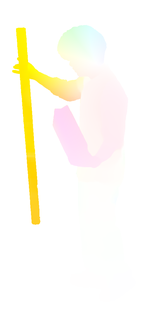}};
\node[img] (FlowF) at (1.03,\iy) {\includegraphics[height=\ih cm]{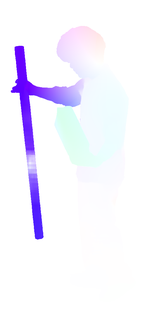}};
\node[lab, anchor=north, text=black!70] at ([yshift=-1pt]FlowB.south) {backward};
\node[lab, anchor=north, text=black!70] at ([yshift=-1pt]FlowF.south) {forward};
\node[img] (Heat) at (2.15,\iy) {\includegraphics[height=\ih cm]{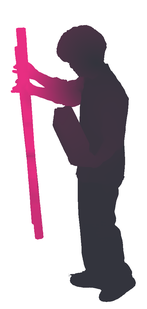}};
\node[img] (Img) at (\ix,\iy) {\includegraphics[height=\ih cm]{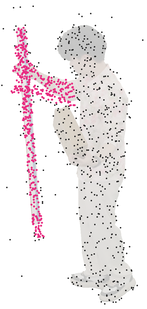}};
\draw[stepA] (1.91,\iy+0.9) -- (2.12,\iy+0.9);
\node[lab, anchor=base, text=black!70] at ($(FlowB.north)!0.5!(FlowF.north)+(0,\ly-\iy-\ih)$) {optical flows};
\node[lab, anchor=base, text=black!70] at ($(Heat.north)+(0,\ly-\iy-\ih)$) {$p_{\text{sampling}}$};
\node[lab, anchor=base, text=black!70] at ($(Img.north)+(0,\ly-\iy-\ih)$) {sampled rays};
\newcommand{\spt}[3]{\draw[#1, line width=0.45pt] (O) -- (\ix+#2*\iw,\iy+#3*\ih) node[circle, fill=#1, inner sep=0.55pt] {};}
\spt{MovePink}{0.133}{0.884}
\spt{MovePink}{0.127}{0.820}
\spt{MovePink}{0.205}{0.741}
\spt{MovePink}{0.328}{0.737}
\spt{MovePink}{0.489}{0.708}
\spt{MovePink}{0.185}{0.700}
\spt{MovePink}{0.412}{0.682}
\spt{MovePink}{0.191}{0.656}
\spt{MovePink}{0.215}{0.561}
\spt{MovePink}{0.189}{0.465}
\spt{MovePink}{0.258}{0.347}
\spt{MovePink}{0.249}{0.265}
\spt{black}{0.533}{0.565}
\spt{black}{0.682}{0.442}
\spt{black}{0.742}{0.029}

\filldraw[draw=tzIn, fill=tzIn!12, fill opacity=0.85, line width=0.6pt, line join=round] (3.72,0.98) -- (3.86,1.08) -- (3.86,1.54) -- (3.72,1.44) -- cycle;
\foreach \c in {(3.72,0.98),(3.86,1.08),(3.86,1.54),(3.72,1.44)} \draw[tzIn, line width=0.6pt] (O) -- \c;
\fill[tzIn] (O) circle (0.045);

\fill[MovePink] (0.35,2.82) circle (0.08); \node[anchor=base west, inner sep=0pt, font=\fontsize{6}{7}\selectfont] at (0.5,2.78) {High Motion};
\fill[black] (2.4,2.82) circle (0.08); \node[anchor=base west, inner sep=0pt, font=\fontsize{6}{7}\selectfont] at (2.55,2.78) {Low Motion};
\end{scope}

\begin{scope}[shift={(1.95,3.2)}]
\draw[panel, tzOurs, line width=0.8pt] (3.9,0) rectangle (12.0,3.2);
\node[tab, fill=tzOurs] at (4.1,3.2) {(b) Sliding-window reconstruction};

\def\bh{1.8cm}
\foreach \k in {0,...,4} {
  \pgfmathsetmacro\bx{4.1+\k*1.4+(\k==0)*0.25-(\k==4)*0.25}
  \ifnum\k=0 \def\bo{0.45}\else\ifnum\k=4 \def\bo{0.45}\else\def\bo{1}\fi\fi
  \node[img, opacity=\bo] (B\k) at (\bx,0.7) {\includegraphics[height=\bh]{\methoddir/b5_\k.png}};
}
\foreach \a/\b in {1/2, 2/3} {
  \draw[flowA, line width=1pt] ([xshift=2pt,yshift=0.42cm]B\a.east) -- ([xshift=-2pt,yshift=0.42cm]B\b.west);
  \draw[flowA, line width=1pt] ([xshift=-2pt,yshift=-0.22cm]B\b.west) -- ([xshift=2pt,yshift=-0.22cm]B\a.east);
  \node[lab, align=center, text=mFlow, inner sep=0pt] at ($([yshift=0.1cm]B\a.east)!0.5!([yshift=0.1cm]B\b.west)$) {flow\\fields};
}
\foreach \a/\b in {1/2, 2/3} {
  \draw[regA] ([xshift=2pt,yshift=-0.7cm]B\a.east) -- ([xshift=-2pt,yshift=-0.7cm]B\b.west);
  \node[lab, anchor=south, text=mReg] at ($([yshift=-0.7cm]B\a.east)!0.5!([yshift=-0.7cm]B\b.west)$) {$\mathcal{L}_{\mathrm{norm}}$};
}
\foreach \k/\l in {0/{frame~$t{-}2$}, 1/{frame~$t{-}1$\\(prev)}, 2/{frame~$t$\\(curr)}, 3/{frame~$t{+}1$\\(next)}, 4/{frame~$t{+}2$}}
  \node[lab, anchor=north, align=center, text=black!70, font=\fontsize{5}{6}\selectfont] at ([yshift=-1pt]B\k.south) {\l};
\draw[tzOurs, rounded corners=4pt, line width=1.4pt] ([shift={(-0.08,-0.52)}]B1.south west) rectangle ([shift={(0.08,0.06)}]B3.north east);
\fill[mSdf!8, rounded corners=3pt] (10.55,0.65) rectangle (11.6,2.65);
\node[img] (Can) at (10.65,0.75) {\includegraphics[height=1.8cm]{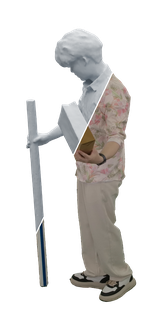}};
\draw[rounded corners=3pt, mSdf, line width=0.8pt] (10.55,0.65) rectangle (11.6,2.65);
\node[lab, anchor=north, text=mSdf] at (11.075,0.6) {\textbf{canonical space}};
\node[lab, text=mDef, anchor=center] at (9.6,3.03) {deformation fields};
\foreach \k/\d in {1/-0.36, 2/0, 3/0.36}
  \draw[defA] ([yshift=0.12cm]B\k.north) .. controls ([yshift=0.4cm]B\k.north) and ([xshift=\d cm,yshift=0.3cm]Can.north) .. ([xshift=\d cm,yshift=5pt]Can.north);
\end{scope}

\begin{scope}[shift={(14.25,3.2)}]
\draw[panel, tzApp, line width=0.8pt] (0,0) rectangle (3.5,3.2);
\node[tab, fill=tzApp] at (0.2,3.2) {(c) Dense correspondences};

\def\Z{1.0}
\node[img, draw=black!20] (Za) at (0.4,1.95) {\includegraphics[width=\Z cm]{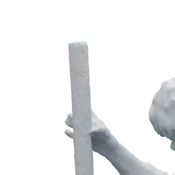}};
\node[img, draw=black!20] (Zb) at (2.1,1.95) {\includegraphics[width=\Z cm]{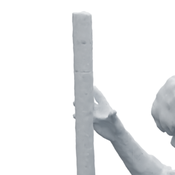}};
\node[lab, anchor=north west, fill=white, inner sep=0.5pt] at ([shift={(1pt,-1pt)}]Za.north west) {frame~$t$};
\node[lab, anchor=north east, fill=white, inner sep=0.5pt] at ([shift={(-1pt,-1pt)}]Zb.north east) {frame~$t{\pm}1$};
\newcommand{\cpt}[7]{\pgfmathsetmacro\hue{mod(#1*0.13,1)}\definecolor{pc}{hsb}{\hue,0.85,0.9}\draw[pc, -{Latex[length=2.5pt]}, line width=0.7pt, shorten >=3.2pt] ($(Za.south west)+(#2*\Z,#3*\Z)$) -- ($(Zb.south west)+(#4*\Z,#5*\Z)$);
  \fill[pc, draw=white, line width=0.3pt] ($(Za.south west)+(#2*\Z,#3*\Z)$) circle (1.7pt);
  \fill[pc, draw=white, line width=0.3pt] ($(Zb.south west)+(#6*\Z,#7*\Z)$) circle (1.5pt);
  \draw[pc, line width=0.7pt] ($(Zb.south west)+(#4*\Z,#5*\Z)$) circle (3pt);}
\cpt{0}{0.4928}{0.4703}{0.5152}{0.6234}{0.5139}{0.6240}
\cpt{4}{0.4567}{0.3325}{0.4782}{0.4585}{0.4787}{0.4570}
\cpt{6}{0.6617}{0.2095}{0.7057}{0.2904}{0.7025}{0.2885}

\draw[-{Latex[length=2.5pt]}, black!70, line width=0.7pt] (0.15,1.76) -- (0.45,1.76);
\node[lab, anchor=west, inner sep=0pt] at (0.52,1.76) {flow fields};
\draw[black!70, line width=0.7pt] (1.75,1.76) circle (3pt); \fill[black!70] (1.75,1.76) circle (1.3pt);
\node[lab, anchor=west, inner sep=0pt] at (1.92,1.76) {nearest neighbor};

\foreach \f/\x in {1/0.62, 25/1.57, 49/2.52}
  \node[img] (C\f) at (\x,0.5) {\includegraphics[height=1.1cm]{\methoddir/c_corr_\f.png}};
\foreach \f in {1, 25, 49} \node[lab, anchor=north, text=black!60] at ([yshift=1pt]C\f.south) {frame \f};
\draw[stepA] ([xshift=1pt]C1.east) -- ([xshift=-1pt]C25.west);
\draw[stepA] ([xshift=1pt]C25.east) -- ([xshift=-1pt]C49.west);
\node[lab, text=black!60] at (1.75,0.12) {Correspondences carried from frame 1};
\end{scope}

\fill[tzOurs] (5.6,4.66) -- (5.8,4.8) -- (5.6,4.94) -- cycle;
\fill[tzApp] (14.0,4.66) -- (14.2,4.8) -- (14.0,4.94) -- cycle;
\end{tikzpicture}

%% file: figures/main_results.tex
\definecolor{ahqBzoom}{HTML}{E82E82}
\newcommand{\ahqBdir}{images}
\newlength{\ahqBcol}\setlength{\ahqBcol}{\dimexpr(0.9\columnwidth-1.6em-4pt)/4\relax}
\newcommand{\ahqBcrop}[7]{\pgfmathsetmacro{\ahqBvx}{#4*#2}\pgfmathsetmacro{\ahqBvy}{(1-#7)*#3}\pgfmathsetmacro{\ahqBvX}{#6*#2}\pgfmathsetmacro{\ahqBvY}{(1-#5)*#3}\edef\ahqBtmp{\noexpand\includegraphics[viewport=\ahqBvx\space\ahqBvy\space\ahqBvX\space\ahqBvY,clip,width=\noexpand\ahqBcol]{#1}}\ahqBtmp}
\newcommand{\ahqBrowahq}{\def\rP{ahq3}\def\rW{696}\def\rH{1170}\def\rBx{0.035}\def\rBa{0.02}\def\rBX{0.60}\def\rBb{0.40}\def\rZs{zoom}\def\rZW{559}\def\rZH{539}\def\rZa{0}\def\rZb{0}\def\rZc{1}\def\rZd{0.778}\def\rXa{0}\def\rXb{0}\def\rXc{0.34}\def\rXd{0.20}}
\newcommand{\ahqBrowdtd}{\def\rP{d4d3}\def\rW{1006}\def\rH{1656}\def\rBx{0.21}\def\rBa{0.16}\def\rBX{0.81}\def\rBb{0.62}\def\rZs{full}\def\rZW{1006}\def\rZH{1656}\def\rZa{0.40}\def\rZb{0.335}\def\rZc{0.62}\def\rZd{0.435}\def\rXa{0.40}\def\rXb{0.335}\def\rXc{0.62}\def\rXd{0.435}}
\newcommand{\ahqBrowtao}{\def\rP{tao3}\def\rW{1392}\def\rH{2160}\def\rBx{0.20}\def\rBa{0.01}\def\rBX{0.80}\def\rBb{0.40}\def\rZs{full}\def\rZW{1392}\def\rZH{2160}\def\rZa{0.596}\def\rZb{0.259}\def\rZc{0.812}\def\rZd{0.363}\def\rXa{0.596}\def\rXb{0.259}\def\rXc{0.812}\def\rXd{0.363}}
\newif\ifahqBzoom\ahqBzoomtrue
\newif\ifahqBbox\ahqBboxtrue
\newcommand{\ahqBcell}[2]{\begin{tikzpicture}
    \node[anchor=south west,inner sep=0] (img) at (0,0) {\ahqBcrop{\ahqBdir/\rP_#1_full.png}{\rW}{\rH}{\rBx}{\rBa}{\rBX}{\rBb}};
    \ifahqBbox\begin{scope}[x={(img.south east)},y={(img.north west)}]
      \pgfmathsetmacro{\ahqBqa}{max(0,(\rXa-\rBx)/(\rBX-\rBx))}\pgfmathsetmacro{\ahqBqc}{min(1,(\rXc-\rBx)/(\rBX-\rBx))}\pgfmathsetmacro{\ahqBqb}{max(0,(\rXb-\rBa)/(\rBb-\rBa))}\pgfmathsetmacro{\ahqBqd}{min(1,(\rXd-\rBa)/(\rBb-\rBa))}\draw[ahqBzoom,line width=0.5pt] (\ahqBqa,{1-\ahqBqb}) rectangle (\ahqBqc,{1-\ahqBqd});
    \end{scope}\fi
    \def\ahqBcd{#2}\ifx\ahqBcd\empty\else\node[font=\tiny,anchor=south east,fill=white,fill opacity=0.8,text opacity=1,inner sep=1pt] at (img.south east) {\ahqBcd};\fi
    \ifahqBzoom\node[anchor=north west,inner sep=0] (zm) at ([yshift=-1.5pt]img.south west) {\ahqBcrop{\ahqBdir/\rP_#1_\rZs.png}{\rZW}{\rZH}{\rZa}{\rZb}{\rZc}{\rZd}};
    \draw[ahqBzoom,line width=0.5pt] (zm.south west) rectangle (zm.north east);\fi
  \end{tikzpicture}}
\newcommand{\ahqBhead}[1]{\makebox[\ahqBcol]{\scriptsize #1}}
\newcommand{\ahqBlabel}[2]{\makebox[1.6em][c]{\rotatebox{90}{\shortstack{\scriptsize\textbf{#1}\\[-1pt]\tiny #2}}}}

\begin{figure}[t]
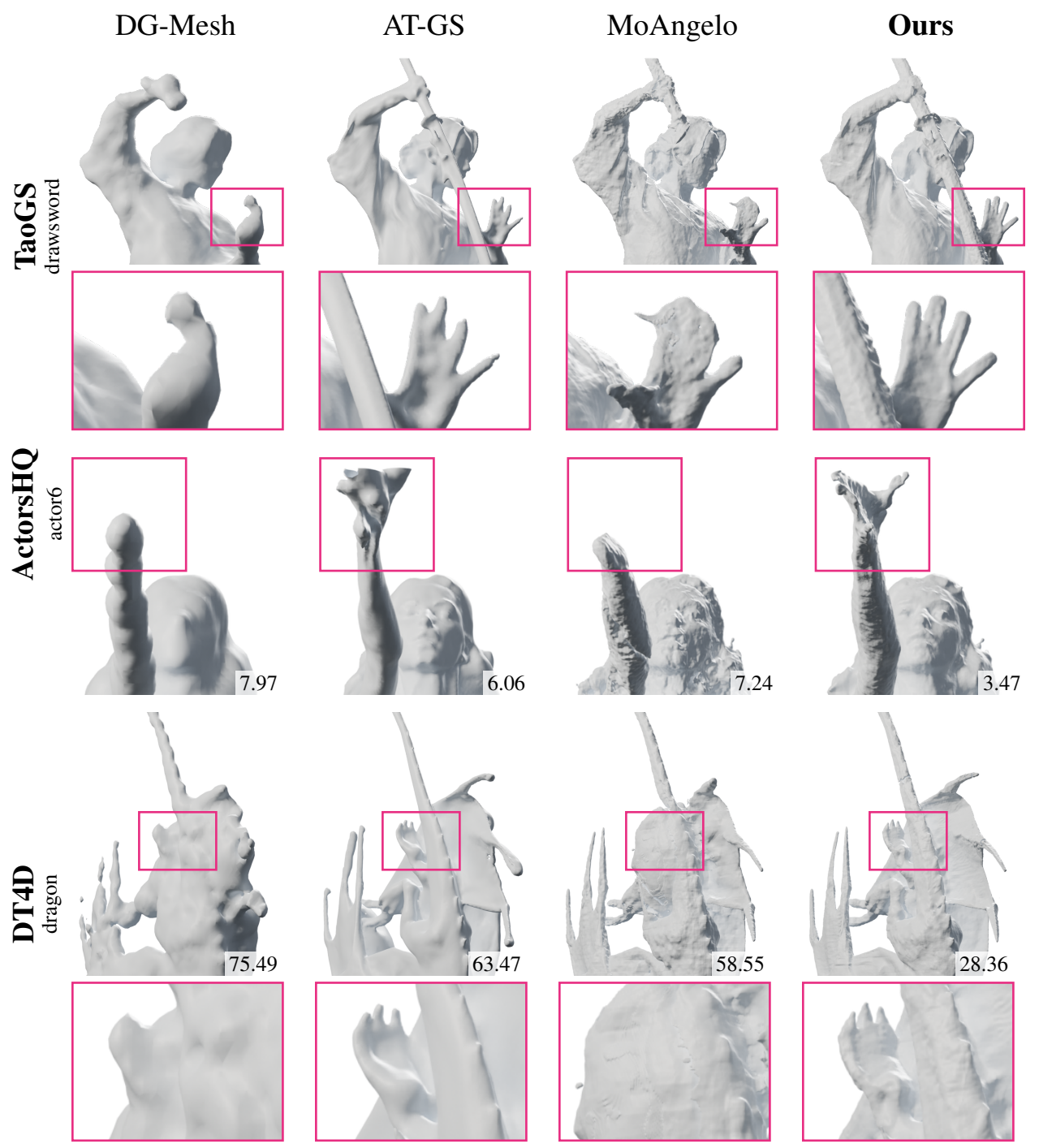

  \centering
  \hspace*{1.6em}\ahqBhead{DG-Mesh}\hfill\ahqBhead{AT-GS}\hfill\ahqBhead{MoAngelo}\hfill\ahqBhead{\textbf{Ours}}\\[2pt]
  \ahqBrowtao\ahqBzoomtrue\ahqBboxtrue\raisebox{0.75\ahqBcol}{\ahqBlabel{TaoGS}{drawsword}}\ahqBcell{dgmesh}{}\hfill\ahqBcell{atgs}{}\hfill\ahqBcell{moangelo}{}\hfill\ahqBcell{ours}{}\\[3pt]
  \ahqBrowahq\ahqBzoomfalse\ahqBboxtrue\raisebox{0.55\ahqBcol}{\ahqBlabel{ActorsHQ}{actor6}}\ahqBcell{dgmesh}{7.97}\hfill\ahqBcell{atgs}{6.06}\hfill\ahqBcell{moangelo}{7.24}\hfill\ahqBcell{ours}{3.47}\\[3pt]
  \ahqBrowdtd\ahqBzoomtrue\raisebox{0.95\ahqBcol}{\ahqBlabel{DT4D}{dragon}}\ahqBcell{dgmesh}{75.49}\hfill\ahqBcell{atgs}{63.47}\hfill\ahqBcell{moangelo}{58.55}\hfill\ahqBcell{ours}{28.36}
\vspace{-4pt}\caption{\textbf{Qualitative comparison} on TaoGS, ActorsHQ and DeformingThings4D; boxes mark the compared regions, enlarged below on TaoGS and DeformingThings4D. Chamfer distance (mm) of the shown frame in the corners.}
  \label{fig:ahq_actor6}
\vspace{-8pt}\end{figure}

%% file: tables/main_comparison_ahq_avg.tex
\newcommand{\crY}{\ding{51}}\newcommand{\crN}{\ding{55}}\newcommand{\crQ}{(\ding{51})}
\begin{table}[t]
\centering
\footnotesize
\setlength{\tabcolsep}{2.5pt}
\resizebox{\linewidth}{!}{\begin{tabular}{@{}lc ccc|cccc@{}}
\toprule
 & & \multicolumn{3}{c|}{ActorsHQ} & \multicolumn{4}{c}{DeformingThings4D} \\
\cmidrule(lr){3-5}\cmidrule(l){6-9}
Method & Corr. & CD$\downarrow$ & NC$\uparrow$ & F@1$\uparrow$ & CD$\downarrow$ & NC$\uparrow$ & F@1$\uparrow$ & TCE$\downarrow$ \\
\midrule
\multicolumn{9}{@{}l}{\emph{Dynamic, SDF-based}} \\
Tensor4D$^\dagger$~\cite{shao2023tensor4d} & \crQ & 8.32 & 0.884 & 0.831 & 16.38 & 0.863 & 0.662 & 6.65 \\
NeuS2 (dynamic)~\cite{wang2023neus2} & \crN & 4.05 & 0.915 & 0.959 & 10.82 & 0.907 & \rankC{0.803} & 5.05 \\
MoAngelo~\cite{ebbed2026moangelo} & \crQ & \rankC{3.62} & 0.916 & \rankC{0.962} & \rankC{9.51} & 0.878 & 0.782 & 5.13 \\
\midrule
\multicolumn{9}{@{}l}{\emph{Dynamic, NeRF-based}} \\
SceNeRFlow~\cite{tretschk2024scenerflow} & \crQ & 16.45 & 0.705 & 0.530 & 37.12 & 0.583 & 0.303 & 18.98 \\
\midrule
\multicolumn{9}{@{}l}{\emph{Dynamic, Gaussian-based}} \\
Dynamic 3DGS~\cite{Luiten2024} & \crQ & 7.29 & 0.735 & 0.798 & 63.69 & 0.725 & 0.528 & 8.65 \\
Dynamic-2DGS$^\dagger$~\cite{zhang2024dynamic2dgaussiansgeometrically} & \crQ & 48.53 & 0.616 & 0.405 & 23.94 & 0.714 & 0.559 & 14.39 \\
AT-GS~\cite{chen2025atgs} & \crQ & 5.18 & \rankC{0.921} & 0.941 & 18.49 & \rankC{0.911} & 0.773 & \rankC{3.18} \\
4DGS~\cite{4dgaussians} & \crQ & 10.65 & 0.696 & 0.652 & 19.57 & 0.709 & 0.486 & 12.98 \\
SC-GS~\cite{huang2024scgs} & \crQ & 6.81 & 0.767 & 0.833 & 12.20 & 0.768 & 0.660 & 6.70 \\
Deformable 3DGS$^\dagger$~\cite{yang2024deformable} & \crQ & 8.08 & 0.744 & 0.748 & 14.90 & 0.747 & 0.564 & 8.81 \\
\midrule
\multicolumn{9}{@{}l}{\emph{Dynamic, mesh-based Gaussians}} \\
GauSTAR~\cite{zheng2025gaustar} & \crQ & 6.46 & 0.895 & 0.915 & 149.20 & 0.790 & 0.569 & 28.40 \\
DG-Mesh$^\dagger$~\cite{Liu2025} & \crQ & 6.80 & 0.871 & 0.828 & 22.82 & 0.826 & 0.585 & 9.88 \\
MaGS$^\dagger$~\cite{ma2024magsreconstructingsimulatingdynamic} & \crY & 15.06 & 0.725 & 0.481 & 26.65 & 0.696 & 0.307 & 10.26 \\
\midrule
\multicolumn{9}{@{}l}{\emph{Per-frame static (no correspondences)}} \\
NeuS2 (static)~\cite{wang2023neus2} & \crN & 4.12 & 0.897 & 0.949 & 10.30 & 0.901 & 0.803 & 4.77 \\
2DGS~\cite{Huang2DGS2024} & \crN & 3.74 & 0.907 & 0.981 & 8.30 & 0.904 & 0.822 & 3.68 \\
PGSR~\cite{chen2024pgsr} & \crN & 2.92 & 0.924 & 0.989 & 15.81 & 0.856 & 0.620 & 8.10 \\
\midrule
\textbf{Ours (Lite)} & \crQ & \rankB{3.17} & \rankB{0.939} & \rankB{0.978} & \rankB{7.90} & \rankB{0.926} & \rankB{0.829} & \rankB{3.16} \\
\textbf{Ours} & \crY & \rankA{3.12} & \rankA{0.940} & \rankA{0.980} & \rankA{7.62} & \rankA{0.930} & \rankA{0.834} & \rankA{2.87} \\
\bottomrule
\end{tabular}}
\vspace{-4pt}\caption{\textbf{Geometry} on 6 ActorsHQ and 10 DeformingThings4D scenes (CD, TCE in mm). \ranklegend{} among the dynamic methods. Corr.: \ding{51}~direct, (\ding{51})~indirect, \ding{55}~none. $^\dagger$fewer scenes (the method failed on the others)}
\label{tab:geometry_avg}
\vspace{-8pt}\end{table}

%% file: sec/4_experiments.tex
\section{Experiments}\label{sec:experiments}

We evaluate the accuracy of our surfaces (\cref{sec:exp_reconstruction}), of their correspondences (\cref{sec:exp_correspondence}), and conduct an ablation study (\cref{sec:exp_ablation}).
We use 50-frame sequences: real captures with ground-truth geometry (ActorsHQ), synthetic ones that add ground-truth correspondences (DeformingThings4D), and fast-moving real captures (TaoGS).

\subsection{Experimental Setup}

\paragraph{Datasets.}
ActorsHQ~\cite{isik2023humanrf} contains real captures of humans with ground-truth meshes; we use 6 sequences.
DeformingThings4D~\cite{li2021deformingthings4d} includes animated synthetic characters, humanoids and animals, whose ground-truth meshes share the same vertices over time; we render 10 sequences from 50 cameras (App.~\ref{sec:supp_data}).
TaoGS~\cite{taogs} contains real captures of humans with fast motion, loose clothing and objects, but no ground-truth geometry; on 6 sequences, we evaluate the correspondences against 2D point tracks in held-out cameras.

\input{tables/correspondence}
\paragraph{Baselines.}
We compare with 13 dynamic reconstruction methods based on SDFs, on a NeRF, on Gaussians or surfels, and on Gaussians attached to a mesh (\Cref{tab:geometry_avg}), and run the static methods NeuS2~\cite{wang2023neus2}, 2DGS~\cite{Huang2DGS2024} and PGSR~\cite{chen2024pgsr} independently per frame as a reference for single-frame accuracy.
As each of our frames is optimized in three windows, MoAngelo, which tracks one frame at a time, is run with three times as many iterations per frame, so that both methods perform the same number of iterations per frame.
4DSurf~\cite{wu20264dsurf}, Space-time 2DGS~\cite{wang2024st2dgs} and DynaSurfGS~\cite{cai2024dynasurfgs} are not compared, as their code is not available, nor is NDR~\cite{cai2022neuralsurface}, which uses a single RGB-D camera. 

\paragraph{Metrics.}
We report the Chamfer distance (CD, mm) to the ground-truth meshes, with its two directions, accuracy (Acc., reconstruction to ground truth) and completeness (Compl.), normal consistency (NC), and the F-score at 1\,cm (F@1).
On DeformingThings4D, the temporal consistency error (TCE) measures how much the offset between a ground-truth surface point and the reconstruction changes between adjacent frames (App.~\ref{sec:supp_tce}).
The normal angle error (NAE) compares nearest-point normals of reconstruction and ground truth (App.~\ref{sec:supp_normal_metrics}).

\input{figures/corr_combined}\subsection{Reconstruction}\label{sec:exp_reconstruction}

Among dynamic methods (\Cref{tab:geometry_avg}), ours achieves the lowest Chamfer distance and the highest normal consistency on both datasets, with the lowest Chamfer distance on every scene (App.~\cref{tab:perscene_cd,tab:perscene_nc,tab:perscene_tce,fig:geo_cmp_1,fig:geo_cmp_2,fig:geo_cmp_3}), and the lowest temporal consistency error on DeformingThings4D.
Against the most accurate dynamic baseline, MoAngelo~\cite{ebbed2026moangelo}, it reduces the Chamfer distance from 3.62 to 3.12\,mm on ActorsHQ and from 9.51 to 7.62\,mm on DeformingThings4D; its temporal consistency error is 2.87\,mm, against 3.18\,mm for the most consistent baseline, AT-GS~\cite{chen2025atgs}.
The dynamic Gaussian-based methods render held-out views slightly worse than ours (App.~\cref{tab:nvs,tab:perscene_psnr,tab:perscene_ssim,tab:perscene_lpips,fig:nvs_cmp_1,fig:nvs_cmp_2}) and have a clearly lower normal consistency, as their geometry is a by-product of image fitting.
The per-frame static methods reach a lower Chamfer distance on ActorsHQ (PGSR 2.92 vs.\ 3.12\,mm) but a higher one on DeformingThings4D (15.81 vs.\ 7.62\,mm) and a higher temporal consistency error (2DGS 3.68 and PGSR 8.10 vs.\ 2.87\,mm).
Ours (Lite), which pairs pixels by optical flow instead of flow fields (\cref{sec:geo-variant}), is almost as accurate (3.17 vs.\ 3.12\,mm), so the flow fields contribute the correspondences (4.14 vs.\ 5.48, \Cref{tab:correspondence}) at 48\% more time per frame (43 vs.\ 29 minutes).
In \Cref{fig:ahq_actor6}, ours recovers thin structures such as fingers and the dragon's raised hand, which the baselines miss or make noisy.

\subsection{Correspondence Estimation}\label{sec:exp_correspondence}
\input{figures/ablation_main}

\paragraph{Protocol.}
On DeformingThings4D, the ground-truth meshes share one topology, so every surface point has a known position in every frame.
We track 5000 points sampled uniformly on the ground-truth surface of frame~1 and report, over all 10 scenes, their mean geodesic error on the ground-truth surface of frame~$k$ (shape scaled to unit area~\cite{kim2011blended}) and the area under the curve (AUC) of the fraction of points below an error threshold, up to 0.1~\cite{cao2023ulrssm}.
The correspondences of each method are read off its own reconstruction (App.~\ref{sec:supp_corr}).
For reference, we also evaluate shape matching~\cite{ovsjanikov2012functional,cao2023ulrssm,bastianxie2024hybrid} on ground-truth and reconstructed meshes (\cref{fig:corr_fig}, right).

\paragraph{Results.}
Among reconstruction methods (\Cref{tab:correspondence,fig:corr_fig}), our correspondences have a mean geodesic error of 4.14, against 7.26 for Dynamic 3DGS~\cite{Luiten2024}, 7.56 for SC-GS~\cite{huang2024scgs} and 7.58 for MoAngelo~\cite{ebbed2026moangelo}; SC-GS and Dynamic 3DGS are more accurate on 5 and 4 of the 10 scenes (median 3.53 and 6.11 vs.\ 3.82), but their surfaces are less accurate than ours on all 10 scenes in CD, NC and TCE (App.~\ref{sec:supp_detailed}).
Of all methods with correspondences, ours has the lowest mean Chamfer distance and correspondence error (\Cref{fig:cd_vs_corr}).
Relating the frames of our own model through its canonical space instead gives 5.08 (\Cref{tab:correspondence}): on the same surfaces, the flow fields reduce the error by 18\%, since the deformations of different frames, optimized in different windows, need not agree in the canonical space.
Lifting the same optical flow onto our meshes instead of the flow fields gives 10.5 (App.~\ref{sec:supp_liftflow}): the accuracy comes from the jointly optimized flow fields, not from the optical flow alone.
On real captures (TaoGS, ActorsHQ), our correspondences projected into held-out cameras deviate from CoTracker3~\cite{karaev2024cotracker3} tracks by a median of 3.20 and 3.21 pixels, against 4.16 and 4.53 for the canonical route, and less than those of every other reconstruction method on the same pairs (App.~\ref{sec:supp_realdata}).
Our mean error is the lowest of all reconstruction methods in every frame (App.~\ref{sec:supp_corr}), and the correspondences transfer a texture or an albedo, which we relight, over the sequence (App.~\ref{sec:supp_applications} and video).

\subsection{Ablation Study}\label{sec:exp_ablation}

\input{tables/ablation}

The ablation (\Cref{tab:ablation,fig:ablation_main,fig:corr_fig}) removes one component at a time.
$\mathcal{L}_{\mathrm{norm}}$ improves the normals: without it, the normal angle error grows on each DeformingThings4D scene, on vampire by more than four times its variation over three seeds (Table~\ref{tab:seeds}), while the Chamfer distance is unchanged on average (on vampire, $\mathcal{L}_{\mathrm{norm}}$ raises it by about 0.2\,mm in exchange for the better normals).
The flow fields provide the correspondences: without them, only the much less accurate canonical route remains, and the normal and temporal consistency errors increase.
$\mathcal{L}_{\mathrm{flow}}$ makes the flow fields follow the motion: without it, $\mathcal{L}_{\mathrm{norm}}$ moves them to points whose normals agree but that do not correspond, which degrades both the correspondences and the geometry.
The motion-guided sampling improves the geometry on every scene and the temporal consistency, while the correspondence error is slightly lower without it, and ARAP and smoothness keep the optimization stable: without them, the geometry collapses on three of the six scenes.
The effects of $\mathcal{L}_{\mathrm{flow}}$ and the sampling exceed the seed variation of our full model (0.04\,mm on an ActorsHQ scene, Table~\ref{tab:seeds}).
As the metrics are averaged over all surface points and frames, local errors change them only slightly; these differences are better visible in \cref{fig:ablation_main}, App.~\ref{sec:supp_detailed} and the video.

%% file: tables/correspondence.tex
\begin{table}[t]
\centering
\footnotesize
\setlength{\tabcolsep}{3pt}
\begin{tabular}{@{}lcccc@{}}
\toprule
Method & All & Early & Late & AUC$\uparrow$ \\
\midrule
Tensor4D~\cite{shao2023tensor4d} & 16.73 & 10.79 & 18.50 & 0.325 \\
SceNeRFlow~\cite{tretschk2024scenerflow} & 17.28 & 5.86 & 20.68 & 0.337 \\
GauSTAR~\cite{zheng2025gaustar} & 17.53 & 6.89 & 20.70 & 0.313 \\
Dynamic 3DGS~\cite{Luiten2024} & 7.26 & 3.48 & 8.38 & 0.622 \\
4DGS~\cite{4dgaussians} & 17.84 & 7.79 & 20.83 & 0.335 \\
SC-GS~\cite{huang2024scgs} & 7.56 & 2.76 & 8.98 & \rankC{0.632} \\
Deformable 3DGS~\cite{yang2024deformable}$^\dagger$ & 11.61 & 3.77 & 13.94 & 0.521 \\
Dynamic-2DGS~\cite{zhang2024dynamic2dgaussiansgeometrically}$^\dagger$ & 15.78 & 8.02 & 18.08 & 0.243 \\
DG-Mesh~\cite{Liu2025}$^\dagger$ & 15.20 & 5.93 & 17.96 & 0.422 \\
MaGS~\cite{ma2024magsreconstructingsimulatingdynamic}$^\dagger$ & 11.32 & 3.71 & 13.59 & 0.530 \\
MoAngelo~\cite{ebbed2026moangelo} & 7.58 & 2.73 & 9.02 & 0.542 \\
\textbf{Ours} (Lite) & \rankC{5.48} & \rankC{2.39} & \rankC{6.40} & 0.622 \\
\textbf{Ours}, canonical route & \rankB{5.08} & \rankB{2.36} & \rankB{5.89} & \rankB{0.645} \\
\textbf{Ours} & \rankA{4.14} & \rankA{1.92} & \rankA{4.80} & \rankA{0.695} \\
\bottomrule
\end{tabular}
\vspace{-4pt}\caption{\textbf{Correspondences on reconstructed surfaces} (DeformingThings4D): geodesic error ($\times 100$) over the 50 frames (All), the first quarter of the sequence (Early) and the rest (Late). \ranklegend. $^\dagger$Fewer scenes (the method failed on the others).
}
\label{tab:correspondence}
\vspace{-8pt}\end{table}

%% file: figures/corr_combined.tex
\begin{figure*}[t]
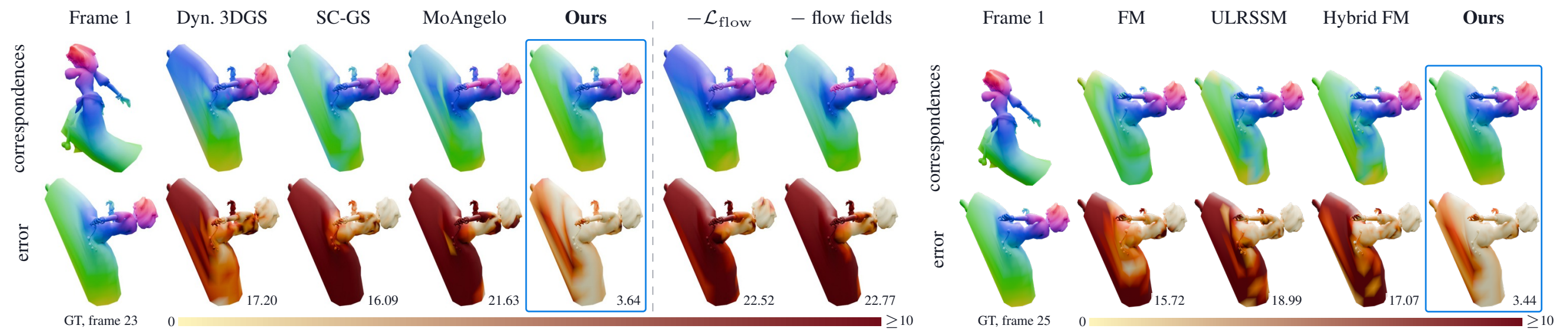

  \centering
  \definecolor{tzIn}{RGB}{74,82,99}\definecolor{tzOurs}{RGB}{14,128,229}\definecolor{tzInk}{RGB}{24,28,40}
  \begin{minipage}[b]{0.585\textwidth}\centering
  
  \newcommand{\corrdir}{imgs/corr_native}
  \newcommand{\abg}{imgs/ablation_fig}
  \resizebox{\linewidth}{!}{\begin{tikzpicture}[x=1cm, y=1cm, font=\scriptsize, text=tzInk,
      img/.style={anchor=south west, inner sep=0}]
    \def\cw{1.52}\def\gap{0.02}\def\lw{0.30}\def\sep{0.16}
    \def\chh{1.705}
    \foreach \c/\n/\t in {0/src/{Frame 1}, 1/d3dg/{Dyn.\ 3DGS}, 2/scgs/{SC-GS}, 3/moangelo/{MoAngelo}, 4/ours/{\textbf{Ours}}}
      \node[anchor=south] at (\lw+\c*\cw+\c*\gap+0.5*\cw, 2*\chh+0.02) {\strut\t};
    \foreach \c/\n in {0/src, 1/d3dg, 2/scgs, 3/moangelo, 4/ours}
      \node[img] at (\lw+\c*\cw+\c*\gap, \chh) {\includegraphics[width=\cw cm]{\corrdir/\n}};
    \foreach \c/\n in {0/ref, 1/err_d3dg, 2/err_scgs, 3/err_moangelo, 4/err_ours}
      \node[img] at (\lw+\c*\cw+\c*\gap, 0) {\includegraphics[width=\cw cm]{\corrdir/\n}};
    \foreach \c/\n/\t in {5/noflow/{$-\mathcal{L}_{\mathrm{flow}}$}, 6/nocross/{$-$ flow fields}} {
      \node[anchor=south] at (\lw+\sep+\c*\cw+\c*\gap+0.5*\cw, 2*\chh+0.02) {\strut\t};
      \node[img] at (\lw+\sep+\c*\cw+\c*\gap, \chh) {\includegraphics[width=\cw cm]{\abg/corr_\n}};
      \node[img] at (\lw+\sep+\c*\cw+\c*\gap, 0) {\includegraphics[width=\cw cm]{\abg/corr_err_\n}};
    }
    \draw[tzIn!50, line width=0.4pt, dashed] (\lw+5*\cw+4.5*\gap+0.5*\sep, -0.02) -- (\lw+5*\cw+4.5*\gap+0.5*\sep, 2*\chh+0.30);
    \draw[tzOurs, line width=0.6pt, rounded corners=1pt] (\lw+4*\cw+4*\gap, -0.02) rectangle (\lw+5*\cw+4*\gap, 2*\chh+0.02);
    \node[rotate=90, anchor=south] at (\lw-0.04, 1.5*\chh) {correspondences};
    \node[rotate=90, anchor=south] at (\lw-0.04, 0.5*\chh) {error};
    \node[font=\tiny, anchor=north, inner sep=1pt] at (\lw+0.5*\cw, -0.03) {GT, frame 23};
    \foreach \c/\v in {1/17.20, 2/16.09, 3/21.63, 4/3.64}
      \node[font=\tiny, anchor=south east, fill=white, fill opacity=0.8, text opacity=1, inner sep=1pt] at (\lw+\c*\cw+\c*\gap+\cw-0.02, 0.02) {\v};
    \foreach \c/\v in {5/22.52, 6/22.77}
      \node[font=\tiny, anchor=south east, fill=white, fill opacity=0.8, text opacity=1, inner sep=1pt] at (\lw+\sep+\c*\cw+\c*\gap+\cw-0.02, 0.02) {\v};
    \shade[left color={rgb,255:red,255;green,247;blue,189}, middle color={rgb,255:red,252;green,153;blue,36}, right color={rgb,255:red,115;green,13;blue,26}] (\lw+\cw+\gap+0.2,-0.20) rectangle (\lw+\sep+7*\cw+6*\gap-0.25,-0.10);
    \node[anchor=east, inner sep=1pt, font=\tiny] at (\lw+\cw+\gap+0.2,-0.15) {0};
    \node[anchor=west, inner sep=1pt, font=\tiny] at (\lw+\sep+7*\cw+6*\gap-0.25,-0.15) {$\geq$10};
  \end{tikzpicture}}
  \end{minipage}\hfill
  \begin{minipage}[b]{0.409\textwidth}\centering
  
  \newcommand{\corrdir}{imgs/corr}
  \resizebox{\linewidth}{!}{\begin{tikzpicture}[x=1cm, y=1cm, font=\scriptsize, text=tzInk,
      img/.style={anchor=south west, inner sep=0}]
    \def\cw{1.49}\def\gap{0.02}\def\lw{0.32}
    \def\chh{1.563}
    \foreach \c/\n/\t in {0/src/{Frame 1}, 1/fm/{FM}, 2/ulrssm/{ULRSSM}, 3/hybrid/{Hybrid FM}, 4/ours/{\textbf{Ours}}}
      \node[anchor=south] at (\lw+\c*\cw+\c*\gap+0.5*\cw, 2*1.705+0.0675) {\strut\t};
    \foreach \c/\n in {0/src, 1/fm, 2/ulrssm, 3/hybrid, 4/ours}
      \node[img] at (\lw+\c*\cw+\c*\gap, \chh) {\includegraphics[width=\cw cm]{\corrdir/\n.png}};
    \foreach \c/\n in {0/ref, 1/err_fm, 2/err_ulrssm, 3/err_hybrid, 4/err_ours}
      \node[img] at (\lw+\c*\cw+\c*\gap, 0) {\includegraphics[width=\cw cm]{\corrdir/\n.png}};
    \draw[tzOurs, line width=0.6pt, rounded corners=1pt] (\lw+4*\cw+4*\gap, -0.02) rectangle (\lw+5*\cw+4*\gap, 2*\chh+0.02);
    \node[rotate=90, anchor=south] at (\lw-0.04, 1.5*\chh) {correspondences};
    \node[rotate=90, anchor=south] at (\lw-0.04, 0.5*\chh) {error};
    
    \node[font=\tiny, anchor=north, inner sep=1pt] at (\lw+0.5*\cw, -0.03) {GT, frame 25};
    \foreach \c/\v in {1/15.72, 2/18.99, 3/17.07, 4/3.44}
      \node[font=\tiny, anchor=south east, fill=white, fill opacity=0.8, text opacity=1, inner sep=1pt] at (\lw+\c*\cw+\c*\gap+\cw-0.02, 0.02) {\v};
    \shade[left color={rgb,255:red,255;green,247;blue,189}, middle color={rgb,255:red,252;green,153;blue,36}, right color={rgb,255:red,115;green,13;blue,26}] (\lw+\cw+\gap+0.2,-0.20) rectangle (\lw+5*\cw+4*\gap-0.25,-0.10);
    \node[anchor=east, inner sep=1pt, font=\tiny] at (\lw+\cw+\gap+0.2,-0.15) {0};
    \node[anchor=west, inner sep=1pt, font=\tiny] at (\lw+5*\cw+4*\gap-0.25,-0.15) {$\geq$10};
  \end{tikzpicture}}
  \end{minipage}
\vspace{-4pt}
  \caption{\textbf{Correspondence evaluation} on DeformingThings4D peasantgirl: points of frame~1 carried to a later frame, colored by their frame-1 position and by their geodesic error ($\times 100$, mean in the corner). Left: reconstruction competitors and ablation. Right: shape matching, run on the ground-truth meshes; over all scenes, chained Hybrid FM is slightly more accurate than ours (App.~\ref{sec:supp_corr}).}
  \label{fig:corr_fig}
\vspace{-8pt}
\end{figure*}

%% file: figures/ablation_main.tex
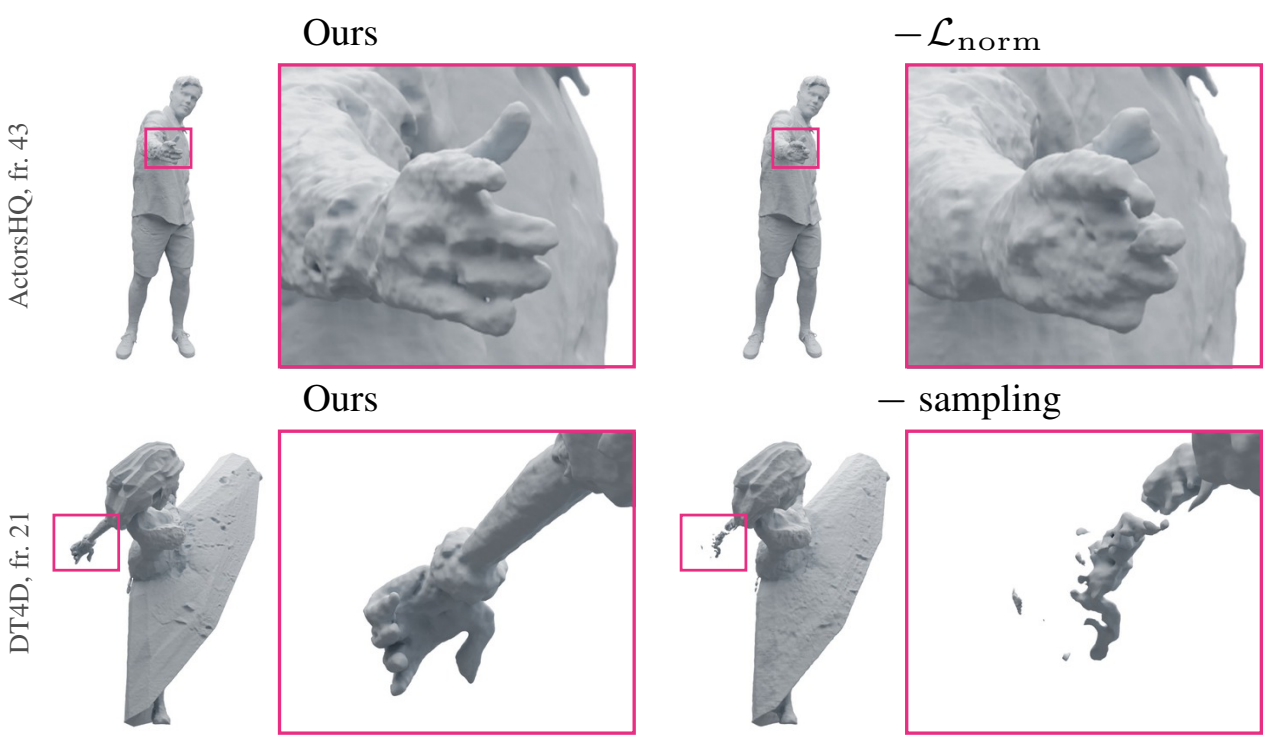
\begin{figure}[t]
\centering
\definecolor{mabA}{RGB}{232,46,130}
\newcommand{\abh}{imgs/ablation_fig2}
\resizebox{\columnwidth}{!}{\begin{tikzpicture}[x=1in,y=1in,font=\scriptsize,
    hd/.style={anchor=south,inner sep=1pt},
    rl/.style={rotate=90,anchor=south,inner sep=1pt,font=\tiny,text=black!70}]
\def\zh{0.75}\def\zw{0.881}\def\g{0.03}\def\pg{0.10}
\def\fs{0.547}\newcommand{\cell}[9]{\pgfmathsetmacro{\fw}{\zh/#5}\pgfmathsetmacro{\fx}{#1+0.5*(\fs-\fw)}\node[anchor=north west,inner sep=0pt] at (\fx,#2) {\includegraphics[height=\zh in]{\abh/full_#3_#4}};
  \draw[mabA,line width=0.5pt] (\fx+#6*\fw,#2-#7*\zh) rectangle (\fx+#8*\fw,#2-#9*\zh);
  \node[anchor=north west,inner sep=0pt] at (#1+\fs+\g,#2) {\includegraphics[width=\zw in]{\abh/zoom_#3_#4}};
  \draw[mabA,line width=0.6pt] (#1+\fs+\g,#2) rectangle (#1+\fs+\g+\zw,#2-\zh);
}
\pgfmathsetmacro{\w}{\fs+\g+\zw}
\pgfmathsetmacro{\yb}{-\zh-0.16}
\cell{0}{0}{a8}{ours}{3.3226}{0.3724}{0.2120}{0.8710}{0.3398}
\cell{\w+\pg}{0}{a8}{nolnorm}{3.3226}{0.3724}{0.2120}{0.8710}{0.3398}
\node[hd] at (0.5*\w,0) {\strut Ours};
\node[hd] at (\w+\pg+0.5*\w,0) {\strut $-\mathcal{L}_{\mathrm{norm}}$};
\node[rl] at (-0.02,-0.5*\zh) {ActorsHQ, fr.\ 43};
\cell{0}{\yb}{pg21}{ours}{1.3709}{0.0304}{0.2777}{0.3246}{0.4604}
\cell{\w+\pg}{\yb}{pg21}{nosampling}{1.3709}{0.0304}{0.2777}{0.3246}{0.4604}
\node[hd] at (0.5*\w,\yb) {\strut Ours};
\node[hd] at (\w+\pg+0.5*\w,\yb) {\strut $-$ sampling};
\node[rl] at (-0.02,\yb-0.5*\zh) {DT4D, fr.\ 21};
\end{tikzpicture}}
\vspace{-4pt}\caption{\textbf{Ablation of the geometry} (\Cref{tab:ablation}): full meshes and close-ups of the boxed regions. Without $\mathcal{L}_{\mathrm{norm}}$, the hand loses its shape (ActorsHQ); without the motion-guided sampling, the moving hand falls apart (DeformingThings4D). More visualizations: App.~\ref{sec:supp_detailed} and video.}
\label{fig:ablation_main}
\vspace{-8pt}\end{figure}

%% file: tables/ablation.tex
\begin{table}[t]
\centering
\footnotesize
\setlength{\tabcolsep}{3pt}
\resizebox{\columnwidth}{!}{\begin{tabular}{@{}lccccccccc@{}}
\toprule
& \multicolumn{4}{c}{ActorsHQ} & \multicolumn{5}{c}{DeformingThings4D} \\
\cmidrule(lr){2-5}\cmidrule(l){6-10}
Variant & CD$\downarrow$ & Acc.$\downarrow$ & Compl.$\downarrow$ & NC$\uparrow$ & CD$\downarrow$ & NC$\uparrow$ & NAE$\downarrow$ & TCE$\downarrow$ & Corr.$\downarrow$ \\
\midrule
\textbf{Ours} (full model) & \rankC{3.30} & \rankA{3.19} & \rankC{3.42} & \rankA{0.934} & \rankB{8.43} & \rankA{0.894} & \rankA{18.6} & \rankA{3.03} & \rankB{3.40} \\
\quad $-$ $\mathcal{L}_{\mathrm{norm}}$ & \rankB{3.29} & \rankB{3.23} & \rankB{3.36} & \rankB{0.933} & \rankC{8.45} & 0.884 & 19.8 & \rankB{3.25} & \rankC{3.42} \\
\quad $-$ $\mathcal{L}_{\mathrm{flow}}$ & 3.47 & \rankC{3.27} & 3.68 & 0.930 & 8.78 & \rankB{0.890} & \rankB{19.2} & 3.34 & 8.69 \\
\quad $-$ flow fields & \rankA{3.27} & \rankA{3.19} & \rankA{3.35} & \rankB{0.933} & \rankA{8.42} & 0.886 & \rankC{19.6} & \rankC{3.33} & 7.69$^*$ \\
\quad $-$ motion-guided sampling & 3.48 & \rankB{3.23} & 3.72 & \rankC{0.932} & 9.56 & \rankC{0.889} & 19.7 & 4.74 & \rankA{3.19} \\
\quad $-$ ARAP, smoothness & 10.18 & 13.56 & 6.81 & 0.812 & 10.73 & 0.825 & 26.3 & 6.79 & 7.79 \\
\bottomrule
\end{tabular}}
\vspace{-4pt}\caption{\textbf{Ablations} on three ActorsHQ and three DeformingThings4D scenes; each row removes one component ($-$ flow fields also removes $\mathcal{L}_{\mathrm{flow}}$ and $\mathcal{L}_{\mathrm{norm}}$). \ranklegend{} per column. $^*$Through the canonical space.}
\label{tab:ablation}
\vspace{-8pt}\end{table}

%% file: sec/conclusions.tex
\section{Conclusion} \label{sec:conclusion}

We presented FlowSurf4D, the first method that provides highly accurate reconstructions with dense correspondences among dynamic methods.
It reconstructs the most accurate surfaces among dynamic methods on every scene, with a 43\% lower mean correspondence error than the next best dynamic method and the closest to 2D point tracks on real captures.
Explicit flow fields between adjacent frames, supervised by optical flow, provide the correspondences and, within a sliding window, enable temporal regularization of adjacent surfaces that improves their normals and temporal consistency, while motion-guided sampling of moving regions makes the surfaces accurate.
The correspondences are accurate enough for texture transfer, relighting and placing the textured reconstructions in virtual environments.

\paragraph{Limitations.}
FlowSurf4D trades running time for accuracy: it gives the most accurate surfaces and the lowest mean correspondence error among dynamic methods but takes 43 minutes per frame on one GPU, so it targets offline applications such as content creation and visual effects. Fewer iterations shorten the running time at some loss of accuracy, while remaining more accurate than most baselines (App.~\ref{sec:supp_iters}). Its correspondences directly benefit from strong flow estimators, while its geometry is robust to the flow quality (App.~\ref{sec:supp_raft}).

\paragraph{Acknowledgments.}
Zorah L\"ahner is supported by the German Research Foundation (DFG) Sachbeihilfe grant LA 5191/2-1. Daniele Baieri is supported by the Alexander von Humboldt Foundation through a postdoctoral fellowship. The authors gratefully acknowledge the access to the Marvin cluster of the University of Bonn.

%% file: supp_floats.tex
\makeatletter
\newif\ifsupp@wide
\let\supp@dblfloat\@dblfloat\def\@dblfloat{\global\supp@widetrue\supp@dblfloat}
\let\supp@xfloat\@xfloat
\newif\ifsuppcolfloat
\newif\ifsuppnofloatpage
\def\@xfloat#1[#2]{\ifsupp@wide\global\supp@widefalse\ifsuppnofloatpage\supp@xfloat{#1}[#2]\else\supp@xfloat{#1}[#2p]\fi\else\ifsuppcolfloat\supp@xfloat{#1}[!htb]\else\supp@xfloat{#1}[H]\fi\fi}
\makeatother
\renewcommand{\dbltopfraction}{0.95}\renewcommand{\dblfloatpagefraction}{0.5}\setcounter{dbltopnumber}{3}
\renewcommand{\topfraction}{0.9}\renewcommand{\textfraction}{0.07}

\renewcommand{\thesection}{\Alph{section}}
\renewcommand{\thefigure}{S\arabic{figure}}
\renewcommand{\thetable}{S\arabic{table}}
\renewcommand{\theequation}{S\arabic{equation}}

%% file: sec/appendix.tex
\providecommand{\FloatBarrier}{}\providecommand{\suppnofloatpagetrue}{}\providecommand{\suppnofloatpagefalse}{}\providecommand{\suppcolfloattrue}{}\providecommand{\suppcolfloatfalse}{}\makeatletter
\@ifundefined{supp@xfloat}{\renewenvironment{figure}[1][]{\par\medskip\noindent\begin{minipage}{\linewidth}\def\@captype{figure}}{\end{minipage}\par\medskip}\renewenvironment{table}[1][]{\par\medskip\noindent\begin{minipage}{\linewidth}\def\@captype{table}}{\end{minipage}\par\medskip}\gdef\suppgallery#1{\g@addto@macro\supp@galleries{\input{#1}}}}{\gdef\suppgallery#1{\input{#1}}}\gdef\supp@galleries{}
\gdef\suppflushgalleries{\supp@galleries}
\makeatother
\section{Notation}\label{sec:supp_notation}
\Cref{tab:notation} lists the symbols used in the main paper.
\input{figures/initalize}

\input{tables/notation}
\input{tables/hyperparameters}
\suppcolfloattrue\input{tables/runtime}\suppcolfloatfalse

\section{Datasets}\label{sec:supp_data}
We render the DeformingThings4D sequences from 50 cameras at $940\times1280$ pixels on a black background, with foreground masks.
The five animals (canie, dragon, fox, grizz, rabbit) have no texture; we colour each surface point by its 3D position in the first frame, and this colour stays fixed to the surface like a texture; all methods receive the same images. No method is given the mapping from colour to position, so the colour acts as an ordinary, smooth surface texture; correspondences are less accurate on the animals than on the textured humanoids for most methods, including ours (5.39 vs.\ 2.88, \Cref{tab:perscene_corr}).
The TaoGS captures have 79--81 cameras and include fast motion, loose clothing and interaction with objects.

\section{Implementation Details}\label{sec:supp_impl}
\Cref{tab:hyperparameters} lists the hyperparameters of our final configuration.
\input{figures/window_alg}

\subsection{Reconstruction and Motion Regularization Losses}\label{sec:supp_losses}
For each frame $\tau$ of the window $W = \{t-1, t, t+1\}$, we sample a mini-batch of $R$ rays and compare the rendered colors and masks (Eq.~\eqref{eq:render-color} of the paper) with the input images $I$ and foreground masks $M$:
\begin{equation}\label{eq:lrgb}
\begin{aligned}
\mathcal{L}_{\text{rgb}} &= \frac{1}{3}\sum_{\tau \in W}\frac{1}{R}\sum^{R}_{i=1}\lVert C_{i,\tau} - I_{i, \tau}\rVert_1, \\
\mathcal{L}_{\text{mask}} &= \frac{1}{3} \sum_{\tau \in W} \frac{1}{R}\sum^{R}_{i=1}\mathrm{BCE}\left(O_{i,\tau}, M_{i,\tau} \right),
\end{aligned}
\end{equation}
where $\mathrm{BCE}$ is the binary cross-entropy, and group them into the data term $\mathcal{L}_{\text{data}} = \lambda_{\text{rgb}}\mathcal{L}_{\text{rgb}} + \lambda_{\text{mask}}\mathcal{L}_{\text{mask}}$\phantomsection\label{eq:ldata}.
To keep the deformed SDF valid, we apply the eikonal constraint both after deformation and in canonical space, where $f_{\text{sdf}}$ is parameterized:
\begin{equation}\label{eq:lsdf}
\begin{aligned}
\mathcal{L}_{\text{sdf}} &= \frac{1}{3}\sum_{\tau \in W} \frac{1}{R}\sum_{x\in \mathbf{x}_{\tau}}\left( \left\lVert \nabla_{x} f_{\text{sdf}}\left(\delta_\tau(x)\right)  \right\rVert_2 - 1 \right)^2, \\
\mathcal{L}^{\star}_{\text{sdf}} &= \frac{1}{3}\sum_{\tau \in W} \frac{1}{R}\sum_{x\in \mathbf{x}_{\tau}}\left( \left\lVert \nabla_{x^\star} f_{\text{sdf}}\left(x^\star\right) \right\rVert_2 - 1 \right)^2, \quad x^\star = \delta_\tau(x).
\end{aligned}
\end{equation}
The first does not imply the second unless the deformation is locally rigid, and the canonical gradient must stay close to unit length because it is normalized to obtain the normals used in $\mathcal{L}_{\mathrm{norm}}$.

\paragraph{Motion regularization.}
We regularize the deformation and flow fields (\Cref{sec:motion-reg}) at the volume-rendered surface point $\bar{x}_{\tau,j}$ of each ray $j$.
For a field $g$ that maps a point to a new position (a deformation field $\delta^t_\tau$ or a flow field $\phi^t_{\tau\to\tau'}$), let $\mathbf{J}_g(x) \in \mathbb{R}^{3\times 3}$ be the Jacobian of the mapped position with respect to $x$, and let $\nabla_x \xi_g(x) \in \mathbb{R}^{6\times 3}$ be the spatial gradient of the field's $\mathfrak{se}(3)$ output.
We encourage locally rigid motion with an as-rigid-as-possible (ARAP) term~\cite{sorkine2007arap}, related to the elastic regularization of~\cite{park2021nerfies}, and spatially smooth fields with a smoothness term:
\begin{align}
\ell_{\mathrm{arap}}(g) &= \frac{1}{R}\sum_{j=1}^{R} \tfrac{1}{9}\big\lVert \mathbf{J}_g(\bar{x}_j)^{\top}\mathbf{J}_g(\bar{x}_j) - \mathbf{I} \big\rVert_F^2, \label{eq:larap}\\
\ell_{\mathrm{smooth}}(g) &= \frac{1}{R}\sum_{j=1}^{R} \tfrac{1}{18}\big\lVert \nabla_x \xi_g(\bar{x}_j) \big\rVert_1. \label{eq:lsmooth}
\end{align}
We average each term over the three deformation fields of the window, giving $\mathcal{L}^{\delta}_{\mathrm{arap}}$ and $\mathcal{L}^{\delta}_{\mathrm{smooth}}$, and over the four flow fields, giving $\mathcal{L}^{\phi}_{\mathrm{arap}}$ and $\mathcal{L}^{\phi}_{\mathrm{smooth}}$; the factors $\tfrac{1}{9}$ and $\tfrac{1}{18}$ average over the $3\times3$ and $6\times3$ entries.
The ARAP term keeps the local motion close to a rotation; since it acts on each point in isolation, the smoothness term penalizes abrupt changes of the motion between nearby points.

\subsection{Motion Factor of the Normal Consistency Loss}\label{sec:supp_motion_factor}
The motion factor that weights each ray in $\mathcal{L}_{\mathrm{norm}}$ (Eq.~\eqref{eq:lnorm} of the paper) is
\begin{equation}\label{eq:motion-factor}
s_{\tau\to\tau',j} = \max\!\left(\frac{r}{\lVert V^{*}_{\tau\to\tau',j} \rVert + r},\ s_{\min}\right),
\end{equation}
where $r$ is the 90th percentile of $\lVert V^{*}_{\tau\to\tau',j} \rVert$ over the batch and $s_{\min} = 0.05$.
It decreases with the optical-flow magnitude at the ray, and the floor $s_{\min}$ keeps fast-moving rays in the loss.

\section{Motion-Guided Importance Sampling}\label{sec:supp_sampling}
This section details the sampling of \Cref{sec:sampling}, illustrated in \Cref{fig:sampling_tech}.

\input{figures/sampling_col}

\paragraph{Sampling regions.}
For each view $i$, we define the core $\mathcal{S}_{i,t}$ as the union of the foreground masks of frames $t-1$ and $t$, which covers where the foreground is in the current frame as well as where it was in the previous one.
We include the previous position because frame $t$ is initialized from the solution of frame $t-1$ (\Cref{sec:window}), so the geometry reconstructed for frame $t-1$ is still present at those pixels, and rays must be sampled there to remove it.
Since the foreground masks are not pixel-accurate, the surface outline does not fall exactly on the core boundary, and sampling only inside the core would leave the part of the outline outside the boundary without rays.
We therefore dilate the core by 8 pixels, resulting in the dilated core $\mathcal{D}_{i,t}$, and define the outer region as $\Omega \setminus \mathcal{D}_{i,t}$, where $\Omega$ is the set of all pixel positions $\mathbf{u}$ of the image.
We cast $90\%$ of the rays of each batch through $\mathcal{D}_{i,t}$, weighted by the motion of each pixel as described below, and the remaining $10\%$ uniformly through the outer region.
The outer region keeps a share of the rays so that floaters appearing far from the foreground are still removed.

\paragraph{Motion score.}
For each pixel $\mathbf{u} \in \mathcal{D}_{i,t}$, we define its motion value as
\begin{equation}
\mu(\mathbf{u}) = \max\big(\|V^*_{t-1 \to t}(\mathbf{u})\|,\ \|V^*_{t \to t-1}(\mathbf{u})\|\big).
\label{eq:motion-value}
\end{equation}
The two flows measure the same motion, but on different pixels.
The backward flow $V^*_{t \to t-1}$ is defined on the pixels of frame $t$, so it captures the motion where the foreground is now.
The forward flow $V^*_{t-1 \to t}$ is defined on the pixels of frame $t-1$, so it captures the motion where the foreground was.
Using both therefore also assigns a motion value to the pixels that a moving part has just left, which are exactly the pixels where outdated geometry must be removed.
We take the maximum rather than the sum so that pixels covered by the foreground in both frames are not counted twice.
We then normalize the motion value to a score in $[0,1]$:
\begin{equation}
\begin{aligned}
\hat{\mu}(\mathbf{u}) &= \min\!\Big(\frac{\mu(\mathbf{u})}{\mu_{\mathrm{ref}}},\ 1\Big), \\
\mu_{\mathrm{ref}} &= \max\!\big(\operatorname{pct}_{95}\{\mu(\mathbf{u}') : \mathbf{u}' \in \mathcal{S}_{i,t}\},\ \mu_{\min}\big),
\end{aligned}
\label{eq:motion-score}
\end{equation}
where $\operatorname{pct}_{95}$ denotes the 95th percentile and $\mu_{\min}$ is a lower bound on the reference scale $\mu_{\mathrm{ref}}$.
Dividing by $\mu_{\mathrm{ref}}$ measures motion relative to the fastest-moving parts of the foreground in the current frame, so that the moving parts of a frame receive high scores whether the frame contains slow or fast motion.
We compute $\mu_{\mathrm{ref}}$ on the core rather than on the whole image, because most background pixels have zero flow and the remaining ones are dominated by flow outliers, and either would distort the scale.
The lower bound $\mu_{\min}$ handles nearly static frames: there, $\mu_{\mathrm{ref}}$ would be set by sub-pixel flow noise, and dividing by it would turn this noise into high scores.
We consider a pixel as moving if it lies in the core and its score exceeds a threshold $\kappa$:
\begin{equation}
\mathcal{A}_{i,t} = \{\mathbf{u} \in \mathcal{S}_{i,t} : \hat{\mu}(\mathbf{u}) > \kappa\}.
\label{eq:moving-set}
\end{equation}
Pixels of the dilation band cannot be moving, since optical flow is least reliable along the silhouette, and a few noisy boundary pixels would otherwise make a static frame appear to contain motion.

\paragraph{Adaptive sampling weights.}
Inside the dilated core, each pixel receives the sampling weight
\begin{equation}
\rho_\alpha(\mathbf{u}) = (1-\alpha) + \alpha\,\hat{\mu}(\mathbf{u}), \qquad \mathbf{u} \in \mathcal{D}_{i,t},
\label{eq:sampling-weight}
\end{equation}
where $\alpha \in [0,1]$ controls the emphasis on motion: $\alpha = 0$ samples $\mathcal{D}_{i,t}$ uniformly, and larger values favour pixels with a higher motion score.
A fixed $\alpha$ would make the share of rays on moving pixels depend on how much of the foreground moves: a waving hand would receive only a small share, while in a frame where the whole body moves the emphasis would have no effect.
We therefore choose $\alpha$ per image, such that the moving pixels receive a target share $\gamma$ of the rays drawn from $\mathcal{D}_{i,t}$.
This share is
\begin{equation}
\psi(\alpha) = \frac{\sum_{\mathbf{u} \in \mathcal{A}_{i,t}} \rho_\alpha(\mathbf{u})}{\sum_{\mathbf{u} \in \mathcal{D}_{i,t}} \rho_\alpha(\mathbf{u})},
\label{eq:moving-share}
\end{equation}
which increases with $\alpha$ as long as the moving pixels have a higher mean score than $\mathcal{D}_{i,t}$ as a whole, as is the case in practice.
We set $\alpha_{i,t}$ to the smallest $\alpha \in [0, \alpha_{\max}]$ with $\psi(\alpha) \geq \gamma$, found by a binary search over $\alpha$.
The three types of motion described above lead to three cases.
If only a small part of the foreground moves, the target may not be reachable, i.e., $\psi(\alpha_{\max}) < \gamma$, and we set $\alpha_{i,t} = \alpha_{\max}$.
If most of the foreground moves, the target is already met without emphasis, i.e., $\psi(0) \geq \gamma$, and we set $\alpha_{i,t} = 0$.
If fewer than a fraction $\eta$ of the core pixels are moving, we treat the frame as static and also set $\alpha_{i,t} = 0$.
The cap $\alpha_{\max} < 1$ keeps a minimum weight of $1-\alpha_{\max}$ on static foreground pixels, which still need to be fitted, and prevents the rays from collapsing onto a few pixels when the moving region is very small.

\paragraph{Sampling distribution.}
Combining the weights with the split between the dilated core and the outer region, the probability of sampling pixel $\mathbf{u}$ is
\begin{equation}
p_{i,t}(\mathbf{u}) =
\begin{cases}
0.9\, \dfrac{\rho_{\alpha_{i,t}}(\mathbf{u})}{\sum_{\mathbf{u}' \in \mathcal{D}_{i,t}} \rho_{\alpha_{i,t}}(\mathbf{u}')} & \text{if } \mathbf{u} \in \mathcal{D}_{i,t}, \\[8pt]
0.1\, \dfrac{1}{|\Omega \setminus \mathcal{D}_{i,t}|} & \text{otherwise},
\end{cases}
\label{eq:sampling-prob}
\end{equation}
and we draw the rays of each batch from $p_{i,t}$ without replacement.
With $\gamma = 0.5$, about $0.9\,\gamma = 45\%$ of all rays land on moving pixels whenever the target is reachable, compared to the fraction of the image they cover under uniform sampling (about $4\%$).
We apply this sampling independently to each frame of the window (\Cref{sec:window}), each with its own preceding frame.
The values of $\mu_{\min}$, $\kappa$, $\eta$, $\gamma$, and $\alpha_{\max}$ are listed in \Cref{tab:hyperparameters}.

\section{Lite Variant}\label{sec:supp_geo}
This section details \emph{Ours (Lite)} (\Cref{sec:geo-variant}), which removes the flow fields and obtains cross-frame pairs from the precomputed optical flow instead.

\paragraph{Ray pairing.}
The rays of the current frame are sampled as in \Cref{sec:sampling}.
A ray through pixel $\mathbf{u}_j$ is paired with the pixel
\begin{equation}
\mathbf{u}'_{\tau',j} = \operatorname{round}\big(\mathbf{u}_j + V^*_{\mathrm{curr}\to\tau'}(\mathbf{u}_j)\big), \qquad \tau' \in \{\mathrm{prev}, \mathrm{next}\},
\end{equation}
of the neighboring frame, and the pair receives the forward--backward confidence
\begin{equation}
c_{\tau',j} = \exp\!\Big(-\tfrac{1}{\sigma_{\mathrm{fb}}}\big\lVert \mathbf{u}'_{\tau',j} + V^*_{\tau'\to\mathrm{curr}}(\mathbf{u}'_{\tau',j}) - \mathbf{u}_j \big\rVert_2\Big),
\end{equation}
which is $1$ when following the optical flow to the neighboring frame and back returns to the starting pixel, and decays with the round-trip error, which is large at occlusions and flow errors.
The rays of the previous and the next frame are exactly these paired pixels, so no additional rays are rendered.

\paragraph{Losses.}
For each $\tau' \in \{\mathrm{prev}, \mathrm{next}\}$, let $j'$ denote the ray paired with ray $j$ of the current frame, and let $\omega_{\tau',j} = O_{\mathrm{curr},j}\, c_{\tau',j}$.
The normal consistency compares the composited canonical normals of the two rays, each obtained through the deformation field of its own frame (\Cref{eq:composited-normals}):
\begin{equation}
\mathcal{L}^{\mathrm{geo}}_{\mathrm{norm}} = \frac{1}{2}\sum_{\tau'} \frac{\sum_{j} \omega_{\tau',j} \big(1 - \bar{n}^{\star\top}_{\mathrm{curr},j}\, \bar{n}^{\star}_{\tau',j'}\big)}{\sum_{j} \omega_{\tau',j}}.
\end{equation}
This normal consistency is the only cross-frame loss of the Lite variant.
As in the full model, gradients only flow into the less converged frame: the $\mathrm{prev}$ side is constant in the $(\mathrm{curr}, \mathrm{prev})$ pair, and the $\mathrm{curr}$ side is constant in the $(\mathrm{curr}, \mathrm{next})$ pair.
The flow loss $\mathcal{L}_{\mathrm{flow}}$ and the flow-field regularizers $\mathcal{L}^{\phi}_{\star}$ are dropped; all other losses and hyperparameters are those of the full model (\Cref{tab:hyperparameters}), except $\lambda_{\mathrm{norm}} = 0.1$ and $\sigma_{\mathrm{fb}} = 3$ pixels.

\section{Temporal Consistency Error}\label{sec:supp_tce}
The ground-truth meshes of DeformingThings4D share their vertices over each sequence.
Let $\mathbf{g}_j^t$ be the position of ground-truth vertex $j$ in frame $t$, $\mathbf{n}_j^t$ its normal, and $w_j^t$ its area, one third of the summed area of its incident faces.
We sample $2\cdot10^6$ points uniformly on the reconstructed mesh $\mathcal{M}_t$ and match $\mathbf{g}_j^t$ to the nearest sample $\mathbf{p}_j^t$ whose face normal $\mathbf{m}_j^t$ differs from $\mathbf{n}_j^t$ by at most $60^\circ$, so that, for example, a point on the hand is not matched to the nearby torso.
The offset of the vertex to the reconstruction is the vector from it to the plane of the matched face, along that face's normal:
\begin{equation}
\mathbf{e}_j^t = \big((\mathbf{p}_j^t - \mathbf{g}_j^t)^\top \mathbf{m}_j^t\big)\,\mathbf{m}_j^t .
\end{equation}
The temporal consistency error between frames $t$ and $t+1$ is the area-weighted mean change of this offset,
\begin{equation}
\mathrm{TCE}_t = \frac{\sum_{j \in \mathcal{J}_t} w_j^{t+1}\, \big\| \mathbf{e}_j^{t+1} - \mathbf{e}_j^t \big\|}{\sum_{j \in \mathcal{J}_t} w_j^{t+1}},
\end{equation}
where $\mathcal{J}_t$ contains the vertices matched in both frames.
We average $\mathrm{TCE}_t$ over all pairs of adjacent frames, then over the scenes.
A surface that is wrong by the same offset in both frames has zero error, and one that is correct in one frame and off by 1\,cm in the next has an error of 1\,cm: the metric measures how much the reconstruction flickers relative to the true surface.
Because the matching uses the ground-truth correspondences, motion along the surface does not change the error, so the TCE does not measure correspondence accuracy, which we evaluate separately (\Cref{sec:supp_corr}).

\paragraph{Normal metrics.}\label{sec:supp_normal_metrics}
The normal angle error (NAE) is the mean angle between the normals of nearest points on the reconstruction and on the ground truth, in both directions (100k samples per mesh).

\section{Detailed Comparison}\label{sec:supp_detailed}
This section gives the per-scene results behind the averages of the main paper: geometry on ActorsHQ and DeformingThings4D and the ablation; the per-scene correspondence results are in \cref{sec:supp_corr}.
All baselines receive the same input as our method, the same cameras, images, foreground masks and frames, and run with the same settings on every scene. MoAngelo~\cite{ebbed2026moangelo}, which optimizes one frame at a time, is run with three times as many iterations per frame as our method, to match the number of optimization iterations per frame of our three-frame windows.
All metrics are computed on the raw extracted meshes of every method, including ours, without removing floaters or any other post-processing.
Some of them fail on individual scenes, mostly on DeformingThings4D, whose characters move fast and have thin parts; we report their results as obtained and checked the cause of every failure in the logs and renders. GauSTAR does not update its tracked mesh on dragon, which stays at its initial shape (Chamfer distance 1244\,mm). The mesh of Tensor4D on dragon is degenerate and fills its bounding box. PGSR renders a translucent shell around the character on canie, fox and rabbit, where its geometry also fails. Deformable 3DGS loses all its Gaussians on abe in both training attempts, so MaGS, which starts from it, has no input there, and the training of DG-Mesh fails on abe on an empty mesh. Dynamic-2DGS collapses during training or fails in meshing on 4 ActorsHQ scenes; on DeformingThings4D its meshing fails on 5 scenes and its training times out on peasantgirl. The characters of DeformingThings4D keep the scale of the dataset, in which a humanoid is about 1.9 units tall; we treat one unit as one metre and report distances in mm.
Where a method gave no result on a scene (--), its average is taken over the remaining scenes and marked with $^\dagger$ in \Cref{tab:geometry_avg}.
\input{tables/perscene_cd}
\input{tables/perscene_nc}
\input{tables/perscene_tce}
\input{figures/geo_gallery}
\input{figures/geo_cmp_gallery}
\input{tables/ablation_detailed}
\Cref{tab:seeds} repeats our full model with three random seeds on one scene per dataset, and the variant without $\mathcal{L}_{\mathrm{norm}}$ on vampire: the Chamfer distance varies by 0.05\,mm or less, below the effects of $\mathcal{L}_{\mathrm{flow}}$ and the motion-guided sampling in \cref{tab:ablation}, and the normal angle error of the two variants on vampire differs by about 2\textdegree{}, more than four times its variation over the seeds.
\input{tables/seeds}
\input{figures/ablation_closeups}

\section{Novel-View Synthesis}\label{sec:supp_nvs}

\input{tables/nvs}
\input{tables/perscene_psnr}
\input{tables/perscene_ssim}
\input{tables/perscene_lpips}
\begingroup\addtocounter{table}{-1}\renewcommand{\thetable}{S\arabic{table}a}\input{tables/nvs_taogs}\endgroup \input{figures/nvs_cmp_gallery}
\input{figures/nvs_cmp_taogs_num}

We hold out 6 cameras per sequence on ActorsHQ and 5 on DeformingThings4D, and report PSNR, SSIM and LPIPS on them over the 50 frames, with the background of the photos set to black using the foreground masks (\Cref{tab:nvs}).
On ActorsHQ, ours renders the held-out views most accurately (33.71\,dB PSNR, against 32.31\,dB for AT-GS and 32.30\,dB for 2DGS); Ours (Lite) is 0.3\,dB lower on ActorsHQ and 0.6\,dB lower on DeformingThings4D.
\Cref{fig:taogs} shows further reconstructions on TaoGS.

\needspace{14\baselineskip}On TaoGS (\Cref{tab:nvs_taogs}, \Cref{fig:nvs_cmp_taogs}), AT-GS renders the held-out views with the highest PSNR among the dynamic methods (33.69 vs.\ 33.36\,dB for ours); ours is second, the most accurate on magic3, umbrella and changeclothes, and has the lowest LPIPS among the dynamic methods (0.045 vs.\ 0.056 for AT-GS). The per-frame static 2DGS and PGSR, fitted to every frame independently, reach 33.90 and 33.61\,dB. Our method targets the accuracy of the reconstructed surfaces and of their correspondences rather than view synthesis; the geometry is compared in \cref{sec:supp_detailed}.

\section{Correspondence Evaluation}\label{sec:supp_corr}
We evaluate on DeformingThings4D, whose ground-truth meshes share one topology over each sequence, so that every surface point has a known position in every frame.
We sample 5000 points uniformly on the ground-truth surface of frame~1.
For every frame~$k$, the error of a point is the geodesic distance on the ground-truth surface of frame~$k$ between the position predicted by a method and the true position, on the shape scaled to unit area~\cite{kim2011blended}.
We report the mean error over the points and frames, and the area under the curve (AUC) of the fraction of points whose error is below a threshold, for thresholds up to 0.1~\cite{cao2023ulrssm}.
We consider two settings.
On \emph{ground-truth surfaces}, our flow fields move the sampled points directly, and the shape-matching methods are run on the ground-truth frames remeshed independently to about 5k vertices, so that the frames do not share vertices~\cite{ren2018continuous}.
On \emph{reconstructed surfaces}, each point is first moved to the closest point of the method's reconstruction of frame~1, and its correspondence is read off the reconstruction of frame~$k$; points farther than 2\,cm from the reconstruction of frame~1 count as not covered, but are scored like all other points, so all reported errors include them.
We compare with functional maps~\cite{ovsjanikov2012functional} refined with ZoomOut~\cite{melzi2019zoomout}, with ULRSSM~\cite{cao2023ulrssm}, an unsupervised learned spectral matcher, and with Hybrid FM~\cite{bastianxie2024hybrid}, which adds an elastic basis to the spectral basis of ULRSSM; both use models pretrained on DeformingThings4D, with the same input and setup.
Each maps frame~1 to frame~$k$ either directly or by chaining the maps between adjacent frames.
We evaluate all 10 scenes over the 50 frames.
MoAngelo~\cite{ebbed2026moangelo} and \emph{Ours (Lite)}, which have no flow fields, obtain correspondences through their canonical frame, and are evaluated in the same way on their own reconstructions.
The canonical route of our full model (\Cref{tab:correspondence}) matches each vertex of frame~$k$ to the query point whose canonical position is nearest under the deformation of frame~$k$.
DG-Mesh~\cite{Liu2025} builds an independent mesh for every frame and has no correspondences of its own, and its training failed on abe, so its row in \Cref{tab:correspondence} is a mean over the other 9 scenes, on which our reconstruction has an error of 4.40.
We obtain correspondences from what it does keep, in two ways, and move each point to its closest point on the reconstruction of frame~1 as above.
In \emph{Gaussian tracking}, the point follows the distance-weighted average displacement, from frame~1 to frame~$k$, of its 4 nearest canonical Gaussians.
In \emph{backward map}, the point is sent to the nearest frame-1 vertex, and its match is the vertex of frame~$k$ closest to it in canonical space under the backward deformation network.
Both snap the final position to the nearest vertex of the frame-$k$ mesh, and neither is tuned. \Cref{tab:correspondence} reports Gaussian tracking; the backward map gives a similar error (15.07 vs.\ 15.20).
\input{tables/correspondence_twostage}
To compare with a two-stage pipeline that first reconstructs every frame and then matches the reconstructions (\Cref{tab:corr_twostage}), we run Hybrid FM, with the same code and pretrained model as on the ground-truth surfaces, on the per-frame meshes of PGSR~\cite{chen2024pgsr} (frame~1 is reconstructed with the same settings as the other frames).
Isotropic remeshing does not converge on these meshes, which contain holes and disconnected fragments, so each frame is reduced to about 5k vertices (4655--5550) by quadric decimation instead, and the triangles are less regular than on the remeshed ground-truth frames.
Points are moved onto the PGSR mesh of frame~1 and scored as above, including the points that are not covered.
PGSR fails to reconstruct canie, dragon and fox (Chamfer distance 12--46\,mm), where only 15\%, 2\% and 35\% of the points are covered; on these scenes the error reflects the failed reconstruction rather than the matcher.
Over all 10 scenes, the mean error is 30.38 for direct and 33.46 for chained matching, against 4.14 for ours; it stays above 7.8 on every scene, including those that PGSR reconstructs well (e.g.\ 27.96 on ninja and 33.50 on peasantgirl, with 93\% and 90\% coverage).
On the 7 scenes that PGSR reconstructs, the mean error is 28.02 for direct and 31.17 for chained matching, against 3.75 for ours.
On poses far from frame~1, many points are matched to the wrong body part, often the mirrored one, and chaining keeps such a mistake in all later frames.
To separate the effect of the decimation from that of the PGSR geometry, we decimate the ground-truth frames in the same way (after subdividing each frame and adding a 1\% jitter, so that the frames do not share their triangulation) and run the same matcher: its mean error rises only from 9.25 to 9.35 for direct and from 3.88 to 5.02 for chained matching (AUC 0.64 and 0.68), against 30.38 and 33.46 on the PGSR meshes, so the decimation explains less than 5\% of the gap, which comes from the reconstructed geometry.
The ninja character is part of the DeformingThings4D data on which the Hybrid FM model was trained.
We evaluate the other dynamic reconstruction methods of \Cref{sec:experiments} in the same way, each with the correspondence its representation provides (\Cref{tab:perscene_corr}): GauSTAR tracks its mesh; for the Gaussian-based methods, a point follows the displacement of its nearest Gaussians; Tensor4D is matched in its canonical space.
Over the 10 scenes, our mean error (4.14) is the lowest; the next are Dynamic 3DGS (7.26), SC-GS (7.56) and MoAngelo (7.58).
Per scene, Dynamic 3DGS is more accurate than ours on abe, canie, fox and grizz, and SC-GS on AJ, fox, grizz, ninja and vampire; on each of these scenes our surface is more accurate than theirs (\Cref{tab:perscene_cd,tab:perscene_nc,tab:perscene_tce}), by up to a factor of 34 (canie: Chamfer distance 3.06 vs.\ 105.07\,mm for Dynamic 3DGS). Their Gaussians follow the motion, but do not form an accurate surface, while our correspondences lie on an accurate surface in every scene.
\IfFileExists{imgs/plots/drift_baselines.pdf}{\input{figures/drift_baselines}}{}
\Cref{fig:drift_baselines} shows how the errors evolve over the sequence.
Our mean error is the lowest of all methods in every frame; it grows mostly over the first 20 frames (from 0.78 in frame~2 to about 4.3 in frame~20) and then slowly, to 5.38 in frame~49, while those of Dynamic 3DGS, SC-GS and MoAngelo grow to about 8--10.
Through the canonical space, our model reaches the same error as with the flow fields only in the last frame (5.40 vs.\ 5.38).
The median over the scenes is lower for SC-GS than for ours from frame~29 on, because SC-GS is very accurate on some scenes and fails on others (\Cref{tab:perscene_corr}).
\input{figures/long200}
On three sequences of 200 frames, reconstructed with 0.2 of the iterations (\Cref{fig:long200}), the Chamfer distance does not grow on Maria and leopard (Chamfer distance 8.03 and 3.62\,mm in the first 50 frames, 7.23 and 3.58\,mm in the last 50); on peasant girl it rises from 10.32 to 12.55\,mm after frame~150, when the girl stands up and her wide skirt, an open single layer in the ground-truth mesh, unfolds; our reconstruction closes the skirt below its hem. The correspondence error keeps growing after frame~50, as expected for correspondences chained over the sequence, from 2.34, 4.70 and 1.25 in the first 50 frames to 10.19, 8.23 and 6.16 in the last 50 frames (peasant girl, Maria, leopard).
\input{figures/cd_vs_corr_detail}
\input{figures/corr_gallery}
\input{tables/perscene_corr}
\input{tables/median_wins}
\input{figures/pck_recon}\needspace{14\baselineskip}\paragraph{Scored map and its quality.}
\input{tables/map_quality}
Our correspondences are scored as described in \cref{sec:correspondence}: each point is moved through the composed forward flow fields and snapped to the nearest vertex of our frame-$k$ mesh only at frame~$k$, not after every step. \Cref{tab:map_quality} reports how close the resulting maps are to one-to-one at frame~49. On average, 97.0\% of the 5000 points land on distinct vertices, against 90.9\% through the canonical space (71.8\% on dragon). In both routes the tracked points end up slightly closer together than their true positions (spacing 0.84 and 0.82) and cover 84.4\% and 83.1\% of the surface that the true positions cover; the flow route covers less than the canonical route on abe, canie, fox, grizz and rabbit, and more on the other five scenes.

\paragraph{Shape matching on the ground-truth surfaces.}
\input{tables/correspondence_gt}
\input{tables/correspondence_detailed}
\input{figures/sm_gt}\input{figures/pck_sm}On the ground-truth surfaces, our flow fields move the points directly, while the shape-matching methods run on the ground-truth frames independently remeshed so that they share no vertices~\cite{ren2018continuous}.
These methods match given surfaces without the images, after the reconstruction. FlowSurf4D is not a shape-matching method: we include them only as a reference for the accuracy of our correspondences, not as competing methods.
Our flow fields are optimized only through our own reconstruction; the ground-truth surfaces are never used. Applied to points on the ground-truth surfaces (\Cref{tab:correspondence_gt}), they are almost as accurate as on our reconstruction (3.99 vs.\ 4.14), so they describe the motion of the scene, not only that of the surface they were optimized with.
Chained Hybrid FM~\cite{bastianxie2024hybrid} has a mean error of 3.88 (AUC 0.761 vs.\ 0.704) and chained ULRSSM 4.73.

\needspace{12\baselineskip}Ours are the most accurate of all methods in the early frames (1.73 vs.\ 2.31 for chained Hybrid FM), while in the late frames small errors of the individual steps accumulate (4.67 vs.\ 4.35).
Per scene (\Cref{tab:correspondence_detailed}), chained Hybrid FM and chained ULRSSM are more accurate than our flow fields on the ground-truth surfaces on 8 of the 10 scenes, and their median errors over the scenes are lower (2.61 and 2.85 vs.\ 3.40); our mean error is lower than that of ULRSSM because of the two scenes with the largest deformations, dragon and peasantgirl.

\paragraph{Shape matching on our meshes.}
We also apply Hybrid FM and ULRSSM to our reconstructed meshes (\Cref{tab:corr_twostage}), decimated to about 5k vertices per frame, mapping frame~1 to frame~$k$ directly or by chaining adjacent frames.
We score all methods on the query points that both our flow fields and the matcher map (80\% of the queries on average; 5.5\% on dragon, which is about 7\,m tall, so that few query points lie within 2\,cm of a vertex of its 5k-vertex input mesh).
On this common subset, our mean error is 3.93 (Early 1.59, Late 4.63, AUC 0.705); Hybrid FM has 24.87 directly and 27.32 chained (AUC 0.320 and 0.251), and ULRSSM 26.68 directly and 30.86 chained (AUC 0.295 and 0.211).
The solver of Hybrid FM fails on zero-area triangles, so the two such triangles of our remeshed AJ frame~5 and peasantgirl frame~48 are removed (no other frame of the 10 scenes has any). Ours is more accurate on every scene; the closest is grizz (2.41 vs.\ 3.83 for chained Hybrid FM).
On reconstructed surfaces, whose meshes differ from frame to frame, the matchers are much less accurate than on the ground-truth surfaces.
\input{figures/sm_twostage}

\section{Flow Fields vs.\ Lifted Optical Flow}\label{sec:supp_liftflow}
\input{tables/liftflow}
Our flow fields are supervised by the FlowSeek optical flow, so one may ask whether that flow alone already gives the correspondences.
\Cref{tab:liftflow} answers this on our own reconstructed surfaces: instead of the flow fields, each tracked point is projected into every camera in which it is visible, moved by that camera's optical flow, lifted to the front-most point of the next mesh at the target pixel, and the per-camera positions are combined by their median and snapped to the next mesh; this is chained over the sequence like our correspondences.
On ground-truth meshes with ground-truth optical flow, this procedure reaches a geodesic error of about 2, the floor set by snapping to the coarse ground-truth vertices.
With the same optical flow and the same surfaces, the lifted flow is less accurate than our flow fields on every scene, and its error grows faster over the sequence: optical flow is a 2D estimate per camera, while the flow fields are 3D maps optimized jointly with the surfaces of adjacent frames.

\section{Correspondences on Real Captures}\label{sec:supp_realdata}
\input{figures/realdata_tracks}
\input{tables/realdata_tracks}
\suppcolfloattrue\input{tables/realdata_tracks_common}\suppcolfloatfalse \suppnofloatpagetrue\input{tables/realdata_tracks_scenes}\suppnofloatpagefalse Real captures have no ground-truth correspondences (and TaoGS~\cite{taogs} no ground-truth geometry), so we compare our correspondences with 2D point tracks in cameras that are not used for training (neither their images nor their optical flow reach any loss), on the 6 TaoGS scenes and the 6 ActorsHQ scenes of \Cref{tab:geometry_avg}.
On TaoGS, we hold out 5 cameras per scene and track, with CoTracker3~\cite{karaev2024cotracker3} (offline model, default settings), a regular grid of about 1500 pixels of the frame-1 foreground mask, eroded by 2 pixels, in each held-out video (44.6k query points in total); on ActorsHQ, we use the 6 test cameras of the novel-view evaluation of each scene, with the same settings (54.0k query points).
Each query pixel is lifted onto the method's frame-1 surface by back-projecting the pixel center with the rendered depth, moved to frame~$k$ by the method's correspondences, and projected into the same camera.
A point is visible if its depth is within 1\,cm of the depth of the method's frame-$k$ surface at its pixel, and a pair is evaluated if the point is visible and CoTracker3 marks it as visible.
A point that is occluded in a frame is skipped there and evaluated again when it reappears, since our correspondences are defined on the surface, not in the image.
On our reconstruction, we compare our correspondences (\Cref{sec:correspondence}) with three alternatives: the canonical route, which matches each frame-$k$ vertex whose canonical position is nearest to that of the query point (as in \Cref{sec:exp_correspondence}); the nearest vertex of the frame-$k$ mesh to the frame-1 point, without any correspondence; and zero motion, which keeps the query pixel.

\needspace{12\baselineskip}We evaluate the other reconstruction methods with the correspondence their representation provides, as on DeformingThings4D (\Cref{sec:supp_corr}): for the Gaussian-based methods and DG-Mesh, a point follows the displacement of its 4 nearest Gaussians (Dynamic 3DGS: the rigid motion of its nearest Gaussian), and their surface is the rendered depth (DG-Mesh: its meshes); MoAngelo is matched in its canonical space; Tensor4D is matched through its flow model in its canonical space, GauSTAR follows its tracked mesh faces, and MaGS keeps the same face and barycentric coordinates on its meshes, which share one topology, as on DeformingThings4D.
AT-GS stores no correspondence; we rerun it with identical settings and record, for every Gaussian, the Gaussian it was cloned or split from, and move a point with the frame-$k$ descendants of its 4 nearest frame-1 Gaussians. As AT-GS prunes and densifies at every frame, only 1--3\% of the frame-1 Gaussians still have descendants at frame 49 on TaoGS, and only 18--44\% of the query points are still visible at frame 49 on ActorsHQ.
On ActorsHQ, SC-GS, Deformable 3DGS, DG-Mesh and Dynamic-2DGS are trained on images shifted by whole pixels so that the principal point is at the image center; we shift their depth and opacity back before lifting (residual offset below 0.5\,px). The rendered depth of SC-GS misses parts of the person in some cameras of 3 ActorsHQ scenes (mask IoU 0.42--0.85, against about 0.95 for the other methods), which lowers its fraction of evaluated pairs.
Dynamic-2DGS fails on magic2 (empty mesh) and does not finish on magic3 and changeclothes, and converges only on actor7 and actor8 of ActorsHQ.
\Cref{tab:realdata_tracks} reports the median pixel error, the mean PCK at 4, 8, 16 and 32 pixels and the fraction of evaluated pairs, averaged over the cameras of each scene and then over the scenes, each method on its own visible pairs; \Cref{tab:realdata_tracks_common} compares each method with ours on the pairs visible to both, and \Cref{tab:realdata_tracks_scenes} gives the per-scene results.
On both datasets, our correspondences have the lowest median error and the highest mean PCK in all three frame ranges.
On TaoGS, they have the highest mean PCK on 5 of the 6 scenes; on bag, the scene with the least motion, Deformable 3DGS, Dynamic 3DGS and 4DGS have a higher mean PCK (92.8, 91.8 and 91.6 vs.\ 91.2). On ActorsHQ, they have the highest mean PCK on all 6 scenes, but on actor7, the scene with the least motion, Deformable 3DGS and 4DGS are within 0.05 of ours.
On the pairs visible to both, our median error is lower than that of every method in every frame range, on both datasets.
Our correspondences have a lower median error than the canonical route on all 30 TaoGS cameras and on 28 of the 36 ActorsHQ cameras; the canonical route is more accurate on all cameras of actor7 and on 2 of actor5.
In frame~2, the median error is 1.18\,px on TaoGS and 1.14\,px on ActorsHQ.
The ActorsHQ images are smaller ($747\times1021$ pixels), so the same pixel thresholds are stricter there.
CoTracker3 is itself an estimate: it runs at $384\times512$ pixels on a crop around the person, so differences of a few pixels at full resolution may reflect its own localization error, and its visibility flag excludes about 38\% of the pairs. Each method is trained once per scene, so small differences, as on magic2, bag and actor7, should be read with caution.

\section{Sensitivity to the Optical Flow}\label{sec:supp_raft}
\input{figures/iters_scaling}\input{figures/environments}\input{tables/raft}
\Cref{tab:raft} trains our full model with optical flow from RAFT~\cite{teed2020raft} instead of FlowSeek~\cite{Poggi_2025_ICCV}, with all other settings unchanged.
The RAFT flow has a several times higher end-point error, yet the geometry barely changes (Chamfer distance and temporal consistency error within 0.3\,mm), whereas the correspondences, which the flow supervises, become less accurate.

\section{Accuracy--Runtime Trade-off}\label{sec:supp_iters}
\suppnofloatpagetrue\input{figures/applications}\suppnofloatpagefalse The running time of our method is set by the iteration counts of its schedule, so it can be traded for accuracy without changing the method.
\Cref{tab:iters} scales all of them by $\nicefrac{1}{3}$ (iterations per window, first window, coarse-to-fine step and consistency warm-up), which reduces the time per frame from 43 to 19 minutes.
The surfaces lose some accuracy but remain more accurate than those of every dynamic baseline on the three scenes tested, except AT-GS on peasantgirl and vampire (\cref{sec:supp_detailed}), including MoAngelo~\cite{ebbed2026moangelo} with three times as many iterations (3.70 vs.\ 3.90, 11.51 vs.\ 14.27 and 12.78 vs.\ 12.82), and the correspondences are almost unchanged (2.60 vs.\ 2.50 and 4.08 vs.\ 4.06).
\Cref{fig:iters_scaling} extends this to scales of 0.2 and 0.5 on two DeformingThings4D scenes: the surface errors fall steadily with more iterations, while the correspondences change little.
\input{tables/iters}The full schedule is needed for the accuracy reported in the main paper: with a third of the iterations, the surfaces lose 8--13\% in Chamfer distance and 20--35\% in temporal consistency, and on DeformingThings4D, AT-GS~\cite{chen2025atgs} would be more accurate on peasantgirl and vampire. The training times of all methods are given in \cref{tab:runtime}.

\section{Applications}\label{sec:supp_applications}
Our correspondences link every point of the first mesh to a point of each later mesh, so an edit made on frame~1 follows the surface through the sequence.
\Cref{fig:texture_transfer} dresses the person on frame~1 only and carries the dressing to the later frames, and \cref{fig:applications} edits the blouse on frame~1.
The same links give one albedo per surface point over all cameras and frames, which we relight (\cref{fig:relighting,fig:applications}).
Since the reconstructions are meshes, they can be placed in virtual environments and rendered with standard graphics tools (\cref{fig:environments}); the animated sequences are in the video on the project page.
\input{figures/texture_transfer}
\input{figures/relighting}
\suppflushgalleries

%% file: figures/initalize.tex
\begin{figure}[t]
\centering
\definecolor{tzIn}{RGB}{74,82,99}\definecolor{tzOurs}{RGB}{14,128,229}\definecolor{tzTeal}{RGB}{32,178,160}\definecolor{tzInk}{RGB}{24,28,40}\colorlet{Renderpath}{tzOurs}\colorlet{Flowpath}{tzTeal}\colorlet{Past}{tzIn}

\resizebox{\columnwidth}{!}{\begin{tikzpicture}[node distance=12mm and 1mm]

\node[scale=0.95, draw=Renderpath, text=Past, rounded corners, anchor=south, thick, dashed] (delta-t-1)  {$\delta^{t-1}_{\mathrm{prev}}$};
\node[scale=0.9, draw=Flowpath, text=Past, rounded corners, anchor=center, align=center, fill=white, thick, right=of delta-t-1, dashed] (phi-prev){$\phi^{t-1}_{\mathrm{prev}\to\mathrm{curr}}$\strut\\$\phi^{t-1}_{\mathrm{curr}\to\mathrm{prev}}$\strut};
\node[scale=0.95, draw=Renderpath, text=Past, rounded corners, anchor=south, thick, right=of phi-prev, dashed] (delta-t) {$\delta^{t-1}_{\mathrm{curr}}$};
\node[scale=0.9, draw=Flowpath, text=Past, rounded corners, anchor=center, align=center, fill=white, thick, right=of delta-t, dashed] (phi-next) {$\phi^{t-1}_{\mathrm{curr}\to\mathrm{next}}$\strut\\$\phi^{t-1}_{\mathrm{next}\to\mathrm{curr}}$\strut};
\node[scale=0.95, draw=Renderpath, text=Past, rounded corners, anchor=south, thick, right=of phi-next, dashed] (delta-t+1)  {$\delta^{t-1}_{\mathrm{next}}$};

\draw[color=Past, dashed, rounded corners] ([xshift=-4pt, yshift=-3pt]delta-t-1.west |- phi-prev.south) rectangle ([xshift=4pt,yshift=3pt]delta-t+1.east |- phi-next.north);

\node[fill=white, inner sep=0, outer sep=0] at ([xshift=-2pt, yshift=-1pt]delta-t-1.west |- phi-prev.south) {\includegraphics[width=0.05\columnwidth, trim=5 5 5 5, clip]{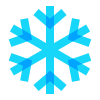}};

\node[scale=0.95, draw=Renderpath, fill=Renderpath!30!white, rounded corners, anchor=south, thick, below=of delta-t] (delta-t-pn)  {$\delta^{t}_{\mathrm{prev}}$};
\node[scale=0.9, draw=Flowpath, fill=Flowpath!30!white, rounded corners, anchor=center, align=center, thick, right=of delta-t-pn] (phi-prev-n){$\phi^{t}_{\mathrm{prev}\to\mathrm{curr}}$\strut\\$\phi^{t}_{\mathrm{curr}\to\mathrm{prev}}$\strut};
\node[scale=0.95, draw=Renderpath, fill=Renderpath!30!white, rounded corners, anchor=south, thick, right=of phi-prev-n] (delta-t-cn) {$\delta^{t}_{\mathrm{curr}}$};
\node[scale=0.9, draw=Flowpath, fill=Flowpath!30!white, rounded corners, anchor=center, align=center, thick, right=of delta-t-cn] (phi-next-n) {$\phi^{t}_{\mathrm{curr}\to\mathrm{next}}$\strut\\$\phi^{t}_{\mathrm{next}\to\mathrm{curr}}$\strut};
\node[scale=0.95, draw=Renderpath, fill=Renderpath!30!white, rounded corners, anchor=south, thick, right=of phi-next-n] (delta-t-nn)  {$\delta^{t}_{\mathrm{next}}$};

\draw[rounded corners, thick] ([xshift=3pt, yshift=3pt]delta-t-pn.west |- phi-prev-n.north) -- ++(-6pt, 0pt) -- ([xshift=-3pt, yshift=-3pt]delta-t-pn.west |- phi-prev-n.south) -- ++(6pt, 0pt);

\draw[rounded corners, thick] ([xshift=-3pt, yshift=3pt]delta-t-nn.east |- phi-prev-n.north) -- ++(6pt, 0pt) -- ([xshift=3pt, yshift=-3pt]delta-t-nn.east |- phi-prev-n.south) -- ++(-6pt, 0pt);

\node[left=3mm of delta-t-pn, align=center] {\textit{shift} \textit{frame}\\\textit{window:}};

\node (rlineref) at ($(delta-t+1)!0.4!(delta-t-cn)$) {};
\node (flineref) at ($(phi-next)!0.55!(phi-prev-n)$) {};

\draw[color=Renderpath, ->] (delta-t) -- node[midway,scale=0.9,fill=white,inner sep=0] {init\strut}  (delta-t-pn);
\draw[color=Renderpath, ->] (delta-t+1) -- (delta-t-cn);
\draw[color=Renderpath, ->, rounded corners] (rlineref.center) -- node[midway,scale=0.9,fill=white,inner sep=0] {init\strut}  (delta-t-nn |- rlineref.center) -- (delta-t-nn);

\draw[color=Flowpath, ->] (phi-next) -- (phi-prev-n);
\draw[color=Flowpath, ->, rounded corners] ([xshift=0.5pt]flineref.center) --  node[midway,scale=0.9,fill=white,inner sep=0] {init\strut} (phi-next-n |- flineref.center) -- (phi-next-n);

\node[above=5mm of delta-t-1, fill=white, inner sep=0pt] (t-2-label) {$t-2$}; 
\node[above=5mm of delta-t, fill=white, inner sep=0pt] (t-1-label) {$t-1$}; 
\node[above=5mm of delta-t+1, fill=white, inner sep=0pt] (t-label) {$t$}; 
\node[fill=white, inner sep=0pt] at (delta-t-nn |- t-label) (t+1-label) {$t+1$}; 

\draw[->>] ([xshift=-10pt, yshift=-6pt]t-2-label.west) -- ([xshift=10pt, yshift=-6pt]t+1-label.east);

\end{tikzpicture}}

\caption{Initialization of the next window after shifting (\Cref{sec:window}).}
\label{fig:init}
\end{figure}

%% file: tables/notation.tex
\newcommand{\notgroup}[1]{\multicolumn{2}{@{}l}{\textit{#1}} \\[1pt]}
\begin{table*}[t]
\centering
\footnotesize
\renewcommand{\arraystretch}{1.12}
\begin{minipage}[t]{0.49\textwidth}
\begin{tabular}[t]{@{}p{0.27\linewidth}p{0.69\linewidth}@{}}
\toprule
Symbol & Meaning \\
\midrule
\notgroup{Input data (\cref{sec:template,sec:optimization})}
$m$, $n$ & number of cameras, number of frames \\
$\Theta=\{\theta_i\}_{i=1}^m$ & camera parameters \\
$I_{i,t}$, $M_{i,t}$ & RGB image and foreground mask of view $i$ at frame $t$ \\
$V^*_{\tau\to\tau'}$ & precomputed optical flow (FlowSeek) from frame $\tau$ to $\tau'$ \\
\midrule
\notgroup{Canonical representation (\cref{sec:template})}
$t_0$ & canonical time step \\
$f_{\text{sdf}}$, $f_{\text{rgb}}$ & canonical signed distance and radiance fields \\
$\mathbf{z}$ & hash-grid feature vector \\
$x$, $v$ & 3D point, viewing direction \\
$d$, $n$, $c$ & signed distance, unit normal, color \\
$f^{(0)}_{\text{sdf,rgb}}$ & canonical model fit to the first frame \\
$f^{(t)}_{\{\text{sdf,rgb}\}}$ & canonical model, as fine-tuned in the window centered at $t$ \\
\midrule
\notgroup{Deformation and flow (\cref{sec:dynamic_scene})}
$\delta_t$ & deformation field, frame $t$ $\to$ canonical space \\
$\xi_t \in \mathfrak{se}(3)$ & transformation predicted by $\delta_t$ \\
$\mathbf{R}_t$, $\mathbf{T}_t$ & rotation and translation obtained from $\xi_t$ \\
$x_t$, $x_t^\star$ & point at frame $t$, and its canonical position \\
$W$ & sliding window $\{t-1, t, t+1\}$ \\
$\mathrm{prev}$, $\mathrm{curr}$, $\mathrm{next}$ & frames $t-1$, $t$, $t+1$ of the window \\
$\tau$, $\tau'$ & frames of the window \\
$\delta^t_{\tau}$ & deformation field of frame $\tau$ in the window at $t$ \\
$\phi^t_{\tau\to\tau'}$ & flow field from frame $\tau$ to $\tau'$ in the window at $t$ \\
\midrule
\notgroup{Volume rendering (\cref{sec:rendering})}
$\mathbf{r}_i$ & camera ray through a sampled pixel \\
$\mathbf{x}_\tau$, $\mathbf{x}^\star_\tau$ & ray samples of frame $\tau$, and their canonical positions \\
$d_\tau$, $n_\tau$, $c_\tau$ & signed distance, normal and color at the samples \\
$\sigma_\tau$ & volume density \\
$w^i_{\tau,j}$ & rendering weight of sample $i$ on ray $j$ \\
$N$ & number of samples per ray \\
$C_{\tau,j}$, $O_{\tau,j}$ & rendered color and soft mask of ray $j$ \\
$\mathbf{v}_{\tau\to\tau'}$ & 3D displacement given by $\phi^t_{\tau\to\tau'}$ \\
$V_{\tau\to\tau',j}$ & rendered 2D optical flow of ray $j$ \\
$P_\tau$ & projection of 3D displacements to the image of frame $\tau$ \\
\bottomrule
\end{tabular}
\end{minipage}\hfill
\begin{minipage}[t]{0.49\textwidth}
\begin{tabular}[t]{@{}p{0.27\linewidth}p{0.69\linewidth}@{}}
\toprule
Symbol & Meaning \\
\midrule
\notgroup{Motion-guided sampling (\cref{sec:supp_sampling})}
$\mathbf{u}$, $\Omega$ & pixel position, set of all pixel positions \\
$\mathcal{S}_{i,t}$ & core: union of the foreground masks of frames $t-1$ and $t$ (view $i$) \\
$\mathcal{D}_{i,t}$ & core dilated by 8 pixels \\
$\Omega \setminus \mathcal{D}_{i,t}$ & outer region \\
$\mu(\mathbf{u})$ & motion value (larger of the two flow magnitudes) \\
$\mu_{\mathrm{ref}}$, $\mu_{\min}$ & reference motion scale and its lower bound \\
$\operatorname{pct}_{95}$ & 95th percentile \\
$\hat\mu(\mathbf{u})$ & normalized motion score in $[0,1]$ \\
$\kappa$ & score threshold for a moving pixel \\
$\mathcal{A}_{i,t}$ & moving pixels \\
$\rho_\alpha(\mathbf{u})$ & sampling weight inside $\mathcal{D}_{i,t}$ \\
$\alpha$, $\alpha_{i,t}$, $\alpha_{\max}$ & motion emphasis, its per-image value, its cap \\
$\psi(\alpha)$ & share of $\mathcal{D}_{i,t}$-rays on moving pixels \\
$\gamma$ & target share of rays on moving pixels \\
$\eta$ & minimum fraction of moving core pixels \\
$p_{i,t}(\mathbf{u})$ & probability of sampling pixel $\mathbf{u}$ \\
\midrule
\notgroup{Losses (\cref{sec:optimization})}
$R$ & rays per frame in a mini-batch \\
$\mathcal{L}_{\text{rgb}}$, $\mathcal{L}_{\text{mask}}$ & color and mask losses \\
$\mathcal{L}_{\text{data}}$ & data term, $\lambda_{\text{rgb}}\mathcal{L}_{\text{rgb}} + \lambda_{\text{mask}}\mathcal{L}_{\text{mask}}$ \\
$\mathcal{L}_{\text{sdf}}$ & eikonal loss after deformation \\
$\mathcal{L}^{\star}_{\text{sdf}}$ & eikonal loss in canonical space \\
$\bar{x}_{\tau,j}$ & volume-rendered surface point of ray $j$ \\
$\mathbf{J}_g$, $\nabla_x \xi_g$ & Jacobian of field $g$, gradient of its $\mathfrak{se}(3)$ output \\
$\ell_{\mathrm{arap}}$, $\ell_{\mathrm{smooth}}$ & ARAP and smoothness terms of one field \\
$\mathcal{L}^{\delta}_{\star}$, $\mathcal{L}^{\phi}_{\star}$ & terms averaged over deformation / flow fields \\
$\mathcal{L}_{\mathrm{flow}}$ & optical flow loss \\
$x_{\tau'} = \phi_{\tau\to\tau'}(x_\tau)$ & point corresponding to $x_\tau$ in frame $\tau'$ \\
$\mathcal{P}$ & the four (source, target) frame pairs of the window \\
$n^\star$ & normal of the canonical SDF \\
$\bar{n}^\star_{\tau,j}$, $\bar{n}^\star_{\tau\to\tau',j}$ & composited canonical normals of ray $j$, direct and through the flow \\
$e_{\tau\to\tau',j}$ & normal disagreement of ray $j$ \\
$\omega_{\tau\to\tau',j}$, $s_{\tau\to\tau',j}$ & ray weight in $\mathcal{L}_{\mathrm{norm}}$, motion factor \\
$r$, $s_{\min}$ & flow-magnitude percentile and floor of the motion factor \\
$\mathcal{L}_{\mathrm{norm}}$ & normal consistency loss \\
$\lambda_\star$ & weight of loss $\mathcal{L}_\star$ \\
$\mathrm{BCE}$ & binary cross-entropy \\
\midrule
\notgroup{Correspondence estimation (\cref{sec:correspondence})}
$\mathcal{M}_t$, $\mathcal{V}_t$ & mesh extracted at frame $t$, its vertex set \\
$\delta_t$, $\phi_{t\to t+1}$ & final deformation and forward flow of frame $t$ \\
$\pi_{s\to t}$ & vertex correspondence from $\mathcal{M}_s$ to $\mathcal{M}_t$ \\
$a_t$ & per-vertex attribute of $\mathcal{M}_t$ (\eg, texture coordinates) \\
\midrule
\notgroup{Lite variant (\cref{sec:geo-variant})}
$\mathbf{u}'_{\tau',j}$ & pixel paired with ray $j$ in frame $\tau'$ \\
$c_{\tau',j}$, $\sigma_{\mathrm{fb}}$ & forward--backward confidence of the pair, its scale \\
$\mathcal{L}^{\mathrm{geo}}_{\mathrm{norm}}$ & normal consistency on ray pairs \\
\bottomrule
\end{tabular}
\end{minipage}
\caption{\textbf{Notation} used in the paper, grouped by the section where each symbol is introduced.}
\label{tab:notation}
\end{table*}

%% file: tables/hyperparameters.tex
\begin{table}[t]
\centering
\footnotesize
\setlength{\tabcolsep}{7pt}
\renewcommand{\arraystretch}{1.0}
\resizebox{\linewidth}{!}{\begin{tabular}{@{}llc@{}}
\toprule
Parameter & Symbol & Value \\
\midrule
\multicolumn{3}{@{}l}{\textit{Architecture}} \\
SDF / deformation hash grid: levels & & 16 \\
\quad resolution (log$_2$), table size (log$_2$) & & 5--11, 20 \\
\quad feature dim.\ (SDF / deformation) & & 4 / 2 \\
SDF MLP (layers $\times$ width) & & $1 \times 64$ \\
Color MLP (layers $\times$ width) & & $2 \times 64$ \\
Deformation MLP (layers $\times$ width) & & $1 \times 64$ \\
Deformation coarse-to-fine: initial levels, step & & 4, 140 iter. \\
\midrule
\multicolumn{3}{@{}l}{\textit{Optimization}} \\
Optimizer, learning rate, weight decay & & AdamW, $10^{-4}$, $10^{-2}$ \\
Learning-rate schedule & & cosine \\
Iterations per window & & 8{,}500 \\
Rays per image & $R$ & 512 \\
Samples per ray (coarse + hierarchical) & $N$ & $64 + 4 \times 16$ \\
\midrule
\multicolumn{3}{@{}l}{\textit{Loss weights}} \\
Color & $\lambda_{\text{rgb}}$ & 1.0 \\
Mask & $\lambda_{\text{mask}}$ & 0.1 \\
Eikonal (deformed / canonical) & $\lambda_{\text{sdf}}$ / $\lambda^{\star}_{\text{sdf}}$ & 0.1 / 0.1 \\
Optical flow & $\lambda_{\text{flow}}$ & $10^{-3}$ \\
Normal consistency & $\lambda_{\text{norm}}$ & 0.05 \\
Deformation ARAP / smoothness & $\lambda^{\delta}_{\mathrm{arap}}$ / $\lambda^{\delta}_{\mathrm{smooth}}$ & 0.05 / 0.005 \\
Flow ARAP / smoothness & $\lambda^{\phi}_{\mathrm{arap}}$ / $\lambda^{\phi}_{\mathrm{smooth}}$ & 0.05 / 0.005 \\
\midrule
\multicolumn{3}{@{}l}{\textit{Motion-guided sampling (\cref{sec:supp_sampling})}} \\
Core dilation & & 8 px \\
Rays in $\mathcal{D}_{i,t}$ / outer region & & 90\% / 10\% \\
Reference percentile & & 95th \\
Lower bound of reference scale & $\mu_{\min}$ & 2 px \\
Moving-pixel threshold & $\kappa$ & 0.25 \\
Minimum moving fraction & $\eta$ & 0.02 \\
Target share & $\gamma$ & 0.5 \\
Cap on motion emphasis & $\alpha_{\max}$ & 0.9 \\
\midrule
\multicolumn{3}{@{}l}{\textit{Correspondence estimation (\cref{sec:correspondence})}} \\
Marching Cubes resolution & & $2048^3$ \\
\midrule
\multicolumn{3}{@{}l}{\textit{Training time on one NVIDIA H200 (approximate)}} \\
One window, Ours / Ours (Lite) & & ${\approx}\,43$ / $29$\,min \\
50-frame sequence, Ours / Ours (Lite) & & ${\approx}\,35$ / $23$\,h \\
\bottomrule
\end{tabular}}
\caption{\textbf{Hyperparameters} of our final configuration, shared by all datasets.}
\label{tab:hyperparameters}
\end{table}

%% file: tables/runtime.tex
\begin{table}[t]
\addtocounter{table}{-1}\renewcommand{\thetable}{S\arabic{table}a}
\centering
\footnotesize
\setlength{\tabcolsep}{4pt}
\begin{tabular}{@{}lc@{}}
\toprule
Method & Training time \\
\midrule
\multicolumn{2}{@{}l}{\emph{Dynamic, SDF-based}} \\
NeuS2 (dynamic)~\cite{wang2023neus2} & 11\,min \\
MoAngelo~\cite{ebbed2026moangelo} & 17.2\,h \\
\midrule
\multicolumn{2}{@{}l}{\emph{Dynamic, Gaussian-based}} \\
Dynamic 3DGS~\cite{Luiten2024} & 27\,min \\
Dynamic-2DGS~\cite{zhang2024dynamic2dgaussiansgeometrically} & 59\,min \\
AT-GS~\cite{chen2025atgs} & 18\,min \\
4DGS~\cite{4dgaussians} & 9\,min \\
SC-GS~\cite{huang2024scgs} & 29\,min \\
Deformable 3DGS~\cite{yang2024deformable} & 11\,min \\
DG-Mesh~\cite{Liu2025} & 39\,min \\
\midrule
\multicolumn{2}{@{}l}{\emph{Per-frame static}} \\
NeuS2 (static)~\cite{wang2023neus2} & 2.1\,h \\
2DGS~\cite{Huang2DGS2024} & 6.6\,h \\
PGSR~\cite{chen2024pgsr} & 11.6\,h \\
\midrule
\textbf{Ours (Lite)} & 23\,h \\
\textbf{Ours} & 35\,h \\
\bottomrule
\end{tabular}
\caption{\textbf{Training time} for one 50-frame sequence on one NVIDIA H200, each method with its official default settings: mean time over an ActorsHQ (actor4\_seq2\_1258) and a DeformingThings4D (peasantgirl\_Martelo2) sequence, without mesh extraction. Per-frame methods: mean of 3 frames $\times\,50$. Tensor4D, SceNeRFlow, GauSTAR and MaGS were trained on another cluster and are not timed here.}
\label{tab:runtime}
\end{table}

%% file: figures/window_alg.tex
\begin{figure}[t]
\centering
\begin{minipage}{\columnwidth}
\scriptsize\raggedright \hrule height 0.8pt\vspace{2pt}
\textbf{Sliding-window optimization:} one window, centered at frame $t$
\vspace{2pt}\hrule\vspace{3pt}
\begin{algorithmic}[1]
\Require all fields after the window centered at $t-1$
\Ensure mesh $\mathcal{M}_{t-1}$ and flow field $\phi_{t-1\to t}$
\Statex \textit{Shift by one frame (\cref{fig:init})}
\State Store the fields of frame $t-2$; they no longer change
\State $\delta_{\mathrm{prev}} \gets \delta_{\mathrm{curr}}$, \ $\delta_{\mathrm{curr}} \gets \delta_{\mathrm{next}}$
\State $\phi_{\mathrm{prev}\to\mathrm{curr}} \gets \phi_{\mathrm{curr}\to\mathrm{next}}$, \ $\phi_{\mathrm{curr}\to\mathrm{prev}} \gets \phi_{\mathrm{next}\to\mathrm{curr}}$
\State $\delta_{\mathrm{next}}, \phi_{\mathrm{curr}\to\mathrm{next}}, \phi_{\mathrm{next}\to\mathrm{curr}}, f_{\mathrm{sdf}}$ and the color field keep their weights
\Statex \textit{Optimize}
\For{$i = 1, \dots, 8{,}500$}
  \State Sample rays of frames $t-1, t, t+1$ (\Cref{sec:sampling})
  \State Compute $\mathcal{L}_{\mathrm{data}}, \mathcal{L}_{\mathrm{sdf}}, \mathcal{L}^{\star}_{\mathrm{sdf}}$ of each frame, $\mathcal{L}_{\mathrm{flow}}$ on the four flow fields, $\mathcal{L}_{\mathrm{norm}}$ on the four pairs, $\mathcal{L}^{\delta}_{\mathrm{arap}}, \mathcal{L}^{\delta}_{\mathrm{smooth}}, \mathcal{L}^{\phi}_{\mathrm{arap}}, \mathcal{L}^{\phi}_{\mathrm{smooth}}$
  \State Per $\mathcal{L}_{\mathrm{norm}}$ pair, fix the normal of the frame optimized in more windows
  \State Update $\delta_{\mathrm{prev}}, \delta_{\mathrm{curr}}, \delta_{\mathrm{next}}$, the four $\phi$, $f_{\mathrm{sdf}}$ and the color field
\EndFor
\Statex \textit{Finalize frame $t-1$}
\State $\mathcal{M}_{t-1} \gets$ mesh of $f_{\mathrm{sdf}}$ deformed by $\delta_{\mathrm{prev}}$
\State $\phi_{t-1\to t} \gets \phi_{\mathrm{prev}\to\mathrm{curr}}$ \Comment{for the correspondences (\Cref{sec:correspondence})}
\end{algorithmic}
\vspace{2pt}\hrule height 0.8pt
\end{minipage}
\caption{\textbf{One window} of the sliding-window optimization (\Cref{sec:window}), centered at frame $t$ (prev $=t-1$, curr $=t$, next $=t+1$).}
\label{fig:window_alg}
\end{figure}

%% file: figures/sampling_col.tex
\begin{figure}[t]
\centering
\definecolor{tzIn}{RGB}{74,82,99}\definecolor{tzOurs}{RGB}{14,128,229}\definecolor{tzApp}{RGB}{226,104,38}
\definecolor{tzPink}{RGB}{232,46,130}\definecolor{tzInk}{RGB}{24,28,40}
\begin{tikzpicture}[x=1cm, y=1cm, font=\scriptsize, text=tzInk,
    img/.style={inner sep=0pt, outer sep=0pt, anchor=south west},
    tcap/.style={inner sep=1pt, anchor=north, font=\fontsize{6.5}{7.5}\selectfont},
    op/.style={inner sep=1pt, font=\fontsize{6}{7}\selectfont, align=center, text=black!70},
    stepA/.style={-{Latex[length=3pt]}, black!60, line width=0.7pt}]
\def\th{1.25cm}
\foreach \n/\f/\x in {P/s_prev/0, C/s_curr/0.8, B/s_flow_bw/2.15, F/s_flow_fw/3.2, M/s_score/4.75, R/s_regions/6.4}
  \node[img] (\n) at (\x,1.75) {\includegraphics[height=\th]{imgs/sampling/\f}};
\node[tcap] at (P.south) {$I_{t-1}$};
\node[tcap] at (C.south) {$I_t$};
\node[tcap] at (B.south) {$V^*_{t\to t-1}$};
\node[tcap] at (F.south) {$V^*_{t-1\to t}$};
\node[tcap] at (M.south) {$\hat\mu$};
\node[tcap] at (R.south) {regions};
\draw[stepA] ([xshift=2pt]C.east) -- ([xshift=-2pt]B.west) node[op, midway, above=1pt] {optical\\flow};
\draw[stepA] ([xshift=2pt]F.east) -- ([xshift=-2pt]M.west) node[op, midway, above=1pt] {max,\\$\div\,\mathrm{pct}_{95}$};
\draw[stepA] ([xshift=2pt]M.east) -- ([xshift=-2pt]R.west) node[op, midway, above=1pt] {masks,\\dilate};
\foreach \c/\t/\y in {tzOurs/{$\mathcal{S}$}/2.85, tzApp/{$\mathcal{D}\setminus\mathcal{S}$}/2.6, black!12/{$\Omega\setminus\mathcal{D}$}/2.35}
  { \fill[\c] (7.25,\y) rectangle +(0.13,0.13); \node[op, anchor=west, inner sep=0.5pt] at (7.4,\y+0.065) {\t}; }
\node[op, anchor=north west, align=left, inner sep=0.5pt] at (7.22,2.2) {90\% of\\rays in $\mathcal{D}$,\\10\% out};
\node[img] (D) at (0,0) {\includegraphics[height=\th]{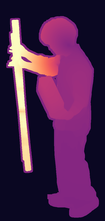}};
\node[tcap] at (D.south) {density $p$};
\begin{axis}[at={(1.85cm,0.22cm)}, anchor=south west, scale only axis, width=3.5cm, height=0.95cm,
    xmin=0, xmax=1, ymin=0, ymax=85, xtick={0,0.5,1}, ytick={0,25,50},
    tick label style={font=\fontsize{6}{7}\selectfont}, label style={font=\fontsize{6}{7}\selectfont},
    xlabel={$\alpha$}, ylabel={moving (\%)}, xlabel shift=-5pt, ylabel shift=-6pt,
    tick align=outside, major tick length=2pt, axis line style={black!60}, every tick/.style={black!60}, clip=false]
  \fill[black!8] (axis cs:0.9,0) rectangle (axis cs:1,85);
  \addplot[tzIn, line width=0.9pt] table[x=alpha, y=share] {imgs/sampling/s_share.dat};
  \draw[tzPink, densely dashed, line width=0.6pt] (axis cs:0,50) -- (axis cs:1,50);
  \node[op, anchor=north west, text=tzPink, inner sep=0.5pt] at (axis cs:0.02,48) {target $\gamma$};
  \node[op, anchor=south east, inner sep=0.5pt] at (axis cs:0.9,85) {$\alpha_{\max}$};
  \draw[tzPink, densely dotted, line width=0.6pt] (axis cs:0.857,0) -- (axis cs:0.857,50);
  \fill[tzPink] (axis cs:0.857,50) circle (1.4pt);
  \node[op, anchor=north east, text=tzPink, inner sep=1pt] at (axis cs:0.83,47) {$\alpha^*$};
  \node[op, anchor=north west, inner sep=1pt, text=tzIn] at (axis cs:0.03,84) {$\rho_\alpha = (1{-}\alpha) + \alpha\hat\mu$};
\end{axis}
\foreach \n/\f/\x in {U/s_rays_uniform/5.95, O/s_rays_ours/7.25}
  \node[img] (\n) at (\x,0) {\includegraphics[height=\th]{imgs/sampling/\f}};
\node[tcap, align=center] at (U.south) {uniform 1.9\%};
\node[tcap, align=center, text=tzPink] at (O.south) {ours 45.0\%};
\end{tikzpicture}
\caption{\textbf{Motion-guided sampling} on \Cref{fig:pipeline}(a): $\alpha^*$ is set so that a share $\gamma$ of the rays in $\mathcal{D}$ hits moving pixels.}
\label{fig:sampling_tech}
\end{figure}

%% file: tables/perscene_cd.tex
\begin{table*}[t]
\centering
\scriptsize
\setlength{\tabcolsep}{2.2pt}
\resizebox{\textwidth}{!}{\begin{tabular}{@{}lccccccc|ccccccccccc@{}}
\toprule
 & \multicolumn{7}{c|}{ActorsHQ} & \multicolumn{11}{c}{DeformingThings4D} \\
Method & a4-1258 & a5-1942 & a6-1081 & a6-1604 & a7-554 & a8-325 & Mean & abe & AJ & canie & dragon & fox & grizz & ninja & peasant. & rabbit & vampire & Mean \\
\midrule
\multicolumn{19}{@{}l}{\emph{Dynamic, SDF-based}} \\
Tensor4D & 7.34 & 12.57 & 6.32 & 5.90 & 9.77 & 8.03 & 8.32 & 10.47 & 10.57 & 9.42 & -- & 8.62 & 44.27 & 6.23 & 22.51 & 5.03 & 30.27 & 16.38$^\dagger$ \\
NeuS2 & \rankC{4.31} & \rankC{4.25} & \rankC{4.56} & \rankB{4.44} & \rankC{3.13} & \rankB{3.59} & \rankC{4.05} & \rankC{7.86} & 9.18 & \rankB{3.72} & 43.74 & \rankB{2.77} & \rankB{6.82} & 6.31 & \rankC{11.66} & 2.28 & 13.89 & \rankC{10.82} \\
MoAngelo & \rankB{3.90} & \rankB{3.14} & \rankB{3.38} & \rankC{4.46} & \rankB{2.79} & \rankC{4.06} & \rankB{3.62} & 8.39 & 8.83 & \rankC{5.51} & \rankB{29.70} & \rankC{2.82} & \rankC{7.17} & \rankB{3.59} & 14.27 & \rankB{1.97} & \rankC{12.82} & \rankB{9.51} \\
\multicolumn{19}{@{}l}{\emph{Dynamic, NeRF-based}} \\
SceNeRFlow & 18.63 & 16.33 & 16.16 & 17.81 & 8.49 & 21.26 & 16.45 & 21.50 & 31.12 & 23.44 & 118.93 & 18.61 & 60.10 & 21.47 & 36.50 & 11.64 & 27.88 & 37.12 \\
\multicolumn{19}{@{}l}{\emph{Dynamic, Gaussian-based}} \\
Dynamic 3DGS & 7.69 & 6.74 & 6.59 & 8.13 & 6.91 & 7.68 & 7.29 & 103.95 & 9.16 & 105.07 & 39.66 & 6.73 & 174.60 & 10.29 & 94.60 & 3.16 & 89.64 & 63.69 \\
Dynamic-2DGS & -- & -- & -- & -- & 26.83 & 70.23 & 48.53$^\dagger$ & 8.71 & 442.92$^\ast$ & -- & 57.16 & 5.96 & -- & -- & -- & -- & -- & 23.94$^\dagger$ \\
AT-GS & 5.27 & 4.95 & 5.15 & 5.39 & 5.44 & 4.92 & 5.18 & \rankB{7.10} & \rankC{8.62} & 5.83 & 116.49 & 4.33 & 9.84 & 7.10 & \rankB{10.70} & \rankC{2.23} & \rankB{12.67} & 18.49 \\
4DGS & 7.93 & 10.57 & 8.91 & 9.37 & 17.32 & 9.83 & 10.65 & 17.31 & 9.78 & 10.02 & 74.97 & 7.25 & 15.46 & 19.35 & 19.73 & 3.69 & 18.11 & 19.57 \\
SC-GS & 8.13 & 5.41 & 6.12 & 7.26 & 5.85 & 8.11 & 6.81 & 7.93 & \rankB{8.19} & 9.55 & \rankC{36.32} & 4.45 & 11.42 & 5.84 & 18.39 & 4.02 & 15.86 & 12.20 \\
Deformable 3DGS & 7.82 & 7.26 & 8.01 & 7.62 & 7.86 & 9.94 & 8.08 & -- & 9.35 & 12.19 & 39.81 & 7.22 & 13.59 & 9.36 & 20.04 & 4.22 & 18.30 & 14.90$^\dagger$ \\
\multicolumn{19}{@{}l}{\emph{Dynamic, mesh-based Gaussian}} \\
GauSTAR & 9.21 & 6.38 & 5.77 & 5.80 & 5.47 & 6.10 & 6.46 & 8.57 & 10.53 & 58.29 & 1243.90 & 7.29 & 62.26 & 5.37 & 47.16 & 5.92 & 42.65 & 149.20 \\
MaGS & 15.83 & 11.80 & 11.33 & 15.13 & 8.74 & 27.54 & 15.06 & -- & 11.03 & 35.35 & 55.38 & 16.77 & 23.66 & 25.15 & 35.02 & 13.94 & 23.59 & 26.65$^\dagger$ \\
DG-Mesh & 7.83 & 5.70 & 6.38 & 7.41 & 4.57 & 8.91 & 6.80 & -- & 10.47 & 15.80 & 82.05 & 6.22 & 20.71 & \rankC{5.27} & 24.05 & 5.69 & 35.13 & 22.82$^\dagger$ \\
\multicolumn{19}{@{}l}{\emph{Per-frame static}} \\
2DGS & 4.07 & 3.60 & 3.86 & 3.85 & 3.44 & 3.63 & 3.74 & 5.98 & 7.61 & 3.92 & 26.86 & 3.09 & 8.26 & 5.55 & 9.25 & 1.61 & 10.85 & 8.30 \\
PGSR & 3.10 & 2.86 & 3.11 & 3.02 & 2.63 & 2.82 & 2.92 & 5.46 & 7.91 & 46.49 & 26.78 & 28.42 & 8.42 & 3.47 & 8.20 & 12.01 & 10.94 & 15.81 \\
NeuS2 (per-frame) & 4.22 & 3.74 & 4.13 & 4.36 & 3.60 & 4.70 & 4.12 & 6.59 & 9.70 & 3.75 & 39.50 & 2.76 & 6.71 & 5.01 & 11.95 & 1.75 & 15.33 & 10.30 \\
\midrule
\textbf{Ours} & \rankA{3.42} & \rankA{3.10} & \rankA{3.08} & \rankA{3.22} & \rankA{2.65} & \rankA{3.28} & \rankA{3.12} & \rankA{5.28} & \rankA{7.45} & \rankA{3.06} & \rankA{25.57} & \rankA{2.18} & \rankA{6.06} & \rankA{3.24} & \rankA{10.20} & \rankA{1.34} & \rankA{11.85} & \rankA{7.62} \\
\bottomrule
\end{tabular}}
\caption{\textbf{Per-scene Chamfer distance} (mm, $\downarrow$) on ActorsHQ and DeformingThings4D. \ranklegend{} among the dynamic methods; --: failed; $^\ast$failed run, left out of the mean; $^\dagger$mean over the scenes with a result.}
\label{tab:perscene_cd}
\end{table*}

%% file: tables/perscene_nc.tex
\begin{table*}[t]
\centering
\scriptsize
\setlength{\tabcolsep}{2.2pt}
\resizebox{\textwidth}{!}{\begin{tabular}{@{}lccccccc|ccccccccccc@{}}
\toprule
 & \multicolumn{7}{c|}{ActorsHQ} & \multicolumn{11}{c}{DeformingThings4D} \\
Method & a4-1258 & a5-1942 & a6-1081 & a6-1604 & a7-554 & a8-325 & Mean & abe & AJ & canie & dragon & fox & grizz & ninja & peasant. & rabbit & vampire & Mean \\
\midrule
\multicolumn{19}{@{}l}{\emph{Dynamic, SDF-based}} \\
Tensor4D & 0.879 & 0.876 & 0.877 & 0.877 & 0.905 & 0.891 & 0.884 & 0.845 & \rankC{0.888} & 0.887 & -- & 0.886 & 0.880 & \rankC{0.941} & 0.824 & 0.880 & 0.732 & 0.863$^\dagger$ \\
NeuS2 & \rankC{0.911} & 0.921 & \rankC{0.897} & \rankC{0.899} & \rankC{0.935} & \rankC{0.926} & 0.915 & \rankC{0.887} & 0.884 & \rankC{0.947} & \rankC{0.868} & \rankC{0.960} & \rankC{0.966} & 0.933 & \rankC{0.879} & \rankC{0.934} & \rankC{0.817} & \rankC{0.907} \\
MoAngelo & \rankC{0.911} & \rankB{0.935} & \rankB{0.908} & 0.885 & \rankB{0.948} & 0.910 & \rankC{0.916} & 0.828 & 0.850 & 0.891 & \rankB{0.871} & 0.944 & 0.963 & \rankB{0.946} & 0.819 & 0.914 & 0.759 & 0.878 \\
\multicolumn{19}{@{}l}{\emph{Dynamic, NeRF-based}} \\
SceNeRFlow & 0.696 & 0.712 & 0.686 & 0.642 & 0.768 & 0.728 & 0.705 & 0.593 & 0.641 & 0.534 & 0.587 & 0.582 & 0.589 & 0.620 & 0.581 & 0.521 & 0.581 & 0.583 \\
\multicolumn{19}{@{}l}{\emph{Dynamic, Gaussian-based}} \\
Dynamic 3DGS & 0.729 & 0.744 & 0.728 & 0.716 & 0.755 & 0.739 & 0.735 & 0.717 & 0.719 & 0.734 & 0.732 & 0.771 & 0.740 & 0.720 & 0.634 & 0.861 & 0.618 & 0.725 \\
Dynamic-2DGS & -- & -- & -- & -- & 0.641 & 0.591 & 0.616$^\dagger$ & 0.710 & 0.533$^\ast$ & -- & 0.653 & 0.780 & -- & -- & -- & -- & -- & 0.714$^\dagger$ \\
AT-GS & \rankB{0.921} & \rankC{0.923} & \rankB{0.908} & \rankB{0.907} & 0.931 & \rankB{0.935} & \rankB{0.921} & \rankB{0.891} & \rankB{0.895} & \rankB{0.953} & 0.812 & \rankB{0.969} & \rankB{0.973} & 0.929 & \rankA{0.893} & \rankB{0.962} & \rankB{0.834} & \rankB{0.911} \\
4DGS & 0.712 & 0.696 & 0.705 & 0.705 & 0.647 & 0.712 & 0.696 & 0.619 & 0.711 & 0.749 & 0.652 & 0.785 & 0.788 & 0.668 & 0.655 & 0.854 & 0.605 & 0.709 \\
SC-GS & 0.760 & 0.797 & 0.767 & 0.751 & 0.769 & 0.756 & 0.767 & 0.753 & 0.794 & 0.752 & 0.776 & 0.840 & 0.793 & 0.824 & 0.677 & 0.844 & 0.633 & 0.768 \\
Deformable 3DGS & 0.747 & 0.759 & 0.736 & 0.741 & 0.726 & 0.754 & 0.744 & -- & 0.735 & 0.745 & 0.745 & 0.804 & 0.788 & 0.737 & 0.689 & 0.849 & 0.630 & 0.747$^\dagger$ \\
\multicolumn{19}{@{}l}{\emph{Dynamic, mesh-based Gaussian}} \\
GauSTAR & 0.874 & 0.894 & 0.890 & 0.890 & 0.915 & 0.907 & 0.895 & 0.875 & 0.868 & 0.660 & 0.573 & 0.924 & 0.767 & \rankC{0.941} & 0.739 & 0.876 & 0.673 & 0.790 \\
MaGS & 0.717 & 0.743 & 0.739 & 0.687 & 0.814 & 0.650 & 0.725 & -- & 0.775 & 0.617 & 0.700 & 0.678 & 0.847 & 0.743 & 0.649 & 0.624 & 0.633 & 0.696$^\dagger$ \\
DG-Mesh & 0.857 & 0.885 & 0.867 & 0.855 & 0.907 & 0.855 & 0.871 & -- & 0.871 & 0.801 & 0.783 & 0.897 & 0.929 & 0.915 & 0.768 & 0.792 & 0.677 & 0.826$^\dagger$ \\
\multicolumn{19}{@{}l}{\emph{Per-frame static}} \\
2DGS & 0.900 & 0.914 & 0.890 & 0.893 & 0.924 & 0.924 & 0.907 & 0.871 & 0.890 & 0.940 & 0.869 & 0.954 & 0.939 & 0.910 & 0.880 & 0.965 & 0.818 & 0.904 \\
PGSR & 0.913 & 0.932 & 0.904 & 0.908 & 0.941 & 0.945 & 0.924 & 0.882 & 0.893 & 0.737 & 0.880 & 0.777 & 0.948 & 0.935 & 0.887 & 0.809 & 0.815 & 0.856 \\
NeuS2 (per-frame) & 0.897 & 0.913 & 0.887 & 0.880 & 0.912 & 0.893 & 0.897 & 0.872 & 0.865 & 0.946 & 0.871 & 0.954 & 0.965 & 0.923 & 0.864 & 0.946 & 0.809 & 0.901 \\
\midrule
\textbf{Ours} & \rankA{0.933} & \rankA{0.945} & \rankA{0.930} & \rankA{0.923} & \rankA{0.960} & \rankA{0.947} & \rankA{0.940} & \rankA{0.906} & \rankA{0.910} & \rankA{0.969} & \rankA{0.900} & \rankA{0.980} & \rankA{0.981} & \rankA{0.959} & \rankB{0.887} & \rankA{0.973} & \rankA{0.837} & \rankA{0.930} \\
\bottomrule
\end{tabular}}
\caption{\textbf{Per-scene normal consistency} ($\uparrow$) on ActorsHQ and DeformingThings4D. \ranklegend{} among the dynamic methods; --: failed; $^\ast$failed run, left out of the mean; $^\dagger$mean over the scenes with a result.}
\label{tab:perscene_nc}
\end{table*}

%% file: tables/perscene_tce.tex
\begin{table*}[t]
\centering
\scriptsize
\setlength{\tabcolsep}{2.2pt}
\footnotesize
\begin{tabular}{@{}lccccccccccc@{}}
\toprule
 & \multicolumn{11}{c}{DeformingThings4D} \\
Method & abe & AJ & canie & dragon & fox & grizz & ninja & peasant. & rabbit & vampire & Mean \\
\midrule
\multicolumn{12}{@{}l}{\emph{Dynamic, SDF-based}} \\
Tensor4D & 5.29 & 4.22 & 6.66 & -- & 5.64 & 6.24 & 2.05 & 12.38 & 2.90 & 14.47 & 6.65$^\dagger$ \\
NeuS2 & \rankC{3.07} & 3.90 & \rankC{2.28} & 20.82 & 1.82 & 3.11 & 2.18 & \rankC{4.07} & 1.69 & \rankC{7.53} & \rankC{5.05} \\
MoAngelo & 3.57 & 4.33 & 3.37 & \rankC{17.07} & \rankC{1.66} & \rankC{3.01} & \rankB{1.08} & 6.84 & \rankC{1.19} & 9.21 & 5.13 \\
\multicolumn{12}{@{}l}{\emph{Dynamic, NeRF-based}} \\
SceNeRFlow & 13.71 & 23.95 & 9.08 & 67.97 & 7.44 & 13.52 & 12.30 & 20.90 & 3.82 & 17.12 & 18.98 \\
\multicolumn{12}{@{}l}{\emph{Dynamic, Gaussian-based}} \\
Dynamic 3DGS & 4.30 & 5.72 & 5.26 & 23.64 & 3.22 & 7.11 & 3.85 & 14.39 & 1.47 & 17.51 & 8.65 \\
Dynamic-2DGS & 5.59 & 46.39$^\ast$ & -- & 34.53 & 3.06 & -- & -- & -- & -- & -- & 14.39$^\dagger$ \\
AT-GS & \rankA{1.60} & \rankB{2.09} & \rankB{1.80} & \rankB{12.86} & \rankB{1.06} & \rankB{2.06} & 1.32 & \rankB{3.01} & \rankB{0.72} & \rankB{5.31} & \rankB{3.18} \\
4DGS & 10.88 & 6.48 & 6.24 & 56.33 & 4.16 & 8.60 & 8.75 & 11.53 & 2.27 & 14.62 & 12.98 \\
SC-GS & 4.22 & 4.42 & 5.06 & 21.36 & 2.08 & 5.14 & 1.96 & 9.69 & 2.01 & 11.03 & 6.70 \\
Deformable 3DGS & -- & 5.26 & 6.73 & 25.05 & 3.40 & 6.50 & 3.46 & 12.44 & 2.34 & 14.12 & 8.81$^\dagger$ \\
\multicolumn{12}{@{}l}{\emph{Dynamic, mesh-based Gaussian}} \\
GauSTAR & 3.37 & 3.77 & 20.70 & 171.44 & 3.03 & 24.37 & \rankC{1.25} & 27.98 & 3.23 & 24.88 & 28.40 \\
MaGS & -- & 3.60 & 12.25 & 22.29 & 5.60 & 7.25 & 7.41 & 15.57 & 4.93 & 13.43 & 10.26$^\dagger$ \\
DG-Mesh & -- & \rankC{3.21} & 9.76 & 28.63 & 3.16 & 4.53 & 1.80 & 11.89 & 3.39 & 22.52 & 9.88$^\dagger$ \\
\multicolumn{12}{@{}l}{\emph{Per-frame static}} \\
2DGS & 2.89 & 3.15 & 1.45 & 12.72 & 0.90 & 2.83 & 2.00 & 3.52 & 0.45 & 6.91 & 3.68 \\
PGSR & 2.20 & 2.94 & 27.71 & 11.09 & 15.70 & 2.64 & 1.09 & 3.64 & 6.84 & 7.16 & 8.10 \\
NeuS2 (per-frame) & 2.57 & 4.49 & 1.90 & 19.22 & 1.29 & 2.45 & 1.94 & 4.12 & 1.07 & 8.60 & 4.77 \\
\midrule
\textbf{Ours} & \rankB{1.68} & \rankA{2.06} & \rankA{1.40} & \rankA{11.11} & \rankA{0.95} & \rankA{1.73} & \rankA{0.93} & \rankA{2.93} & \rankA{0.67} & \rankA{5.23} & \rankA{2.87} \\
\bottomrule
\end{tabular}
\caption{\textbf{Per-scene temporal consistency error} (mm, $\downarrow$) on DeformingThings4D (\cref{sec:supp_tce}). \ranklegend{} among the dynamic methods; --: failed; $^\ast$failed run, left out of the mean; $^\dagger$mean over the scenes with a result.}
\label{tab:perscene_tce}
\end{table*}

%% file: figures/geo_gallery.tex
\begin{figure*}[p]
\centering
\definecolor{tzInk}{RGB}{24,28,40}
\begin{tikzpicture}[x=1cm, y=1cm, font=\scriptsize, text=tzInk, cell/.style={anchor=center, inner sep=0},
    val/.style={font=\tiny, anchor=south east, fill=white, fill opacity=0.8, text opacity=1, inner sep=1pt}]
\begin{scope}[shift={(0,0.000)}]
\node[anchor=south west, inner sep=0, font=\footnotesize] at (0.400,5.850) {actor4\_seq2\_1258};
\node[anchor=south, inner sep=1pt] at (1.238,5.540) {2};
\node[cell] at (1.238,4.140) {\includegraphics[width=0.890cm,height=2.217cm]{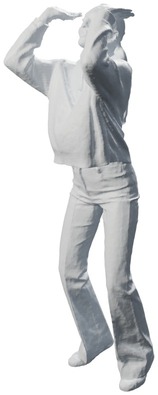}};
\node[val] at (2.055,2.780) {3.00};
\node[cell] at (1.238,1.380) {\includegraphics[width=0.890cm,height=2.217cm]{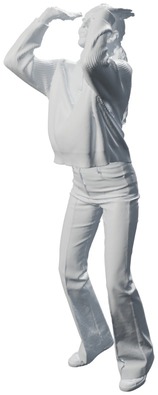}};
\node[anchor=south, inner sep=1pt] at (2.913,5.540) {13};
\node[cell] at (2.913,4.140) {\includegraphics[width=1.044cm,height=2.700cm]{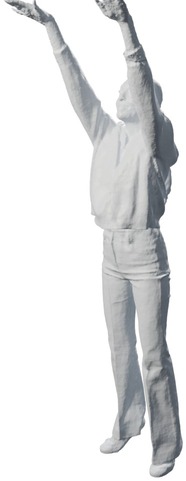}};
\node[val] at (3.730,2.780) {3.29};
\node[cell] at (2.913,1.380) {\includegraphics[width=1.044cm,height=2.700cm]{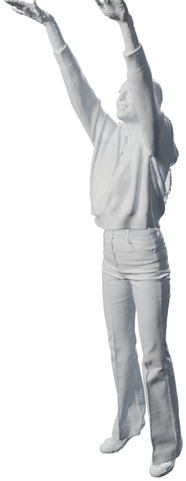}};
\node[anchor=south, inner sep=1pt] at (4.588,5.540) {25};
\node[cell] at (4.588,4.140) {\includegraphics[width=1.252cm,height=2.150cm]{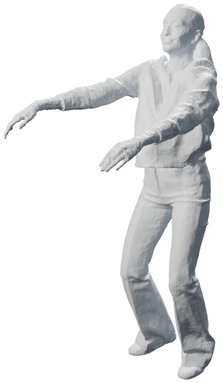}};
\node[val] at (5.405,2.780) {3.43};
\node[cell] at (4.588,1.380) {\includegraphics[width=1.252cm,height=2.150cm]{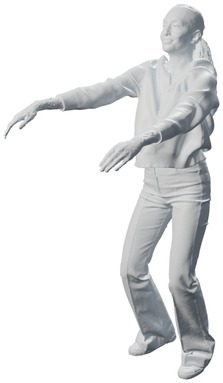}};
\node[anchor=south, inner sep=1pt] at (6.263,5.540) {37};
\node[cell] at (6.263,4.140) {\includegraphics[width=0.935cm,height=2.220cm]{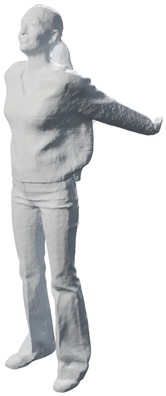}};
\node[val] at (7.080,2.780) {3.61};
\node[cell] at (6.263,1.380) {\includegraphics[width=0.935cm,height=2.220cm]{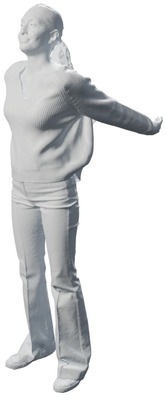}};
\node[anchor=south, inner sep=1pt] at (7.938,5.540) {49};
\node[cell] at (7.938,4.140) {\includegraphics[width=0.817cm,height=2.206cm]{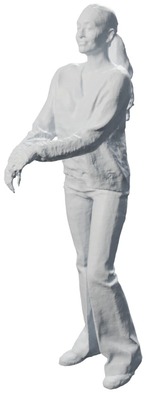}};
\node[val] at (8.755,2.780) {3.40};
\node[cell] at (7.938,1.380) {\includegraphics[width=0.817cm,height=2.206cm]{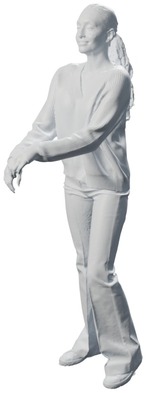}};
\node[rotate=90, anchor=south, inner sep=1pt, font=\scriptsize\bfseries] at (0.400,4.140) {Ours};
\node[rotate=90, anchor=south, inner sep=1pt] at (0.400,1.380) {GT};
\draw[black!15, line width=0.3pt] (0.400,2.760) -- (8.775,2.760);
\end{scope}
\begin{scope}[shift={(0,0.000)}]
\node[anchor=south west, inner sep=0, font=\footnotesize] at (9.025,5.850) {actor5\_seq2\_1942};
\node[anchor=south, inner sep=1pt] at (9.863,5.540) {2};
\node[cell] at (9.863,4.140) {\includegraphics[width=1.154cm,height=2.377cm]{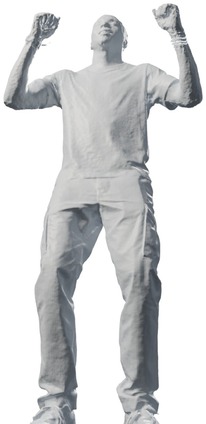}};
\node[val] at (10.680,2.780) {2.81};
\node[cell] at (9.863,1.380) {\includegraphics[width=1.154cm,height=2.377cm]{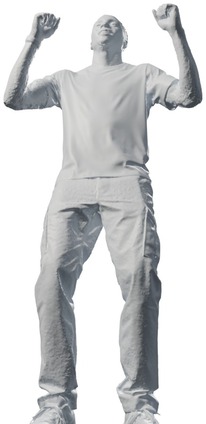}};
\node[anchor=south, inner sep=1pt] at (11.538,5.540) {13};
\node[cell] at (11.538,4.140) {\includegraphics[width=1.083cm,height=2.536cm]{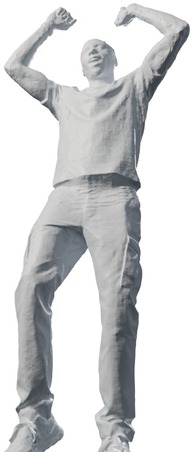}};
\node[val] at (12.355,2.780) {3.11};
\node[cell] at (11.538,1.380) {\includegraphics[width=1.083cm,height=2.536cm]{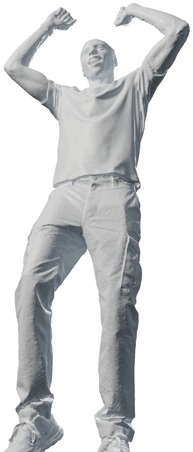}};
\node[anchor=south, inner sep=1pt] at (13.213,5.540) {25};
\node[cell] at (13.213,4.140) {\includegraphics[width=0.971cm,height=2.700cm]{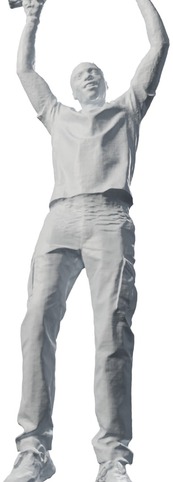}};
\node[val] at (14.030,2.780) {3.11};
\node[cell] at (13.213,1.380) {\includegraphics[width=0.971cm,height=2.700cm]{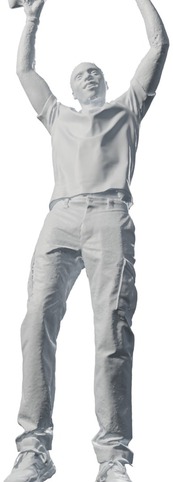}};
\node[anchor=south, inner sep=1pt] at (14.887,5.540) {37};
\node[cell] at (14.887,4.140) {\includegraphics[width=0.897cm,height=2.578cm]{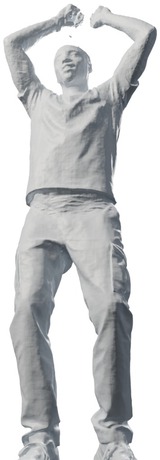}};
\node[val] at (15.705,2.780) {3.24};
\node[cell] at (14.887,1.380) {\includegraphics[width=0.897cm,height=2.578cm]{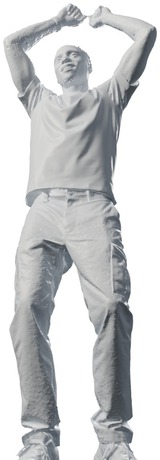}};
\node[anchor=south, inner sep=1pt] at (16.562,5.540) {49};
\node[cell] at (16.562,4.140) {\includegraphics[width=1.101cm,height=2.427cm]{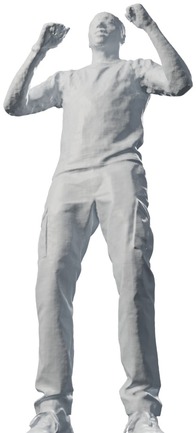}};
\node[val] at (17.380,2.780) {3.02};
\node[cell] at (16.562,1.380) {\includegraphics[width=1.101cm,height=2.427cm]{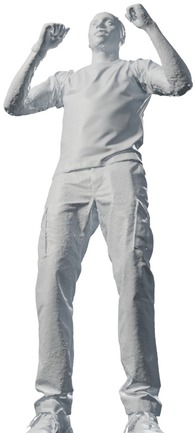}};
\draw[black!15, line width=0.3pt] (9.025,2.760) -- (17.400,2.760);
\end{scope}
\begin{scope}[shift={(0,-6.200)}]
\node[anchor=south west, inner sep=0, font=\footnotesize] at (0.400,5.850) {actor6\_seq2\_1081};
\node[anchor=south, inner sep=1pt] at (1.238,5.540) {2};
\node[cell] at (1.238,4.140) {\includegraphics[width=0.906cm,height=2.687cm]{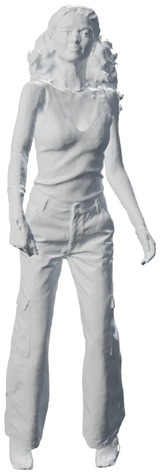}};
\node[val] at (2.055,2.780) {2.97};
\node[cell] at (1.238,1.380) {\includegraphics[width=0.906cm,height=2.687cm]{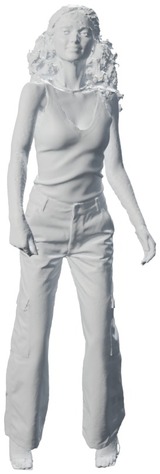}};
\node[anchor=south, inner sep=1pt] at (2.913,5.540) {13};
\node[cell] at (2.913,4.140) {\includegraphics[width=1.071cm,height=2.584cm]{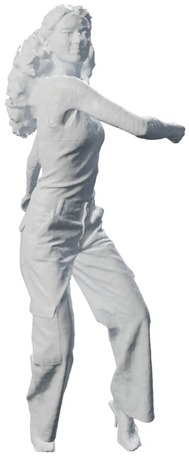}};
\node[val] at (3.730,2.780) {3.20};
\node[cell] at (2.913,1.380) {\includegraphics[width=1.071cm,height=2.584cm]{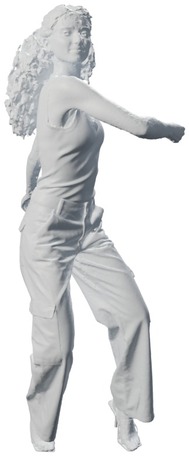}};
\node[anchor=south, inner sep=1pt] at (4.588,5.540) {25};
\node[cell] at (4.588,4.140) {\includegraphics[width=0.777cm,height=2.657cm]{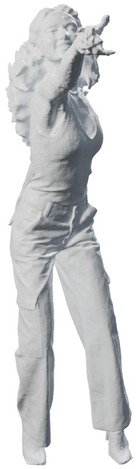}};
\node[val] at (5.405,2.780) {3.13};
\node[cell] at (4.588,1.380) {\includegraphics[width=0.777cm,height=2.657cm]{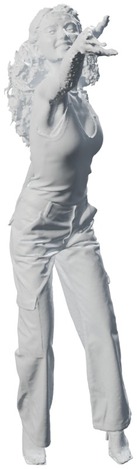}};
\node[anchor=south, inner sep=1pt] at (6.263,5.540) {37};
\node[cell] at (6.263,4.140) {\includegraphics[width=1.345cm,height=2.700cm]{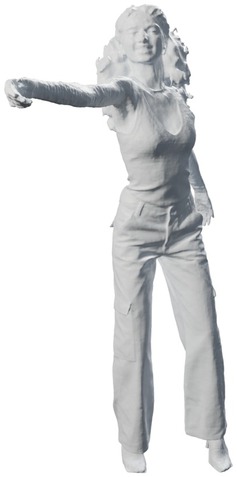}};
\node[val] at (7.080,2.780) {3.02};
\node[cell] at (6.263,1.380) {\includegraphics[width=1.345cm,height=2.700cm]{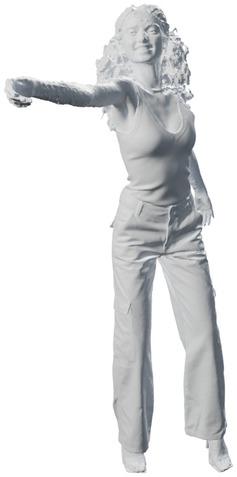}};
\node[anchor=south, inner sep=1pt] at (7.938,5.540) {49};
\node[cell] at (7.938,4.140) {\includegraphics[width=1.110cm,height=2.690cm]{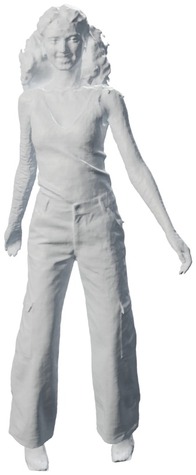}};
\node[val] at (8.755,2.780) {3.15};
\node[cell] at (7.938,1.380) {\includegraphics[width=1.110cm,height=2.690cm]{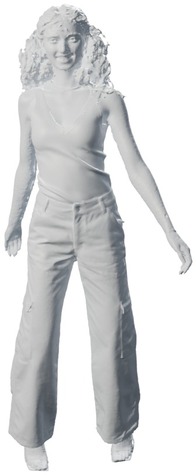}};
\node[rotate=90, anchor=south, inner sep=1pt, font=\scriptsize\bfseries] at (0.400,4.140) {Ours};
\node[rotate=90, anchor=south, inner sep=1pt] at (0.400,1.380) {GT};
\draw[black!15, line width=0.3pt] (0.400,2.760) -- (8.775,2.760);
\end{scope}
\begin{scope}[shift={(0,-6.200)}]
\node[anchor=south west, inner sep=0, font=\footnotesize] at (9.025,5.850) {actor6\_seq2\_1604};
\node[anchor=south, inner sep=1pt] at (9.863,5.540) {2};
\node[cell] at (9.863,4.140) {\includegraphics[width=0.923cm,height=2.639cm]{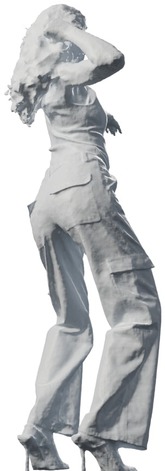}};
\node[val] at (10.680,2.780) {3.04};
\node[cell] at (9.863,1.380) {\includegraphics[width=0.923cm,height=2.639cm]{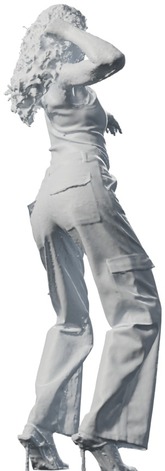}};
\node[anchor=south, inner sep=1pt] at (11.538,5.540) {13};
\node[cell] at (11.538,4.140) {\includegraphics[width=1.148cm,height=2.651cm]{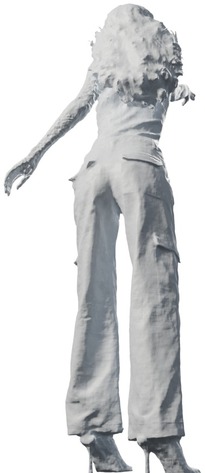}};
\node[val] at (12.355,2.780) {3.34};
\node[cell] at (11.538,1.380) {\includegraphics[width=1.148cm,height=2.651cm]{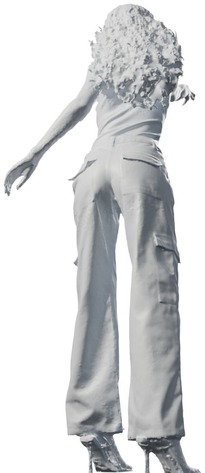}};
\node[anchor=south, inner sep=1pt] at (13.213,5.540) {25};
\node[cell] at (13.213,4.140) {\includegraphics[width=1.096cm,height=2.606cm]{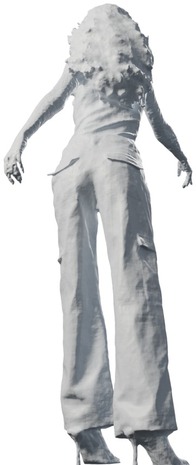}};
\node[val] at (14.030,2.780) {3.13};
\node[cell] at (13.213,1.380) {\includegraphics[width=1.096cm,height=2.606cm]{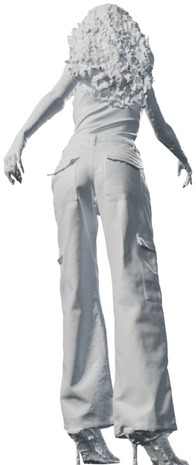}};
\node[anchor=south, inner sep=1pt] at (14.887,5.540) {37};
\node[cell] at (14.887,4.140) {\includegraphics[width=0.966cm,height=2.642cm]{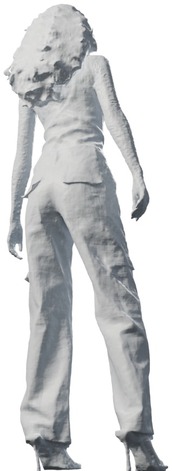}};
\node[val] at (15.705,2.780) {3.15};
\node[cell] at (14.887,1.380) {\includegraphics[width=0.966cm,height=2.642cm]{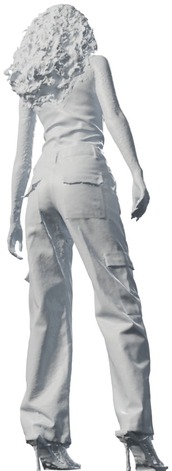}};
\node[anchor=south, inner sep=1pt] at (16.562,5.540) {49};
\node[cell] at (16.562,4.140) {\includegraphics[width=0.863cm,height=2.700cm]{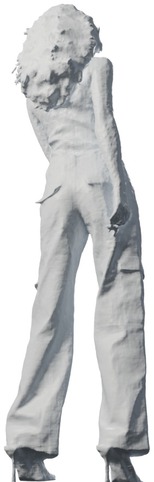}};
\node[val] at (17.380,2.780) {3.36};
\node[cell] at (16.562,1.380) {\includegraphics[width=0.863cm,height=2.700cm]{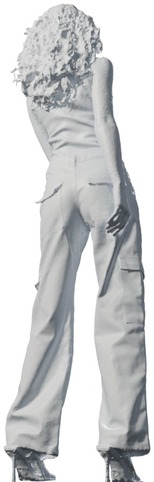}};
\draw[black!15, line width=0.3pt] (9.025,2.760) -- (17.400,2.760);
\end{scope}
\begin{scope}[shift={(0,-12.400)}]
\node[anchor=south west, inner sep=0, font=\footnotesize] at (0.400,5.850) {actor7\_seq1\_554};
\node[anchor=south, inner sep=1pt] at (1.238,5.540) {2};
\node[cell] at (1.238,4.140) {\includegraphics[width=1.283cm,height=2.384cm]{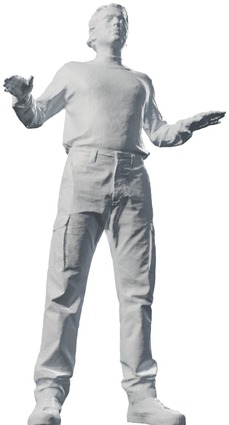}};
\node[val] at (2.055,2.780) {2.59};
\node[cell] at (1.238,1.380) {\includegraphics[width=1.283cm,height=2.384cm]{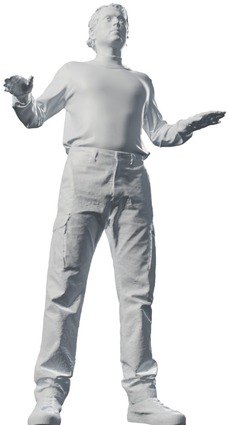}};
\node[anchor=south, inner sep=1pt] at (2.913,5.540) {13};
\node[cell] at (2.913,4.140) {\includegraphics[width=1.193cm,height=2.379cm]{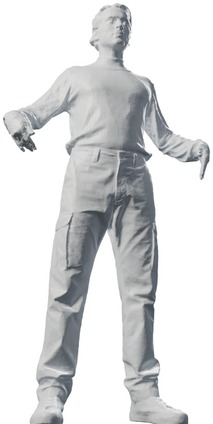}};
\node[val] at (3.730,2.780) {2.61};
\node[cell] at (2.913,1.380) {\includegraphics[width=1.193cm,height=2.379cm]{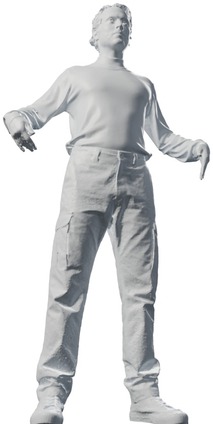}};
\node[anchor=south, inner sep=1pt] at (4.588,5.540) {25};
\node[cell] at (4.588,4.140) {\includegraphics[width=1.566cm,height=2.379cm]{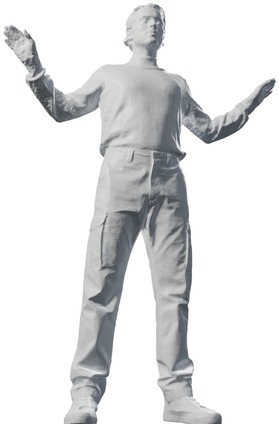}};
\node[val] at (5.405,2.780) {2.62};
\node[cell] at (4.588,1.380) {\includegraphics[width=1.566cm,height=2.379cm]{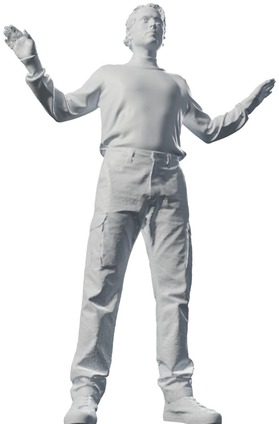}};
\node[anchor=south, inner sep=1pt] at (6.263,5.540) {37};
\node[cell] at (6.263,4.140) {\includegraphics[width=1.664cm,height=2.384cm]{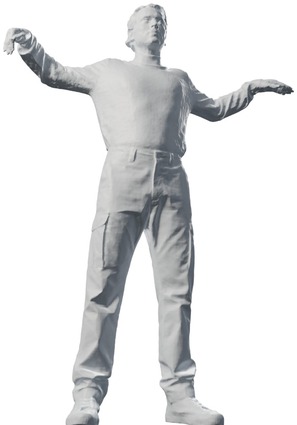}};
\node[val] at (7.080,2.780) {2.67};
\node[cell] at (6.263,1.380) {\includegraphics[width=1.664cm,height=2.384cm]{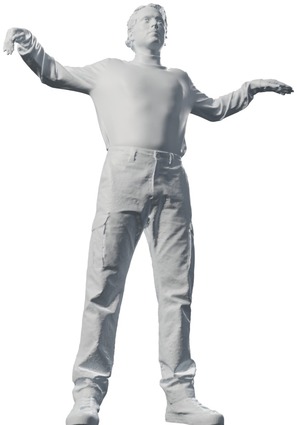}};
\node[anchor=south, inner sep=1pt] at (7.938,5.540) {49};
\node[cell] at (7.938,4.140) {\includegraphics[width=1.157cm,height=2.386cm]{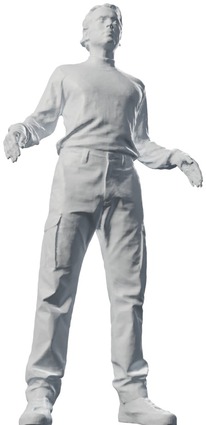}};
\node[val] at (8.755,2.780) {2.70};
\node[cell] at (7.938,1.380) {\includegraphics[width=1.157cm,height=2.386cm]{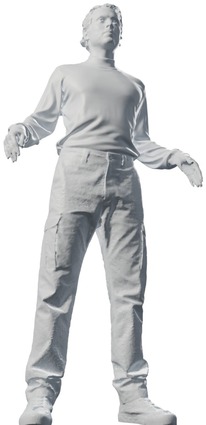}};
\node[rotate=90, anchor=south, inner sep=1pt, font=\scriptsize\bfseries] at (0.400,4.140) {Ours};
\node[rotate=90, anchor=south, inner sep=1pt] at (0.400,1.380) {GT};
\draw[black!15, line width=0.3pt] (0.400,2.760) -- (8.775,2.760);
\end{scope}
\begin{scope}[shift={(0,-12.400)}]
\node[anchor=south west, inner sep=0, font=\footnotesize] at (9.025,5.850) {actor8\_seq2\_325};
\node[anchor=south, inner sep=1pt] at (9.863,5.540) {2};
\node[cell] at (9.863,4.140) {\includegraphics[width=0.858cm,height=2.632cm]{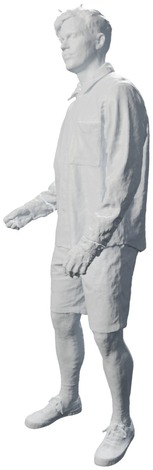}};
\node[val] at (10.680,2.780) {3.00};
\node[cell] at (9.863,1.380) {\includegraphics[width=0.858cm,height=2.632cm]{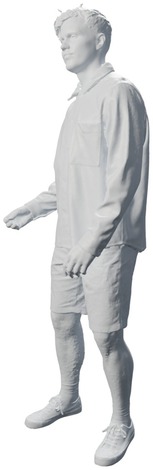}};
\node[anchor=south, inner sep=1pt] at (11.538,5.540) {13};
\node[cell] at (11.538,4.140) {\includegraphics[width=1.261cm,height=2.373cm]{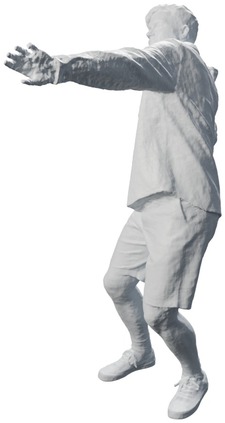}};
\node[val] at (12.355,2.780) {3.33};
\node[cell] at (11.538,1.380) {\includegraphics[width=1.261cm,height=2.373cm]{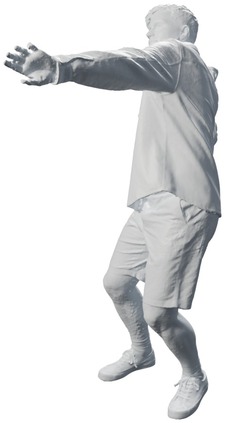}};
\node[anchor=south, inner sep=1pt] at (13.213,5.540) {25};
\node[cell] at (13.213,4.140) {\includegraphics[width=1.509cm,height=2.700cm]{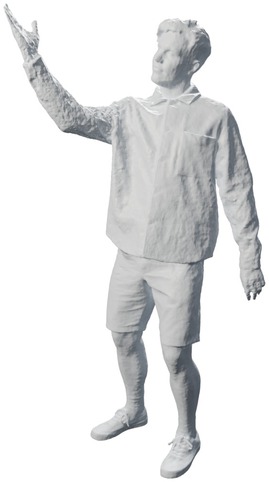}};
\node[val] at (14.030,2.780) {3.33};
\node[cell] at (13.213,1.380) {\includegraphics[width=1.509cm,height=2.700cm]{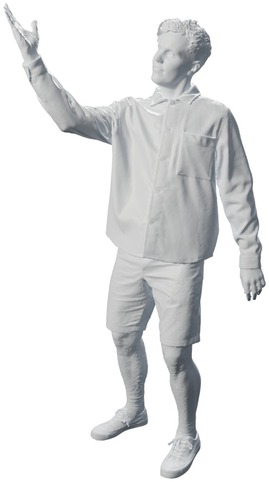}};
\node[anchor=south, inner sep=1pt] at (14.887,5.540) {37};
\node[cell] at (14.887,4.140) {\includegraphics[width=1.474cm,height=2.596cm]{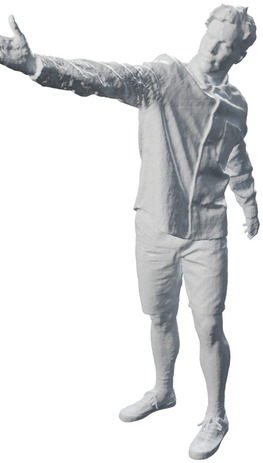}};
\node[val] at (15.705,2.780) {3.28};
\node[cell] at (14.887,1.380) {\includegraphics[width=1.474cm,height=2.596cm]{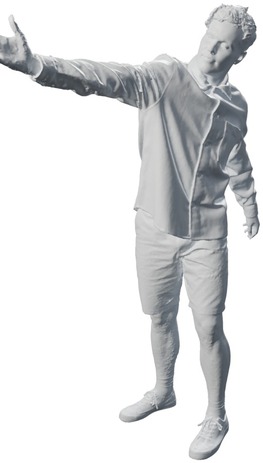}};
\node[anchor=south, inner sep=1pt] at (16.562,5.540) {49};
\node[cell] at (16.562,4.140) {\includegraphics[width=0.921cm,height=2.637cm]{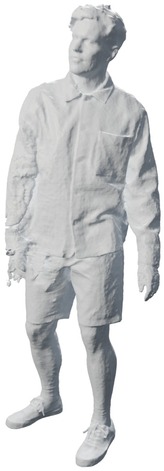}};
\node[val] at (17.380,2.780) {4.41};
\node[cell] at (16.562,1.380) {\includegraphics[width=0.921cm,height=2.637cm]{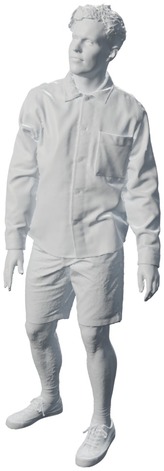}};
\draw[black!15, line width=0.3pt] (9.025,2.760) -- (17.400,2.760);
\end{scope}
\end{tikzpicture}
\caption{\textbf{Our reconstructions over the sequence}, on every scene of ActorsHQ (this figure), DeformingThings4D and TaoGS (next three figures), at frames 2, 13, 25, 37 and 49, with the ground truth below and the Chamfer distance of the frame in mm in the corner (\Cref{tab:geometry_avg}; frame~1 is not evaluated). Each frame is cropped to the reconstruction, with the same crop for ours and the ground truth; all frames of a scene are shown at the same scale. One dataset camera per scene: among those that see the whole body in every frame, the front-facing one with the largest mean silhouette (actor5: no front-facing camera sees the whole body in every frame; Cam005 is used, which leaves 4--5\% of the surface outside the image, e.g.\ the raised hands at frame~25).}
\label{fig:geo_gallery_1}
\end{figure*}

\begin{figure*}[p]
\centering
\definecolor{tzInk}{RGB}{24,28,40}
\begin{tikzpicture}[x=1cm, y=1cm, font=\scriptsize, text=tzInk, cell/.style={anchor=center, inner sep=0},
    val/.style={font=\tiny, anchor=south east, fill=white, fill opacity=0.8, text opacity=1, inner sep=1pt}]
\begin{scope}[shift={(0,0.000)}]
\node[anchor=south west, inner sep=0, font=\footnotesize] at (0.400,6.450) {abe\_CoverToStand};
\node[anchor=south, inner sep=1pt] at (1.238,6.140) {2};
\node[cell] at (1.238,4.590) {\includegraphics[width=1.109cm,height=2.127cm]{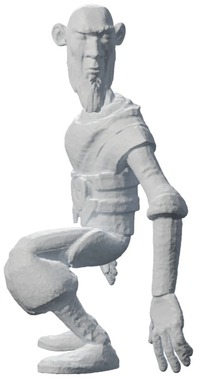}};
\node[val] at (2.055,3.080) {5.12};
\node[cell] at (1.238,1.530) {\includegraphics[width=1.109cm,height=2.127cm]{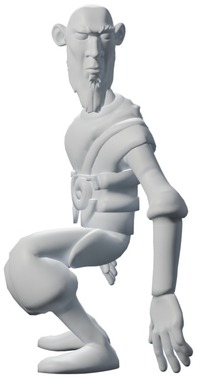}};
\node[anchor=south, inner sep=1pt] at (2.913,6.140) {13};
\node[cell] at (2.913,4.590) {\includegraphics[width=1.506cm,height=2.134cm]{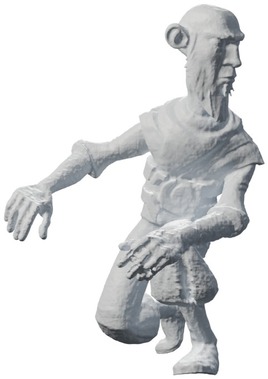}};
\node[val] at (3.730,3.080) {5.82};
\node[cell] at (2.913,1.530) {\includegraphics[width=1.506cm,height=2.134cm]{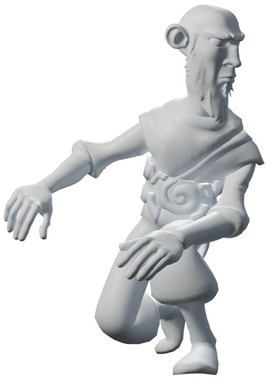}};
\node[anchor=south, inner sep=1pt] at (4.588,6.140) {25};
\node[cell] at (4.588,4.590) {\includegraphics[width=1.724cm,height=2.408cm]{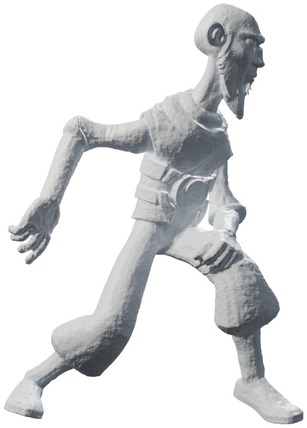}};
\node[val] at (5.405,3.080) {5.31};
\node[cell] at (4.588,1.530) {\includegraphics[width=1.724cm,height=2.408cm]{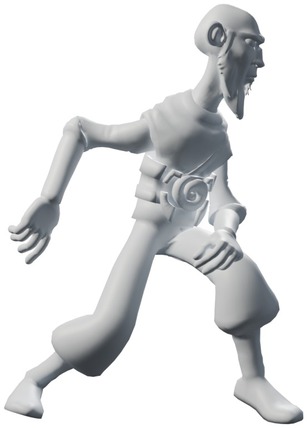}};
\node[anchor=south, inner sep=1pt] at (6.263,6.140) {37};
\node[cell] at (6.263,4.590) {\includegraphics[width=1.296cm,height=2.430cm]{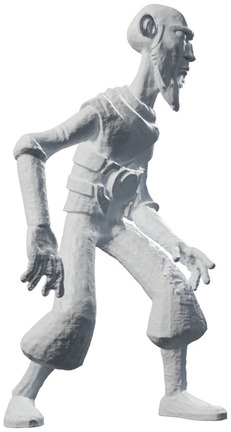}};
\node[val] at (7.080,3.080) {5.02};
\node[cell] at (6.263,1.530) {\includegraphics[width=1.296cm,height=2.430cm]{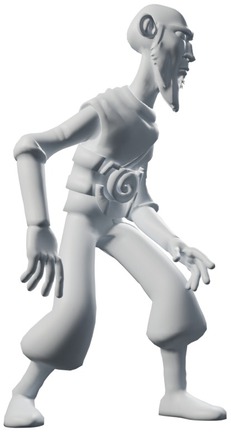}};
\node[anchor=south, inner sep=1pt] at (7.938,6.140) {49};
\node[cell] at (7.938,4.590) {\includegraphics[width=1.296cm,height=2.505cm]{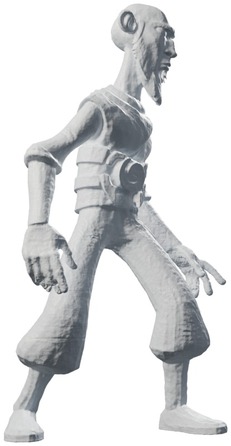}};
\node[val] at (8.755,3.080) {4.89};
\node[cell] at (7.938,1.530) {\includegraphics[width=1.296cm,height=2.505cm]{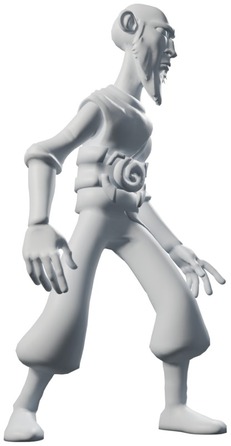}};
\node[rotate=90, anchor=south, inner sep=1pt, font=\scriptsize\bfseries] at (0.400,4.590) {Ours};
\node[rotate=90, anchor=south, inner sep=1pt] at (0.400,1.530) {GT};
\draw[black!15, line width=0.3pt] (0.400,3.060) -- (8.775,3.060);
\end{scope}
\begin{scope}[shift={(0,0.000)}]
\node[anchor=south west, inner sep=0, font=\footnotesize] at (9.025,6.450) {AJ\_BigStomachHit};
\node[anchor=south, inner sep=1pt] at (9.863,6.140) {2};
\node[cell] at (9.863,4.590) {\includegraphics[width=1.231cm,height=2.785cm]{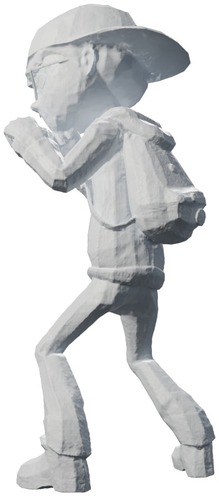}};
\node[val] at (10.680,3.080) {7.53};
\node[cell] at (9.863,1.530) {\includegraphics[width=1.231cm,height=2.785cm]{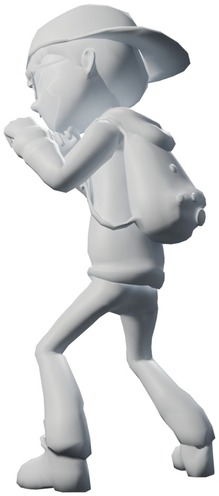}};
\node[anchor=south, inner sep=1pt] at (11.538,6.140) {13};
\node[cell] at (11.538,4.590) {\includegraphics[width=1.640cm,height=2.050cm]{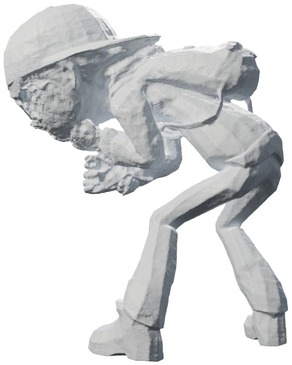}};
\node[val] at (12.355,3.080) {8.10};
\node[cell] at (11.538,1.530) {\includegraphics[width=1.640cm,height=2.050cm]{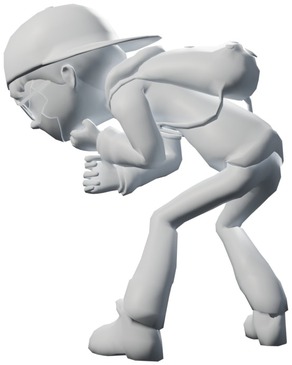}};
\node[anchor=south, inner sep=1pt] at (13.213,6.140) {25};
\node[cell] at (13.213,4.590) {\includegraphics[width=1.590cm,height=2.067cm]{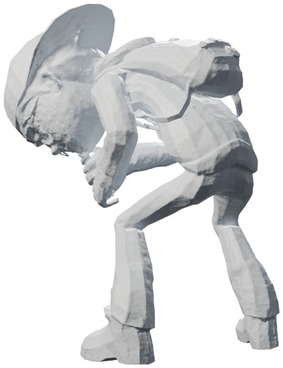}};
\node[val] at (14.030,3.080) {7.35};
\node[cell] at (13.213,1.530) {\includegraphics[width=1.590cm,height=2.067cm]{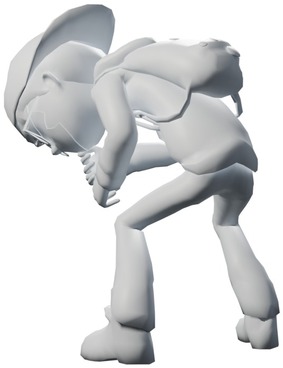}};
\node[anchor=south, inner sep=1pt] at (14.887,6.140) {37};
\node[cell] at (14.887,4.590) {\includegraphics[width=1.353cm,height=2.619cm]{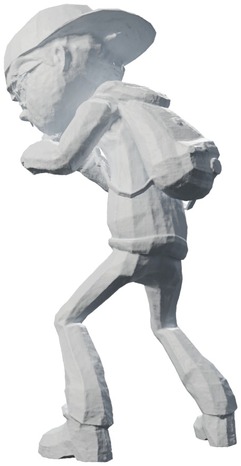}};
\node[val] at (15.705,3.080) {7.17};
\node[cell] at (14.887,1.530) {\includegraphics[width=1.353cm,height=2.619cm]{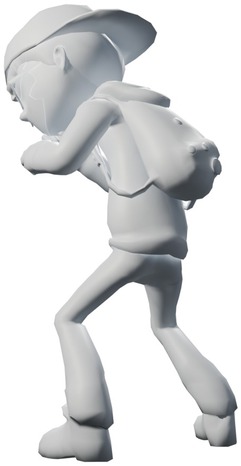}};
\node[anchor=south, inner sep=1pt] at (16.562,6.140) {49};
\node[cell] at (16.562,4.590) {\includegraphics[width=1.237cm,height=2.776cm]{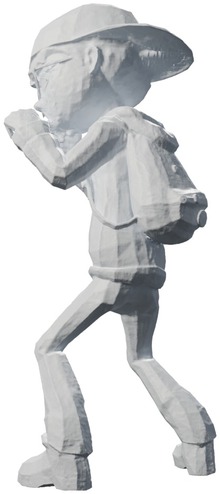}};
\node[val] at (17.380,3.080) {7.16};
\node[cell] at (16.562,1.530) {\includegraphics[width=1.237cm,height=2.776cm]{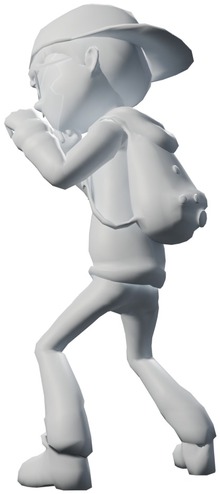}};
\draw[black!15, line width=0.3pt] (9.025,3.060) -- (17.400,3.060);
\end{scope}
\begin{scope}[shift={(0,-6.800)}]
\node[anchor=south west, inner sep=0, font=\footnotesize] at (0.400,6.450) {ninja\_KettlebellSwing};
\node[anchor=south, inner sep=1pt] at (1.238,6.140) {2};
\node[cell] at (1.238,4.590) {\includegraphics[width=1.179cm,height=2.264cm]{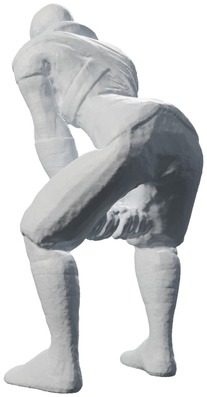}};
\node[val] at (2.055,3.080) {4.00};
\node[cell] at (1.238,1.530) {\includegraphics[width=1.179cm,height=2.264cm]{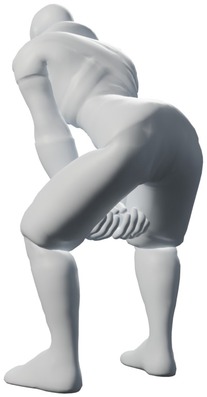}};
\node[anchor=south, inner sep=1pt] at (2.913,6.140) {13};
\node[cell] at (2.913,4.590) {\includegraphics[width=1.079cm,height=2.877cm]{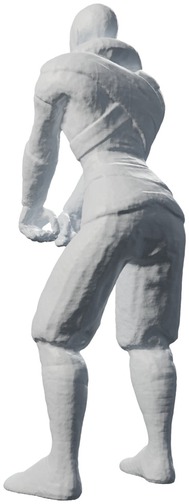}};
\node[val] at (3.730,3.080) {3.24};
\node[cell] at (2.913,1.530) {\includegraphics[width=1.079cm,height=2.877cm]{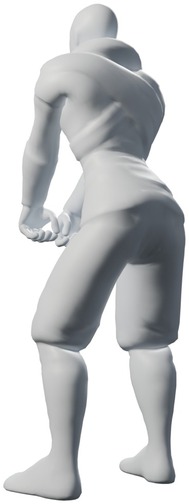}};
\node[anchor=south, inner sep=1pt] at (4.588,6.140) {25};
\node[cell] at (4.588,4.590) {\includegraphics[width=1.023cm,height=3.000cm]{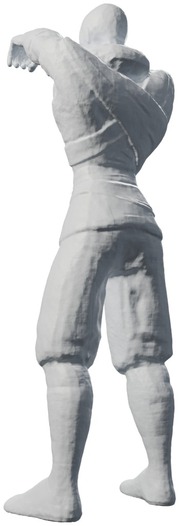}};
\node[val] at (5.405,3.080) {2.92};
\node[cell] at (4.588,1.530) {\includegraphics[width=1.023cm,height=3.000cm]{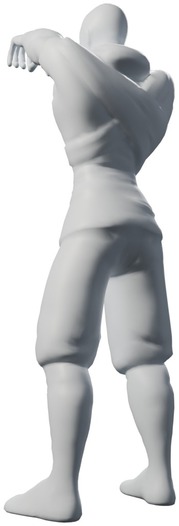}};
\node[anchor=south, inner sep=1pt] at (6.263,6.140) {37};
\node[cell] at (6.263,4.590) {\includegraphics[width=1.079cm,height=2.950cm]{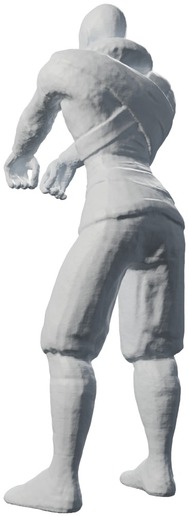}};
\node[val] at (7.080,3.080) {2.92};
\node[cell] at (6.263,1.530) {\includegraphics[width=1.079cm,height=2.950cm]{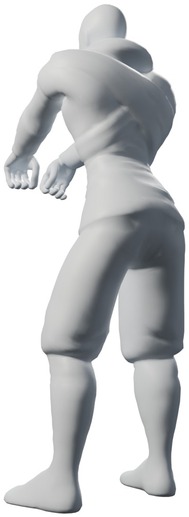}};
\node[anchor=south, inner sep=1pt] at (7.938,6.140) {49};
\node[cell] at (7.938,4.590) {\includegraphics[width=1.173cm,height=2.270cm]{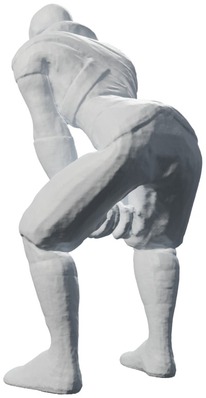}};
\node[val] at (8.755,3.080) {3.76};
\node[cell] at (7.938,1.530) {\includegraphics[width=1.173cm,height=2.270cm]{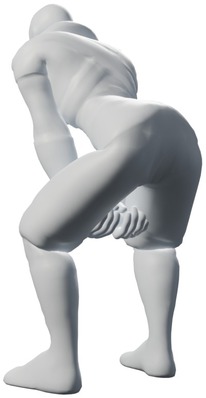}};
\node[rotate=90, anchor=south, inner sep=1pt, font=\scriptsize\bfseries] at (0.400,4.590) {Ours};
\node[rotate=90, anchor=south, inner sep=1pt] at (0.400,1.530) {GT};
\draw[black!15, line width=0.3pt] (0.400,3.060) -- (8.775,3.060);
\end{scope}
\begin{scope}[shift={(0,-6.800)}]
\node[anchor=south west, inner sep=0, font=\footnotesize] at (9.025,6.450) {vampire\_MmaKick};
\node[anchor=south, inner sep=1pt] at (9.863,6.140) {2};
\node[cell] at (9.863,4.590) {\includegraphics[width=1.197cm,height=2.261cm]{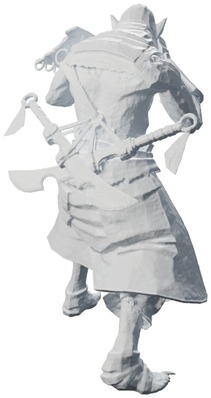}};
\node[val] at (10.680,3.080) {10.70};
\node[cell] at (9.863,1.530) {\includegraphics[width=1.197cm,height=2.261cm]{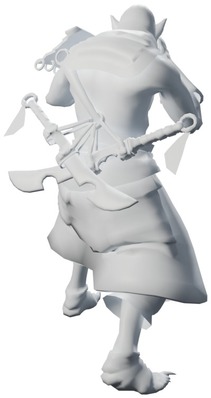}};
\node[anchor=south, inner sep=1pt] at (11.538,6.140) {13};
\node[cell] at (11.538,4.590) {\includegraphics[width=2.033cm,height=1.886cm]{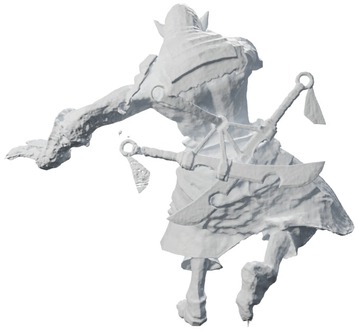}};
\node[val] at (12.355,3.080) {11.10};
\node[cell] at (11.538,1.530) {\includegraphics[width=2.033cm,height=1.886cm]{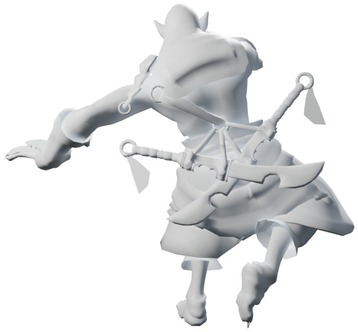}};
\node[anchor=south, inner sep=1pt] at (13.213,6.140) {25};
\node[cell] at (13.213,4.590) {\includegraphics[width=1.197cm,height=1.679cm]{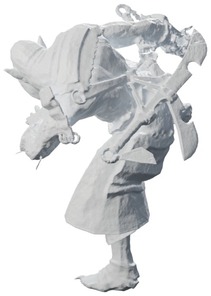}};
\node[val] at (14.030,3.080) {12.08};
\node[cell] at (13.213,1.530) {\includegraphics[width=1.197cm,height=1.679cm]{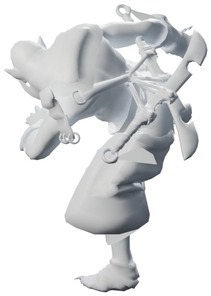}};
\node[anchor=south, inner sep=1pt] at (14.887,6.140) {37};
\node[cell] at (14.887,4.590) {\includegraphics[width=1.230cm,height=2.153cm]{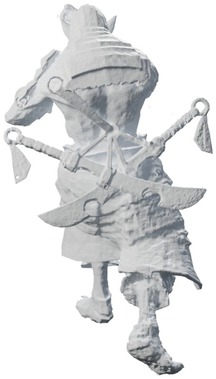}};
\node[val] at (15.705,3.080) {12.13};
\node[cell] at (14.887,1.530) {\includegraphics[width=1.230cm,height=2.153cm]{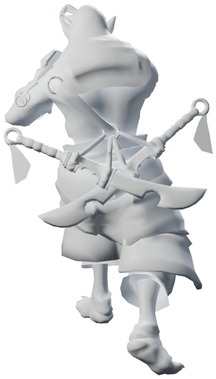}};
\node[anchor=south, inner sep=1pt] at (16.562,6.140) {49};
\node[cell] at (16.562,4.590) {\includegraphics[width=1.203cm,height=2.264cm]{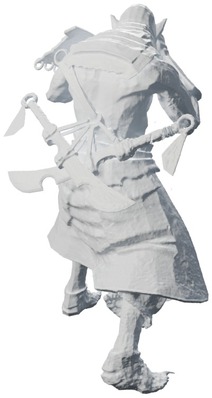}};
\node[val] at (17.380,3.080) {11.77};
\node[cell] at (16.562,1.530) {\includegraphics[width=1.203cm,height=2.264cm]{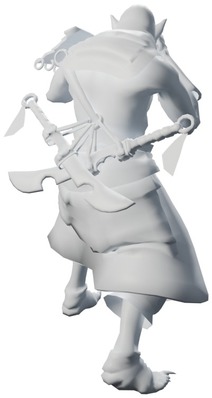}};
\draw[black!15, line width=0.3pt] (9.025,3.060) -- (17.400,3.060);
\end{scope}
\begin{scope}[shift={(0,-13.600)}]
\node[anchor=south west, inner sep=0, font=\footnotesize] at (0.400,6.450) {peasantgirl\_Martelo2};
\node[anchor=south, inner sep=1pt] at (1.238,6.140) {2};
\node[cell] at (1.238,4.590) {\includegraphics[width=1.481cm,height=1.864cm]{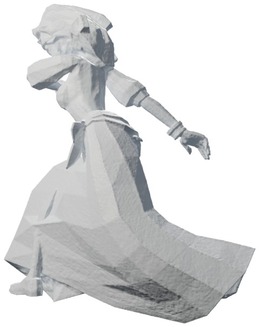}};
\node[val] at (2.055,3.080) {6.93};
\node[cell] at (1.238,1.530) {\includegraphics[width=1.481cm,height=1.864cm]{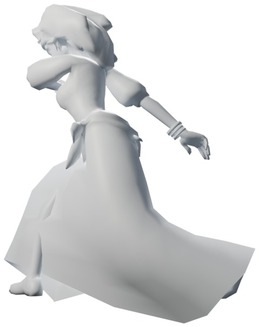}};
\node[anchor=south, inner sep=1pt] at (2.913,6.140) {13};
\node[cell] at (2.913,4.590) {\includegraphics[width=1.539cm,height=1.873cm]{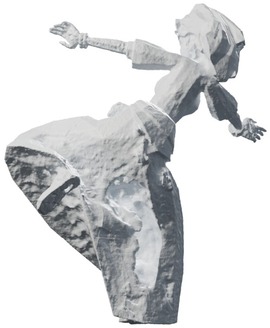}};
\node[val] at (3.730,3.080) {13.09};
\node[cell] at (2.913,1.530) {\includegraphics[width=1.539cm,height=1.873cm]{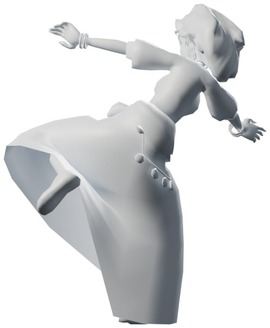}};
\node[anchor=south, inner sep=1pt] at (4.588,6.140) {25};
\node[cell] at (4.588,4.590) {\includegraphics[width=1.691cm,height=1.703cm]{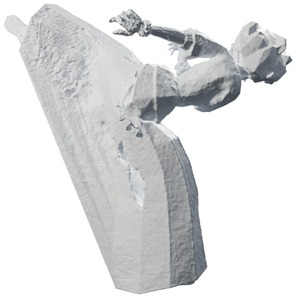}};
\node[val] at (5.405,3.080) {12.31};
\node[cell] at (4.588,1.530) {\includegraphics[width=1.691cm,height=1.703cm]{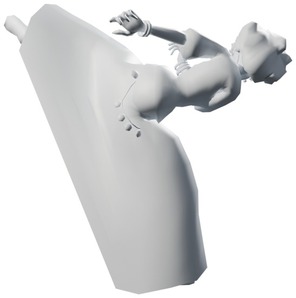}};
\node[anchor=south, inner sep=1pt] at (6.263,6.140) {37};
\node[cell] at (6.263,4.590) {\includegraphics[width=1.533cm,height=1.870cm]{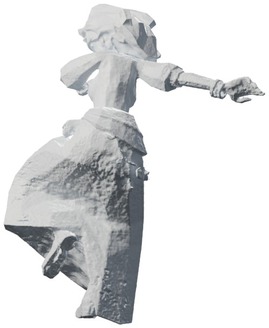}};
\node[val] at (7.080,3.080) {9.51};
\node[cell] at (6.263,1.530) {\includegraphics[width=1.533cm,height=1.870cm]{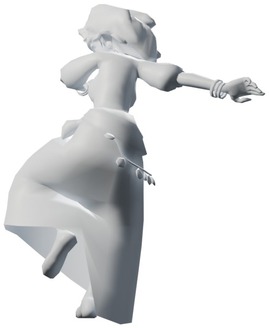}};
\node[anchor=south, inner sep=1pt] at (7.938,6.140) {49};
\node[cell] at (7.938,4.590) {\includegraphics[width=1.552cm,height=1.861cm]{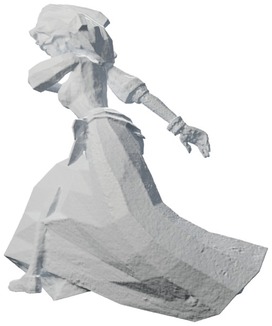}};
\node[val] at (8.755,3.080) {7.02};
\node[cell] at (7.938,1.530) {\includegraphics[width=1.552cm,height=1.861cm]{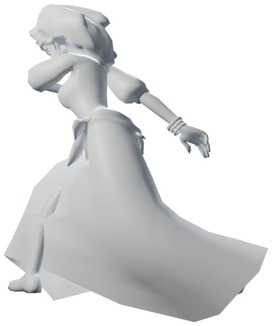}};
\node[rotate=90, anchor=south, inner sep=1pt, font=\scriptsize\bfseries] at (0.400,4.590) {Ours};
\node[rotate=90, anchor=south, inner sep=1pt] at (0.400,1.530) {GT};
\draw[black!15, line width=0.3pt] (0.400,3.060) -- (8.775,3.060);
\end{scope}
\begin{scope}[shift={(0,-13.600)}]
\node[anchor=south west, inner sep=0, font=\footnotesize] at (9.025,6.450) {dragonOF2\_act11};
\node[anchor=south, inner sep=1pt] at (9.863,6.140) {2};
\node[cell] at (9.863,4.590) {\includegraphics[width=1.027cm,height=1.666cm]{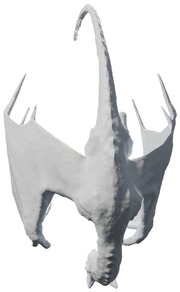}};
\node[val] at (10.680,3.080) {22.19};
\node[cell] at (9.863,1.530) {\includegraphics[width=1.027cm,height=1.666cm]{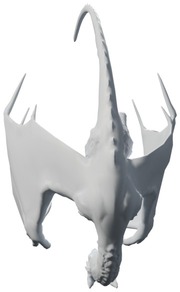}};
\node[anchor=south, inner sep=1pt] at (11.538,6.140) {13};
\node[cell] at (11.538,4.590) {\includegraphics[width=0.973cm,height=2.602cm]{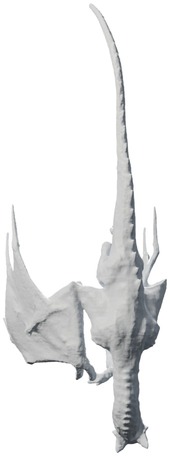}};
\node[val] at (12.355,3.080) {21.67};
\node[cell] at (11.538,1.530) {\includegraphics[width=0.973cm,height=2.602cm]{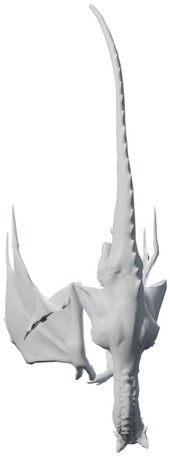}};
\node[anchor=south, inner sep=1pt] at (13.213,6.140) {25};
\node[cell] at (13.213,4.590) {\includegraphics[width=1.313cm,height=2.739cm]{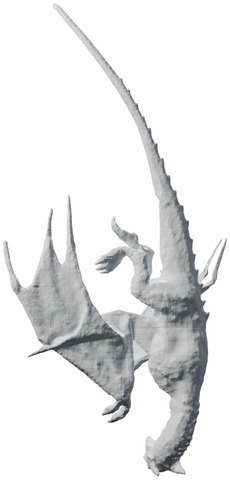}};
\node[val] at (14.030,3.080) {22.55};
\node[cell] at (13.213,1.530) {\includegraphics[width=1.313cm,height=2.739cm]{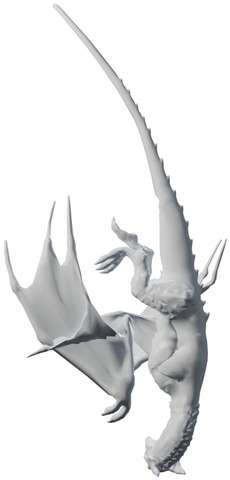}};
\node[anchor=south, inner sep=1pt] at (14.887,6.140) {37};
\node[cell] at (14.887,4.590) {\includegraphics[width=1.615cm,height=2.751cm]{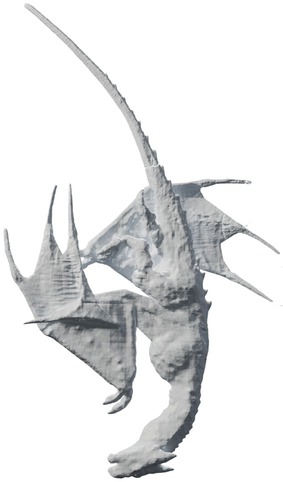}};
\node[val] at (15.705,3.080) {31.58};
\node[cell] at (14.887,1.530) {\includegraphics[width=1.615cm,height=2.751cm]{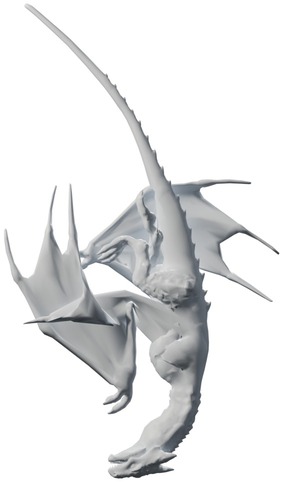}};
\node[anchor=south, inner sep=1pt] at (16.562,6.140) {49};
\node[cell] at (16.562,4.590) {\includegraphics[width=1.615cm,height=2.745cm]{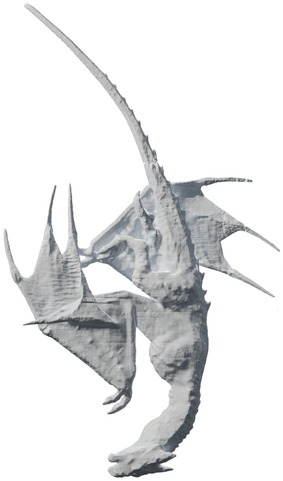}};
\node[val] at (17.380,3.080) {28.36};
\node[cell] at (16.562,1.530) {\includegraphics[width=1.615cm,height=2.745cm]{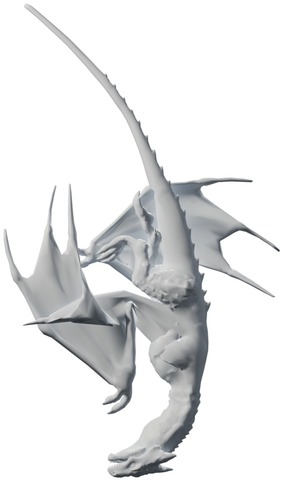}};
\draw[black!15, line width=0.3pt] (9.025,3.060) -- (17.400,3.060);
\end{scope}
\end{tikzpicture}
\caption{\textbf{Our reconstructions over the sequence} (continued), DeformingThings4D.}
\label{fig:geo_gallery_2}
\end{figure*}

\begin{figure*}[p]
\centering
\definecolor{tzInk}{RGB}{24,28,40}
\begin{tikzpicture}[x=1cm, y=1cm, font=\scriptsize, text=tzInk, cell/.style={anchor=center, inner sep=0},
    val/.style={font=\tiny, anchor=south east, fill=white, fill opacity=0.8, text opacity=1, inner sep=1pt}]
\begin{scope}[shift={(0,0.000)}]
\node[anchor=south west, inner sep=0, font=\footnotesize] at (0.400,4.750) {grizzRSS\_Attack5};
\node[anchor=south, inner sep=1pt] at (2.100,4.440) {2};
\node[cell] at (2.100,3.315) {\includegraphics[width=2.756cm,height=1.817cm]{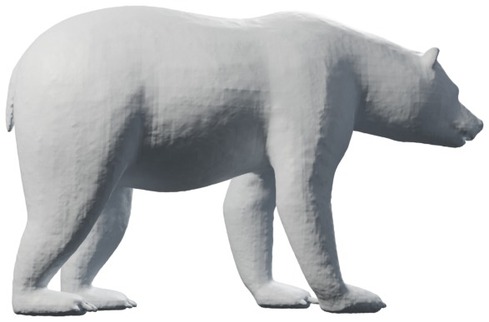}};
\node[val] at (3.780,2.230) {6.10};
\node[cell] at (2.100,1.105) {\includegraphics[width=2.756cm,height=1.817cm]{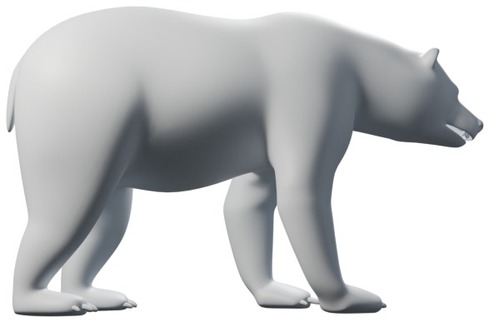}};
\node[anchor=south, inner sep=1pt] at (5.500,4.440) {13};
\node[cell] at (5.500,3.315) {\includegraphics[width=2.727cm,height=2.150cm]{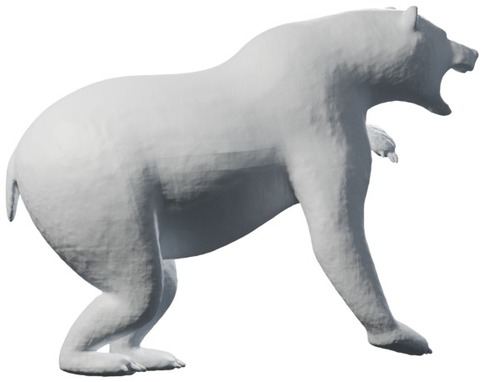}};
\node[val] at (7.180,2.230) {5.85};
\node[cell] at (5.500,1.105) {\includegraphics[width=2.727cm,height=2.150cm]{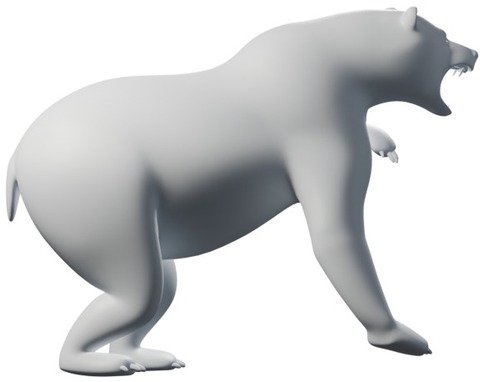}};
\node[anchor=south, inner sep=1pt] at (8.900,4.440) {25};
\node[cell] at (8.900,3.315) {\includegraphics[width=3.474cm,height=1.878cm]{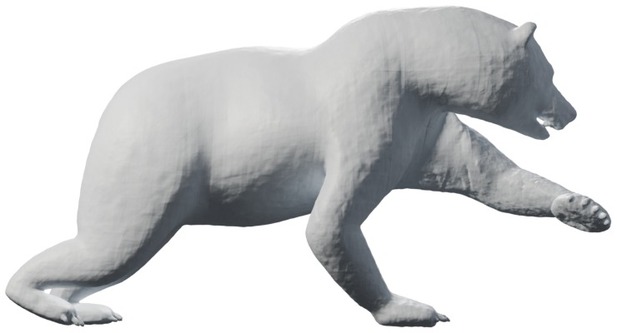}};
\node[val] at (10.580,2.230) {6.14};
\node[cell] at (8.900,1.105) {\includegraphics[width=3.474cm,height=1.878cm]{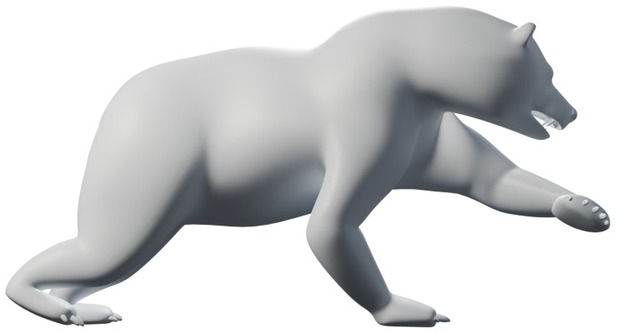}};
\node[anchor=south, inner sep=1pt] at (12.300,4.440) {37};
\node[cell] at (12.300,3.315) {\includegraphics[width=2.859cm,height=1.760cm]{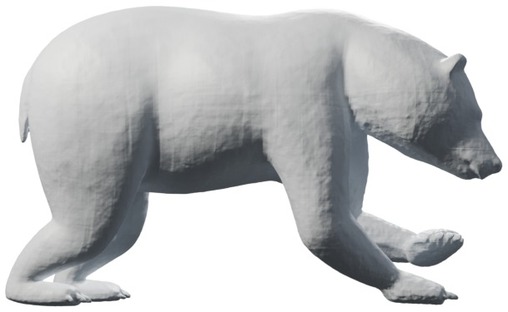}};
\node[val] at (13.980,2.230) {6.19};
\node[cell] at (12.300,1.105) {\includegraphics[width=2.859cm,height=1.760cm]{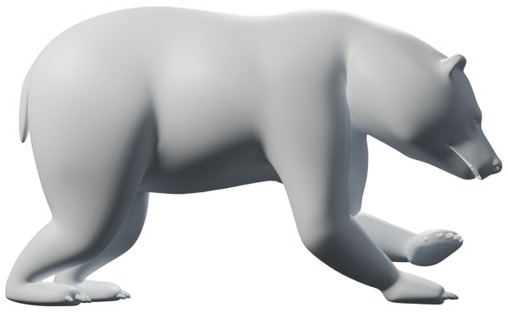}};
\node[anchor=south, inner sep=1pt] at (15.700,4.440) {49};
\node[cell] at (15.700,3.315) {\includegraphics[width=2.770cm,height=1.812cm]{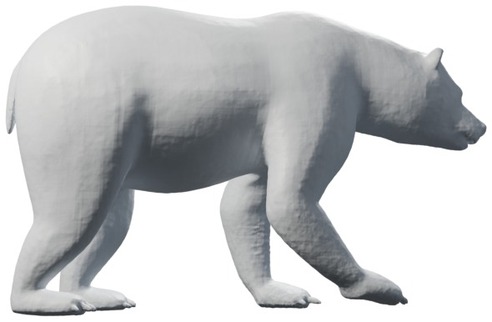}};
\node[val] at (17.380,2.230) {6.10};
\node[cell] at (15.700,1.105) {\includegraphics[width=2.770cm,height=1.812cm]{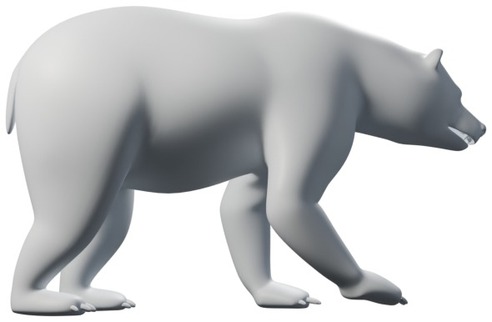}};
\node[rotate=90, anchor=south, inner sep=1pt, font=\scriptsize\bfseries] at (0.400,3.315) {Ours};
\node[rotate=90, anchor=south, inner sep=1pt] at (0.400,1.105) {GT};
\draw[black!15, line width=0.3pt] (0.400,2.210) -- (17.400,2.210);
\end{scope}
\begin{scope}[shift={(0,-5.100)}]
\node[anchor=south west, inner sep=0, font=\footnotesize] at (0.400,4.750) {rabbitFKP\_Damaged2};
\node[anchor=south, inner sep=1pt] at (2.100,4.440) {2};
\node[cell] at (2.100,3.315) {\includegraphics[width=1.844cm,height=2.046cm]{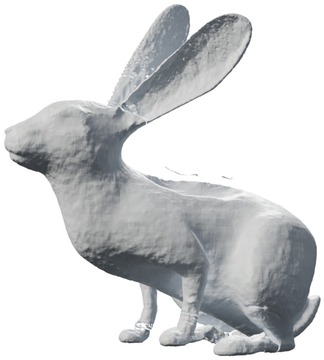}};
\node[val] at (3.780,2.230) {1.78};
\node[cell] at (2.100,1.105) {\includegraphics[width=1.844cm,height=2.046cm]{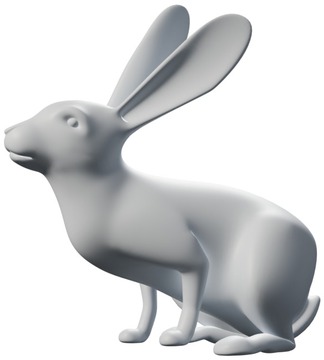}};
\node[anchor=south, inner sep=1pt] at (5.500,4.440) {13};
\node[cell] at (5.500,3.315) {\includegraphics[width=1.405cm,height=1.642cm]{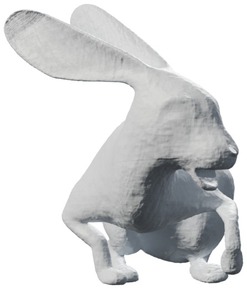}};
\node[val] at (7.180,2.230) {1.53};
\node[cell] at (5.500,1.105) {\includegraphics[width=1.405cm,height=1.642cm]{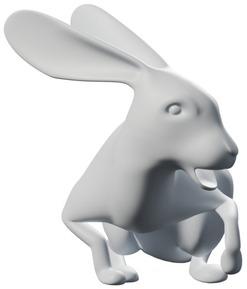}};
\node[anchor=south, inner sep=1pt] at (8.900,4.440) {25};
\node[cell] at (8.900,3.315) {\includegraphics[width=1.831cm,height=1.395cm]{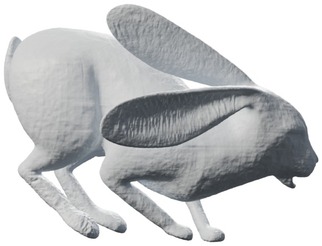}};
\node[val] at (10.580,2.230) {1.29};
\node[cell] at (8.900,1.105) {\includegraphics[width=1.831cm,height=1.395cm]{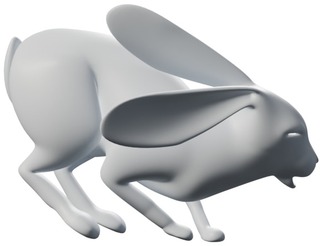}};
\node[anchor=south, inner sep=1pt] at (12.300,4.440) {37};
\node[cell] at (12.300,3.315) {\includegraphics[width=1.869cm,height=1.519cm]{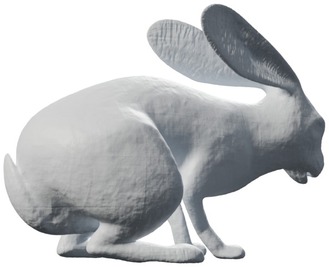}};
\node[val] at (13.980,2.230) {1.25};
\node[cell] at (12.300,1.105) {\includegraphics[width=1.869cm,height=1.519cm]{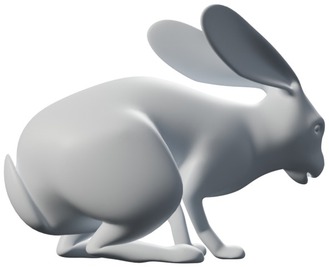}};
\node[anchor=south, inner sep=1pt] at (15.700,4.440) {49};
\node[cell] at (15.700,3.315) {\includegraphics[width=1.904cm,height=2.150cm]{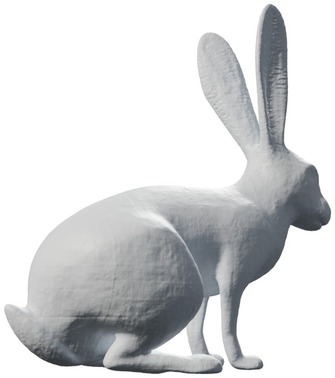}};
\node[val] at (17.380,2.230) {1.26};
\node[cell] at (15.700,1.105) {\includegraphics[width=1.904cm,height=2.150cm]{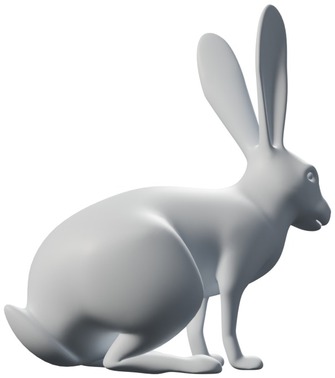}};
\node[rotate=90, anchor=south, inner sep=1pt, font=\scriptsize\bfseries] at (0.400,3.315) {Ours};
\node[rotate=90, anchor=south, inner sep=1pt] at (0.400,1.105) {GT};
\draw[black!15, line width=0.3pt] (0.400,2.210) -- (17.400,2.210);
\end{scope}
\begin{scope}[shift={(0,-10.200)}]
\node[anchor=south west, inner sep=0, font=\footnotesize] at (0.400,4.750) {canieLTT\_Damaged4};
\node[anchor=south, inner sep=1pt] at (2.100,4.440) {2};
\node[cell] at (2.100,3.315) {\includegraphics[width=3.447cm,height=1.626cm]{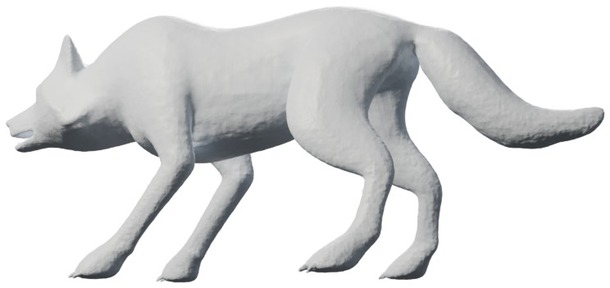}};
\node[val] at (3.780,2.230) {2.67};
\node[cell] at (2.100,1.105) {\includegraphics[width=3.447cm,height=1.626cm]{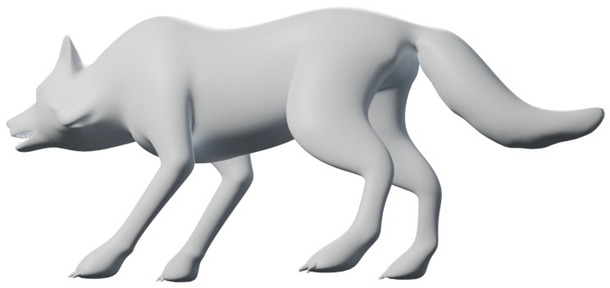}};
\node[anchor=south, inner sep=1pt] at (5.500,4.440) {13};
\node[cell] at (5.500,3.315) {\includegraphics[width=2.998cm,height=2.150cm]{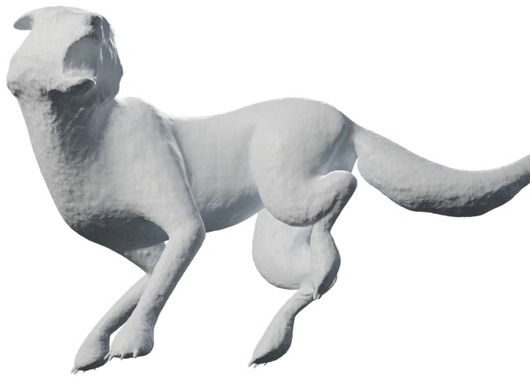}};
\node[val] at (7.180,2.230) {3.03};
\node[cell] at (5.500,1.105) {\includegraphics[width=2.998cm,height=2.150cm]{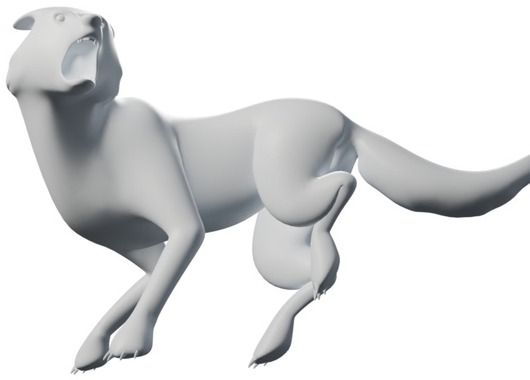}};
\node[anchor=south, inner sep=1pt] at (8.900,4.440) {25};
\node[cell] at (8.900,3.315) {\includegraphics[width=2.048cm,height=1.872cm]{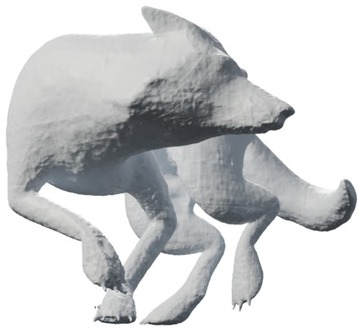}};
\node[val] at (10.580,2.230) {3.16};
\node[cell] at (8.900,1.105) {\includegraphics[width=2.048cm,height=1.872cm]{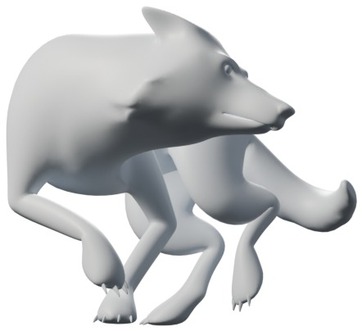}};
\node[anchor=south, inner sep=1pt] at (12.300,4.440) {37};
\node[cell] at (12.300,3.315) {\includegraphics[width=2.155cm,height=1.770cm]{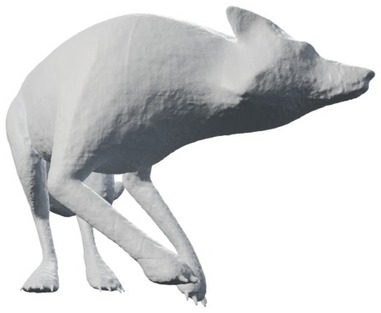}};
\node[val] at (13.980,2.230) {3.18};
\node[cell] at (12.300,1.105) {\includegraphics[width=2.155cm,height=1.770cm]{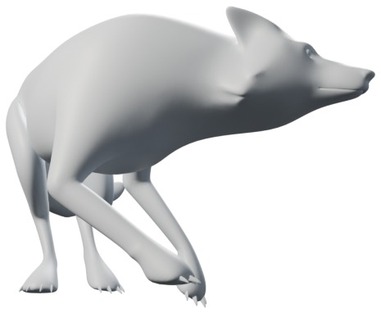}};
\node[anchor=south, inner sep=1pt] at (15.700,4.440) {49};
\node[cell] at (15.700,3.315) {\includegraphics[width=3.072cm,height=1.650cm]{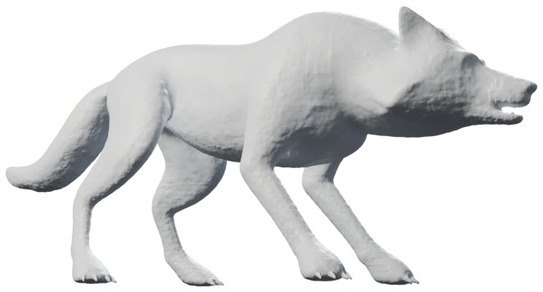}};
\node[val] at (17.380,2.230) {2.83};
\node[cell] at (15.700,1.105) {\includegraphics[width=3.072cm,height=1.650cm]{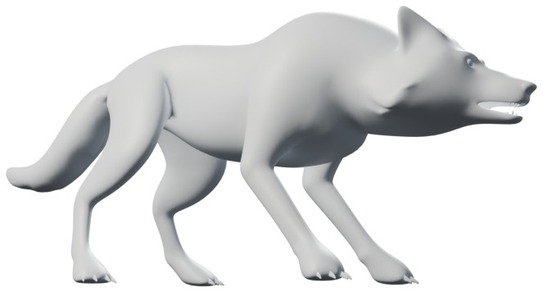}};
\node[rotate=90, anchor=south, inner sep=1pt, font=\scriptsize\bfseries] at (0.400,3.315) {Ours};
\node[rotate=90, anchor=south, inner sep=1pt] at (0.400,1.105) {GT};
\draw[black!15, line width=0.3pt] (0.400,2.210) -- (17.400,2.210);
\end{scope}
\begin{scope}[shift={(0,-15.300)}]
\node[anchor=south west, inner sep=0, font=\footnotesize] at (0.400,4.750) {foxWDFS\_Damaged1};
\node[anchor=south, inner sep=1pt] at (2.100,4.440) {2};
\node[cell] at (2.100,3.315) {\includegraphics[width=3.374cm,height=2.119cm]{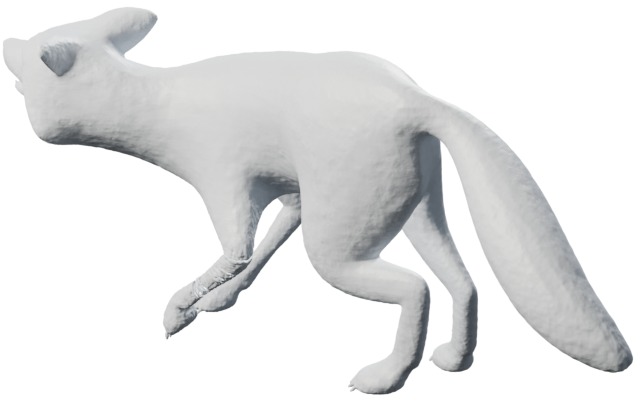}};
\node[val] at (3.780,2.230) {2.15};
\node[cell] at (2.100,1.105) {\includegraphics[width=3.374cm,height=2.119cm]{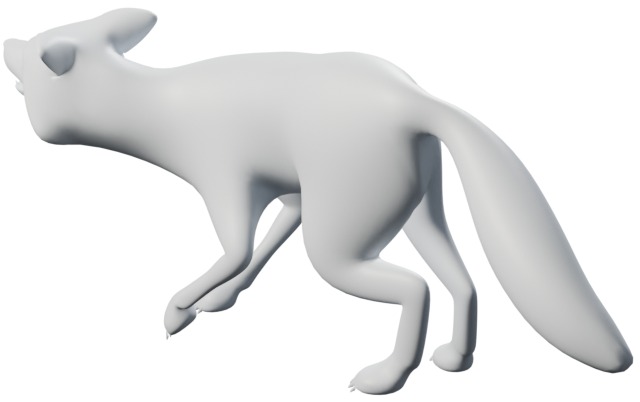}};
\node[anchor=south, inner sep=1pt] at (5.500,4.440) {13};
\node[cell] at (5.500,3.315) {\includegraphics[width=2.763cm,height=1.677cm]{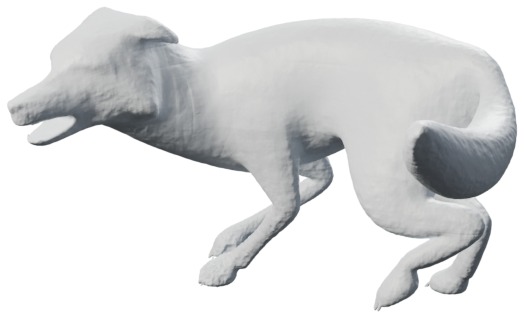}};
\node[val] at (7.180,2.230) {2.15};
\node[cell] at (5.500,1.105) {\includegraphics[width=2.763cm,height=1.677cm]{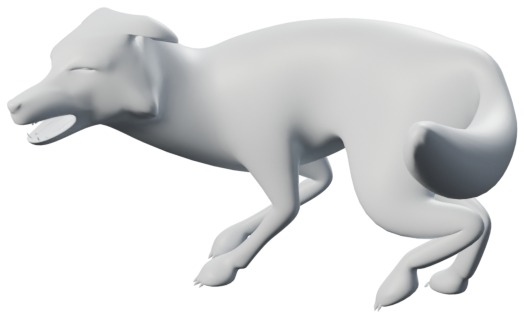}};
\node[anchor=south, inner sep=1pt] at (8.900,4.440) {25};
\node[cell] at (8.900,3.315) {\includegraphics[width=3.063cm,height=1.729cm]{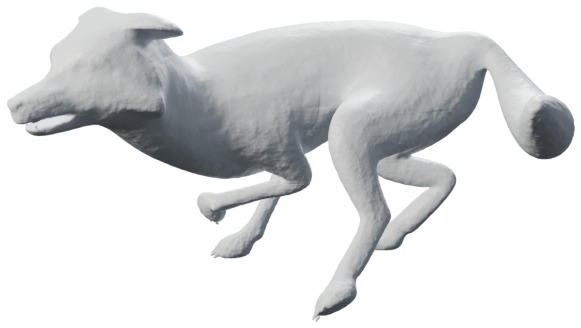}};
\node[val] at (10.580,2.230) {2.17};
\node[cell] at (8.900,1.105) {\includegraphics[width=3.063cm,height=1.729cm]{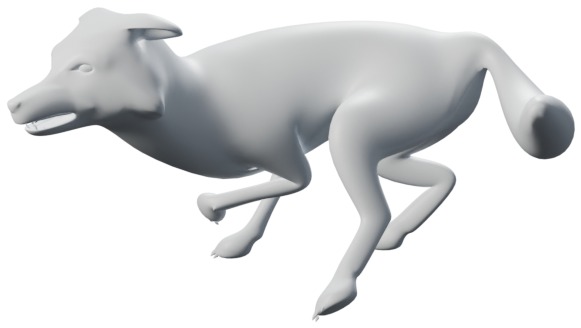}};
\node[anchor=south, inner sep=1pt] at (12.300,4.440) {37};
\node[cell] at (12.300,3.315) {\includegraphics[width=3.617cm,height=1.814cm]{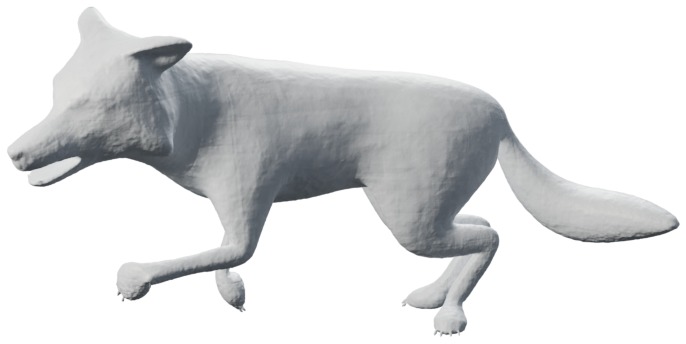}};
\node[val] at (13.980,2.230) {2.16};
\node[cell] at (12.300,1.105) {\includegraphics[width=3.617cm,height=1.814cm]{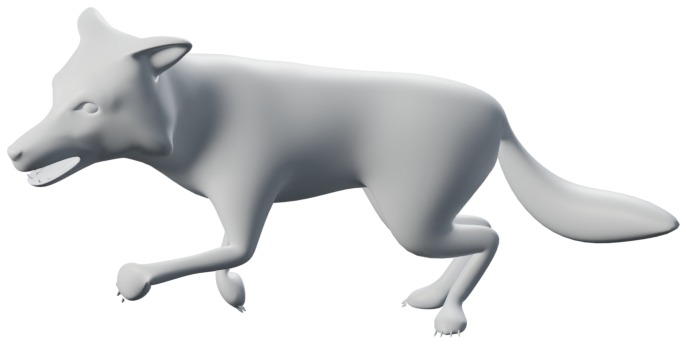}};
\node[anchor=south, inner sep=1pt] at (15.700,4.440) {49};
\node[cell] at (15.700,3.315) {\includegraphics[width=2.926cm,height=1.988cm]{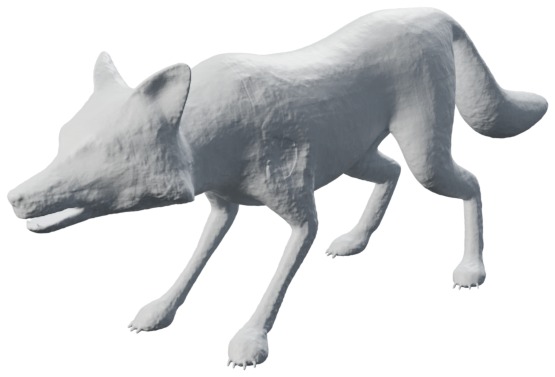}};
\node[val] at (17.380,2.230) {2.20};
\node[cell] at (15.700,1.105) {\includegraphics[width=2.926cm,height=1.988cm]{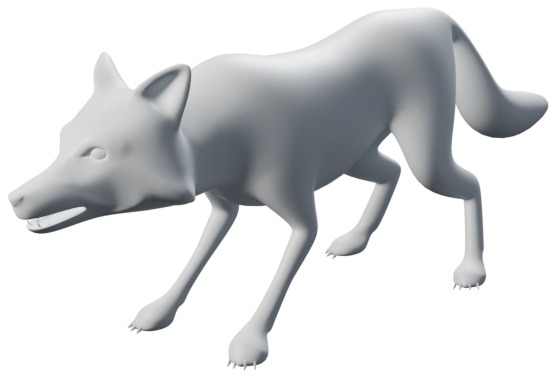}};
\node[rotate=90, anchor=south, inner sep=1pt, font=\scriptsize\bfseries] at (0.400,3.315) {Ours};
\node[rotate=90, anchor=south, inner sep=1pt] at (0.400,1.105) {GT};
\draw[black!15, line width=0.3pt] (0.400,2.210) -- (17.400,2.210);
\end{scope}
\end{tikzpicture}
\caption{\textbf{Our reconstructions over the sequence} (continued), DeformingThings4D.}
\label{fig:geo_gallery_3}
\end{figure*}

\begin{figure*}[p]
\centering
\definecolor{tzInk}{RGB}{24,28,40}
\begin{tikzpicture}[x=1cm, y=1cm, font=\scriptsize, text=tzInk]
\begin{scope}[shift={(0,-3.495)}]
\node[anchor=south west, inner sep=0, font=\footnotesize] at (0.400,3.825) {magic2};
\node[rotate=90, anchor=south, inner sep=1pt, font=\scriptsize\bfseries] at (0.400,1.747) {Ours};
\node[anchor=south west, inner sep=0] at (0.400,0) {\includegraphics[height=3.495cm]{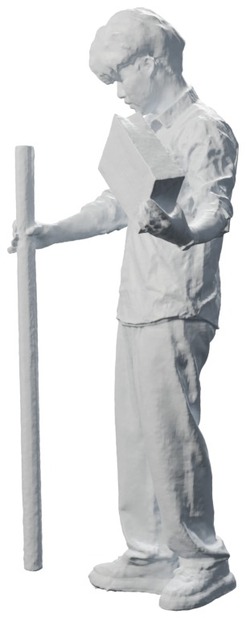}};
\node[anchor=south, inner sep=1pt] at (1.097,3.515) {2};
\node[anchor=south west, inner sep=0] at (1.914,0) {\includegraphics[height=3.495cm]{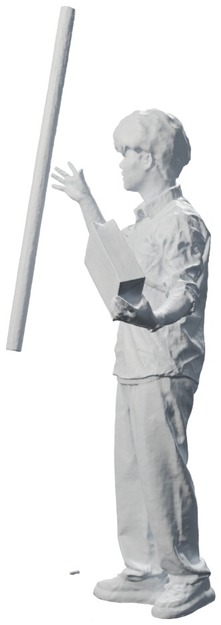}};
\node[anchor=south, inner sep=1pt] at (2.533,3.515) {13};
\node[anchor=south west, inner sep=0] at (3.271,0) {\includegraphics[height=3.495cm]{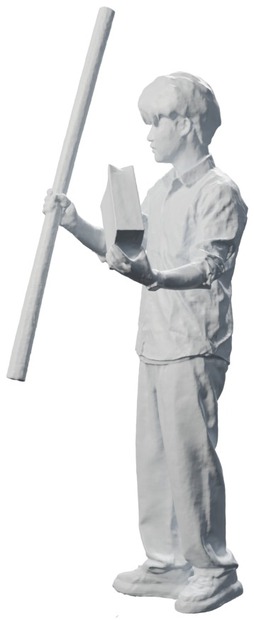}};
\node[anchor=south, inner sep=1pt] at (3.985,3.515) {25};
\node[anchor=south west, inner sep=0] at (4.819,0) {\includegraphics[height=3.495cm]{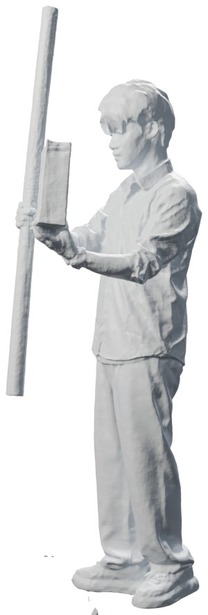}};
\node[anchor=south, inner sep=1pt] at (5.411,3.515) {37};
\node[anchor=south west, inner sep=0] at (6.123,0) {\includegraphics[height=3.495cm]{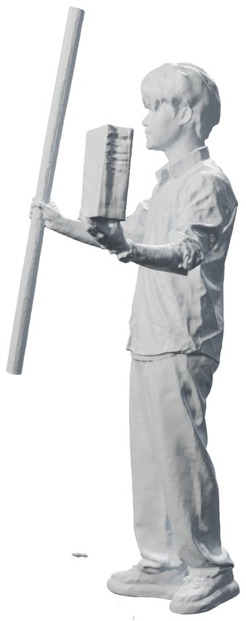}};
\node[anchor=south, inner sep=1pt] at (6.813,3.515) {49};
\node[anchor=south west, inner sep=0, font=\footnotesize] at (7.954,3.825) {magic3};
\node[anchor=south west, inner sep=0] at (7.954,0) {\includegraphics[height=3.495cm]{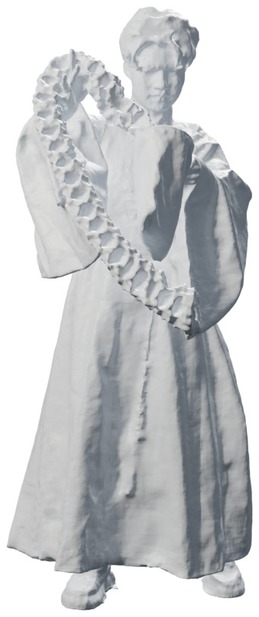}};
\node[anchor=south, inner sep=1pt] at (8.687,3.515) {2};
\node[anchor=south west, inner sep=0] at (9.540,0) {\includegraphics[height=3.495cm]{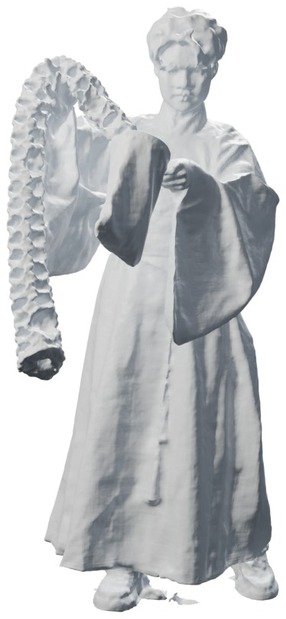}};
\node[anchor=south, inner sep=1pt] at (10.347,3.515) {13};
\node[anchor=south west, inner sep=0] at (11.274,0) {\includegraphics[height=3.495cm]{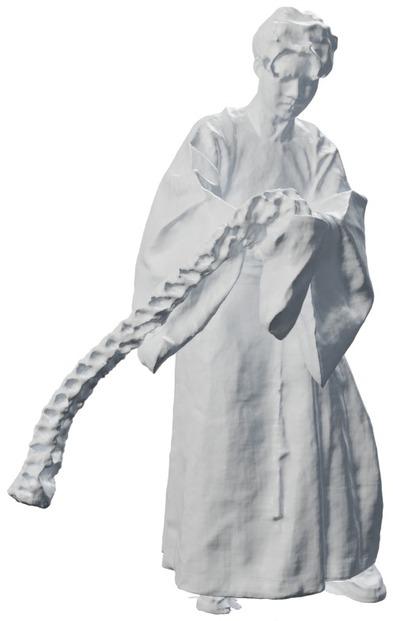}};
\node[anchor=south, inner sep=1pt] at (12.380,3.515) {25};
\node[anchor=south west, inner sep=0] at (13.607,0) {\includegraphics[height=3.495cm]{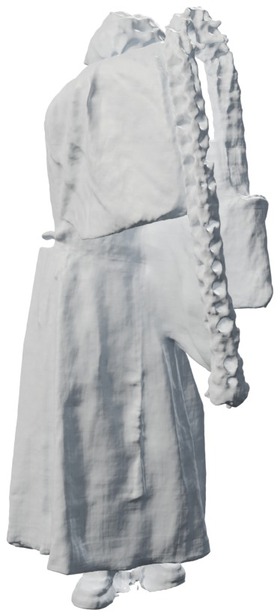}};
\node[anchor=south, inner sep=1pt] at (14.402,3.515) {37};
\node[anchor=south west, inner sep=0] at (15.317,0) {\includegraphics[height=3.495cm]{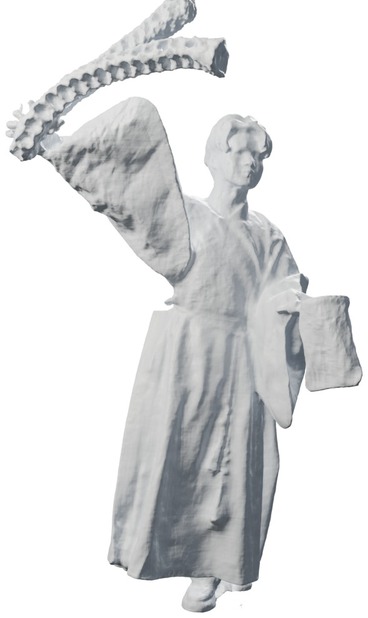}};
\node[anchor=south, inner sep=1pt] at (16.358,3.515) {49};
\end{scope}
\begin{scope}[shift={(0,-6.728)}]
\node[anchor=south west, inner sep=0, font=\footnotesize] at (0.400,2.883) {umbrella};
\node[rotate=90, anchor=south, inner sep=1pt, font=\scriptsize\bfseries] at (0.400,1.277) {Ours};
\node[anchor=south west, inner sep=0] at (0.400,0) {\includegraphics[height=2.553cm]{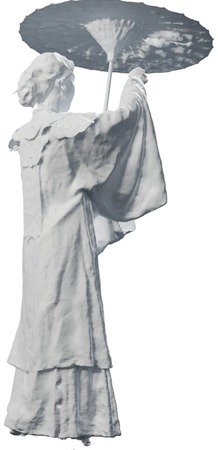}};
\node[anchor=south, inner sep=1pt] at (1.019,2.573) {2};
\node[anchor=south west, inner sep=0] at (1.758,0) {\includegraphics[height=2.553cm]{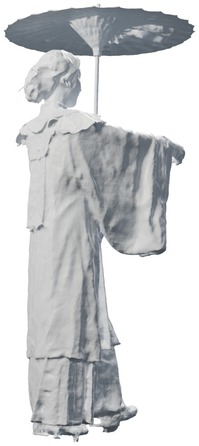}};
\node[anchor=south, inner sep=1pt] at (2.326,2.573) {13};
\node[anchor=south west, inner sep=0] at (3.015,0) {\includegraphics[height=2.553cm]{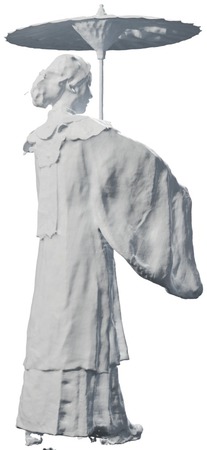}};
\node[anchor=south, inner sep=1pt] at (3.601,2.573) {25};
\node[anchor=south west, inner sep=0] at (4.307,0) {\includegraphics[height=2.553cm]{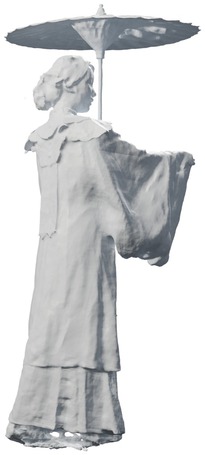}};
\node[anchor=south, inner sep=1pt] at (4.882,2.573) {37};
\node[anchor=south west, inner sep=0] at (5.577,0) {\includegraphics[height=2.553cm]{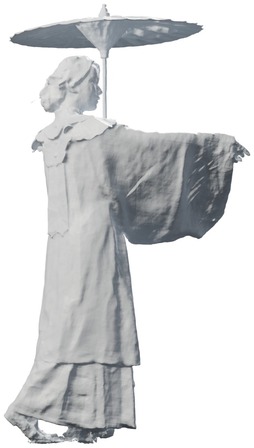}};
\node[anchor=south, inner sep=1pt] at (6.300,2.573) {48};
\node[anchor=south west, inner sep=0, font=\footnotesize] at (7.474,2.883) {drawsword};
\node[anchor=south west, inner sep=0] at (7.474,0) {\includegraphics[height=2.553cm]{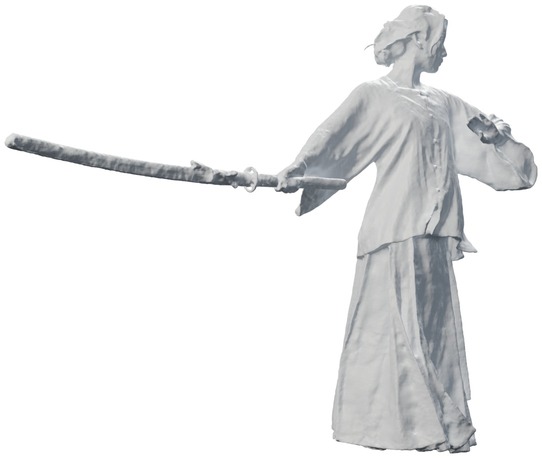}};
\node[anchor=south, inner sep=1pt] at (8.986,2.573) {2};
\node[anchor=south west, inner sep=0] at (10.619,0) {\includegraphics[height=2.553cm]{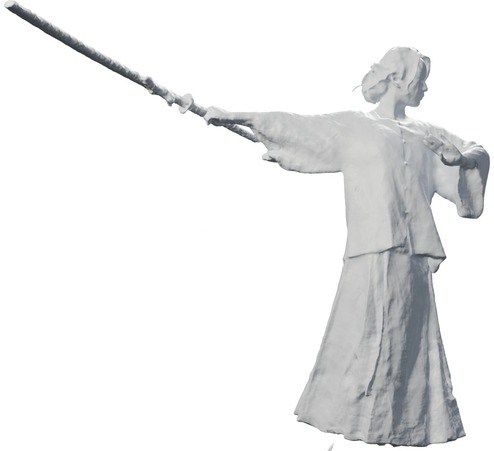}};
\node[anchor=south, inner sep=1pt] at (12.017,2.573) {13};
\node[anchor=south west, inner sep=0] at (13.536,0) {\includegraphics[height=2.553cm]{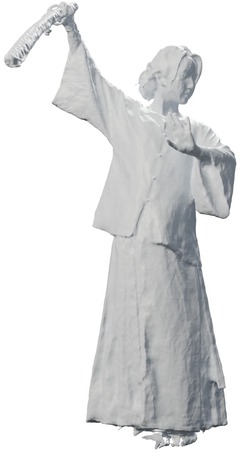}};
\node[anchor=south, inner sep=1pt] at (14.211,2.573) {25};
\node[anchor=south west, inner sep=0] at (15.007,0) {\includegraphics[height=2.553cm]{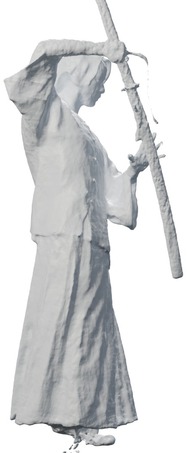}};
\node[anchor=south, inner sep=1pt] at (15.533,2.573) {37};
\node[anchor=south west, inner sep=0] at (16.180,0) {\includegraphics[height=2.553cm]{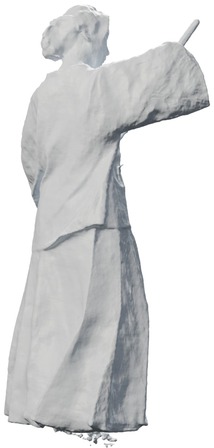}};
\node[anchor=south, inner sep=1pt] at (16.790,2.573) {49};
\end{scope}
\begin{scope}[shift={(0,-9.693)}]
\node[anchor=south west, inner sep=0, font=\footnotesize] at (0.400,2.615) {bag};
\node[rotate=90, anchor=south, inner sep=1pt, font=\scriptsize\bfseries] at (0.400,1.143) {Ours};
\node[anchor=south west, inner sep=0] at (0.400,0) {\includegraphics[height=2.285cm]{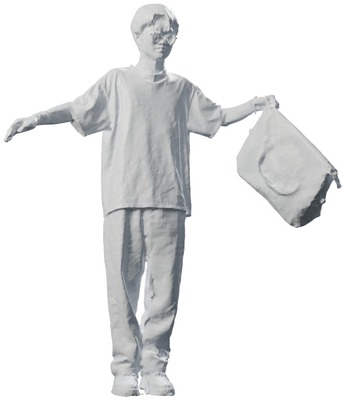}};
\node[anchor=south, inner sep=1pt] at (1.388,2.305) {2};
\node[anchor=south west, inner sep=0] at (2.496,0) {\includegraphics[height=2.285cm]{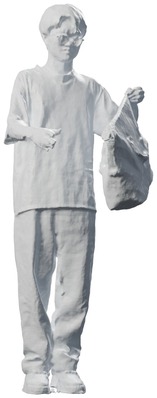}};
\node[anchor=south, inner sep=1pt] at (2.946,2.305) {13};
\node[anchor=south west, inner sep=0] at (3.517,0) {\includegraphics[height=2.285cm]{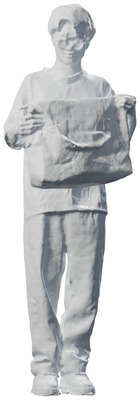}};
\node[anchor=south, inner sep=1pt] at (3.915,2.305) {25};
\node[anchor=south west, inner sep=0] at (4.434,0) {\includegraphics[height=2.285cm]{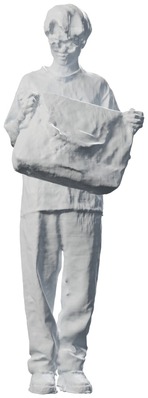}};
\node[anchor=south, inner sep=1pt] at (4.859,2.305) {37};
\node[anchor=south west, inner sep=0] at (5.403,0) {\includegraphics[height=2.285cm]{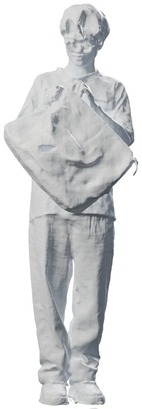}};
\node[anchor=south, inner sep=1pt] at (5.799,2.305) {49};
\node[anchor=south west, inner sep=0, font=\footnotesize] at (6.644,2.615) {changeclothes};
\node[anchor=south west, inner sep=0] at (6.644,0) {\includegraphics[height=2.285cm]{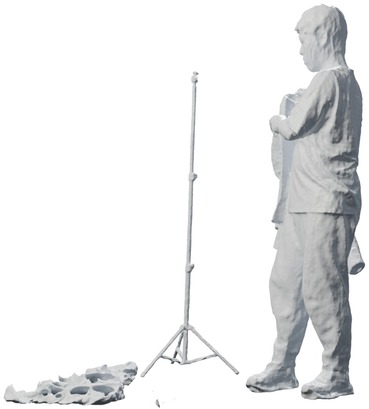}};
\node[anchor=south, inner sep=1pt] at (7.676,2.305) {2};
\node[anchor=south west, inner sep=0] at (8.828,0) {\includegraphics[height=2.285cm]{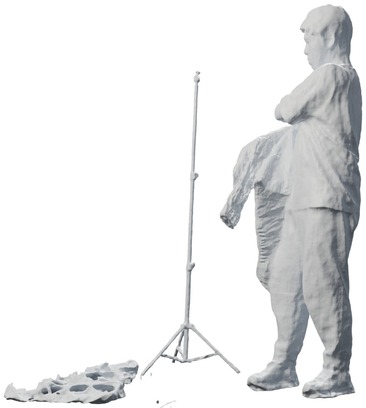}};
\node[anchor=south, inner sep=1pt] at (9.853,2.305) {13};
\node[anchor=south west, inner sep=0] at (10.997,0) {\includegraphics[height=2.285cm]{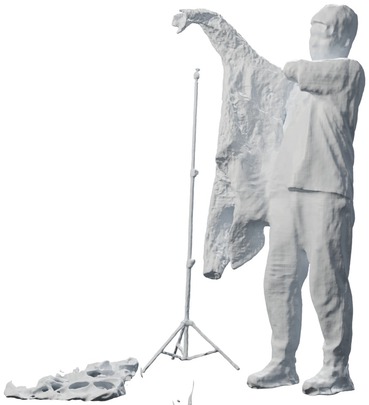}};
\node[anchor=south, inner sep=1pt] at (12.039,2.305) {25};
\node[anchor=south west, inner sep=0] at (13.201,0) {\includegraphics[height=2.285cm]{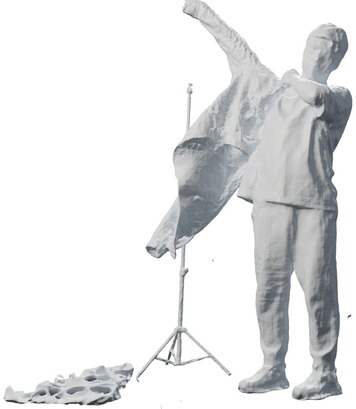}};
\node[anchor=south, inner sep=1pt] at (14.196,2.305) {37};
\node[anchor=south west, inner sep=0] at (15.311,0) {\includegraphics[height=2.285cm]{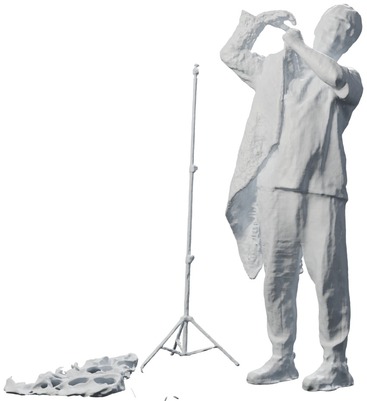}};
\node[anchor=south, inner sep=1pt] at (16.355,2.305) {49};
\end{scope}
\end{tikzpicture}
\caption{\textbf{Our reconstructions over the sequence} (continued), TaoGS. TaoGS provides no ground-truth geometry. Drawsword: small fragments far from the actor, caused by errors in the input foreground masks, are removed for display.}
\label{fig:taogs}
\end{figure*}

%% file: figures/geo_cmp_gallery.tex
\begin{figure*}[p]
\centering
\begin{tikzpicture}[x=1cm, y=1cm, font=\scriptsize, cell/.style={anchor=north west, inner sep=0},
    val/.style={font=\tiny, anchor=south east, fill=white, fill opacity=0.85, text opacity=1, inner sep=1pt}]
\draw[black!45, line width=0.4pt] (0.050,0.420) -- (3.302,0.420);
\node[anchor=south, inner sep=1pt, font=\scriptsize\itshape] at (1.676,0.420) {Static (per frame)};
\draw[black!45, line width=0.4pt] (3.562,0.420) -- (15.414,0.420);
\node[anchor=south, inner sep=1pt, font=\scriptsize\itshape] at (9.488,0.420) {Dynamic};
\draw[black!40, dashed, line width=0.5pt] (3.432,0.720) -- (3.432,-18.243);
\node[anchor=south, inner sep=1pt] at (0.828,0.000) {2DGS};
\node[anchor=south, inner sep=1pt] at (2.524,0.000) {PGSR};
\node[anchor=south, inner sep=1pt] at (4.340,0.000) {NeuS2};
\node[anchor=south, inner sep=1pt] at (6.036,0.000) {AT-GS};
\node[anchor=south, inner sep=1pt] at (7.732,0.000) {SC-GS};
\node[anchor=south, inner sep=1pt] at (9.428,0.000) {DG-Mesh};
\node[anchor=south, inner sep=1pt] at (11.124,0.000) {GauSTAR};
\node[anchor=south, inner sep=1pt] at (12.820,0.000) {MoAngelo};
\node[anchor=south, inner sep=1pt] at (14.636,0.000) {\color[HTML]{0E80E5}\textbf{Ours}};
\node[anchor=south, inner sep=1pt] at (16.452,0.000) {GT};
\node[anchor=north west, inner sep=1pt, fill=white, font=\scriptsize\itshape] at (0.000,-0.140) {DeformingThings4D rabbit};
\node[cell] at (0.000,-0.480) {\includegraphics[width=1.656cm]{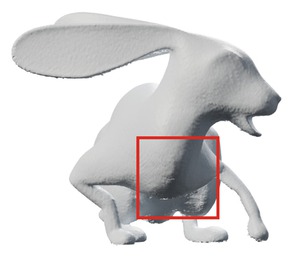}};
\node[val] at (1.656,-1.932) {1.66};
\node[cell] at (0.000,-1.982) {\includegraphics[width=1.656cm]{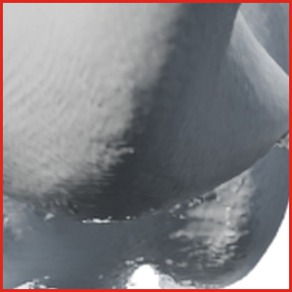}};
\node[cell] at (1.696,-0.480) {\includegraphics[width=1.656cm]{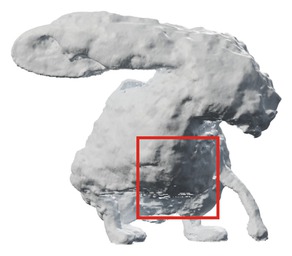}};
\node[val] at (3.352,-1.932) {8.78};
\node[cell] at (1.696,-1.982) {\includegraphics[width=1.656cm]{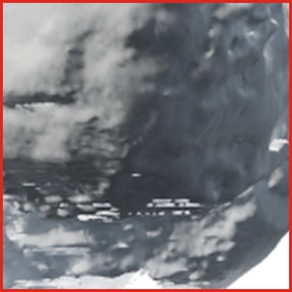}};
\node[cell] at (3.512,-0.480) {\includegraphics[width=1.656cm]{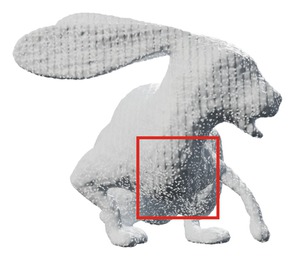}};
\node[val] at (5.168,-1.932) {2.38};
\node[cell] at (3.512,-1.982) {\includegraphics[width=1.656cm]{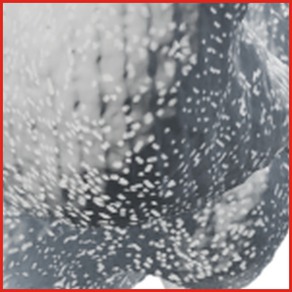}};
\node[cell] at (5.208,-0.480) {\includegraphics[width=1.656cm]{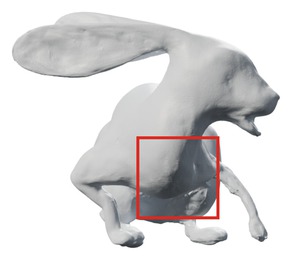}};
\node[val] at (6.864,-1.932) {2.50};
\node[cell] at (5.208,-1.982) {\includegraphics[width=1.656cm]{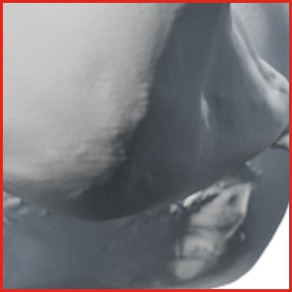}};
\node[cell] at (6.904,-0.480) {\includegraphics[width=1.656cm]{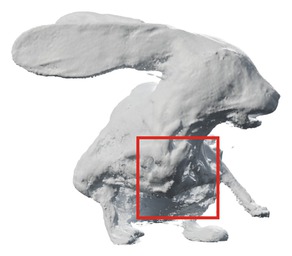}};
\node[val] at (8.560,-1.932) {4.71};
\node[cell] at (6.904,-1.982) {\includegraphics[width=1.656cm]{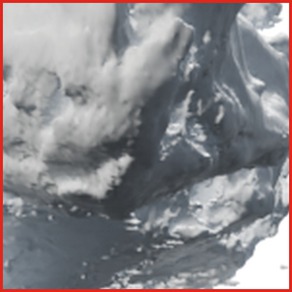}};
\node[cell] at (8.600,-0.480) {\includegraphics[width=1.656cm]{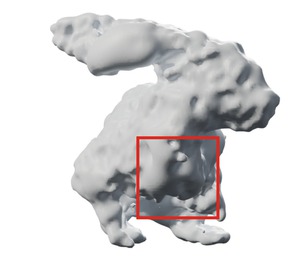}};
\node[val] at (10.256,-1.932) {7.21};
\node[cell] at (8.600,-1.982) {\includegraphics[width=1.656cm]{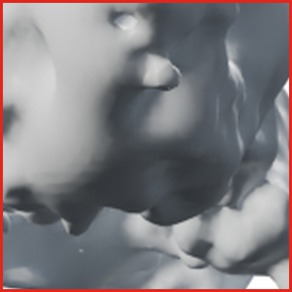}};
\node[cell] at (10.296,-0.480) {\includegraphics[width=1.656cm]{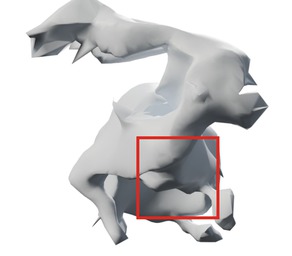}};
\node[val] at (11.952,-1.932) {7.43};
\node[cell] at (10.296,-1.982) {\includegraphics[width=1.656cm]{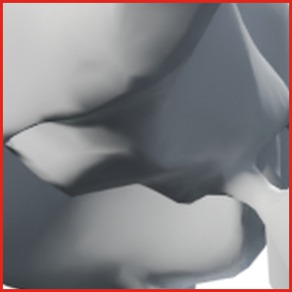}};
\node[cell] at (11.992,-0.480) {\includegraphics[width=1.656cm]{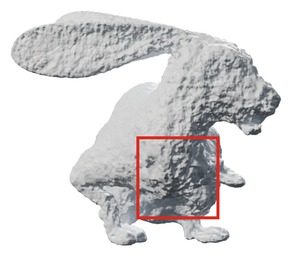}};
\node[val] at (13.648,-1.932) {2.70};
\node[cell] at (11.992,-1.982) {\includegraphics[width=1.656cm]{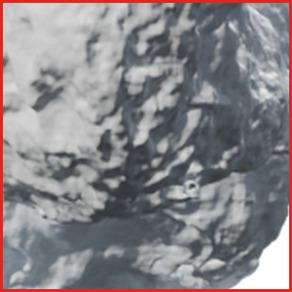}};
\node[cell] at (13.808,-0.480) {\includegraphics[width=1.656cm]{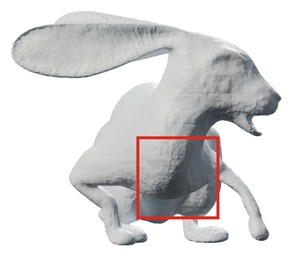}};
\node[val] at (15.464,-1.932) {\color[HTML]{0E80E5}\bfseries 1.37};
\node[cell] at (13.808,-1.982) {\includegraphics[width=1.656cm]{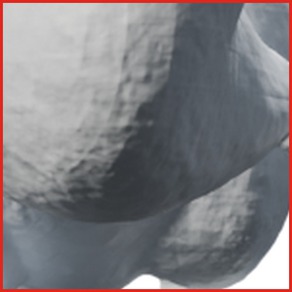}};
\node[cell] at (15.624,-0.480) {\includegraphics[width=1.656cm]{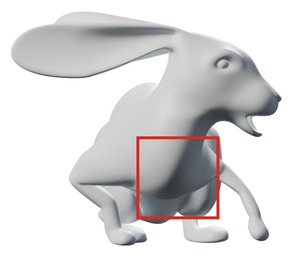}};
\node[cell] at (15.624,-1.982) {\includegraphics[width=1.656cm]{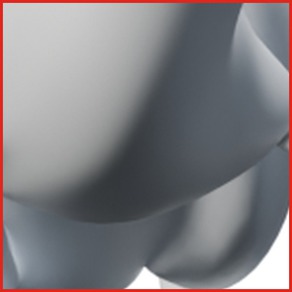}};
\node[anchor=north west, inner sep=1pt, fill=white, font=\scriptsize\itshape] at (0.000,-3.878) {DeformingThings4D fox};
\node[cell] at (0.000,-4.218) {\includegraphics[width=1.656cm]{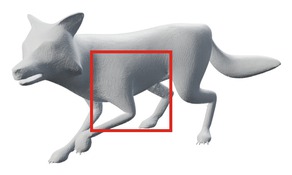}};
\node[val] at (1.656,-5.213) {3.14};
\node[cell] at (0.000,-5.263) {\includegraphics[width=1.656cm]{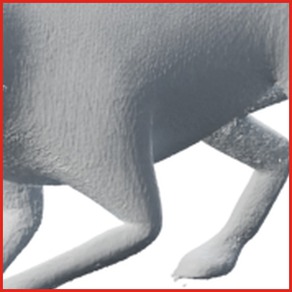}};
\node[cell] at (1.696,-4.218) {\includegraphics[width=1.656cm]{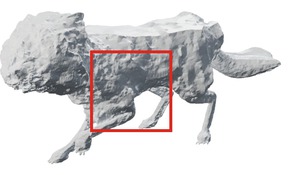}};
\node[val] at (3.352,-5.213) {25.51};
\node[cell] at (1.696,-5.263) {\includegraphics[width=1.656cm]{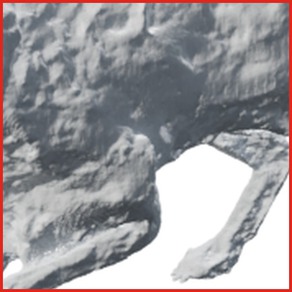}};
\node[cell] at (3.512,-4.218) {\includegraphics[width=1.656cm]{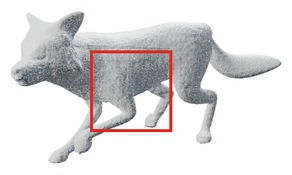}};
\node[val] at (5.168,-5.213) {3.16};
\node[cell] at (3.512,-5.263) {\includegraphics[width=1.656cm]{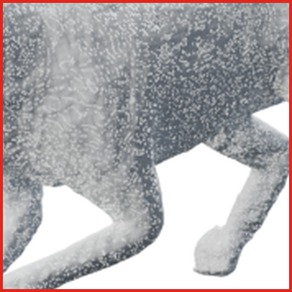}};
\node[cell] at (5.208,-4.218) {\includegraphics[width=1.656cm]{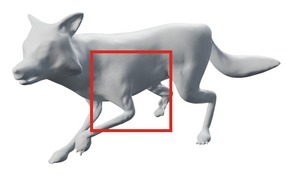}};
\node[val] at (6.864,-5.213) {4.67};
\node[cell] at (5.208,-5.263) {\includegraphics[width=1.656cm]{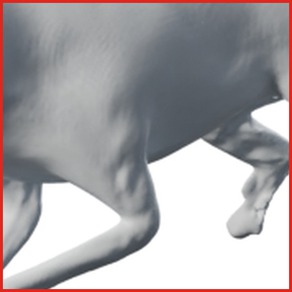}};
\node[cell] at (6.904,-4.218) {\includegraphics[width=1.656cm]{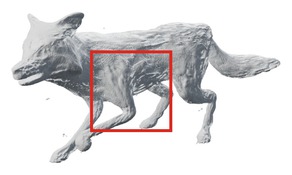}};
\node[val] at (8.560,-5.213) {4.32};
\node[cell] at (6.904,-5.263) {\includegraphics[width=1.656cm]{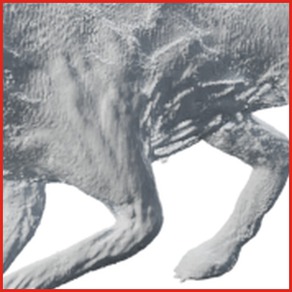}};
\node[cell] at (8.600,-4.218) {\includegraphics[width=1.656cm]{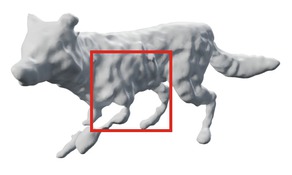}};
\node[val] at (10.256,-5.213) {6.50};
\node[cell] at (8.600,-5.263) {\includegraphics[width=1.656cm]{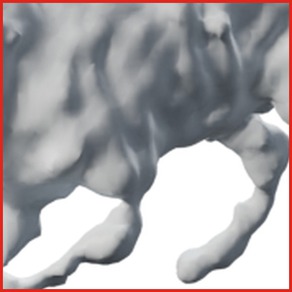}};
\node[cell] at (10.296,-4.218) {\includegraphics[width=1.656cm]{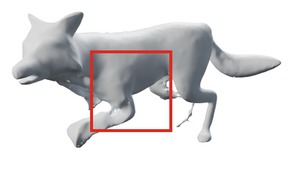}};
\node[val] at (11.952,-5.213) {10.86};
\node[cell] at (10.296,-5.263) {\includegraphics[width=1.656cm]{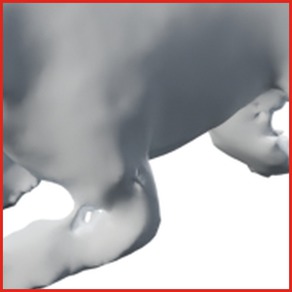}};
\node[cell] at (11.992,-4.218) {\includegraphics[width=1.656cm]{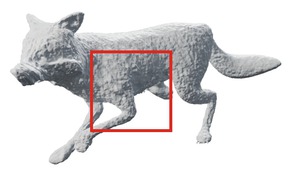}};
\node[val] at (13.648,-5.213) {4.53};
\node[cell] at (11.992,-5.263) {\includegraphics[width=1.656cm]{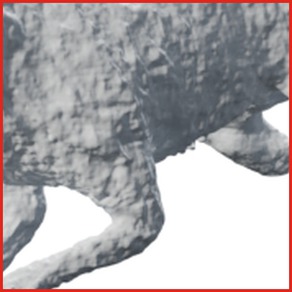}};
\node[cell] at (13.808,-4.218) {\includegraphics[width=1.656cm]{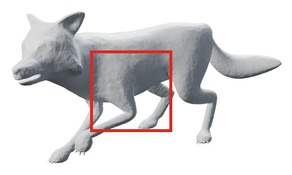}};
\node[val] at (15.464,-5.213) {\color[HTML]{0E80E5}\bfseries 2.15};
\node[cell] at (13.808,-5.263) {\includegraphics[width=1.656cm]{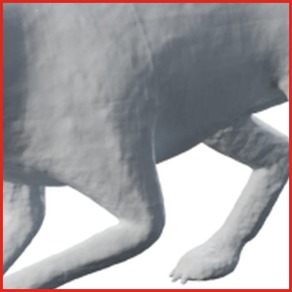}};
\node[cell] at (15.624,-4.218) {\includegraphics[width=1.656cm]{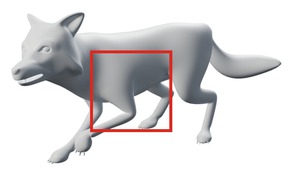}};
\node[cell] at (15.624,-5.263) {\includegraphics[width=1.656cm]{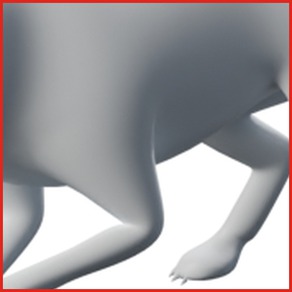}};
\node[anchor=north west, inner sep=1pt, fill=white, font=\scriptsize\itshape] at (0.000,-7.159) {DeformingThings4D canie};
\node[cell] at (0.000,-7.499) {\includegraphics[width=1.656cm]{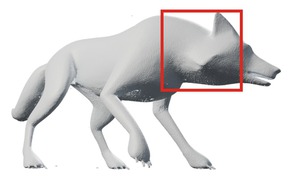}};
\node[val] at (1.656,-8.520) {3.75};
\node[cell] at (0.000,-8.570) {\includegraphics[width=1.656cm]{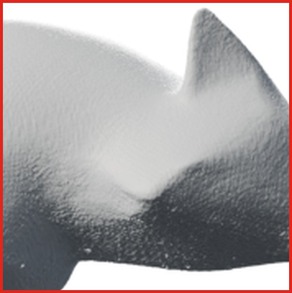}};
\node[cell] at (1.696,-7.499) {\includegraphics[width=1.656cm]{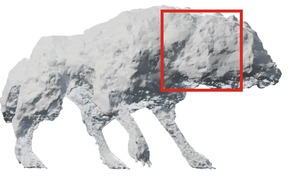}};
\node[val] at (3.352,-8.520) {37.33};
\node[cell] at (1.696,-8.570) {\includegraphics[width=1.656cm]{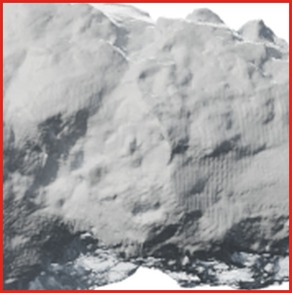}};
\node[cell] at (3.512,-7.499) {\includegraphics[width=1.656cm]{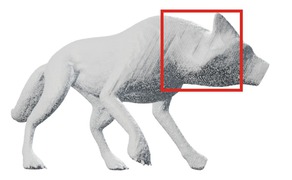}};
\node[val] at (5.168,-8.520) {4.04};
\node[cell] at (3.512,-8.570) {\includegraphics[width=1.656cm]{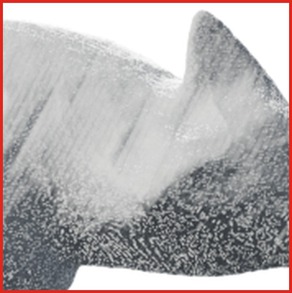}};
\node[cell] at (5.208,-7.499) {\includegraphics[width=1.656cm]{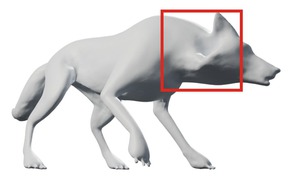}};
\node[val] at (6.864,-8.520) {5.66};
\node[cell] at (5.208,-8.570) {\includegraphics[width=1.656cm]{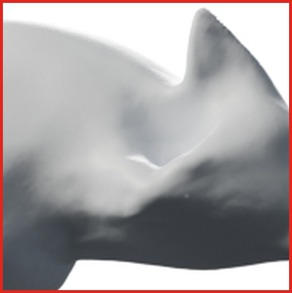}};
\node[cell] at (6.904,-7.499) {\includegraphics[width=1.656cm]{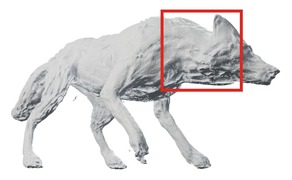}};
\node[val] at (8.560,-8.520) {7.11};
\node[cell] at (6.904,-8.570) {\includegraphics[width=1.656cm]{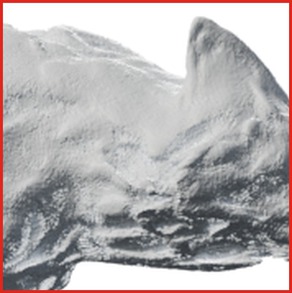}};
\node[cell] at (8.600,-7.499) {\includegraphics[width=1.656cm]{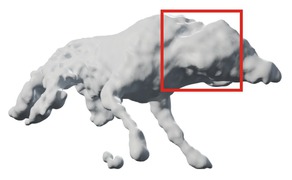}};
\node[val] at (10.256,-8.520) {12.41};
\node[cell] at (8.600,-8.570) {\includegraphics[width=1.656cm]{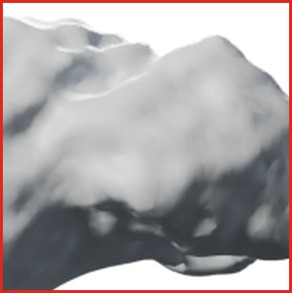}};
\node[cell] at (10.296,-7.499) {\includegraphics[width=1.656cm]{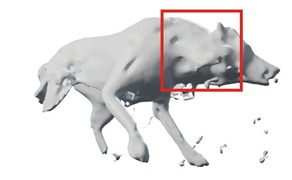}};
\node[val] at (11.952,-8.520) {20.27};
\node[cell] at (10.296,-8.570) {\includegraphics[width=1.656cm]{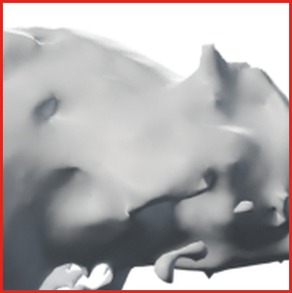}};
\node[cell] at (11.992,-7.499) {\includegraphics[width=1.656cm]{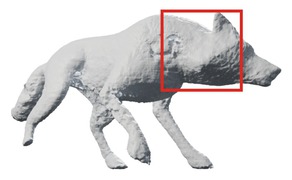}};
\node[val] at (13.648,-8.520) {8.11};
\node[cell] at (11.992,-8.570) {\includegraphics[width=1.656cm]{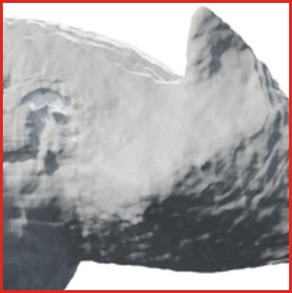}};
\node[cell] at (13.808,-7.499) {\includegraphics[width=1.656cm]{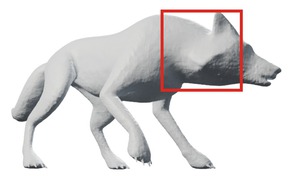}};
\node[val] at (15.464,-8.520) {\color[HTML]{0E80E5}\bfseries 2.98};
\node[cell] at (13.808,-8.570) {\includegraphics[width=1.656cm]{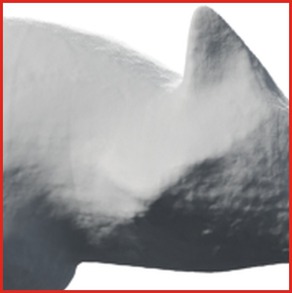}};
\node[cell] at (15.624,-7.499) {\includegraphics[width=1.656cm]{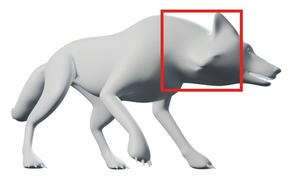}};
\node[cell] at (15.624,-8.570) {\includegraphics[width=1.656cm]{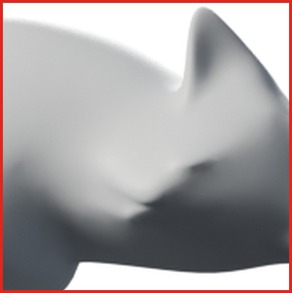}};
\node[anchor=north west, inner sep=1pt, fill=white, font=\scriptsize\itshape] at (0.000,-10.476) {DeformingThings4D grizz};
\node[cell] at (0.000,-10.816) {\includegraphics[width=1.656cm]{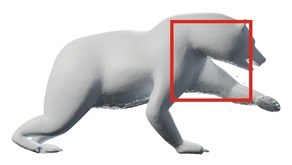}};
\node[val] at (1.656,-11.755) {8.18};
\node[cell] at (0.000,-11.805) {\includegraphics[width=1.656cm]{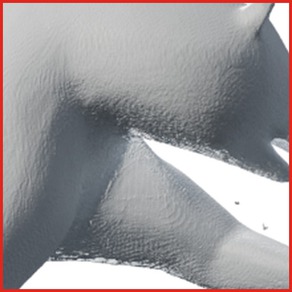}};
\node[cell] at (1.696,-10.816) {\includegraphics[width=1.656cm]{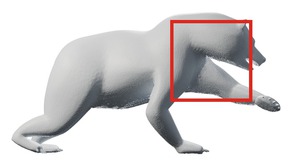}};
\node[val] at (3.352,-11.755) {8.37};
\node[cell] at (1.696,-11.805) {\includegraphics[width=1.656cm]{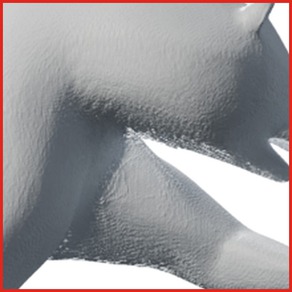}};
\node[cell] at (3.512,-10.816) {\includegraphics[width=1.656cm]{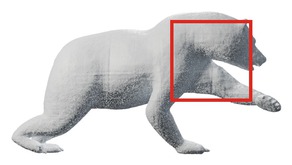}};
\node[val] at (5.168,-11.755) {8.51};
\node[cell] at (3.512,-11.805) {\includegraphics[width=1.656cm]{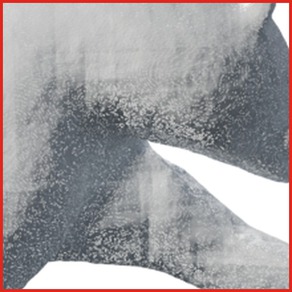}};
\node[cell] at (5.208,-10.816) {\includegraphics[width=1.656cm]{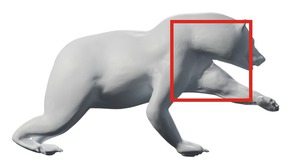}};
\node[val] at (6.864,-11.755) {10.47};
\node[cell] at (5.208,-11.805) {\includegraphics[width=1.656cm]{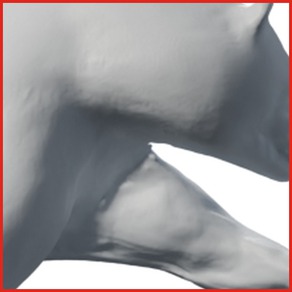}};
\node[cell] at (6.904,-10.816) {\includegraphics[width=1.656cm]{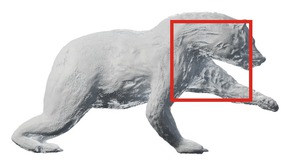}};
\node[val] at (8.560,-11.755) {12.71};
\node[cell] at (6.904,-11.805) {\includegraphics[width=1.656cm]{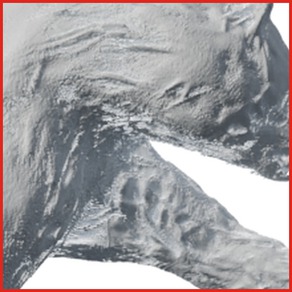}};
\node[cell] at (8.600,-10.816) {\includegraphics[width=1.656cm]{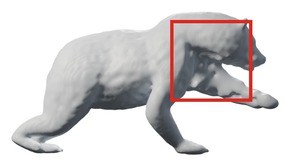}};
\node[val] at (10.256,-11.755) {19.31};
\node[cell] at (8.600,-11.805) {\includegraphics[width=1.656cm]{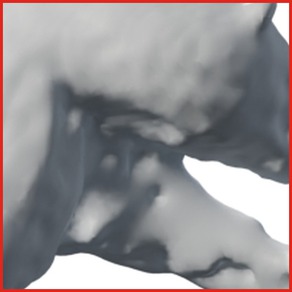}};
\node[cell] at (10.296,-10.816) {\includegraphics[width=1.656cm]{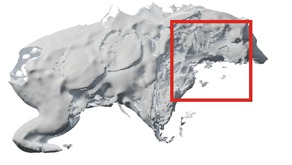}};
\node[val] at (11.952,-11.755) {79.24};
\node[cell] at (10.296,-11.805) {\includegraphics[width=1.656cm]{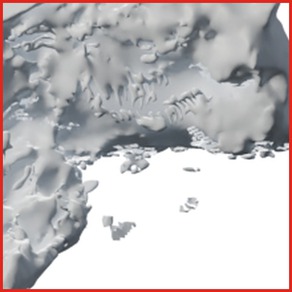}};
\node[cell] at (11.992,-10.816) {\includegraphics[width=1.656cm]{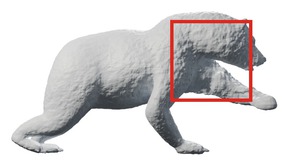}};
\node[val] at (13.648,-11.755) {9.18};
\node[cell] at (11.992,-11.805) {\includegraphics[width=1.656cm]{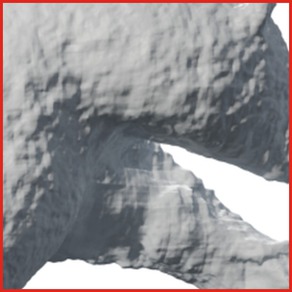}};
\node[cell] at (13.808,-10.816) {\includegraphics[width=1.656cm]{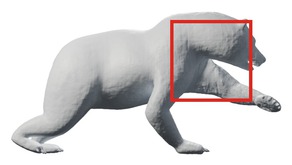}};
\node[val] at (15.464,-11.755) {\color[HTML]{0E80E5}\bfseries 6.14};
\node[cell] at (13.808,-11.805) {\includegraphics[width=1.656cm]{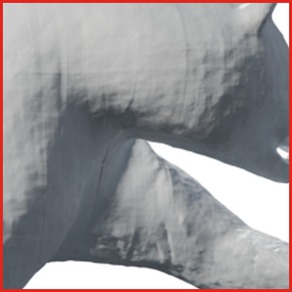}};
\node[cell] at (15.624,-10.816) {\includegraphics[width=1.656cm]{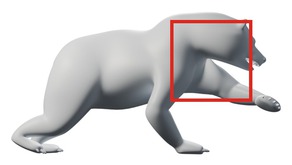}};
\node[cell] at (15.624,-11.805) {\includegraphics[width=1.656cm]{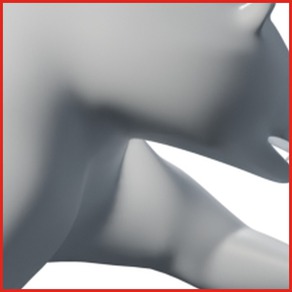}};
\node[anchor=north west, inner sep=1pt, fill=white, font=\scriptsize\itshape] at (0.000,-13.701) {DeformingThings4D dragon};
\node[cell] at (0.000,-14.041) {\includegraphics[width=1.656cm]{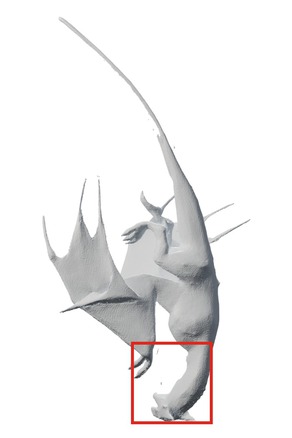}};
\node[val] at (1.656,-16.528) {26.10};
\node[cell] at (0.000,-16.578) {\includegraphics[width=1.656cm]{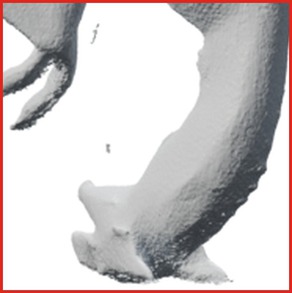}};
\node[cell] at (1.696,-14.041) {\includegraphics[width=1.656cm]{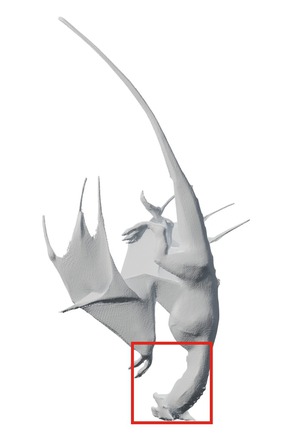}};
\node[val] at (3.352,-16.528) {25.84};
\node[cell] at (1.696,-16.578) {\includegraphics[width=1.656cm]{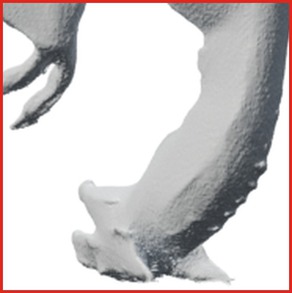}};
\node[cell] at (3.512,-14.041) {\includegraphics[width=1.656cm]{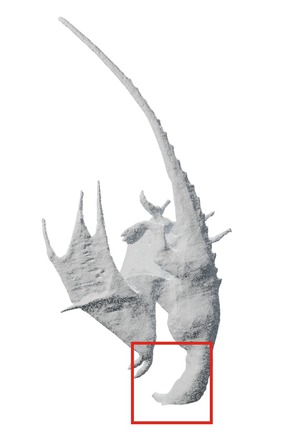}};
\node[val] at (5.168,-16.528) {43.41};
\node[cell] at (3.512,-16.578) {\includegraphics[width=1.656cm]{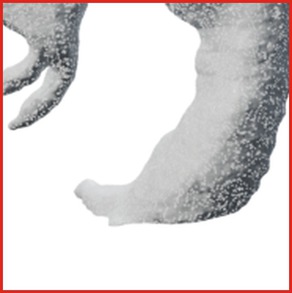}};
\node[cell] at (5.208,-14.041) {\includegraphics[width=1.656cm]{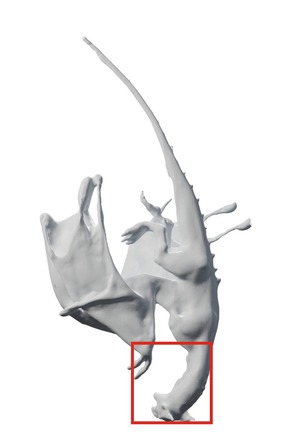}};
\node[val] at (6.864,-16.528) {85.07};
\node[cell] at (5.208,-16.578) {\includegraphics[width=1.656cm]{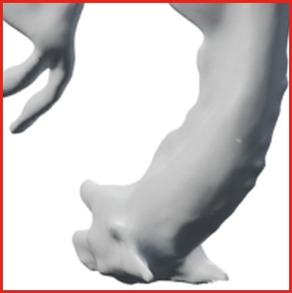}};
\node[cell] at (6.904,-14.041) {\includegraphics[width=1.656cm]{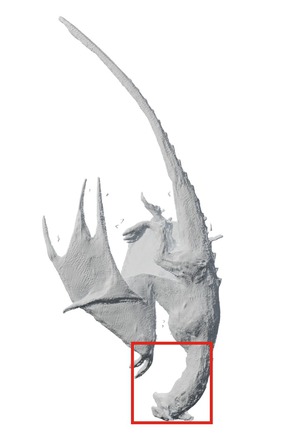}};
\node[val] at (8.560,-16.528) {34.49};
\node[cell] at (6.904,-16.578) {\includegraphics[width=1.656cm]{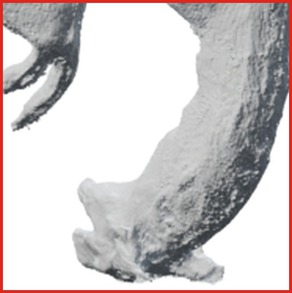}};
\node[cell] at (8.600,-14.041) {\includegraphics[width=1.656cm]{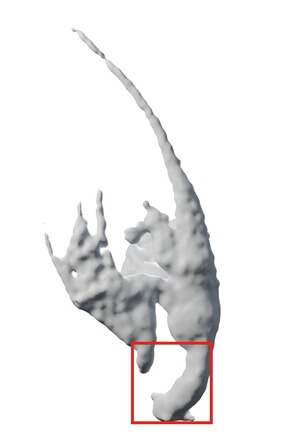}};
\node[val] at (10.256,-16.528) {85.42};
\node[cell] at (8.600,-16.578) {\includegraphics[width=1.656cm]{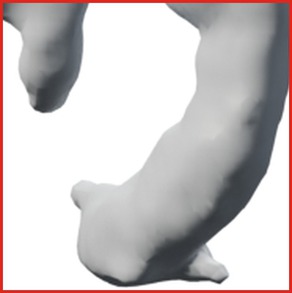}};
\node[cell] at (10.296,-14.041) {\includegraphics[width=1.656cm]{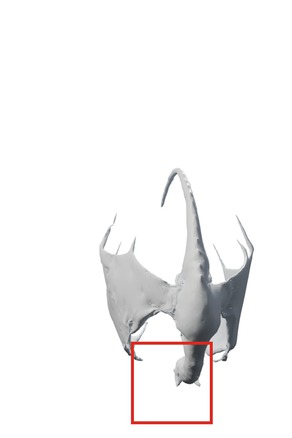}};
\node[val] at (11.952,-16.528) {1328.86};
\node[cell] at (10.296,-16.578) {\includegraphics[width=1.656cm]{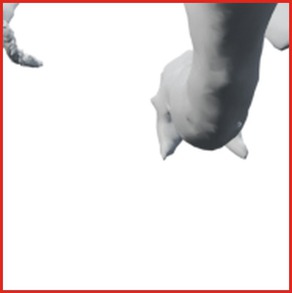}};
\node[cell] at (11.992,-14.041) {\includegraphics[width=1.656cm]{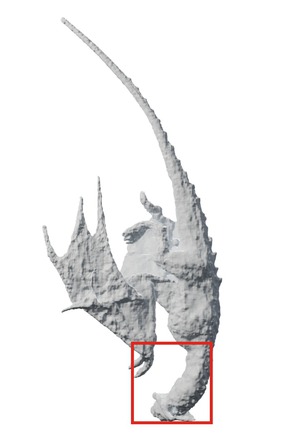}};
\node[val] at (13.648,-16.528) {34.13};
\node[cell] at (11.992,-16.578) {\includegraphics[width=1.656cm]{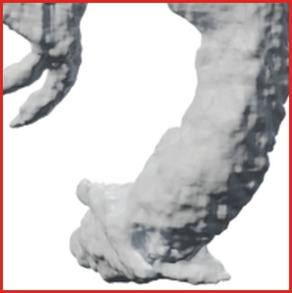}};
\node[cell] at (13.808,-14.041) {\includegraphics[width=1.656cm]{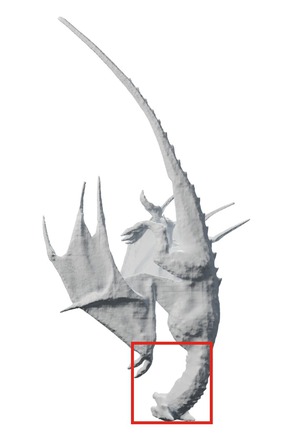}};
\node[val] at (15.464,-16.528) {\color[HTML]{0E80E5}\bfseries 22.57};
\node[cell] at (13.808,-16.578) {\includegraphics[width=1.656cm]{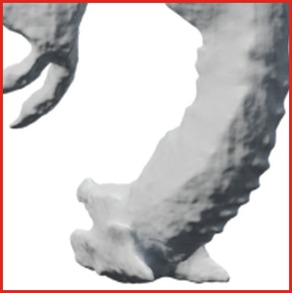}};
\node[cell] at (15.624,-14.041) {\includegraphics[width=1.656cm]{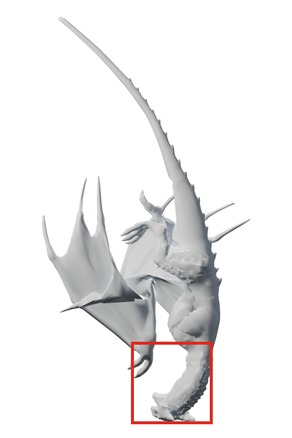}};
\node[cell] at (15.624,-16.578) {\includegraphics[width=1.656cm]{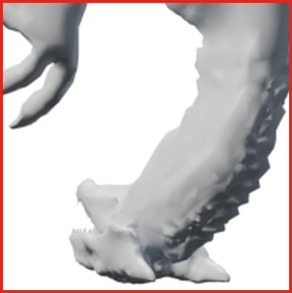}};
\end{tikzpicture}
\caption{\textbf{Geometry compared with the baselines} (DeformingThings4D). Numbers: Chamfer distance of the frame (mm), as in \cref{tab:geometry_avg}. The red box marks the enlarged region (same box for every method). Left to right: per-frame static methods (no correspondences), dynamic methods, Ours, ground truth.}
\label{fig:geo_cmp_1}
\end{figure*}

\begin{figure*}[p]
\centering
\begin{tikzpicture}[x=1cm, y=1cm, font=\scriptsize, cell/.style={anchor=north west, inner sep=0},
    val/.style={font=\tiny, anchor=south east, fill=white, fill opacity=0.85, text opacity=1, inner sep=1pt}]
\draw[black!45, line width=0.4pt] (0.050,0.420) -- (3.302,0.420);
\node[anchor=south, inner sep=1pt, font=\scriptsize\itshape] at (1.676,0.420) {Static (per frame)};
\draw[black!45, line width=0.4pt] (3.562,0.420) -- (15.414,0.420);
\node[anchor=south, inner sep=1pt, font=\scriptsize\itshape] at (9.488,0.420) {Dynamic};
\draw[black!40, dashed, line width=0.5pt] (3.432,0.720) -- (3.432,-18.456);
\node[anchor=south, inner sep=1pt] at (0.828,0.000) {2DGS};
\node[anchor=south, inner sep=1pt] at (2.524,0.000) {PGSR};
\node[anchor=south, inner sep=1pt] at (4.340,0.000) {NeuS2};
\node[anchor=south, inner sep=1pt] at (6.036,0.000) {AT-GS};
\node[anchor=south, inner sep=1pt] at (7.732,0.000) {SC-GS};
\node[anchor=south, inner sep=1pt] at (9.428,0.000) {DG-Mesh};
\node[anchor=south, inner sep=1pt] at (11.124,0.000) {GauSTAR};
\node[anchor=south, inner sep=1pt] at (12.820,0.000) {MoAngelo};
\node[anchor=south, inner sep=1pt] at (14.636,0.000) {\color[HTML]{0E80E5}\textbf{Ours}};
\node[anchor=south, inner sep=1pt] at (16.452,0.000) {GT};
\node[anchor=north west, inner sep=1pt, fill=white, font=\scriptsize\itshape] at (0.000,-0.140) {ActorsHQ actor6\_seq2\_1604};
\node[cell] at (0.000,-0.480) {\includegraphics[width=1.656cm]{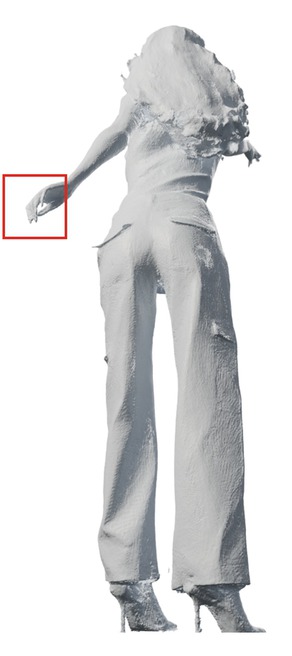}};
\node[val] at (1.656,-4.203) {3.86};
\node[cell] at (0.000,-4.253) {\includegraphics[width=1.656cm]{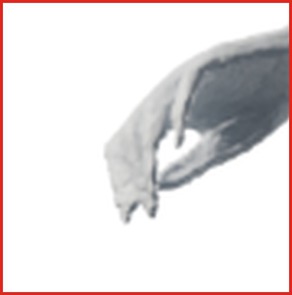}};
\node[cell] at (1.696,-0.480) {\includegraphics[width=1.656cm]{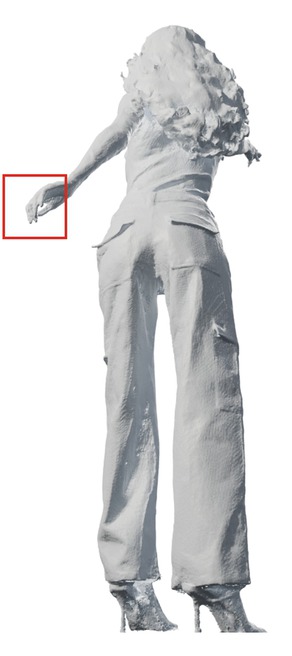}};
\node[val] at (3.352,-4.203) {3.06};
\node[cell] at (1.696,-4.253) {\includegraphics[width=1.656cm]{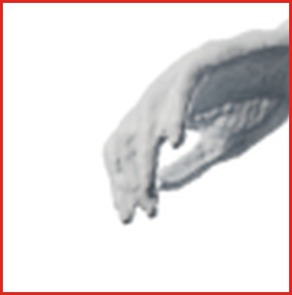}};
\node[cell] at (3.512,-0.480) {\includegraphics[width=1.656cm]{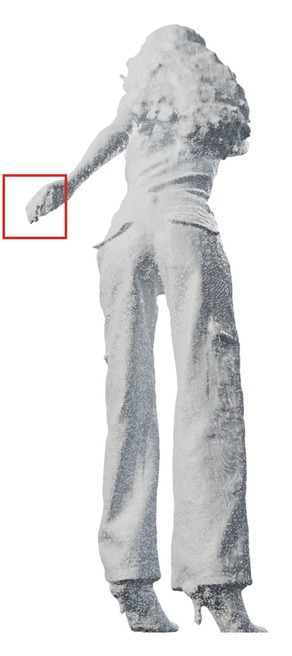}};
\node[val] at (5.168,-4.203) {4.96};
\node[cell] at (3.512,-4.253) {\includegraphics[width=1.656cm]{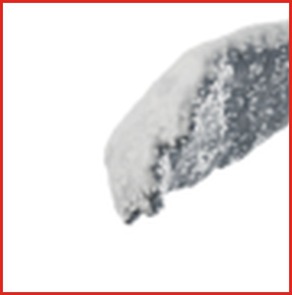}};
\node[cell] at (5.208,-0.480) {\includegraphics[width=1.656cm]{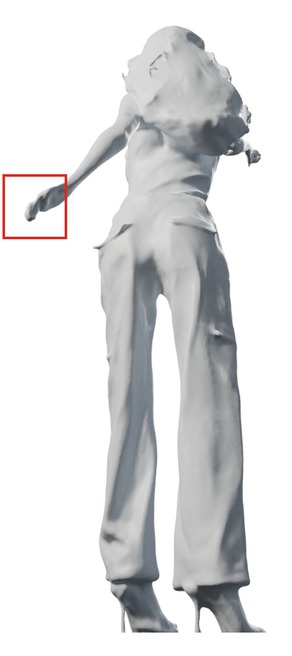}};
\node[val] at (6.864,-4.203) {5.88};
\node[cell] at (5.208,-4.253) {\includegraphics[width=1.656cm]{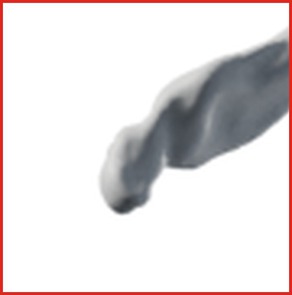}};
\node[cell] at (6.904,-0.480) {\includegraphics[width=1.656cm]{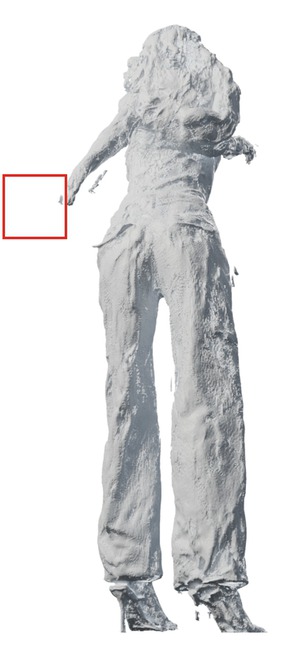}};
\node[val] at (8.560,-4.203) {6.73};
\node[cell] at (6.904,-4.253) {\includegraphics[width=1.656cm]{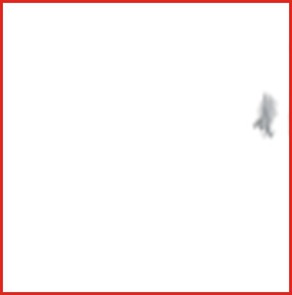}};
\node[cell] at (8.600,-0.480) {\includegraphics[width=1.656cm]{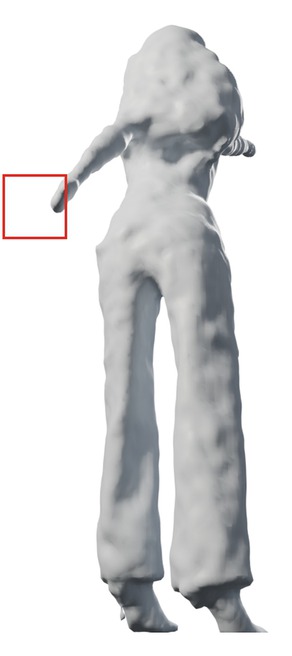}};
\node[val] at (10.256,-4.203) {6.82};
\node[cell] at (8.600,-4.253) {\includegraphics[width=1.656cm]{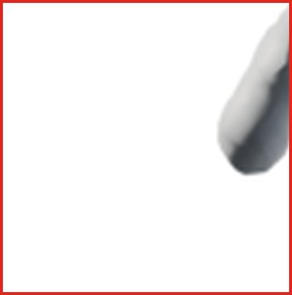}};
\node[cell] at (10.296,-0.480) {\includegraphics[width=1.656cm]{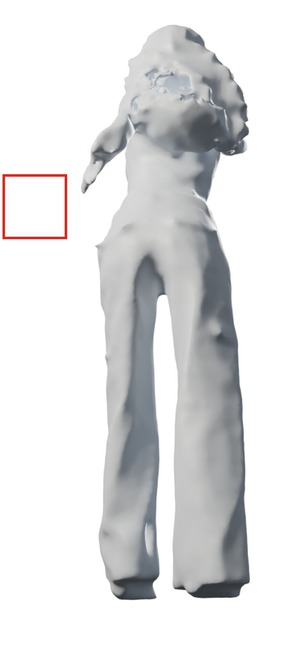}};
\node[val] at (11.952,-4.203) {9.44};
\node[cell] at (10.296,-4.253) {\includegraphics[width=1.656cm]{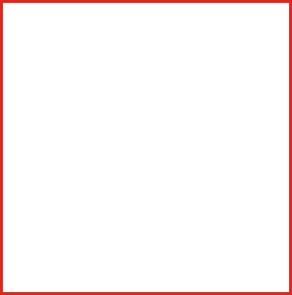}};
\node[cell] at (11.992,-0.480) {\includegraphics[width=1.656cm]{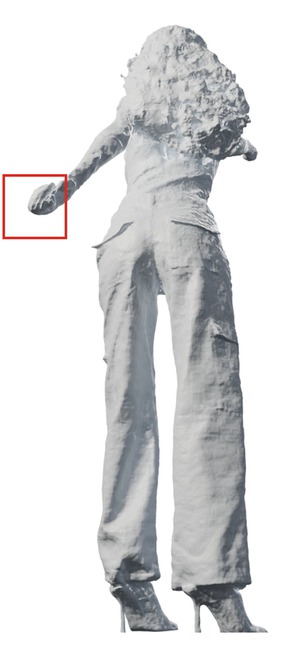}};
\node[val] at (13.648,-4.203) {5.48};
\node[cell] at (11.992,-4.253) {\includegraphics[width=1.656cm]{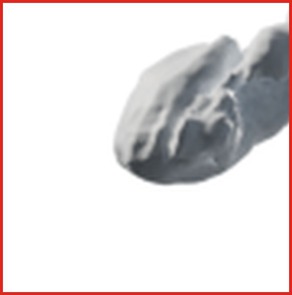}};
\node[cell] at (13.808,-0.480) {\includegraphics[width=1.656cm]{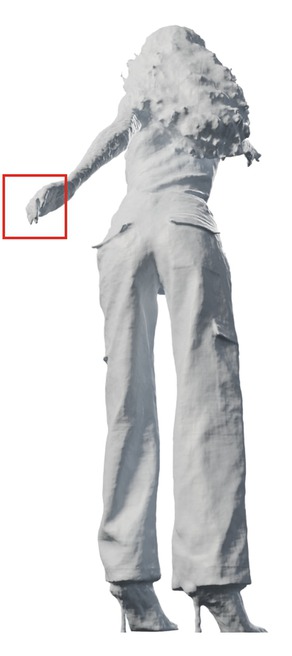}};
\node[val] at (15.464,-4.203) {\color[HTML]{0E80E5}\bfseries 3.22};
\node[cell] at (13.808,-4.253) {\includegraphics[width=1.656cm]{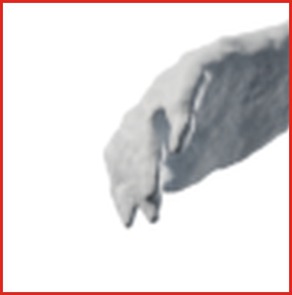}};
\node[cell] at (15.624,-0.480) {\includegraphics[width=1.656cm]{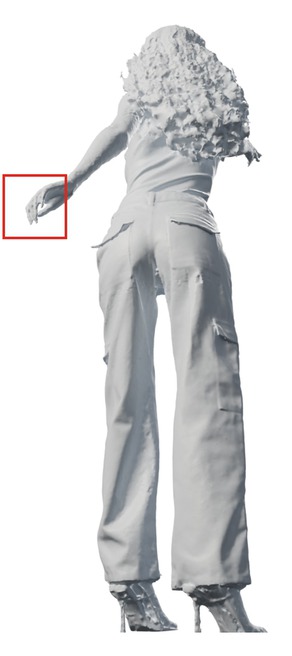}};
\node[cell] at (15.624,-4.253) {\includegraphics[width=1.656cm]{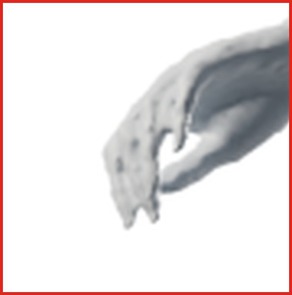}};
\node[anchor=north west, inner sep=1pt, fill=white, font=\scriptsize\itshape] at (0.000,-6.167) {ActorsHQ actor6\_seq2\_1081};
\node[cell] at (0.000,-6.507) {\includegraphics[width=1.656cm]{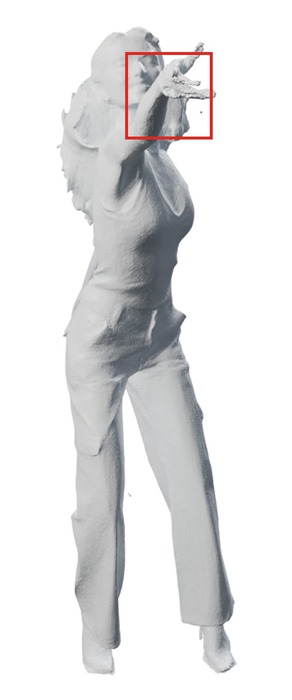}};
\node[val] at (1.656,-10.485) {3.82};
\node[cell] at (0.000,-10.535) {\includegraphics[width=1.656cm]{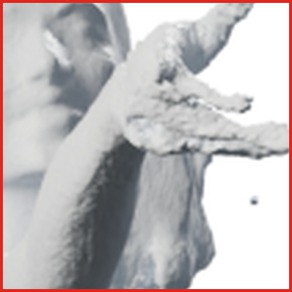}};
\node[cell] at (1.696,-6.507) {\includegraphics[width=1.656cm]{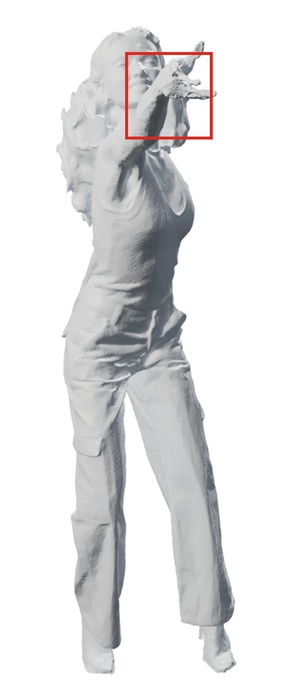}};
\node[val] at (3.352,-10.485) {3.08};
\node[cell] at (1.696,-10.535) {\includegraphics[width=1.656cm]{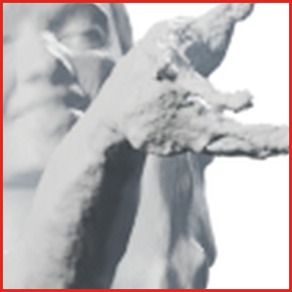}};
\node[cell] at (3.512,-6.507) {\includegraphics[width=1.656cm]{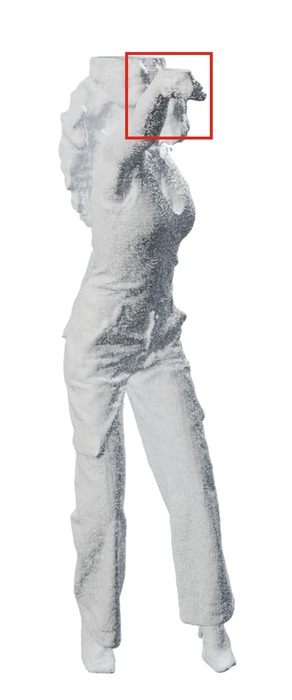}};
\node[val] at (5.168,-10.485) {4.44};
\node[cell] at (3.512,-10.535) {\includegraphics[width=1.656cm]{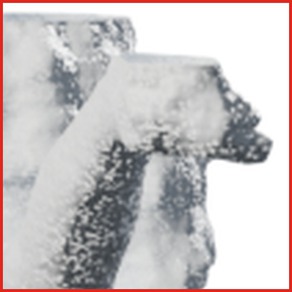}};
\node[cell] at (5.208,-6.507) {\includegraphics[width=1.656cm]{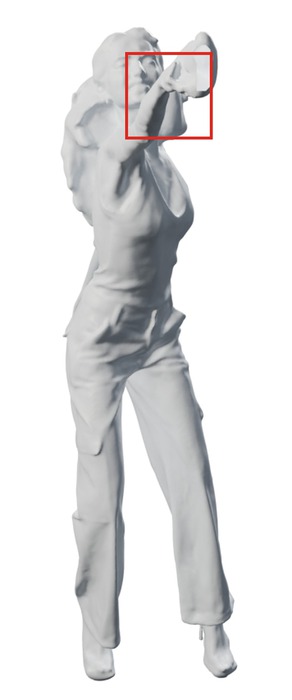}};
\node[val] at (6.864,-10.485) {5.01};
\node[cell] at (5.208,-10.535) {\includegraphics[width=1.656cm]{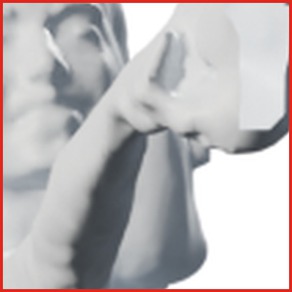}};
\node[cell] at (6.904,-6.507) {\includegraphics[width=1.656cm]{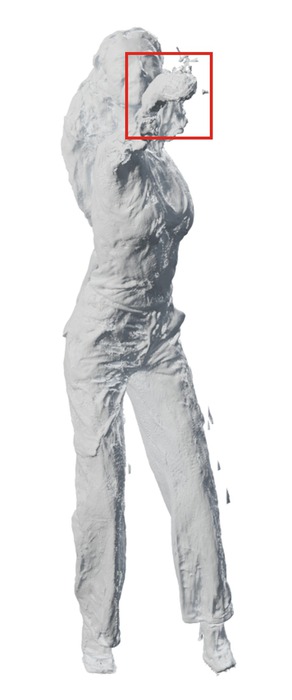}};
\node[val] at (8.560,-10.485) {5.62};
\node[cell] at (6.904,-10.535) {\includegraphics[width=1.656cm]{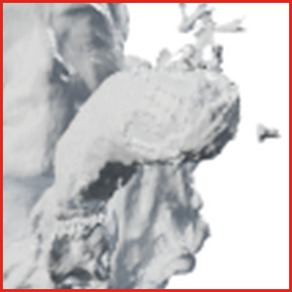}};
\node[cell] at (8.600,-6.507) {\includegraphics[width=1.656cm]{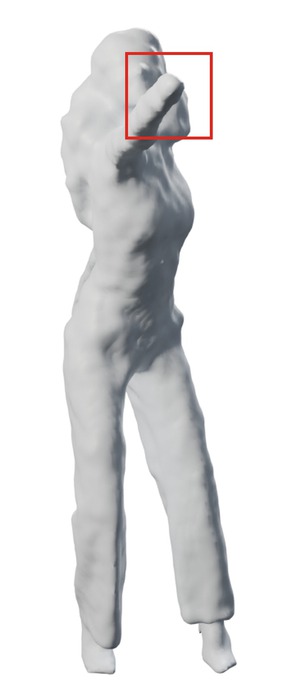}};
\node[val] at (10.256,-10.485) {6.41};
\node[cell] at (8.600,-10.535) {\includegraphics[width=1.656cm]{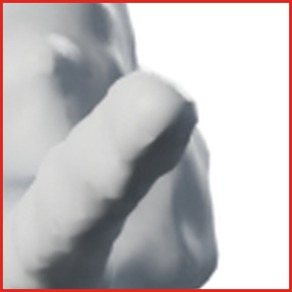}};
\node[cell] at (10.296,-6.507) {\includegraphics[width=1.656cm]{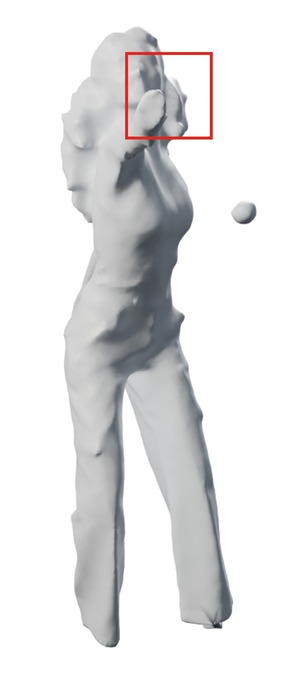}};
\node[val] at (11.952,-10.485) {7.27};
\node[cell] at (10.296,-10.535) {\includegraphics[width=1.656cm]{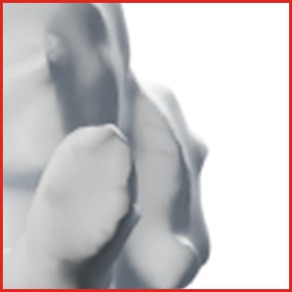}};
\node[cell] at (11.992,-6.507) {\includegraphics[width=1.656cm]{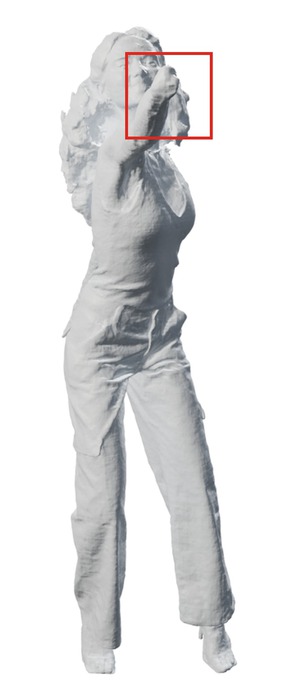}};
\node[val] at (13.648,-10.485) {3.99};
\node[cell] at (11.992,-10.535) {\includegraphics[width=1.656cm]{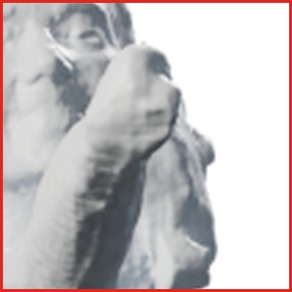}};
\node[cell] at (13.808,-6.507) {\includegraphics[width=1.656cm]{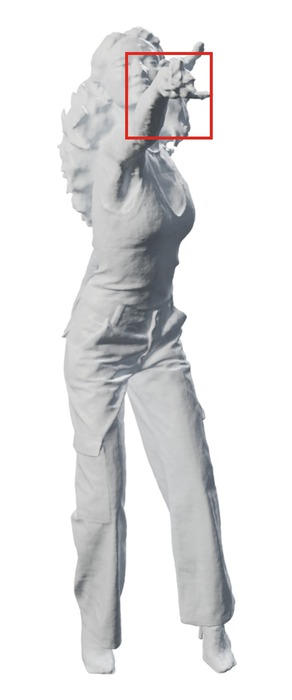}};
\node[val] at (15.464,-10.485) {\color[HTML]{0E80E5}\bfseries 3.13};
\node[cell] at (13.808,-10.535) {\includegraphics[width=1.656cm]{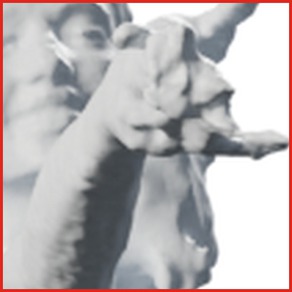}};
\node[cell] at (15.624,-6.507) {\includegraphics[width=1.656cm]{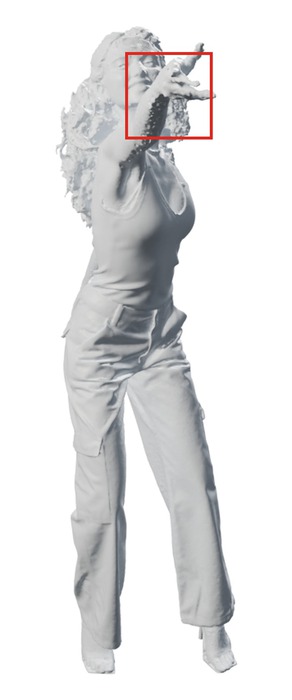}};
\node[cell] at (15.624,-10.535) {\includegraphics[width=1.656cm]{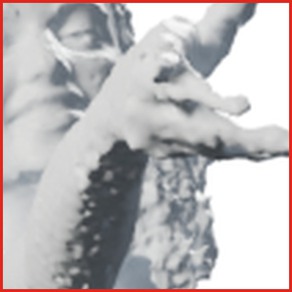}};
\node[anchor=north west, inner sep=1pt, fill=white, font=\scriptsize\itshape] at (0.000,-12.431) {ActorsHQ actor4\_seq2\_1258};
\node[cell] at (0.000,-12.771) {\includegraphics[width=1.656cm]{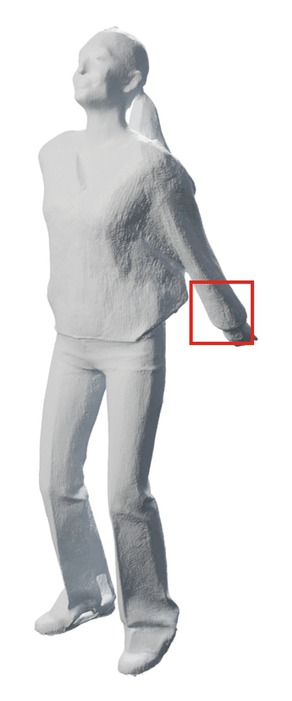}};
\node[val] at (1.656,-16.750) {4.24};
\node[cell] at (0.000,-16.800) {\includegraphics[width=1.656cm]{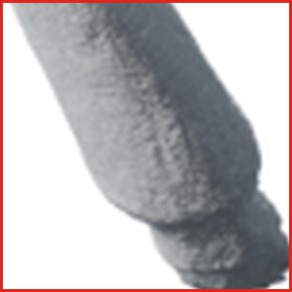}};
\node[cell] at (1.696,-12.771) {\includegraphics[width=1.656cm]{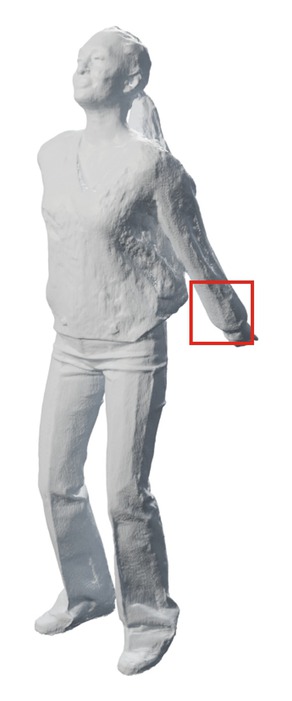}};
\node[val] at (3.352,-16.750) {3.46};
\node[cell] at (1.696,-16.800) {\includegraphics[width=1.656cm]{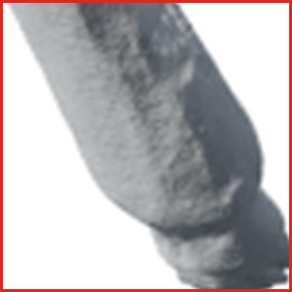}};
\node[cell] at (3.512,-12.771) {\includegraphics[width=1.656cm]{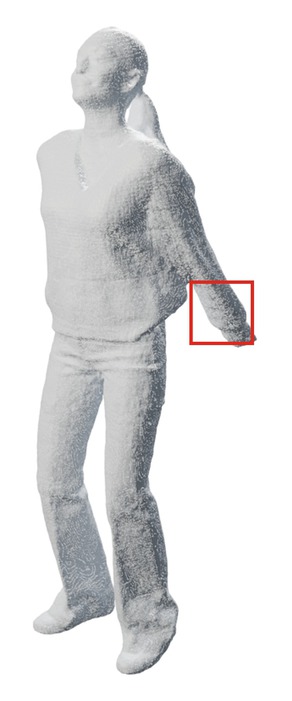}};
\node[val] at (5.168,-16.750) {4.96};
\node[cell] at (3.512,-16.800) {\includegraphics[width=1.656cm]{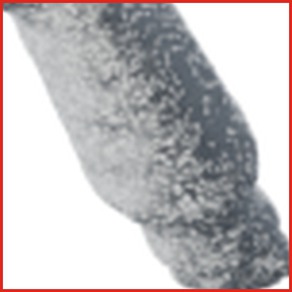}};
\node[cell] at (5.208,-12.771) {\includegraphics[width=1.656cm]{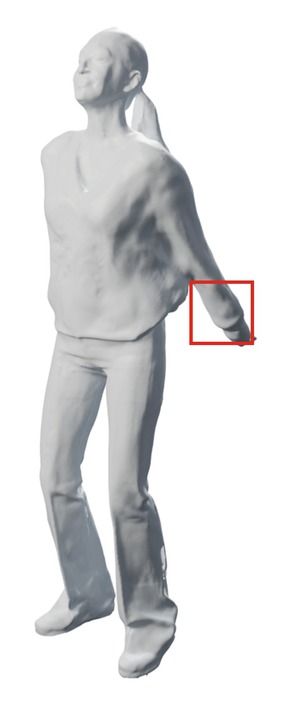}};
\node[val] at (6.864,-16.750) {5.95};
\node[cell] at (5.208,-16.800) {\includegraphics[width=1.656cm]{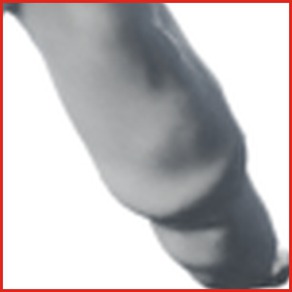}};
\node[cell] at (6.904,-12.771) {\includegraphics[width=1.656cm]{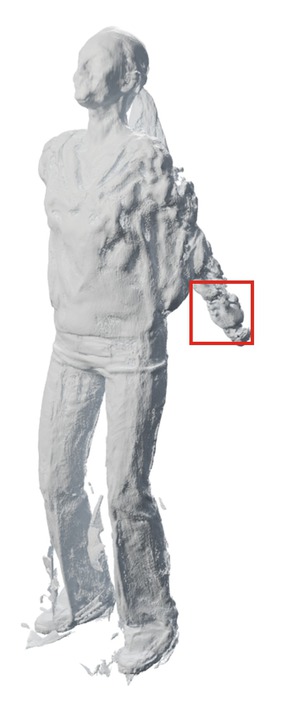}};
\node[val] at (8.560,-16.750) {7.99};
\node[cell] at (6.904,-16.800) {\includegraphics[width=1.656cm]{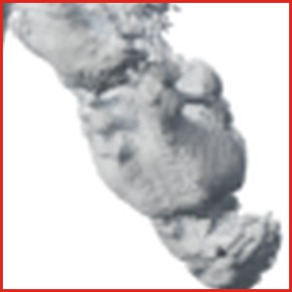}};
\node[cell] at (8.600,-12.771) {\includegraphics[width=1.656cm]{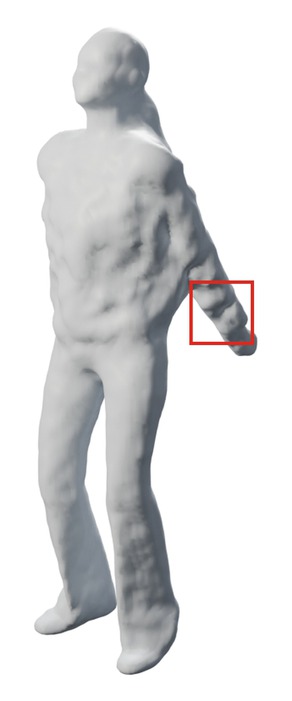}};
\node[val] at (10.256,-16.750) {8.07};
\node[cell] at (8.600,-16.800) {\includegraphics[width=1.656cm]{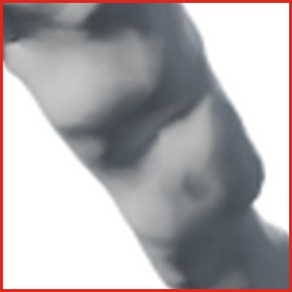}};
\node[cell] at (10.296,-12.771) {\includegraphics[width=1.656cm]{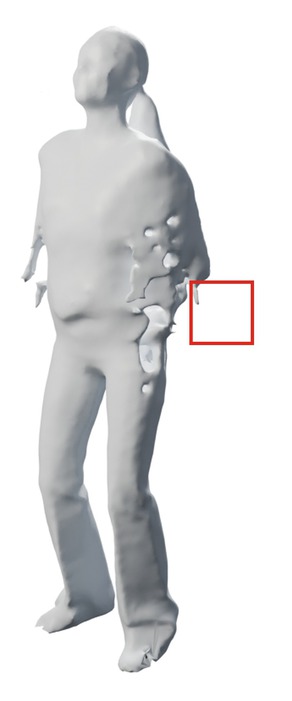}};
\node[val] at (11.952,-16.750) {13.16};
\node[cell] at (10.296,-16.800) {\includegraphics[width=1.656cm]{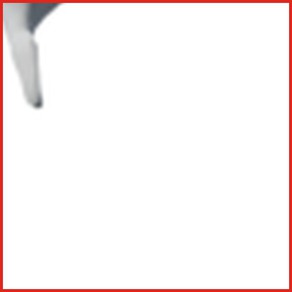}};
\node[cell] at (11.992,-12.771) {\includegraphics[width=1.656cm]{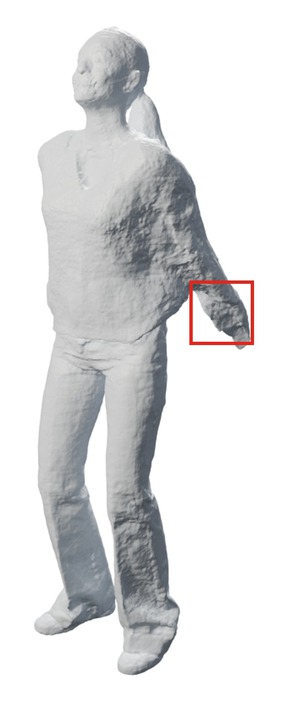}};
\node[val] at (13.648,-16.750) {5.50};
\node[cell] at (11.992,-16.800) {\includegraphics[width=1.656cm]{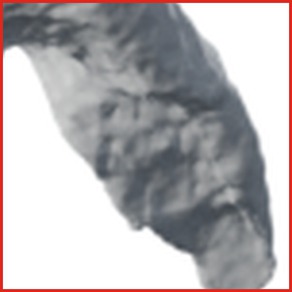}};
\node[cell] at (13.808,-12.771) {\includegraphics[width=1.656cm]{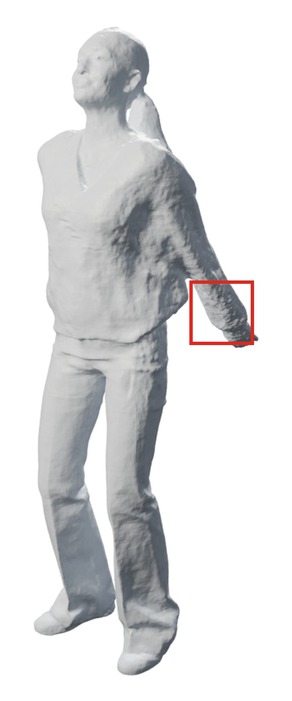}};
\node[val] at (15.464,-16.750) {\color[HTML]{0E80E5}\bfseries 3.80};
\node[cell] at (13.808,-16.800) {\includegraphics[width=1.656cm]{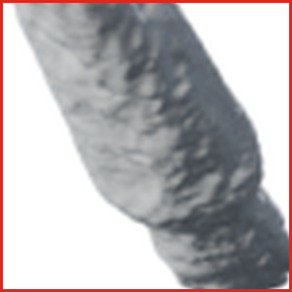}};
\node[cell] at (15.624,-12.771) {\includegraphics[width=1.656cm]{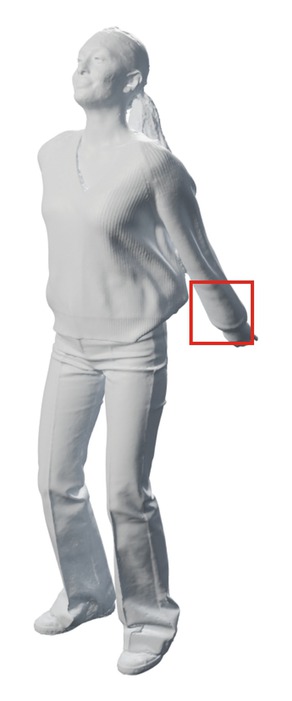}};
\node[cell] at (15.624,-16.800) {\includegraphics[width=1.656cm]{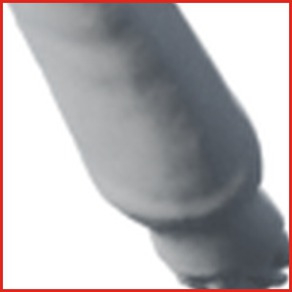}};
\end{tikzpicture}
\caption{\textbf{Geometry compared with the baselines} (ActorsHQ). Numbers: Chamfer distance of the frame (mm), as in \cref{tab:geometry_avg}. The red box marks the enlarged region (same box for every method). Left to right: per-frame static methods (no correspondences), dynamic methods, Ours, ground truth.}
\label{fig:geo_cmp_2}
\end{figure*}

\begin{figure*}[p]
\centering
\begin{tikzpicture}[x=1cm, y=1cm, font=\scriptsize, cell/.style={anchor=north west, inner sep=0},
    val/.style={font=\tiny, anchor=south east, fill=white, fill opacity=0.85, text opacity=1, inner sep=1pt}]
\node[anchor=south, inner sep=1pt] at (1.413,0.000) {NeuS2};
\node[anchor=south, inner sep=1pt] at (4.280,0.000) {AT-GS};
\node[anchor=south, inner sep=1pt] at (7.147,0.000) {DG-Mesh};
\node[anchor=south, inner sep=1pt] at (10.013,0.000) {GauSTAR};
\node[anchor=south, inner sep=1pt] at (12.880,0.000) {MoAngelo};
\node[anchor=south, inner sep=1pt] at (15.867,0.000) {\color[HTML]{0E80E5}\textbf{Ours}};
\node[anchor=north west, inner sep=1pt, fill=white, font=\scriptsize\itshape] at (0.000,-0.140) {TaoGS drawsword};
\node[cell] at (0.000,-0.480) {\includegraphics[width=2.827cm]{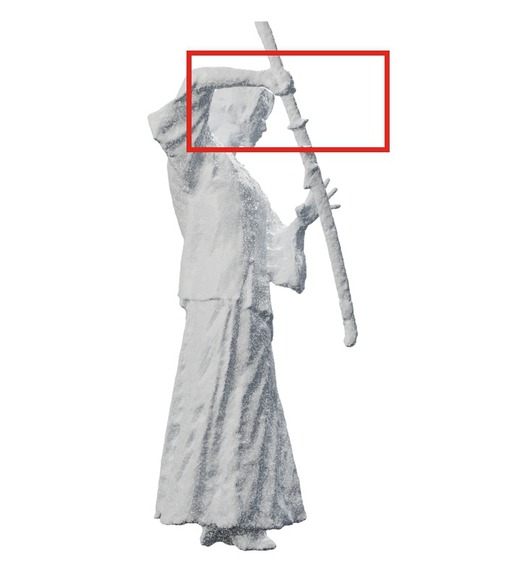}};
\node[cell] at (0.000,-3.782) {\includegraphics[width=2.827cm]{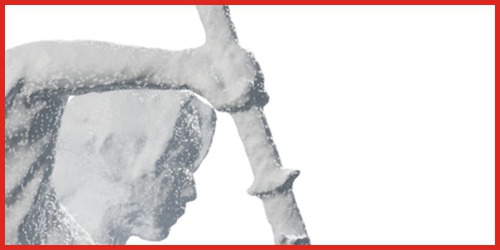}};
\node[cell] at (2.867,-0.480) {\includegraphics[width=2.827cm]{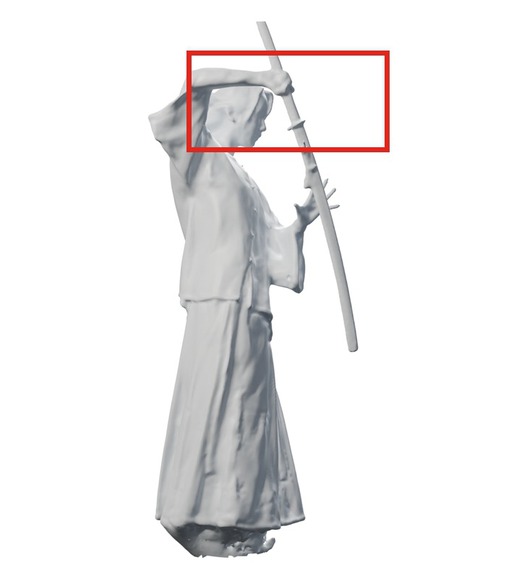}};
\node[cell] at (2.867,-3.782) {\includegraphics[width=2.827cm]{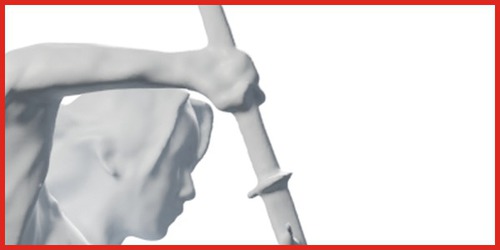}};
\node[cell] at (5.733,-0.480) {\includegraphics[width=2.827cm]{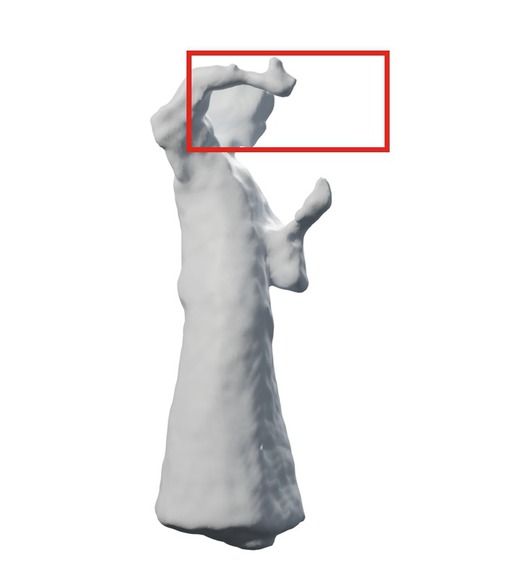}};
\node[cell] at (5.733,-3.782) {\includegraphics[width=2.827cm]{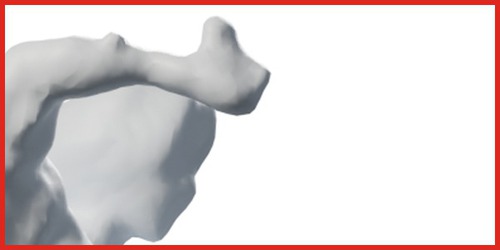}};
\node[cell] at (8.600,-0.480) {\includegraphics[width=2.827cm]{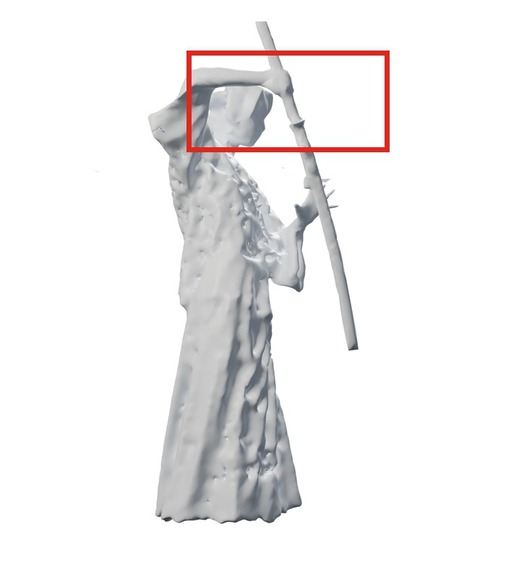}};
\node[cell] at (8.600,-3.782) {\includegraphics[width=2.827cm]{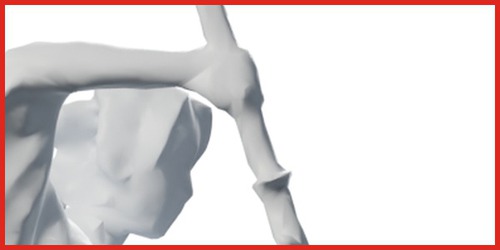}};
\node[cell] at (11.467,-0.480) {\includegraphics[width=2.827cm]{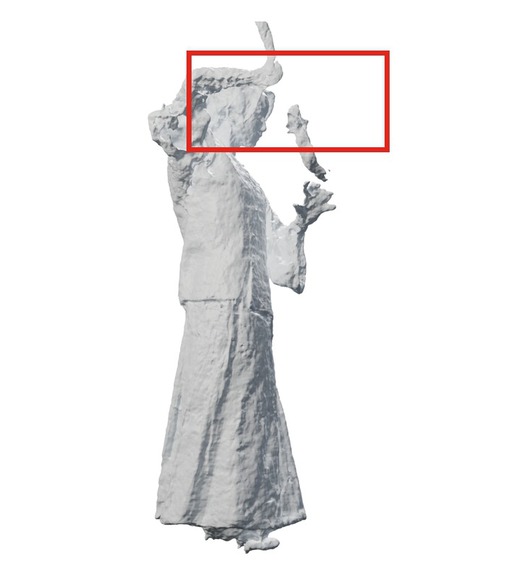}};
\node[cell] at (11.467,-3.782) {\includegraphics[width=2.827cm]{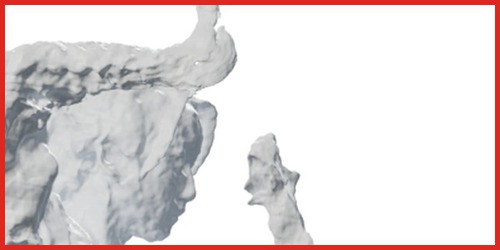}};
\node[cell] at (14.453,-0.480) {\includegraphics[width=2.827cm]{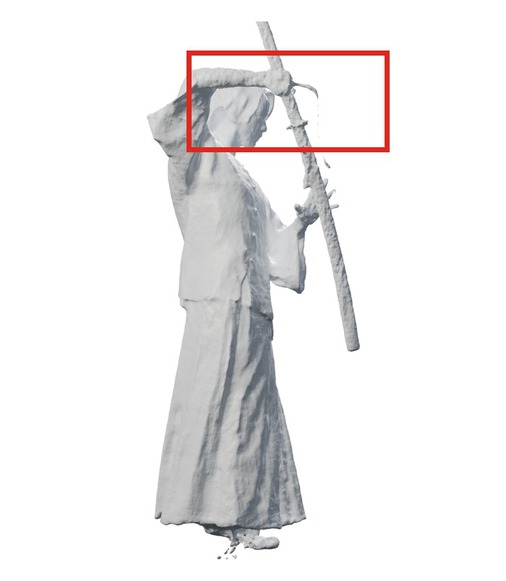}};
\node[cell] at (14.453,-3.782) {\includegraphics[width=2.827cm]{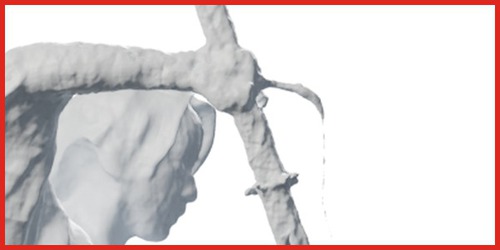}};
\node[anchor=north west, inner sep=1pt, fill=white, font=\scriptsize\itshape] at (0.000,-5.435) {TaoGS changeclothes};
\node[cell] at (0.000,-5.775) {\includegraphics[width=2.827cm]{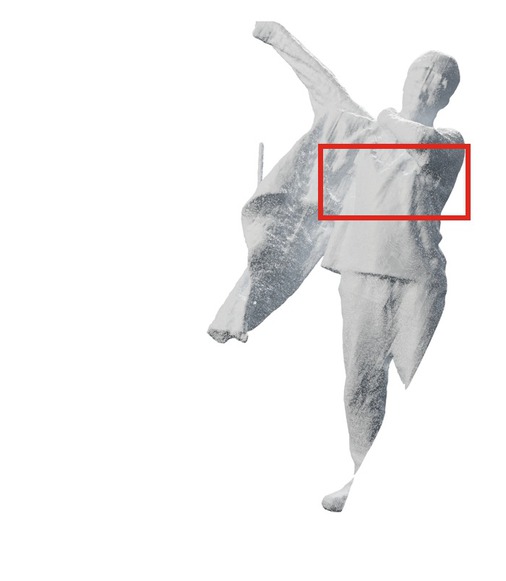}};
\node[cell] at (0.000,-9.073) {\includegraphics[width=2.827cm]{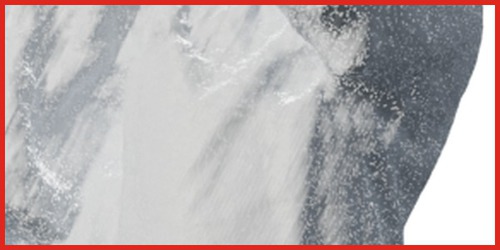}};
\node[cell] at (2.867,-5.775) {\includegraphics[width=2.827cm]{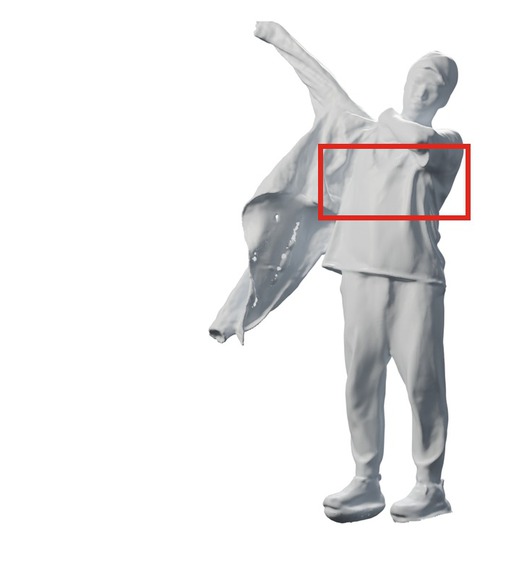}};
\node[cell] at (2.867,-9.073) {\includegraphics[width=2.827cm]{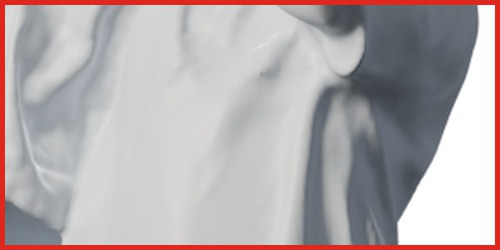}};
\node[cell] at (5.733,-5.775) {\includegraphics[width=2.827cm]{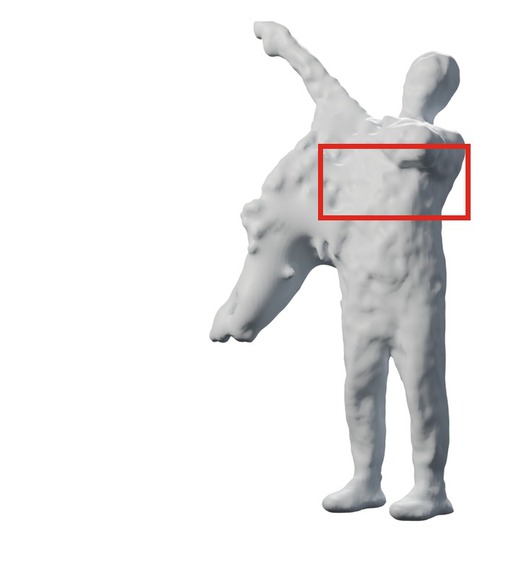}};
\node[cell] at (5.733,-9.073) {\includegraphics[width=2.827cm]{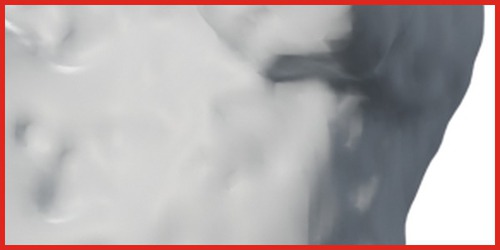}};
\node[cell] at (8.600,-5.775) {\includegraphics[width=2.827cm]{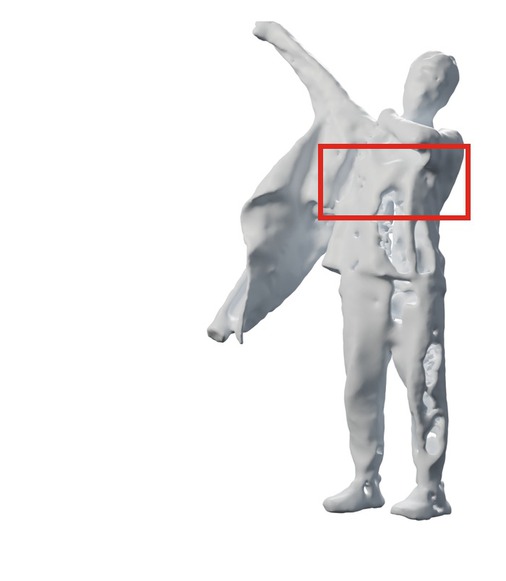}};
\node[cell] at (8.600,-9.073) {\includegraphics[width=2.827cm]{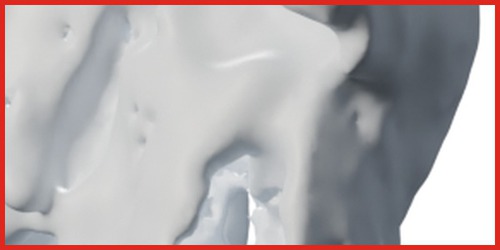}};
\node[cell] at (11.467,-5.775) {\includegraphics[width=2.827cm]{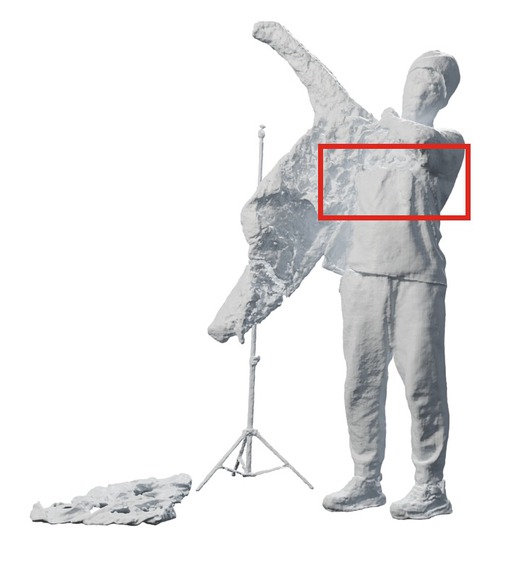}};
\node[cell] at (11.467,-9.073) {\includegraphics[width=2.827cm]{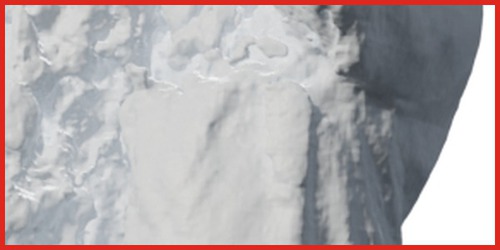}};
\node[cell] at (14.453,-5.775) {\includegraphics[width=2.827cm]{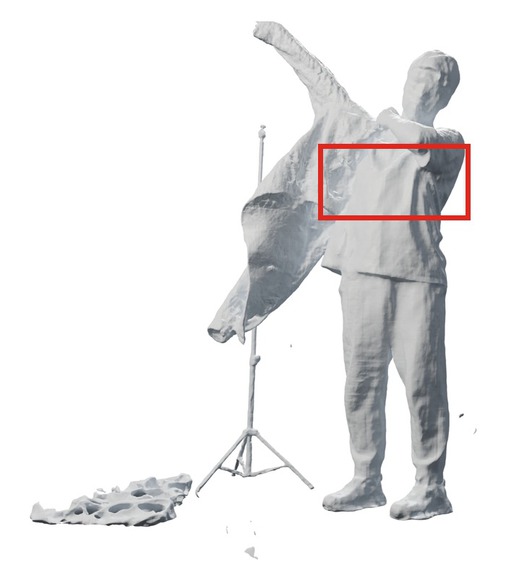}};
\node[cell] at (14.453,-9.073) {\includegraphics[width=2.827cm]{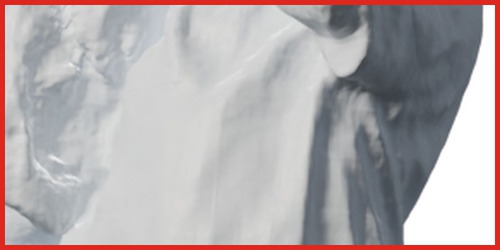}};
\node[anchor=north west, inner sep=1pt, fill=white, font=\scriptsize\itshape] at (0.000,-10.726) {TaoGS umbrella};
\node[cell] at (0.000,-11.066) {\includegraphics[width=2.827cm]{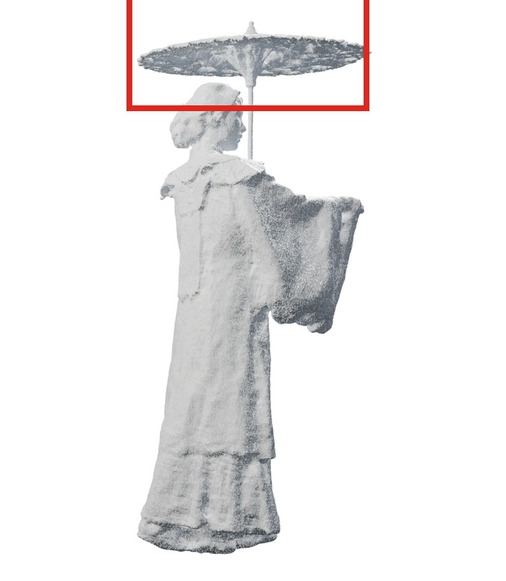}};
\node[cell] at (0.000,-14.369) {\includegraphics[width=2.827cm]{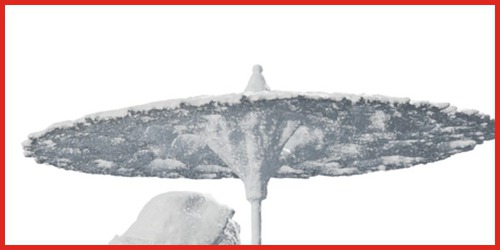}};
\node[cell] at (2.867,-11.066) {\includegraphics[width=2.827cm]{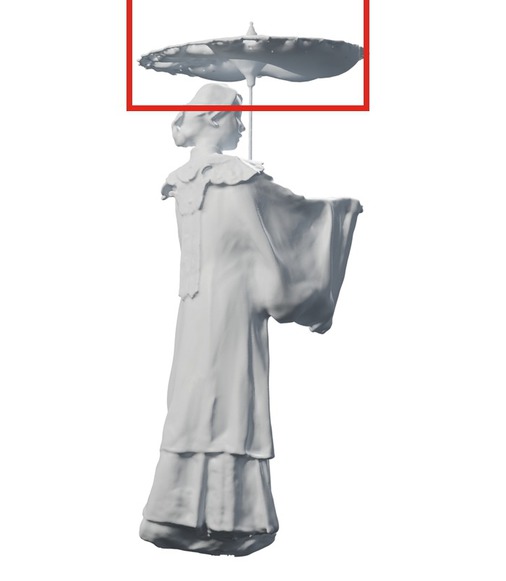}};
\node[cell] at (2.867,-14.369) {\includegraphics[width=2.827cm]{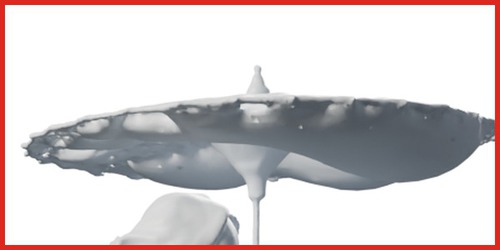}};
\node[cell] at (5.733,-11.066) {\includegraphics[width=2.827cm]{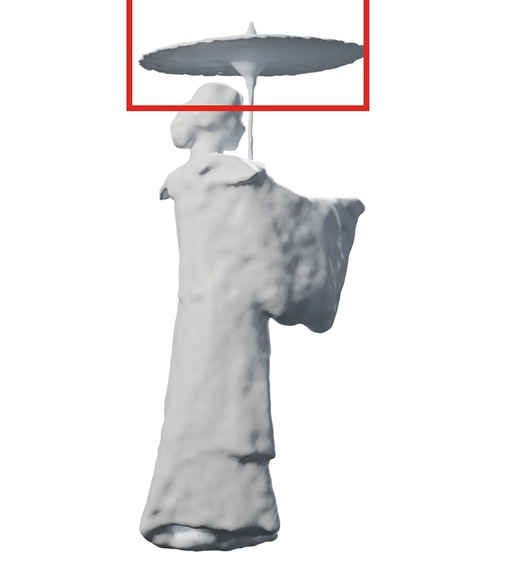}};
\node[cell] at (5.733,-14.369) {\includegraphics[width=2.827cm]{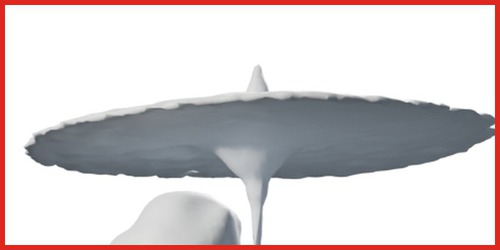}};
\node[cell] at (8.600,-11.066) {\includegraphics[width=2.827cm]{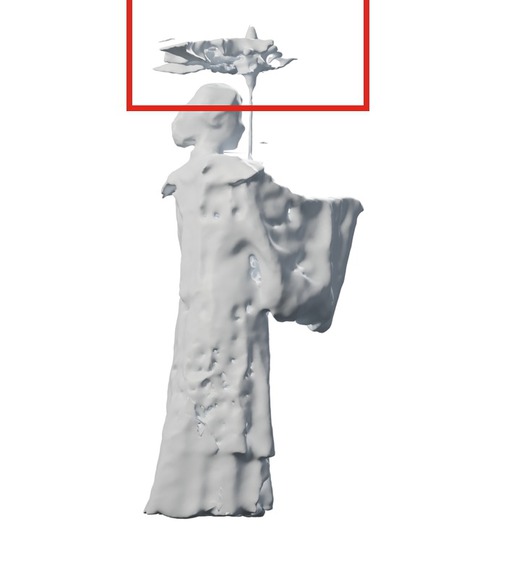}};
\node[cell] at (8.600,-14.369) {\includegraphics[width=2.827cm]{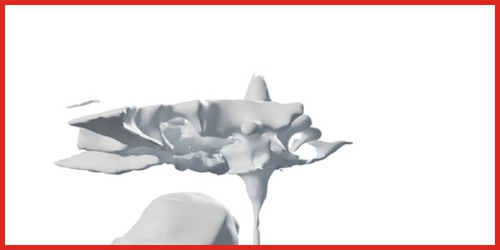}};
\node[cell] at (11.467,-11.066) {\includegraphics[width=2.827cm]{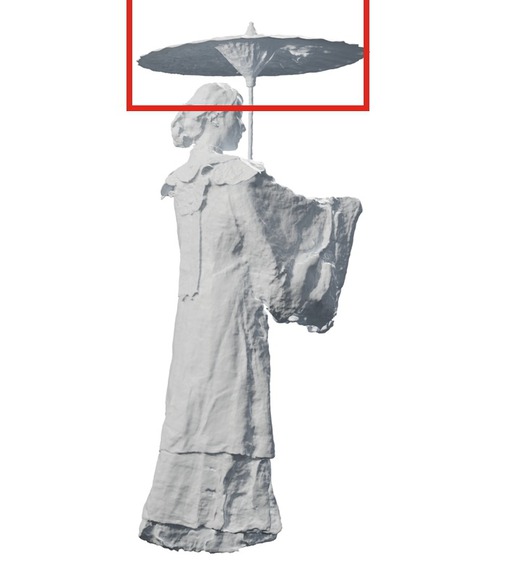}};
\node[cell] at (11.467,-14.369) {\includegraphics[width=2.827cm]{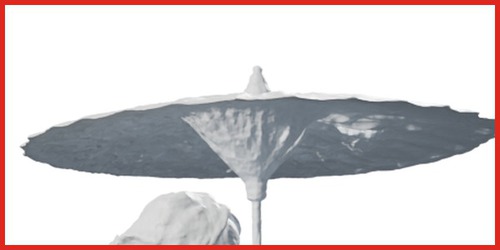}};
\node[cell] at (14.453,-11.066) {\includegraphics[width=2.827cm]{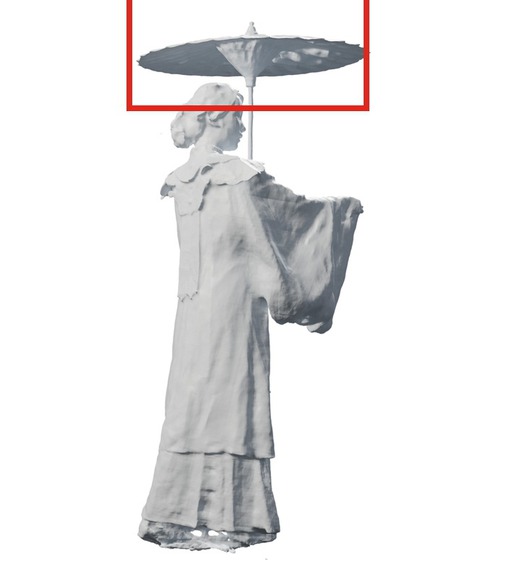}};
\node[cell] at (14.453,-14.369) {\includegraphics[width=2.827cm]{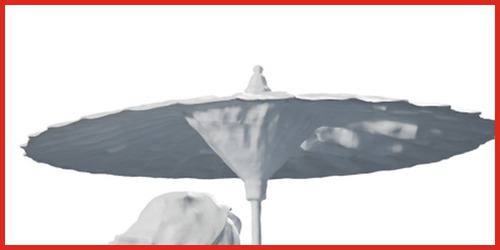}};
\node[anchor=north west, inner sep=1pt, fill=white, font=\scriptsize\itshape] at (0.000,-16.022) {TaoGS bag};
\node[cell] at (0.000,-16.362) {\includegraphics[width=2.827cm]{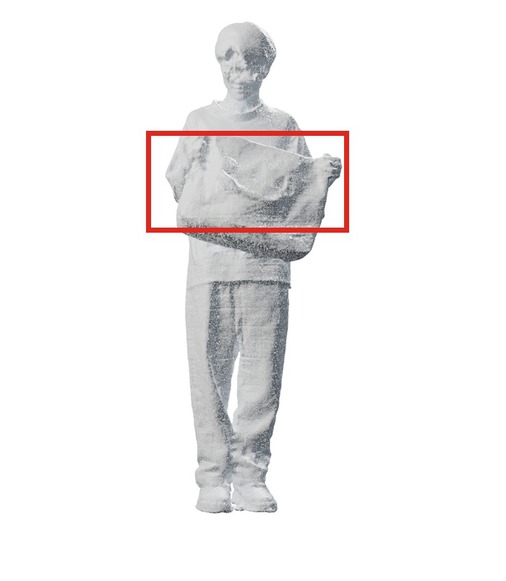}};
\node[cell] at (0.000,-19.660) {\includegraphics[width=2.827cm]{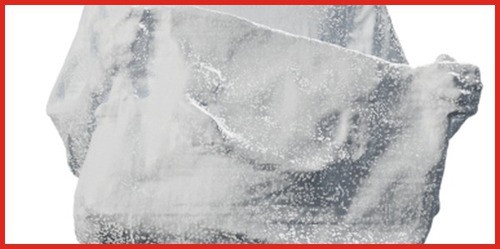}};
\node[cell] at (2.867,-16.362) {\includegraphics[width=2.827cm]{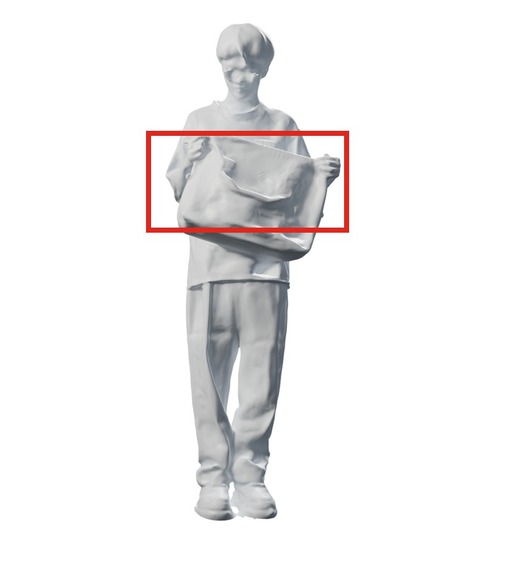}};
\node[cell] at (2.867,-19.660) {\includegraphics[width=2.827cm]{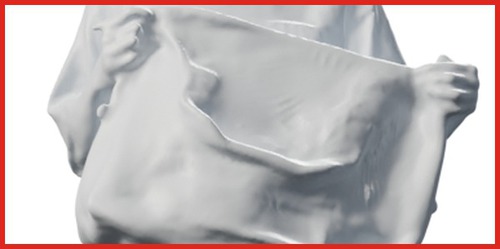}};
\node[cell] at (5.733,-16.362) {\includegraphics[width=2.827cm]{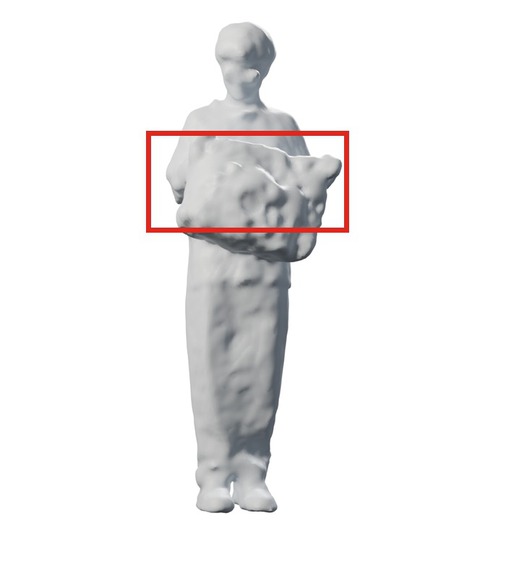}};
\node[cell] at (5.733,-19.660) {\includegraphics[width=2.827cm]{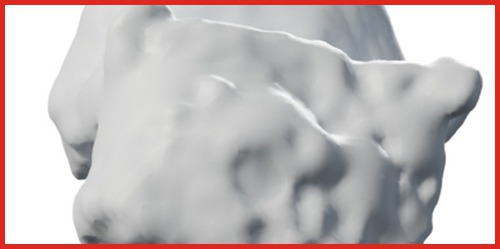}};
\node[cell] at (8.600,-16.362) {\includegraphics[width=2.827cm]{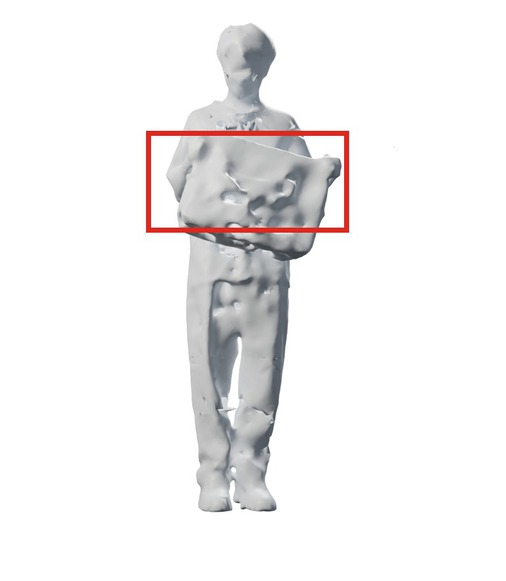}};
\node[cell] at (8.600,-19.660) {\includegraphics[width=2.827cm]{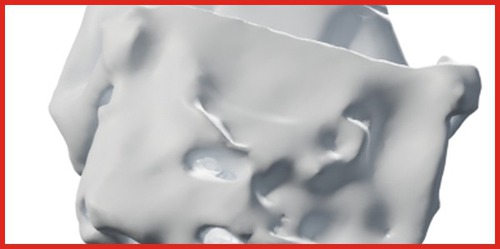}};
\node[cell] at (11.467,-16.362) {\includegraphics[width=2.827cm]{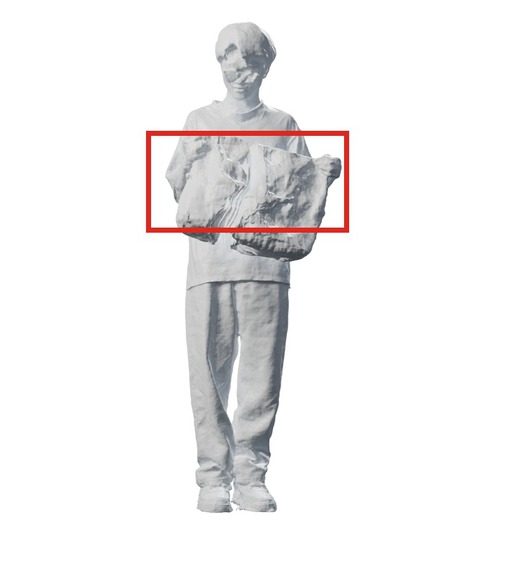}};
\node[cell] at (11.467,-19.660) {\includegraphics[width=2.827cm]{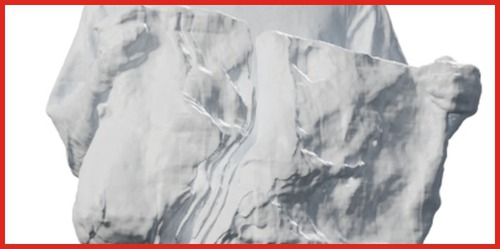}};
\node[cell] at (14.453,-16.362) {\includegraphics[width=2.827cm]{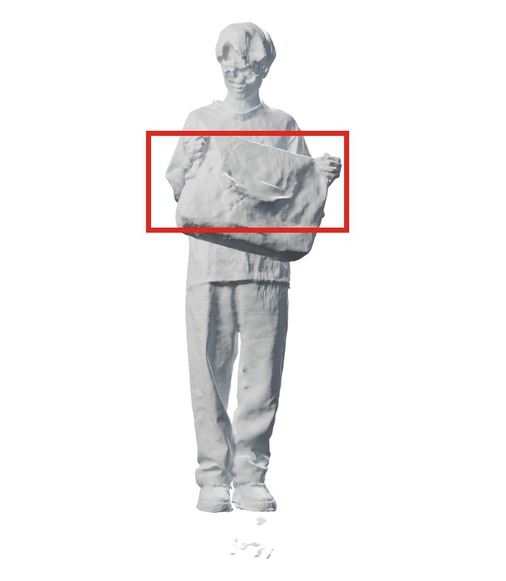}};
\node[cell] at (14.453,-19.660) {\includegraphics[width=2.827cm]{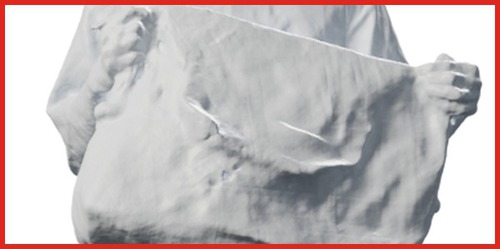}};
\end{tikzpicture}
\caption{\textbf{Geometry compared with the baselines on TaoGS} (real captures, no ground-truth geometry). Baselines shown: NeuS2, AT-GS, DG-Mesh, GauSTAR and MoAngelo. The red box marks the enlarged region (same box for every method).}
\label{fig:geo_cmp_3}
\end{figure*}

%% file: tables/ablation_detailed.tex
\begin{table}[t]
\centering
\caption{\textbf{Cumulative ablation on 6 ActorsHQ scenes}, CD (mm) and NC, complementing the ablation of \Cref{tab:ablation}, which removes one component at a time on 3 of these scenes. Each column removes one more component than the column to its left, so $-$ sampling here also removes $\mathcal{L}_{\mathrm{norm}}$ (earlier runs); on the 3 scenes of \Cref{tab:ablation} it gives 3.43\,mm, against 3.48\,mm when only the sampling is removed: without the sampling, $\mathcal{L}_{\mathrm{norm}}$ does not lower the Chamfer distance. \textbf{Bold}: best per scene.}
\label{tab:ablation_detailed}

\resizebox{\columnwidth}{!}{\begin{tabular}{c|cc|cc|cc}
\toprule
& \multicolumn{2}{c|}{\textbf{Ours} (full)} & \multicolumn{2}{c|}{$-$ $\mathcal{L}_{\mathrm{norm}}$} & \multicolumn{2}{c}{$-$ sampling} \\
\cmidrule(lr){2-3}
\cmidrule(lr){4-5}
\cmidrule(lr){6-7}
Scene & CD $\downarrow$ & NC $\uparrow$ & CD $\downarrow$ & NC $\uparrow$ & CD $\downarrow$ & NC $\uparrow$ \\
\midrule
actor4\_seq2\_1258 & 3.42 & \textbf{0.9335} & \textbf{3.40} & 0.9321 & 3.57 & 0.9294 \\
actor5\_seq2\_1942 & 3.10 & \textbf{0.9449} & \textbf{3.06} & 0.9427 & 3.21 & 0.9378 \\
actor6\_seq2\_1081 & 3.08 & \textbf{0.9297} & \textbf{3.06} & 0.9277 & 3.16 & 0.9258 \\
actor6\_seq2\_1604 & 3.22 & \textbf{0.9235} & \textbf{3.21} & 0.9210 & 3.34 & 0.9173 \\
actor7\_seq1\_554 & \textbf{2.65} & \textbf{0.9601} & 2.66 & 0.9594 & 2.69 & 0.9575 \\
actor8\_seq2\_325 & 3.28 & \textbf{0.9465} & \textbf{3.27} & 0.9444 & 3.37 & 0.9438 \\
\midrule
Average & 3.12 & \textbf{0.9397} & \textbf{3.11} & 0.9379 & 3.22 & 0.9353 \\
\bottomrule
\end{tabular}}
\end{table}

%% file: tables/seeds.tex
\begin{table}[t]
\centering
\addtocounter{table}{-1}\renewcommand{\thetable}{S\arabic{table}a}
\footnotesize
\setlength{\tabcolsep}{3pt}
\resizebox{\columnwidth}{!}{\begin{tabular}{@{}llccccc@{}}
\toprule
Scene & Variant & Metric & Seed 0 & Seed 1 & Seed 2 & Range \\
\midrule
ActorsHQ actor4\_1258 & Ours & CD$\downarrow$ & 3.42 & 3.37 & 3.38 & 0.04 \\
 & & NC$\uparrow$ & 0.934 & 0.935 & 0.935 & 0.002 \\
\midrule
DT4D vampire & Ours & CD$\downarrow$ & 11.85 & 11.89 & 11.84 & 0.05 \\
 & & NC$\uparrow$ & 0.837 & 0.835 & 0.835 & 0.002 \\
 & & NAE$\downarrow$ & 25.5 & 25.7 & 25.8 & 0.2 \\
 & & TCE$\downarrow$ & 5.23 & 5.27 & 5.21 & 0.06 \\
\cmidrule(l){2-7}
 & $-$ $\mathcal{L}_{\mathrm{norm}}$ & CD$\downarrow$ & 11.64 & 11.60 & 11.65 & 0.05 \\
 & & NC$\uparrow$ & 0.815 & 0.818 & 0.819 & 0.004 \\
 & & NAE$\downarrow$ & 27.7 & 27.5 & 27.3 & 0.4 \\
 & & TCE$\downarrow$ & 5.60 & 5.65 & 5.69 & 0.09 \\
\bottomrule
\end{tabular}}\caption{\textbf{Variation over random seeds.} Seed 0 is the run reported in the main paper; seeds 1 and 2 repeat it with the same configuration. Range: maximum minus minimum over the three seeds. CD and TCE in mm, NAE in degrees.}
\label{tab:seeds}
\end{table}

%% file: figures/ablation_closeups.tex
\begin{figure*}[t]
\centering\scriptsize\setlength{\tabcolsep}{0pt}\definecolor{tzOurs}{RGB}{14,128,229}\definecolor{mabA}{RGB}{232,46,130}
{\small\textbf{Geometry}}\par\smallskip
{\footnotesize\textbf{(a) Without the motion-guided sampling}}\par\smallskip
\begin{tikzpicture}[x=1in,y=1in,font=\scriptsize]
\node[anchor=south,inner sep=1pt] at (0.6672,0) {\strut \textcolor{tzOurs}{Ours}};
\node[anchor=north west,inner sep=0pt] at (0.0000,0) {\includegraphics[height=0.76in]{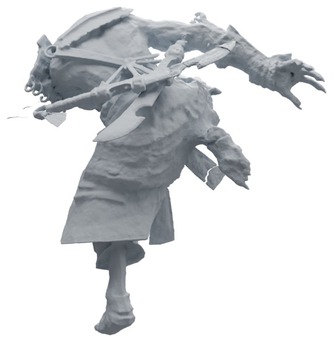}};
\draw[mabA,line width=0.5pt] (0.2696,-0.0892) rectangle (0.4566,-0.3386);
\node[anchor=north west,inner sep=0pt] at (0.7643,0) {\includegraphics[width=0.5700in]{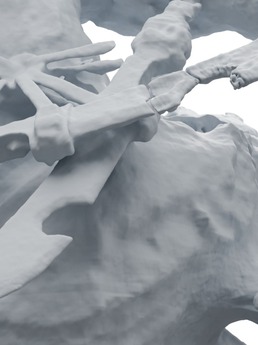}};
\draw[mabA,line width=0.6pt] (0.7643,0) rectangle (1.3343,-0.7600);
\node[anchor=south,inner sep=1pt] at (2.0715,0) {\strut $-$ sampling};
\node[anchor=north west,inner sep=0pt] at (1.4043,0) {\includegraphics[height=0.76in]{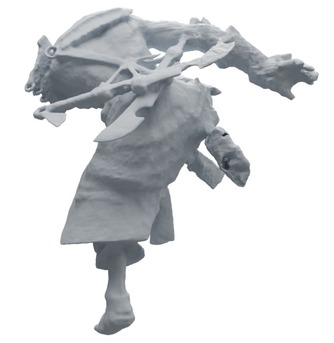}};
\draw[mabA,line width=0.5pt] (1.6739,-0.0892) rectangle (1.8609,-0.3386);
\node[anchor=north west,inner sep=0pt] at (2.1686,0) {\includegraphics[width=0.5700in]{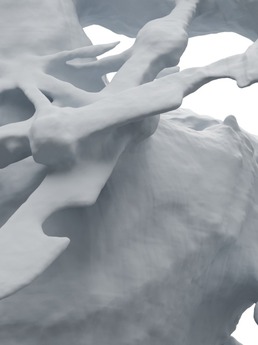}};
\draw[mabA,line width=0.6pt] (2.1686,0) rectangle (2.7386,-0.7600);
\node[anchor=north,inner sep=1pt] at (1.3693,-0.7800) {vampire, frame 17};
\end{tikzpicture}
\hfill
\begin{tikzpicture}[x=1in,y=1in,font=\scriptsize]
\node[anchor=south,inner sep=1pt] at (0.4605,0) {\strut \textcolor{tzOurs}{Ours}};
\node[anchor=north west,inner sep=0pt] at (0.0000,0) {\includegraphics[height=0.76in]{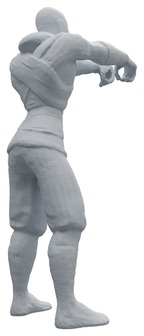}};
\draw[mabA,line width=0.5pt] (0.2129,-0.0830) rectangle (0.3174,-0.2223);
\node[anchor=north west,inner sep=0pt] at (0.3510,0) {\includegraphics[width=0.5700in]{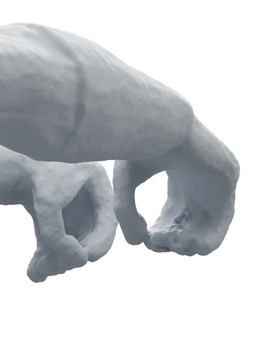}};
\draw[mabA,line width=0.6pt] (0.3510,0) rectangle (0.9210,-0.7600);
\node[anchor=south,inner sep=1pt] at (1.4515,0) {\strut $-$ sampling};
\node[anchor=north west,inner sep=0pt] at (0.9910,0) {\includegraphics[height=0.76in]{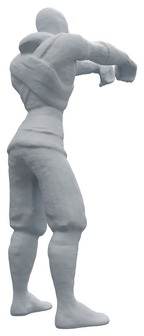}};
\draw[mabA,line width=0.5pt] (1.2040,-0.0830) rectangle (1.3084,-0.2223);
\node[anchor=north west,inner sep=0pt] at (1.3420,0) {\includegraphics[width=0.5700in]{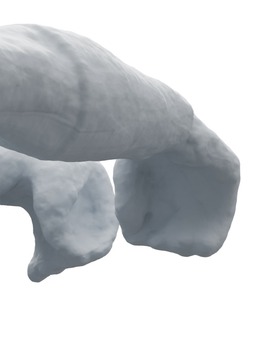}};
\draw[mabA,line width=0.6pt] (1.3420,0) rectangle (1.9120,-0.7600);
\node[anchor=north,inner sep=1pt] at (0.9560,-0.7800) {ninja, frame 33};
\end{tikzpicture}
\hfill
\begin{tikzpicture}[x=1in,y=1in,font=\scriptsize]
\node[anchor=south,inner sep=1pt] at (0.4559,0) {\strut \textcolor{tzOurs}{Ours}};
\node[anchor=north west,inner sep=0pt] at (0.0000,0) {\includegraphics[height=0.76in]{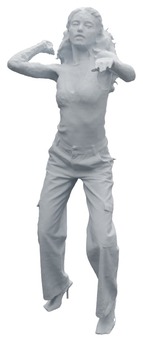}};
\draw[mabA,line width=0.5pt] (0.0147,-0.0621) rectangle (0.1183,-0.2003);
\node[anchor=north west,inner sep=0pt] at (0.3417,0) {\includegraphics[width=0.5700in]{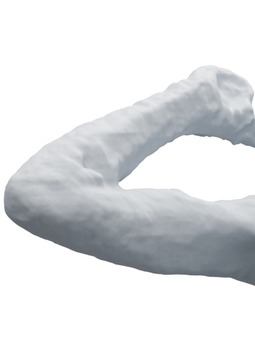}};
\draw[mabA,line width=0.6pt] (0.3417,0) rectangle (0.9117,-0.7600);
\node[anchor=south,inner sep=1pt] at (1.4376,0) {\strut $-$ sampling};
\node[anchor=north west,inner sep=0pt] at (0.9817,0) {\includegraphics[height=0.76in]{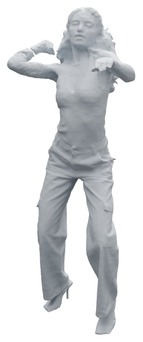}};
\draw[mabA,line width=0.5pt] (0.9964,-0.0621) rectangle (1.1001,-0.2003);
\node[anchor=north west,inner sep=0pt] at (1.3234,0) {\includegraphics[width=0.5700in]{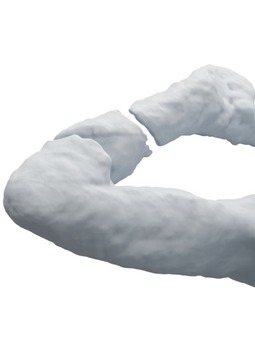}};
\draw[mabA,line width=0.6pt] (1.3234,0) rectangle (1.8934,-0.7600);
\node[anchor=north,inner sep=1pt] at (0.9467,-0.7800) {actor6\_seq2\_1604, frame 6};
\end{tikzpicture}
\par\medskip
\begin{tikzpicture}[x=1in,y=1in,font=\scriptsize]
\node[anchor=south,inner sep=1pt] at (0.5060,0) {\strut \textcolor{tzOurs}{Ours}};
\node[anchor=north west,inner sep=0pt] at (0.0000,0) {\includegraphics[height=0.76in]{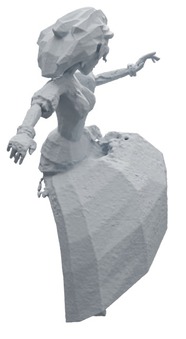}};
\draw[mabA,line width=0.5pt] (0.0101,-0.1895) rectangle (0.1925,-0.4327);
\node[anchor=north west,inner sep=0pt] at (0.4421,0) {\includegraphics[width=0.5700in]{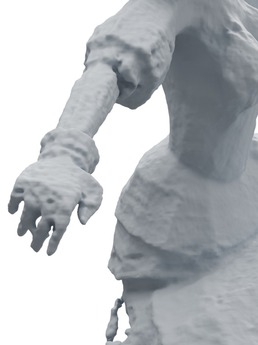}};
\draw[mabA,line width=0.6pt] (0.4421,0) rectangle (1.0121,-0.7600);
\node[anchor=south,inner sep=1pt] at (1.5881,0) {\strut $-$ sampling};
\node[anchor=north west,inner sep=0pt] at (1.0821,0) {\includegraphics[height=0.76in]{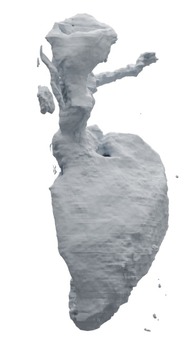}};
\draw[mabA,line width=0.5pt] (1.0922,-0.1895) rectangle (1.2746,-0.4327);
\node[anchor=north west,inner sep=0pt] at (1.5242,0) {\includegraphics[width=0.5700in]{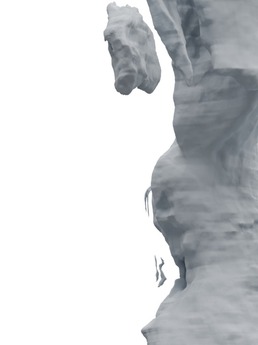}};
\draw[mabA,line width=0.6pt] (1.5242,0) rectangle (2.0942,-0.7600);
\node[anchor=north,inner sep=1pt] at (1.0471,-0.7800) {peasantgirl, frame 34};
\end{tikzpicture}
\hfill
\begin{tikzpicture}[x=1in,y=1in,font=\scriptsize]
\node[anchor=south,inner sep=1pt] at (0.4946,0) {\strut \textcolor{tzOurs}{Ours}};
\node[anchor=north west,inner sep=0pt] at (0.0000,0) {\includegraphics[height=0.76in]{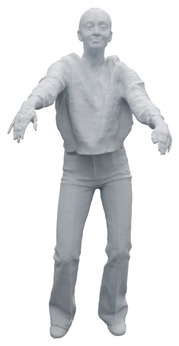}};
\draw[mabA,line width=0.5pt] (0.0078,-0.1970) rectangle (0.1099,-0.3331);
\node[anchor=north west,inner sep=0pt] at (0.4191,0) {\includegraphics[width=0.5700in]{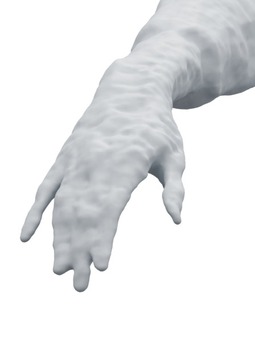}};
\draw[mabA,line width=0.6pt] (0.4191,0) rectangle (0.9891,-0.7600);
\node[anchor=south,inner sep=1pt] at (1.5537,0) {\strut $-$ sampling};
\node[anchor=north west,inner sep=0pt] at (1.0591,0) {\includegraphics[height=0.76in]{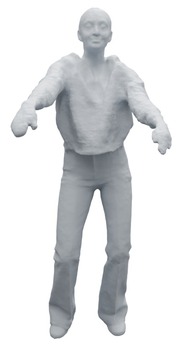}};
\draw[mabA,line width=0.5pt] (1.0669,-0.1970) rectangle (1.1690,-0.3331);
\node[anchor=north west,inner sep=0pt] at (1.4782,0) {\includegraphics[width=0.5700in]{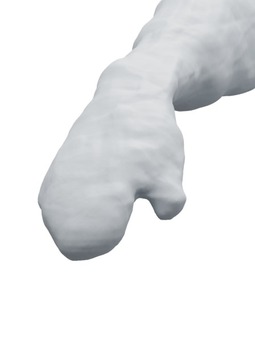}};
\draw[mabA,line width=0.6pt] (1.4782,0) rectangle (2.0482,-0.7600);
\node[anchor=north,inner sep=1pt] at (1.0241,-0.7800) {actor4\_seq2\_1258, frame 25};
\end{tikzpicture}
\hfill
\begin{tikzpicture}[x=1in,y=1in,font=\scriptsize]
\node[anchor=south,inner sep=1pt] at (0.4812,0) {\strut \textcolor{tzOurs}{Ours}};
\node[anchor=north west,inner sep=0pt] at (0.0000,0) {\includegraphics[height=0.76in]{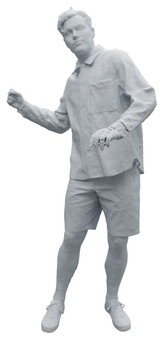}};
\draw[mabA,line width=0.5pt] (0.2063,-0.2706) rectangle (0.3074,-0.4055);
\node[anchor=north west,inner sep=0pt] at (0.3923,0) {\includegraphics[width=0.5700in]{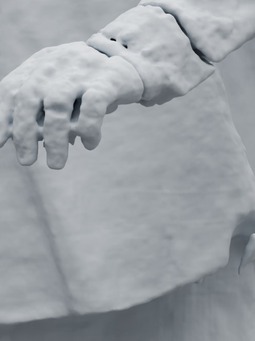}};
\draw[mabA,line width=0.6pt] (0.3923,0) rectangle (0.9623,-0.7600);
\node[anchor=south,inner sep=1pt] at (1.5135,0) {\strut $-$ sampling};
\node[anchor=north west,inner sep=0pt] at (1.0323,0) {\includegraphics[height=0.76in]{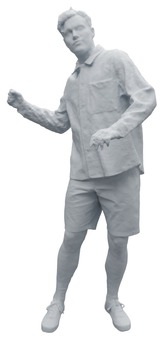}};
\draw[mabA,line width=0.5pt] (1.2386,-0.2706) rectangle (1.3397,-0.4055);
\node[anchor=north west,inner sep=0pt] at (1.4247,0) {\includegraphics[width=0.5700in]{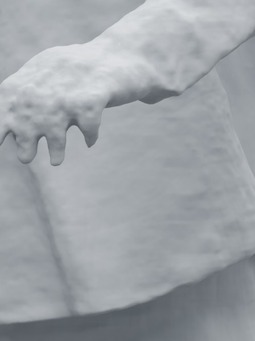}};
\draw[mabA,line width=0.6pt] (1.4247,0) rectangle (1.9947,-0.7600);
\node[anchor=north,inner sep=1pt] at (0.9973,-0.7800) {actor8\_seq2\_325, frame 6};
\end{tikzpicture}
\par\medskip
{\footnotesize\textbf{(b) Without $\mathcal{L}_{\mathrm{norm}}$}}\par\smallskip
\begin{tikzpicture}[x=1in,y=1in,font=\scriptsize]
\node[anchor=south,inner sep=1pt] at (0.6582,0) {\strut \textcolor{tzOurs}{Ours}};
\node[anchor=north west,inner sep=0pt] at (0.0000,0) {\includegraphics[height=0.76in]{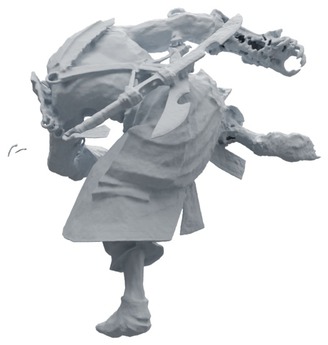}};
\draw[mabA,line width=0.5pt] (0.3233,-0.1036) rectangle (0.5091,-0.3514);
\node[anchor=north west,inner sep=0pt] at (0.7465,0) {\includegraphics[width=0.5700in]{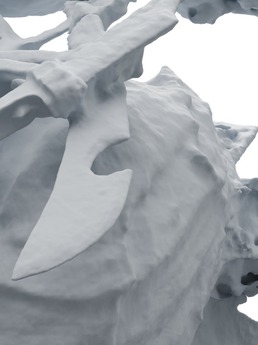}};
\draw[mabA,line width=0.6pt] (0.7465,0) rectangle (1.3165,-0.7600);
\node[anchor=south,inner sep=1pt] at (2.0447,0) {\strut $-\mathcal{L}_{\mathrm{norm}}$};
\node[anchor=north west,inner sep=0pt] at (1.3865,0) {\includegraphics[height=0.76in]{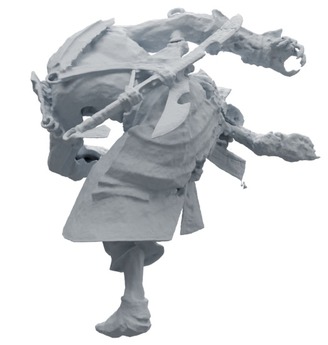}};
\draw[mabA,line width=0.5pt] (1.7097,-0.1036) rectangle (1.8956,-0.3514);
\node[anchor=north west,inner sep=0pt] at (2.1329,0) {\includegraphics[width=0.5700in]{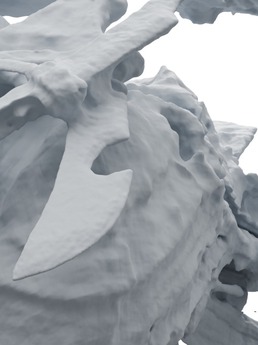}};
\draw[mabA,line width=0.6pt] (2.1329,0) rectangle (2.7029,-0.7600);
\node[anchor=north,inner sep=1pt] at (1.3515,-0.7800) {vampire, frame 25};
\end{tikzpicture}
\hfill
\begin{tikzpicture}[x=1in,y=1in,font=\scriptsize]
\node[anchor=south,inner sep=1pt] at (0.4774,0) {\strut \textcolor{tzOurs}{Ours}};
\node[anchor=north west,inner sep=0pt] at (0.0000,0) {\includegraphics[height=0.76in]{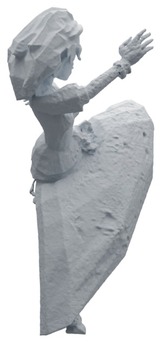}};
\draw[mabA,line width=0.5pt] (0.1646,-0.2094) rectangle (0.3551,-0.4635);
\node[anchor=north west,inner sep=0pt] at (0.3849,0) {\includegraphics[width=0.5700in]{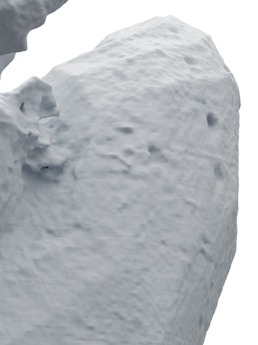}};
\draw[mabA,line width=0.6pt] (0.3849,0) rectangle (0.9549,-0.7600);
\node[anchor=south,inner sep=1pt] at (1.5023,0) {\strut $-\mathcal{L}_{\mathrm{norm}}$};
\node[anchor=north west,inner sep=0pt] at (1.0249,0) {\includegraphics[height=0.76in]{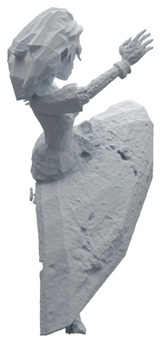}};
\draw[mabA,line width=0.5pt] (1.1894,-0.2094) rectangle (1.3800,-0.4635);
\node[anchor=north west,inner sep=0pt] at (1.4098,0) {\includegraphics[width=0.5700in]{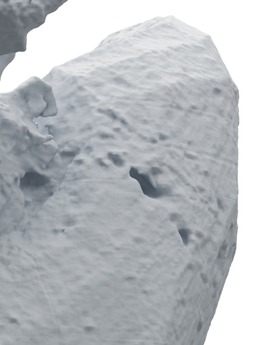}};
\draw[mabA,line width=0.6pt] (1.4098,0) rectangle (1.9798,-0.7600);
\node[anchor=north,inner sep=1pt] at (0.9899,-0.7800) {peasantgirl, frame 16};
\end{tikzpicture}
\hfill
\begin{tikzpicture}[x=1in,y=1in,font=\scriptsize]
\node[anchor=south,inner sep=1pt] at (0.4608,0) {\strut \textcolor{tzOurs}{Ours}};
\node[anchor=north west,inner sep=0pt] at (0.0000,0) {\includegraphics[height=0.76in]{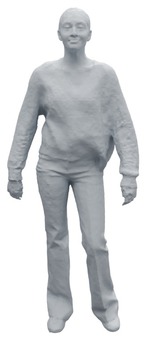}};
\draw[mabA,line width=0.5pt] (0.0160,-0.1712) rectangle (0.2069,-0.4258);
\node[anchor=north west,inner sep=0pt] at (0.3517,0) {\includegraphics[width=0.5700in]{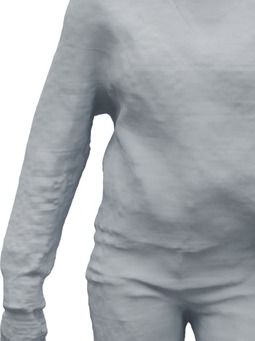}};
\draw[mabA,line width=0.6pt] (0.3517,0) rectangle (0.9217,-0.7600);
\node[anchor=south,inner sep=1pt] at (1.4525,0) {\strut $-\mathcal{L}_{\mathrm{norm}}$};
\node[anchor=north west,inner sep=0pt] at (0.9917,0) {\includegraphics[height=0.76in]{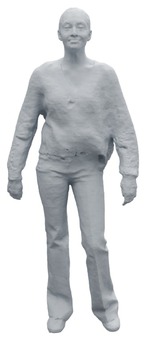}};
\draw[mabA,line width=0.5pt] (1.0077,-0.1712) rectangle (1.1986,-0.4258);
\node[anchor=north west,inner sep=0pt] at (1.3434,0) {\includegraphics[width=0.5700in]{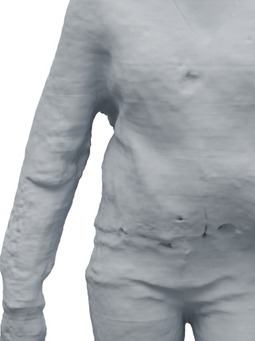}};
\draw[mabA,line width=0.6pt] (1.3434,0) rectangle (1.9134,-0.7600);
\node[anchor=north,inner sep=1pt] at (0.9567,-0.7800) {actor4\_seq2\_1258, frame 43};
\end{tikzpicture}
\par\medskip
\begin{tikzpicture}[x=1in,y=1in,font=\scriptsize]
\node[anchor=south,inner sep=1pt] at (0.4063,0) {\strut \textcolor{tzOurs}{Ours}};
\node[anchor=north west,inner sep=0pt] at (0.0000,0) {\includegraphics[height=0.76in]{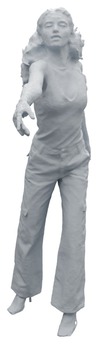}};
\draw[mabA,line width=0.5pt] (0.0187,-0.0819) rectangle (0.2089,-0.3356);
\node[anchor=north west,inner sep=0pt] at (0.2425,0) {\includegraphics[width=0.5700in]{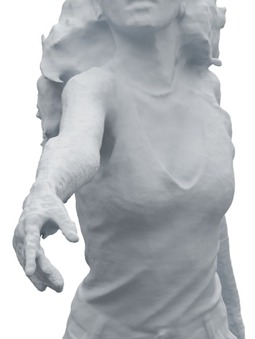}};
\draw[mabA,line width=0.6pt] (0.2425,0) rectangle (0.8125,-0.7600);
\node[anchor=south,inner sep=1pt] at (1.2888,0) {\strut $-\mathcal{L}_{\mathrm{norm}}$};
\node[anchor=north west,inner sep=0pt] at (0.8825,0) {\includegraphics[height=0.76in]{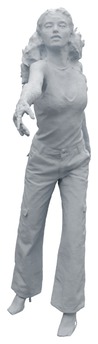}};
\draw[mabA,line width=0.5pt] (0.9012,-0.0819) rectangle (1.0914,-0.3356);
\node[anchor=north west,inner sep=0pt] at (1.1250,0) {\includegraphics[width=0.5700in]{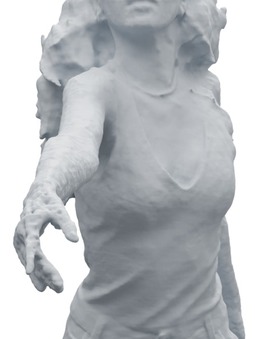}};
\draw[mabA,line width=0.6pt] (1.1250,0) rectangle (1.6950,-0.7600);
\node[anchor=north,inner sep=1pt] at (0.8475,-0.7800) {actor6\_seq2\_1604, frame 25};
\end{tikzpicture}
\hfill
\begin{tikzpicture}[x=1in,y=1in,font=\scriptsize]
\node[anchor=south,inner sep=1pt] at (0.4292,0) {\strut \textcolor{tzOurs}{Ours}};
\node[anchor=north west,inner sep=0pt] at (0.0000,0) {\includegraphics[height=0.76in]{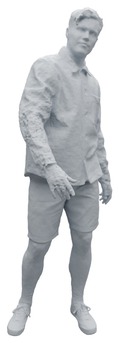}};
\draw[mabA,line width=0.5pt] (0.0257,-0.1776) rectangle (0.2174,-0.4333);
\node[anchor=north west,inner sep=0pt] at (0.2884,0) {\includegraphics[width=0.5700in]{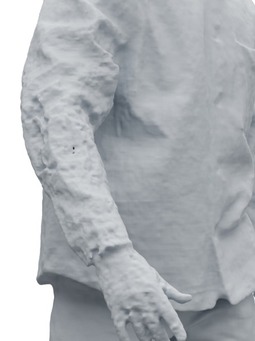}};
\draw[mabA,line width=0.6pt] (0.2884,0) rectangle (0.8584,-0.7600);
\node[anchor=south,inner sep=1pt] at (1.3576,0) {\strut $-\mathcal{L}_{\mathrm{norm}}$};
\node[anchor=north west,inner sep=0pt] at (0.9284,0) {\includegraphics[height=0.76in]{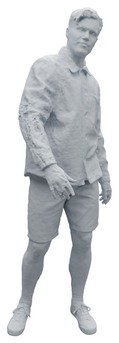}};
\draw[mabA,line width=0.5pt] (0.9541,-0.1776) rectangle (1.1458,-0.4333);
\node[anchor=north west,inner sep=0pt] at (1.2168,0) {\includegraphics[width=0.5700in]{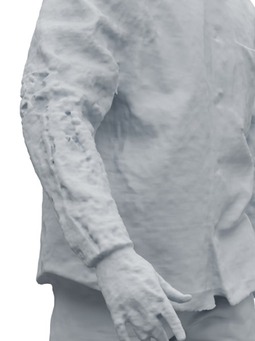}};
\draw[mabA,line width=0.6pt] (1.2168,0) rectangle (1.7868,-0.7600);
\node[anchor=north,inner sep=1pt] at (0.8934,-0.7800) {actor8\_seq2\_325, frame 47};
\end{tikzpicture}
\hfill
\begin{tikzpicture}[x=1in,y=1in,font=\scriptsize]
\node[anchor=south,inner sep=1pt] at (0.6532,0) {\strut \textcolor{tzOurs}{Ours}};
\node[anchor=north west,inner sep=0pt] at (0.0000,0) {\includegraphics[height=0.76in]{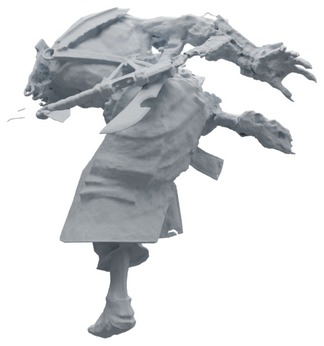}};
\draw[mabA,line width=0.5pt] (0.2845,-0.0809) rectangle (0.4743,-0.3340);
\node[anchor=north west,inner sep=0pt] at (0.7365,0) {\includegraphics[width=0.5700in]{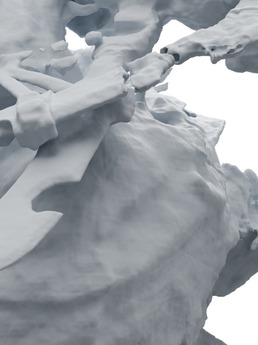}};
\draw[mabA,line width=0.6pt] (0.7365,0) rectangle (1.3065,-0.7600);
\node[anchor=south,inner sep=1pt] at (2.0297,0) {\strut $-\mathcal{L}_{\mathrm{norm}}$};
\node[anchor=north west,inner sep=0pt] at (1.3765,0) {\includegraphics[height=0.76in]{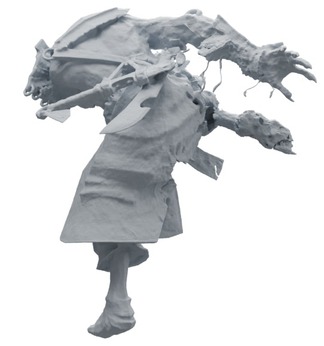}};
\draw[mabA,line width=0.5pt] (1.6610,-0.0809) rectangle (1.8508,-0.3340);
\node[anchor=north west,inner sep=0pt] at (2.1129,0) {\includegraphics[width=0.5700in]{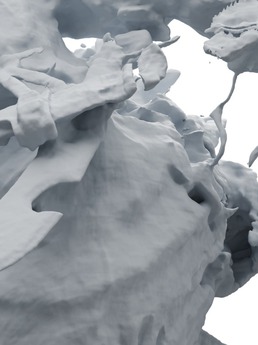}};
\draw[mabA,line width=0.6pt] (2.1129,0) rectangle (2.6829,-0.7600);
\node[anchor=north,inner sep=1pt] at (1.3415,-0.7800) {vampire, frame 18};
\end{tikzpicture}
\par\medskip
{\small\textbf{Correspondences}}\par\smallskip{\footnotesize\textbf{(c) Without $\mathcal{L}_{\mathrm{flow}}$}}\par\smallskip
\begin{tabular}{@{}c@{\hspace{2pt}}cc@{\hspace{14pt}}c@{\hspace{2pt}}cc@{}}
vampire, frame 1 & \textcolor{tzOurs}{Ours}, frame 22 & $-\mathcal{L}_{\mathrm{flow}}$, frame 22 & ninja, frame 1 & \textcolor{tzOurs}{Ours}, frame 26 & $-\mathcal{L}_{\mathrm{flow}}$, frame 26\\
\includegraphics[width=0.135\textwidth]{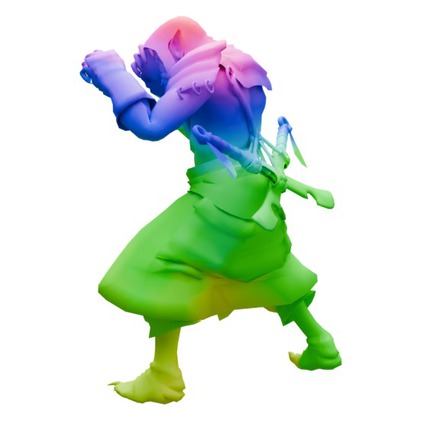} & \includegraphics[width=0.135\textwidth]{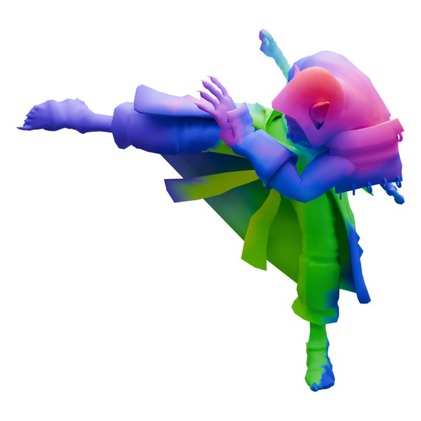} & \includegraphics[width=0.135\textwidth]{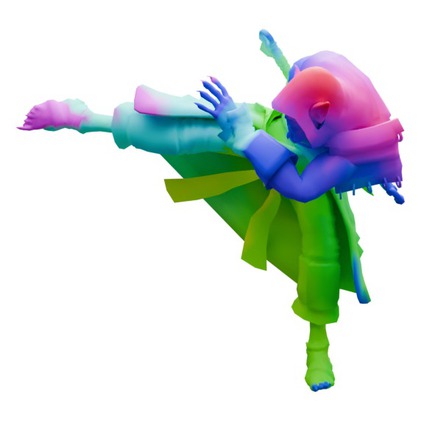} & \includegraphics[width=0.135\textwidth]{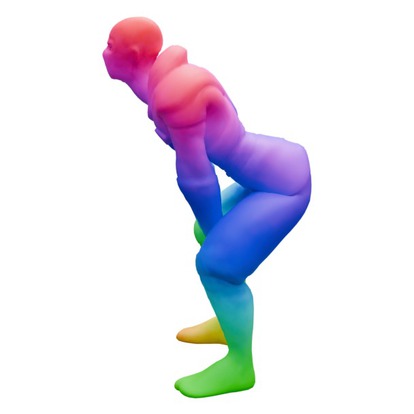} & \includegraphics[width=0.135\textwidth]{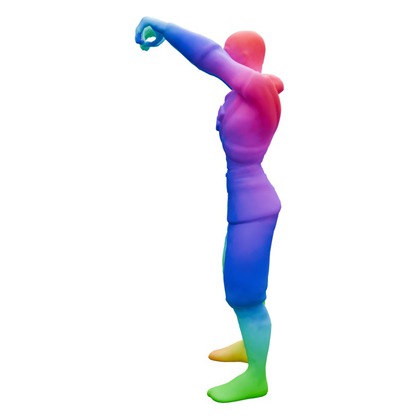} & \includegraphics[width=0.135\textwidth]{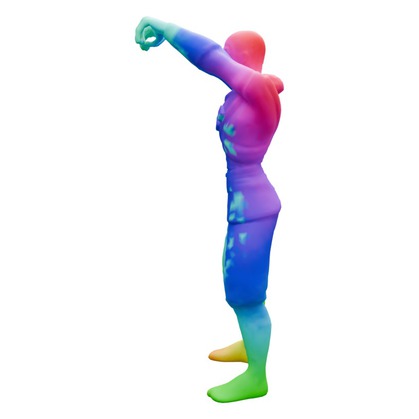}\\
\begin{tabular}[b]{@{}c@{}}geodesic error ($\times$100)\\[1pt]\begin{tikzpicture}\shade[left color={rgb,255:red,255;green,247;blue,189}, middle color={rgb,255:red,252;green,153;blue,36}, right color={rgb,255:red,115;green,13;blue,26}] (0,0) rectangle (1.8,0.12);\node[anchor=north,font=\tiny,inner sep=1pt] at (0,0){0};\node[anchor=north,font=\tiny,inner sep=1pt] at (1.8,0){$\geq$10};\end{tikzpicture}\\[14pt]\end{tabular} & \begin{tikzpicture}\node[inner sep=0]{\includegraphics[width=0.135\textwidth]{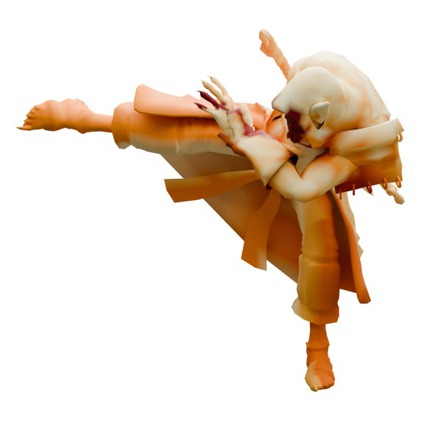}};\node[anchor=south east,font=\tiny,inner sep=1pt] at (current bounding box.south east){\textcolor{tzOurs}{5.81}};\end{tikzpicture} & \begin{tikzpicture}\node[inner sep=0]{\includegraphics[width=0.135\textwidth]{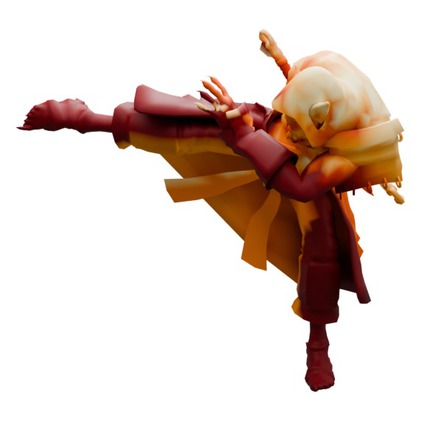}};\node[anchor=south east,font=\tiny,inner sep=1pt] at (current bounding box.south east){8.63};\end{tikzpicture} &  & \begin{tikzpicture}\node[inner sep=0]{\includegraphics[width=0.135\textwidth]{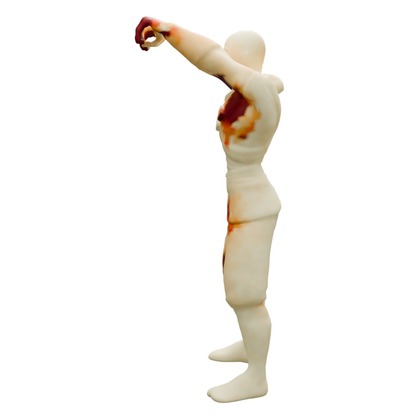}};\node[anchor=south east,font=\tiny,inner sep=1pt] at (current bounding box.south east){\textcolor{tzOurs}{4.40}};\end{tikzpicture} & \begin{tikzpicture}\node[inner sep=0]{\includegraphics[width=0.135\textwidth]{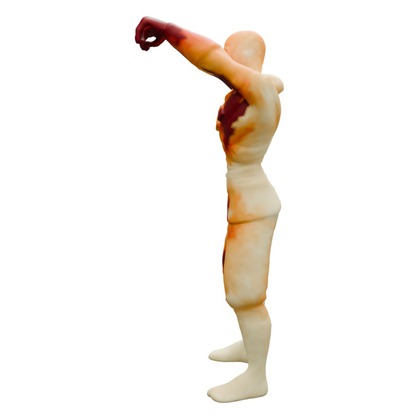}};\node[anchor=south east,font=\tiny,inner sep=1pt] at (current bounding box.south east){8.51};\end{tikzpicture}\\
\end{tabular}\par
\caption{\textbf{Ablation of the geometry and the correspondences} (\cref{tab:ablation}). Geometry: full meshes of our model and of the variant without (a) the motion-guided sampling or (b) $\mathcal{L}_{\mathrm{norm}}$, from the same camera, and close-ups of the boxed regions (those shown in the video on the project page). Correspondences (c): points of frame~1 carried to later frames by our model and by the variant without $\mathcal{L}_{\mathrm{flow}}$, colored by their frame-1 position (top) and by their geodesic error ($\times 100$, mean in the corner, bottom); \cref{fig:corr_fig} of the paper shows peasantgirl. ActorsHQ has no ground-truth correspondences.}
\label{fig:ablation_closeups}
\end{figure*}

%% file: tables/nvs.tex
\begin{table}[t]
\centering
\footnotesize
\setlength{\tabcolsep}{2.5pt}
\resizebox{\columnwidth}{!}{\begin{tabular}{@{}lccc|ccc@{}}
\toprule
 & \multicolumn{3}{c|}{ActorsHQ} & \multicolumn{3}{c}{DeformingThings4D} \\
\cmidrule(lr){2-4}\cmidrule(l){5-7}
Method & PSNR$\uparrow$ & SSIM$\uparrow$ & LPIPS$\downarrow$ & PSNR$\uparrow$ & SSIM$\uparrow$ & LPIPS$\downarrow$ \\
\midrule
\multicolumn{7}{@{}l}{\emph{Dynamic, SDF-based}} \\
Tensor4D & 26.14 & 0.925 & 0.135 & 25.91 & 0.958 & 0.073 \\
NeuS2 & 31.73 & 0.965 & 0.052 & 32.75 & 0.984 & 0.034 \\
MoAngelo & 32.28 & \rankC{0.968} & \rankC{0.045} & 32.09 & 0.982 & \rankC{0.032} \\
\midrule
\multicolumn{7}{@{}l}{\emph{Dynamic, NeRF-based}} \\
SceNeRFlow & 22.10 & 0.900 & 0.145 & 22.60 & 0.940 & 0.095 \\
\midrule
\multicolumn{7}{@{}l}{\emph{Dynamic, Gaussian-based}} \\
Dynamic 3DGS & 30.55 & 0.951 & 0.115 & 28.87 & 0.965 & 0.069 \\
Dynamic-2DGS & 20.27$^\dagger$ & 0.873$^\dagger$ & 0.221$^\dagger$ & 29.50$^\dagger$ & 0.977$^\dagger$ & 0.052$^\dagger$ \\
AT-GS & \rankC{32.31} & 0.964 & 0.080 & \rankB{34.40} & \rankC{0.987} & 0.036 \\
4DGS & 29.19 & 0.940 & 0.114 & 30.19 & 0.972 & 0.067 \\
SC-GS & 28.82 & 0.947 & 0.105 & 32.84 & 0.982 & 0.044 \\
Deformable 3DGS & 28.98 & 0.947 & 0.106 & 32.01$^\dagger$ & 0.980$^\dagger$ & 0.048$^\dagger$ \\
\midrule
\multicolumn{7}{@{}l}{\emph{Dynamic, mesh-based Gaussian}} \\
GauSTAR & 26.89 & 0.955 & 0.071 & 25.43 & 0.963 & 0.075 \\
DG-Mesh (mesh) & 26.28 & 0.937 & 0.095 & 26.86$^\dagger$ & 0.969$^\dagger$ & 0.052$^\dagger$ \\
DG-Mesh (Gaussians) & 19.21 & 0.575 & 0.239 & 23.18$^\dagger$ & 0.830$^\dagger$ & 0.117$^\dagger$ \\
MaGS & 29.44 & 0.950 & 0.086 & 32.67$^\dagger$ & 0.983$^\dagger$ & 0.037$^\dagger$ \\
\midrule
\multicolumn{7}{@{}l}{\emph{Per-frame static}} \\
2DGS & 32.30 & 0.967 & 0.036 & 35.50 & 0.990 & 0.024 \\
PGSR & 30.25 & 0.961 & 0.041 & 30.36 & 0.978 & 0.042 \\
\midrule
\textbf{Ours} (Lite) & \rankB{33.44} & \rankB{0.974} & \rankB{0.036} & \rankC{33.97} & \rankB{0.988} & \rankB{0.021} \\
\textbf{Ours} & \rankA{33.71} & \rankA{0.975} & \rankA{0.035} & \rankA{34.54} & \rankA{0.989} & \rankA{0.019} \\
\bottomrule
\end{tabular}}
\caption{\textbf{Novel-view synthesis} on held-out cameras (ActorsHQ: 6 per scene, 6 scenes; DeformingThings4D: 5 per scene, 10 scenes); PSNR in dB. \ranklegend{} among the dynamic methods. --: failed. DG-Mesh is shown both as renderings of its mesh and of its Gaussians. $^\dagger$Averaged over the scenes with a result: Dynamic-2DGS: 2 of 6 ActorsHQ scenes and 3 of 10 DeformingThings4D scenes; Deformable 3DGS: 9 of 10 DeformingThings4D scenes; DG-Mesh (mesh): 9 of 10 DeformingThings4D scenes; DG-Mesh (Gaussians): 9 of 10 DeformingThings4D scenes; MaGS: 9 of 10 DeformingThings4D scenes.}
\label{tab:nvs}
\end{table}

%% file: tables/perscene_psnr.tex
\begin{table*}[t]
\centering
\scriptsize
\setlength{\tabcolsep}{2.2pt}
\resizebox{\textwidth}{!}{\begin{tabular}{@{}lccccccc|ccccccccccc@{}}
\toprule
 & \multicolumn{7}{c|}{ActorsHQ} & \multicolumn{11}{c}{DeformingThings4D} \\
Method & a4-1258 & a5-1942 & a6-1081 & a6-1604 & a7-554 & a8-325 & Mean & abe & AJ & canie & dragon & fox & grizz & ninja & peasant. & rabbit & vampire & Mean \\
\midrule
\multicolumn{19}{@{}l}{\emph{Dynamic, SDF-based}} \\
Tensor4D & 25.05 & 26.05 & 26.30 & 25.77 & 29.03 & 24.66 & 26.14 & 27.05 & 29.11 & 24.97 & 16.38 & 24.43 & 28.29 & 30.00 & 28.11 & 23.09 & 27.69 & 25.91 \\
NeuS2 & 30.99 & 31.67 & 31.32 & 31.01 & 33.42 & \rankC{31.99} & 31.73 & 32.76 & 32.94 & 33.81 & 32.17 & 33.10 & 34.96 & 32.64 & 34.05 & 30.66 & 30.44 & 32.75 \\
MoAngelo & 31.38 & \rankB{33.50} & 32.35 & 31.29 & \rankC{34.02} & 31.13 & 32.28 & 31.33 & 32.52 & 31.20 & 31.12 & 32.60 & 34.12 & 33.05 & 32.72 & 31.85 & 30.41 & 32.09 \\
\multicolumn{19}{@{}l}{\emph{Dynamic, NeRF-based}} \\
SceNeRFlow & 20.02 & 22.44 & 21.46 & 20.34 & 26.82 & 21.50 & 22.10 & 23.30 & 23.93 & 19.57 & 20.76 & 20.33 & 23.02 & 25.13 & 25.21 & 18.62 & 26.10 & 22.60 \\
\multicolumn{19}{@{}l}{\emph{Dynamic, Gaussian-based}} \\
Dynamic 3DGS & 29.64 & 30.58 & 30.65 & 30.64 & 31.86 & 29.96 & 30.55 & 30.94 & 32.10 & 29.36 & 25.66 & 32.73 & 24.15 & 31.05 & 25.10 & 31.23 & 26.42 & 28.87 \\
Dynamic-2DGS & -- & -- & -- & -- & 22.46 & 18.09 & 20.27$^\dagger$ & 28.91 & 18.26$^\ast$ & -- & 28.53 & 31.04 & -- & -- & -- & -- & -- & 29.50$^\dagger$ \\
AT-GS & \rankC{31.65} & 32.43 & \rankC{32.58} & \rankC{32.26} & 33.15 & 31.79 & \rankC{32.31} & \rankC{33.18} & \rankC{34.35} & \rankA{35.76} & \rankA{35.25} & \rankA{35.81} & \rankB{36.01} & 32.11 & \rankB{35.30} & \rankA{34.95} & \rankC{31.27} & \rankB{34.40} \\
4DGS & 28.23 & 29.44 & 28.93 & 28.67 & 31.47 & 28.41 & 29.19 & 28.24 & 31.00 & 30.67 & 28.88 & 30.69 & 31.74 & 30.96 & 31.60 & 28.98 & 29.16 & 30.19 \\
SC-GS & 28.07 & 29.41 & 28.75 & 27.99 & 30.74 & 27.94 & 28.82 & 32.20 & \rankA{34.93} & 32.00 & \rankC{33.64} & 33.41 & 34.66 & \rankB{33.65} & 32.50 & 30.76 & 30.69 & 32.84 \\
Deformable 3DGS & 28.10 & 29.11 & 29.19 & 28.73 & 30.83 & 27.90 & 28.98 & -- & 32.70 & 31.50 & 32.47 & 32.95 & 34.54 & 32.70 & 31.11 & 30.91 & 29.21 & 32.01$^\dagger$ \\
\multicolumn{19}{@{}l}{\emph{Dynamic, mesh-based Gaussian}} \\
GauSTAR & 26.26 & 26.24 & 28.41 & 28.04 & 26.03 & 26.36 & 26.89 & 29.95 & 31.02 & 21.42 & 16.15 & 26.74 & 20.62 & 30.08 & 25.51 & 27.61 & 25.16 & 25.43 \\
MaGS & 27.64 & 29.18 & 30.60 & 29.84 & 31.13 & 28.27 & 29.44 & -- & 33.78 & 32.64 & 33.45 & 34.13 & \rankC{35.71} & 31.71 & 30.53 & 32.18 & 29.86 & 32.67$^\dagger$ \\
DG-Mesh (mesh) & 25.33 & 27.32 & 26.09 & 25.36 & 28.91 & 24.65 & 26.28 & -- & 29.34 & 23.22 & 25.91 & 26.68 & 29.50 & 29.26 & 28.18 & 23.49 & 26.18 & 26.86$^\dagger$ \\
DG-Mesh (Gaussians) & 20.28 & 18.76 & 19.74 & 19.11 & 18.53 & 18.83 & 19.21 & -- & 22.00 & 22.73 & 23.34 & 24.38 & 23.01 & 22.03 & 24.77 & 21.63 & 24.73 & 23.18$^\dagger$ \\
\multicolumn{19}{@{}l}{\emph{Per-frame static}} \\
2DGS & 31.32 & 32.02 & 32.61 & 32.56 & 33.23 & 32.07 & 32.30 & 33.98 & 35.45 & 36.74 & 35.08 & 36.48 & 37.04 & 33.49 & 36.89 & 36.61 & 33.26 & 35.50 \\
PGSR & 29.24 & 30.33 & 30.44 & 30.44 & 31.24 & 29.82 & 30.25 & 33.43 & 34.66 & 18.89 & 35.12 & 20.94 & 37.01 & 32.83 & 36.62 & 20.80 & 33.29 & 30.36 \\
\midrule
\textbf{Ours} (Lite) & \rankB{32.25} & \rankC{33.21} & \rankB{33.46} & \rankB{33.11} & \rankA{35.44} & \rankB{33.16} & \rankB{33.44} & \rankB{33.43} & 34.23 & \rankC{34.53} & 32.89 & \rankC{34.66} & 35.29 & \rankC{33.16} & \rankC{35.16} & \rankC{33.99} & \rankB{32.40} & \rankC{33.97} \\
\textbf{Ours} & \rankA{32.70} & \rankA{33.59} & \rankA{33.75} & \rankA{33.51} & \rankB{35.43} & \rankA{33.31} & \rankA{33.71} & \rankA{33.69} & \rankB{34.53} & \rankB{35.24} & \rankB{33.91} & \rankB{35.50} & \rankA{36.32} & \rankA{33.66} & \rankA{35.57} & \rankB{34.32} & \rankA{32.68} & \rankA{34.54} \\
\bottomrule
\end{tabular}}
\caption{\textbf{Per-scene novel-view synthesis, PSNR} (dB, $\uparrow$) on the held-out cameras. \ranklegend{} among the dynamic methods; --: failed; $^\dagger$mean over the scenes with a result; $^\ast$failed run, left out of the mean. DG-Mesh is rendered both from its mesh and from its Gaussians.}
\label{tab:perscene_psnr}
\end{table*}

%% file: tables/perscene_ssim.tex
\begin{table*}[t]
\centering
\scriptsize
\setlength{\tabcolsep}{2.2pt}
\resizebox{\textwidth}{!}{\begin{tabular}{@{}lccccccc|ccccccccccc@{}}
\toprule
 & \multicolumn{7}{c|}{ActorsHQ} & \multicolumn{11}{c}{DeformingThings4D} \\
Method & a4-1258 & a5-1942 & a6-1081 & a6-1604 & a7-554 & a8-325 & Mean & abe & AJ & canie & dragon & fox & grizz & ninja & peasant. & rabbit & vampire & Mean \\
\midrule
\multicolumn{19}{@{}l}{\emph{Dynamic, SDF-based}} \\
Tensor4D & 0.920 & 0.917 & 0.940 & 0.930 & 0.939 & 0.906 & 0.925 & 0.945 & 0.956 & 0.978 & 0.921 & 0.978 & 0.986 & 0.960 & 0.949 & 0.967 & 0.938 & 0.958 \\
NeuS2 & 0.956 & 0.961 & 0.969 & 0.968 & \rankC{0.969} & \rankB{0.965} & 0.965 & 0.977 & 0.978 & \rankC{0.993} & 0.989 & 0.993 & \rankC{0.994} & 0.974 & \rankB{0.983} & 0.988 & 0.974 & 0.984 \\
MoAngelo & \rankC{0.957} & \rankC{0.969} & \rankC{0.975} & \rankC{0.972} & \rankB{0.974} & \rankC{0.960} & \rankC{0.968} & 0.971 & 0.978 & 0.991 & 0.988 & 0.993 & \rankC{0.994} & \rankC{0.978} & 0.973 & \rankC{0.991} & 0.967 & 0.982 \\
\multicolumn{19}{@{}l}{\emph{Dynamic, NeRF-based}} \\
SceNeRFlow & 0.896 & 0.890 & 0.911 & 0.892 & 0.929 & 0.880 & 0.900 & 0.916 & 0.930 & 0.955 & 0.952 & 0.960 & 0.968 & 0.918 & 0.925 & 0.949 & 0.929 & 0.940 \\
\multicolumn{19}{@{}l}{\emph{Dynamic, Gaussian-based}} \\
Dynamic 3DGS & 0.944 & 0.945 & 0.961 & 0.960 & 0.951 & 0.942 & 0.951 & 0.967 & 0.970 & 0.978 & 0.945 & 0.989 & 0.967 & 0.963 & 0.943 & 0.986 & 0.943 & 0.965 \\
Dynamic-2DGS & -- & -- & -- & -- & 0.908 & 0.838 & 0.873$^\dagger$ & 0.961 & 0.887$^\ast$ & -- & 0.980 & 0.990 & -- & -- & -- & -- & -- & 0.977$^\dagger$ \\
AT-GS & 0.955 & 0.961 & 0.972 & 0.971 & 0.965 & \rankC{0.960} & 0.964 & \rankC{0.981} & \rankC{0.984} & \rankB{0.994} & \rankA{0.993} & \rankB{0.995} & \rankC{0.994} & 0.974 & \rankA{0.985} & \rankB{0.993} & \rankB{0.979} & \rankC{0.987} \\
4DGS & 0.934 & 0.933 & 0.948 & 0.947 & 0.951 & 0.925 & 0.940 & 0.948 & 0.968 & 0.986 & 0.977 & 0.987 & 0.989 & 0.963 & 0.959 & 0.983 & 0.956 & 0.972 \\
SC-GS & 0.942 & 0.946 & 0.956 & 0.952 & 0.954 & 0.934 & 0.947 & 0.974 & \rankC{0.984} & 0.988 & \rankC{0.991} & 0.990 & 0.993 & \rankC{0.978} & 0.968 & 0.985 & 0.967 & 0.982 \\
Deformable 3DGS & 0.940 & 0.941 & 0.958 & 0.955 & 0.954 & 0.934 & 0.947 & -- & 0.976 & 0.988 & 0.989 & 0.992 & 0.993 & 0.973 & 0.962 & 0.986 & 0.960 & 0.980$^\dagger$ \\
\multicolumn{19}{@{}l}{\emph{Dynamic, mesh-based Gaussian}} \\
GauSTAR & 0.943 & 0.949 & 0.965 & 0.963 & 0.957 & 0.954 & 0.955 & 0.972 & 0.975 & 0.960 & 0.925 & 0.984 & 0.964 & 0.968 & 0.954 & 0.982 & 0.948 & 0.963 \\
MaGS & 0.939 & 0.943 & 0.965 & 0.961 & 0.956 & 0.936 & 0.950 & -- & 0.982 & 0.991 & \rankC{0.991} & \rankC{0.994} & \rankB{0.995} & 0.973 & 0.962 & 0.990 & 0.966 & 0.983$^\dagger$ \\
DG-Mesh (mesh) & 0.933 & 0.935 & 0.946 & 0.944 & 0.946 & 0.917 & 0.937 & -- & 0.962 & 0.978 & 0.976 & 0.985 & 0.990 & 0.960 & 0.949 & 0.977 & 0.940 & 0.969$^\dagger$ \\
DG-Mesh (Gaussians) & 0.693 & 0.579 & 0.585 & 0.622 & 0.463 & 0.511 & 0.575 & -- & 0.702 & 0.931 & 0.825 & 0.885 & 0.830 & 0.709 & 0.835 & 0.865 & 0.888 & 0.830$^\dagger$ \\
\multicolumn{19}{@{}l}{\emph{Per-frame static}} \\
2DGS & 0.956 & 0.962 & 0.974 & 0.974 & 0.967 & 0.968 & 0.967 & 0.984 & 0.987 & 0.995 & 0.993 & 0.995 & 0.995 & 0.979 & 0.990 & 0.995 & 0.985 & 0.990 \\
PGSR & 0.950 & 0.957 & 0.967 & 0.967 & 0.962 & 0.960 & 0.961 & 0.981 & 0.984 & 0.951 & 0.993 & 0.966 & 0.995 & 0.974 & 0.989 & 0.964 & 0.983 & 0.978 \\
\midrule
\textbf{Ours} (Lite) & \rankB{0.963} & \rankB{0.970} & \rankB{0.980} & \rankB{0.978} & \rankA{0.980} & \rankA{0.973} & \rankB{0.974} & \rankB{0.982} & \rankB{0.985} & \rankB{0.994} & 0.990 & \rankB{0.995} & \rankB{0.995} & \rankB{0.980} & \rankC{0.982} & \rankA{0.994} & \rankC{0.978} & \rankB{0.988} \\
\textbf{Ours} & \rankA{0.964} & \rankA{0.972} & \rankA{0.981} & \rankA{0.979} & \rankA{0.980} & \rankA{0.973} & \rankA{0.975} & \rankA{0.984} & \rankA{0.986} & \rankA{0.995} & \rankB{0.992} & \rankA{0.996} & \rankA{0.996} & \rankA{0.982} & \rankA{0.985} & \rankA{0.994} & \rankA{0.980} & \rankA{0.989} \\
\bottomrule
\end{tabular}}
\caption{\textbf{Per-scene novel-view synthesis, SSIM} ($\uparrow$) on the held-out cameras. \ranklegend{} among the dynamic methods; --: failed; $^\dagger$mean over the scenes with a result; $^\ast$failed run, left out of the mean. DG-Mesh is rendered both from its mesh and from its Gaussians.}
\label{tab:perscene_ssim}
\end{table*}

%% file: tables/perscene_lpips.tex
\begin{table*}[t]
\centering
\scriptsize
\setlength{\tabcolsep}{2.2pt}
\resizebox{\textwidth}{!}{\begin{tabular}{@{}lccccccc|ccccccccccc@{}}
\toprule
 & \multicolumn{7}{c|}{ActorsHQ} & \multicolumn{11}{c}{DeformingThings4D} \\
Method & a4-1258 & a5-1942 & a6-1081 & a6-1604 & a7-554 & a8-325 & Mean & abe & AJ & canie & dragon & fox & grizz & ninja & peasant. & rabbit & vampire & Mean \\
\midrule
\multicolumn{19}{@{}l}{\emph{Dynamic, SDF-based}} \\
Tensor4D & 0.135 & 0.150 & 0.121 & 0.136 & 0.112 & 0.159 & 0.135 & 0.073 & 0.069 & 0.045 & 0.167 & 0.042 & 0.024 & 0.069 & 0.072 & 0.072 & 0.095 & 0.073 \\
NeuS2 & 0.054 & 0.055 & 0.050 & 0.050 & \rankC{0.047} & \rankC{0.053} & 0.052 & 0.041 & 0.051 & \rankC{0.016} & 0.025 & 0.015 & 0.015 & 0.067 & 0.033 & \rankC{0.025} & 0.052 & 0.034 \\
MoAngelo & \rankC{0.051} & \rankC{0.045} & \rankC{0.039} & \rankC{0.044} & \rankB{0.038} & \rankC{0.053} & \rankC{0.045} & 0.041 & 0.046 & 0.018 & 0.026 & \rankC{0.012} & \rankC{0.012} & \rankC{0.051} & 0.038 & \rankB{0.015} & 0.056 & \rankC{0.032} \\
\multicolumn{19}{@{}l}{\emph{Dynamic, NeRF-based}} \\
SceNeRFlow & 0.147 & 0.162 & 0.137 & 0.160 & 0.109 & 0.157 & 0.145 & 0.105 & 0.103 & 0.091 & 0.101 & 0.072 & 0.065 & 0.134 & 0.088 & 0.110 & 0.085 & 0.095 \\
\multicolumn{19}{@{}l}{\emph{Dynamic, Gaussian-based}} \\
Dynamic 3DGS & 0.111 & 0.129 & 0.101 & 0.100 & 0.115 & 0.133 & 0.115 & 0.063 & 0.071 & 0.061 & 0.065 & 0.034 & 0.063 & 0.097 & 0.092 & 0.050 & 0.096 & 0.069 \\
Dynamic-2DGS & -- & -- & -- & -- & 0.176 & 0.266 & 0.221$^\dagger$ & 0.061 & 0.169$^\ast$ & -- & 0.061 & 0.034 & -- & -- & -- & -- & -- & 0.052$^\dagger$ \\
AT-GS & 0.083 & 0.090 & 0.069 & 0.071 & 0.078 & 0.088 & 0.080 & \rankC{0.039} & 0.043 & 0.025 & 0.028 & 0.022 & 0.027 & 0.072 & \rankC{0.032} & 0.030 & \rankC{0.046} & 0.036 \\
4DGS & 0.112 & 0.129 & 0.105 & 0.111 & 0.087 & 0.142 & 0.114 & 0.098 & 0.070 & 0.048 & 0.070 & 0.038 & 0.036 & 0.099 & 0.067 & 0.054 & 0.086 & 0.067 \\
SC-GS & 0.097 & 0.101 & 0.104 & 0.116 & 0.079 & 0.130 & 0.105 & 0.046 & \rankC{0.035} & 0.045 & 0.034 & 0.037 & 0.034 & 0.052 & 0.051 & 0.051 & 0.056 & 0.044 \\
Deformable 3DGS & 0.106 & 0.122 & 0.090 & 0.099 & 0.083 & 0.134 & 0.106 & -- & 0.052 & 0.043 & 0.037 & 0.028 & 0.027 & 0.063 & 0.060 & 0.048 & 0.075 & 0.048$^\dagger$ \\
\multicolumn{19}{@{}l}{\emph{Dynamic, mesh-based Gaussian}} \\
GauSTAR & 0.073 & 0.078 & 0.066 & 0.061 & 0.073 & 0.071 & 0.071 & \rankC{0.039} & 0.043 & 0.104 & 0.193 & 0.026 & 0.088 & 0.060 & 0.077 & 0.036 & 0.085 & 0.075 \\
MaGS & 0.093 & 0.097 & 0.066 & 0.075 & 0.067 & 0.118 & 0.086 & -- & 0.039 & 0.030 & \rankC{0.024} & 0.017 & 0.018 & 0.063 & 0.056 & 0.029 & 0.061 & 0.037$^\dagger$ \\
DG-Mesh (mesh) & 0.087 & 0.096 & 0.088 & 0.094 & 0.084 & 0.121 & 0.095 & -- & 0.060 & 0.045 & 0.048 & 0.022 & 0.014 & 0.075 & 0.063 & 0.042 & 0.096 & 0.052$^\dagger$ \\
DG-Mesh (Gaussians) & 0.207 & 0.254 & 0.226 & 0.235 & 0.233 & 0.278 & 0.239 & -- & 0.145 & 0.097 & 0.115 & 0.076 & 0.084 & 0.176 & 0.112 & 0.116 & 0.134 & 0.117$^\dagger$ \\
\multicolumn{19}{@{}l}{\emph{Per-frame static}} \\
2DGS & 0.034 & 0.043 & 0.033 & 0.031 & 0.038 & 0.038 & 0.036 & 0.029 & 0.031 & 0.015 & 0.021 & 0.013 & 0.017 & 0.051 & 0.021 & 0.015 & 0.030 & 0.024 \\
PGSR & 0.040 & 0.047 & 0.036 & 0.036 & 0.042 & 0.043 & 0.041 & 0.028 & 0.032 & 0.088 & 0.023 & 0.057 & 0.020 & 0.051 & 0.023 & 0.070 & 0.031 & 0.042 \\
\midrule
\textbf{Ours} (Lite) & \rankB{0.042} & \rankB{0.043} & \rankB{0.031} & \rankB{0.034} & \rankA{0.031} & \rankB{0.036} & \rankB{0.036} & \rankB{0.025} & \rankB{0.033} & \rankB{0.009} & \rankB{0.019} & \rankB{0.008} & \rankB{0.009} & \rankB{0.042} & \rankB{0.025} & \rankA{0.010} & \rankB{0.035} & \rankB{0.021} \\
\textbf{Ours} & \rankA{0.040} & \rankA{0.040} & \rankA{0.030} & \rankA{0.032} & \rankA{0.031} & \rankA{0.035} & \rankA{0.035} & \rankA{0.022} & \rankA{0.030} & \rankA{0.008} & \rankA{0.014} & \rankA{0.007} & \rankA{0.008} & \rankA{0.035} & \rankA{0.022} & \rankA{0.010} & \rankA{0.034} & \rankA{0.019} \\
\bottomrule
\end{tabular}}
\caption{\textbf{Per-scene novel-view synthesis, LPIPS} ($\downarrow$) on the held-out cameras. \ranklegend{} among the dynamic methods; --: failed; $^\dagger$mean over the scenes with a result; $^\ast$failed run, left out of the mean. DG-Mesh is rendered both from its mesh and from its Gaussians.}
\label{tab:perscene_lpips}
\end{table*}

%% file: tables/nvs_taogs.tex
\begin{table*}[t]
\centering
\footnotesize
\setlength{\tabcolsep}{4pt}
\begin{tabular}{@{}l|cccccc|ccc@{}}
\toprule
 & \multicolumn{6}{c|}{PSNR$\uparrow$ per scene} & \multicolumn{3}{c}{Average} \\
\cmidrule(lr){2-7}\cmidrule(l){8-10}
Method & magic2 & magic3 & drawsword & umbrella & bag & changeclothes & PSNR$\uparrow$ & SSIM$\uparrow$ & LPIPS$\downarrow$ \\
\midrule
\multicolumn{10}{@{}l}{\emph{Dynamic, SDF-based}} \\
Tensor4D & 28.58 & 29.63 & 27.53 & 27.19 & 34.91 & -- & 29.57$^\dagger$ & 0.955$^\dagger$ & 0.080$^\dagger$ \\
NeuS2 & 33.93 & 34.37 & \rankC{32.46} & 29.54 & 37.74 & 28.16 & \rankC{32.70} & \rankC{0.969} & \rankC{0.055} \\
MoAngelo & 34.04 & \rankC{34.38} & 31.40 & \rankC{30.15} & 37.78 & \rankC{28.27} & 32.67 & \rankC{0.969} & \rankB{0.051} \\
\midrule
\multicolumn{10}{@{}l}{\emph{Dynamic, NeRF-based}} \\
SceNeRFlow & 25.51 & 28.69 & 24.45 & 25.47 & 30.89 & 23.49 & 26.42 & 0.934 & 0.093 \\
\midrule
\multicolumn{10}{@{}l}{\emph{Dynamic, Gaussian-based}} \\
Dynamic 3DGS & 32.88 & 33.67 & 32.42 & 28.94 & 36.98 & 27.42 & 32.05 & 0.963 & 0.083 \\
Dynamic-2DGS & -- & -- & 27.33 & 28.36 & 29.83 & -- & 28.51$^\dagger$ & 0.958$^\dagger$ & 0.080$^\dagger$ \\
AT-GS & \rankA{34.88} & \rankB{35.26} & \rankA{33.97} & \rankB{30.51} & \rankA{38.71} & \rankB{28.78} & \rankA{33.69} & \rankA{0.974} & 0.056 \\
4DGS & 31.17 & 32.05 & 28.72 & 29.08 & 37.53 & 26.92 & 30.91 & 0.960 & 0.083 \\
SC-GS & \rankC{34.19} & 33.92 & 28.21 & 29.97 & \rankB{38.26} & 26.62 & 31.86 & 0.967 & 0.067 \\
Deformable 3DGS & 33.35 & 32.94 & 29.91 & 29.89 & \rankC{38.24} & 26.91 & 31.87 & 0.967 & 0.069 \\
\midrule
\multicolumn{10}{@{}l}{\emph{Dynamic, mesh-based Gaussian}} \\
GauSTAR & 29.20 & 32.83 & 30.41 & 24.96 & -- & -- & 29.35$^\dagger$ & 0.963$^\dagger$ & 0.053$^\dagger$ \\
DG-Mesh (mesh) & 30.04 & 29.02 & 26.99 & 27.35 & 35.12 & 25.33 & 28.98 & 0.956 & 0.080 \\
DG-Mesh (Gaussians) & 21.97 & 23.41 & 24.56 & 19.02 & 25.69 & 23.04 & 22.95 & 0.788 & 0.143 \\
MaGS & 34.13 & 32.58 & 29.96 & 29.80 & 38.20 & 26.73 & 31.90 & 0.968 & 0.063 \\
\midrule
\multicolumn{10}{@{}l}{\emph{Per-frame static}} \\
2DGS & 35.18 & 36.02 & 33.94 & 30.52 & 38.66 & 29.10 & 33.90 & 0.977 & 0.036 \\
PGSR & 34.74 & 35.72 & 33.60 & 30.35 & 38.29 & 28.95 & 33.61 & 0.977 & 0.035 \\
\midrule
\textbf{Ours} & \rankB{34.48} & \rankA{35.36} & \rankB{32.50} & \rankA{30.52} & 38.17 & \rankA{29.10} & \rankB{33.36} & \rankB{0.972} & \rankA{0.045} \\
\bottomrule
\end{tabular}
\caption{\textbf{Novel-view synthesis on TaoGS}: 5 held-out cameras per scene (the cameras of the point-track evaluation, \Cref{tab:realdata_tracks_scenes}), 50 frames, background of the photos set to black with the foreground masks, as in \Cref{tab:nvs}; PSNR in dB. \ranklegend{} among the dynamic methods. --: failed (GauSTAR on bag and changeclothes learns no colour and renders a uniformly green silhouette; Tensor4D renders no foreground on changeclothes; Dynamic-2DGS produced no final model on magic2, magic3 and changeclothes). $^\dagger$Averaged over the scenes with a result, not ranked: Tensor4D: 5 of 6 scenes; Dynamic-2DGS: 3 of 6 scenes; GauSTAR: 4 of 6 scenes.}
\label{tab:nvs_taogs}
\end{table*}

%% file: figures/nvs_cmp_gallery.tex
\begin{figure*}[p]
\centering
\begin{tikzpicture}[x=1cm, y=1cm, font=\scriptsize, cell/.style={anchor=north west, inner sep=0},
    val/.style={font=\tiny, anchor=south east, fill=white, fill opacity=0.85, text opacity=1, inner sep=1pt}]
\draw[black!45, line width=0.4pt] (0.050,0.420) -- (3.302,0.420);
\node[anchor=south, inner sep=1pt, font=\scriptsize\itshape] at (1.676,0.420) {Static (per frame)};
\draw[black!45, line width=0.4pt] (3.562,0.420) -- (15.414,0.420);
\node[anchor=south, inner sep=1pt, font=\scriptsize\itshape] at (9.488,0.420) {Dynamic};
\draw[black!40, dashed, line width=0.5pt] (3.432,0.720) -- (3.432,-18.281);
\node[anchor=south, inner sep=1pt] at (0.828,0.000) {2DGS};
\node[anchor=south, inner sep=1pt] at (2.524,0.000) {PGSR};
\node[anchor=south, inner sep=1pt] at (4.340,0.000) {NeuS2};
\node[anchor=south, inner sep=1pt] at (6.036,0.000) {AT-GS};
\node[anchor=south, inner sep=1pt] at (7.732,0.000) {SC-GS};
\node[anchor=south, inner sep=1pt] at (9.428,0.000) {DG-Mesh};
\node[anchor=south, inner sep=1pt] at (11.124,0.000) {GauSTAR};
\node[anchor=south, inner sep=1pt] at (12.820,0.000) {MoAngelo};
\node[anchor=south, inner sep=1pt] at (14.636,0.000) {\color[HTML]{0E80E5}\textbf{Ours}};
\node[anchor=south, inner sep=1pt] at (16.452,0.000) {GT};
\node[anchor=north west, inner sep=1pt, fill=white, font=\scriptsize\itshape] at (0,-0.140) {DeformingThings4D vampire, held-out camera Cam035};
\node[cell] at (0.000,-0.480) {\includegraphics[width=1.656cm]{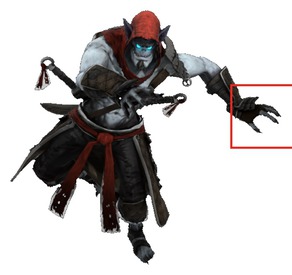}};
\node[val] at (1.656,-2.040) {32.70\,/\,0.026};
\node[cell] at (0.000,-2.090) {\includegraphics[width=1.656cm]{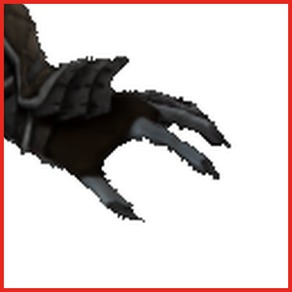}};
\node[cell] at (1.696,-0.480) {\includegraphics[width=1.656cm]{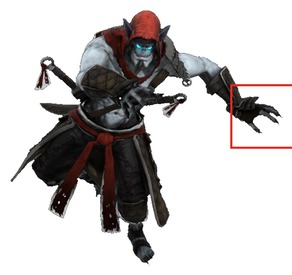}};
\node[val] at (3.352,-2.040) {32.52\,/\,0.028};
\node[cell] at (1.696,-2.090) {\includegraphics[width=1.656cm]{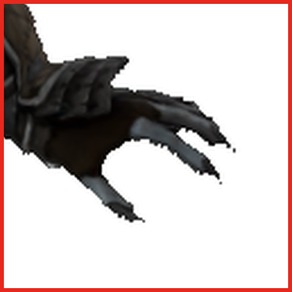}};
\node[cell] at (3.512,-0.480) {\includegraphics[width=1.656cm]{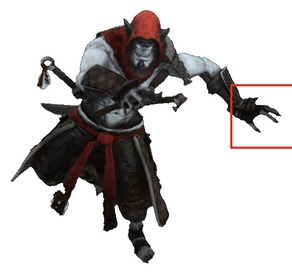}};
\node[val] at (5.168,-2.040) {28.27\,/\,0.056};
\node[cell] at (3.512,-2.090) {\includegraphics[width=1.656cm]{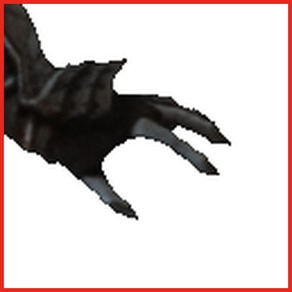}};
\node[cell] at (5.208,-0.480) {\includegraphics[width=1.656cm]{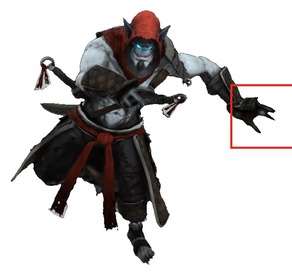}};
\node[val] at (6.864,-2.040) {29.20\,/\,0.050};
\node[cell] at (5.208,-2.090) {\includegraphics[width=1.656cm]{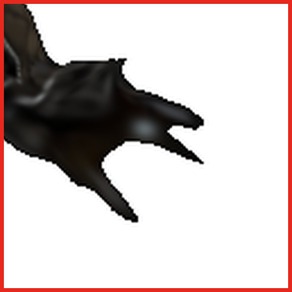}};
\node[cell] at (6.904,-0.480) {\includegraphics[width=1.656cm]{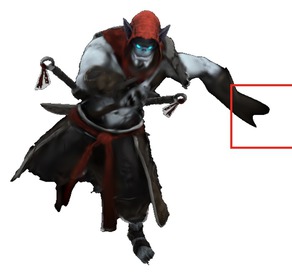}};
\node[val] at (8.560,-2.040) {28.47\,/\,0.066};
\node[cell] at (6.904,-2.090) {\includegraphics[width=1.656cm]{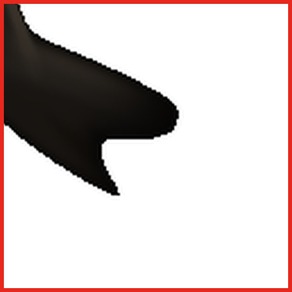}};
\node[cell] at (8.600,-0.480) {\includegraphics[width=1.656cm]{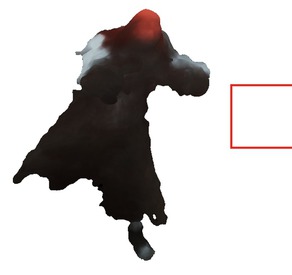}};
\node[val] at (10.256,-2.040) {24.52\,/\,0.112};
\node[cell] at (8.600,-2.090) {\includegraphics[width=1.656cm]{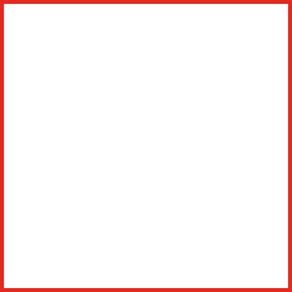}};
\node[cell] at (10.296,-0.480) {\includegraphics[width=1.656cm]{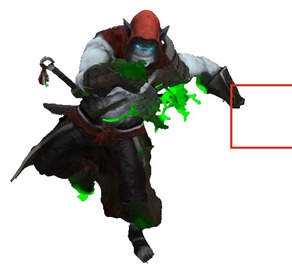}};
\node[val] at (11.952,-2.040) {25.30\,/\,0.081};
\node[cell] at (10.296,-2.090) {\includegraphics[width=1.656cm]{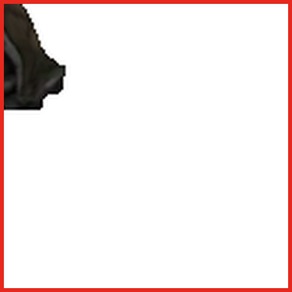}};
\node[cell] at (11.992,-0.480) {\includegraphics[width=1.656cm]{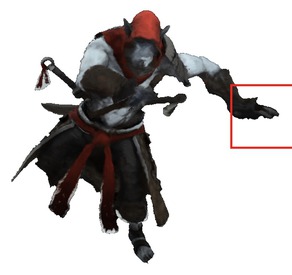}};
\node[val] at (13.648,-2.040) {29.07\,/\,0.060};
\node[cell] at (11.992,-2.090) {\includegraphics[width=1.656cm]{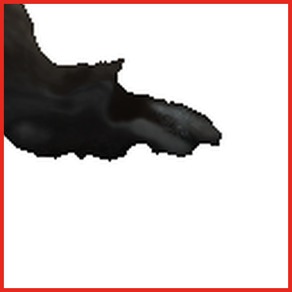}};
\node[cell] at (13.808,-0.480) {\includegraphics[width=1.656cm]{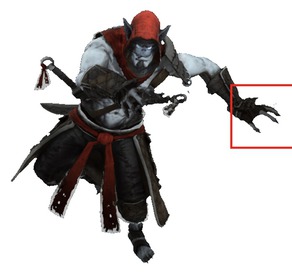}};
\node[val] at (15.464,-2.040) {\color[HTML]{0E80E5}\bfseries 32.79\,/\,0.032};
\node[cell] at (13.808,-2.090) {\includegraphics[width=1.656cm]{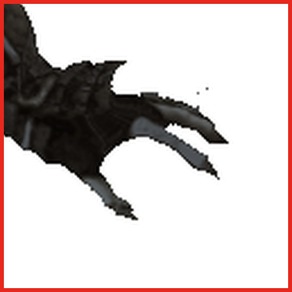}};
\node[cell] at (15.624,-0.480) {\includegraphics[width=1.656cm]{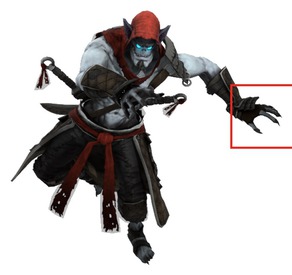}};
\node[cell] at (15.624,-2.090) {\includegraphics[width=1.656cm]{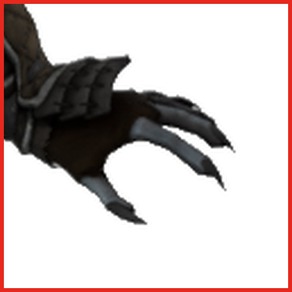}};
\node[anchor=north west, inner sep=1pt, fill=white, font=\scriptsize\itshape] at (0,-3.986) {DeformingThings4D peasantgirl, held-out camera Cam005};
\node[cell] at (0.000,-4.326) {\includegraphics[width=1.656cm]{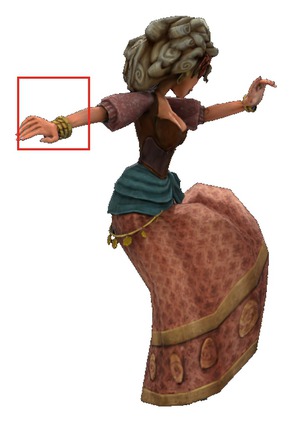}};
\node[val] at (1.656,-6.718) {35.55\,/\,0.022};
\node[cell] at (0.000,-6.768) {\includegraphics[width=1.656cm]{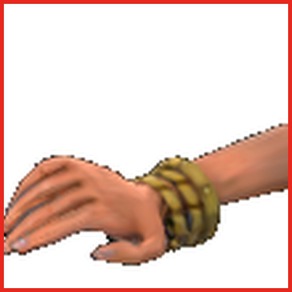}};
\node[cell] at (1.696,-4.326) {\includegraphics[width=1.656cm]{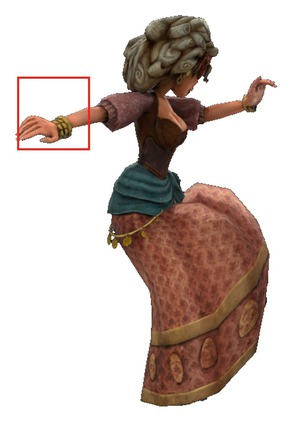}};
\node[val] at (3.352,-6.718) {35.46\,/\,0.025};
\node[cell] at (1.696,-6.768) {\includegraphics[width=1.656cm]{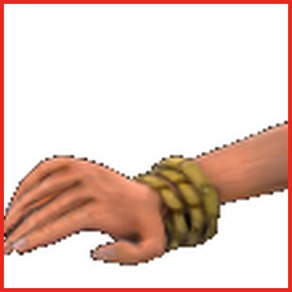}};
\node[cell] at (3.512,-4.326) {\includegraphics[width=1.656cm]{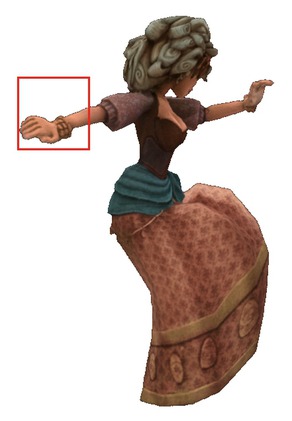}};
\node[val] at (5.168,-6.718) {33.27\,/\,0.037};
\node[cell] at (3.512,-6.768) {\includegraphics[width=1.656cm]{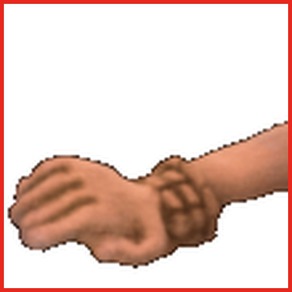}};
\node[cell] at (5.208,-4.326) {\includegraphics[width=1.656cm]{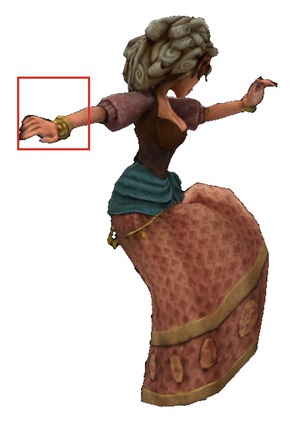}};
\node[val] at (6.864,-6.718) {33.30\,/\,0.039};
\node[cell] at (5.208,-6.768) {\includegraphics[width=1.656cm]{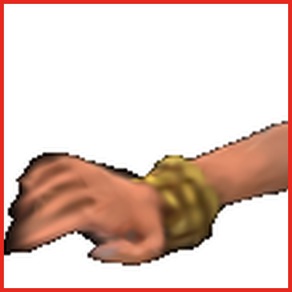}};
\node[cell] at (6.904,-4.326) {\includegraphics[width=1.656cm]{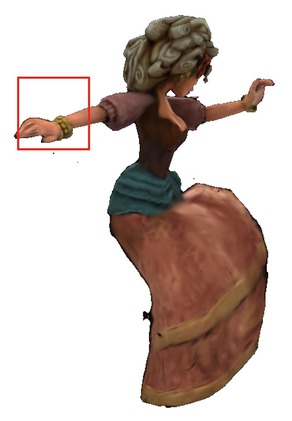}};
\node[val] at (8.560,-6.718) {31.43\,/\,0.058};
\node[cell] at (6.904,-6.768) {\includegraphics[width=1.656cm]{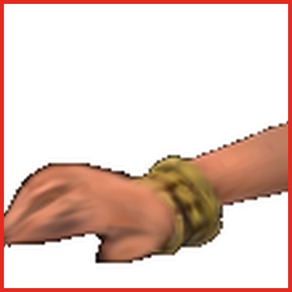}};
\node[cell] at (8.600,-4.326) {\includegraphics[width=1.656cm]{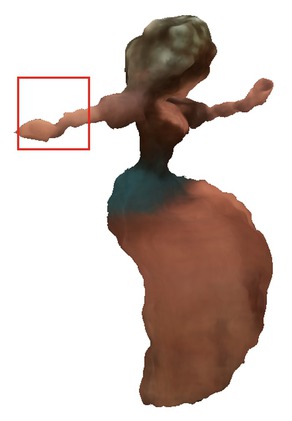}};
\node[val] at (10.256,-6.718) {28.23\,/\,0.062};
\node[cell] at (8.600,-6.768) {\includegraphics[width=1.656cm]{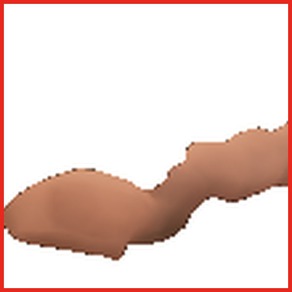}};
\node[cell] at (10.296,-4.326) {\includegraphics[width=1.656cm]{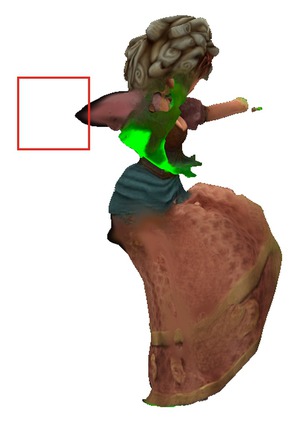}};
\node[val] at (11.952,-6.718) {25.86\,/\,0.082};
\node[cell] at (10.296,-6.768) {\includegraphics[width=1.656cm]{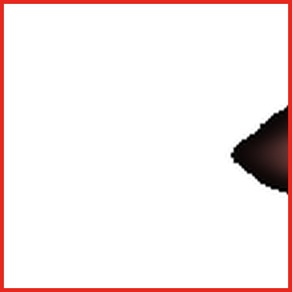}};
\node[cell] at (11.992,-4.326) {\includegraphics[width=1.656cm]{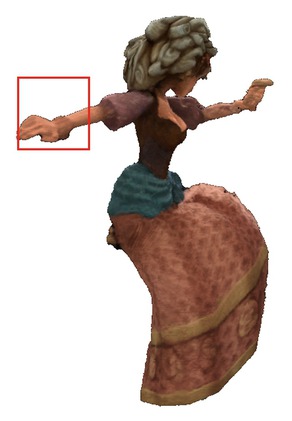}};
\node[val] at (13.648,-6.718) {32.45\,/\,0.039};
\node[cell] at (11.992,-6.768) {\includegraphics[width=1.656cm]{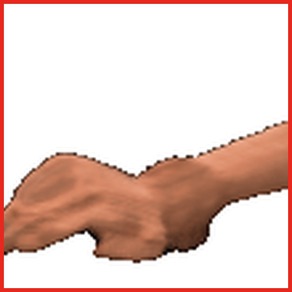}};
\node[cell] at (13.808,-4.326) {\includegraphics[width=1.656cm]{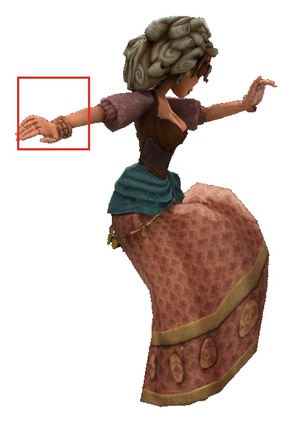}};
\node[val] at (15.464,-6.718) {\color[HTML]{0E80E5}\bfseries 35.33\,/\,0.019};
\node[cell] at (13.808,-6.768) {\includegraphics[width=1.656cm]{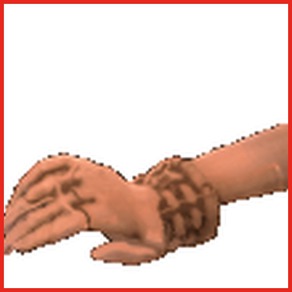}};
\node[cell] at (15.624,-4.326) {\includegraphics[width=1.656cm]{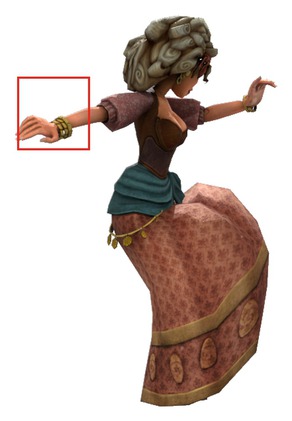}};
\node[cell] at (15.624,-6.768) {\includegraphics[width=1.656cm]{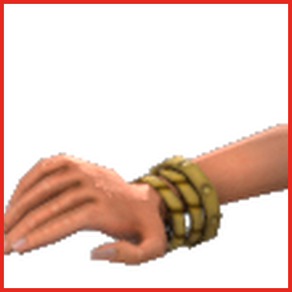}};
\node[anchor=north west, inner sep=1pt, fill=white, font=\scriptsize\itshape] at (0,-8.664) {DeformingThings4D abe, held-out camera Cam005};
\node[cell] at (0.000,-9.004) {\includegraphics[width=1.656cm]{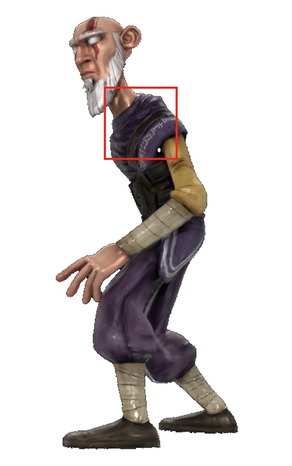}};
\node[val] at (1.656,-11.652) {33.95\,/\,0.024};
\node[cell] at (0.000,-11.702) {\includegraphics[width=1.656cm]{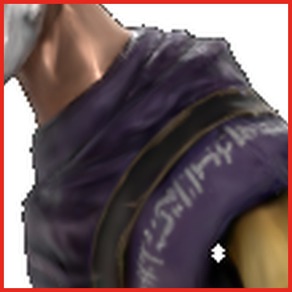}};
\node[cell] at (1.696,-9.004) {\includegraphics[width=1.656cm]{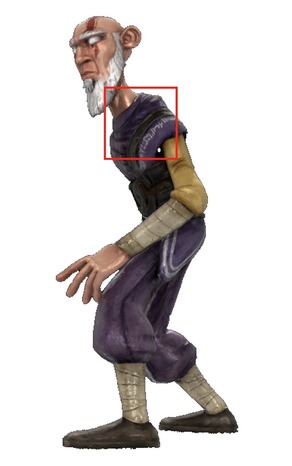}};
\node[val] at (3.352,-11.652) {33.31\,/\,0.024};
\node[cell] at (1.696,-11.702) {\includegraphics[width=1.656cm]{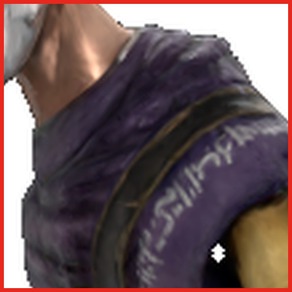}};
\node[cell] at (3.512,-9.004) {\includegraphics[width=1.656cm]{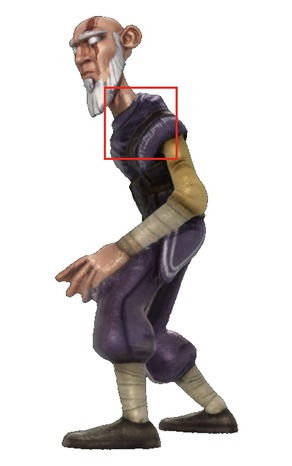}};
\node[val] at (5.168,-11.652) {33.06\,/\,0.034};
\node[cell] at (3.512,-11.702) {\includegraphics[width=1.656cm]{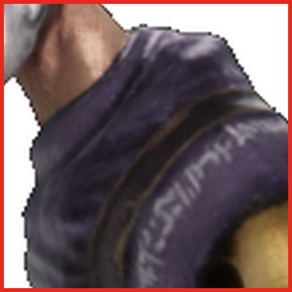}};
\node[cell] at (5.208,-9.004) {\includegraphics[width=1.656cm]{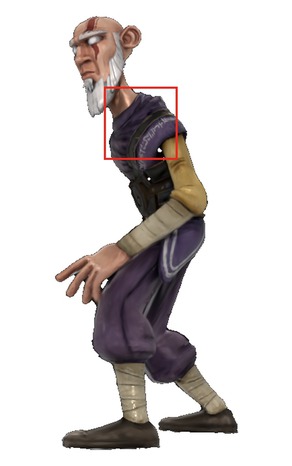}};
\node[val] at (6.864,-11.652) {33.30\,/\,0.032};
\node[cell] at (5.208,-11.702) {\includegraphics[width=1.656cm]{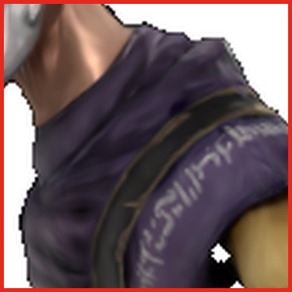}};
\node[cell] at (6.904,-9.004) {\includegraphics[width=1.656cm]{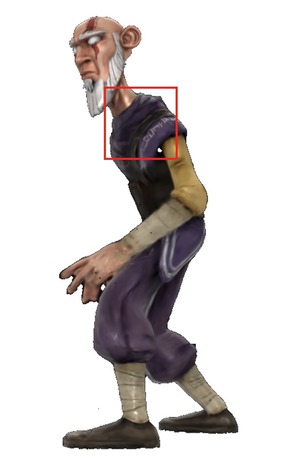}};
\node[val] at (8.560,-11.652) {32.88\,/\,0.035};
\node[cell] at (6.904,-11.702) {\includegraphics[width=1.656cm]{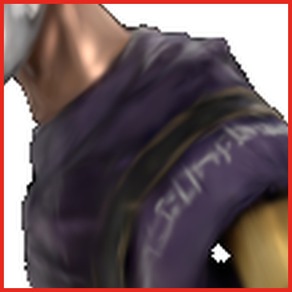}};
\node[draw=black!25, fill=black!4, minimum width=1.656cm, minimum height=2.648cm, anchor=north west, inner sep=0, font=\tiny, align=center] at (8.600,-9.004) {Method\\failed};
\node[draw=black!25, fill=black!4, minimum width=1.656cm, minimum height=1.656cm, anchor=north west, inner sep=0, font=\tiny, align=center] at (8.600,-11.702) {Method\\failed};
\node[cell] at (10.296,-9.004) {\includegraphics[width=1.656cm]{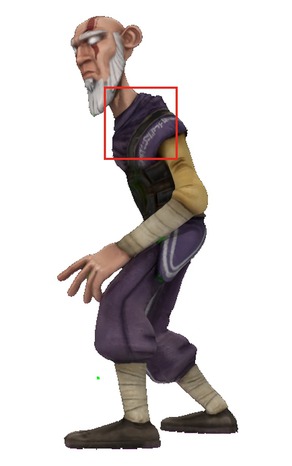}};
\node[val] at (11.952,-11.652) {30.25\,/\,0.032};
\node[cell] at (10.296,-11.702) {\includegraphics[width=1.656cm]{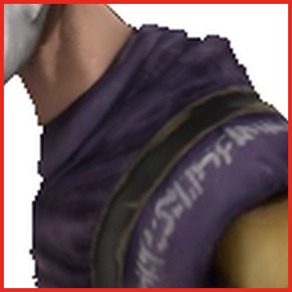}};
\node[cell] at (11.992,-9.004) {\includegraphics[width=1.656cm]{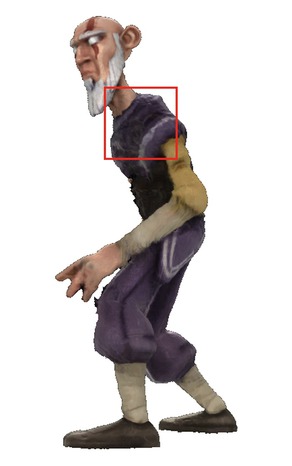}};
\node[val] at (13.648,-11.652) {32.00\,/\,0.034};
\node[cell] at (11.992,-11.702) {\includegraphics[width=1.656cm]{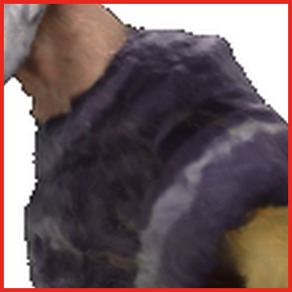}};
\node[cell] at (13.808,-9.004) {\includegraphics[width=1.656cm]{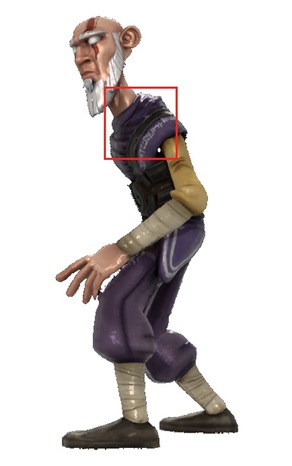}};
\node[val] at (15.464,-11.652) {\color[HTML]{0E80E5}\bfseries 34.74\,/\,0.018};
\node[cell] at (13.808,-11.702) {\includegraphics[width=1.656cm]{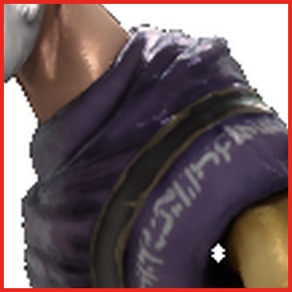}};
\node[cell] at (15.624,-9.004) {\includegraphics[width=1.656cm]{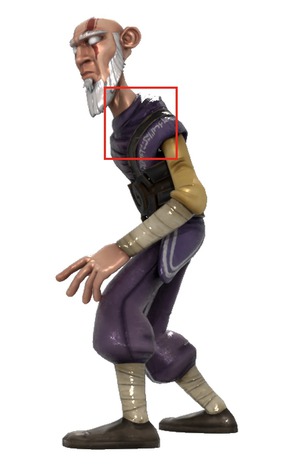}};
\node[cell] at (15.624,-11.702) {\includegraphics[width=1.656cm]{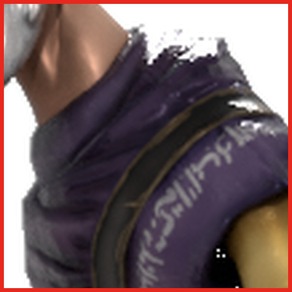}};
\node[anchor=north west, inner sep=1pt, fill=white, font=\scriptsize\itshape] at (0,-13.598) {DeformingThings4D ninja, held-out camera Cam045};
\node[cell] at (0.000,-13.938) {\includegraphics[width=1.656cm]{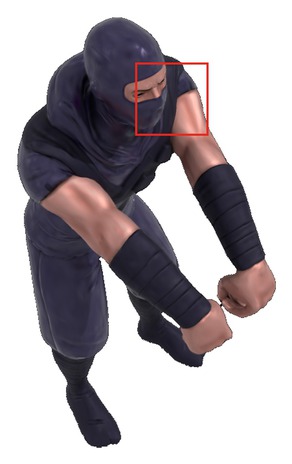}};
\node[val] at (1.656,-16.575) {35.10\,/\,0.039};
\node[cell] at (0.000,-16.625) {\includegraphics[width=1.656cm]{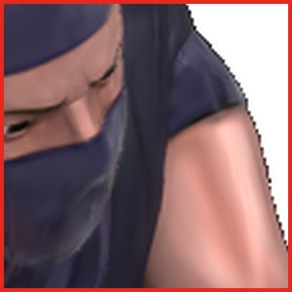}};
\node[cell] at (1.696,-13.938) {\includegraphics[width=1.656cm]{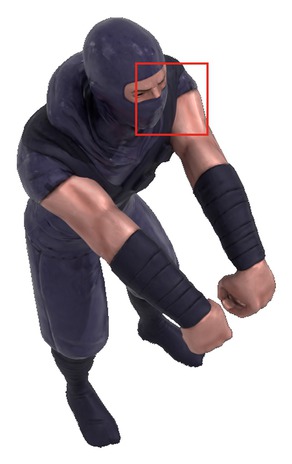}};
\node[val] at (3.352,-16.575) {34.26\,/\,0.039};
\node[cell] at (1.696,-16.625) {\includegraphics[width=1.656cm]{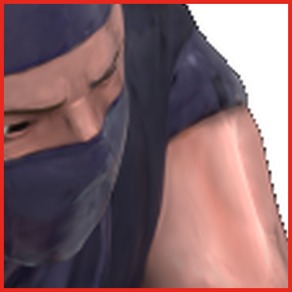}};
\node[cell] at (3.512,-13.938) {\includegraphics[width=1.656cm]{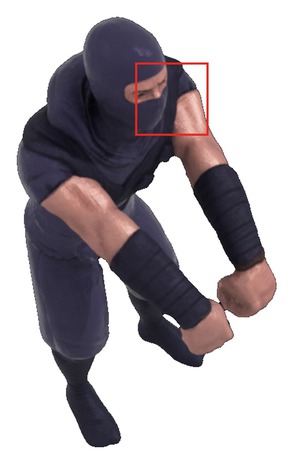}};
\node[val] at (5.168,-16.575) {33.49\,/\,0.058};
\node[cell] at (3.512,-16.625) {\includegraphics[width=1.656cm]{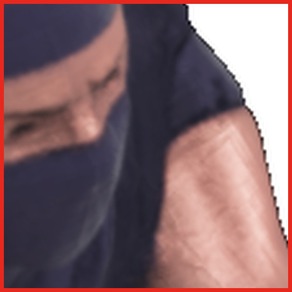}};
\node[cell] at (5.208,-13.938) {\includegraphics[width=1.656cm]{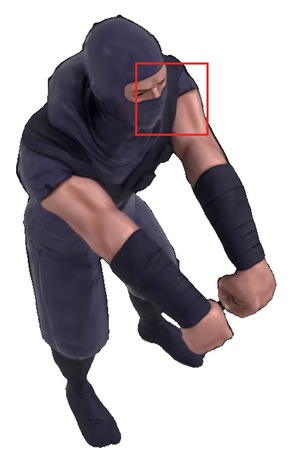}};
\node[val] at (6.864,-16.575) {32.83\,/\,0.059};
\node[cell] at (5.208,-16.625) {\includegraphics[width=1.656cm]{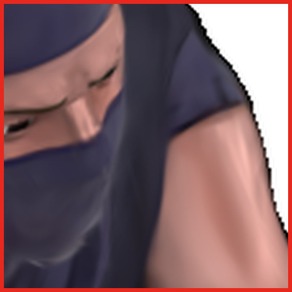}};
\node[cell] at (6.904,-13.938) {\includegraphics[width=1.656cm]{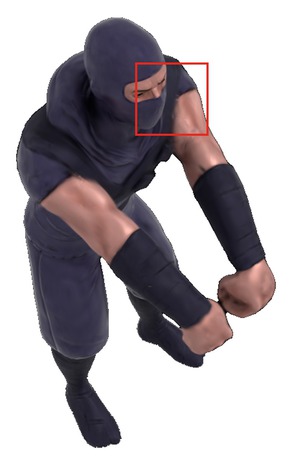}};
\node[val] at (8.560,-16.575) {33.82\,/\,0.046};
\node[cell] at (6.904,-16.625) {\includegraphics[width=1.656cm]{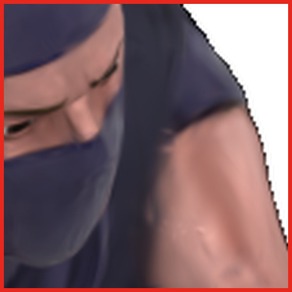}};
\node[cell] at (8.600,-13.938) {\includegraphics[width=1.656cm]{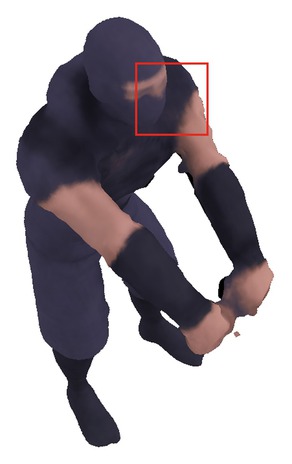}};
\node[val] at (10.256,-16.575) {28.48\,/\,0.064};
\node[cell] at (8.600,-16.625) {\includegraphics[width=1.656cm]{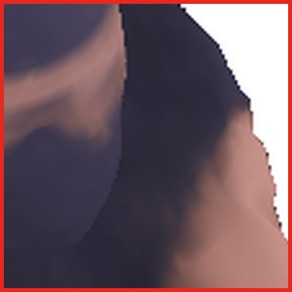}};
\node[cell] at (10.296,-13.938) {\includegraphics[width=1.656cm]{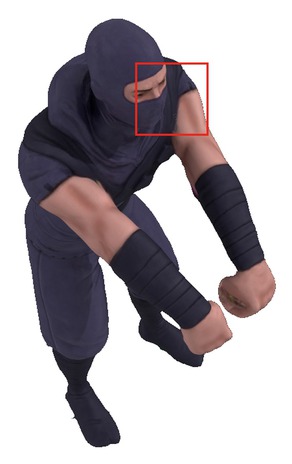}};
\node[val] at (11.952,-16.575) {30.70\,/\,0.049};
\node[cell] at (10.296,-16.625) {\includegraphics[width=1.656cm]{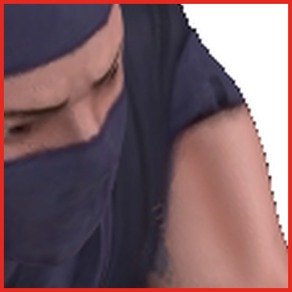}};
\node[cell] at (11.992,-13.938) {\includegraphics[width=1.656cm]{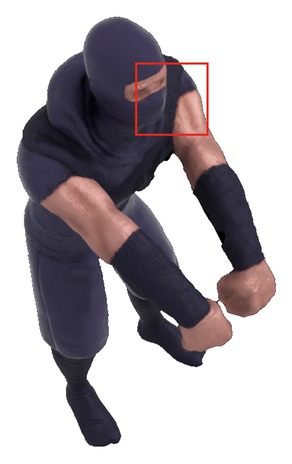}};
\node[val] at (13.648,-16.575) {33.22\,/\,0.050};
\node[cell] at (11.992,-16.625) {\includegraphics[width=1.656cm]{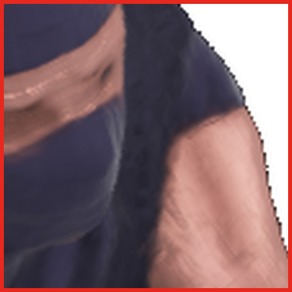}};
\node[cell] at (13.808,-13.938) {\includegraphics[width=1.656cm]{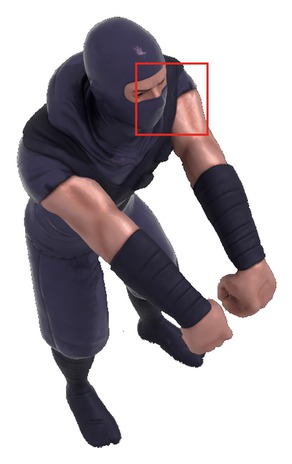}};
\node[val] at (15.464,-16.575) {\color[HTML]{0E80E5}\bfseries 35.12\,/\,0.029};
\node[cell] at (13.808,-16.625) {\includegraphics[width=1.656cm]{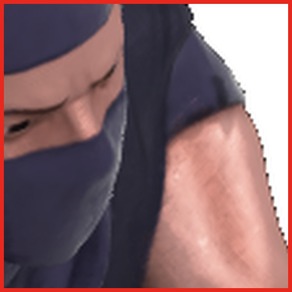}};
\node[cell] at (15.624,-13.938) {\includegraphics[width=1.656cm]{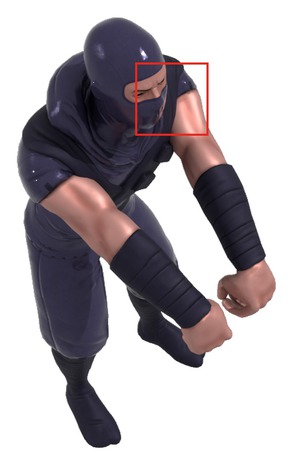}};
\node[cell] at (15.624,-16.625) {\includegraphics[width=1.656cm]{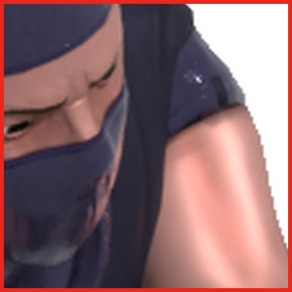}};
\end{tikzpicture}
\caption{\textbf{Novel-view synthesis compared with the baselines} (DeformingThings4D): renders in unseen views, i.e.\ held-out cameras not used for training. Numbers: PSNR (dB) / LPIPS of the image, as in \cref{tab:nvs}, computed as in the evaluation on a black background (GT: the held-out photo with the background masked out); shown here on white. The red box marks the enlarged region (same box for every method). Left to right: per-frame static methods, dynamic methods, Ours, ground truth.}
\label{fig:nvs_cmp_1}
\end{figure*}

\begin{figure*}[p]
\centering
\begin{tikzpicture}[x=1cm, y=1cm, font=\scriptsize, cell/.style={anchor=north west, inner sep=0},
    val/.style={font=\tiny, anchor=south east, fill=white, fill opacity=0.85, text opacity=1, inner sep=1pt}]
\draw[black!45, line width=0.4pt] (0.050,0.420) -- (3.302,0.420);
\node[anchor=south, inner sep=1pt, font=\scriptsize\itshape] at (1.676,0.420) {Static (per frame)};
\draw[black!45, line width=0.4pt] (3.562,0.420) -- (15.414,0.420);
\node[anchor=south, inner sep=1pt, font=\scriptsize\itshape] at (9.488,0.420) {Dynamic};
\draw[black!40, dashed, line width=0.5pt] (3.432,0.720) -- (3.432,-16.698);
\node[anchor=south, inner sep=1pt] at (0.828,0.000) {2DGS};
\node[anchor=south, inner sep=1pt] at (2.524,0.000) {PGSR};
\node[anchor=south, inner sep=1pt] at (4.340,0.000) {NeuS2};
\node[anchor=south, inner sep=1pt] at (6.036,0.000) {AT-GS};
\node[anchor=south, inner sep=1pt] at (7.732,0.000) {SC-GS};
\node[anchor=south, inner sep=1pt] at (9.428,0.000) {DG-Mesh};
\node[anchor=south, inner sep=1pt] at (11.124,0.000) {GauSTAR};
\node[anchor=south, inner sep=1pt] at (12.820,0.000) {MoAngelo};
\node[anchor=south, inner sep=1pt] at (14.636,0.000) {\color[HTML]{0E80E5}\textbf{Ours}};
\node[anchor=south, inner sep=1pt] at (16.452,0.000) {GT};
\node[anchor=north west, inner sep=1pt, fill=white, font=\scriptsize\itshape] at (0,-0.140) {ActorsHQ actor8\_seq2\_325, held-out camera Cam127};
\node[cell] at (0.000,-0.480) {\includegraphics[width=1.656cm]{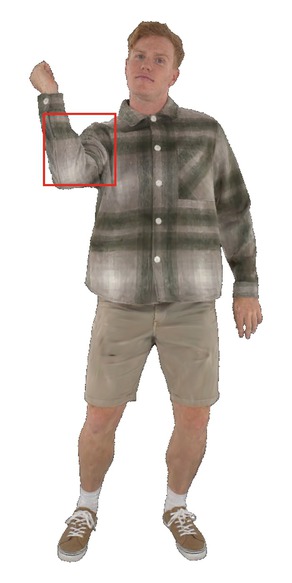}};
\node[val] at (1.656,-3.792) {32.10\,/\,0.037};
\node[cell] at (0.000,-3.842) {\includegraphics[width=1.656cm]{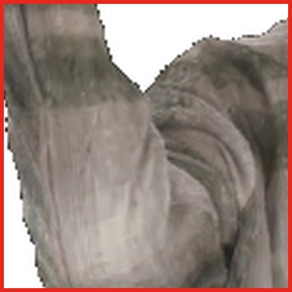}};
\node[cell] at (1.696,-0.480) {\includegraphics[width=1.656cm]{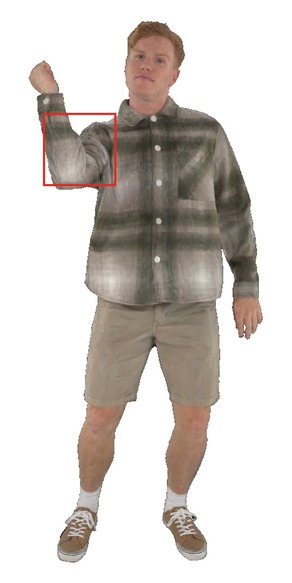}};
\node[val] at (3.352,-3.792) {29.77\,/\,0.042};
\node[cell] at (1.696,-3.842) {\includegraphics[width=1.656cm]{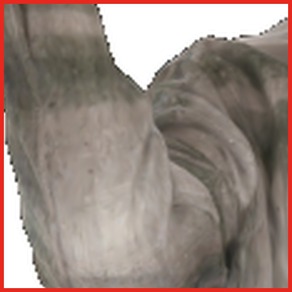}};
\node[cell] at (3.512,-0.480) {\includegraphics[width=1.656cm]{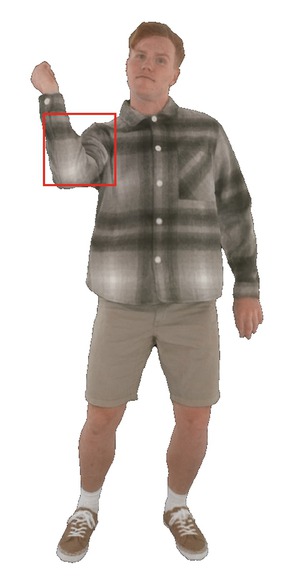}};
\node[val] at (5.168,-3.792) {31.15\,/\,0.048};
\node[cell] at (3.512,-3.842) {\includegraphics[width=1.656cm]{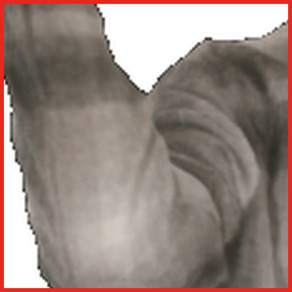}};
\node[cell] at (5.208,-0.480) {\includegraphics[width=1.656cm]{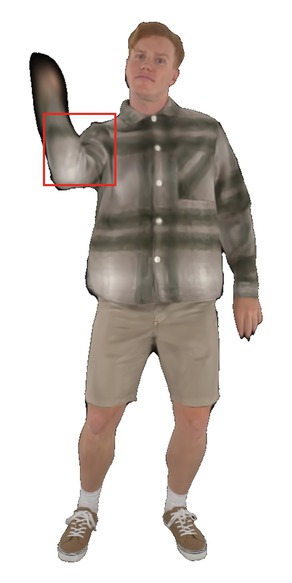}};
\node[val] at (6.864,-3.792) {30.79\,/\,0.098};
\node[cell] at (5.208,-3.842) {\includegraphics[width=1.656cm]{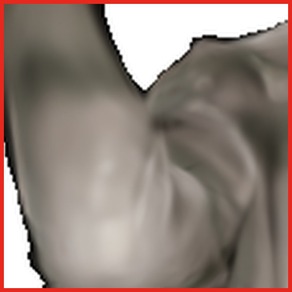}};
\node[cell] at (6.904,-0.480) {\includegraphics[width=1.656cm]{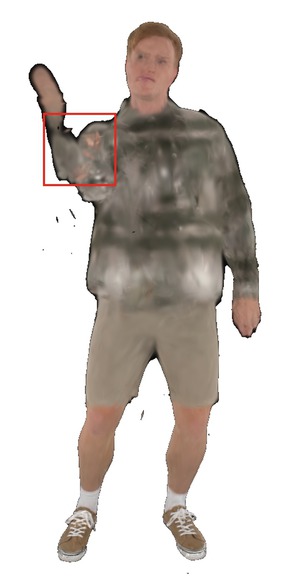}};
\node[val] at (8.560,-3.792) {25.91\,/\,0.147};
\node[cell] at (6.904,-3.842) {\includegraphics[width=1.656cm]{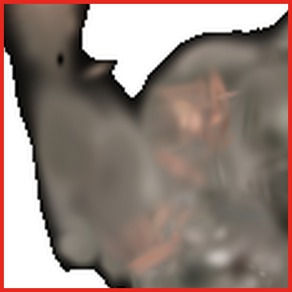}};
\node[cell] at (8.600,-0.480) {\includegraphics[width=1.656cm]{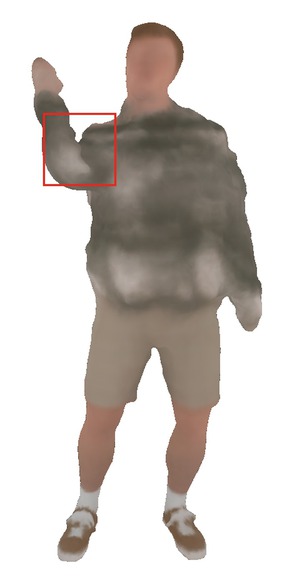}};
\node[val] at (10.256,-3.792) {24.55\,/\,0.124};
\node[cell] at (8.600,-3.842) {\includegraphics[width=1.656cm]{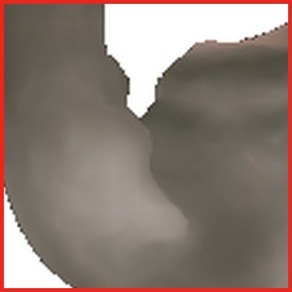}};
\node[cell] at (10.296,-0.480) {\includegraphics[width=1.656cm]{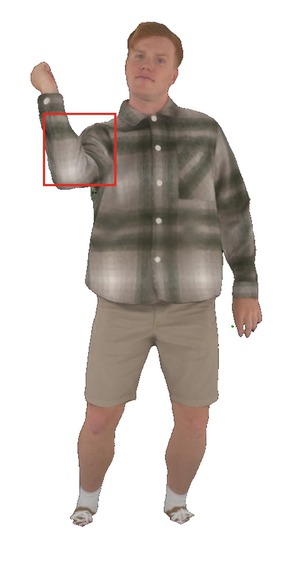}};
\node[val] at (11.952,-3.792) {25.68\,/\,0.070};
\node[cell] at (10.296,-3.842) {\includegraphics[width=1.656cm]{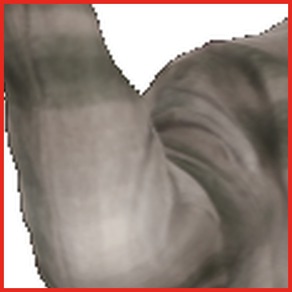}};
\node[cell] at (11.992,-0.480) {\includegraphics[width=1.656cm]{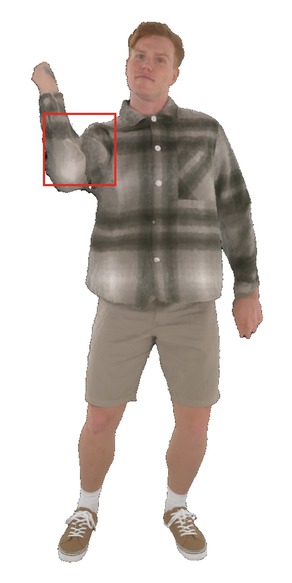}};
\node[val] at (13.648,-3.792) {31.17\,/\,0.045};
\node[cell] at (11.992,-3.842) {\includegraphics[width=1.656cm]{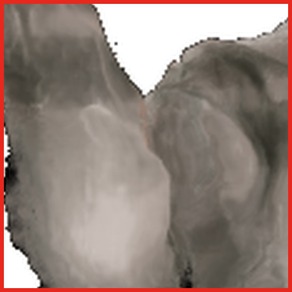}};
\node[cell] at (13.808,-0.480) {\includegraphics[width=1.656cm]{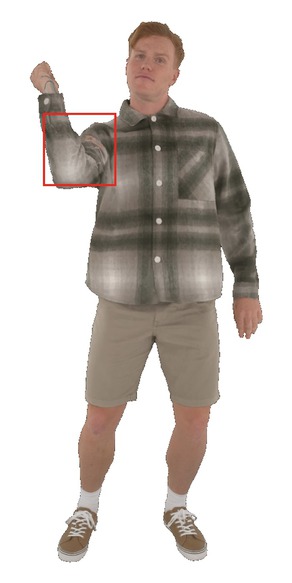}};
\node[val] at (15.464,-3.792) {\color[HTML]{0E80E5}\bfseries 34.92\,/\,0.026};
\node[cell] at (13.808,-3.842) {\includegraphics[width=1.656cm]{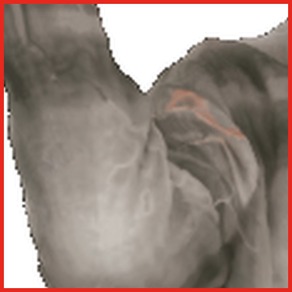}};
\node[cell] at (15.624,-0.480) {\includegraphics[width=1.656cm]{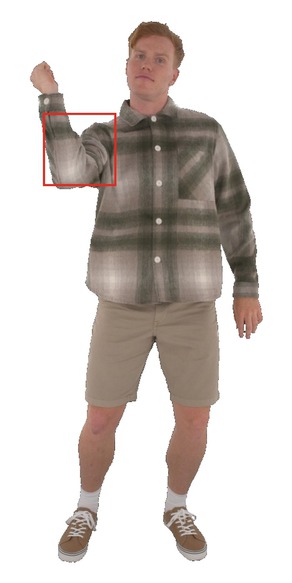}};
\node[cell] at (15.624,-3.842) {\includegraphics[width=1.656cm]{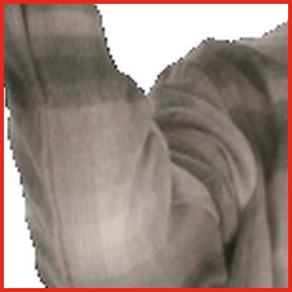}};
\node[anchor=north west, inner sep=1pt, fill=white, font=\scriptsize\itshape] at (0,-5.738) {ActorsHQ actor6\_seq2\_1081, held-out camera Cam024};
\node[cell] at (0.000,-6.078) {\includegraphics[width=1.656cm]{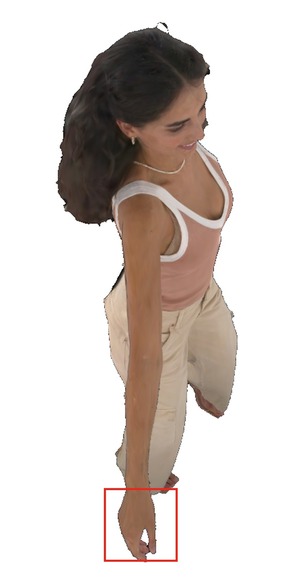}};
\node[val] at (1.656,-9.390) {35.05\,/\,0.029};
\node[cell] at (0.000,-9.440) {\includegraphics[width=1.656cm]{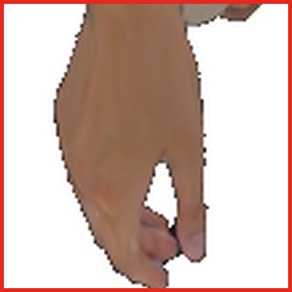}};
\node[cell] at (1.696,-6.078) {\includegraphics[width=1.656cm]{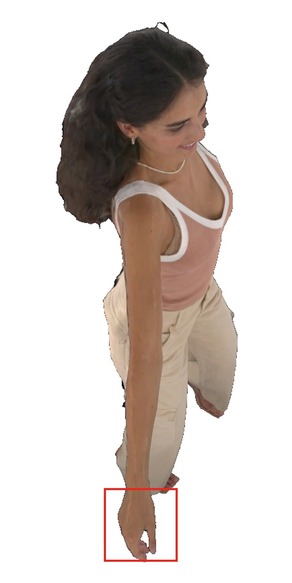}};
\node[val] at (3.352,-9.390) {32.65\,/\,0.030};
\node[cell] at (1.696,-9.440) {\includegraphics[width=1.656cm]{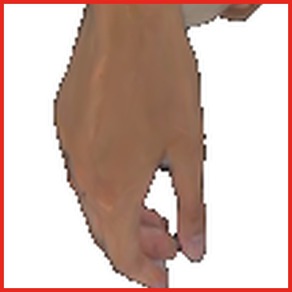}};
\node[cell] at (3.512,-6.078) {\includegraphics[width=1.656cm]{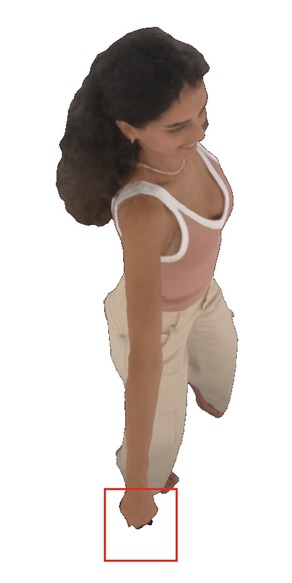}};
\node[val] at (5.168,-9.390) {30.33\,/\,0.054};
\node[cell] at (3.512,-9.440) {\includegraphics[width=1.656cm]{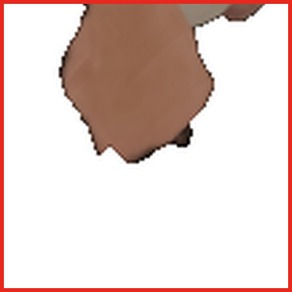}};
\node[cell] at (5.208,-6.078) {\includegraphics[width=1.656cm]{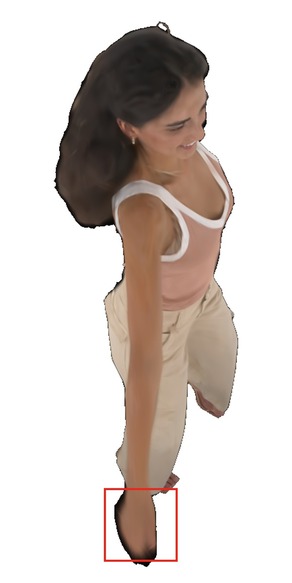}};
\node[val] at (6.864,-9.390) {33.39\,/\,0.068};
\node[cell] at (5.208,-9.440) {\includegraphics[width=1.656cm]{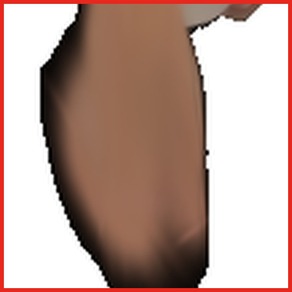}};
\node[cell] at (6.904,-6.078) {\includegraphics[width=1.656cm]{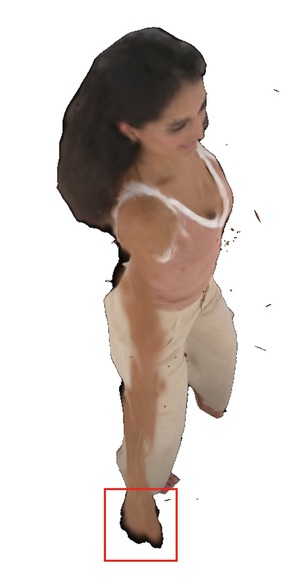}};
\node[val] at (8.560,-9.390) {28.74\,/\,0.110};
\node[cell] at (6.904,-9.440) {\includegraphics[width=1.656cm]{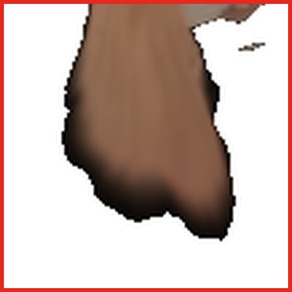}};
\node[cell] at (8.600,-6.078) {\includegraphics[width=1.656cm]{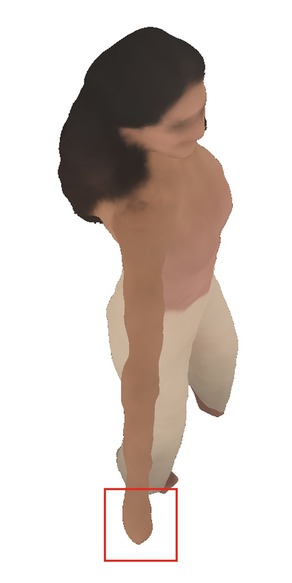}};
\node[val] at (10.256,-9.390) {25.96\,/\,0.091};
\node[cell] at (8.600,-9.440) {\includegraphics[width=1.656cm]{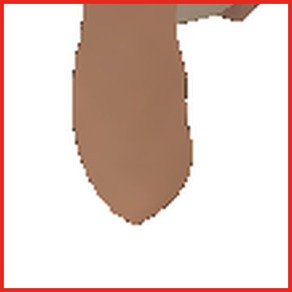}};
\node[cell] at (10.296,-6.078) {\includegraphics[width=1.656cm]{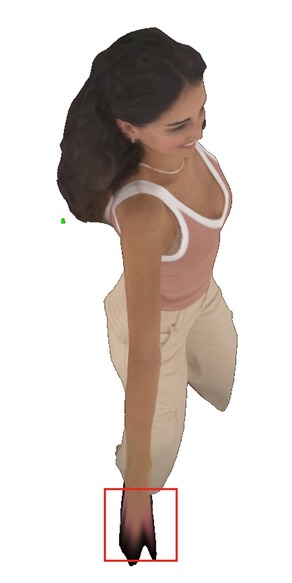}};
\node[val] at (11.952,-9.390) {28.43\,/\,0.079};
\node[cell] at (10.296,-9.440) {\includegraphics[width=1.656cm]{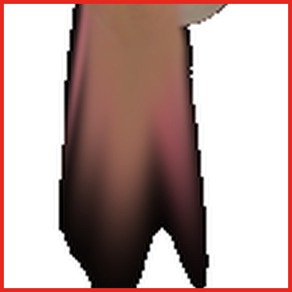}};
\node[cell] at (11.992,-6.078) {\includegraphics[width=1.656cm]{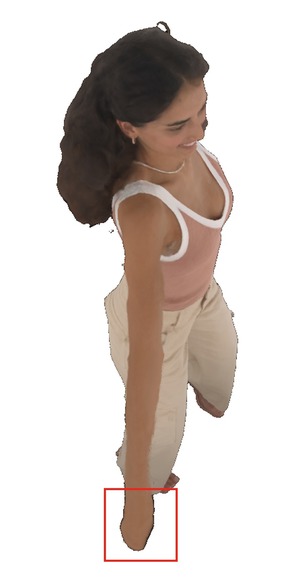}};
\node[val] at (13.648,-9.390) {32.74\,/\,0.040};
\node[cell] at (11.992,-9.440) {\includegraphics[width=1.656cm]{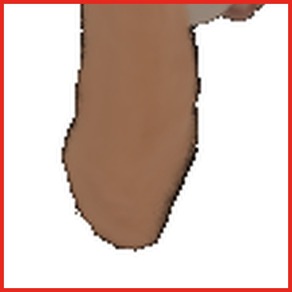}};
\node[cell] at (13.808,-6.078) {\includegraphics[width=1.656cm]{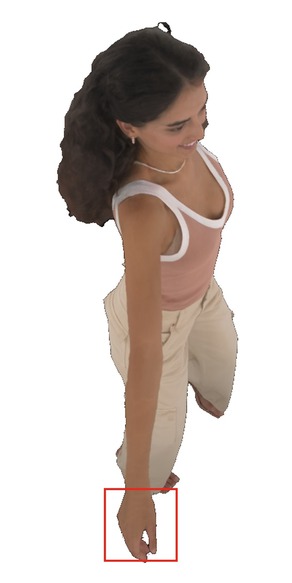}};
\node[val] at (15.464,-9.390) {\color[HTML]{0E80E5}\bfseries 36.51\,/\,0.028};
\node[cell] at (13.808,-9.440) {\includegraphics[width=1.656cm]{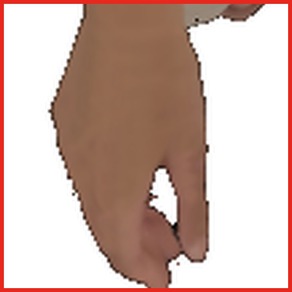}};
\node[cell] at (15.624,-6.078) {\includegraphics[width=1.656cm]{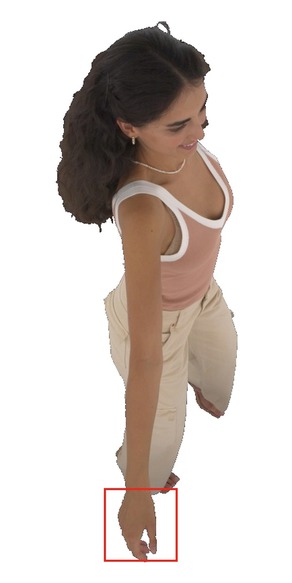}};
\node[cell] at (15.624,-9.440) {\includegraphics[width=1.656cm]{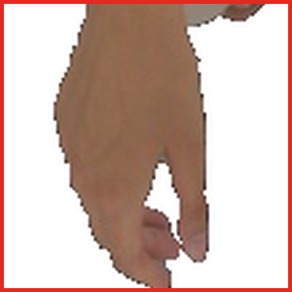}};
\node[anchor=north west, inner sep=1pt, fill=white, font=\scriptsize\itshape] at (0,-11.336) {ActorsHQ actor6\_seq2\_1604, held-out camera Cam152};
\node[cell] at (0.000,-11.676) {\includegraphics[width=1.656cm]{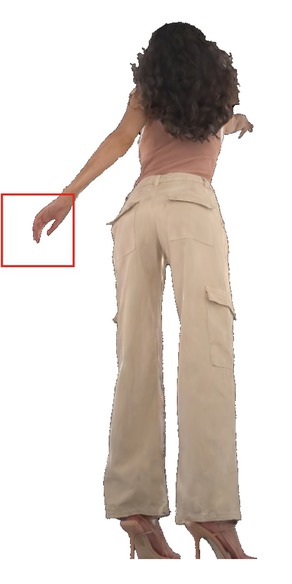}};
\node[val] at (1.656,-14.992) {31.38\,/\,0.032};
\node[cell] at (0.000,-15.042) {\includegraphics[width=1.656cm]{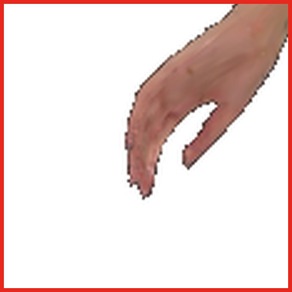}};
\node[cell] at (1.696,-11.676) {\includegraphics[width=1.656cm]{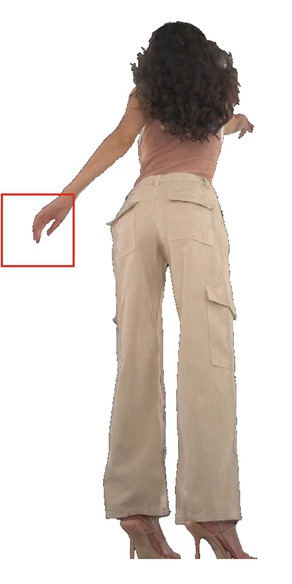}};
\node[val] at (3.352,-14.992) {28.87\,/\,0.041};
\node[cell] at (1.696,-15.042) {\includegraphics[width=1.656cm]{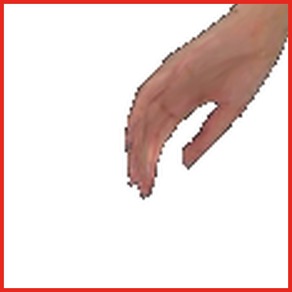}};
\node[cell] at (3.512,-11.676) {\includegraphics[width=1.656cm]{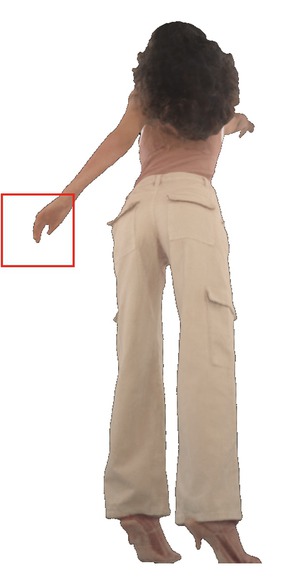}};
\node[val] at (5.168,-14.992) {29.46\,/\,0.059};
\node[cell] at (3.512,-15.042) {\includegraphics[width=1.656cm]{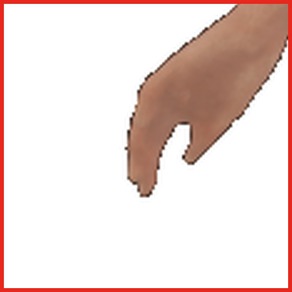}};
\node[cell] at (5.208,-11.676) {\includegraphics[width=1.656cm]{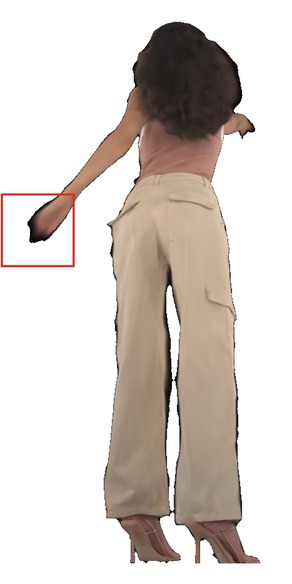}};
\node[val] at (6.864,-14.992) {28.95\,/\,0.104};
\node[cell] at (5.208,-15.042) {\includegraphics[width=1.656cm]{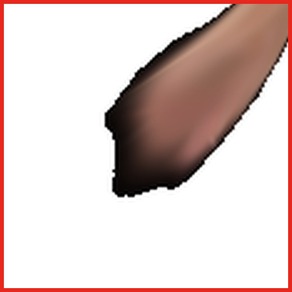}};
\node[cell] at (6.904,-11.676) {\includegraphics[width=1.656cm]{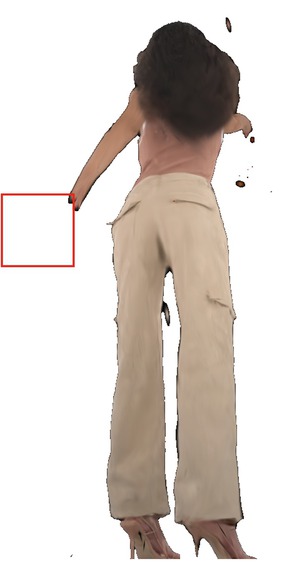}};
\node[val] at (8.560,-14.992) {25.85\,/\,0.123};
\node[cell] at (6.904,-15.042) {\includegraphics[width=1.656cm]{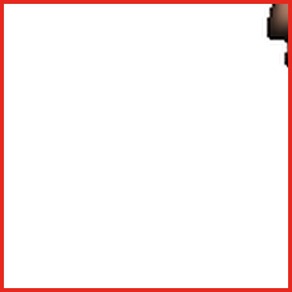}};
\node[cell] at (8.600,-11.676) {\includegraphics[width=1.656cm]{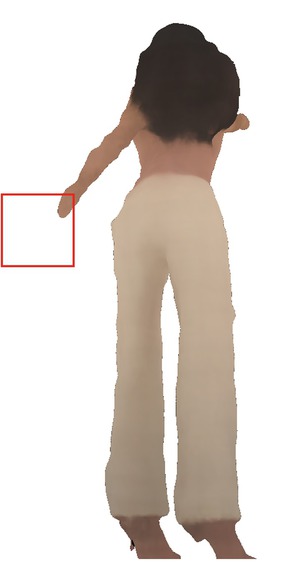}};
\node[val] at (10.256,-14.992) {24.98\,/\,0.097};
\node[cell] at (8.600,-15.042) {\includegraphics[width=1.656cm]{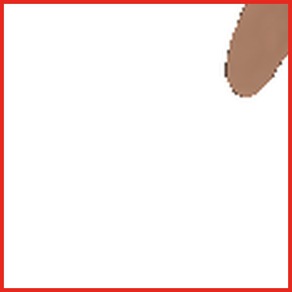}};
\node[cell] at (10.296,-11.676) {\includegraphics[width=1.656cm]{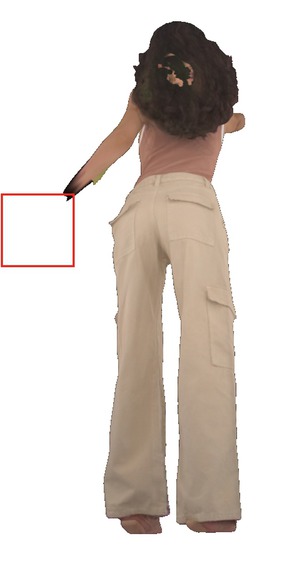}};
\node[val] at (11.952,-14.992) {25.34\,/\,0.080};
\node[cell] at (10.296,-15.042) {\includegraphics[width=1.656cm]{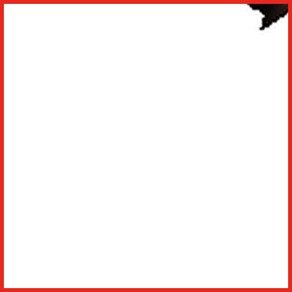}};
\node[cell] at (11.992,-11.676) {\includegraphics[width=1.656cm]{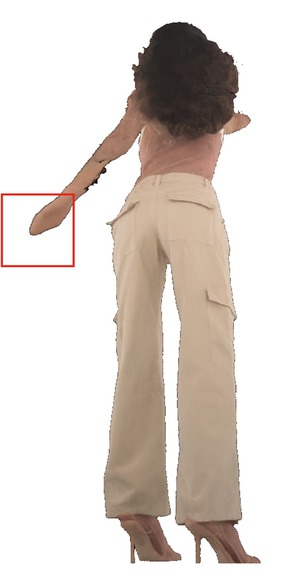}};
\node[val] at (13.648,-14.992) {29.28\,/\,0.054};
\node[cell] at (11.992,-15.042) {\includegraphics[width=1.656cm]{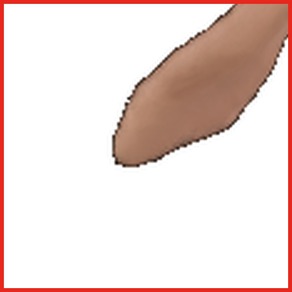}};
\node[cell] at (13.808,-11.676) {\includegraphics[width=1.656cm]{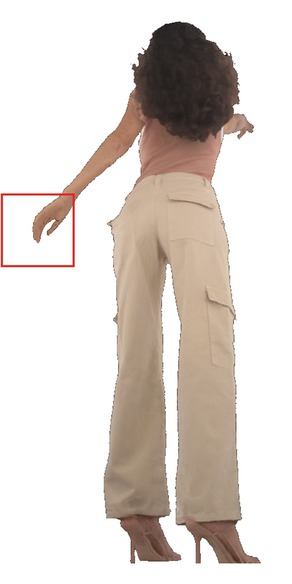}};
\node[val] at (15.464,-14.992) {\color[HTML]{0E80E5}\bfseries 32.47\,/\,0.039};
\node[cell] at (13.808,-15.042) {\includegraphics[width=1.656cm]{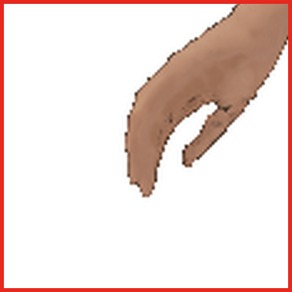}};
\node[cell] at (15.624,-11.676) {\includegraphics[width=1.656cm]{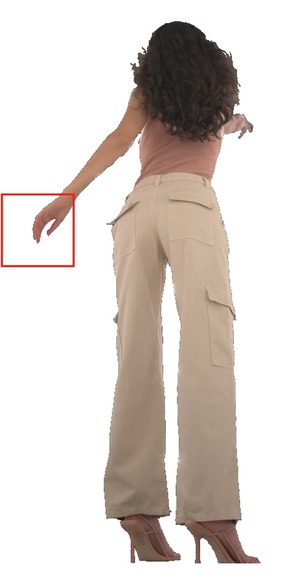}};
\node[cell] at (15.624,-15.042) {\includegraphics[width=1.656cm]{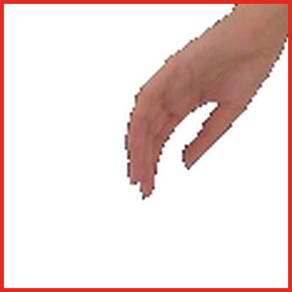}};
\end{tikzpicture}
\caption{\textbf{Novel-view synthesis compared with the baselines} (ActorsHQ): renders in unseen views, i.e.\ held-out cameras not used for training. Numbers: PSNR (dB) / LPIPS of the image, as in \cref{tab:nvs}, computed as in the evaluation on a black background (GT: the held-out photo with the background masked out); shown here on white. The red box marks the enlarged region (same box for every method). Left to right: per-frame static methods, dynamic methods, Ours, ground truth.}
\label{fig:nvs_cmp_2}
\end{figure*}

%% file: figures/nvs_cmp_taogs_num.tex
\begingroup\addtocounter{figure}{-1}\renewcommand{\thefigure}{S\arabic{figure}a}\input{figures/nvs_cmp_taogs}\endgroup

%% file: figures/nvs_cmp_taogs.tex
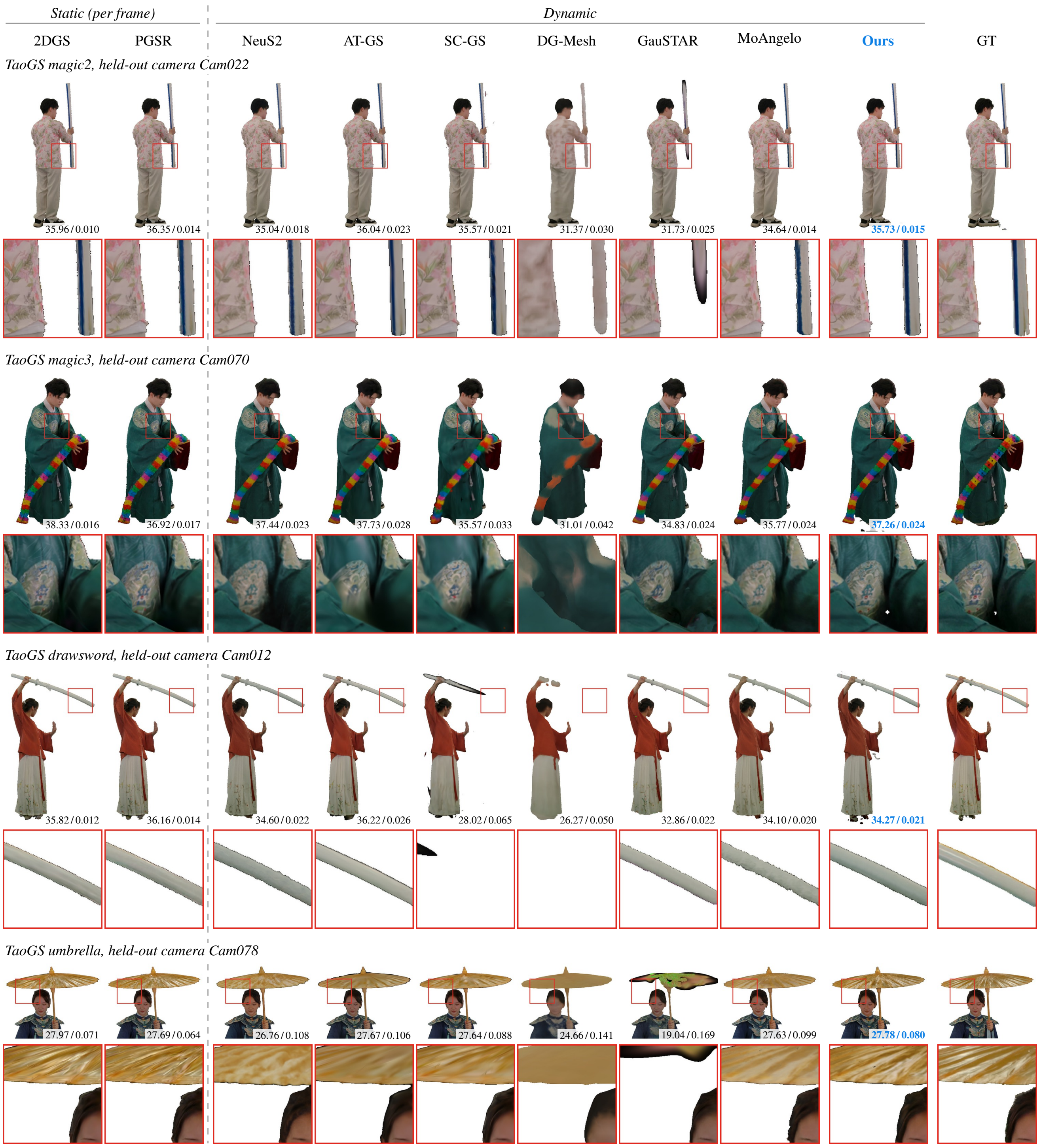
\begin{figure*}[p]
\centering
\begin{tikzpicture}[x=1cm, y=1cm, font=\scriptsize, cell/.style={anchor=north west, inner sep=0},
    val/.style={font=\tiny, anchor=south east, fill=white, fill opacity=0.85, text opacity=1, inner sep=1pt}]
\draw[black!45, line width=0.4pt] (0.050,0.420) -- (3.302,0.420);
\node[anchor=south, inner sep=1pt, font=\scriptsize\itshape] at (1.676,0.420) {Static (per frame)};
\draw[black!45, line width=0.4pt] (3.562,0.420) -- (15.414,0.420);
\node[anchor=south, inner sep=1pt, font=\scriptsize\itshape] at (9.488,0.420) {Dynamic};
\draw[black!40, dashed, line width=0.5pt] (3.432,0.720) -- (3.432,-18.302);
\node[anchor=south, inner sep=1pt] at (0.828,0.000) {2DGS};
\node[anchor=south, inner sep=1pt] at (2.524,0.000) {PGSR};
\node[anchor=south, inner sep=1pt] at (4.340,0.000) {NeuS2};
\node[anchor=south, inner sep=1pt] at (6.036,0.000) {AT-GS};
\node[anchor=south, inner sep=1pt] at (7.732,0.000) {SC-GS};
\node[anchor=south, inner sep=1pt] at (9.428,0.000) {DG-Mesh};
\node[anchor=south, inner sep=1pt] at (11.124,0.000) {GauSTAR};
\node[anchor=south, inner sep=1pt] at (12.820,0.000) {MoAngelo};
\node[anchor=south, inner sep=1pt] at (14.636,0.000) {\color[HTML]{0E80E5}\textbf{Ours}};
\node[anchor=south, inner sep=1pt] at (16.452,0.000) {GT};
\node[anchor=north west, inner sep=1pt, fill=white, font=\scriptsize\itshape] at (0,-0.140) {TaoGS magic2, held-out camera Cam022};
\node[cell] at (0.000,-0.480) {\includegraphics[width=1.656cm]{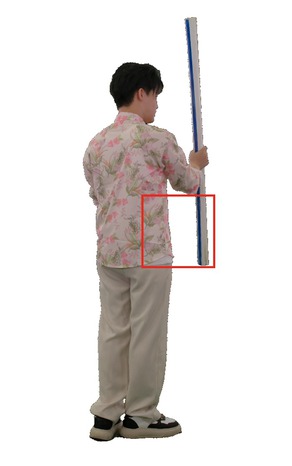}};
\node[val] at (1.656,-3.128) {35.96\,/\,0.010};
\node[cell] at (0.000,-3.178) {\includegraphics[width=1.656cm]{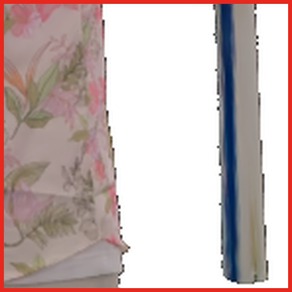}};
\node[cell] at (1.696,-0.480) {\includegraphics[width=1.656cm]{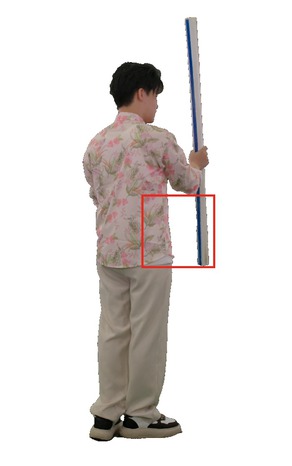}};
\node[val] at (3.352,-3.128) {36.35\,/\,0.014};
\node[cell] at (1.696,-3.178) {\includegraphics[width=1.656cm]{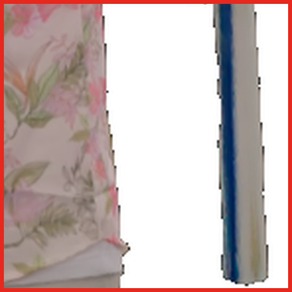}};
\node[cell] at (3.512,-0.480) {\includegraphics[width=1.656cm]{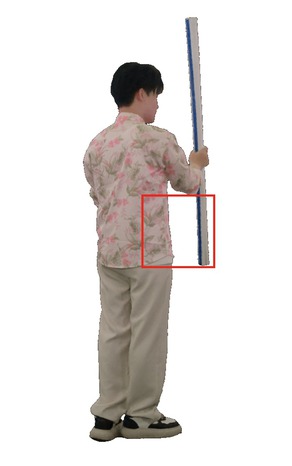}};
\node[val] at (5.168,-3.128) {35.04\,/\,0.018};
\node[cell] at (3.512,-3.178) {\includegraphics[width=1.656cm]{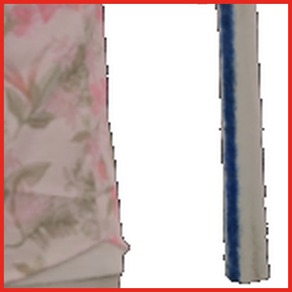}};
\node[cell] at (5.208,-0.480) {\includegraphics[width=1.656cm]{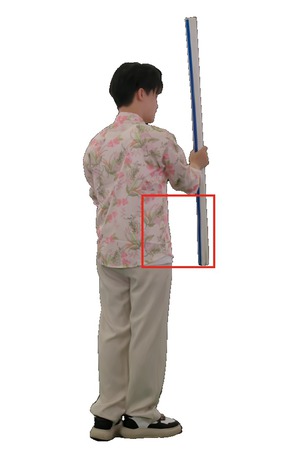}};
\node[val] at (6.864,-3.128) {36.04\,/\,0.023};
\node[cell] at (5.208,-3.178) {\includegraphics[width=1.656cm]{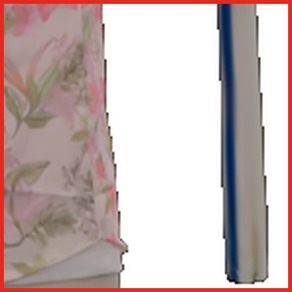}};
\node[cell] at (6.904,-0.480) {\includegraphics[width=1.656cm]{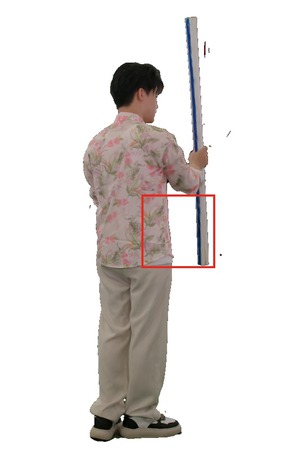}};
\node[val] at (8.560,-3.128) {35.57\,/\,0.021};
\node[cell] at (6.904,-3.178) {\includegraphics[width=1.656cm]{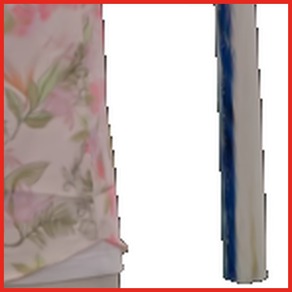}};
\node[cell] at (8.600,-0.480) {\includegraphics[width=1.656cm]{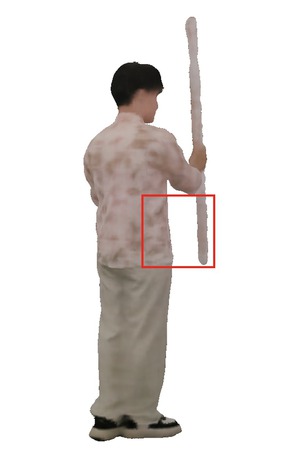}};
\node[val] at (10.256,-3.128) {31.37\,/\,0.030};
\node[cell] at (8.600,-3.178) {\includegraphics[width=1.656cm]{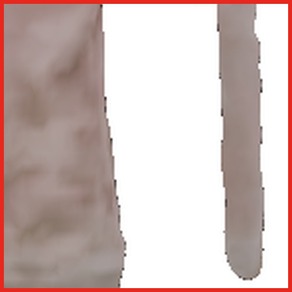}};
\node[cell] at (10.296,-0.480) {\includegraphics[width=1.656cm]{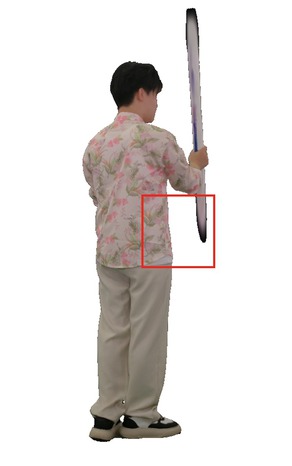}};
\node[val] at (11.952,-3.128) {31.73\,/\,0.025};
\node[cell] at (10.296,-3.178) {\includegraphics[width=1.656cm]{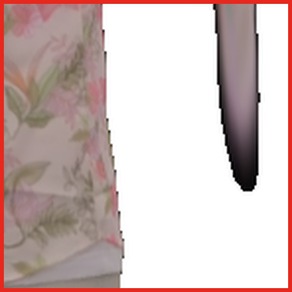}};
\node[cell] at (11.992,-0.480) {\includegraphics[width=1.656cm]{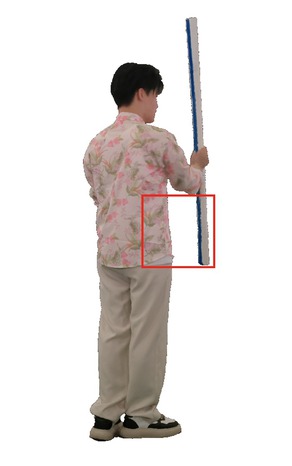}};
\node[val] at (13.648,-3.128) {34.64\,/\,0.014};
\node[cell] at (11.992,-3.178) {\includegraphics[width=1.656cm]{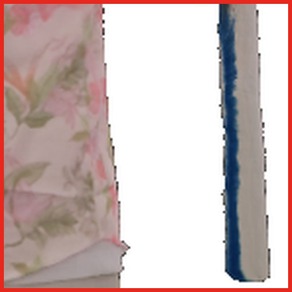}};
\node[cell] at (13.808,-0.480) {\includegraphics[width=1.656cm]{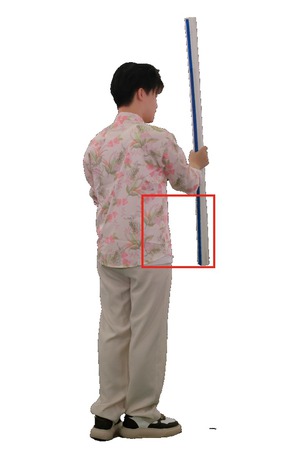}};
\node[val] at (15.464,-3.128) {\color[HTML]{0E80E5}\bfseries 35.73\,/\,0.015};
\node[cell] at (13.808,-3.178) {\includegraphics[width=1.656cm]{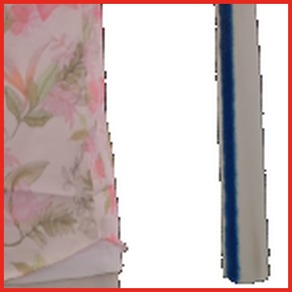}};
\node[cell] at (15.624,-0.480) {\includegraphics[width=1.656cm]{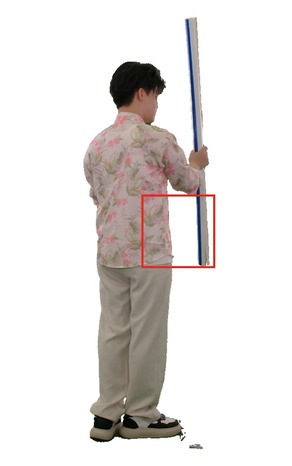}};
\node[cell] at (15.624,-3.178) {\includegraphics[width=1.656cm]{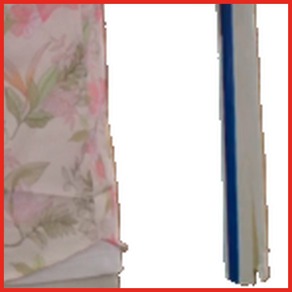}};
\node[anchor=north west, inner sep=1pt, fill=white, font=\scriptsize\itshape] at (0,-5.074) {TaoGS magic3, held-out camera Cam070};
\node[cell] at (0.000,-5.414) {\includegraphics[width=1.656cm]{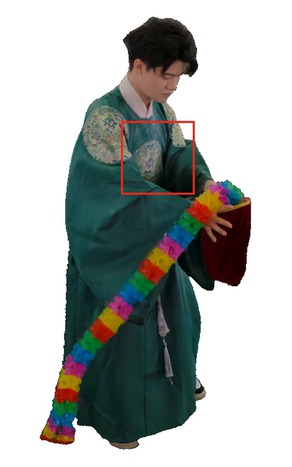}};
\node[val] at (1.656,-8.064) {38.33\,/\,0.016};
\node[cell] at (0.000,-8.114) {\includegraphics[width=1.656cm]{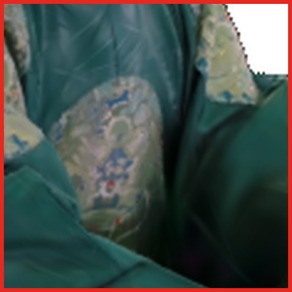}};
\node[cell] at (1.696,-5.414) {\includegraphics[width=1.656cm]{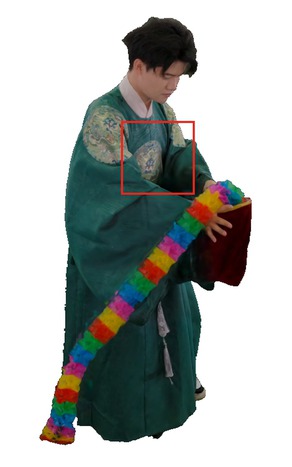}};
\node[val] at (3.352,-8.064) {36.92\,/\,0.017};
\node[cell] at (1.696,-8.114) {\includegraphics[width=1.656cm]{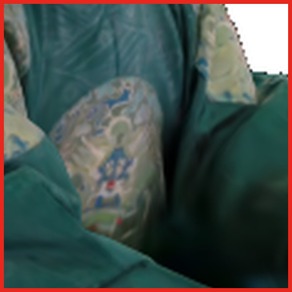}};
\node[cell] at (3.512,-5.414) {\includegraphics[width=1.656cm]{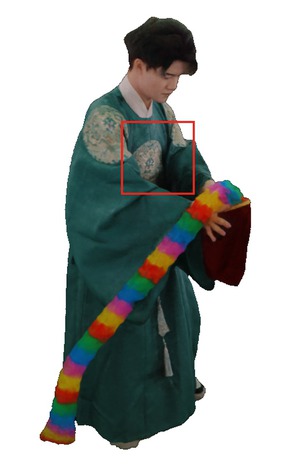}};
\node[val] at (5.168,-8.064) {37.44\,/\,0.023};
\node[cell] at (3.512,-8.114) {\includegraphics[width=1.656cm]{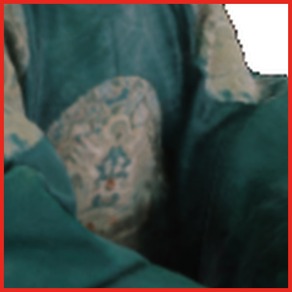}};
\node[cell] at (5.208,-5.414) {\includegraphics[width=1.656cm]{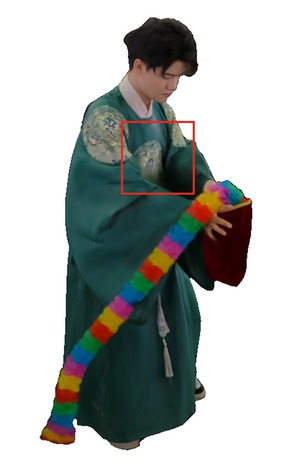}};
\node[val] at (6.864,-8.064) {37.73\,/\,0.028};
\node[cell] at (5.208,-8.114) {\includegraphics[width=1.656cm]{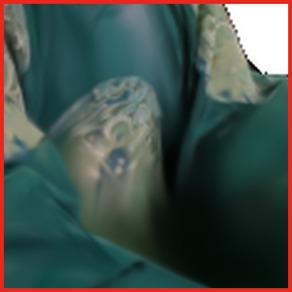}};
\node[cell] at (6.904,-5.414) {\includegraphics[width=1.656cm]{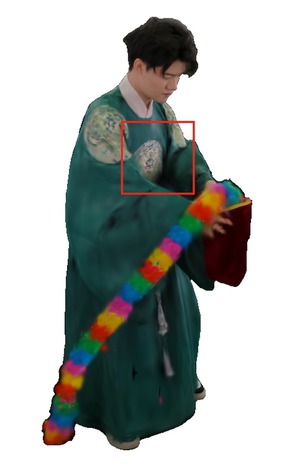}};
\node[val] at (8.560,-8.064) {35.57\,/\,0.033};
\node[cell] at (6.904,-8.114) {\includegraphics[width=1.656cm]{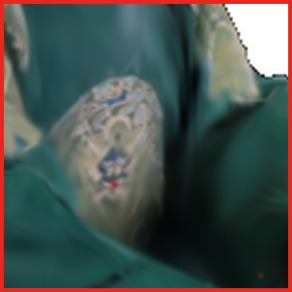}};
\node[cell] at (8.600,-5.414) {\includegraphics[width=1.656cm]{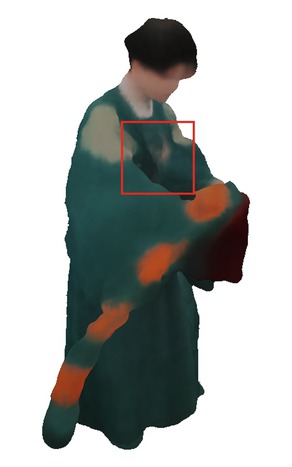}};
\node[val] at (10.256,-8.064) {31.01\,/\,0.042};
\node[cell] at (8.600,-8.114) {\includegraphics[width=1.656cm]{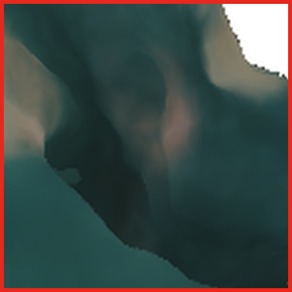}};
\node[cell] at (10.296,-5.414) {\includegraphics[width=1.656cm]{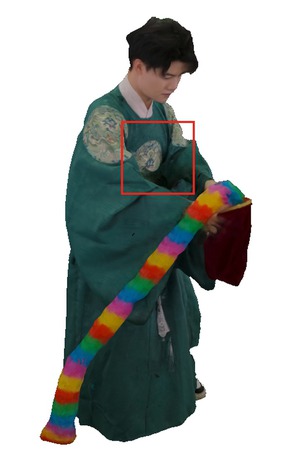}};
\node[val] at (11.952,-8.064) {34.83\,/\,0.024};
\node[cell] at (10.296,-8.114) {\includegraphics[width=1.656cm]{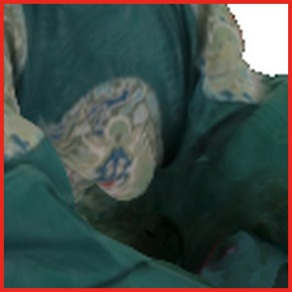}};
\node[cell] at (11.992,-5.414) {\includegraphics[width=1.656cm]{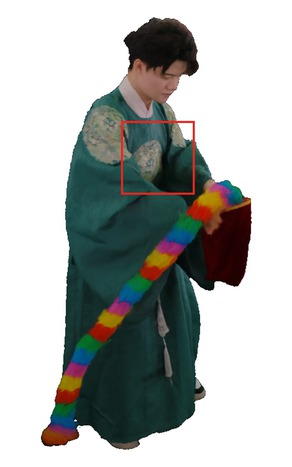}};
\node[val] at (13.648,-8.064) {35.77\,/\,0.024};
\node[cell] at (11.992,-8.114) {\includegraphics[width=1.656cm]{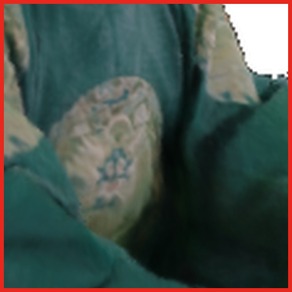}};
\node[cell] at (13.808,-5.414) {\includegraphics[width=1.656cm]{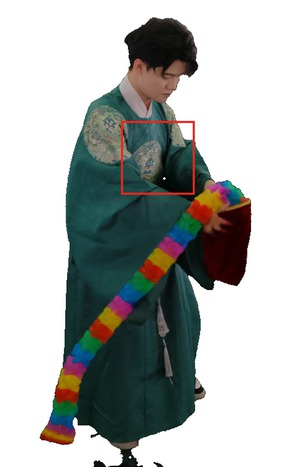}};
\node[val] at (15.464,-8.064) {\color[HTML]{0E80E5}\bfseries 37.26\,/\,0.024};
\node[cell] at (13.808,-8.114) {\includegraphics[width=1.656cm]{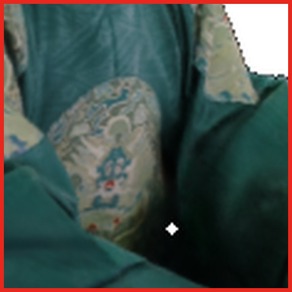}};
\node[cell] at (15.624,-5.414) {\includegraphics[width=1.656cm]{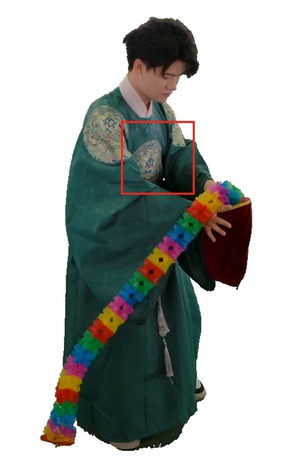}};
\node[cell] at (15.624,-8.114) {\includegraphics[width=1.656cm]{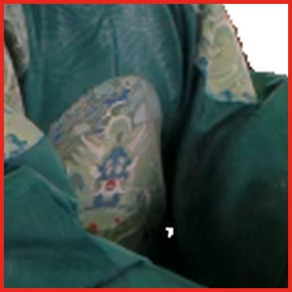}};
\node[anchor=north west, inner sep=1pt, fill=white, font=\scriptsize\itshape] at (0,-10.010) {TaoGS drawsword, held-out camera Cam012};
\node[cell] at (0.000,-10.350) {\includegraphics[width=1.656cm]{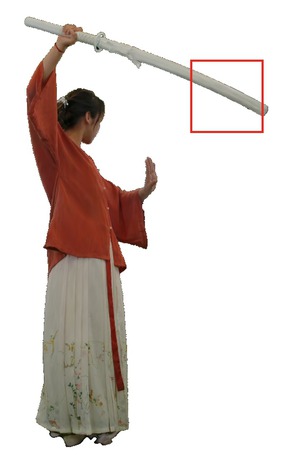}};
\node[val] at (1.656,-12.999) {35.82\,/\,0.012};
\node[cell] at (0.000,-13.049) {\includegraphics[width=1.656cm]{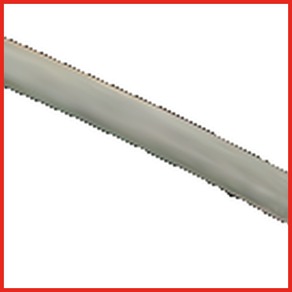}};
\node[cell] at (1.696,-10.350) {\includegraphics[width=1.656cm]{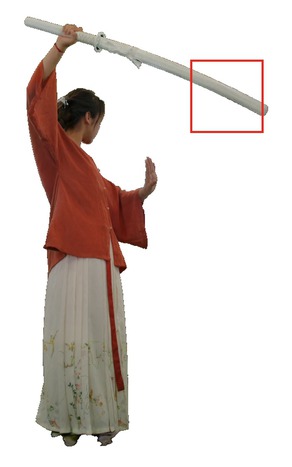}};
\node[val] at (3.352,-12.999) {36.16\,/\,0.014};
\node[cell] at (1.696,-13.049) {\includegraphics[width=1.656cm]{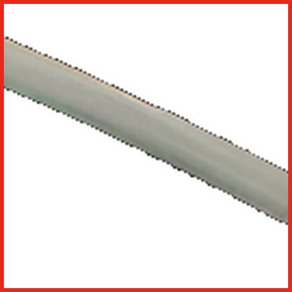}};
\node[cell] at (3.512,-10.350) {\includegraphics[width=1.656cm]{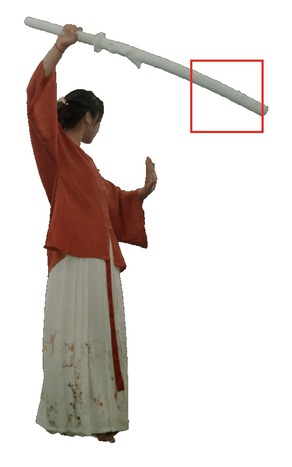}};
\node[val] at (5.168,-12.999) {34.60\,/\,0.022};
\node[cell] at (3.512,-13.049) {\includegraphics[width=1.656cm]{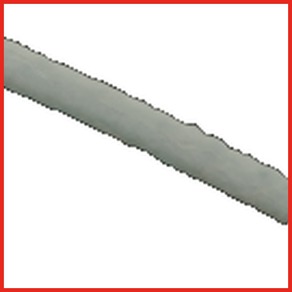}};
\node[cell] at (5.208,-10.350) {\includegraphics[width=1.656cm]{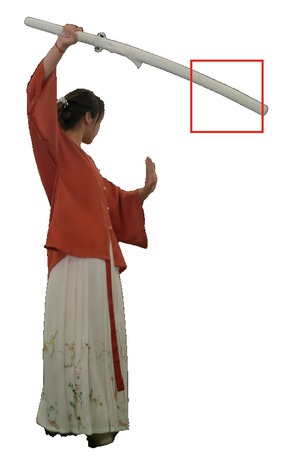}};
\node[val] at (6.864,-12.999) {36.22\,/\,0.026};
\node[cell] at (5.208,-13.049) {\includegraphics[width=1.656cm]{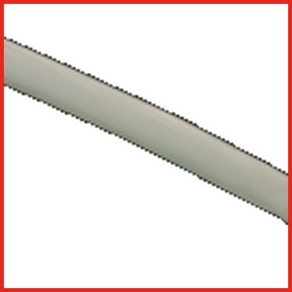}};
\node[cell] at (6.904,-10.350) {\includegraphics[width=1.656cm]{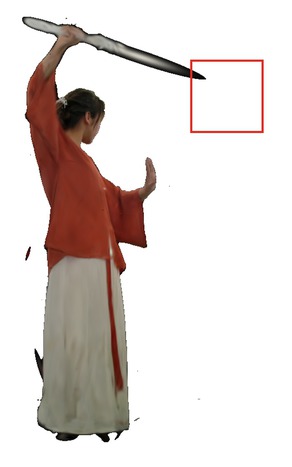}};
\node[val] at (8.560,-12.999) {28.02\,/\,0.065};
\node[cell] at (6.904,-13.049) {\includegraphics[width=1.656cm]{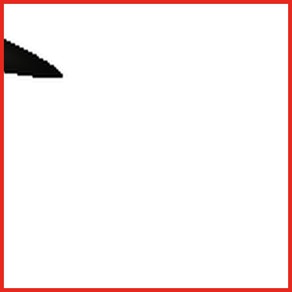}};
\node[cell] at (8.600,-10.350) {\includegraphics[width=1.656cm]{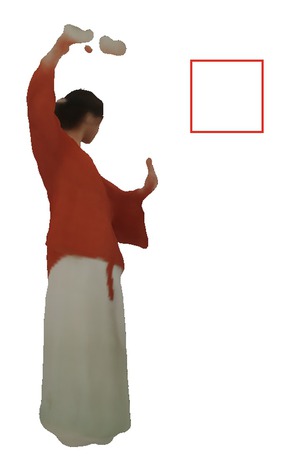}};
\node[val] at (10.256,-12.999) {26.27\,/\,0.050};
\node[cell] at (8.600,-13.049) {\includegraphics[width=1.656cm]{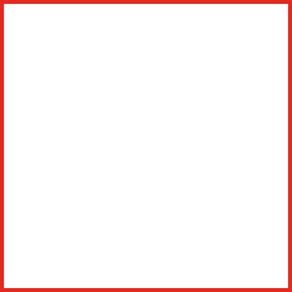}};
\node[cell] at (10.296,-10.350) {\includegraphics[width=1.656cm]{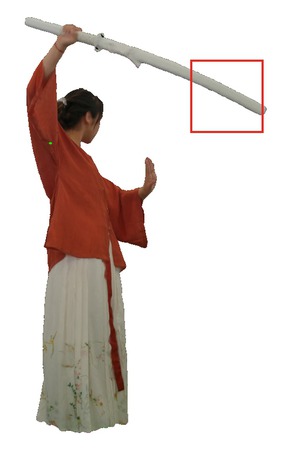}};
\node[val] at (11.952,-12.999) {32.86\,/\,0.022};
\node[cell] at (10.296,-13.049) {\includegraphics[width=1.656cm]{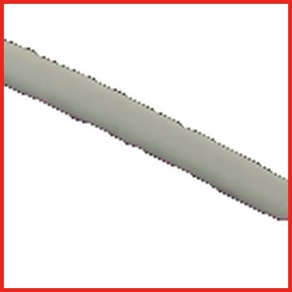}};
\node[cell] at (11.992,-10.350) {\includegraphics[width=1.656cm]{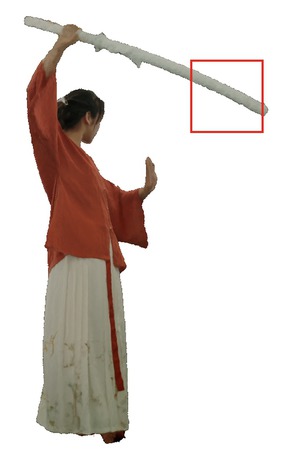}};
\node[val] at (13.648,-12.999) {34.10\,/\,0.020};
\node[cell] at (11.992,-13.049) {\includegraphics[width=1.656cm]{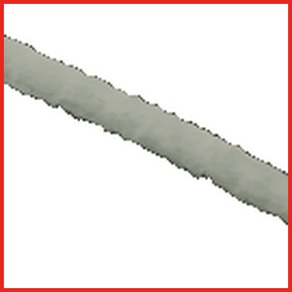}};
\node[cell] at (13.808,-10.350) {\includegraphics[width=1.656cm]{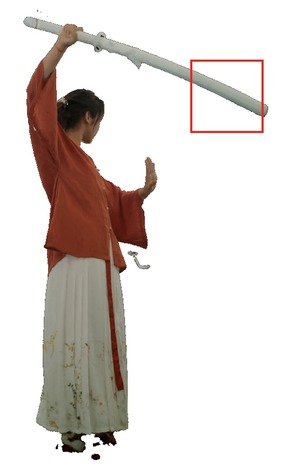}};
\node[val] at (15.464,-12.999) {\color[HTML]{0E80E5}\bfseries 34.27\,/\,0.021};
\node[cell] at (13.808,-13.049) {\includegraphics[width=1.656cm]{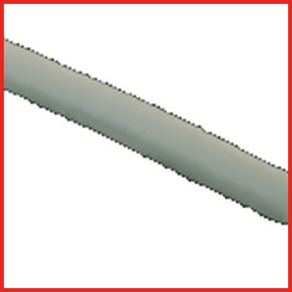}};
\node[cell] at (15.624,-10.350) {\includegraphics[width=1.656cm]{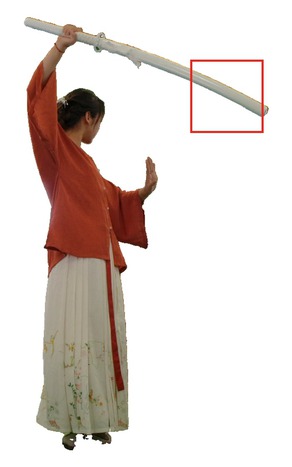}};
\node[cell] at (15.624,-13.049) {\includegraphics[width=1.656cm]{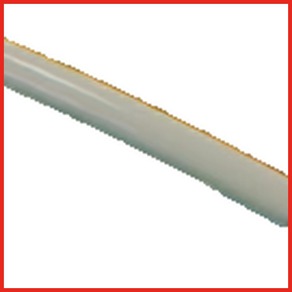}};
\node[anchor=north west, inner sep=1pt, fill=white, font=\scriptsize\itshape] at (0,-14.945) {TaoGS umbrella, held-out camera Cam078};
\node[cell] at (0.000,-15.285) {\includegraphics[width=1.656cm]{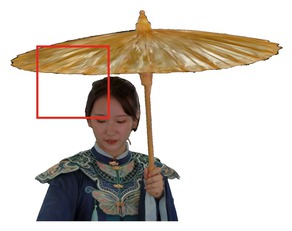}};
\node[val] at (1.656,-16.596) {27.97\,/\,0.071};
\node[cell] at (0.000,-16.646) {\includegraphics[width=1.656cm]{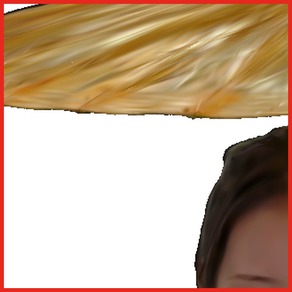}};
\node[cell] at (1.696,-15.285) {\includegraphics[width=1.656cm]{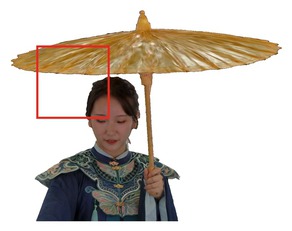}};
\node[val] at (3.352,-16.596) {27.69\,/\,0.064};
\node[cell] at (1.696,-16.646) {\includegraphics[width=1.656cm]{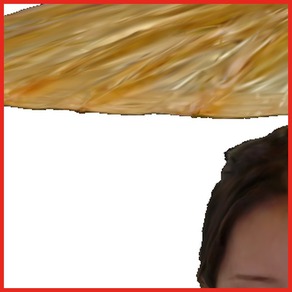}};
\node[cell] at (3.512,-15.285) {\includegraphics[width=1.656cm]{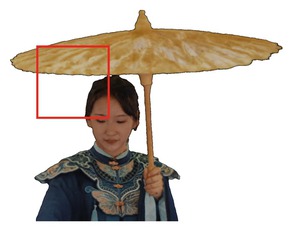}};
\node[val] at (5.168,-16.596) {26.76\,/\,0.108};
\node[cell] at (3.512,-16.646) {\includegraphics[width=1.656cm]{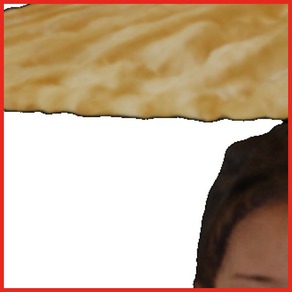}};
\node[cell] at (5.208,-15.285) {\includegraphics[width=1.656cm]{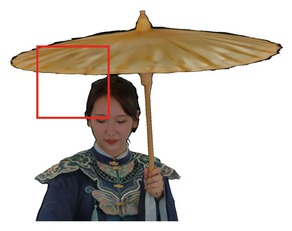}};
\node[val] at (6.864,-16.596) {27.67\,/\,0.106};
\node[cell] at (5.208,-16.646) {\includegraphics[width=1.656cm]{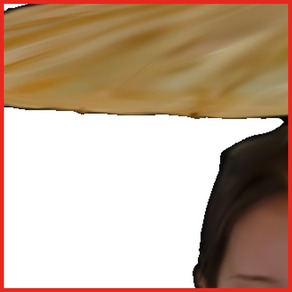}};
\node[cell] at (6.904,-15.285) {\includegraphics[width=1.656cm]{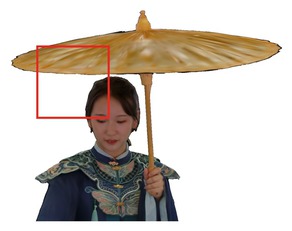}};
\node[val] at (8.560,-16.596) {27.64\,/\,0.088};
\node[cell] at (6.904,-16.646) {\includegraphics[width=1.656cm]{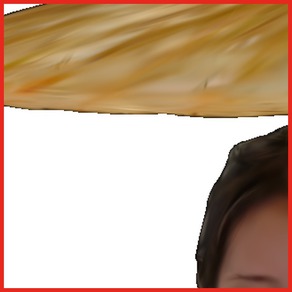}};
\node[cell] at (8.600,-15.285) {\includegraphics[width=1.656cm]{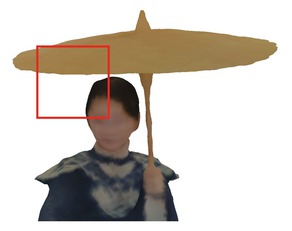}};
\node[val] at (10.256,-16.596) {24.66\,/\,0.141};
\node[cell] at (8.600,-16.646) {\includegraphics[width=1.656cm]{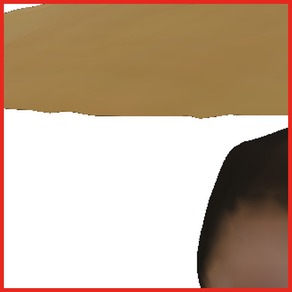}};
\node[cell] at (10.296,-15.285) {\includegraphics[width=1.656cm]{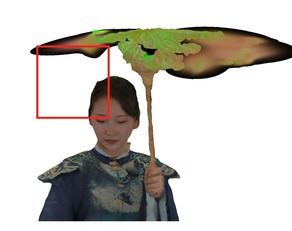}};
\node[val] at (11.952,-16.596) {19.04\,/\,0.169};
\node[cell] at (10.296,-16.646) {\includegraphics[width=1.656cm]{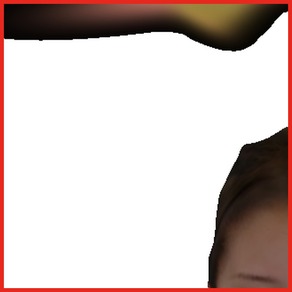}};
\node[cell] at (11.992,-15.285) {\includegraphics[width=1.656cm]{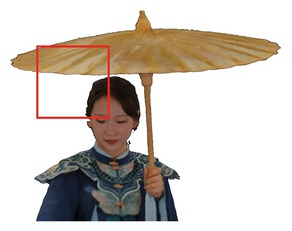}};
\node[val] at (13.648,-16.596) {27.63\,/\,0.099};
\node[cell] at (11.992,-16.646) {\includegraphics[width=1.656cm]{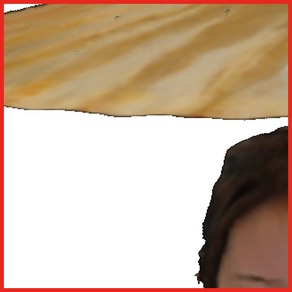}};
\node[cell] at (13.808,-15.285) {\includegraphics[width=1.656cm]{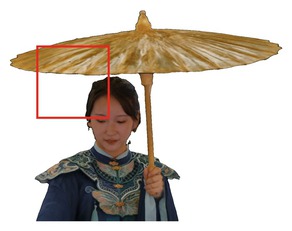}};
\node[val] at (15.464,-16.596) {\color[HTML]{0E80E5}\bfseries 27.78\,/\,0.080};
\node[cell] at (13.808,-16.646) {\includegraphics[width=1.656cm]{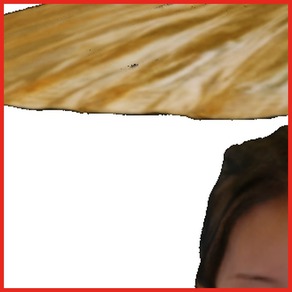}};
\node[cell] at (15.624,-15.285) {\includegraphics[width=1.656cm]{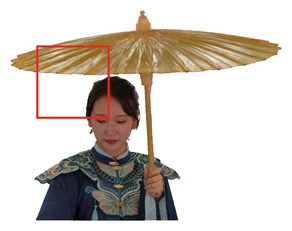}};
\node[cell] at (15.624,-16.646) {\includegraphics[width=1.656cm]{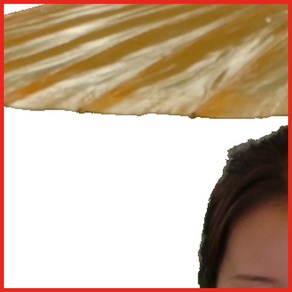}};
\end{tikzpicture}
\caption{\textbf{Novel-view synthesis compared with the baselines} (TaoGS): renders in unseen views, i.e.\ held-out cameras not used for training. Numbers: PSNR (dB) / LPIPS of the image, as in \cref{tab:nvs_taogs}, computed as in the evaluation on a black background (GT: the held-out photo with the background masked out); shown here on white. The red box marks the enlarged region (same box for every method). Left to right: per-frame static methods, dynamic methods, Ours, ground truth. Frame 25 of each scene; per scene, the held-out camera on which Ours has the median PSNR of its 5 cameras.}
\label{fig:nvs_cmp_taogs}
\end{figure*}

%% file: tables/correspondence_twostage.tex
\begin{table}[t]
\centering
\footnotesize
\setlength{\tabcolsep}{3pt}
\begin{tabular}{@{}lcccc@{}}
\toprule
Method & All & Early & Late & AUC$\uparrow$ \\
\midrule
PGSR~\cite{chen2024pgsr} + Hybrid FM, direct & 30.38 & 26.47 & 31.54 & 0.211 \\
PGSR~\cite{chen2024pgsr} + Hybrid FM, chained & 33.46 & 24.42 & 36.15 & 0.144 \\
\textbf{Ours} & \rankA{4.14} & \rankA{1.92} & \rankA{4.80} & \rankA{0.695} \\
\midrule
Ours + Hybrid FM, direct$^\ddagger$ & 24.87 & 20.76 & 26.09 & 0.320 \\
Ours + ULRSSM, direct$^\ddagger$ & 26.68 & 22.41 & 27.96 & 0.295 \\
\textbf{Ours}$^\ddagger$ & \rankA{3.93} & \rankA{1.59} & \rankA{4.63} & \rankA{0.705} \\
\bottomrule
\end{tabular}
\caption{\textbf{Two-stage pipelines} (DeformingThings4D): shape matching applied to reconstructed meshes, of PGSR and of our method; columns as in \Cref{tab:correspondence}. \bestlegend{} in each setting. $^\ddagger$On the points that both our flow fields and the matcher map.}
\label{tab:corr_twostage}
\end{table}

%% file: figures/drift_baselines.tex
\begin{figure*}[t]
\centering
\includegraphics[width=\textwidth]{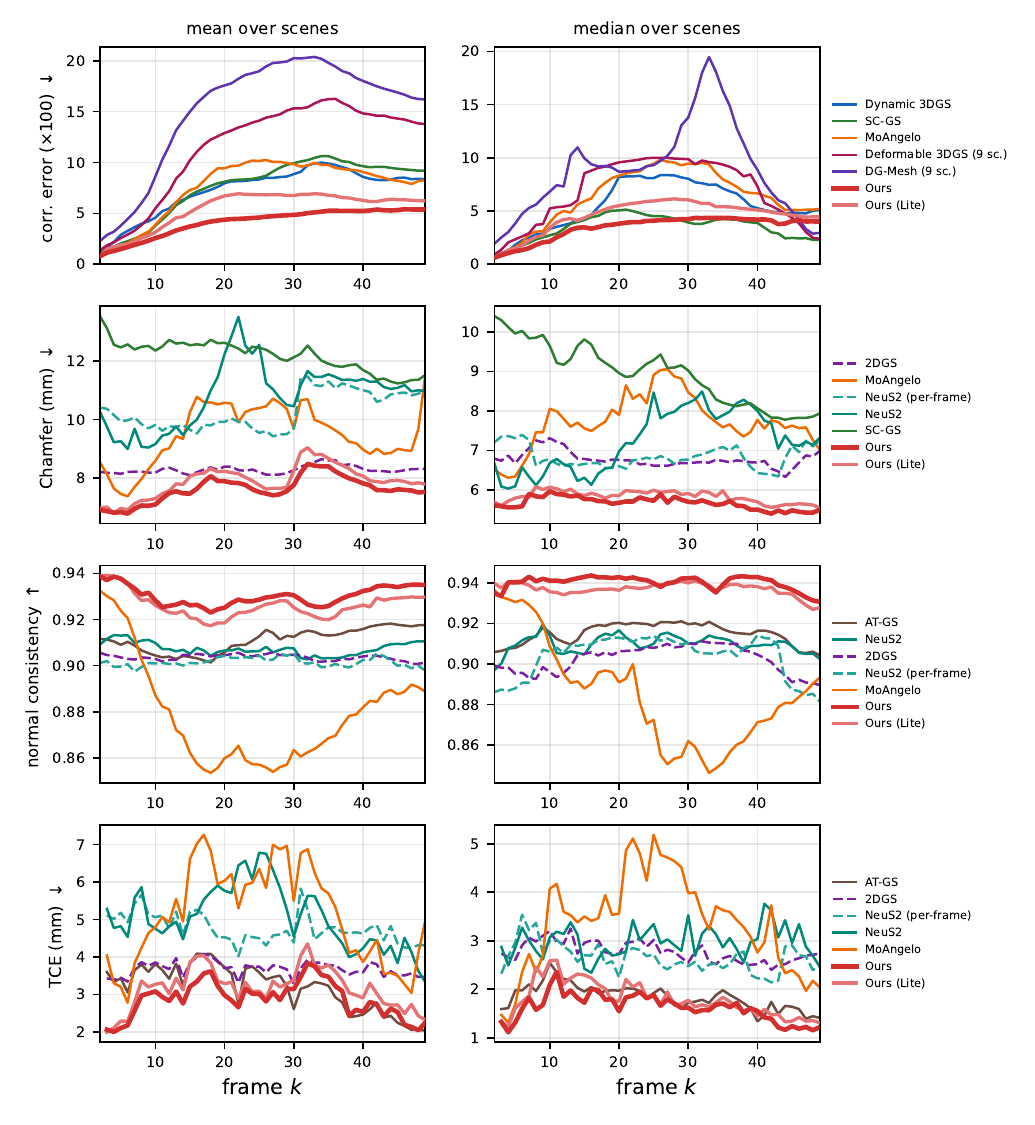}
\caption{\textbf{Accuracy over the sequence} on DeformingThings4D: correspondence error (geodesic error $\times 100$ on each method's own reconstruction, as in \Cref{tab:correspondence}), Chamfer distance, normal consistency and temporal consistency error (TCE; at frame~$k$, between frames $k-1$ and $k$), mean and median over the 10 scenes. Each row shows ours, Ours (Lite) and the five other methods with the best value of that metric over all frames and scenes (among the methods with at least 9 scenes); per-frame methods are dashed. Every curve averages to its value in \Cref{tab:correspondence,tab:perscene_cd,tab:perscene_nc,tab:perscene_tce}.}
\label{fig:drift_baselines}
\end{figure*}

%% file: figures/long200.tex
\begingroup\addtocounter{figure}{-1}\renewcommand{\thefigure}{S\arabic{figure}a}
\begin{figure*}[t]
\centering
\includegraphics[width=\textwidth]{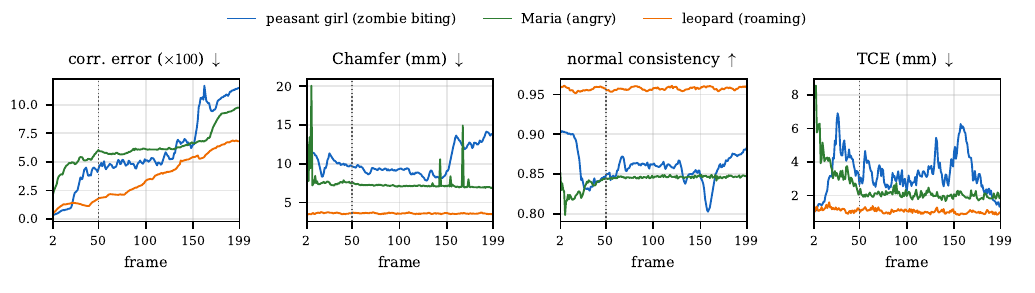}
\caption{\textbf{Accuracy over 200 frames} on three DeformingThings4D sequences of 200 frames, reconstructed with our full model with all iteration counts of the schedule scaled by 0.2 (\cref{sec:supp_iters}): correspondence error (geodesic error $\times 100$ on our reconstruction, as in \Cref{tab:correspondence}), Chamfer distance, normal consistency and temporal consistency error (TCE; at frame~$k$, between frames $k-1$ and $k$). The dotted line marks frame~50, the sequence length of all other evaluations.}
\label{fig:long200}
\end{figure*}
\endgroup

%% file: figures/cd_vs_corr_detail.tex
\begin{figure}[t]
\centering
\includegraphics[width=\columnwidth]{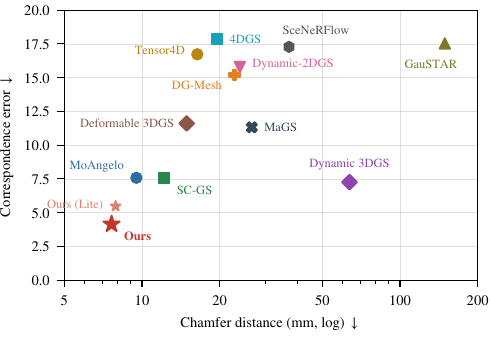}
\caption{\textbf{Surface accuracy vs.\ correspondence accuracy} on DeformingThings4D: mean Chamfer distance (\Cref{tab:geometry_avg}) and mean geodesic correspondence error ($\times 100$, \Cref{tab:correspondence}) of each method, both $\downarrow$.}
\label{fig:cd_vs_corr_detail}
\end{figure}

%% file: tables/perscene_corr.tex
\begin{table*}[t]
\centering
\scriptsize
\setlength{\tabcolsep}{2.2pt}
\footnotesize
\begin{tabular}{@{}lccccccccccc@{}}
\toprule
 & \multicolumn{11}{c}{DeformingThings4D} \\
Method & abe & AJ & canie & dragon & fox & grizz & ninja & peasant. & rabbit & vampire & Mean \\
\midrule
\multicolumn{12}{@{}l}{\emph{Dynamic, SDF-based}} \\
Tensor4D & 13.77 & 5.41 & 34.35 & 32.07 & 25.57 & 3.91 & 7.10 & 14.08 & 24.45 & 6.62 & 16.73 \\
MoAngelo & 2.69 & 8.01 & 10.03 & 10.54 & 4.11 & 3.02 & 6.15 & \rankC{12.31} & 11.45 & 7.44 & 7.58 \\
\multicolumn{12}{@{}l}{\emph{Dynamic, NeRF-based}} \\
SceNeRFlow & 14.53 & 13.12 & 33.74 & 14.56 & 10.54 & 10.79 & 12.02 & 24.07 & 24.54 & 14.90 & 17.28 \\
\multicolumn{12}{@{}l}{\emph{Dynamic, Gaussian-based}} \\
Dynamic 3DGS & \rankB{1.20} & 8.16 & \rankA{4.05} & 13.42 & \rankA{2.57} & \rankB{1.70} & \rankC{3.74} & 14.57 & \rankC{10.77} & 12.43 & \rankC{7.26} \\
Dynamic-2DGS & 18.94 & 20.17 & -- & 15.59 & 8.41 & -- & -- & -- & -- & -- & 15.78$^\dagger$ \\
4DGS & 14.20 & 8.83 & 33.17 & 30.61 & 14.33 & 6.36 & 9.57 & 14.22 & 31.02 & 16.09 & 17.84 \\
SC-GS & 1.96 & \rankA{1.83} & 24.99 & \rankB{8.31} & \rankC{3.83} & \rankA{1.44} & \rankA{3.22} & \rankB{9.18} & 17.89 & \rankA{2.92} & 7.56 \\
Deformable 3DGS & -- & 2.53 & 30.23 & 9.73 & 6.26 & 2.48 & 6.99 & 13.13 & 26.99 & 6.12 & 11.61$^\dagger$ \\
\multicolumn{12}{@{}l}{\emph{Dynamic, mesh-based Gaussian}} \\
GauSTAR & 6.41 & 14.95 & 33.78 & 31.24 & 16.19 & 9.16 & 7.08 & 17.89 & 23.92 & 14.70 & 17.53 \\
MaGS & -- & \rankB{2.44} & 29.95 & 9.38 & 6.10 & \rankC{2.35} & 6.80 & 13.01 & 26.28 & 5.59 & 11.32$^\dagger$ \\
DG-Mesh & -- & 4.59 & 36.99 & 13.06 & 8.11 & 3.20 & 6.98 & 13.81 & 32.84 & 17.24 & 15.20$^\dagger$ \\
\midrule
\textbf{Ours} (Lite) & \rankA{0.85} & 4.06 & \rankB{4.77} & \rankC{8.34} & \rankB{3.37} & 3.45 & 4.87 & 12.81 & \rankA{7.70} & \rankC{4.56} & \rankB{5.48} \\
\textbf{Ours} & \rankC{1.77} & \rankC{2.45} & \rankC{6.90} & \rankA{4.24} & 3.99 & 2.47 & \rankB{3.64} & \rankA{2.50} & \rankB{9.33} & \rankB{4.06} & \rankA{4.14} \\
\bottomrule
\end{tabular}
\caption{\textbf{Per-scene correspondence error} on DeformingThings4D: mean geodesic error ($\times 100$, $\downarrow$) over the 50 frames, all query points, each method with its own correspondences (\cref{sec:supp_corr}). \ranklegend{} among the dynamic methods; --: failed; $^\dagger$mean over the scenes with a result.}
\label{tab:perscene_corr}
\end{table*}

%% file: tables/median_wins.tex
\begin{table}[t]
\centering
\footnotesize
\setlength{\tabcolsep}{3pt}
\resizebox{\columnwidth}{!}{\begin{tabular}{@{}lcc|cc|cc@{}}
\toprule
& \multicolumn{2}{c|}{ActorsHQ CD} & \multicolumn{2}{c|}{DT4D CD} & \multicolumn{2}{c}{DT4D corr.} \\
Method & Med. & Ours better & Med. & Ours better & Med. & Ours better \\
\midrule
Tensor4D & 7.68 & 6/6 & 10.47 & 9/9 & 13.93 & 10/10 \\
NeuS2 & 4.28 & 6/6 & 7.34 & 10/10 & -- & -- \\
MoAngelo & 3.64 & 6/6 & 7.78 & 10/10 & 7.72 & 10/10 \\
SceNeRFlow & 17.07 & 6/6 & 25.66 & 10/10 & 14.54 & 10/10 \\
Dynamic 3DGS & 7.29 & 6/6 & 64.65 & 10/10 & 6.11 & 6/10 \\
Dynamic-2DGS & 48.53 & 2/2 & 8.71$^\ast$ & 3/3 & 17.27 & 4/4 \\
AT-GS & 5.21 & 6/6 & 7.86 & 10/10 & -- & -- \\
4DGS & 9.60 & 6/6 & 16.38 & 10/10 & 14.28 & 10/10 \\
SC-GS & 6.69 & 6/6 & 8.87 & 10/10 & 3.53 & 5/10 \\
Deformable 3DGS & 7.84 & 6/6 & 12.19 & 9/9 & 6.99 & 9/9 \\
GauSTAR & 5.95 & 6/6 & 26.59 & 10/10 & 15.57 & 10/10 \\
DG-Mesh & 6.89 & 6/6 & 15.80 & 9/9 & 13.06 & 9/9 \\
MaGS & 13.46 & 6/6 & 23.66 & 9/9 & 6.80 & 7/9 \\
2DGS & 3.74 & 6/6 & 6.79 & 8/10 & -- & -- \\
PGSR & 2.94 & 1/6 & 9.68 & 8/10 & -- & -- \\
NeuS2 (per-frame) & 4.17 & 6/6 & 6.65 & 10/10 & -- & -- \\
\midrule
\textbf{Ours} & 3.16 & -- & 5.67 & -- & 3.82 & -- \\
\bottomrule
\end{tabular}}\caption{\textbf{Medians over scenes and win counts.} Med.: median over the scenes of the method (CD in mm; correspondence: geodesic error $\times 100$ on each method's own reconstruction); Ours better: number of scenes, among those with a result for the method, on which ours is lower. From the per-scene tables \Cref{tab:perscene_cd,tab:perscene_corr} and their ActorsHQ counterparts. --: no correspondences (NeuS2 and the per-frame methods) or not evaluated on DeformingThings4D (AT-GS, whose correspondences are evaluated on real captures, \cref{sec:supp_realdata}). $^\ast$Without its failed AJ geometry run; its AJ correspondences are scored.}
\label{tab:median_wins}
\end{table}

%% file: figures/pck_recon.tex
\begin{figure*}[t]
\centering
\includegraphics[width=\textwidth]{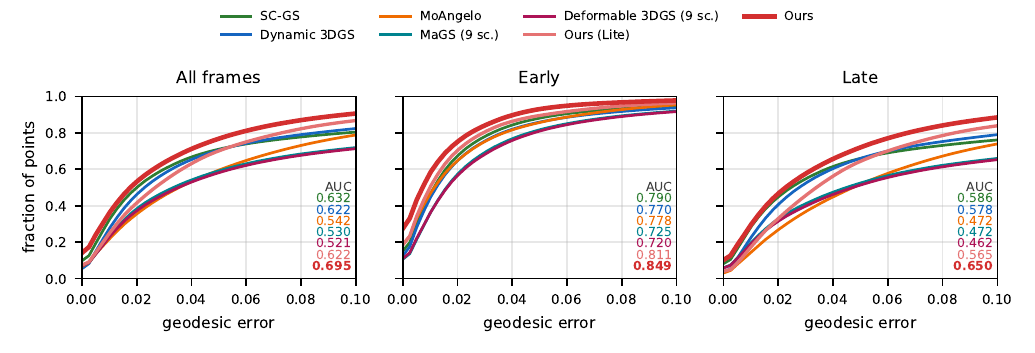}
\caption{\textbf{Cumulative correspondence error on reconstructed surfaces} (DeformingThings4D, 10 scenes): fraction of the points of frame~1 whose geodesic error in frame~$k$ (shape scaled to unit area) is below a threshold, each method on its own reconstruction (as in \Cref{tab:correspondence}), pooled over the frames of each range; numbers: AUC (best in bold). Shown: ours, Ours (Lite) and the five other methods with the highest AUC over all frames.}
\label{fig:pck_recon}
\end{figure*}

%% file: tables/map_quality.tex
\begin{table}[t]
\centering
\addtocounter{table}{-1}\renewcommand{\thetable}{S\arabic{table}a}
\footnotesize
\setlength{\tabcolsep}{3.5pt}
\begin{tabular}{@{}l|cc|cc|cc@{}}
\toprule
 & \multicolumn{2}{c|}{Distinct (\%)$\uparrow$} & \multicolumn{2}{c|}{Spacing$\rightarrow$1} & \multicolumn{2}{c}{Coverage (\%)$\uparrow$} \\
Scene & flow & canon. & flow & canon. & flow & canon. \\
\midrule
abe & 97.6 & 97.9 & 0.93 & 0.96 & 94.5 & 98.2 \\
AJ & 95.5 & 92.3 & 0.89 & 0.84 & 81.1 & 75.0 \\
canie & 96.8 & 90.9 & 0.70 & 0.86 & 71.1 & 78.9 \\
dragon & 97.4 & 71.8 & 0.79 & 0.27 & 88.1 & 57.8 \\
fox & 98.2 & 96.7 & 0.85 & 0.97 & 83.6 & 92.8 \\
grizz & 98.6 & 97.5 & 0.95 & 0.98 & 90.0 & 93.3 \\
ninja & 99.2 & 99.1 & 0.98 & 0.97 & 93.8 & 92.6 \\
peasantgirl & 97.9 & 83.0 & 0.91 & 0.81 & 85.4 & 80.7 \\
rabbit & 95.5 & 86.9 & 0.62 & 0.77 & 69.9 & 76.3 \\
vampire & 93.7 & 92.5 & 0.80 & 0.78 & 87.1 & 85.8 \\
\midrule
Mean & 97.0 & 90.9 & 0.84 & 0.82 & 84.4 & 83.1 \\
\bottomrule
\end{tabular}
\caption{\textbf{Quality of the correspondence maps} on DeformingThings4D at frame~49, for our correspondences (flow) and the canonical route of our model (canon.), on the 5000 query points of \Cref{tab:correspondence}. Distinct: share of the points that land on distinct vertices of our frame-49 mesh (100: one-to-one on the points). Spacing: median over the points of the distance to the nearest other tracked point, divided by the same distance for the true positions (1: no clustering). Coverage: share of the ground-truth surface of frame~49 within $r$ of a tracked point, relative to the same share for the true positions; $r$ is twice the median spacing of the true positions in frame~1.}
\label{tab:map_quality}
\end{table}

%% file: tables/correspondence_gt.tex
\begin{table}[t]
\centering
\footnotesize
\setlength{\tabcolsep}{3pt}
\begin{tabular}{@{}lcccc@{}}
\toprule
Method & All & Early & Late & AUC$\uparrow$ \\
\midrule
FM~\cite{ovsjanikov2012functional}, chained & 7.66 & 4.22 & 8.69 & 0.667 \\
ULRSSM~\cite{cao2023ulrssm}, chained & \rankC{4.73} & \rankC{2.95} & \rankC{5.25} & \rankB{0.735} \\
Hybrid FM~\cite{bastianxie2024hybrid}, chained & \rankA{3.88} & \rankB{2.31} & \rankA{4.35} & \rankA{0.761} \\
\textbf{Ours} & \rankB{3.99} & \rankA{1.73} & \rankB{4.67} & \rankC{0.704} \\
\bottomrule
\end{tabular}
\caption{\textbf{Correspondences on ground-truth surfaces}, a reference for shape matching, which needs these surfaces; columns as in \Cref{tab:correspondence}; \ranklegend.}
\label{tab:correspondence_gt}
\end{table}

%% file: tables/correspondence_detailed.tex
\begin{table*}[t]
\centering
\footnotesize
\setlength{\tabcolsep}{3pt}
\resizebox{\textwidth}{!}{\begin{tabular}{@{}lcccccccccc@{}}
\toprule
Scene & FM, direct & FM, chained & ULRSSM, direct & ULRSSM, chained & Hybrid, direct & Hybrid, chained & MoAngelo (rec.) & \textbf{Ours} Lite (rec.) & \textbf{Ours} (GT) & \textbf{Ours} (rec.) \\
\midrule
abe\_CoverToStand & 1.92 & \rankC{0.78} & 1.15 & \rankB{0.74} & 1.00 & \rankA{0.73} & 2.69 & 0.85 & 1.72 & 1.77 \\
AJ\_BigStomachHit & 12.23 & 4.77 & 4.67 & \rankB{1.91} & 3.43 & \rankA{1.60} & 8.01 & 4.06 & \rankC{2.23} & 2.45 \\
canieLTT\_Damaged4 & 16.26 & 13.88 & 10.43 & \rankB{4.21} & 9.80 & \rankA{2.99} & 10.03 & \rankC{4.77} & 6.94 & 6.90 \\
dragonOF2\_act11 & 41.60 & 7.81 & 33.36 & 10.36 & 32.72 & \rankC{5.65} & 10.54 & 8.34 & \rankA{4.14} & \rankB{4.24} \\
foxWDFS\_Damaged1 & 10.06 & 5.96 & 8.68 & \rankC{3.42} & 7.25 & \rankA{2.97} & 4.11 & \rankB{3.37} & 3.97 & 3.99 \\
grizzRSS\_Attack5 & 10.49 & 4.26 & 7.91 & \rankB{2.27} & 7.50 & \rankA{2.25} & 3.02 & 3.45 & \rankC{2.38} & 2.47 \\
ninja\_KettlebellSwing & 36.00 & 12.22 & \rankC{1.93} & \rankA{1.89} & \rankB{1.90} & \rankB{1.90} & 6.15 & 4.87 & 3.45 & 3.64 \\
peasantgirl\_Martelo2 & 23.24 & \rankC{11.36} & 17.50 & 13.84 & 16.95 & 13.17 & 12.31 & 12.81 & \rankA{2.46} & \rankB{2.50} \\
rabbitFKP\_Damaged2 & 18.45 & 10.94 & 8.13 & \rankB{6.68} & \rankC{6.77} & \rankA{5.95} & 11.45 & 7.70 & 9.30 & 9.33 \\
vampire\_MmaKick & 11.71 & 4.66 & 4.75 & \rankB{1.94} & 5.23 & \rankA{1.65} & 7.44 & 4.56 & \rankC{3.35} & 4.06 \\
\midrule
Average & 18.20 & 7.66 & 9.85 & 4.73 & 9.25 & \rankA{3.88} & 7.58 & 5.48 & \rankB{3.99} & \rankC{4.14} \\
\bottomrule
\end{tabular}}
\caption{\textbf{Per-scene correspondence accuracy on DeformingThings4D}, mean geodesic error ($\times 100$, shape scaled to unit area, $\downarrow$). GT: on the ground-truth surfaces; rec.: on the method's own reconstruction (ours, Ours Lite or MoAngelo). Hybrid: Hybrid FM~\cite{bastianxie2024hybrid}. \ranklegend{} per scene.}
\label{tab:correspondence_detailed}
\end{table*}

%% file: figures/sm_gt.tex
\begin{figure*}[p]
\centering
\definecolor{tzOurs}{RGB}{14,128,229}\definecolor{tzInk}{RGB}{24,28,40}
\begin{tikzpicture}[x=1cm, y=1cm, font=\scriptsize, text=tzInk, img/.style={anchor=south west, inner sep=0},
    val/.style={font=\tiny, anchor=south east, fill=white, fill opacity=0.8, text opacity=1, inner sep=1pt}]
\node[anchor=south west, inner sep=0, font=\footnotesize] at (0.400,16.982) {peasantgirl\_Martelo2};
\node[anchor=south, inner sep=1pt] at (1.201,16.642) {frame 1};
\node[anchor=south, inner sep=1pt] at (2.803,16.642) {13};
\node[anchor=south, inner sep=1pt] at (4.406,16.642) {25};
\node[anchor=south, inner sep=1pt] at (6.008,16.642) {37};
\node[anchor=south, inner sep=1pt] at (7.610,16.642) {49};
\node[img] at (0.400,14.775) {\includegraphics[width=1.602cm,height=1.847cm]{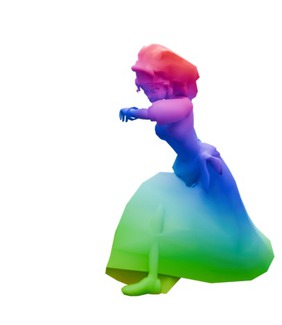}};
\node[img] at (2.002,14.775) {\includegraphics[width=1.602cm,height=1.847cm]{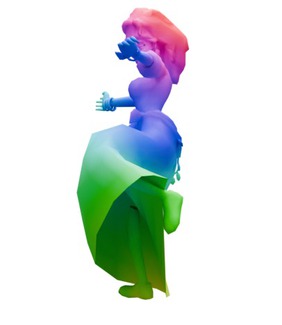}};
\node[img] at (3.605,14.775) {\includegraphics[width=1.602cm,height=1.847cm]{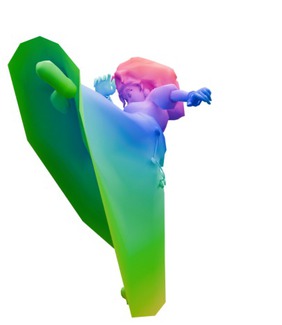}};
\node[img] at (5.207,14.775) {\includegraphics[width=1.602cm,height=1.847cm]{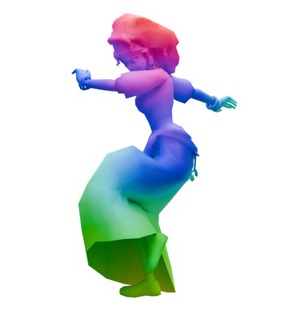}};
\node[img] at (6.809,14.775) {\includegraphics[width=1.602cm,height=1.847cm]{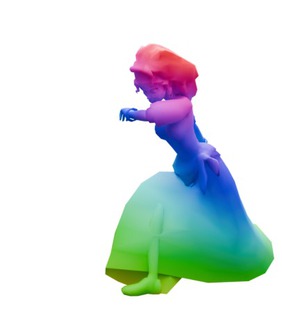}};
\node[rotate=90, anchor=south, inner sep=1pt] at (0.400,15.699) {GT};
\node[rotate=90, anchor=south, inner sep=1pt, align=center, font=\tiny] at (0.400,12.929) {FM};
\node[img] at (0.400,12.929) {\includegraphics[width=1.602cm,height=1.847cm]{imgs/sm_fig/peasantgirl/src}};
\node[img] at (0.400,11.082) {\includegraphics[width=1.602cm,height=1.847cm]{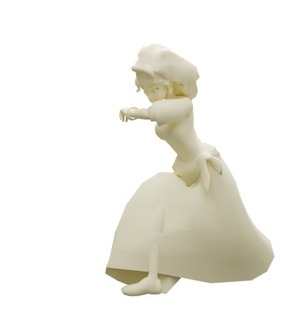}};
\node[val] at (1.982,11.102) {0};
\node[img] at (2.002,12.929) {\includegraphics[width=1.602cm,height=1.847cm]{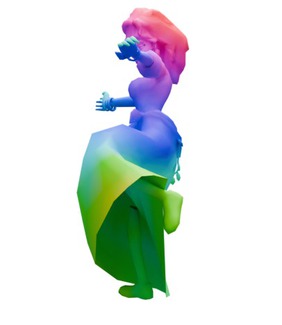}};
\node[img] at (2.002,11.082) {\includegraphics[width=1.602cm,height=1.847cm]{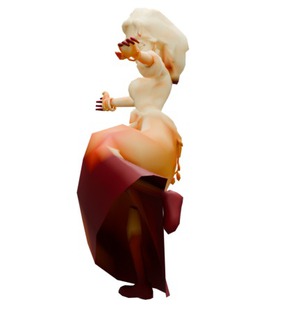}};
\node[val] at (3.585,11.102) {12.19};
\node[img] at (3.605,12.929) {\includegraphics[width=1.602cm,height=1.847cm]{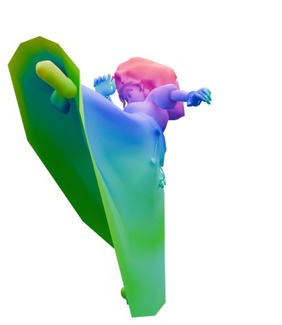}};
\node[img] at (3.605,11.082) {\includegraphics[width=1.602cm,height=1.847cm]{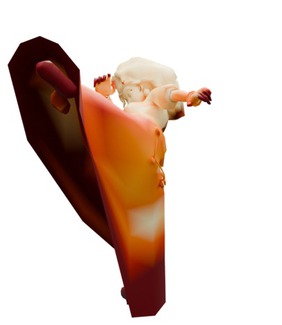}};
\node[val] at (5.187,11.102) {15.72};
\node[img] at (5.207,12.929) {\includegraphics[width=1.602cm,height=1.847cm]{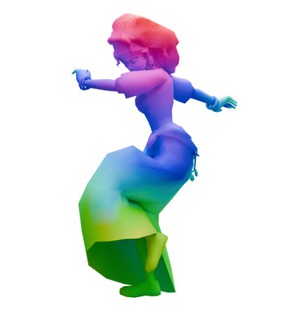}};
\node[img] at (5.207,11.082) {\includegraphics[width=1.602cm,height=1.847cm]{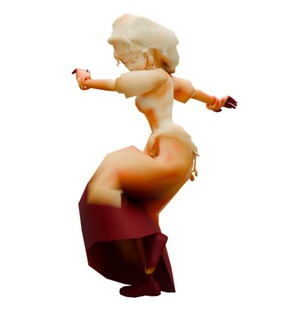}};
\node[val] at (6.789,11.102) {11.32};
\node[img] at (6.809,12.929) {\includegraphics[width=1.602cm,height=1.847cm]{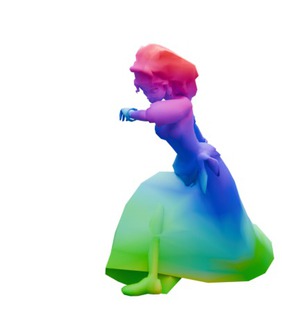}};
\node[img] at (6.809,11.082) {\includegraphics[width=1.602cm,height=1.847cm]{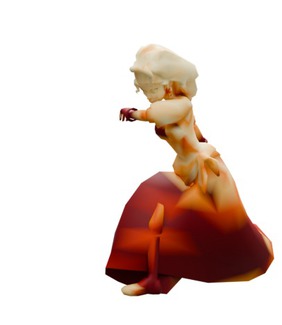}};
\node[val] at (8.391,11.102) {12.99};
\node[rotate=90, anchor=south, inner sep=1pt, align=center, font=\tiny] at (0.400,9.235) {ULRSSM};
\node[img] at (0.400,9.235) {\includegraphics[width=1.602cm,height=1.847cm]{imgs/sm_fig/peasantgirl/src}};
\node[img] at (0.400,7.388) {\includegraphics[width=1.602cm,height=1.847cm]{imgs/sm_fig/peasantgirl/err_src}};
\node[val] at (1.982,7.408) {0};
\node[img] at (2.002,9.235) {\includegraphics[width=1.602cm,height=1.847cm]{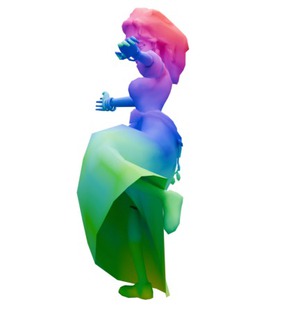}};
\node[img] at (2.002,7.388) {\includegraphics[width=1.602cm,height=1.847cm]{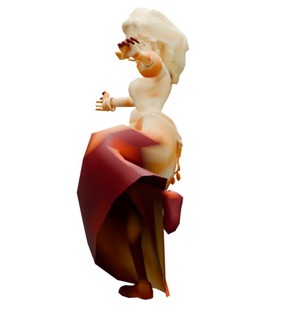}};
\node[val] at (3.585,7.408) {13.98};
\node[img] at (3.605,9.235) {\includegraphics[width=1.602cm,height=1.847cm]{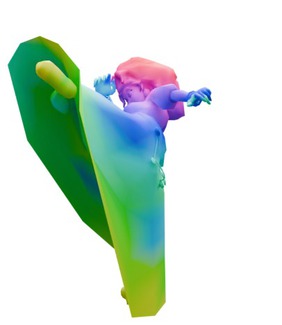}};
\node[img] at (3.605,7.388) {\includegraphics[width=1.602cm,height=1.847cm]{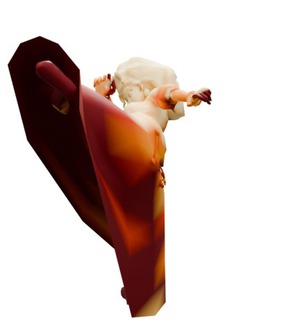}};
\node[val] at (5.187,7.408) {18.99};
\node[img] at (5.207,9.235) {\includegraphics[width=1.602cm,height=1.847cm]{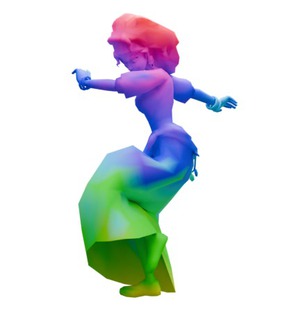}};
\node[img] at (5.207,7.388) {\includegraphics[width=1.602cm,height=1.847cm]{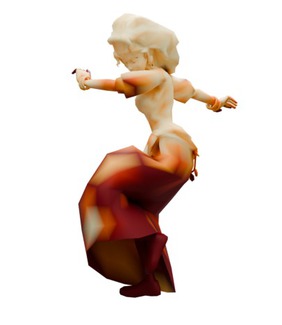}};
\node[val] at (6.789,7.408) {15.05};
\node[img] at (6.809,9.235) {\includegraphics[width=1.602cm,height=1.847cm]{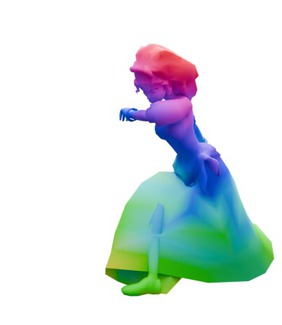}};
\node[img] at (6.809,7.388) {\includegraphics[width=1.602cm,height=1.847cm]{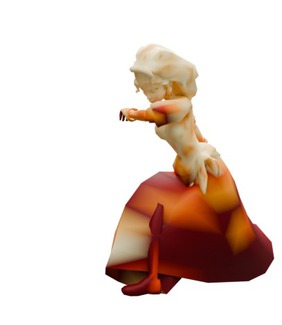}};
\node[val] at (8.391,7.408) {15.94};
\node[rotate=90, anchor=south, inner sep=1pt, align=center, font=\tiny] at (0.400,5.541) {Hybrid FM};
\node[img] at (0.400,5.541) {\includegraphics[width=1.602cm,height=1.847cm]{imgs/sm_fig/peasantgirl/src}};
\node[img] at (0.400,3.694) {\includegraphics[width=1.602cm,height=1.847cm]{imgs/sm_fig/peasantgirl/err_src}};
\node[val] at (1.982,3.714) {0};
\node[img] at (2.002,5.541) {\includegraphics[width=1.602cm,height=1.847cm]{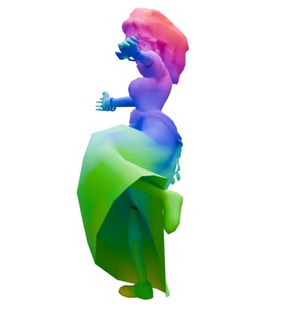}};
\node[img] at (2.002,3.694) {\includegraphics[width=1.602cm,height=1.847cm]{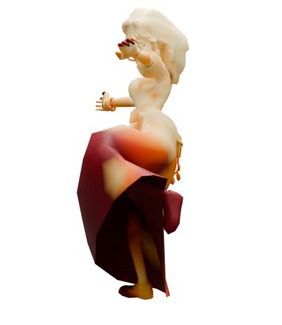}};
\node[val] at (3.585,3.714) {15.80};
\node[img] at (3.605,5.541) {\includegraphics[width=1.602cm,height=1.847cm]{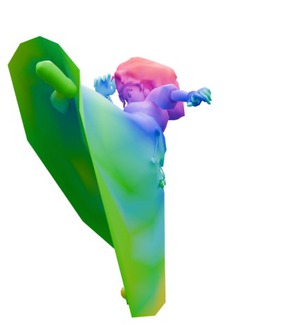}};
\node[img] at (3.605,3.694) {\includegraphics[width=1.602cm,height=1.847cm]{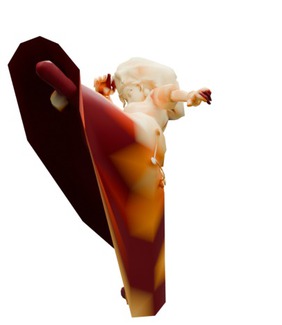}};
\node[val] at (5.187,3.714) {17.07};
\node[img] at (5.207,5.541) {\includegraphics[width=1.602cm,height=1.847cm]{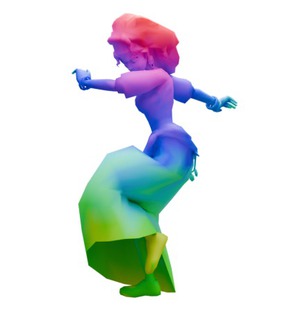}};
\node[img] at (5.207,3.694) {\includegraphics[width=1.602cm,height=1.847cm]{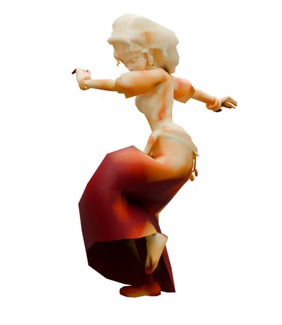}};
\node[val] at (6.789,3.714) {14.51};
\node[img] at (6.809,5.541) {\includegraphics[width=1.602cm,height=1.847cm]{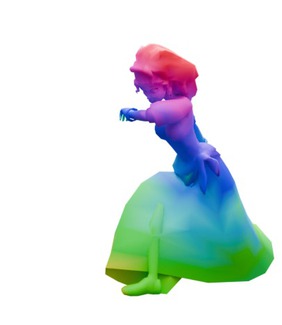}};
\node[img] at (6.809,3.694) {\includegraphics[width=1.602cm,height=1.847cm]{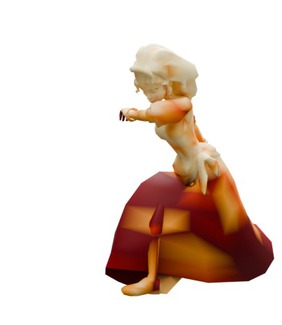}};
\node[val] at (8.391,3.714) {14.95};
\node[rotate=90, anchor=south, inner sep=1pt, align=center, font=\tiny] at (0.400,1.847) {\textbf{Ours}};
\node[img] at (0.400,1.847) {\includegraphics[width=1.602cm,height=1.847cm]{imgs/sm_fig/peasantgirl/src}};
\node[img] at (0.400,0.000) {\includegraphics[width=1.602cm,height=1.847cm]{imgs/sm_fig/peasantgirl/err_src}};
\node[val] at (1.982,0.020) {0};
\node[img] at (2.002,1.847) {\includegraphics[width=1.602cm,height=1.847cm]{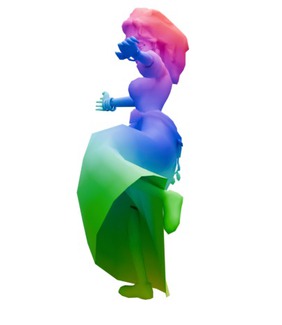}};
\node[img] at (2.002,0.000) {\includegraphics[width=1.602cm,height=1.847cm]{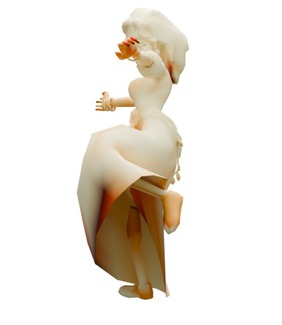}};
\node[val] at (3.585,0.020) {\textbf{1.65}};
\node[img] at (3.605,1.847) {\includegraphics[width=1.602cm,height=1.847cm]{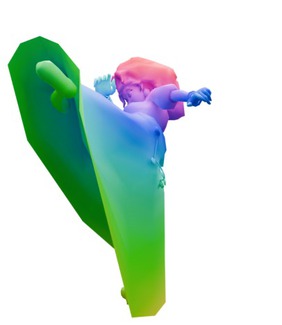}};
\node[img] at (3.605,0.000) {\includegraphics[width=1.602cm,height=1.847cm]{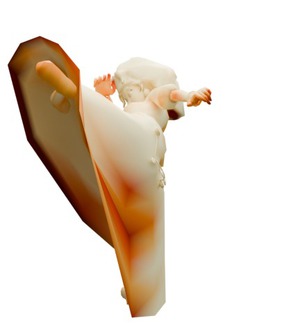}};
\node[val] at (5.187,0.020) {\textbf{3.44}};
\node[img] at (5.207,1.847) {\includegraphics[width=1.602cm,height=1.847cm]{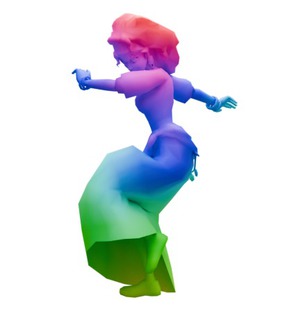}};
\node[img] at (5.207,0.000) {\includegraphics[width=1.602cm,height=1.847cm]{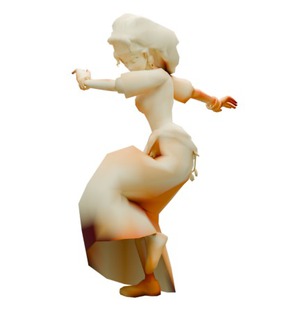}};
\node[val] at (6.789,0.020) {\textbf{2.38}};
\node[img] at (6.809,1.847) {\includegraphics[width=1.602cm,height=1.847cm]{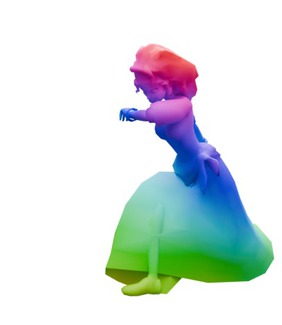}};
\node[img] at (6.809,0.000) {\includegraphics[width=1.602cm,height=1.847cm]{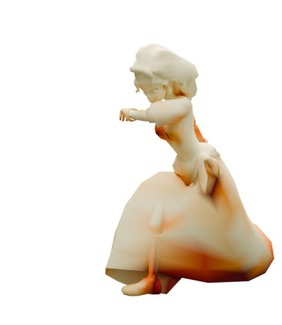}};
\node[val] at (8.391,0.020) {\textbf{3.59}};
\draw[tzOurs, line width=0.6pt, rounded corners=1pt] (0.380,-0.020) rectangle (8.431,3.714);
\node[anchor=south west, inner sep=0, font=\footnotesize] at (8.711,16.982) {vampire\_MmaKick};
\node[anchor=south, inner sep=1pt] at (9.580,16.642) {frame 1};
\node[anchor=south, inner sep=1pt] at (11.318,16.642) {13};
\node[anchor=south, inner sep=1pt] at (13.056,16.642) {25};
\node[anchor=south, inner sep=1pt] at (14.793,16.642) {37};
\node[anchor=south, inner sep=1pt] at (16.531,16.642) {49};
\node[img] at (8.711,14.775) {\includegraphics[width=1.738cm,height=1.847cm]{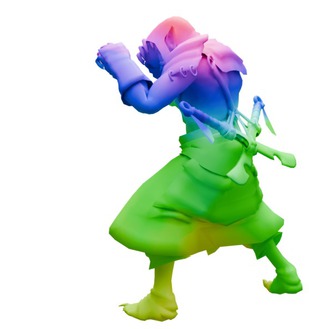}};
\node[img] at (10.449,14.775) {\includegraphics[width=1.738cm,height=1.847cm]{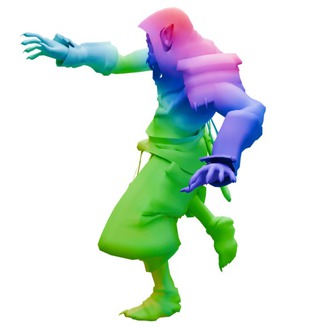}};
\node[img] at (12.187,14.775) {\includegraphics[width=1.738cm,height=1.847cm]{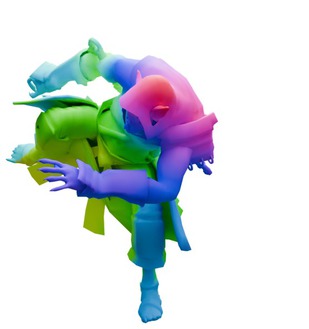}};
\node[img] at (13.925,14.775) {\includegraphics[width=1.738cm,height=1.847cm]{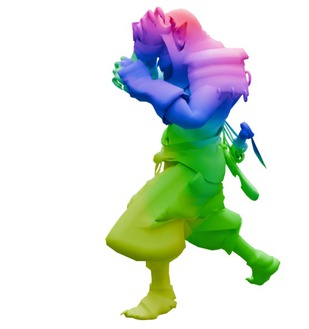}};
\node[img] at (15.662,14.775) {\includegraphics[width=1.738cm,height=1.847cm]{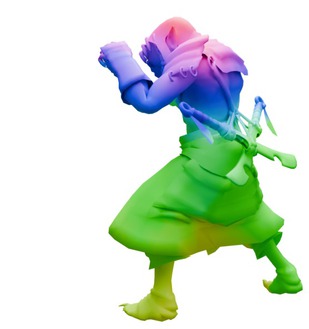}};
\node[img] at (8.711,12.929) {\includegraphics[width=1.738cm,height=1.847cm]{imgs/sm_fig/vampire/src}};
\node[img] at (8.711,11.082) {\includegraphics[width=1.738cm,height=1.847cm]{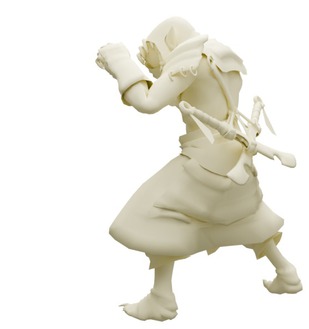}};
\node[val] at (10.429,11.102) {0};
\node[img] at (10.449,12.929) {\includegraphics[width=1.738cm,height=1.847cm]{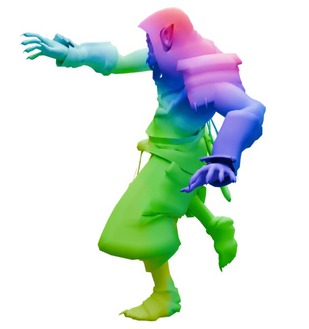}};
\node[img] at (10.449,11.082) {\includegraphics[width=1.738cm,height=1.847cm]{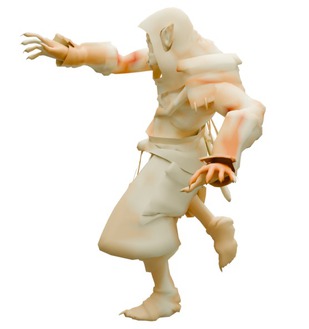}};
\node[val] at (12.167,11.102) {\textbf{1.08}};
\node[img] at (12.187,12.929) {\includegraphics[width=1.738cm,height=1.847cm]{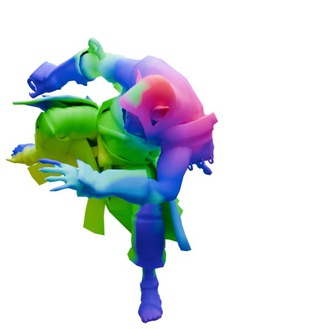}};
\node[img] at (12.187,11.082) {\includegraphics[width=1.738cm,height=1.847cm]{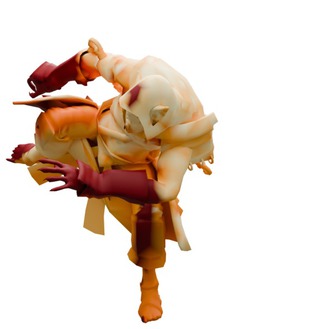}};
\node[val] at (13.905,11.102) {6.96};
\node[img] at (13.925,12.929) {\includegraphics[width=1.738cm,height=1.847cm]{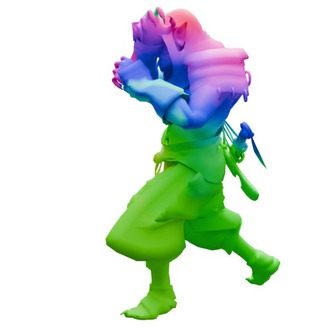}};
\node[img] at (13.925,11.082) {\includegraphics[width=1.738cm,height=1.847cm]{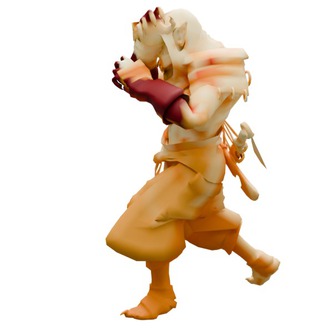}};
\node[val] at (15.642,11.102) {6.80};
\node[img] at (15.662,12.929) {\includegraphics[width=1.738cm,height=1.847cm]{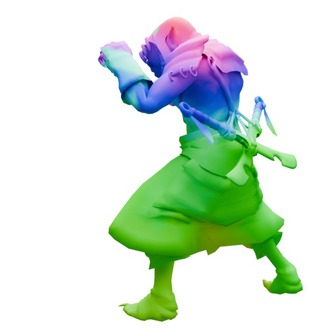}};
\node[img] at (15.662,11.082) {\includegraphics[width=1.738cm,height=1.847cm]{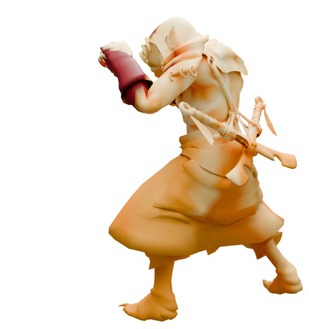}};
\node[val] at (17.380,11.102) {6.94};
\node[img] at (8.711,9.235) {\includegraphics[width=1.738cm,height=1.847cm]{imgs/sm_fig/vampire/src}};
\node[img] at (8.711,7.388) {\includegraphics[width=1.738cm,height=1.847cm]{imgs/sm_fig/vampire/err_src}};
\node[val] at (10.429,7.408) {0};
\node[img] at (10.449,9.235) {\includegraphics[width=1.738cm,height=1.847cm]{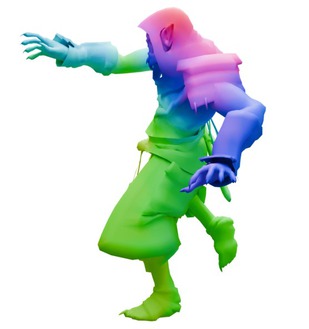}};
\node[img] at (10.449,7.388) {\includegraphics[width=1.738cm,height=1.847cm]{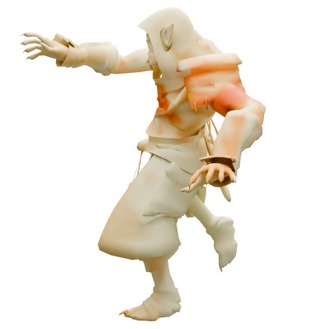}};
\node[val] at (12.167,7.408) {1.59};
\node[img] at (12.187,9.235) {\includegraphics[width=1.738cm,height=1.847cm]{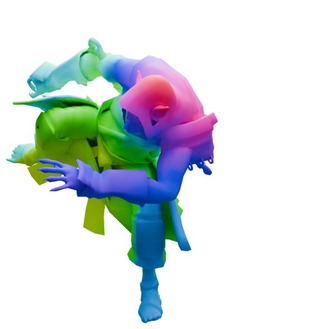}};
\node[img] at (12.187,7.388) {\includegraphics[width=1.738cm,height=1.847cm]{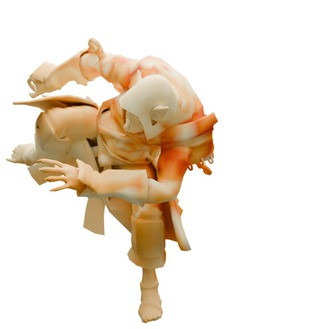}};
\node[val] at (13.905,7.408) {2.32};
\node[img] at (13.925,9.235) {\includegraphics[width=1.738cm,height=1.847cm]{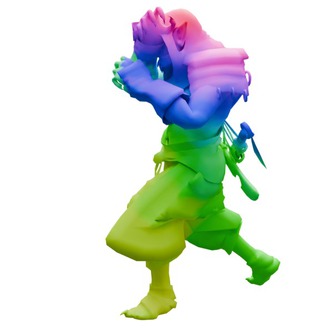}};
\node[img] at (13.925,7.388) {\includegraphics[width=1.738cm,height=1.847cm]{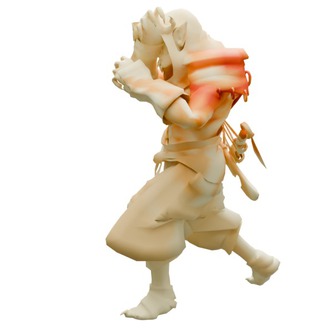}};
\node[val] at (15.642,7.408) {2.32};
\node[img] at (15.662,9.235) {\includegraphics[width=1.738cm,height=1.847cm]{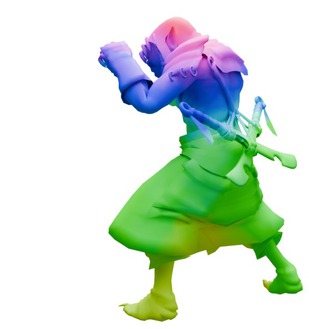}};
\node[img] at (15.662,7.388) {\includegraphics[width=1.738cm,height=1.847cm]{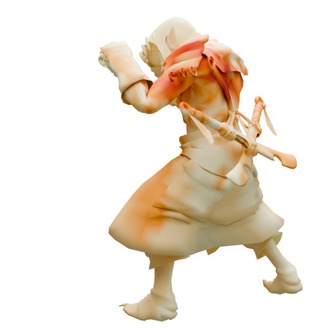}};
\node[val] at (17.380,7.408) {2.35};
\node[img] at (8.711,5.541) {\includegraphics[width=1.738cm,height=1.847cm]{imgs/sm_fig/vampire/src}};
\node[img] at (8.711,3.694) {\includegraphics[width=1.738cm,height=1.847cm]{imgs/sm_fig/vampire/err_src}};
\node[val] at (10.429,3.714) {0};
\node[img] at (10.449,5.541) {\includegraphics[width=1.738cm,height=1.847cm]{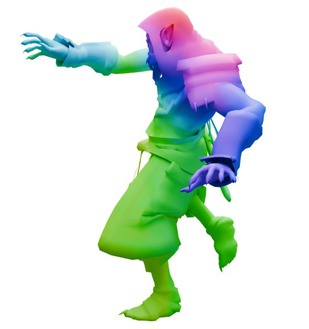}};
\node[img] at (10.449,3.694) {\includegraphics[width=1.738cm,height=1.847cm]{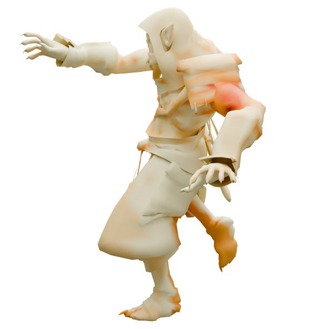}};
\node[val] at (12.167,3.714) {1.22};
\node[img] at (12.187,5.541) {\includegraphics[width=1.738cm,height=1.847cm]{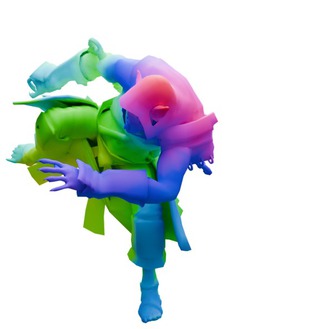}};
\node[img] at (12.187,3.694) {\includegraphics[width=1.738cm,height=1.847cm]{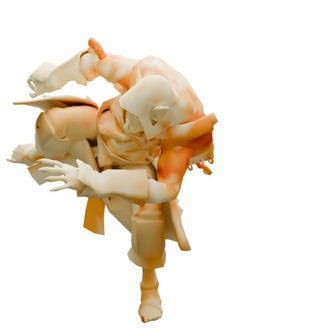}};
\node[val] at (13.905,3.714) {\textbf{1.95}};
\node[img] at (13.925,5.541) {\includegraphics[width=1.738cm,height=1.847cm]{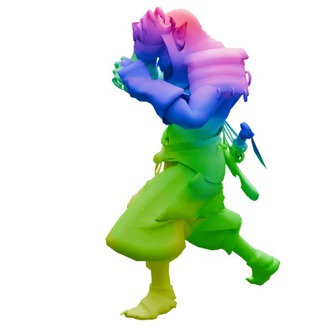}};
\node[img] at (13.925,3.694) {\includegraphics[width=1.738cm,height=1.847cm]{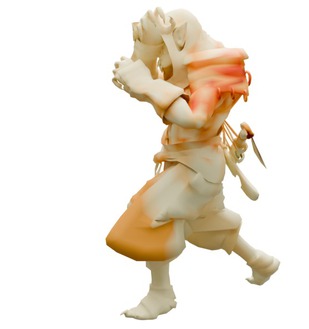}};
\node[val] at (15.642,3.714) {\textbf{2.00}};
\node[img] at (15.662,5.541) {\includegraphics[width=1.738cm,height=1.847cm]{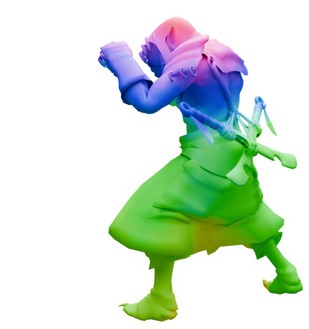}};
\node[img] at (15.662,3.694) {\includegraphics[width=1.738cm,height=1.847cm]{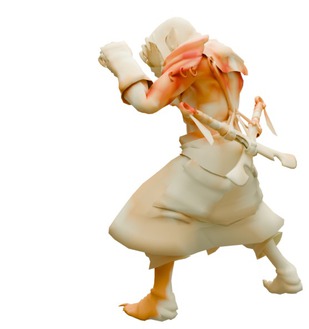}};
\node[val] at (17.380,3.714) {\textbf{2.04}};
\node[img] at (8.711,1.847) {\includegraphics[width=1.738cm,height=1.847cm]{imgs/sm_fig/vampire/src}};
\node[img] at (8.711,0.000) {\includegraphics[width=1.738cm,height=1.847cm]{imgs/sm_fig/vampire/err_src}};
\node[val] at (10.429,0.020) {0};
\node[img] at (10.449,1.847) {\includegraphics[width=1.738cm,height=1.847cm]{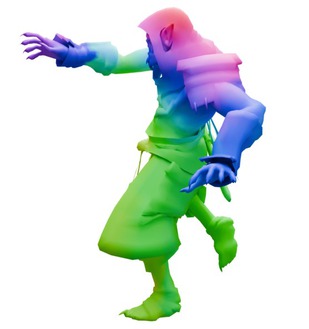}};
\node[img] at (10.449,0.000) {\includegraphics[width=1.738cm,height=1.847cm]{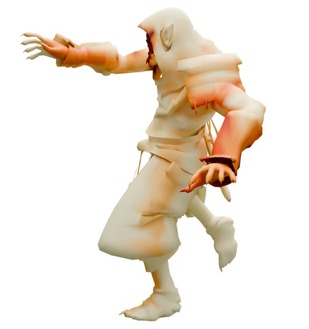}};
\node[val] at (12.167,0.020) {2.79};
\node[img] at (12.187,1.847) {\includegraphics[width=1.738cm,height=1.847cm]{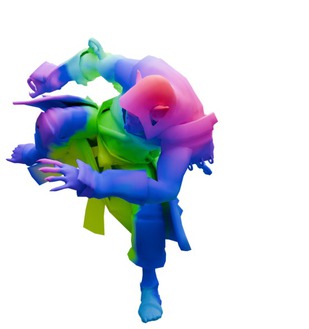}};
\node[img] at (12.187,0.000) {\includegraphics[width=1.738cm,height=1.847cm]{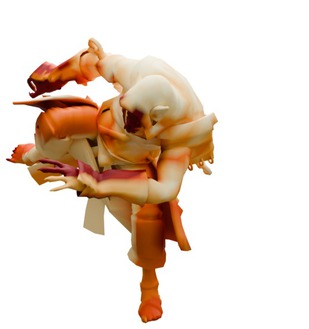}};
\node[val] at (13.905,0.020) {4.33};
\node[img] at (13.925,1.847) {\includegraphics[width=1.738cm,height=1.847cm]{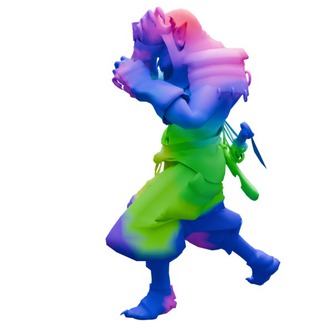}};
\node[img] at (13.925,0.000) {\includegraphics[width=1.738cm,height=1.847cm]{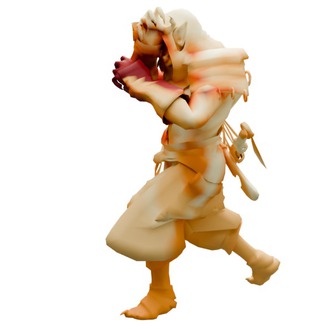}};
\node[val] at (15.642,0.020) {4.01};
\node[img] at (15.662,1.847) {\includegraphics[width=1.738cm,height=1.847cm]{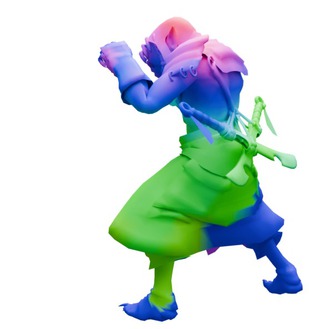}};
\node[img] at (15.662,0.000) {\includegraphics[width=1.738cm,height=1.847cm]{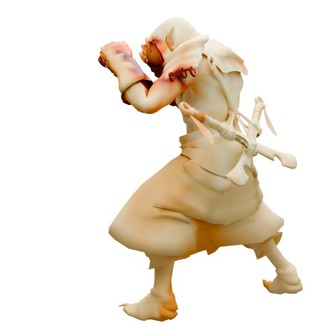}};
\node[val] at (17.380,0.020) {3.91};
\draw[tzOurs, line width=0.6pt, rounded corners=1pt] (8.691,-0.020) rectangle (17.420,3.714);
\shade[left color={rgb,255:red,255;green,247;blue,189}, middle color={rgb,255:red,252;green,153;blue,36}, right color={rgb,255:red,115;green,13;blue,26}] (5.220,-0.350) rectangle (12.180,-0.250);
\node[anchor=east, inner sep=1pt, font=\tiny] at (5.220,-0.300) {geodesic error 0};
\node[anchor=west, inner sep=1pt, font=\tiny] at (12.180,-0.300) {$\geq$10};
\end{tikzpicture}
\caption{\textbf{Shape matching on ground-truth surfaces}: points of frame~1 carried to frames 13, 25, 37 and 49 by functional maps (FM), ULRSSM and Hybrid FM, each chained over adjacent frames and run on the independently remeshed ground-truth frames, and by our flow fields, drawn on the ground-truth surface of that frame as in \Cref{fig:corr_gallery_1}: colored by their frame-1 position and by their geodesic error ($\times 100$; mean over all query points of the frame in the corner, lowest in bold). Per-scene means over all frames: \Cref{tab:correspondence_detailed}.}
\label{fig:sm_gt}
\end{figure*}

%% file: figures/pck_sm.tex
\begin{figure*}[t]
\centering
\includegraphics[width=\textwidth]{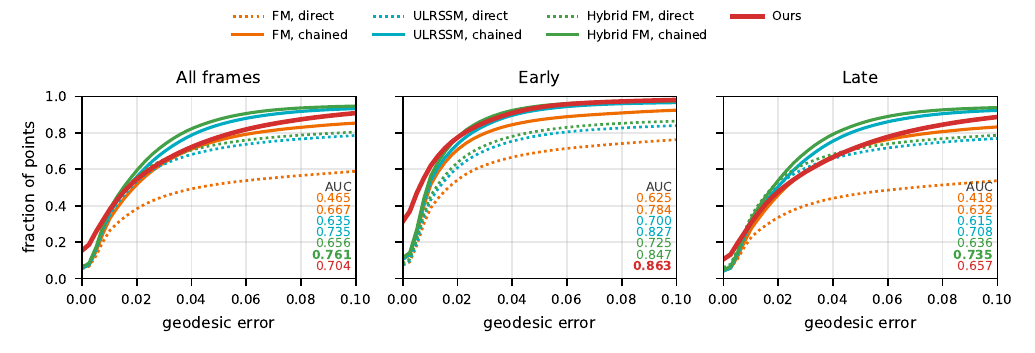}
\caption{\textbf{Cumulative correspondence error} on the ground-truth surfaces of DeformingThings4D (10 scenes): fraction of the points of frame~1 whose geodesic error in frame~$k$ (shape scaled to unit area) is below a threshold, pooled over the frames of each range; numbers: area under the curve (AUC, as in \Cref{tab:correspondence_gt}; best in bold). The shape-matching methods match each pair of ground-truth surfaces, frame~1 to frame~$k$ directly (dotted) or chained through consecutive frames (solid); they are shown as a reference for the accuracy of our correspondences.}
\label{fig:pck_sm}
\end{figure*}

%% file: figures/sm_twostage.tex
\begin{figure*}[p]
\centering
\definecolor{tzOurs}{RGB}{14,128,229}\definecolor{tzInk}{RGB}{24,28,40}
\begin{tikzpicture}[x=1cm, y=1cm, font=\scriptsize, text=tzInk, img/.style={anchor=south west, inner sep=0},
    val/.style={font=\tiny, anchor=south east, fill=white, fill opacity=0.8, text opacity=1, inner sep=1pt}]
\node[anchor=south west, inner sep=0, font=\footnotesize] at (0.400,19.160) {peasantgirl\_Martelo2};
\node[anchor=south, inner sep=1pt] at (1.027,18.820) {frame 1};
\node[anchor=south, inner sep=1pt] at (2.282,18.820) {13};
\node[anchor=south, inner sep=1pt] at (3.536,18.820) {25};
\node[anchor=south, inner sep=1pt] at (4.791,18.820) {37};
\node[anchor=south, inner sep=1pt] at (6.046,18.820) {49};
\node[img] at (0.400,17.354) {\includegraphics[width=1.255cm,height=1.446cm]{imgs/sm_fig/peasantgirl/src}};
\node[img] at (1.655,17.354) {\includegraphics[width=1.255cm,height=1.446cm]{imgs/sm_fig/peasantgirl/ref_13}};
\node[img] at (2.909,17.354) {\includegraphics[width=1.255cm,height=1.446cm]{imgs/sm_fig/peasantgirl/ref_25}};
\node[img] at (4.164,17.354) {\includegraphics[width=1.255cm,height=1.446cm]{imgs/sm_fig/peasantgirl/ref_37}};
\node[img] at (5.418,17.354) {\includegraphics[width=1.255cm,height=1.446cm]{imgs/sm_fig/peasantgirl/ref_49}};
\node[rotate=90, anchor=south, inner sep=1pt] at (0.400,18.077) {GT};
\node[rotate=90, anchor=south, inner sep=1pt, align=center, font=\tiny] at (0.400,15.185) {PGSR + Hybrid FM};
\node[img] at (1.655,15.908) {\includegraphics[width=1.255cm,height=1.446cm]{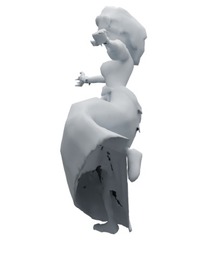}};
\node[img] at (2.909,15.908) {\includegraphics[width=1.255cm,height=1.446cm]{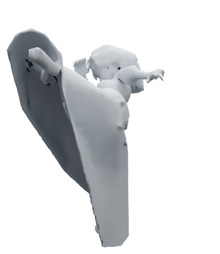}};
\node[img] at (4.164,15.908) {\includegraphics[width=1.255cm,height=1.446cm]{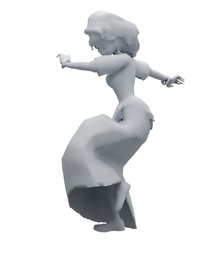}};
\node[img] at (5.418,15.908) {\includegraphics[width=1.255cm,height=1.446cm]{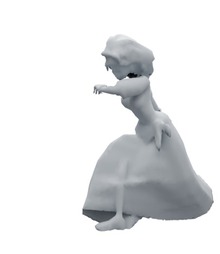}};
\node[font=\tiny, text=black!55, align=center] at (1.027,16.631) {input\\mesh};
\node[img] at (0.400,14.462) {\includegraphics[width=1.255cm,height=1.446cm]{imgs/sm_fig/peasantgirl/src}};
\node[img] at (0.400,13.015) {\includegraphics[width=1.255cm,height=1.446cm]{imgs/sm_fig/peasantgirl/err_src}};
\node[val] at (1.635,13.035) {0};
\node[img] at (1.655,14.462) {\includegraphics[width=1.255cm,height=1.446cm]{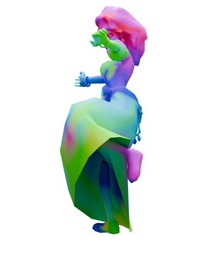}};
\node[img] at (1.655,13.015) {\includegraphics[width=1.255cm,height=1.446cm]{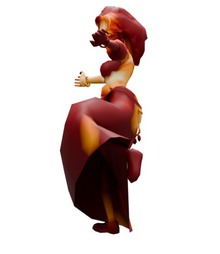}};
\node[val] at (2.889,13.035) {33.75};
\node[img] at (2.909,14.462) {\includegraphics[width=1.255cm,height=1.446cm]{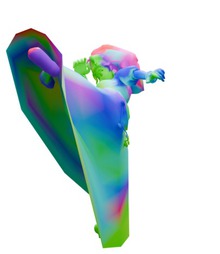}};
\node[img] at (2.909,13.015) {\includegraphics[width=1.255cm,height=1.446cm]{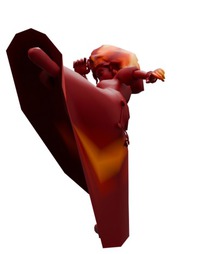}};
\node[val] at (4.144,13.035) {39.27};
\node[img] at (4.164,14.462) {\includegraphics[width=1.255cm,height=1.446cm]{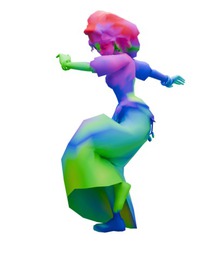}};
\node[img] at (4.164,13.015) {\includegraphics[width=1.255cm,height=1.446cm]{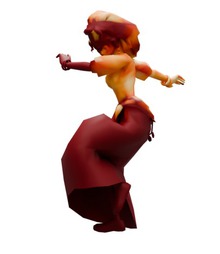}};
\node[val] at (5.398,13.035) {29.00};
\node[img] at (5.418,14.462) {\includegraphics[width=1.255cm,height=1.446cm]{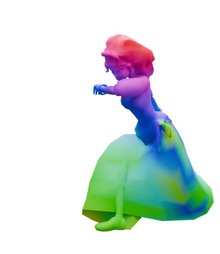}};
\node[img] at (5.418,13.015) {\includegraphics[width=1.255cm,height=1.446cm]{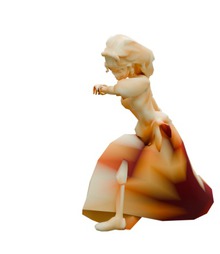}};
\node[val] at (6.653,13.035) {8.92};
\node[rotate=90, anchor=south, inner sep=1pt, align=center, font=\tiny] at (0.400,10.846) {Our mesh + ULRSSM};
\node[img] at (1.655,11.569) {\includegraphics[width=1.255cm,height=1.446cm]{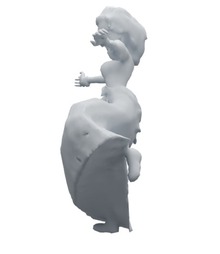}};
\node[img] at (2.909,11.569) {\includegraphics[width=1.255cm,height=1.446cm]{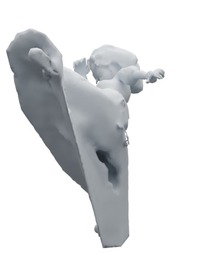}};
\node[img] at (4.164,11.569) {\includegraphics[width=1.255cm,height=1.446cm]{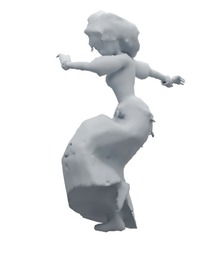}};
\node[img] at (5.418,11.569) {\includegraphics[width=1.255cm,height=1.446cm]{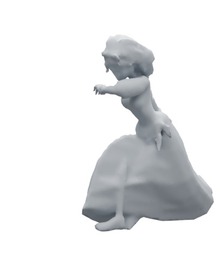}};
\node[font=\tiny, text=black!55, align=center] at (1.027,12.292) {input\\mesh};
\node[img] at (0.400,10.123) {\includegraphics[width=1.255cm,height=1.446cm]{imgs/sm_fig/peasantgirl/src}};
\node[img] at (0.400,8.677) {\includegraphics[width=1.255cm,height=1.446cm]{imgs/sm_fig/peasantgirl/err_src}};
\node[val] at (1.635,8.697) {0};
\node[img] at (1.655,10.123) {\includegraphics[width=1.255cm,height=1.446cm]{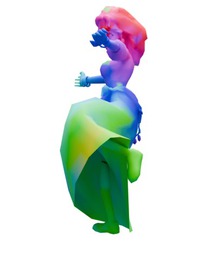}};
\node[img] at (1.655,8.677) {\includegraphics[width=1.255cm,height=1.446cm]{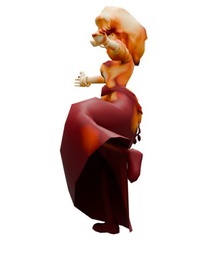}};
\node[val] at (2.889,8.697) {22.46};
\node[img] at (2.909,10.123) {\includegraphics[width=1.255cm,height=1.446cm]{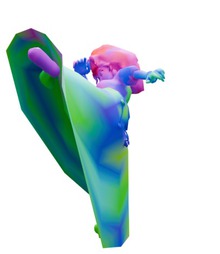}};
\node[img] at (2.909,8.677) {\includegraphics[width=1.255cm,height=1.446cm]{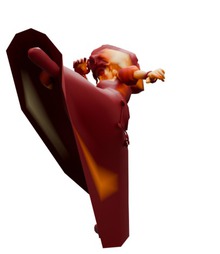}};
\node[val] at (4.144,8.697) {30.15};
\node[img] at (4.164,10.123) {\includegraphics[width=1.255cm,height=1.446cm]{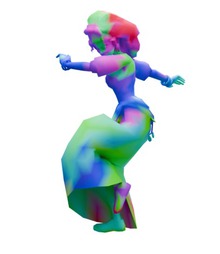}};
\node[img] at (4.164,8.677) {\includegraphics[width=1.255cm,height=1.446cm]{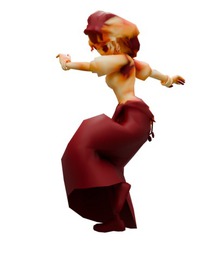}};
\node[val] at (5.398,8.697) {29.16};
\node[img] at (5.418,10.123) {\includegraphics[width=1.255cm,height=1.446cm]{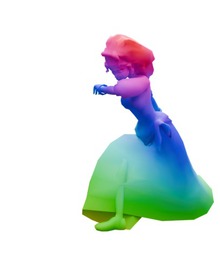}};
\node[img] at (5.418,8.677) {\includegraphics[width=1.255cm,height=1.446cm]{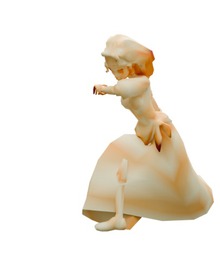}};
\node[val] at (6.653,8.697) {2.06};
\node[rotate=90, anchor=south, inner sep=1pt, align=center, font=\tiny] at (0.400,6.508) {Our mesh + Hybrid FM};
\node[img] at (1.655,7.231) {\includegraphics[width=1.255cm,height=1.446cm]{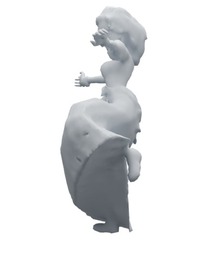}};
\node[img] at (2.909,7.231) {\includegraphics[width=1.255cm,height=1.446cm]{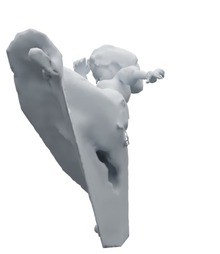}};
\node[img] at (4.164,7.231) {\includegraphics[width=1.255cm,height=1.446cm]{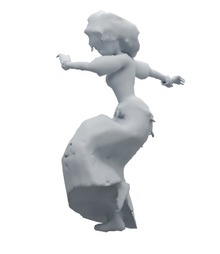}};
\node[img] at (5.418,7.231) {\includegraphics[width=1.255cm,height=1.446cm]{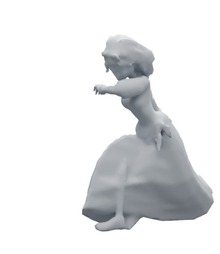}};
\node[font=\tiny, text=black!55, align=center] at (1.027,7.954) {input\\mesh};
\node[img] at (0.400,5.785) {\includegraphics[width=1.255cm,height=1.446cm]{imgs/sm_fig/peasantgirl/src}};
\node[img] at (0.400,4.338) {\includegraphics[width=1.255cm,height=1.446cm]{imgs/sm_fig/peasantgirl/err_src}};
\node[val] at (1.635,4.358) {0};
\node[img] at (1.655,5.785) {\includegraphics[width=1.255cm,height=1.446cm]{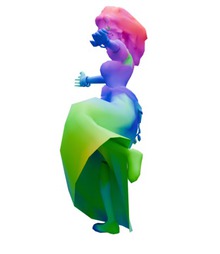}};
\node[img] at (1.655,4.338) {\includegraphics[width=1.255cm,height=1.446cm]{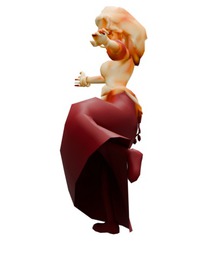}};
\node[val] at (2.889,4.358) {23.22};
\node[img] at (2.909,5.785) {\includegraphics[width=1.255cm,height=1.446cm]{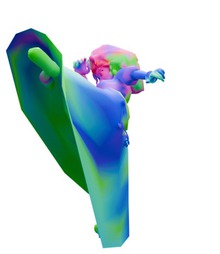}};
\node[img] at (2.909,4.338) {\includegraphics[width=1.255cm,height=1.446cm]{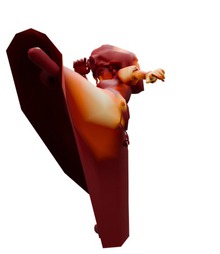}};
\node[val] at (4.144,4.358) {31.95};
\node[img] at (4.164,5.785) {\includegraphics[width=1.255cm,height=1.446cm]{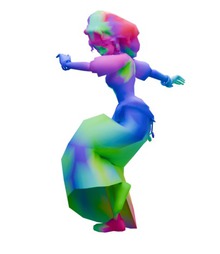}};
\node[img] at (4.164,4.338) {\includegraphics[width=1.255cm,height=1.446cm]{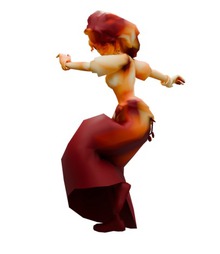}};
\node[val] at (5.398,4.358) {27.85};
\node[img] at (5.418,5.785) {\includegraphics[width=1.255cm,height=1.446cm]{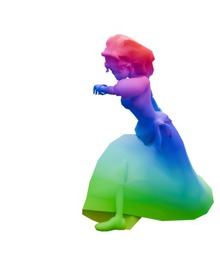}};
\node[img] at (5.418,4.338) {\includegraphics[width=1.255cm,height=1.446cm]{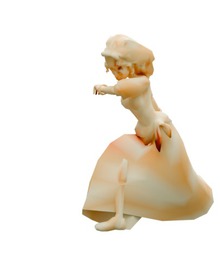}};
\node[val] at (6.653,4.358) {\textbf{1.61}};
\node[rotate=90, anchor=south, inner sep=1pt, align=center, font=\tiny] at (0.400,2.169) {\textbf{Ours}};
\node[img] at (1.655,2.892) {\includegraphics[width=1.255cm,height=1.446cm]{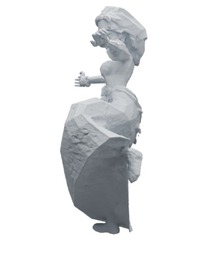}};
\node[img] at (2.909,2.892) {\includegraphics[width=1.255cm,height=1.446cm]{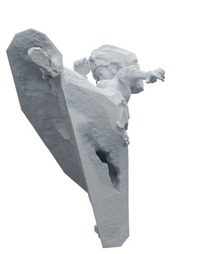}};
\node[img] at (4.164,2.892) {\includegraphics[width=1.255cm,height=1.446cm]{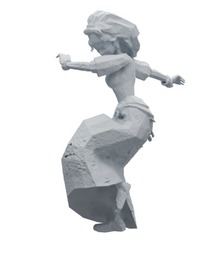}};
\node[img] at (5.418,2.892) {\includegraphics[width=1.255cm,height=1.446cm]{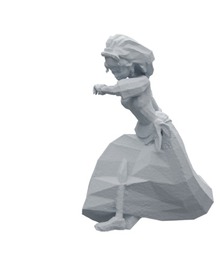}};
\node[font=\tiny, text=black!55, align=center] at (1.027,3.615) {input\\mesh};
\node[img] at (0.400,1.446) {\includegraphics[width=1.255cm,height=1.446cm]{imgs/sm_fig/peasantgirl/src}};
\node[img] at (0.400,0.000) {\includegraphics[width=1.255cm,height=1.446cm]{imgs/sm_fig/peasantgirl/err_src}};
\node[val] at (1.635,0.020) {0};
\node[img] at (1.655,1.446) {\includegraphics[width=1.255cm,height=1.446cm]{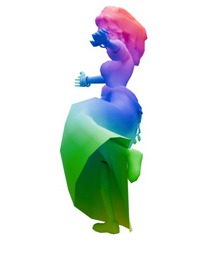}};
\node[img] at (1.655,0.000) {\includegraphics[width=1.255cm,height=1.446cm]{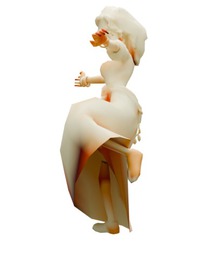}};
\node[val] at (2.889,0.020) {\textbf{1.87}};
\node[img] at (2.909,1.446) {\includegraphics[width=1.255cm,height=1.446cm]{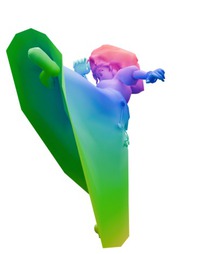}};
\node[img] at (2.909,0.000) {\includegraphics[width=1.255cm,height=1.446cm]{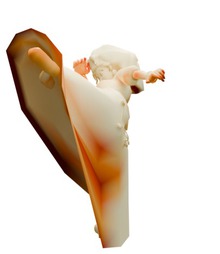}};
\node[val] at (4.144,0.020) {\textbf{3.47}};
\node[img] at (4.164,1.446) {\includegraphics[width=1.255cm,height=1.446cm]{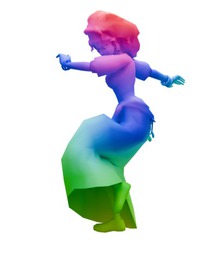}};
\node[img] at (4.164,0.000) {\includegraphics[width=1.255cm,height=1.446cm]{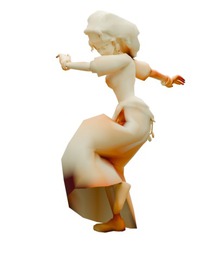}};
\node[val] at (5.398,0.020) {\textbf{2.40}};
\node[img] at (5.418,1.446) {\includegraphics[width=1.255cm,height=1.446cm]{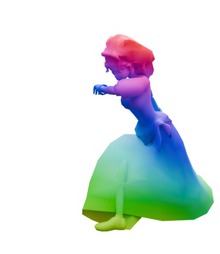}};
\node[img] at (5.418,0.000) {\includegraphics[width=1.255cm,height=1.446cm]{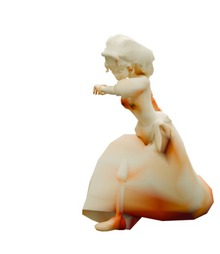}};
\node[val] at (6.653,0.020) {3.56};
\draw[tzOurs, line width=0.6pt, rounded corners=1pt] (0.380,-0.020) rectangle (6.693,4.358);
\node[anchor=south west, inner sep=0, font=\footnotesize] at (6.973,19.160) {vampire\_MmaKick};
\node[anchor=south, inner sep=1pt] at (7.653,18.820) {frame 1};
\node[anchor=south, inner sep=1pt] at (9.014,18.820) {13};
\node[anchor=south, inner sep=1pt] at (10.374,18.820) {25};
\node[anchor=south, inner sep=1pt] at (11.735,18.820) {37};
\node[anchor=south, inner sep=1pt] at (13.096,18.820) {49};
\node[img] at (6.973,17.354) {\includegraphics[width=1.361cm,height=1.446cm]{imgs/sm_fig/vampire/src}};
\node[img] at (8.334,17.354) {\includegraphics[width=1.361cm,height=1.446cm]{imgs/sm_fig/vampire/ref_13}};
\node[img] at (9.694,17.354) {\includegraphics[width=1.361cm,height=1.446cm]{imgs/sm_fig/vampire/ref_25}};
\node[img] at (11.055,17.354) {\includegraphics[width=1.361cm,height=1.446cm]{imgs/sm_fig/vampire/ref_37}};
\node[img] at (12.415,17.354) {\includegraphics[width=1.361cm,height=1.446cm]{imgs/sm_fig/vampire/ref_49}};
\node[img] at (8.334,15.908) {\includegraphics[width=1.361cm,height=1.446cm]{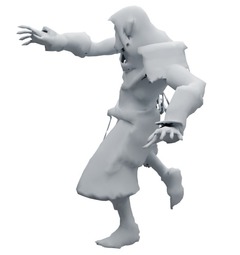}};
\node[img] at (9.694,15.908) {\includegraphics[width=1.361cm,height=1.446cm]{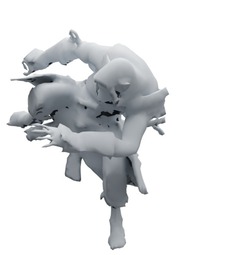}};
\node[img] at (11.055,15.908) {\includegraphics[width=1.361cm,height=1.446cm]{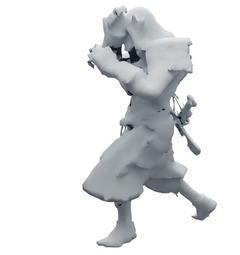}};
\node[img] at (12.415,15.908) {\includegraphics[width=1.361cm,height=1.446cm]{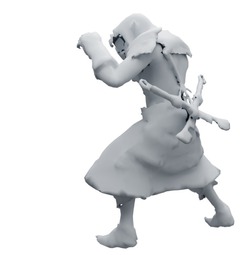}};
\node[font=\tiny, text=black!55, align=center] at (7.653,16.631) {input\\mesh};
\node[img] at (6.973,14.462) {\includegraphics[width=1.361cm,height=1.446cm]{imgs/sm_fig/vampire/src}};
\node[img] at (6.973,13.015) {\includegraphics[width=1.361cm,height=1.446cm]{imgs/sm_fig/vampire/err_src}};
\node[val] at (8.314,13.035) {0};
\node[img] at (8.334,14.462) {\includegraphics[width=1.361cm,height=1.446cm]{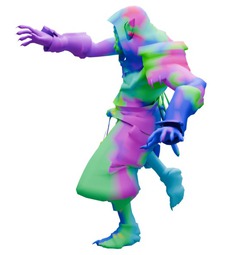}};
\node[img] at (8.334,13.015) {\includegraphics[width=1.361cm,height=1.446cm]{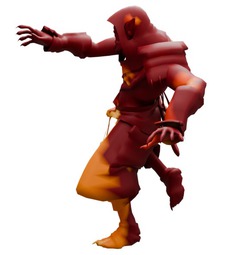}};
\node[val] at (9.674,13.035) {41.65};
\node[img] at (9.694,14.462) {\includegraphics[width=1.361cm,height=1.446cm]{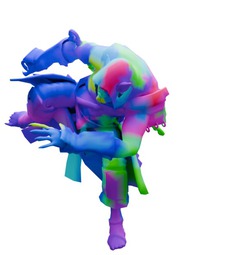}};
\node[img] at (9.694,13.015) {\includegraphics[width=1.361cm,height=1.446cm]{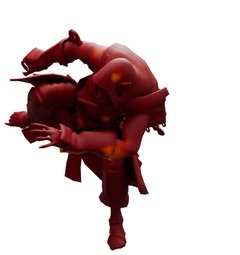}};
\node[val] at (11.035,13.035) {44.92};
\node[img] at (11.055,14.462) {\includegraphics[width=1.361cm,height=1.446cm]{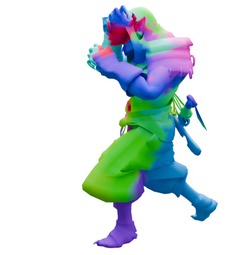}};
\node[img] at (11.055,13.015) {\includegraphics[width=1.361cm,height=1.446cm]{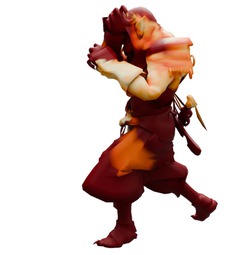}};
\node[val] at (12.395,13.035) {21.48};
\node[img] at (12.415,14.462) {\includegraphics[width=1.361cm,height=1.446cm]{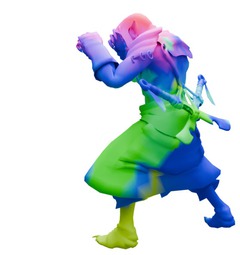}};
\node[img] at (12.415,13.015) {\includegraphics[width=1.361cm,height=1.446cm]{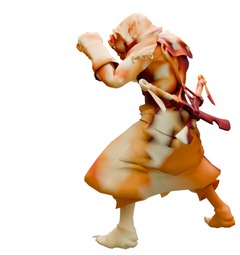}};
\node[val] at (13.756,13.035) {10.28};
\node[img] at (8.334,11.569) {\includegraphics[width=1.361cm,height=1.446cm]{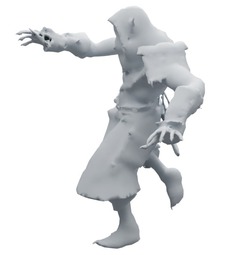}};
\node[img] at (9.694,11.569) {\includegraphics[width=1.361cm,height=1.446cm]{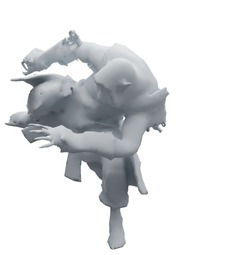}};
\node[img] at (11.055,11.569) {\includegraphics[width=1.361cm,height=1.446cm]{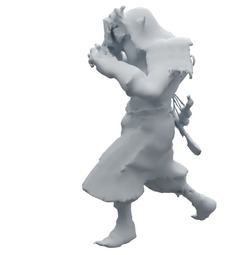}};
\node[img] at (12.415,11.569) {\includegraphics[width=1.361cm,height=1.446cm]{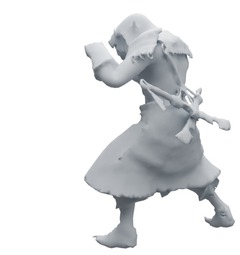}};
\node[font=\tiny, text=black!55, align=center] at (7.653,12.292) {input\\mesh};
\node[img] at (6.973,10.123) {\includegraphics[width=1.361cm,height=1.446cm]{imgs/sm_fig/vampire/src}};
\node[img] at (6.973,8.677) {\includegraphics[width=1.361cm,height=1.446cm]{imgs/sm_fig/vampire/err_src}};
\node[val] at (8.314,8.697) {0};
\node[img] at (8.334,10.123) {\includegraphics[width=1.361cm,height=1.446cm]{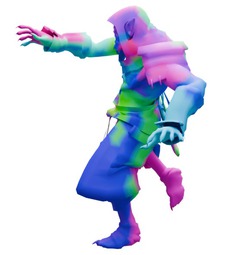}};
\node[img] at (8.334,8.677) {\includegraphics[width=1.361cm,height=1.446cm]{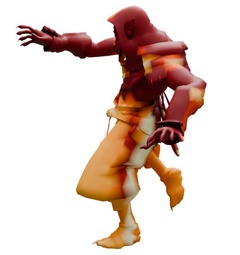}};
\node[val] at (9.674,8.697) {34.40};
\node[img] at (9.694,10.123) {\includegraphics[width=1.361cm,height=1.446cm]{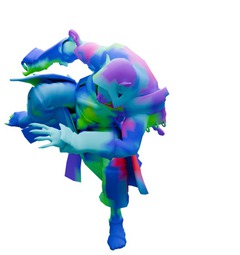}};
\node[img] at (9.694,8.677) {\includegraphics[width=1.361cm,height=1.446cm]{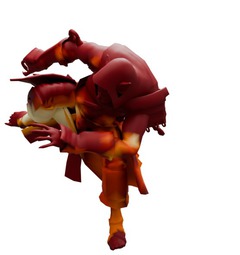}};
\node[val] at (11.035,8.697) {43.72};
\node[img] at (11.055,10.123) {\includegraphics[width=1.361cm,height=1.446cm]{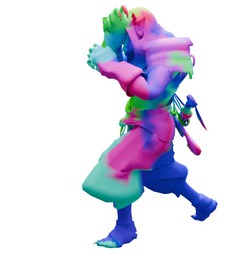}};
\node[img] at (11.055,8.677) {\includegraphics[width=1.361cm,height=1.446cm]{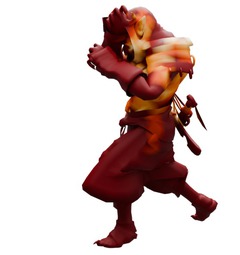}};
\node[val] at (12.395,8.697) {32.61};
\node[img] at (12.415,10.123) {\includegraphics[width=1.361cm,height=1.446cm]{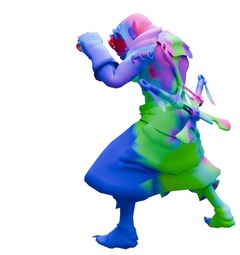}};
\node[img] at (12.415,8.677) {\includegraphics[width=1.361cm,height=1.446cm]{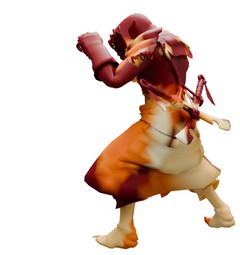}};
\node[val] at (13.756,8.697) {28.70};
\node[img] at (8.334,7.231) {\includegraphics[width=1.361cm,height=1.446cm]{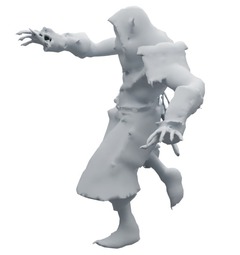}};
\node[img] at (9.694,7.231) {\includegraphics[width=1.361cm,height=1.446cm]{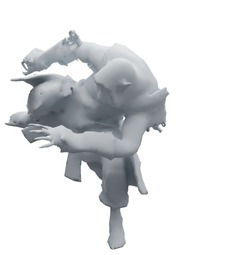}};
\node[img] at (11.055,7.231) {\includegraphics[width=1.361cm,height=1.446cm]{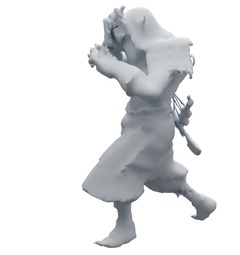}};
\node[img] at (12.415,7.231) {\includegraphics[width=1.361cm,height=1.446cm]{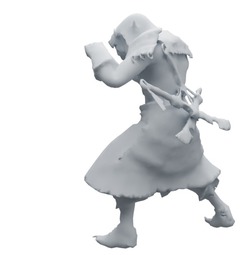}};
\node[font=\tiny, text=black!55, align=center] at (7.653,7.954) {input\\mesh};
\node[img] at (6.973,5.785) {\includegraphics[width=1.361cm,height=1.446cm]{imgs/sm_fig/vampire/src}};
\node[img] at (6.973,4.338) {\includegraphics[width=1.361cm,height=1.446cm]{imgs/sm_fig/vampire/err_src}};
\node[val] at (8.314,4.358) {0};
\node[img] at (8.334,5.785) {\includegraphics[width=1.361cm,height=1.446cm]{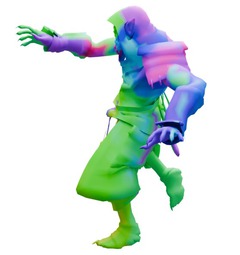}};
\node[img] at (8.334,4.338) {\includegraphics[width=1.361cm,height=1.446cm]{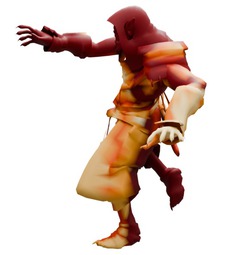}};
\node[val] at (9.674,4.358) {27.16};
\node[img] at (9.694,5.785) {\includegraphics[width=1.361cm,height=1.446cm]{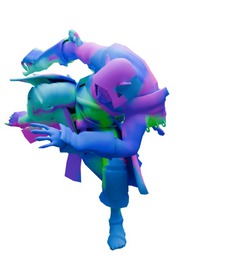}};
\node[img] at (9.694,4.338) {\includegraphics[width=1.361cm,height=1.446cm]{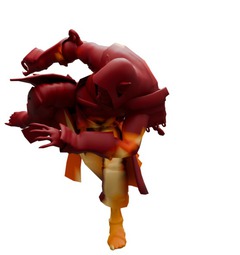}};
\node[val] at (11.035,4.358) {42.24};
\node[img] at (11.055,5.785) {\includegraphics[width=1.361cm,height=1.446cm]{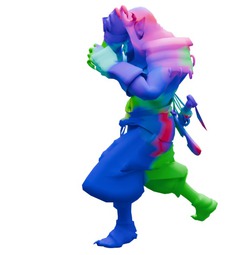}};
\node[img] at (11.055,4.338) {\includegraphics[width=1.361cm,height=1.446cm]{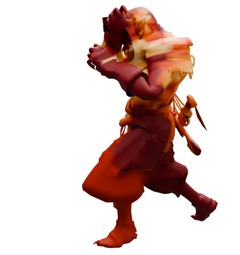}};
\node[val] at (12.395,4.358) {19.15};
\node[img] at (12.415,5.785) {\includegraphics[width=1.361cm,height=1.446cm]{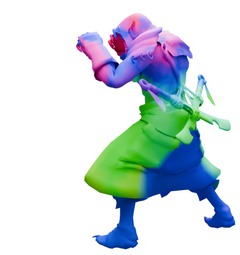}};
\node[img] at (12.415,4.338) {\includegraphics[width=1.361cm,height=1.446cm]{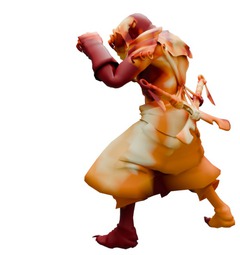}};
\node[val] at (13.756,4.358) {15.59};
\node[img] at (8.334,2.892) {\includegraphics[width=1.361cm,height=1.446cm]{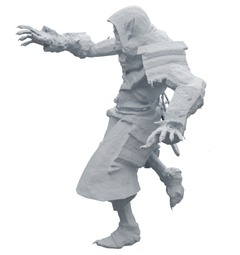}};
\node[img] at (9.694,2.892) {\includegraphics[width=1.361cm,height=1.446cm]{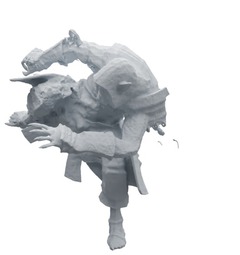}};
\node[img] at (11.055,2.892) {\includegraphics[width=1.361cm,height=1.446cm]{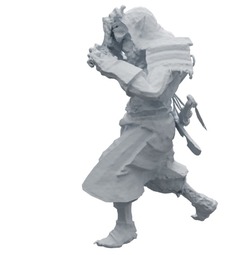}};
\node[img] at (12.415,2.892) {\includegraphics[width=1.361cm,height=1.446cm]{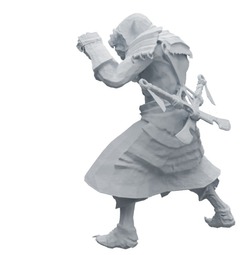}};
\node[font=\tiny, text=black!55, align=center] at (7.653,3.615) {input\\mesh};
\node[img] at (6.973,1.446) {\includegraphics[width=1.361cm,height=1.446cm]{imgs/sm_fig/vampire/src}};
\node[img] at (6.973,0.000) {\includegraphics[width=1.361cm,height=1.446cm]{imgs/sm_fig/vampire/err_src}};
\node[val] at (8.314,0.020) {0};
\node[img] at (8.334,1.446) {\includegraphics[width=1.361cm,height=1.446cm]{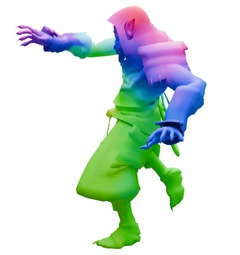}};
\node[img] at (8.334,0.000) {\includegraphics[width=1.361cm,height=1.446cm]{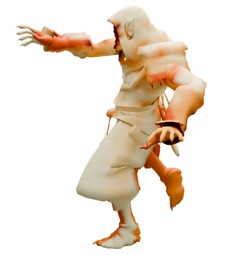}};
\node[val] at (9.674,0.020) {\textbf{3.42}};
\node[img] at (9.694,1.446) {\includegraphics[width=1.361cm,height=1.446cm]{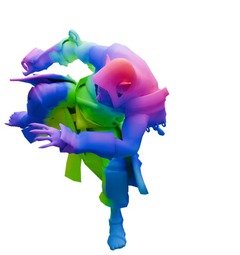}};
\node[img] at (9.694,0.000) {\includegraphics[width=1.361cm,height=1.446cm]{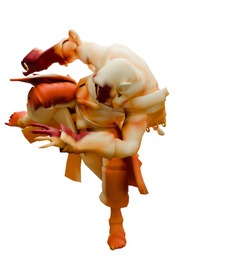}};
\node[val] at (11.035,0.020) {\textbf{5.27}};
\node[img] at (11.055,1.446) {\includegraphics[width=1.361cm,height=1.446cm]{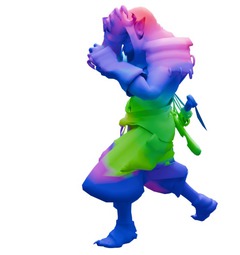}};
\node[img] at (11.055,0.000) {\includegraphics[width=1.361cm,height=1.446cm]{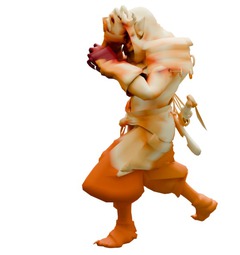}};
\node[val] at (12.395,0.020) {\textbf{4.55}};
\node[img] at (12.415,1.446) {\includegraphics[width=1.361cm,height=1.446cm]{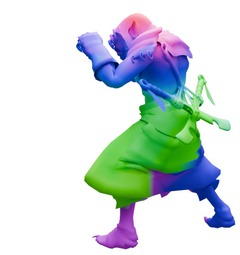}};
\node[img] at (12.415,0.000) {\includegraphics[width=1.361cm,height=1.446cm]{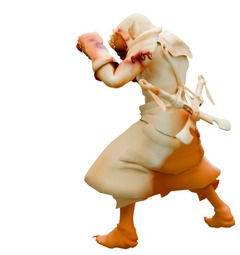}};
\node[val] at (13.756,0.020) {\textbf{4.45}};
\draw[tzOurs, line width=0.6pt, rounded corners=1pt] (6.953,-0.020) rectangle (13.796,4.358);
\shade[left color={rgb,255:red,255;green,247;blue,189}, middle color={rgb,255:red,252;green,153;blue,36}, right color={rgb,255:red,115;green,13;blue,26}] (4.133,-0.350) rectangle (9.643,-0.250);
\node[anchor=east, inner sep=1pt, font=\tiny] at (4.133,-0.300) {geodesic error 0};
\node[anchor=west, inner sep=1pt, font=\tiny] at (9.643,-0.300) {$\geq$10};
\end{tikzpicture}
\caption{\textbf{Two-stage pipelines on reconstructed surfaces}: shape matching (direct, frame~1 to frame~$k$) applied to per-frame reconstructions, the PGSR meshes decimated to about 5k vertices and our meshes remeshed to 5k--17k vertices (\emph{input mesh} rows), against the correspondences of our flow fields, read off our full-resolution reconstruction (its input mesh row); drawn and scored as in \Cref{fig:sm_gt}, all query points included. Points within 2\,cm of the frame-1 input mesh: PGSR 90\% / 55\%, our mesh 97\% / 60\%, ours 98\% / 71\% (peasantgirl / vampire); the others are scored like all points. For Hybrid FM, the two zero-area triangles of our remeshed peasantgirl frame~48 are removed, as in \Cref{tab:corr_twostage}.}
\label{fig:sm_twostage}
\end{figure*}

%% file: tables/liftflow.tex
\begin{table}[t]
\centering
\footnotesize
\setlength{\tabcolsep}{2.5pt}
\begin{tabular}{@{}l|cccc|cccc@{}}
\toprule
 & \multicolumn{4}{c|}{Ours (flow fields)} & \multicolumn{4}{c}{Lifted optical flow} \\
Scene & All & Early & Late & AUC$\uparrow$ & All & Early & Late & AUC$\uparrow$ \\
\midrule
abe & \textbf{1.77} & \textbf{0.63} & \textbf{2.11} & \textbf{0.823} & 8.18 & 3.10 & 9.68 & 0.370 \\
AJ & \textbf{2.45} & \textbf{1.50} & \textbf{2.74} & \textbf{0.790} & 16.50 & 6.52 & 19.46 & 0.410 \\
canie & \textbf{6.90} & \textbf{2.48} & \textbf{8.22} & \textbf{0.462} & 9.67 & 3.96 & 11.37 & 0.367 \\
dragon & \textbf{4.24} & \textbf{1.78} & \textbf{4.98} & \textbf{0.683} & 20.25 & 10.58 & 23.12 & 0.401 \\
fox & \textbf{3.99} & \textbf{1.95} & \textbf{4.60} & \textbf{0.607} & 6.55 & 2.81 & 7.66 & 0.450 \\
grizz & \textbf{2.47} & \textbf{1.07} & \textbf{2.88} & \textbf{0.762} & 4.95 & 1.90 & 5.85 & 0.605 \\
ninja & \textbf{3.64} & \textbf{2.75} & \textbf{3.91} & \textbf{0.889} & 6.80 & 3.98 & 7.64 & 0.652 \\
peasantgirl & \textbf{2.50} & \textbf{0.59} & \textbf{3.07} & \textbf{0.837} & 5.22 & 2.03 & 6.16 & 0.637 \\
rabbit & \textbf{9.33} & \textbf{4.95} & \textbf{10.63} & \textbf{0.345} & 12.35 & 6.10 & 14.21 & 0.264 \\
vampire & \textbf{4.06} & \textbf{1.47} & \textbf{4.83} & \textbf{0.754} & 14.61 & 7.38 & 16.76 & 0.424 \\
\midrule
Mean & \textbf{4.14} & \textbf{1.92} & \textbf{4.80} & \textbf{0.695} & 10.51 & 4.84 & 12.19 & 0.458 \\
\bottomrule
\end{tabular}
\caption{\textbf{Flow fields vs.\ lifted optical flow} on DeformingThings4D, on the same reconstructed surfaces (ours): geodesic error ($\times 100$, $\downarrow$) over all, early and late frames, as in \Cref{tab:correspondence}. Lifted optical flow: each point is moved by the FlowSeek optical flow of every camera that sees it, lifted to the next mesh, and chained, without flow fields. \textbf{Bold}: best of the two.}
\label{tab:liftflow}
\end{table}

%% file: figures/realdata_tracks.tex
\begin{figure*}[t]
\centering
\definecolor{tzInk}{RGB}{24,28,40}
\begin{tikzpicture}[x=1cm, y=1cm, font=\scriptsize, text=tzInk, cell/.style={anchor=center, inner sep=0}]
\node[anchor=south, inner sep=1pt] at (2.145,0.02) {frame 1};
\node[anchor=south, inner sep=1pt] at (5.535,0.02) {13};
\node[anchor=south, inner sep=1pt] at (8.925,0.02) {25};
\node[anchor=south, inner sep=1pt] at (12.315,0.02) {37};
\node[anchor=south, inner sep=1pt] at (15.705,0.02) {49};
\node[rotate=90, anchor=south, inner sep=1pt] at (0.450,-1.625) {magic2};
\node[cell] at (2.145,-1.625) {\includegraphics[width=3.290cm,height=3.170cm,keepaspectratio]{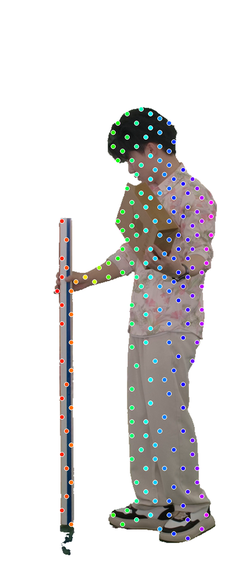}};
\node[cell] at (5.535,-1.625) {\includegraphics[width=3.290cm,height=3.170cm,keepaspectratio]{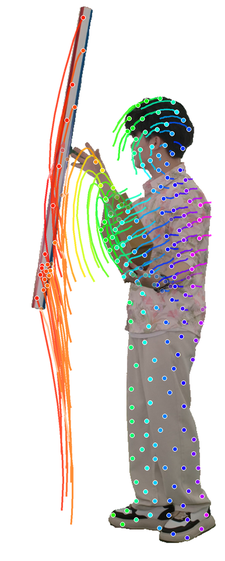}};
\node[cell] at (8.925,-1.625) {\includegraphics[width=3.290cm,height=3.170cm,keepaspectratio]{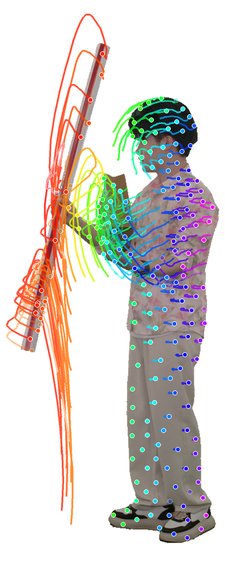}};
\node[cell] at (12.315,-1.625) {\includegraphics[width=3.290cm,height=3.170cm,keepaspectratio]{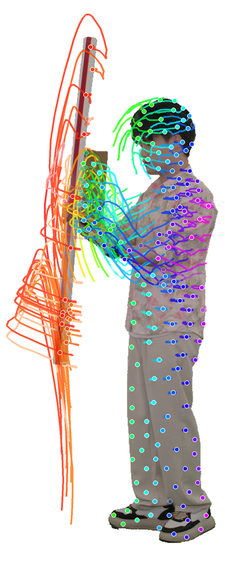}};
\node[cell] at (15.705,-1.625) {\includegraphics[width=3.290cm,height=3.170cm,keepaspectratio]{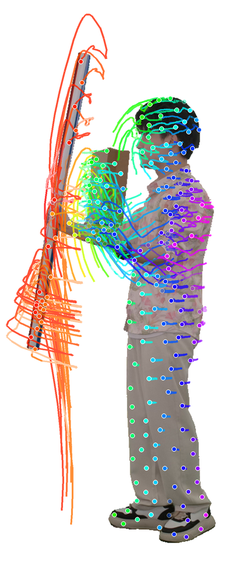}};
\node[rotate=90, anchor=south, inner sep=1pt] at (0.450,-4.875) {magic3};
\node[cell] at (2.145,-4.875) {\includegraphics[width=3.290cm,height=3.170cm,keepaspectratio]{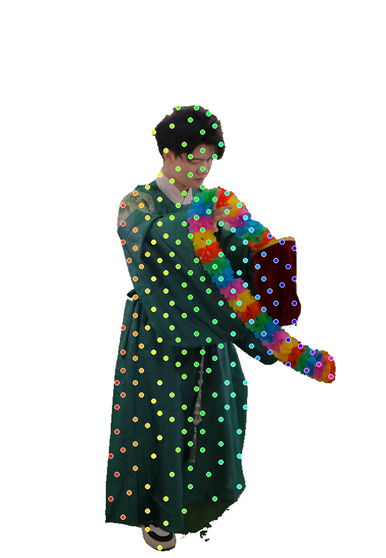}};
\node[cell] at (5.535,-4.875) {\includegraphics[width=3.290cm,height=3.170cm,keepaspectratio]{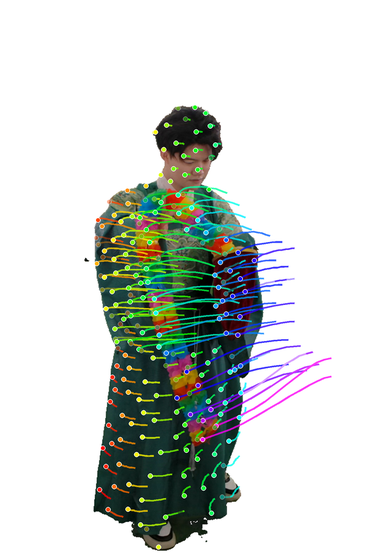}};
\node[cell] at (8.925,-4.875) {\includegraphics[width=3.290cm,height=3.170cm,keepaspectratio]{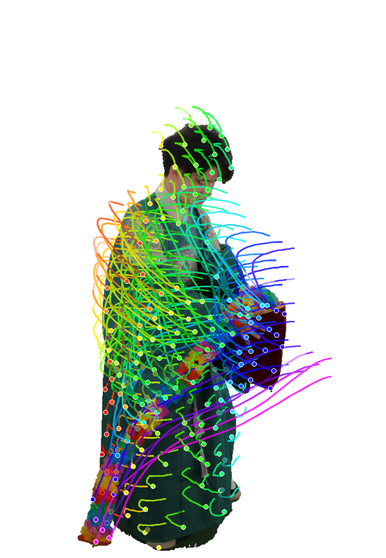}};
\node[cell] at (12.315,-4.875) {\includegraphics[width=3.290cm,height=3.170cm,keepaspectratio]{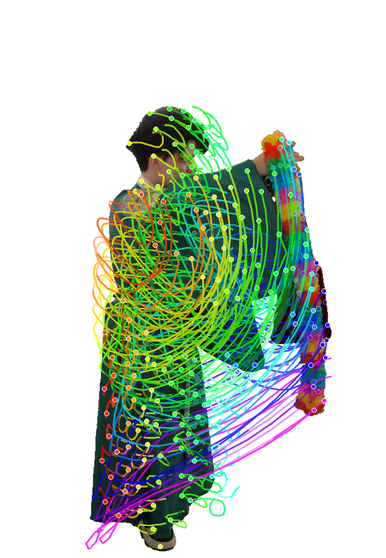}};
\node[cell] at (15.705,-4.875) {\includegraphics[width=3.290cm,height=3.170cm,keepaspectratio]{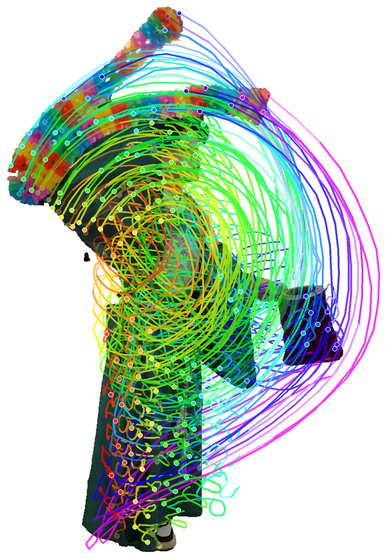}};
\draw[black!15, line width=0.3pt] (0.450,-3.250) -- (17.400,-3.250);
\node[rotate=90, anchor=south, inner sep=1pt] at (0.450,-8.125) {drawsword};
\node[cell] at (2.145,-8.125) {\includegraphics[width=3.290cm,height=3.170cm,keepaspectratio]{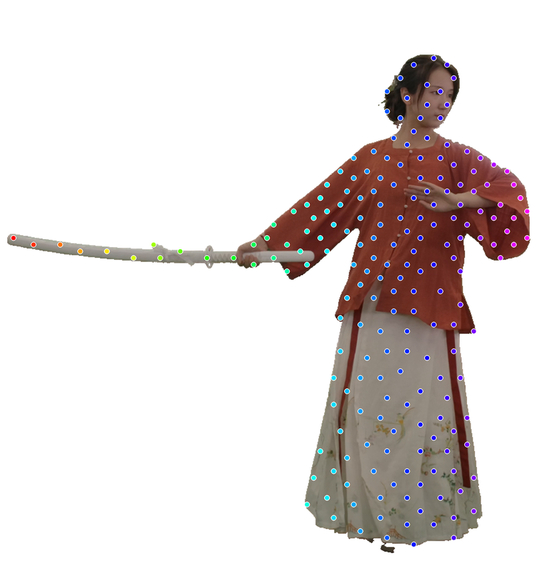}};
\node[cell] at (5.535,-8.125) {\includegraphics[width=3.290cm,height=3.170cm,keepaspectratio]{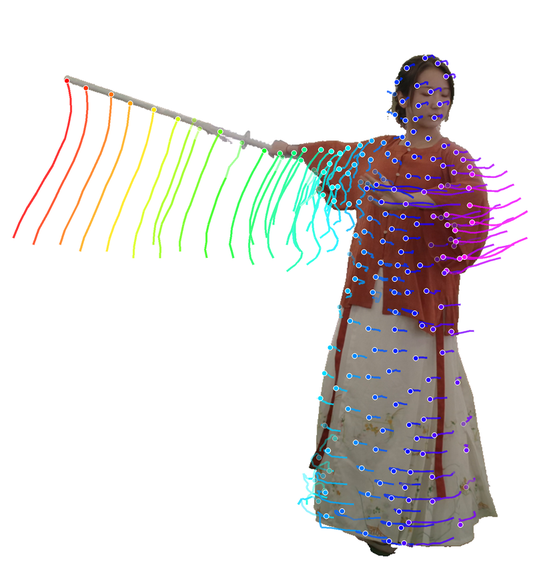}};
\node[cell] at (8.925,-8.125) {\includegraphics[width=3.290cm,height=3.170cm,keepaspectratio]{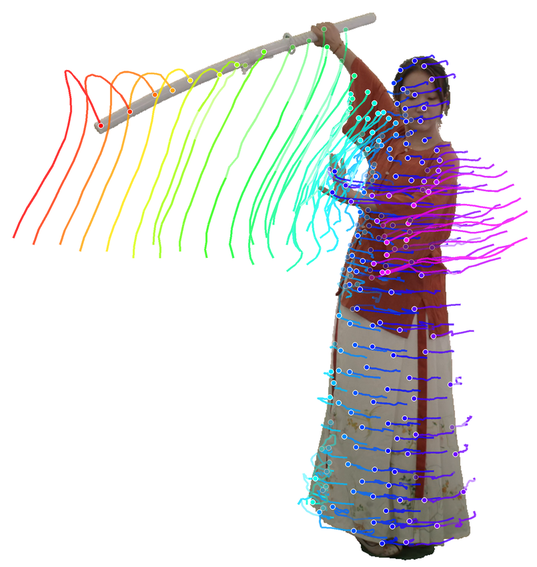}};
\node[cell] at (12.315,-8.125) {\includegraphics[width=3.290cm,height=3.170cm,keepaspectratio]{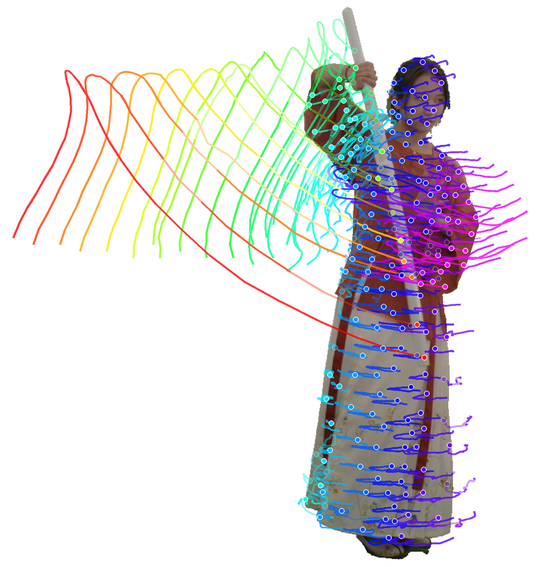}};
\node[cell] at (15.705,-8.125) {\includegraphics[width=3.290cm,height=3.170cm,keepaspectratio]{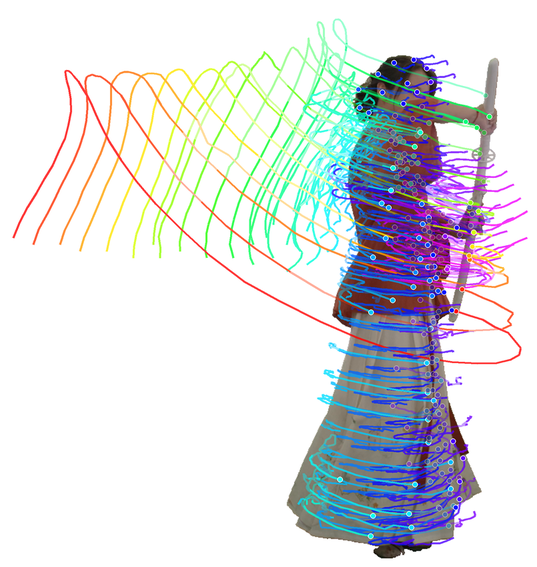}};
\draw[black!15, line width=0.3pt] (0.450,-6.500) -- (17.400,-6.500);
\node[rotate=90, anchor=south, inner sep=1pt] at (0.450,-11.375) {umbrella};
\node[cell] at (2.145,-11.375) {\includegraphics[width=3.290cm,height=3.170cm,keepaspectratio]{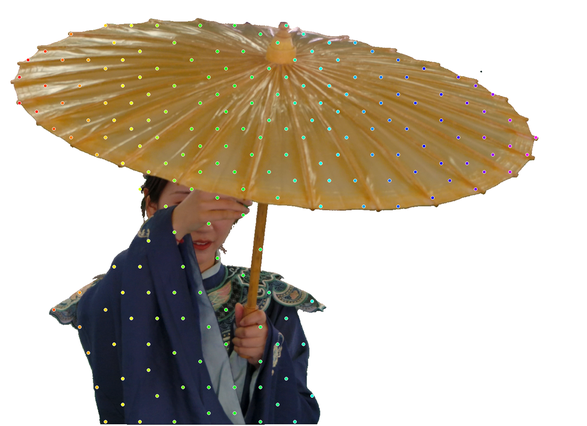}};
\node[cell] at (5.535,-11.375) {\includegraphics[width=3.290cm,height=3.170cm,keepaspectratio]{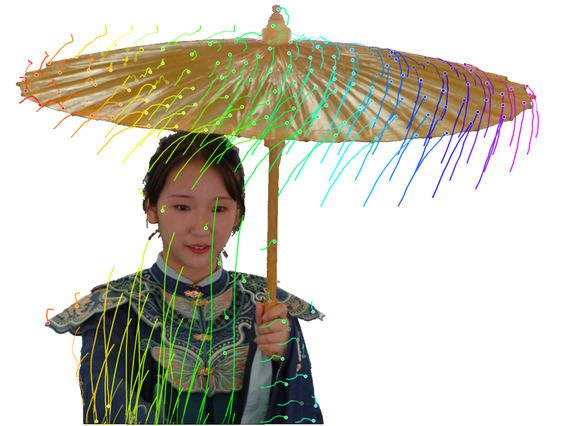}};
\node[cell] at (8.925,-11.375) {\includegraphics[width=3.290cm,height=3.170cm,keepaspectratio]{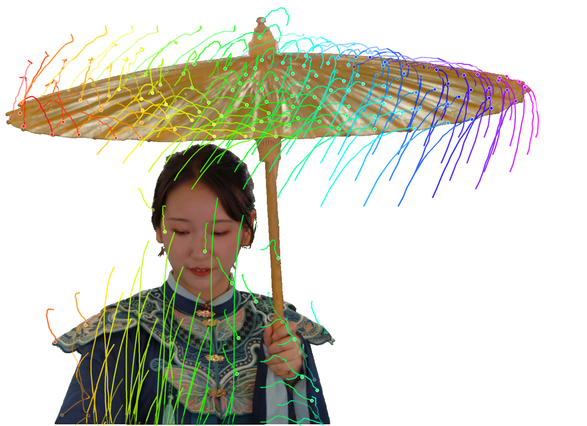}};
\node[cell] at (12.315,-11.375) {\includegraphics[width=3.290cm,height=3.170cm,keepaspectratio]{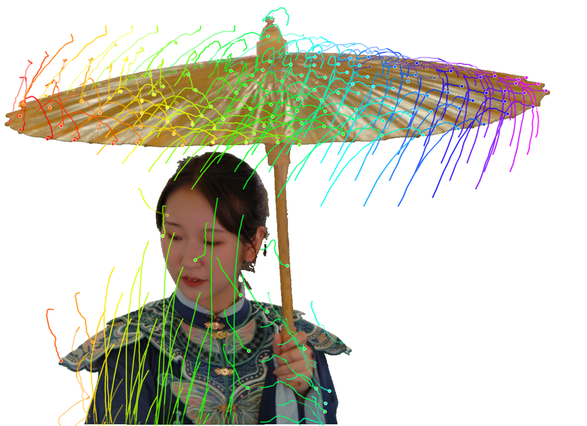}};
\node[cell] at (15.705,-11.375) {\includegraphics[width=3.290cm,height=3.170cm,keepaspectratio]{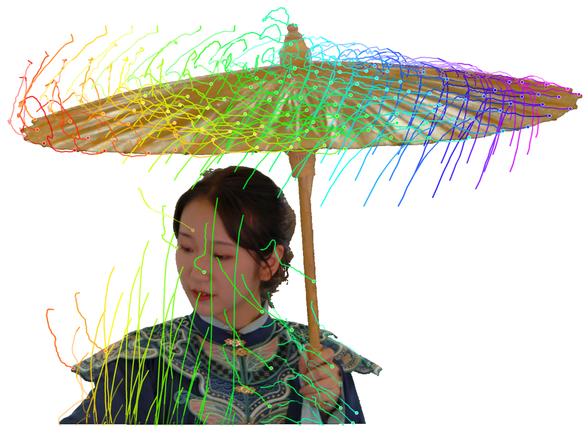}};
\draw[black!15, line width=0.3pt] (0.450,-9.750) -- (17.400,-9.750);
\node[rotate=90, anchor=south, inner sep=1pt] at (0.450,-14.625) {bag};
\node[cell] at (2.145,-14.625) {\includegraphics[width=3.290cm,height=3.170cm,keepaspectratio]{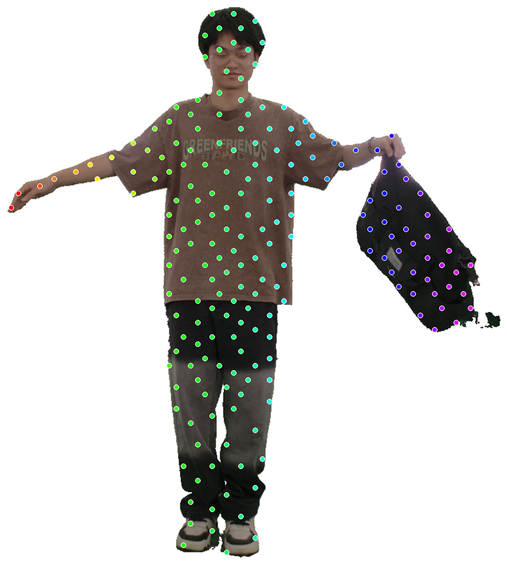}};
\node[cell] at (5.535,-14.625) {\includegraphics[width=3.290cm,height=3.170cm,keepaspectratio]{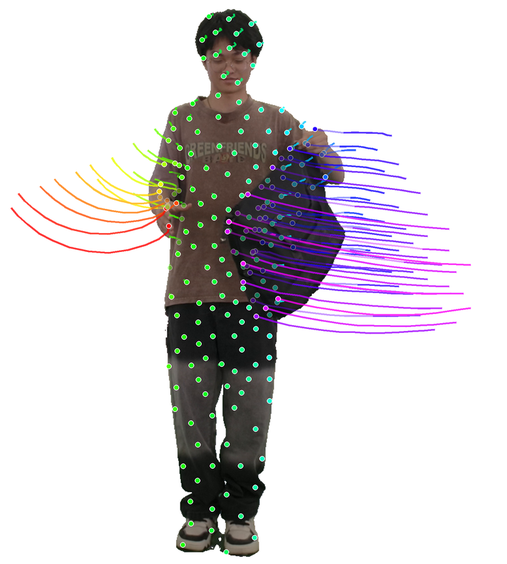}};
\node[cell] at (8.925,-14.625) {\includegraphics[width=3.290cm,height=3.170cm,keepaspectratio]{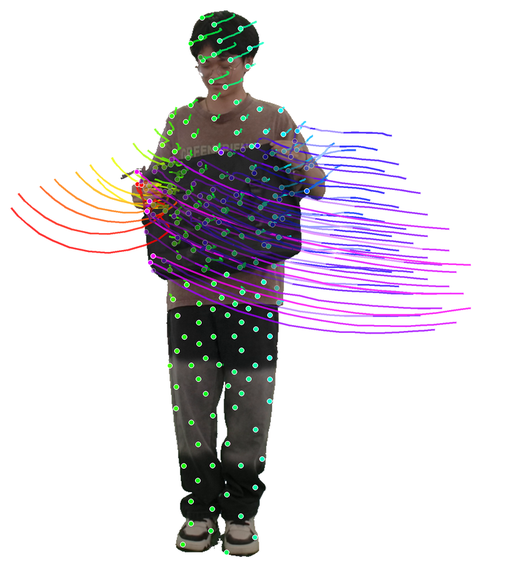}};
\node[cell] at (12.315,-14.625) {\includegraphics[width=3.290cm,height=3.170cm,keepaspectratio]{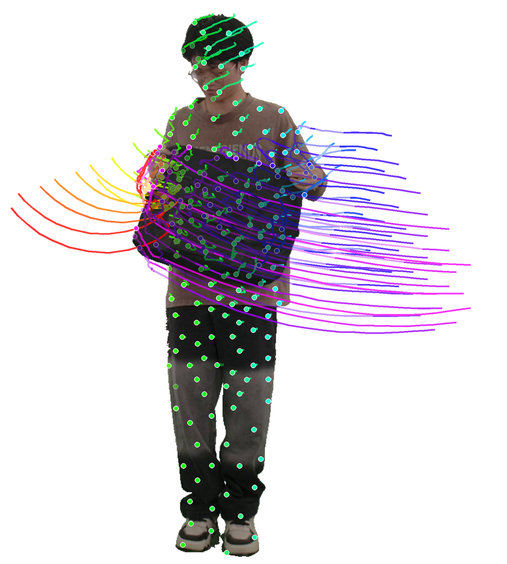}};
\node[cell] at (15.705,-14.625) {\includegraphics[width=3.290cm,height=3.170cm,keepaspectratio]{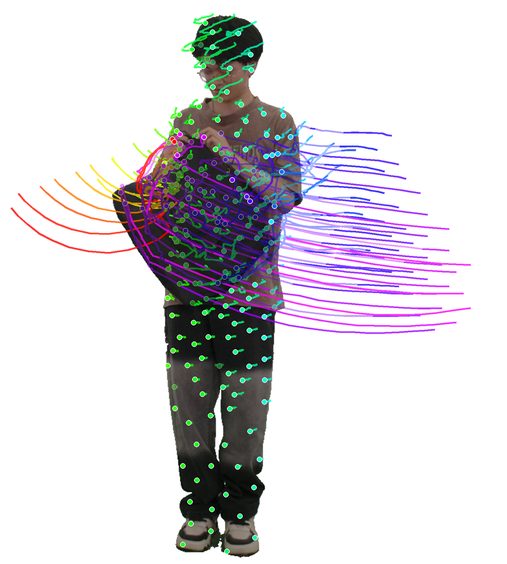}};
\draw[black!15, line width=0.3pt] (0.450,-13.000) -- (17.400,-13.000);
\node[rotate=90, anchor=south, inner sep=1pt] at (0.450,-17.875) {changeclothes};
\node[cell] at (2.145,-17.875) {\includegraphics[width=3.290cm,height=3.170cm,keepaspectratio]{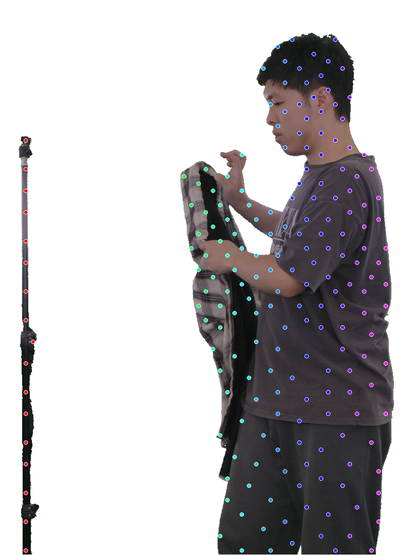}};
\node[cell] at (5.535,-17.875) {\includegraphics[width=3.290cm,height=3.170cm,keepaspectratio]{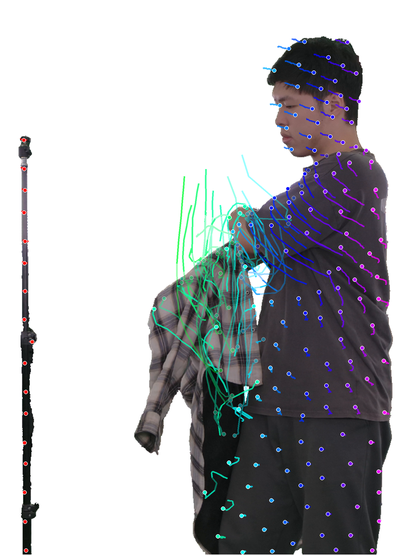}};
\node[cell] at (8.925,-17.875) {\includegraphics[width=3.290cm,height=3.170cm,keepaspectratio]{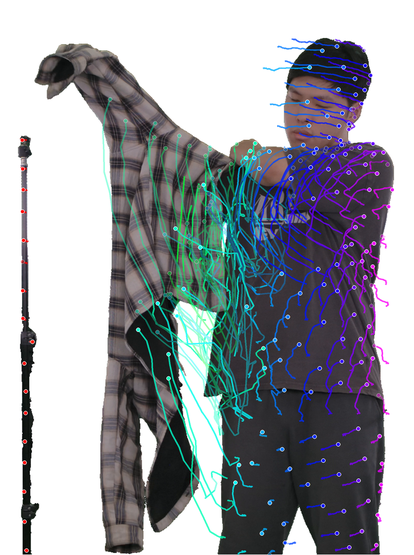}};
\node[cell] at (12.315,-17.875) {\includegraphics[width=3.290cm,height=3.170cm,keepaspectratio]{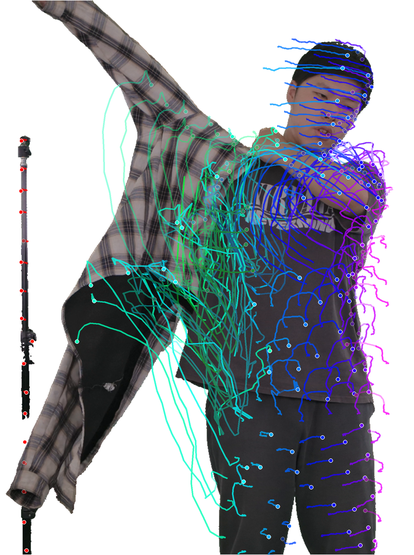}};
\node[cell] at (15.705,-17.875) {\includegraphics[width=3.290cm,height=3.170cm,keepaspectratio]{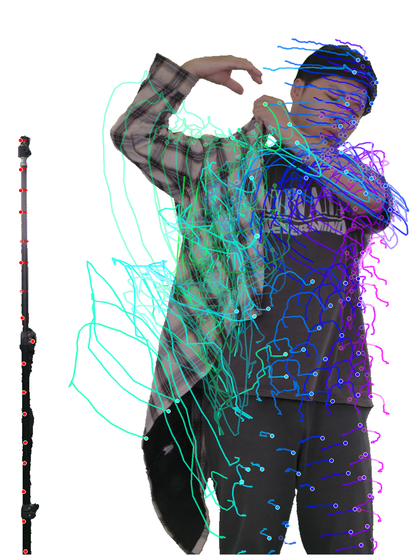}};
\draw[black!15, line width=0.3pt] (0.450,-16.250) -- (17.400,-16.250);
\end{tikzpicture}
\caption{\textbf{Our correspondences on real captures} (TaoGS), projected into a camera that was not used for training: 200 points of frame~1, colored by their frame-1 position, at later frames with their whole trajectory since frame~1, drawn as in the video on the project page; points and trajectory segments are fainter where the point is occluded on our reconstruction. Accuracy against CoTracker3 tracks: \Cref{tab:realdata_tracks,tab:realdata_tracks_common}.}
\label{fig:corr_realdata}
\end{figure*}

%% file: tables/realdata_tracks.tex
\begin{table*}[t]
\centering
\footnotesize
\setlength{\tabcolsep}{5pt}
\begin{tabular}{@{}lccc@{\hspace{10pt}}ccc@{\hspace{10pt}}ccc@{}}
\toprule
& \multicolumn{3}{c}{All} & \multicolumn{3}{c}{Early} & \multicolumn{3}{c}{Late} \\
\cmidrule(lr){2-4}\cmidrule(lr){5-7}\cmidrule(l){8-10}
Method & $\delta_{\mathrm{avg}}\uparrow$ & Med.$\downarrow$ & Vis. & $\delta_{\mathrm{avg}}\uparrow$ & Med.$\downarrow$ & Vis. & $\delta_{\mathrm{avg}}\uparrow$ & Med.$\downarrow$ & Vis. \\
\midrule
\multicolumn{10}{@{}l}{\textbf{TaoGS}} \\
\multicolumn{10}{@{}l}{\emph{Our reconstruction, other correspondences}} \\
Zero motion & 43.0 & 28.49 & 64 & 59.4 & 10.96 & 84 & 34.9 & 42.04 & 58 \\
Nearest vertex & 46.8 & 20.66 & 55 & 64.8 & 8.47 & 73 & 37.3 & 32.32 & 50 \\
Ours, canonical route & \rankB{75.8} & \rankB{4.46} & 61 & \rankB{87.6} & \rankB{2.59} & 82 & \rankB{69.6} & \rankB{6.24} & 55 \\
\midrule
\multicolumn{10}{@{}l}{\emph{Other reconstruction methods, their own correspondences}} \\
Dynamic 3DGS~\cite{Luiten2024} & 73.7 & 5.56 & 46 & 85.3 & 3.15 & 67 & 66.4 & 7.95 & 40 \\
4DGS~\cite{4dgaussians} & 67.6 & 7.32 & 39 & 78.5 & 4.38 & 58 & 58.9 & 12.73 & 33 \\
SC-GS~\cite{huang2024scgs} & \rankC{74.6} & 5.63 & 47 & 84.6 & 3.00 & 70 & 67.3 & 11.04 & 41 \\
Deformable 3DGS~\cite{yang2024deformable} & 73.6 & 5.46 & 48 & 84.3 & 3.21 & 71 & 65.8 & 9.67 & 42 \\
Dynamic-2DGS~\cite{zhang2024dynamic2dgaussiansgeometrically}$^\dagger$ & 57.2 & 10.83 & 39 & 68.0 & 6.82 & 59 & 49.0 & 17.34 & 33 \\
AT-GS~\cite{chen2025atgs}$^\ddagger$ & 62.2 & 8.16 & 34 & 75.4 & 5.10 & 59 & 51.7 & 11.98 & 27 \\
DG-Mesh~\cite{Liu2025} & 67.1 & 7.79 & 51 & 79.0 & 4.37 & 74 & 59.3 & 13.28 & 44 \\
GauSTAR~\cite{zheng2025gaustar} & 70.9 & 6.25 & 22 & 79.4 & 4.65 & 43 & 59.7 & 11.32 & 16 \\
MaGS~\cite{ma2024magsreconstructingsimulatingdynamic} & 69.9 & 6.71 & 55 & 82.4 & 3.46 & 74 & 62.8 & 10.91 & 50 \\
Tensor4D~\cite{shao2023tensor4d} & 47.4 & 27.12 & 49 & 54.8 & 21.73 & 63 & 43.7 & 29.91 & 44 \\
MoAngelo~\cite{ebbed2026moangelo} & 74.0 & \rankC{5.34} & 61 & \rankC{85.6} & \rankC{2.92} & 82 & \rankC{67.6} & \rankC{7.42} & 55 \\
\midrule
\textbf{Ours} & \rankA{82.2} & \rankA{3.42} & 62 & \rankA{90.8} & \rankA{2.12} & 82 & \rankA{77.6} & \rankA{4.50} & 56 \\
\midrule
\multicolumn{10}{@{}l}{\textbf{ActorsHQ}} \\
\multicolumn{10}{@{}l}{\emph{Our reconstruction, other correspondences}} \\
Zero motion & 27.0 & 40.63 & 63 & 33.2 & 30.48 & 77 & 24.4 & 44.95 & 58 \\
Nearest vertex & 33.6 & 30.46 & 52 & 39.7 & 24.01 & 66 & 30.2 & 34.89 & 47 \\
Ours, canonical route & \rankB{74.8} & \rankB{5.00} & 61 & \rankC{84.3} & \rankB{3.07} & 75 & \rankB{70.5} & \rankB{6.13} & 57 \\
\midrule
\multicolumn{10}{@{}l}{\emph{Other reconstruction methods, their own correspondences}} \\
Dynamic 3DGS & \rankC{73.0} & \rankC{5.62} & 45 & 79.4 & 4.26 & 60 & \rankC{69.8} & \rankC{6.46} & 41 \\
4DGS & 61.8 & 8.97 & 36 & 67.3 & 6.86 & 48 & 57.3 & 11.70 & 32 \\
SC-GS & 70.6 & 6.30 & 45 & 75.3 & 4.76 & 59 & 67.7 & 7.52 & 41 \\
Deformable 3DGS & 66.9 & 7.62 & 45 & 71.5 & 5.99 & 58 & 64.3 & 8.74 & 41 \\
Dynamic-2DGS$^\dagger$ & 54.2 & 22.72 & 48 & 56.0 & 20.06 & 57 & 53.8 & 23.81 & 46 \\
AT-GS$^\ddagger$ & 60.5 & 8.97 & 30 & 72.5 & 5.68 & 52 & 51.6 & 12.30 & 24 \\
DG-Mesh & 66.1 & 7.74 & 49 & 70.0 & 6.20 & 62 & 64.1 & 8.72 & 45 \\
GauSTAR & 71.0 & 5.83 & 40 & \rankB{85.5} & \rankC{3.09} & 63 & 55.1 & 17.01 & 33 \\
MaGS & 64.3 & 8.57 & 52 & 68.6 & 6.76 & 65 & 62.3 & 9.56 & 49 \\
Tensor4D & 54.3 & 12.01 & 59 & 58.2 & 10.21 & 72 & 52.6 & 12.99 & 56 \\
MoAngelo & 69.7 & 6.09 & 60 & 79.4 & 3.79 & 75 & 65.3 & 7.48 & 56 \\
\midrule
\textbf{Ours} & \rankA{83.1} & \rankA{3.42} & 61 & \rankA{87.4} & \rankA{2.58} & 75 & \rankA{81.2} & \rankA{3.82} & 57 \\
\bottomrule
\end{tabular}
\caption{\textbf{Correspondences on real captures}: each method's correspondences from frame~1 to frame~$k$, projected into held-out cameras and compared with CoTracker3 tracks over the 50 frames (All), the first quarter of the sequence (Early) and the rest (Late); TaoGS: 6 scenes, 5 held-out cameras each, $1920\times1080$; ActorsHQ: 6 scenes, 6 held-out cameras each, about $747\times1021$, so the same pixel thresholds are stricter. $\delta_{\mathrm{avg}}$: mean PCK at 4, 8, 16 and 32\,px (\%); Med.: median error (px); Vis.: percentage of all (query, frame) pairs that are evaluated, i.e.\ visible for CoTracker3 and on the method's own surface. Each method is evaluated on its own visible pairs; \Cref{tab:realdata_tracks_common} compares each method with ours on the same pairs. \ranklegend{} per dataset. $^\dagger$TaoGS: 3 scenes (drawsword, umbrella, bag), on which ours has 81.2 / 3.65 / 60; ActorsHQ: 2 scenes (actor7, actor8), on which ours has 88.4 / 2.41 / 72. $^\ddagger$Correspondence from the lineage of its Gaussians (see text).}
\label{tab:realdata_tracks}
\end{table*}

%% file: tables/realdata_tracks_common.tex
\begin{table}[t]
\centering
\footnotesize
\setlength{\tabcolsep}{2.5pt}
\begin{tabular}{@{}lccccccc@{}}
\toprule
& & \multicolumn{2}{c}{All} & \multicolumn{2}{c}{Early} & \multicolumn{2}{c}{Late} \\
\cmidrule(lr){3-4}\cmidrule(lr){5-6}\cmidrule(l){7-8}
Method & Vis. & Med. & Ours & Med. & Ours & Med. & Ours \\
\midrule
\multicolumn{8}{@{}l}{\textbf{TaoGS}} \\
Zero motion & 53 & 24.60 & \textbf{3.20} & 9.36 & \textbf{1.97} & 39.53 & \textbf{4.30} \\
Nearest vertex & 53 & 20.36 & \textbf{3.20} & 8.38 & \textbf{1.97} & 32.41 & \textbf{4.30} \\
Ours, canonical route & 53 & 4.16 & \textbf{3.20} & 2.43 & \textbf{1.97} & 5.79 & \textbf{4.30} \\
\midrule
Dynamic 3DGS & 46 & 5.52 & \textbf{3.07} & 3.13 & \textbf{2.01} & 7.91 & \textbf{4.17} \\
4DGS & 38 & 7.27 & \textbf{2.68} & 4.35 & \textbf{1.82} & 12.72 & \textbf{3.71} \\
SC-GS & 47 & 5.60 & \textbf{2.85} & 2.98 & \textbf{1.92} & 11.19 & \textbf{3.92} \\
Deformable 3DGS & 48 & 5.41 & \textbf{2.90} & 3.19 & \textbf{1.94} & 9.73 & \textbf{3.95} \\
Dynamic-2DGS$^\dagger$ & 38 & 10.74 & \textbf{3.01} & 6.77 & \textbf{2.13} & 17.55 & \textbf{3.92} \\
AT-GS$^\ddagger$ & 34 & 8.13 & \textbf{2.62} & 5.08 & \textbf{1.87} & 11.96 & \textbf{3.78} \\
DG-Mesh & 50 & 7.78 & \textbf{2.98} & 4.34 & \textbf{1.97} & 13.52 & \textbf{4.03} \\
GauSTAR & 22 & 6.21 & \textbf{2.04} & 4.64 & \textbf{1.63} & 11.01 & \textbf{3.67} \\
MaGS & 54 & 6.65 & \textbf{3.29} & 3.41 & \textbf{2.06} & 10.89 & \textbf{4.38} \\
Tensor4D & 47 & 27.01 & \textbf{3.53} & 21.71 & \textbf{2.07} & 29.77 & \textbf{4.87} \\
MoAngelo & 60 & 5.29 & \textbf{3.35} & 2.89 & \textbf{2.10} & 7.38 & \textbf{4.43} \\
\midrule
\multicolumn{8}{@{}l}{\textbf{ActorsHQ}} \\
Zero motion & 50 & 37.44 & \textbf{3.21} & 27.55 & \textbf{2.41} & 42.90 & \textbf{3.69} \\
Nearest vertex & 50 & 30.22 & \textbf{3.21} & 23.71 & \textbf{2.41} & 34.72 & \textbf{3.69} \\
Ours, canonical route & 50 & 4.53 & \textbf{3.21} & 2.79 & \textbf{2.41} & 5.80 & \textbf{3.69} \\
\midrule
Dynamic 3DGS & 45 & 5.60 & \textbf{3.16} & 4.25 & \textbf{2.37} & 6.44 & \textbf{3.60} \\
4DGS & 35 & 8.96 & \textbf{2.77} & 6.84 & \textbf{2.05} & 11.69 & \textbf{3.35} \\
SC-GS & 45 & 6.29 & \textbf{2.99} & 4.75 & \textbf{2.23} & 7.52 & \textbf{3.47} \\
Deformable 3DGS & 45 & 7.61 & \textbf{3.00} & 5.97 & \textbf{2.22} & 8.73 & \textbf{3.45} \\
Dynamic-2DGS$^\dagger$ & 48 & 22.37 & \textbf{2.18} & 19.86 & \textbf{1.66} & 23.25 & \textbf{2.47} \\
AT-GS$^\ddagger$ & 30 & 8.96 & \textbf{2.77} & 5.67 & \textbf{2.15} & 12.29 & \textbf{3.37} \\
DG-Mesh & 49 & 7.70 & \textbf{3.08} & 6.18 & \textbf{2.29} & 8.67 & \textbf{3.51} \\
GauSTAR & 40 & 5.82 & \textbf{2.90} & 3.07 & \textbf{2.38} & 17.00 & \textbf{3.58} \\
MaGS & 52 & 8.54 & \textbf{3.26} & 6.74 & \textbf{2.46} & 9.53 & \textbf{3.65} \\
Tensor4D & 58 & 12.03 & \textbf{3.39} & 10.21 & \textbf{2.55} & 13.02 & \textbf{3.78} \\
MoAngelo & 60 & 6.05 & \textbf{3.38} & 3.76 & \textbf{2.55} & 7.44 & \textbf{3.79} \\
\bottomrule
\end{tabular}
\caption{\textbf{Correspondences on real captures, same pairs}: median error (px) of each method (Med.) and of ours (Ours) on the pairs visible to CoTracker3, to ours and to that method (for our variants: to all four of them); Vis.: percentage of all pairs evaluated; \textbf{Bold}: best of each pair. Mean over the 6 scenes ($^\dagger$TaoGS 3, ActorsHQ 2 scenes).}
\label{tab:realdata_tracks_common}
\end{table}

%% file: tables/realdata_tracks_scenes.tex
\begin{table*}[t]
\centering
\footnotesize
\setlength{\tabcolsep}{4pt}
\begin{tabular}{@{}lcccccccccccc@{}}
\toprule
\textbf{TaoGS} & \multicolumn{2}{c}{magic2} & \multicolumn{2}{c}{magic3} & \multicolumn{2}{c}{drawsword} & \multicolumn{2}{c}{umbrella} & \multicolumn{2}{c}{bag} & \multicolumn{2}{c}{changeclothes} \\
\cmidrule(lr){2-3}\cmidrule(lr){4-5}\cmidrule(lr){6-7}\cmidrule(lr){8-9}\cmidrule(lr){10-11}\cmidrule(lr){12-13}
Method & $\delta_{\mathrm{avg}}\uparrow$ & Med.$\downarrow$ & $\delta_{\mathrm{avg}}\uparrow$ & Med.$\downarrow$ & $\delta_{\mathrm{avg}}\uparrow$ & Med.$\downarrow$ & $\delta_{\mathrm{avg}}\uparrow$ & Med.$\downarrow$ & $\delta_{\mathrm{avg}}\uparrow$ & Med.$\downarrow$ & $\delta_{\mathrm{avg}}\uparrow$ & Med.$\downarrow$ \\
\midrule
Zero motion & 58.1 & 8.52 & 26.0 & 41.25 & 31.4 & 28.06 & 15.9 & 77.75 & 72.6 & 4.39 & 54.0 & 10.98 \\
Nearest vertex & 62.0 & 7.36 & 29.4 & 33.72 & 31.8 & 26.73 & 23.5 & 42.64 & 77.2 & 3.47 & 56.7 & 10.07 \\
Ours, canonical route & 86.9 & 2.60 & \rankB{77.9} & \rankB{4.30} & \rankC{64.4} & \rankC{7.19} & \rankB{76.8} & \rankB{4.35} & 83.9 & 2.15 & 65.2 & 6.17 \\
\midrule
Dynamic 3DGS & 84.6 & 3.35 & 74.1 & 5.15 & 59.0 & 9.66 & 65.3 & 7.17 & \rankB{91.8} & 1.93 & 67.5 & 6.11 \\
4DGS & 83.2 & 3.20 & 60.7 & 7.53 & 46.9 & 13.76 & 59.7 & 11.84 & \rankC{91.6} & \rankA{1.35} & 63.5 & 6.24 \\
SC-GS & \rankB{90.0} & \rankC{2.25} & 73.6 & 4.95 & 57.6 & 8.40 & 63.7 & 12.15 & 90.9 & \rankC{1.45} & \rankC{71.5} & \rankB{4.58} \\
Deformable 3DGS & \rankC{89.4} & 2.35 & 70.6 & 4.96 & 54.5 & 9.72 & 66.9 & 8.78 & \rankA{92.8} & \rankB{1.42} & 67.2 & 5.52 \\
Dynamic-2DGS$^\dagger$ & -- & -- & -- & -- & 47.7 & 13.55 & 54.1 & 13.11 & 69.9 & 5.82 & -- & -- \\
AT-GS$^\ddagger$ & 68.5 & 6.37 & 61.6 & 7.97 & 56.9 & 9.65 & 51.8 & 11.61 & 73.6 & 5.18 & 60.6 & 8.17 \\
DG-Mesh & 85.8 & 2.99 & 57.5 & 8.77 & 51.5 & 11.29 & 59.5 & 13.12 & 90.1 & 1.82 & 58.2 & 8.76 \\
GauSTAR & 73.9 & 5.33 & 69.7 & 6.03 & \rankB{65.6} & \rankB{6.84} & 56.2 & 10.69 & 84.6 & 3.65 & \rankB{75.3} & \rankC{4.94} \\
MaGS & 88.9 & \rankB{2.16} & 61.0 & 7.81 & 49.4 & 12.85 & 66.6 & 9.23 & 90.3 & 1.66 & 63.1 & 6.57 \\
Tensor4D & 77.7 & 4.50 & 59.0 & 8.85 & 35.6 & 18.64 & 15.5 & 50.53 & 84.7 & 3.14 & 12.0 & 77.05 \\
MoAngelo & 87.6 & 2.46 & \rankC{74.4} & \rankC{4.83} & 57.2 & 9.60 & \rankC{73.3} & \rankC{5.15} & 89.8 & 2.04 & 61.6 & 7.97 \\
\midrule
\textbf{Ours} & \rankA{91.3} & \rankA{2.13} & \rankA{80.8} & \rankA{3.69} & \rankA{71.5} & \rankA{5.40} & \rankA{81.0} & \rankA{3.90} & 91.2 & 1.64 & \rankA{77.6} & \rankA{3.75} \\
\midrule
\textbf{ActorsHQ} & \multicolumn{2}{c}{actor4 (1258)} & \multicolumn{2}{c}{actor5 (1942)} & \multicolumn{2}{c}{actor6 (1081)} & \multicolumn{2}{c}{actor6 (1604)} & \multicolumn{2}{c}{actor7 (554)} & \multicolumn{2}{c}{actor8 (325)} \\
\cmidrule(lr){2-3}\cmidrule(lr){4-5}\cmidrule(lr){6-7}\cmidrule(lr){8-9}\cmidrule(lr){10-11}\cmidrule(lr){12-13}
Method & $\delta_{\mathrm{avg}}\uparrow$ & Med.$\downarrow$ & $\delta_{\mathrm{avg}}\uparrow$ & Med.$\downarrow$ & $\delta_{\mathrm{avg}}\uparrow$ & Med.$\downarrow$ & $\delta_{\mathrm{avg}}\uparrow$ & Med.$\downarrow$ & $\delta_{\mathrm{avg}}\uparrow$ & Med.$\downarrow$ & $\delta_{\mathrm{avg}}\uparrow$ & Med.$\downarrow$ \\
\midrule
Zero motion & 15.9 & 36.96 & 12.8 & 41.26 & 14.3 & 50.67 & 13.9 & 65.84 & 78.8 & 4.00 & 26.3 & 45.07 \\
Nearest vertex & 22.6 & 29.00 & 17.7 & 36.27 & 23.3 & 36.44 & 21.1 & 45.80 & 84.0 & 2.98 & 32.7 & 32.27 \\
Ours, canonical route & 57.2 & 9.68 & \rankB{75.0} & \rankB{4.39} & \rankB{77.6} & \rankB{4.33} & 66.4 & \rankC{6.59} & \rankC{93.3} & \rankA{1.19} & \rankB{79.4} & \rankB{3.80} \\
\midrule
Dynamic 3DGS & \rankB{73.1} & \rankB{5.56} & 64.0 & 7.47 & \rankC{73.1} & \rankC{5.58} & \rankC{69.1} & 6.62 & 88.8 & 2.39 & \rankC{70.1} & 6.09 \\
4DGS & 59.7 & 8.86 & 57.5 & 9.78 & 49.2 & 12.77 & 50.2 & 12.55 & \rankB{93.5} & \rankC{1.33} & 60.4 & 8.52 \\
SC-GS & \rankC{72.4} & 5.76 & 69.6 & 6.25 & 63.8 & 7.76 & 56.3 & 10.48 & 92.8 & 1.56 & 68.4 & 6.02 \\
Deformable 3DGS & 69.9 & 6.55 & 61.0 & 8.76 & 65.0 & 7.70 & 52.3 & 12.58 & \rankA{93.6} & 1.39 & 59.6 & 8.75 \\
Dynamic-2DGS$^\dagger$ & -- & -- & -- & -- & -- & -- & -- & -- & 82.4 & 3.47 & 26.0 & 41.96 \\
AT-GS$^\ddagger$ & 57.4 & 9.67 & 54.6 & 10.52 & 59.1 & 9.09 & 55.2 & 10.94 & 76.7 & 4.78 & 60.2 & 8.80 \\
DG-Mesh & 65.8 & 7.26 & 64.8 & 7.71 & 64.8 & 7.70 & 53.0 & 12.14 & 92.6 & 1.71 & 55.5 & 9.93 \\
GauSTAR & 71.6 & \rankC{5.68} & 63.7 & 7.56 & 68.5 & 6.59 & \rankB{77.9} & \rankA{3.28} & 78.6 & 4.41 & 65.8 & 7.43 \\
MaGS & 68.1 & 6.78 & 61.3 & 8.67 & 63.6 & 8.08 & 47.5 & 15.02 & 92.5 & 1.37 & 52.7 & 11.49 \\
Tensor4D & 40.9 & 16.03 & 55.4 & 10.20 & 51.3 & 11.46 & 33.4 & 21.34 & 89.1 & 2.21 & 55.8 & 10.83 \\
MoAngelo & 62.1 & 7.98 & \rankC{70.4} & \rankC{5.58} & 67.8 & 6.53 & 58.8 & 9.24 & 91.1 & \rankB{1.21} & 68.1 & \rankC{5.98} \\
\midrule
\textbf{Ours} & \rankA{79.5} & \rankA{4.18} & \rankA{79.0} & \rankA{3.98} & \rankA{82.9} & \rankA{3.46} & \rankA{80.4} & \rankB{4.07} & \rankA{93.6} & 1.63 & \rankA{83.2} & \rankA{3.20} \\
\bottomrule
\end{tabular}
\caption{\textbf{Correspondences on real captures, per scene}: mean over the held-out cameras of each scene, each method on its own visible pairs; metrics as in \Cref{tab:realdata_tracks}; \ranklegend{} per scene and dataset. $^\dagger$On TaoGS, training fails on magic2 (empty mesh) and does not finish on magic3 and changeclothes; on ActorsHQ, it converges only on actor7 and actor8. $^\ddagger$Correspondence from the lineage of its Gaussians. --: failed ($^\dagger$).}
\label{tab:realdata_tracks_scenes}
\end{table*}

%% file: figures/iters_scaling.tex
\begin{figure*}[t]
\centering
\includegraphics[width=\textwidth]{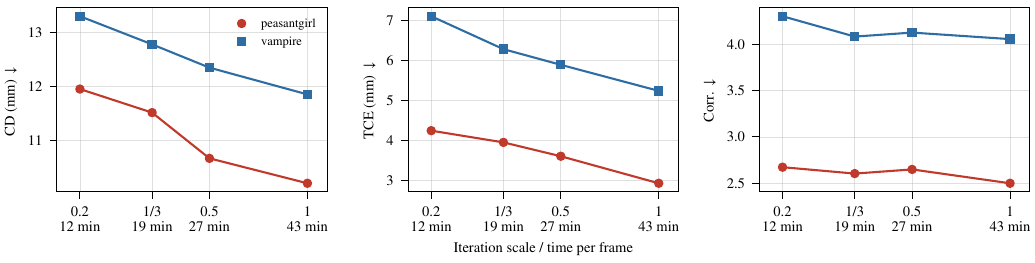}
\caption{\textbf{Accuracy vs.\ running time.} Chamfer distance, temporal consistency error and correspondence error of our method on two DeformingThings4D scenes when all iteration counts of the schedule are scaled by 0.2, $\nicefrac{1}{3}$, 0.5 or 1 (time per frame below each scale). The surface errors fall steadily with more iterations; the correspondences change little.}
\label{fig:iters_scaling}
\end{figure*}

%% file: figures/environments.tex
\begin{figure*}[t]
\centering
\setlength{\tabcolsep}{1pt}\renewcommand{\arraystretch}{0.6}
\begin{tabular}{cc}\includegraphics[width=0.494\textwidth]{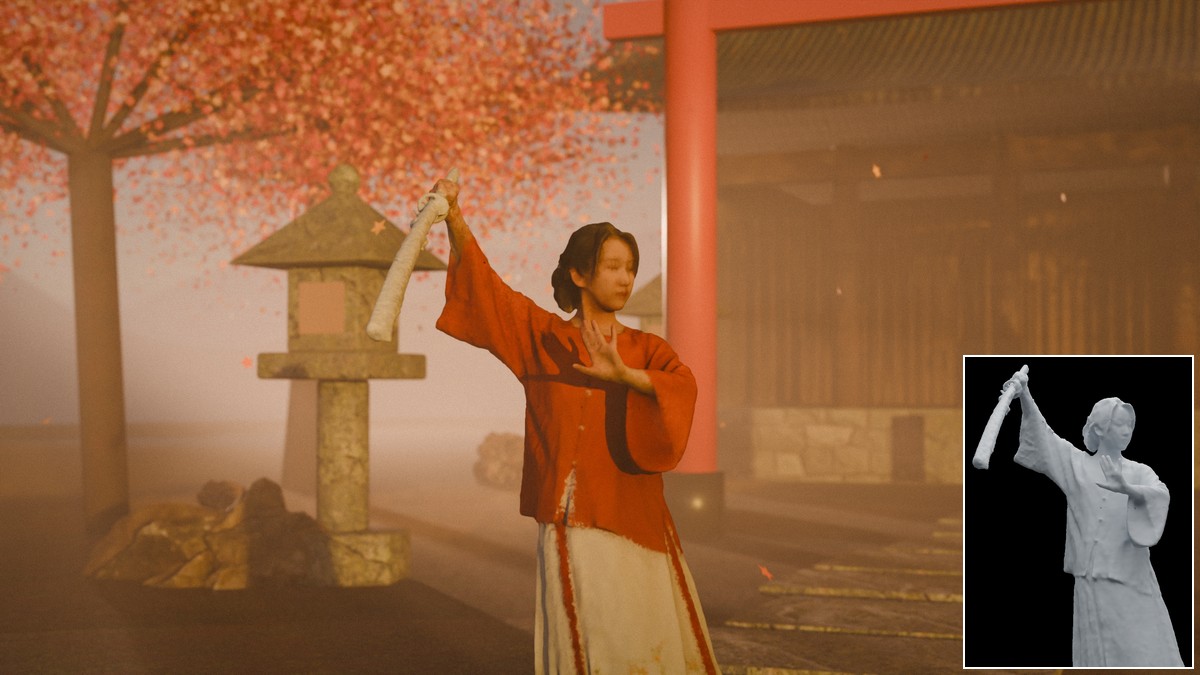} &
\includegraphics[width=0.494\textwidth]{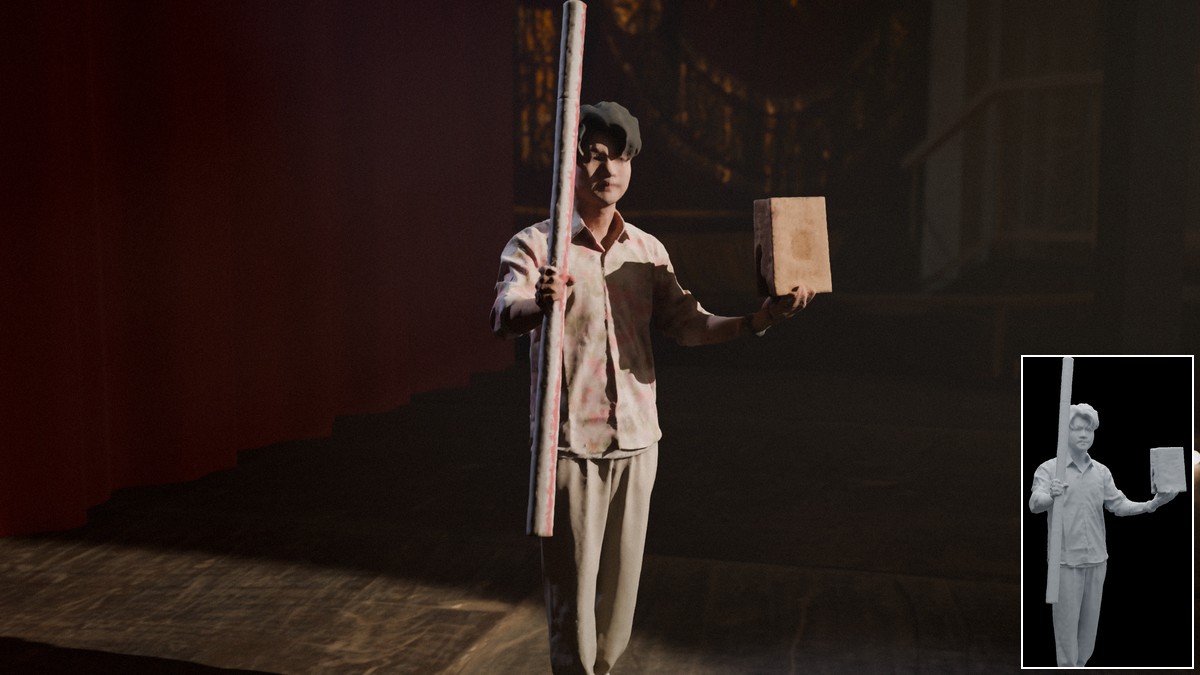} \\
\includegraphics[width=0.494\textwidth]{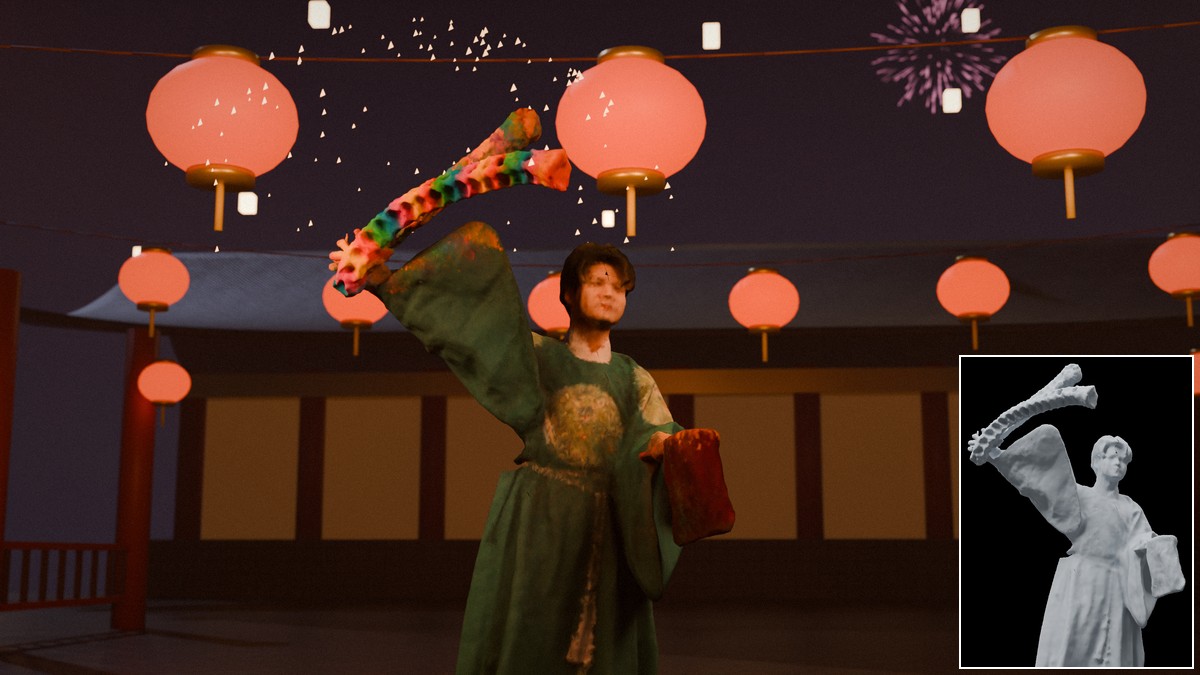} &
\includegraphics[width=0.494\textwidth]{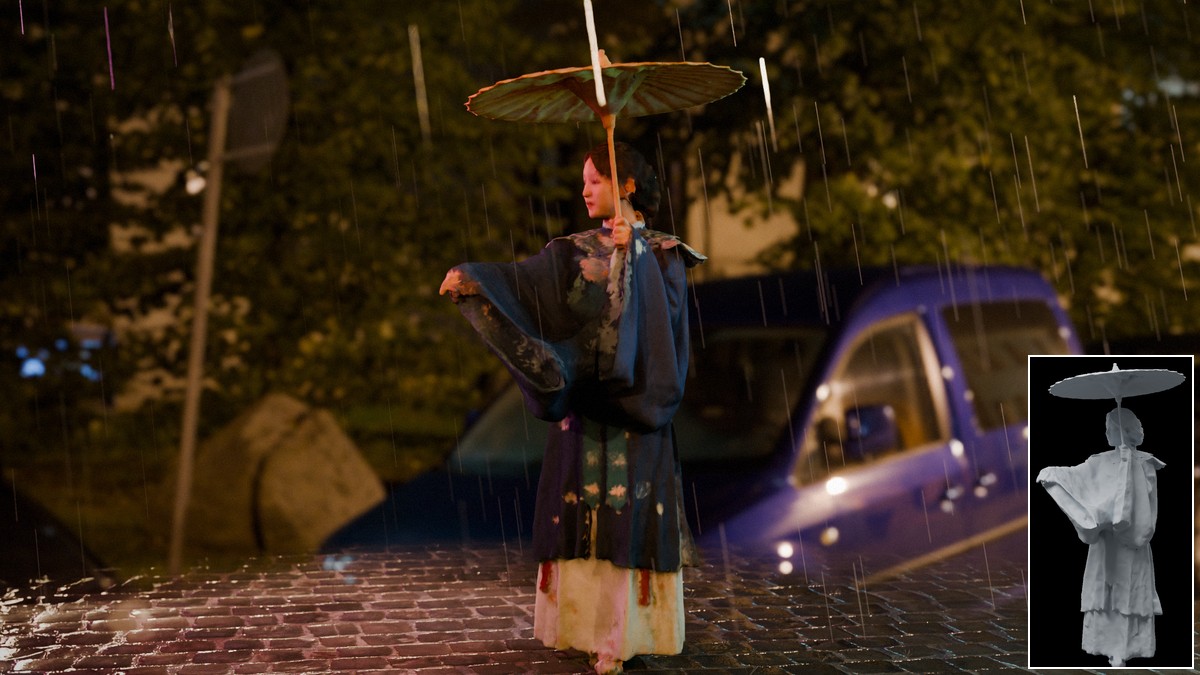} \\
\end{tabular}
\caption{\textbf{Content creation.} Our textured 4D reconstructions of real captures placed in virtual environments and rendered with standard graphics tools; the inset shows our reconstructed geometry of the same frame. The animated sequences are in the video on the project page.}
\label{fig:environments}
\end{figure*}

%% file: tables/raft.tex
\begin{table*}[t]
\centering
\footnotesize
\setlength{\tabcolsep}{3pt}
\begin{tabular}{@{}llccccc@{}}
\toprule
Scene & Flow & Flow EPE (px)$\downarrow$ & CD$\downarrow$ & TCE$\downarrow$ & Corr.$\downarrow$ \\
\midrule
peasantgirl & FlowSeek & \rankA{1.34} & \rankA{10.20} & \rankA{2.93} & \rankA{2.50} \\
            & RAFT     & 8.17 & 10.43 & 3.11 & 14.49 \\
\midrule
vampire     & FlowSeek & \rankA{1.61} & 11.85 & \rankA{5.23} & \rankA{4.06} \\
            & RAFT     & 5.95 & \rankA{11.82} & 5.37 & 6.13 \\
\midrule
ninja       & FlowSeek & -- & \rankA{3.24} & \rankA{0.93} & \rankA{3.64} \\
            & RAFT     & -- & 3.25 & 0.94 & 4.60 \\
\midrule
actor4\_1258 & FlowSeek & -- & 3.42 & -- & -- \\
            & RAFT     & -- & \rankA{3.37} & -- & -- \\
\midrule
actor8\_325 & FlowSeek & -- & 3.28 & -- & -- \\
            & RAFT     & -- & \rankA{3.23} & -- & -- \\
\bottomrule
\end{tabular}
\caption{\textbf{Sensitivity to the input optical flow} on DeformingThings4D (peasantgirl, vampire, ninja) and ActorsHQ (actor4, actor8): our full model supervised by RAFT~\cite{teed2020raft} instead of FlowSeek~\cite{Poggi_2025_ICCV} optical flow. Flow EPE: end-point error of the input flow against the ground-truth flow; CD and TCE in mm; Corr.: geodesic error ($\times 100$) as in \Cref{tab:correspondence}. \bestlegend{} of the two per scene. ActorsHQ has no ground-truth flow or correspondences, and no flow EPE was computed for ninja (--).}
\label{tab:raft}
\end{table*}

%% file: figures/applications.tex
\begin{figure*}[t]
  \centering
  \definecolor{tzIn}{RGB}{74,82,99}\definecolor{tzOurs}{RGB}{14,128,229}\definecolor{tzApp}{RGB}{226,104,38}\definecolor{tzInk}{RGB}{24,28,40}
  \newcommand{\appdir}{imgs/app}
  \begin{tikzpicture}[x=1cm, y=1cm, font=\scriptsize, text=tzInk, img/.style={anchor=south west, inner sep=0}]
    \def\cw{3.55}\def\chh{3.59}\def\gap{0.06}\def\lw{0.6}
    \foreach \r/\k/\t in {0/photo/{Input}, 1/dots/{Edited on frame 1}, 2/warm/{Relit, warm light}, 3/cool/{Relit, cool light}} {
      \pgfmathsetmacro{\yy}{(3-\r)*(\chh+\gap)}
      \node[rotate=90, anchor=south, align=center, text width=\chh cm] at (\lw-0.1, \yy+0.5*\chh) {\t};
      \foreach \c/\n in {0/1, 1/17, 2/33, 3/49}
        \node[img] at (\lw+\c*\cw+\c*\gap, \yy) {\includegraphics[width=\cw cm]{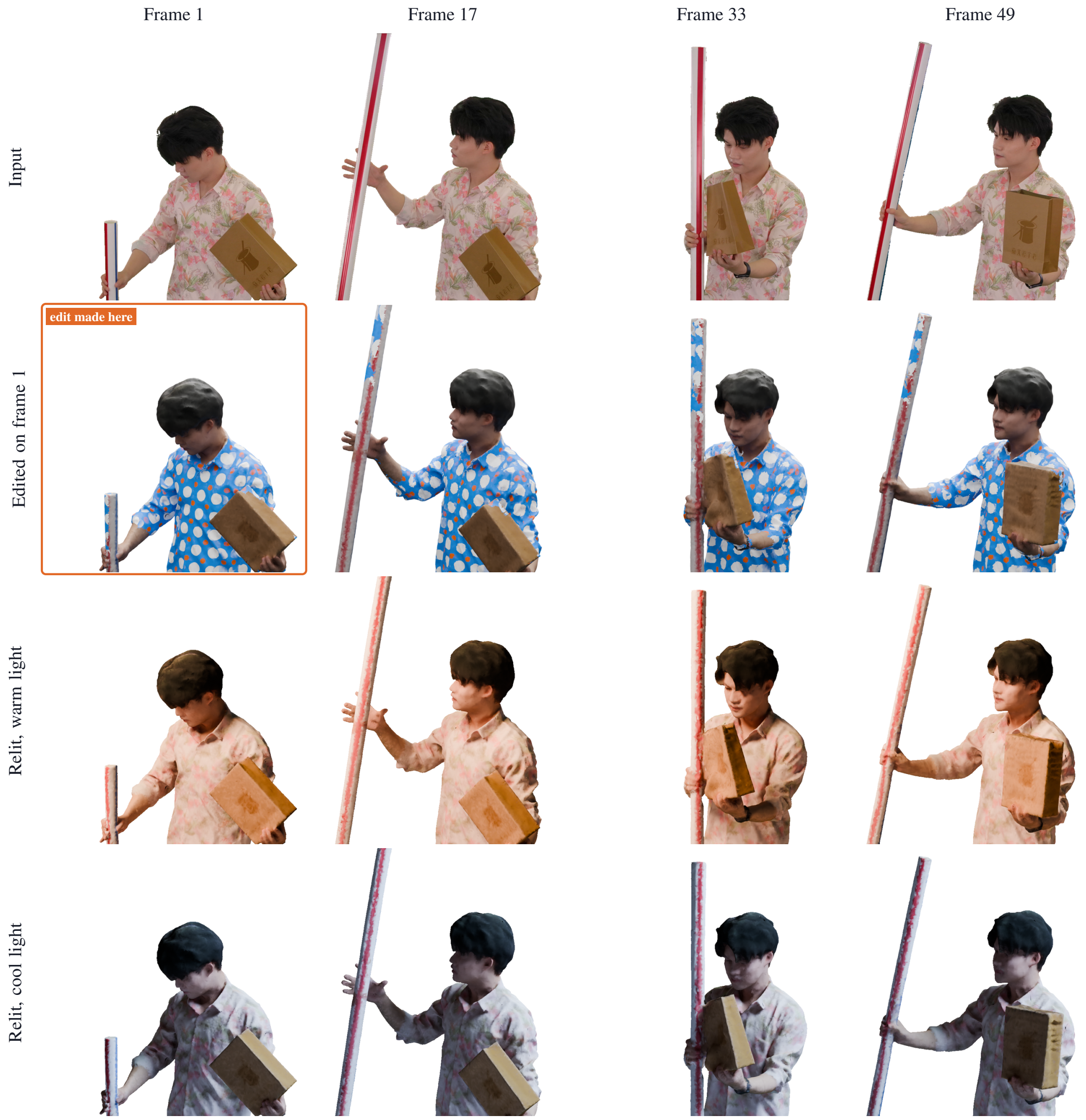}};
    }
    \foreach \c/\n in {0/1, 1/17, 2/33, 3/49}
      \node[anchor=south] at (\lw+\c*\cw+\c*\gap+0.5*\cw, 4*\chh+3*\gap+0.05) {Frame \n};
    \draw[tzApp, line width=0.8pt, rounded corners=1.5pt] (\lw, 2*\chh+2*\gap-0.02) rectangle (\lw+\cw, 3*\chh+2*\gap+0.02);
    \node[anchor=north west, fill=tzApp, text=white, font=\tiny\bfseries, inner sep=1.5pt] at (\lw+0.05, 3*\chh+2*\gap-0.03) {edit made here};
  \end{tikzpicture}
  \caption{\textbf{Applications of the correspondences} on TaoGS~\cite{taogs}. The blouse is edited on frame~1 only; the edit follows the surface through frames 17, 33 and 49 because every point of the first mesh is linked to a point of each later mesh. The same links give one albedo over the sequence, which is relit with a warm and a cool key light.}
  \label{fig:applications}
\end{figure*}

%% file: tables/iters.tex
\begin{table*}[t]
\centering
\footnotesize
\setlength{\tabcolsep}{3pt}
\begin{tabular}{@{}llcccc@{}}
\toprule
Scene & Method & Iterations & CD$\downarrow$ & TCE$\downarrow$ & Corr.$\downarrow$ \\
\midrule
peasantgirl & Ours                    & $1\times$             & \rankA{10.20} & \rankA{2.93} & \rankA{2.50} \\
            & Ours, $\nicefrac{1}{3}$ iterations & $\nicefrac{1}{3}\times$ & 11.51 & 3.95 & 2.60 \\
            & \textit{Change}         &                       & \textit{+12.8\%} & \textit{+34.8\%} & \textit{+4.0\%} \\
\midrule
vampire     & Ours                    & $1\times$             & \rankA{11.85} & \rankA{5.23} & \rankA{4.06} \\
            & Ours, $\nicefrac{1}{3}$ iterations & $\nicefrac{1}{3}\times$ & 12.78 & 6.28 & 4.08 \\
            & \textit{Change}         &                       & \textit{+7.8\%} & \textit{+20.1\%} & \textit{+0.5\%} \\
\midrule
actor4\_1258 & Ours                   & $1\times$             & \rankA{3.42} & -- & -- \\
            & Ours, $\nicefrac{1}{3}$ iterations & $\nicefrac{1}{3}\times$ & 3.70 & -- & -- \\
            & \textit{Change}         &                       & \textit{+8.2\%} & -- & -- \\
\bottomrule
\end{tabular}
\caption{\textbf{Accuracy--runtime trade-off.} Our model with all iteration counts of its schedule scaled by $\nicefrac{1}{3}$ (19 instead of 43 minutes per frame) against the full model (Iterations: iteration counts relative to the full model); Change: relative increase of the error. Chamfer distance (CD) in mm. TCE: temporal consistency error in mm. --: not evaluated (ActorsHQ has no ground-truth correspondences). \bestlegend}
\label{tab:iters}
\end{table*}

%% file: figures/texture_transfer.tex
\begin{figure*}[t]
\centering
\definecolor{tzInk}{RGB}{24,28,40}
\begin{tikzpicture}[x=1cm, y=1cm, font=\scriptsize, text=tzInk, cell/.style={anchor=center, inner sep=0}]
\node[anchor=south, inner sep=1pt] at (2.238,-0.360) {frame 1};
\node[anchor=south, inner sep=1pt] at (5.624,-0.360) {13};
\node[anchor=south, inner sep=1pt] at (9.010,-0.360) {25};
\node[anchor=south, inner sep=1pt] at (12.396,-0.360) {37};
\node[anchor=south, inner sep=1pt] at (15.782,-0.360) {49 (48 for umbrella)};
\node[rotate=90, anchor=south, inner sep=1pt] at (0.500,-2.129) {drawsword};
\node[cell, anchor=south] at (2.238,-3.838) {\includegraphics[width=3.236cm]{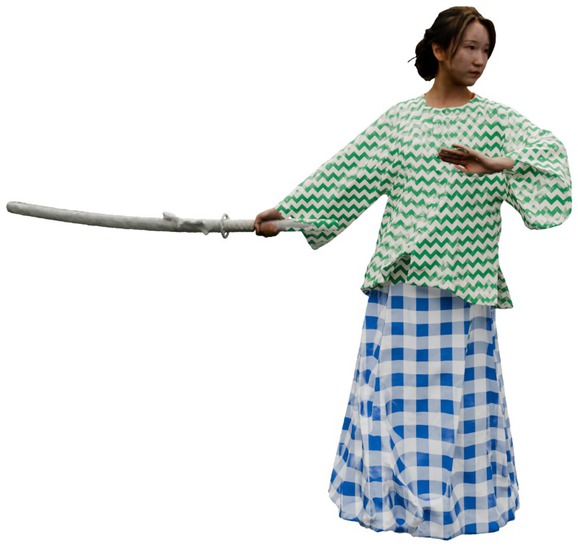}};
\node[cell, anchor=south] at (5.624,-3.838) {\includegraphics[width=2.706cm]{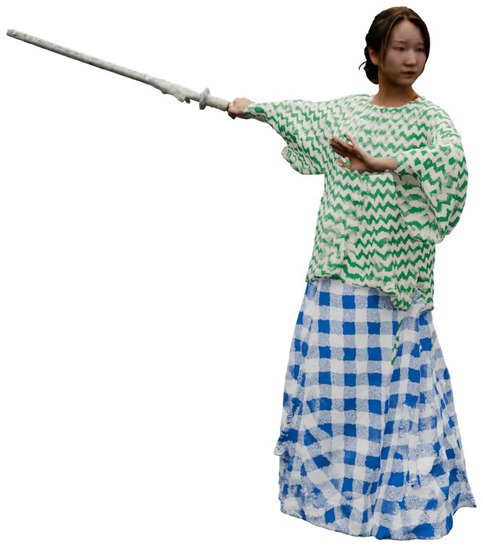}};
\node[cell, anchor=south] at (9.010,-3.838) {\includegraphics[width=2.362cm]{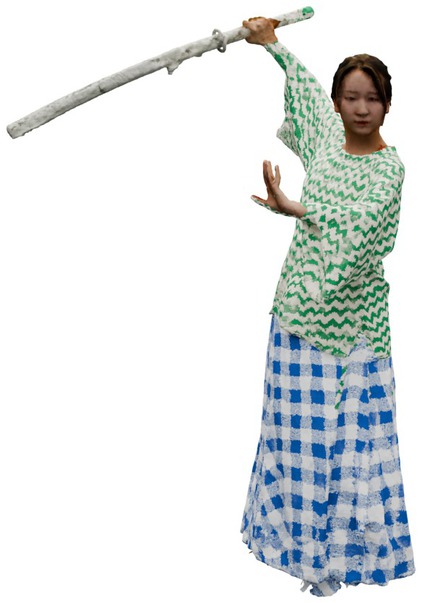}};
\node[cell, anchor=south] at (12.396,-3.838) {\includegraphics[width=1.133cm]{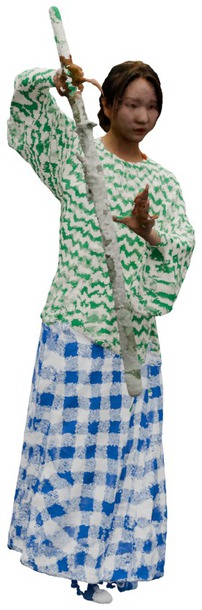}};
\node[cell, anchor=south] at (15.782,-3.838) {\includegraphics[width=1.359cm]{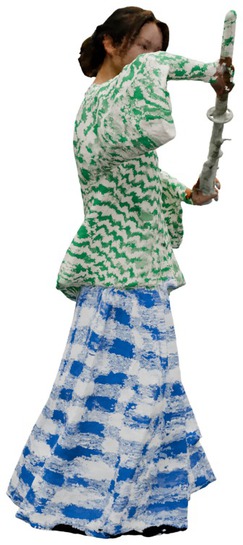}};
\node[rotate=90, anchor=south, inner sep=1pt] at (0.500,-6.415) {magic2};
\node[cell, anchor=south] at (2.238,-8.872) {\includegraphics[width=1.561cm]{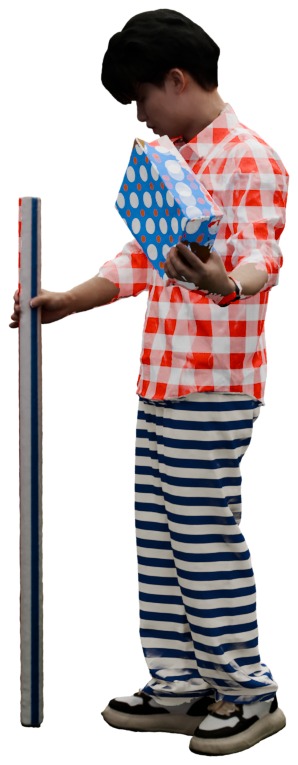}};
\node[cell, anchor=south] at (5.624,-8.872) {\includegraphics[width=1.739cm]{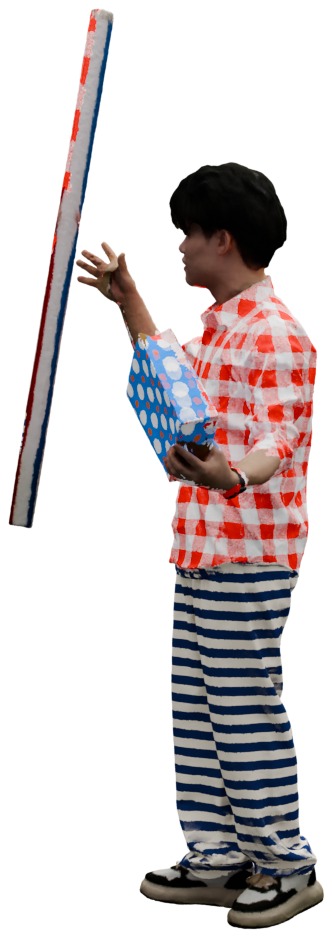}};
\node[cell, anchor=south] at (9.010,-8.872) {\includegraphics[width=1.875cm]{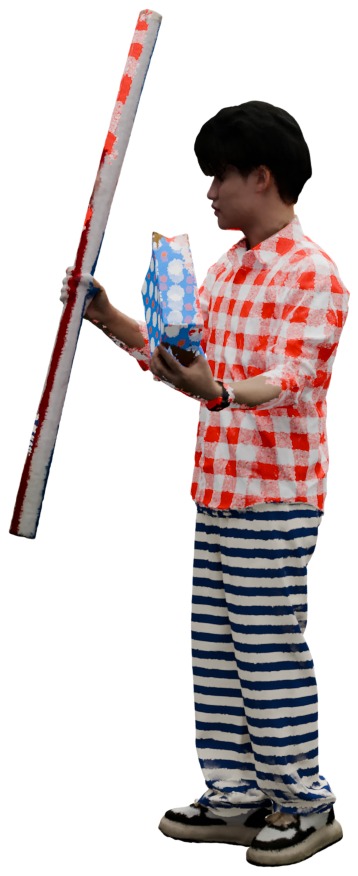}};
\node[cell, anchor=south] at (12.396,-8.872) {\includegraphics[width=1.572cm]{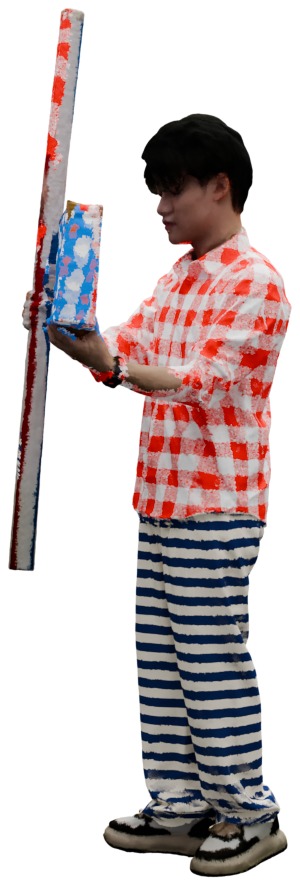}};
\node[cell, anchor=south] at (15.782,-8.872) {\includegraphics[width=1.781cm]{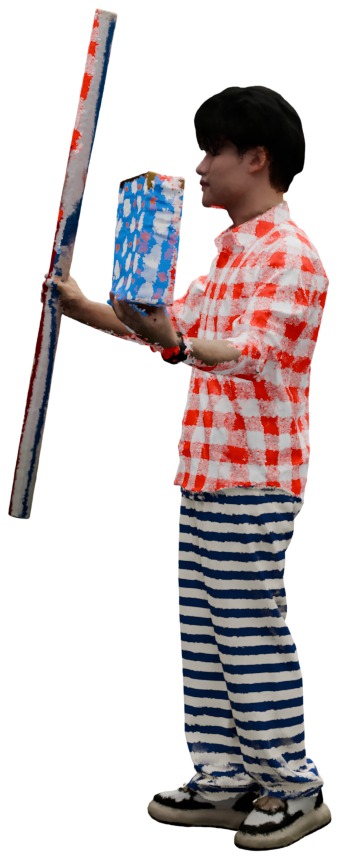}};
\draw[black!15, line width=0.3pt] (0.620,-3.898) -- (17.400,-3.898);
\node[rotate=90, anchor=south, inner sep=1pt] at (0.500,-11.449) {magic3};
\node[cell, anchor=south] at (2.238,-13.906) {\includegraphics[width=1.696cm]{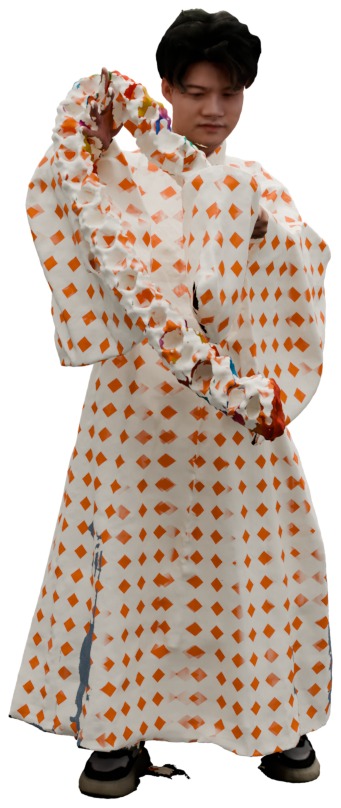}};
\node[cell, anchor=south] at (5.624,-13.906) {\includegraphics[width=1.860cm]{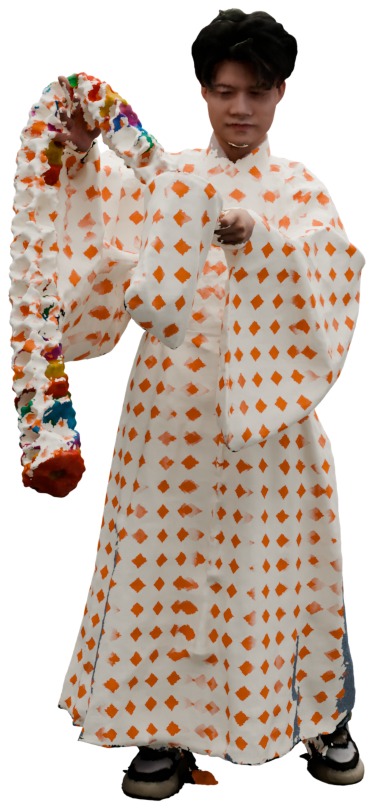}};
\node[cell, anchor=south] at (9.010,-13.906) {\includegraphics[width=2.472cm]{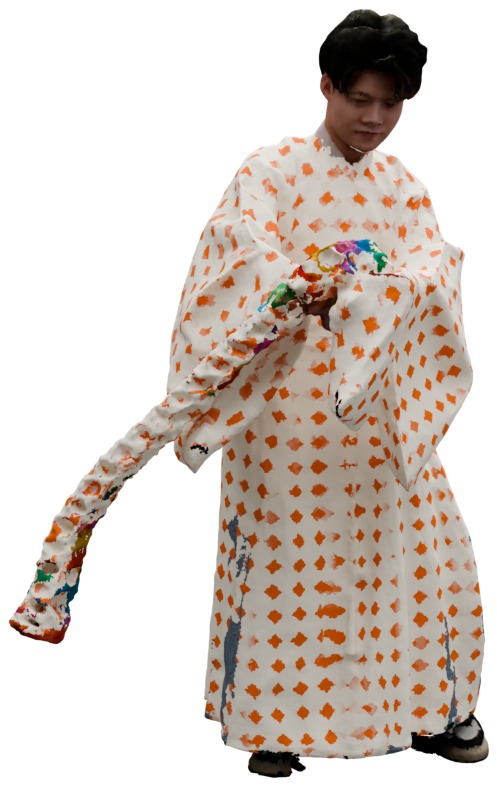}};
\node[cell, anchor=south] at (12.396,-13.906) {\includegraphics[width=1.860cm]{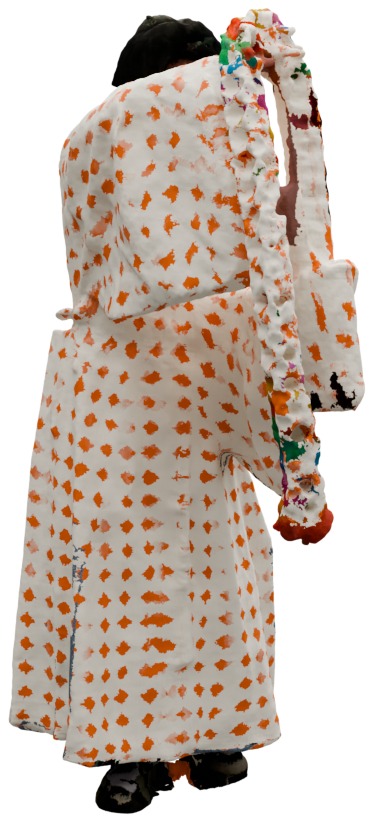}};
\node[cell, anchor=south] at (15.782,-13.906) {\includegraphics[width=2.929cm]{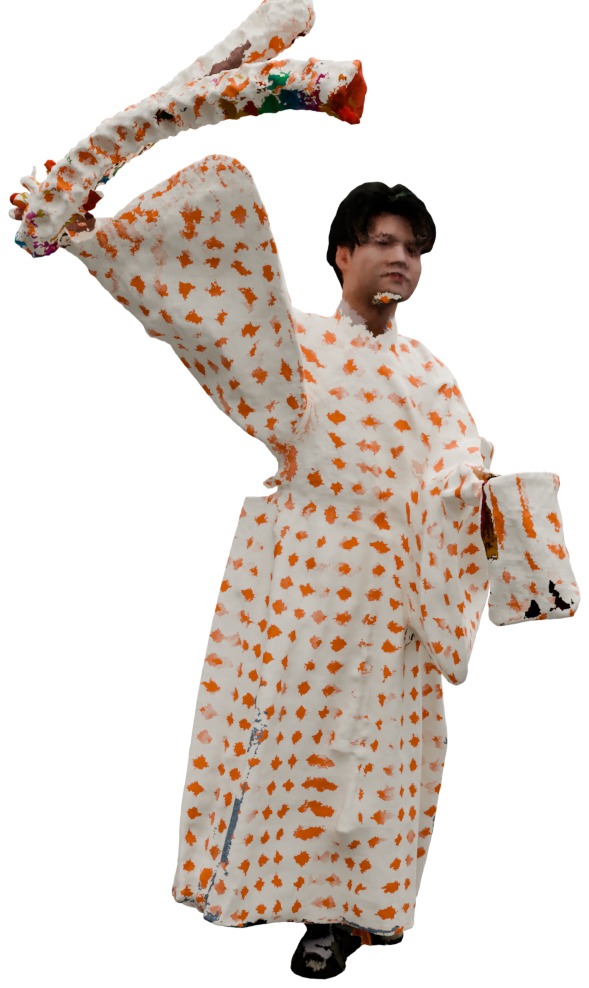}};
\draw[black!15, line width=0.3pt] (0.620,-8.932) -- (17.400,-8.932);
\node[rotate=90, anchor=south, inner sep=1pt] at (0.500,-16.483) {umbrella};
\node[cell, anchor=south] at (2.238,-18.940) {\includegraphics[width=2.360cm]{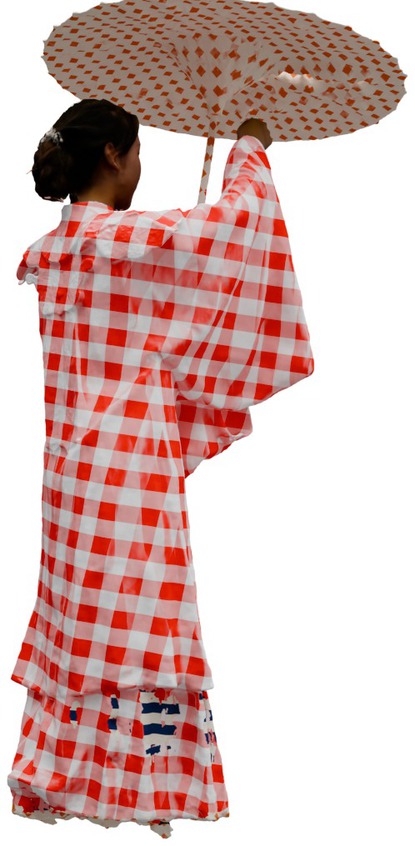}};
\node[cell, anchor=south] at (5.624,-18.940) {\includegraphics[width=2.189cm]{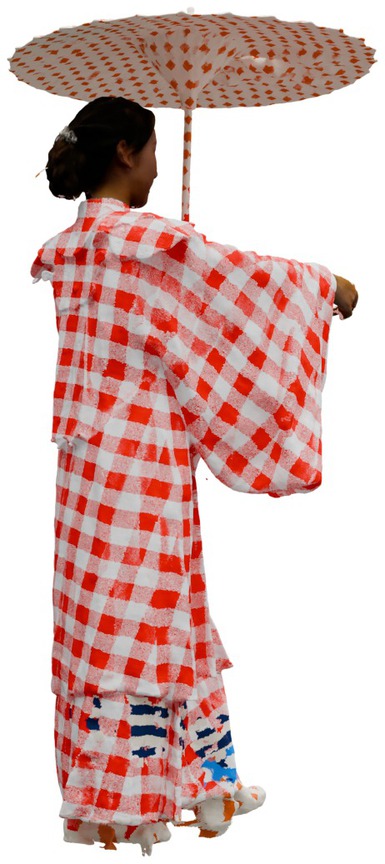}};
\node[cell, anchor=south] at (9.010,-18.940) {\includegraphics[width=2.203cm]{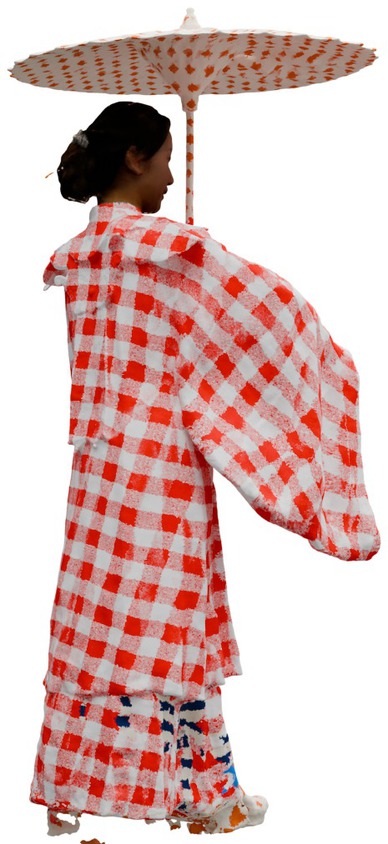}};
\node[cell, anchor=south] at (12.396,-18.940) {\includegraphics[width=2.185cm]{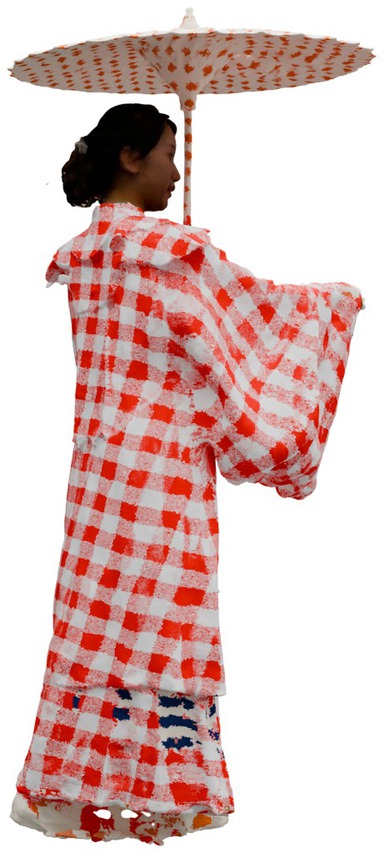}};
\node[cell, anchor=south] at (15.782,-18.940) {\includegraphics[width=2.711cm]{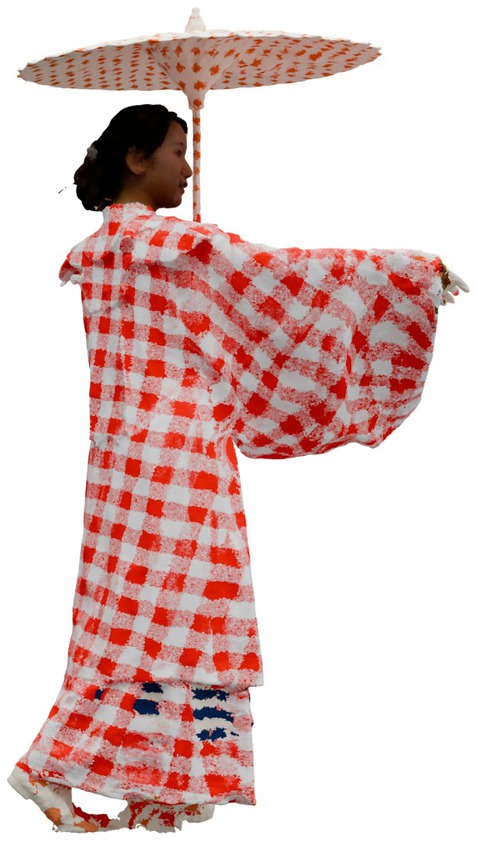}};
\draw[black!15, line width=0.3pt] (0.620,-13.966) -- (17.400,-13.966);
\end{tikzpicture}
\caption{\textbf{Texture transfer} on TaoGS: the person is dressed on frame~1 only, with one pattern per part (top, bottom, held objects; parts from a clothes parser on the frame-1 images), and our correspondences carry the dressing to the later frames. drawsword is seen from a camera that was not used for training.}
\label{fig:texture_transfer}
\end{figure*}

%% file: figures/relighting.tex
\begin{figure*}[t]
\centering
\definecolor{tzInk}{RGB}{24,28,40}
\begin{tikzpicture}[x=1cm, y=1cm, font=\scriptsize, text=tzInk, cell/.style={anchor=center, inner sep=0}]
\node[anchor=south, inner sep=1pt] at (1.829,0.02) {input};
\node[anchor=south, inner sep=1pt] at (4.668,0.02) {albedo};
\node[anchor=south, inner sep=1pt] at (7.506,0.02) {warm, left};
\node[anchor=south, inner sep=1pt] at (10.344,0.02) {cool, right};
\node[anchor=south, inner sep=1pt] at (13.182,0.02) {white, top};
\node[anchor=south, inner sep=1pt] at (16.021,0.02) {white, side};
\node[rotate=90, anchor=south, inner sep=1pt] at (0.450,-1.625) {magic2};
\node[cell] at (1.829,-1.625) {\includegraphics[width=2.758cm,height=3.190cm,keepaspectratio]{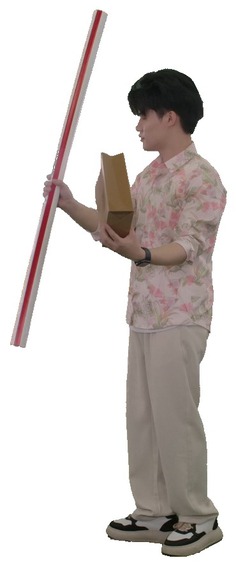}};
\node[cell] at (4.668,-1.625) {\includegraphics[width=2.758cm,height=3.190cm,keepaspectratio]{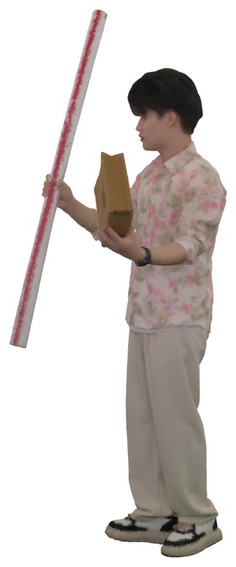}};
\node[cell] at (7.506,-1.625) {\includegraphics[width=2.758cm,height=3.190cm,keepaspectratio]{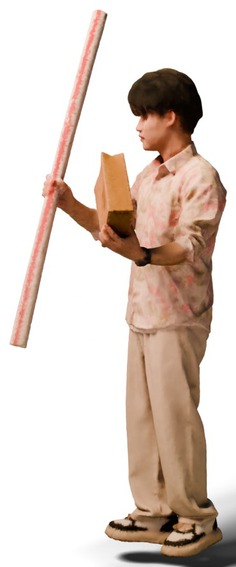}};
\node[cell] at (10.344,-1.625) {\includegraphics[width=2.758cm,height=3.190cm,keepaspectratio]{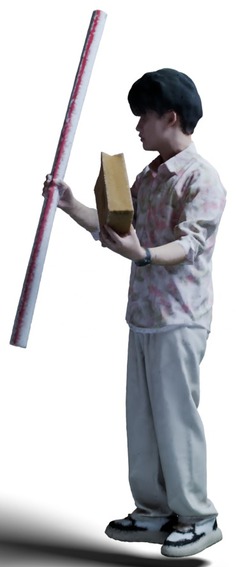}};
\node[cell] at (13.182,-1.625) {\includegraphics[width=2.758cm,height=3.190cm,keepaspectratio]{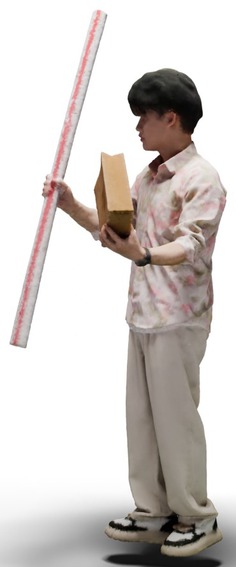}};
\node[cell] at (16.021,-1.625) {\includegraphics[width=2.758cm,height=3.190cm,keepaspectratio]{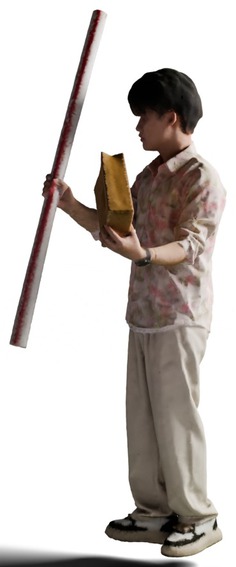}};
\node[rotate=90, anchor=south, inner sep=1pt] at (0.450,-4.875) {magic3};
\node[cell] at (1.829,-4.875) {\includegraphics[width=2.758cm,height=3.190cm,keepaspectratio]{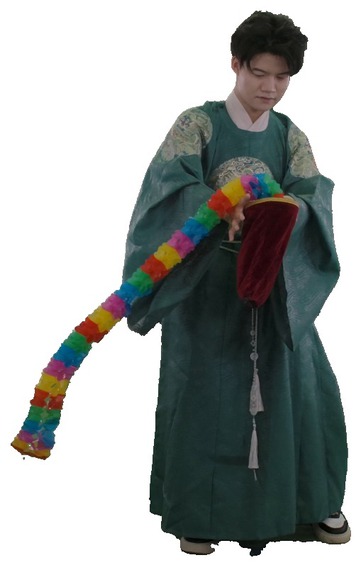}};
\node[cell] at (4.668,-4.875) {\includegraphics[width=2.758cm,height=3.190cm,keepaspectratio]{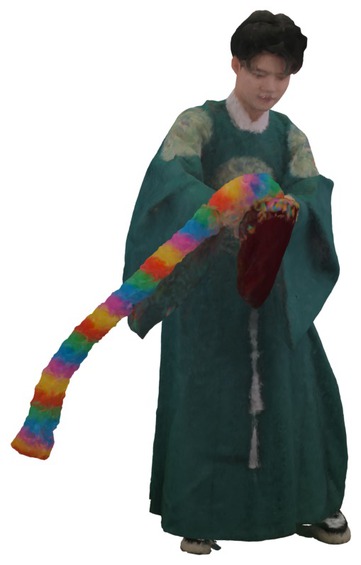}};
\node[cell] at (7.506,-4.875) {\includegraphics[width=2.758cm,height=3.190cm,keepaspectratio]{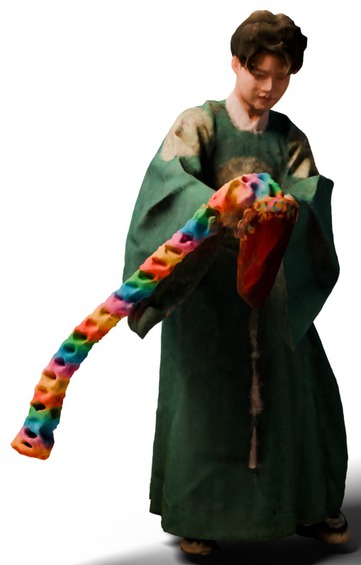}};
\node[cell] at (10.344,-4.875) {\includegraphics[width=2.758cm,height=3.190cm,keepaspectratio]{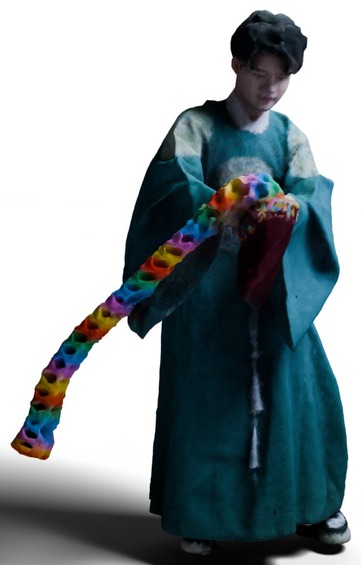}};
\node[cell] at (13.182,-4.875) {\includegraphics[width=2.758cm,height=3.190cm,keepaspectratio]{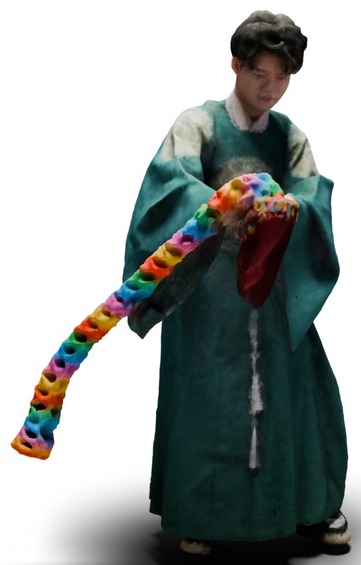}};
\node[cell] at (16.021,-4.875) {\includegraphics[width=2.758cm,height=3.190cm,keepaspectratio]{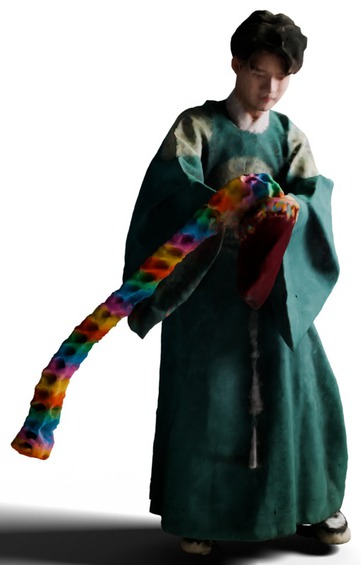}};
\draw[black!15, line width=0.3pt] (0.450,-3.250) -- (17.400,-3.250);
\node[rotate=90, anchor=south, inner sep=1pt] at (0.450,-8.125) {drawsword};
\node[cell] at (1.829,-8.125) {\includegraphics[width=2.758cm,height=3.190cm,keepaspectratio]{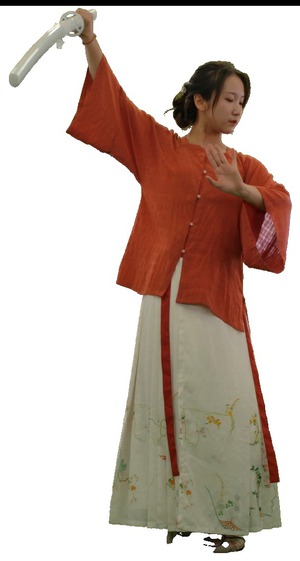}};
\node[cell] at (4.668,-8.125) {\includegraphics[width=2.758cm,height=3.190cm,keepaspectratio]{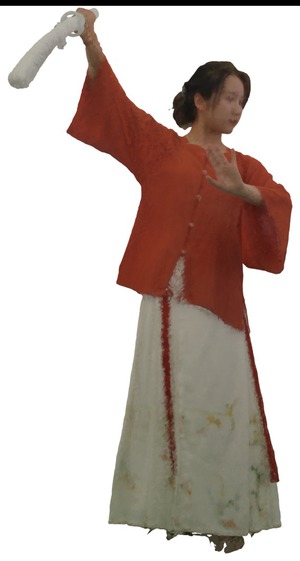}};
\node[cell] at (7.506,-8.125) {\includegraphics[width=2.758cm,height=3.190cm,keepaspectratio]{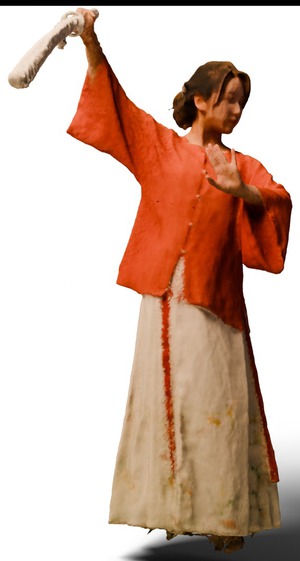}};
\node[cell] at (10.344,-8.125) {\includegraphics[width=2.758cm,height=3.190cm,keepaspectratio]{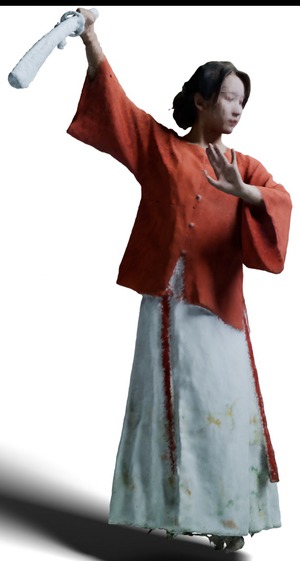}};
\node[cell] at (13.182,-8.125) {\includegraphics[width=2.758cm,height=3.190cm,keepaspectratio]{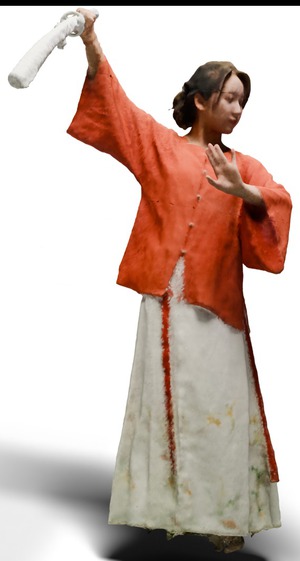}};
\node[cell] at (16.021,-8.125) {\includegraphics[width=2.758cm,height=3.190cm,keepaspectratio]{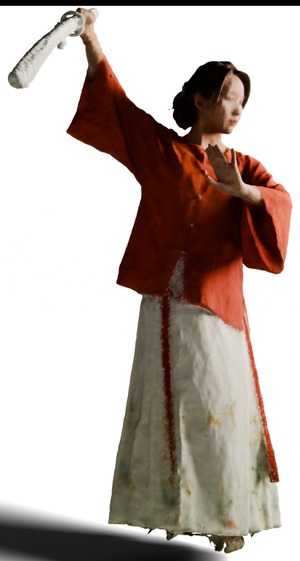}};
\draw[black!15, line width=0.3pt] (0.450,-6.500) -- (17.400,-6.500);
\node[rotate=90, anchor=south, inner sep=1pt] at (0.450,-11.375) {umbrella};
\node[cell] at (1.829,-11.375) {\includegraphics[width=2.758cm,height=3.190cm,keepaspectratio]{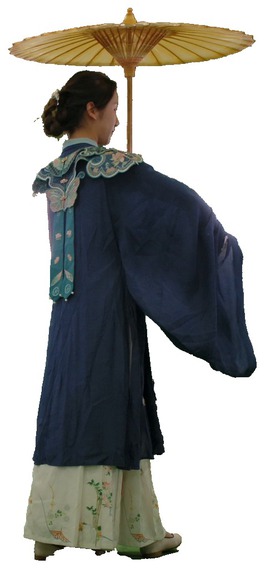}};
\node[cell] at (4.668,-11.375) {\includegraphics[width=2.758cm,height=3.190cm,keepaspectratio]{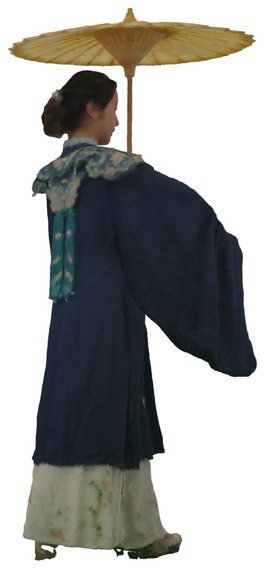}};
\node[cell] at (7.506,-11.375) {\includegraphics[width=2.758cm,height=3.190cm,keepaspectratio]{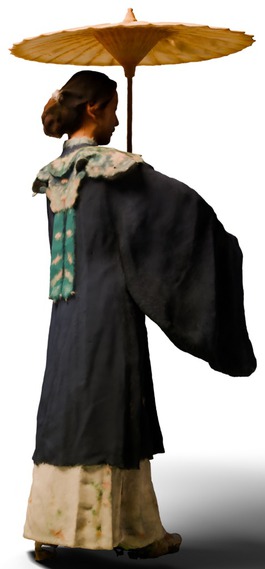}};
\node[cell] at (10.344,-11.375) {\includegraphics[width=2.758cm,height=3.190cm,keepaspectratio]{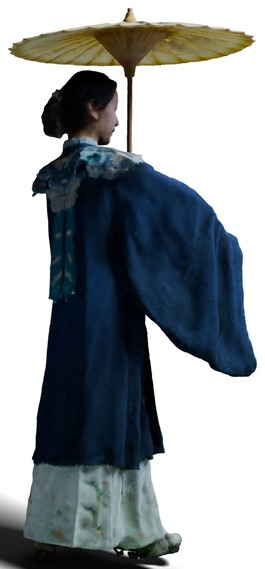}};
\node[cell] at (13.182,-11.375) {\includegraphics[width=2.758cm,height=3.190cm,keepaspectratio]{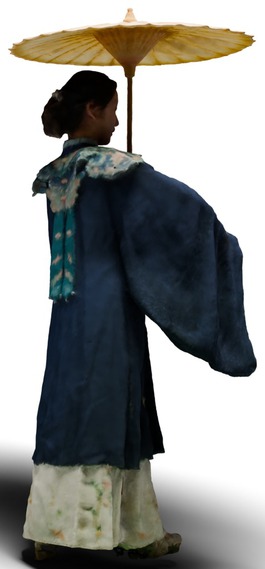}};
\node[cell] at (16.021,-11.375) {\includegraphics[width=2.758cm,height=3.190cm,keepaspectratio]{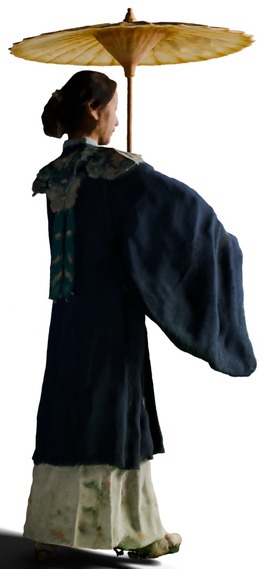}};
\draw[black!15, line width=0.3pt] (0.450,-9.750) -- (17.400,-9.750);
\node[rotate=90, anchor=south, inner sep=1pt] at (0.450,-14.625) {bag};
\node[cell] at (1.829,-14.625) {\includegraphics[width=2.758cm,height=3.190cm,keepaspectratio]{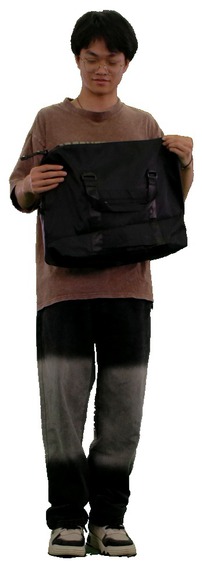}};
\node[cell] at (4.668,-14.625) {\includegraphics[width=2.758cm,height=3.190cm,keepaspectratio]{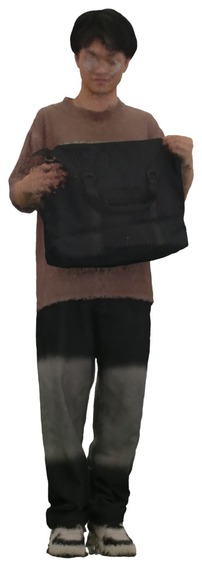}};
\node[cell] at (7.506,-14.625) {\includegraphics[width=2.758cm,height=3.190cm,keepaspectratio]{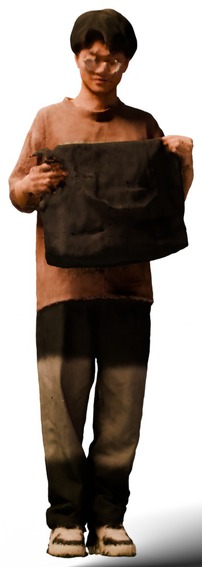}};
\node[cell] at (10.344,-14.625) {\includegraphics[width=2.758cm,height=3.190cm,keepaspectratio]{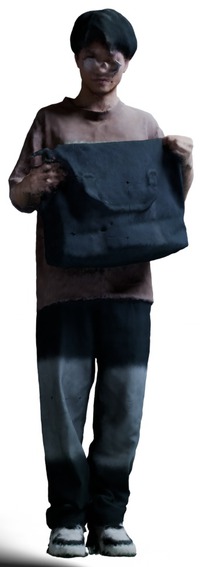}};
\node[cell] at (13.182,-14.625) {\includegraphics[width=2.758cm,height=3.190cm,keepaspectratio]{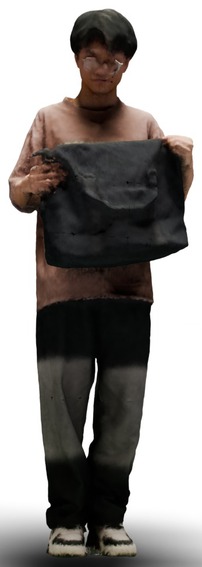}};
\node[cell] at (16.021,-14.625) {\includegraphics[width=2.758cm,height=3.190cm,keepaspectratio]{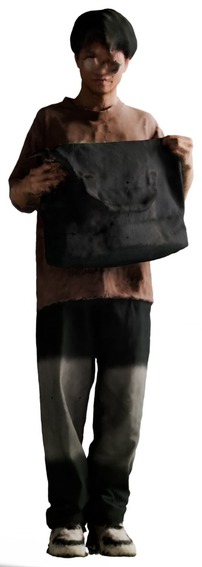}};
\draw[black!15, line width=0.3pt] (0.450,-13.000) -- (17.400,-13.000);
\node[rotate=90, anchor=south, inner sep=1pt] at (0.450,-17.875) {changeclothes};
\node[cell] at (1.829,-17.875) {\includegraphics[width=2.758cm,height=3.190cm,keepaspectratio]{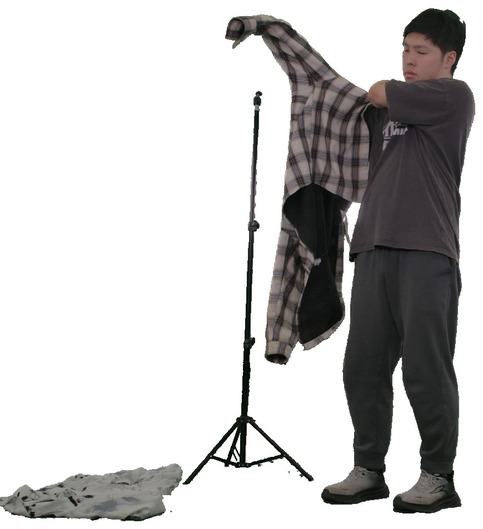}};
\node[cell] at (4.668,-17.875) {\includegraphics[width=2.758cm,height=3.190cm,keepaspectratio]{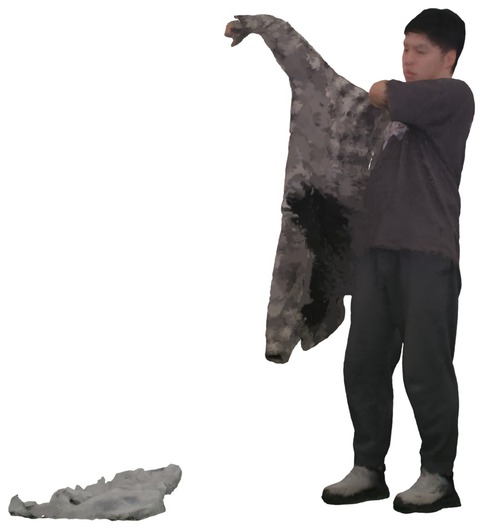}};
\node[cell] at (7.506,-17.875) {\includegraphics[width=2.758cm,height=3.190cm,keepaspectratio]{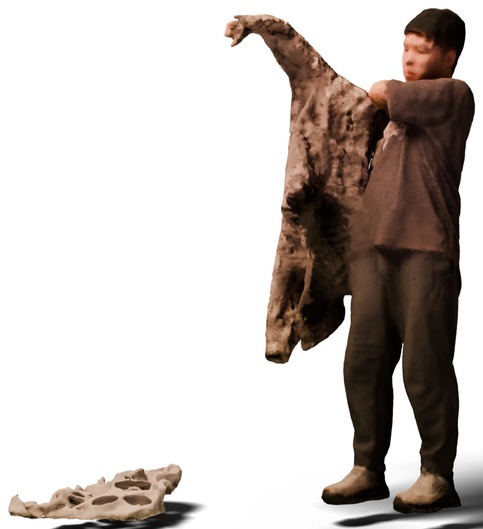}};
\node[cell] at (10.344,-17.875) {\includegraphics[width=2.758cm,height=3.190cm,keepaspectratio]{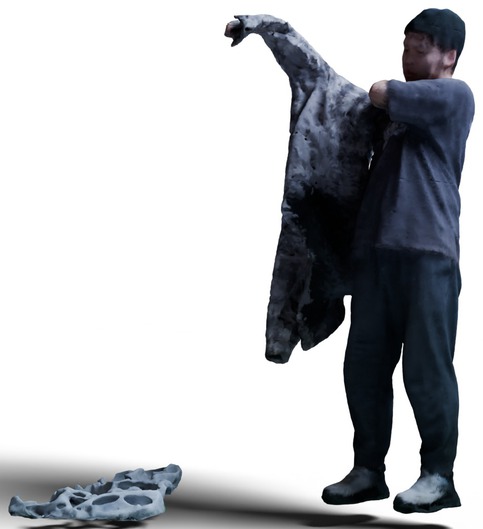}};
\node[cell] at (13.182,-17.875) {\includegraphics[width=2.758cm,height=3.190cm,keepaspectratio]{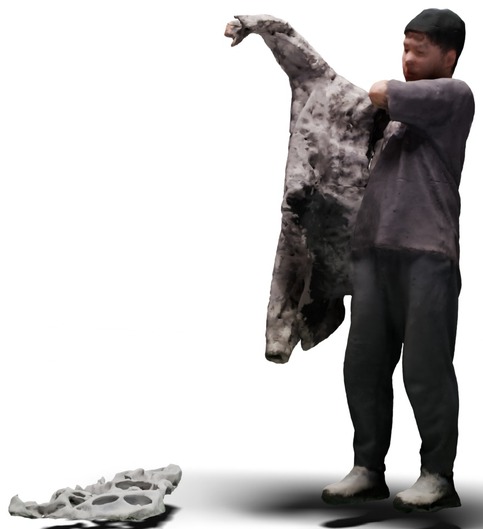}};
\node[cell] at (16.021,-17.875) {\includegraphics[width=2.758cm,height=3.190cm,keepaspectratio]{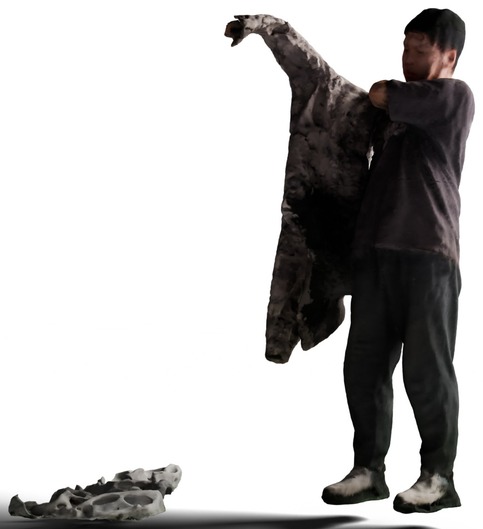}};
\draw[black!15, line width=0.3pt] (0.450,-16.250) -- (17.400,-16.250);
\draw[black!30, line width=0.4pt] (6.087,0.3) -- (6.087,-19.500);
\end{tikzpicture}
\caption{\textbf{Relighting} on TaoGS (frame~25). The albedo is the color of each surface point averaged over all cameras and frames through our correspondences; it is relit with four key lights. Input photo for reference.}
\label{fig:relighting}
\end{figure*}